\documentclass[11pt]{article}
\usepackage[T1]{fontenc}
\usepackage[utf8]{inputenc}
\usepackage{lmodern,microtype}
\DisableLigatures[-]{encoding=T1,family=tt*}
\usepackage[margin=1in]{geometry}
\usepackage{amsmath,amssymb,amsthm,mathtools}
\usepackage{booktabs,longtable,array,enumitem,graphicx}
\usepackage{tikz}
\usetikzlibrary{arrows.meta,positioning,calc}
\usepackage[hidelinks,unicode]{hyperref}
\hypersetup{pdftitle={Sharp Logarithmic-Product Bounds for List Replicability of Sign Matrices},pdfauthor={Guangjian Zhang},pdfsubject={Learning theory and topological combinatorics}}
\allowdisplaybreaks
\newtheorem{theorem}{Theorem}[section]
\newtheorem{lemma}[theorem]{Lemma}
\newtheorem{proposition}[theorem]{Proposition}
\newtheorem{corollary}[theorem]{Corollary}
\theoremstyle{definition}
\newtheorem{definition}[theorem]{Definition}
\newcommand{\proofstep}[1]{\par\smallskip\noindent\textbf{#1}\enspace}

\newenvironment{resultenv}[3]{\refstepcounter{theorem}\label{#2}\par\medskip\noindent\textbf{#1 \thetheorem}\def\thmtmp{#3}\ifx\thmtmp\empty\else\ (#3)\fi\textbf{.}\ \itshape\ignorespaces}{\par\medskip}

\title{Sharp Logarithmic-Product Bounds for List Replicability of Sign Matrices}
\author{Guangjian Zhang\\\small\texttt{zgj1226029469@outlook.com}}
\date{}
\begin{document}
\maketitle
\begin{abstract}
List replicability measures how many accurate hypotheses suffice to contain
the output of a learning rule with arbitrarily high confidence. We prove
that every nonempty class of $m$ binary concepts on $n$ domain points has
list replicability at most $805\ln(2m)\ln(2n)$. This resolves the
logarithmic-product conjecture of Blondal, Hatami, Hatami, Lalov, and
Tretiak. Hanner matrices give lower bounds of the same order, for square
matrices and for explicit families with unequal logarithmic aspect ratios.
The upper bound is obtained by constructing nested acyclic sets in the faces
of a cube. A critical-point estimate supplies highly connected regions,
weighted cell attachments keep the sets compatible across faces, and a cubical
carrier theorem converts them into accurate finite covers and learning rules.
Appending one constant coordinate turns list replicability into an exact
geometric quantity, the face width, and graph-orientation classes satisfy exact
product formulas. In particular, there are explicit classes with list
replicability $4k+1$ and equivariant index $3k$, so the additive gap between the
two is unbounded. We also develop entropy-tilting and two-sided Gibbs
constructions, give combinatorial upper bounds adapted to structured matrices,
compute exact values on classical families, and study the index of Paley
matrices.
\end{abstract}


\clearpage
\setcounter{tocdepth}{1}
\begingroup
\setlength{\parskip}{0pt}
\makeatletter
\renewcommand*\l@section{\@dottedtocline{1}{0em}{1.9em}}
\makeatother
\tableofcontents
\endgroup
\clearpage
\section{Introduction}
A finite binary concept class can be written as a sign matrix. Its rows
are the possible target functions, and its columns are the domain
points. Ordinary realizable learning asks for an accurate function from
random labeled observations. List replicability imposes an additional
requirement: for each data distribution, a bounded list of accurate
functions must contain the output with arbitrarily high confidence.
The list may depend on the distribution, but its size must be uniform over
all realizable distributions.

This requirement connects learning to finite-dimensional topology.
Stable output choices produce covers with bounded multiplicity, and
topological obstructions force that multiplicity to be large.
Chase, Moran, and Yehudayoff~\cite{CMY23} and Dixon, Pavan, Vander Woude,
and Vinodchandran~\cite{DPVWV23} developed the learning-theoretic framework.
The local Borsuk--Ulam theorem of Chase, Chornomaz, Moran, and
Yehudayoff~\cite{CCMY24} supplies general bounds. The sign-complex
approach of Frick, Hosseini, and Vasileuski~\cite{FHV26} and its
connection to list replicability~\cite{BHHLT26} place these bounds in
the same setting as sign rank and equivariant index.

\subsection{Main results}
The basic quantitative question is how large the list replicability of a matrix
with $m$ rows and $n$ columns can be. Write $\operatorname{lr}(C)$ for the list
replicability number of a class $C$, the least list size that can be achieved at
every accuracy; Section~\ref{sec:model} gives the precise definitions.
Before~\cite{Zhang26} the general upper bounds were linear:
$\operatorname{lr}(C)\le\min\{1+\lfloor m/2\rfloor,n\}$,
by~\cite[Theorem E]{CCMY24} and~\cite[Lemma 4]{CMY23}.
Blondal, Hatami, Hatami, Lalov, and Tretiak~\cite[Conjecture 1.17]{BHHLT26}
conjectured an upper bound proportional to $\log m\log n$.
We prove this conjecture.

\begin{theorem}[Logarithmic-product bound]\label{thm:main}
For every nonempty finite class $C\subseteq\{\pm1\}^n$, $n\ge1$, with
$m=|C|$,
\[
 \operatorname{lr}(C)\le805\ln(2m)\ln(2n).
\]
Consequently, for every $N\times M$ sign matrix $B$ with $N,M\ge2$,
\[
 \operatorname{lr}(B)\le1548\log_2N\log_2M.
\]
The same bounds hold for the closed-cover invariant $\mathrm{LR}$ of
Section~\ref{sec:model}.
\end{theorem}
The bound is uniform over all positive accuracies. It controls the number of
possible accurate outputs of a learning rule, while the sample size of the rule
depends on the class, the accuracy, and the confidence.
Since $\operatorname{lr}\le\operatorname{signrk}$~\cite{BHH26a}, the theorem
also shows that list replicability, like the index~\cite{FHMNV26v1}, cannot
certify a sign-rank lower bound above $O(\log^2N)$ on $N\times N$ matrices.

The upper bound has the correct order for square matrices and along explicit
rectangular families, the Hanner matrices with unequal logarithmic aspect ratios.
The following lower bounds hold at each fixed error below $1/2$.

\begin{theorem}[Quantitative Hanner families]\label{thm:main-lower}
For every $N\ge2$, there is an $N\times N$ sign matrix $B$ such
that, for every $0<\varepsilon<1/2$,
\[
 \operatorname{lr}(B,\varepsilon)\ge\frac1{72}\log_2^2N.
\]
For each fixed $a\in(0,1)$ and every integer
$t\ge\lceil4/\min(a,1-a)^2\rceil$, there is an explicit matrix $B_t$ with
$N_t$ rows and $M_t$ columns such that, for every $0<\varepsilon<1/2$
and with $d=2^t$,
\[
 \frac1{16}\log_2N_t\log_2M_t
 \le\frac d2\le\operatorname{lr}(B_t,\varepsilon)
 \le\operatorname{lr}(B_t)\le d,
\]
and $\log M_t=\Theta(d^a)$, $\log N_t=\Theta(d^{1-a})$.
\end{theorem}
This removes the $\log\log N$ factor from the lower bound
$\Omega(\log^2N/\log\log N)$ of~\cite[Proposition 1.16]{BHHLT26}, which
originates in~\cite{CMW25}. Thus the largest list replicability of an
$N\times N$ sign matrix is $\Theta(\log^2N)$. This settles the question
of~\cite[Section 1.4]{BHHLT26} on its growth.

Theorems~\ref{thm:main} and~\ref{thm:main-lower} concern list replicability itself,
which is not a function of the equivariant index of the sign complex
(Proposition~\ref{res:5.8}; Section~\ref{sec:model} recalls these notions).
By~\cite[Theorem 3.1]{BHHLT26},
$\operatorname{Ind}S(C)\le2\operatorname{lr}(C)-1$. The following result shows
that this inequality has no additive converse.

\begin{theorem}[Unbounded index gap]\label{thm:intro-gap}
For every $k\ge1$ there is an explicit class $D_k$ with $32^k$ concepts on
$6k+1$ domain points such that
\[
 \operatorname{lr}(D_k)=4k+1,\qquad\operatorname{Ind}S(D_k)=3k.
\]
\end{theorem}
Theorem~\ref{thm:intro-gap} is part of Theorem~\ref{thm:index-gap}. It answers
negatively the question of~\cite[Section 9.2]{Zhang26} whether
$\operatorname{lr}\le\operatorname{Ind}+O(1)$.

\subsection{Obstacles to the upper bound}
There are three obstacles to passing from a low-dimensional
representation of a class to a learning rule.

First, a geometric representation must preserve the coordinate
constraints that make its output accurate. A map of the sign complex
to a small sphere controls the equivariant index, but it does not by itself
provide a rule for choosing Boolean predictions. We instead
construct maps that preserve the faces of a cube. A coordinate fixed
on a face remains fixed after the map, which is exactly the condition
needed for the output to agree with sufficiently frequent observations.

Second, constructing these maps separately on individual faces is
insufficient, because the choices must agree on intersections of faces.
The proof therefore builds one nested family of auxiliary sets, in finitely
many stages. Each stage has its own highly connected regions of vectors, and a
convex combination of vectors from several stages must lie in the region of the
correct later stage. Cells carry weights that record their stages, and the
additivity of these weights under convex combination makes this bookkeeping
compatible with the geometry.

Third, a topological cover has to produce one learning rule that works
uniformly over distributions. This conversion costs a positive margin in the
loss. Compactness gives a single neighborhood radius at which the
cover multiplicity remains bounded, and empirical frequencies then identify
one of the permitted outputs. Proposition~\ref{prop:cover-rule} carries out this
step with explicit accuracy and confidence parameters.

\subsection{Further results}
Theorem~\ref{thm:main} is uniform over all matrices. The other results of this
paper describe what the bound measures and what additional structure can achieve.

Anchoring a class, that is, appending one constant coordinate, identifies its list
replicability with a geometric face-separation invariant, the face width.
Graph orientations give exact values and product
formulas, among them the classes of Theorem~\ref{thm:intro-gap}. Cubes,
Sylvester Hadamard classes, projective planes, and restricted output families give
further exact comparisons.

Several other constructions retain information that
Theorem~\ref{thm:main} does not provide. Entropy tilting transports
accurate covers without a boosting factor. A two-sided Gibbs
construction lowers the fixed-scale metric entropy of a continuous encoding of the
class from order $\ln^2m\ln n$, the net size in~\cite{Zhang26}
and~\cite[Proposition 2.5]{BHHLTpoly26}, to order $\ln^{3/2}m\ln n$. Closure chains,
switching binary rank, and bounded error intersections give bounds adapted to
structured matrices. We also establish limits on what specific encodings can
achieve, in particular the power maps on which the entropy-tilting argument
rests, and we study Paley matrices as the basic explicit pseudorandom example.

\subsection{Relation to earlier work}
An earlier preprint of the author~\cite{Zhang26} gave the first
polylogarithmic upper bound on list replicability,
$2^{31}\ln^2(2m)\ln(2n)$, through nonlinear encoding and entropy
tilting. It also contained the Hanner lower bounds, the exact families,
the combinatorial bounds, and the Paley analysis developed below.
Theorem~\ref{thm:main} removes the additional factor $\ln(2m)$ by a
different argument, the compatible-carrier construction, which is
developed here in full together with further structural results.

Blondal, Hatami, Hatami, Lalov, and Tretiak~\cite{BHHLTpoly26} later gave a
short self-contained proof of the $O(\log^2m\log n)$ bound.
Their paper~\cite{BHH26b} gives the covering-dimension formulation of list
replicability and the list values of extremal classes, and their
sphere-covering theorem~\cite{BHH26c} gives the cube value at every fixed error.

Blondal, P.~Hatami, Lalov, and Tretiak~\cite{BHLT26} bound list
replicability in terms of the Littlestone dimension $d$ and the dual VC
dimension $v^*$. They prove $\operatorname{lr}\le(d+2)^d$ and
$\operatorname{lr}\le2^{O(dv^*)}$ on countable domains, an exponential
improvement on the bound $2^{2^{O(d)}}$ that follows from the work of Bun,
Livni, and Moran~\cite{BLM20}. These bounds and Theorem~\ref{thm:main} are
incomparable. Theirs are constant on classes of bounded Littlestone
dimension and apply to infinite classes, and Theorem~\ref{thm:main} is
polylogarithmic in $m$ and $n$ for every finite matrix, including those with
$d\approx\log_2m$.

Section~\ref{app:static} studies a different quantity, the
conservative-label parameter $q(C)$, which bounds list replicability from
above but is not bounded by it. A bound on $q(C)$ in terms of $d$ alone gives
the exact suprema $2$ and $4$ of list replicability on finite classes at $d=1,2$,
and $\operatorname{lr}(C)\le22$ at $d=3$, below the
values $3$, $16$, and $125$ of $(d+2)^d$; from $d=4$ on, the argument of~\cite{BHLT26}
gives the smaller values (its stated bound $(d+2)^d$ from $d=5$ on). Their bounds on list replicability appeared first, and
we found this bound on $q(C)$ independently of their work.
Section~\ref{sec:static-comparison} gives the comparison.

The critical-point estimate used for Theorem~\ref{thm:main} is based on the
smoothed norm geometry in the topological Figiel--Lindenstrauss--Milman
theorem of Frick, Hosseini, Myzelev, Narnapatti, and
Vasileuski~\cite{FHMNV26v1}. Their theorem (\cite[Theorem 1]{FHMNV26v1};
\cite[Theorem 2]{FHMNV26} with a different proof and constant $4$) bounds
equivariant index for strongly regular CW complexes. Their paper appeared first. We found independently the index bound
of~\cite{Zhang26}, which has an additional logarithmic factor, and the lower bound for
the index from Hanner polytopes given in the same preprint. In the present proof, logarithmic
barriers impose the coordinate constraints, and the compatible-carrier construction
turns the connected regions into a bound for list replicability.

Our construction adjoins convex hulls so that the fibers of its projection to the
cube are convex. This convex-fiber construction is related to
Vassiliev's simplicial resolutions~\cite{Vas92}. The count of multiple points
in our construction starts from the viewpoint of Goryunov and Mond~\cite{GM93}, who study
image cohomology through multiple-point spaces. Kalai and
Meshulam~\cite{KM08} bound the Leray number of a simplicial image in
terms of fiber multiplicity. In our construction the weighted attachments control
these multiplicities while maintaining the coordinate-face constraints.

The classical convex-geometric background is the theorem of Figiel,
Lindenstrauss, and Milman~\cite{FLM77}. The Hanner construction
of~\cite[Example 3.7 and Remark 2]{FLM77} shows its asymptotic sharpness.
For the index of free simplicial complexes this sharpness appeared first
in~\cite[Remark 9]{FHMNV26v1}. The lower bounds for
list replicability in Theorem~\ref{thm:main-lower}, for total matrices and at every
fixed error, are built on the same construction; see also
Milo~\cite{Milo25a,Milo25b,Milo26}.

\subsection{Organization}
Section~\ref{sec:model} fixes the definitions, and
Section~\ref{sec:construction} outlines the proof of Theorem~\ref{thm:main}.
Section~\ref{sec:structure} states the other results of
Parts~\ref{part:lower} and~\ref{part:topology}, Section~\ref{sec:methods} states those
of Parts~\ref{part:analytic} and~\ref{part:limits}, and Section~\ref{sec:open} lists
open questions. Section~\ref{ssec:2} collects the notation for sign complexes, closed
covers, and power maps used in the rest of the paper. Complete proofs
occupy Parts~\ref{part:upper}--\ref{part:limits}.
Part~\ref{part:upper} proves Theorem~\ref{thm:main}.
Part~\ref{part:lower} proves Theorem~\ref{thm:main-lower} and computes exact
values on basic families.
Part~\ref{part:topology} develops face separation and topological criteria
and proves Theorem~\ref{thm:index-gap}.
Part~\ref{part:analytic} contains the entropy-tilting and combinatorial upper
bounds and applies the combinatorial bounds to Paley matrices. Part~\ref{part:limits} establishes
limits of specific representations.
The proof of Theorem~\ref{thm:main} uses no computer calculation. A few exact
values in Parts~\ref{part:lower}, \ref{part:topology},
and~\ref{part:analytic} rest on, or are confirmed by, finite computations in exact
arithmetic, which Appendix~\ref{sec:computation} lists.

\section{Definitions and basic comparisons}\label{sec:model}
This section fixes the definitions used to state the results of the paper.
Let $C\subseteq\{\pm1\}^n$ be a finite nonempty class. A distribution
$\mu$ on $[n]\times\{\pm1\}$ is realizable by $C$ if some $c\in C$
has $\mu\{(j,y):c(j)\ne y\}=0$. Write $\Delta_C$
for the compact space of such distributions. For any Boolean function
$h$, its population loss is
\[
 \operatorname{loss}_\mu(h)=\Pr_{(j,y)\sim\mu}\{h(j)\ne y\}.
\]
A learning rule takes a finite labeled sample and returns a Boolean
function. Outputs may lie outside $C$.

\begin{definition}[List replicability]
Fix $\varepsilon>0$. A rule $A$ is $(\varepsilon,b)$-list replicable if, for every
$\delta>0$, there is a sample size $T(\delta)$ such that for every
realizable $\mu$ a list $\mathcal L_\mu$ of at most $b$ hypotheses
satisfies
\[
 \operatorname{loss}_\mu(h)\le\varepsilon\quad(h\in\mathcal L_\mu),
 \qquad
 \Pr_{S\sim\mu^{T(\delta)}}\{A(S)\in\mathcal L_\mu\}\ge1-\delta.
\]
The probability includes any randomness of the rule.
Let $\operatorname{lr}(C,\varepsilon)$ be the least such $b$, and put
$\operatorname{lr}(C)=\sup_{\varepsilon>0}\operatorname{lr}(C,\varepsilon)$.
We call $\operatorname{lr}(C)$ the list replicability number of $C$, or briefly
its list number. For a matrix $B$, these quantities refer to its row class $C_B$.
\end{definition}
The rules constructed for Theorem~\ref{thm:main} are deterministic, and their
lists are fixed by $\mu$ and $\varepsilon$. Repeated rows or columns can be
removed without increasing the list number. Unless otherwise specified, classes
are total. A partial row has undefined entries, written $*$; a distribution
realized by that row is supported only on its defined labeled points. In the definition one rule serves all $\delta$, while the list $\mathcal L_\mu$ may depend on $\delta$; Section~\ref{ssec:5.1} compares this with the model of~\cite{BHH26b}, which allows one rule for each pair $(\varepsilon,\delta)$.

A labeled point $(j,s)\in[n]\times\{\pm1\}$ is called an atom when it is regarded as a point of the sample space, and a literal when it is regarded as specifying the face $y_j=s$ of the cube $[-1,1]^n$.

Our upper bounds are proved by constructing closed covers of $\Delta_C$. Let
$R_1,\ldots,R_t$ be a finite closed cover of $\Delta_C$. Its multiplicity is the
maximum number of members containing a single point. Suppose each member has a
Boolean label $h_j$ that is accurate throughout $R_j$. Denote the smallest
multiplicity of such a cover at error $\alpha$ by $\mathrm{LR}(C,\alpha)$ and its
supremum over positive $\alpha$ by $\mathrm{LR}(C)$.
The two quantities compare as follows:
\begin{equation}\label{eq:cover-learning}
 \mathrm{LR}(C,\alpha)\le\operatorname{lr}(C,\alpha),\qquad
 \operatorname{lr}(C,\alpha+\tau)\le\mathrm{LR}(C,\alpha)
 \quad(\tau>0).
\end{equation}
In particular $\mathrm{LR}(C)=\operatorname{lr}(C)$.
The first inequality is proved in Theorem~\ref{res:b}, following
the learner-to-cover construction of~\cite[Theorem 7]{CMY23},
\cite[Corollary 23]{CCMY24}, and \cite[Theorem 4.1]{BHH26b}; see also
\cite[Theorem 2.11]{BHHLT26}. The second is the cover-to-learner direction of
the same characterization~\cite[Theorem 7]{CMY23};
Proposition~\ref{prop:cover-rule} proves it with a positive margin and a
uniform sample bound. Whenever this paper converts a closed cover into a rule,
it leaves such a positive margin in the error.

The general linear bound for finite classes is the following.
\begin{theorem}[{\cite[Theorem E]{CCMY24}}]\label{thm:baseline}
Every nonempty finite total class satisfies
$\operatorname{lr}(C)\le1+\lfloor |C|/2\rfloor$.
\end{theorem}
We also compare the list number with two invariants of sign matrices. The sign
complex $S(C)$ contains the labeled-coordinate simplices of the rows and their
negatives, and it has the involution that flips every label. Its equivariant index
$\operatorname{Ind}S(C)$, also written $\operatorname{Ind}(C)$, is the least $d$ for which it maps
equivariantly to $S^d$. For $0<\varepsilon<1/2$,
\[
 \operatorname{Ind}S(C)\le2\operatorname{lr}(C,\varepsilon)-1,
 \qquad \operatorname{lr}(C)\le\operatorname{signrk}(C).
\]
The first inequality is due to~\cite{BHHLT26} and the second
to~\cite{BHH26a}. Section~\ref{ssec:2} fixes the topological
conventions and the additional notation needed in the proofs.

We write $m=|C|$ and $n$ for the domain size. In the matrix formulation
we write $N$ for the number of rows and $M$ for the number of columns, so that
$N\ge m$, with equality for distinct rows, and $M=n$. We write $\ln$ for the
natural logarithm and $\log_2$ for the logarithm in base two.

\section{Overview of the proof of the upper bound}\label{sec:construction}
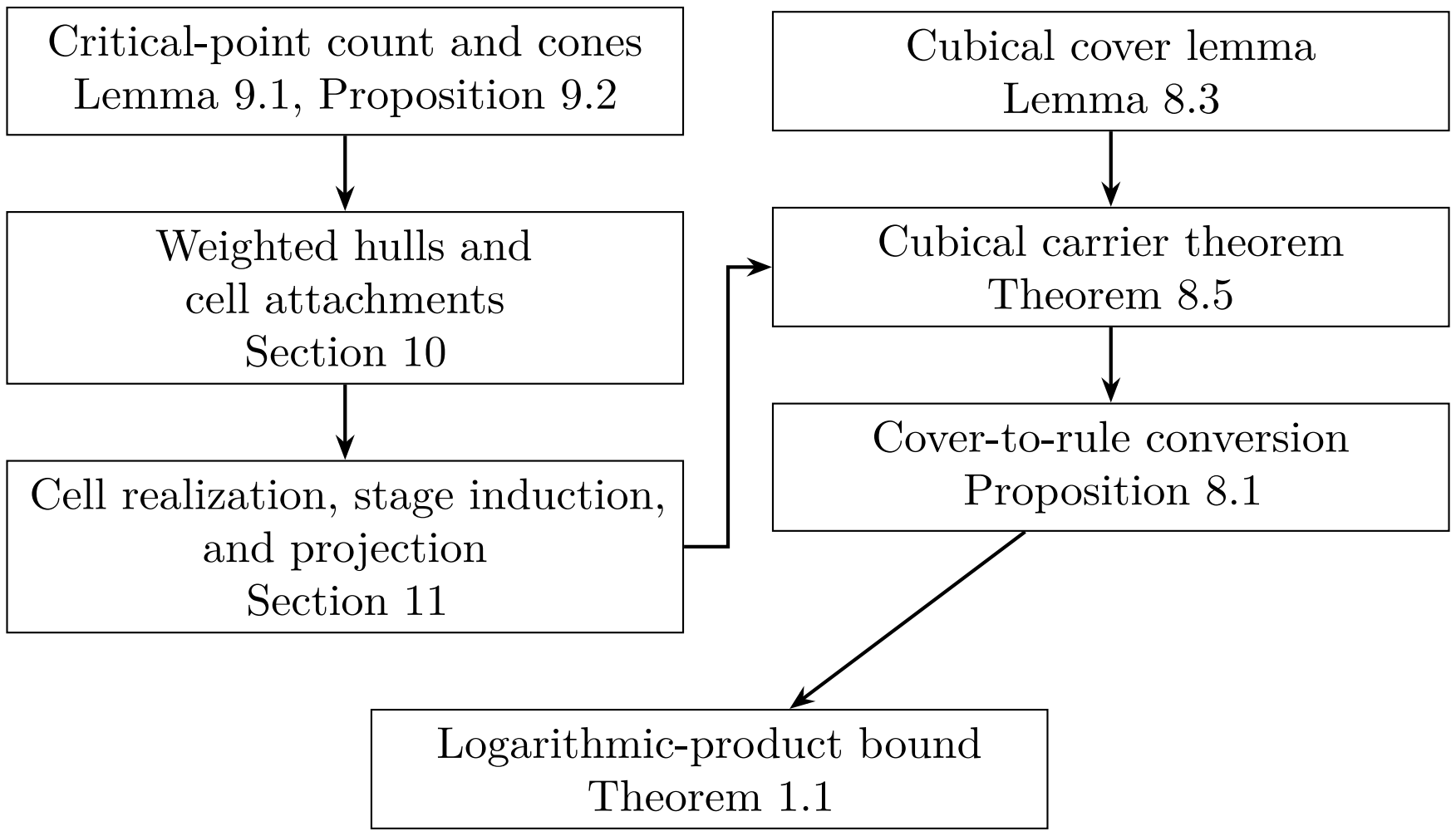
\begin{figure}[htbp]
\centering
\begin{tikzpicture}[node distance=6mm and 7mm,
 box/.style={draw,rectangle,align=center,text width=5.1cm,inner sep=4pt,font=\small},
 arr/.style={-{Stealth[length=2mm]},thick}]
\node[box] (m) {Critical-point count and cones\\Lemma~\ref{lem:morse-count}, Proposition~\ref{prop:cones}};
\node[box,below=of m] (a) {Weighted hulls and cell attachments\\Section~\ref{app:hulls}};
\node[box,below=of a] (s) {Cell realization, stage induction,\\and projection\\Section~\ref{app:construction}};
\node[box,right=of m] (cc) {Cubical cover lemma\\Lemma~\ref{lem:cube-cover}};
\node[box,below=of cc] (car) {Cubical carrier theorem\\Theorem~\ref{thm:carrier}};
\node[box,below=of car] (rule) {Cover-to-rule conversion\\Proposition~\ref{prop:cover-rule}};
\node[box,below=of s,xshift=2.9cm] (main) {Logarithmic-product bound\\Theorem~\ref{thm:main}};
\draw[arr] (m)--(a);\draw[arr] (a)--(s);
\draw[arr] (cc)--(car);
\draw[arr] (s.east)--++(.35,0)|-(car.west);
\draw[arr] (car)--(rule);
\draw[arr] (rule)--(main);
\end{tikzpicture}
\caption{Main dependencies in the proof of Theorem~\ref{thm:main}. The stage
induction and the projection construct the input required by the carrier theorem.
That theorem uses the cubical cover lemma, and the resulting cover gives a rule
by Proposition~\ref{prop:cover-rule}. In the small-dimensional case,
the cubical cover lemma and the cover-to-rule conversion suffice.}
\label{fig:sharp-dependencies}
\end{figure}

This section outlines the proof of Theorem~\ref{thm:main}, from a sign matrix
to a learning rule. Figure~\ref{fig:sharp-dependencies} shows how the main
statements depend on one another. Let $A$ be the matrix whose rows are the members
of $C$, and put $Q=[-1,1]^n$. A literal $(j,s)$ specifies the face $y_j=s$, and a
set of literals is consistent if it uses each coordinate at most once.
For a consistent set $S$ of literals, write $Q_S$ for their common
face and $m_S=n-|S|$ for its dimension. Define
\[
 N[C]=\bigcup_{c\in C}\{y\in\partial Q:c_jy_j\ge0\text{ for every }j\}.
\]
This is the region of the cube boundary whose coordinate signs agree
with at least one concept. A face-preserving map keeps every coordinate
already fixed at $+1$ or $-1$ unchanged.

\subsection{The geometric reduction}
We seek a face-preserving map from $N[C]$ to a low skeleton of $Q$. Two results
explain why such a map suffices and what its construction requires.

\paragraph{From a map to a list.}
By Lemma~\ref{lem:cube-cover}, a continuous face-preserving map $N[C]\to Q^{(d)}$
gives a closed cover of $\Delta_C$ with multiplicity at most $d+1$ and arbitrarily
small positive loss. The cover is labeled by Boolean vertices of a triangulation of
the cube skeleton, and its labels agree with every labeled coordinate whose mass
exceeds a chosen threshold. Proposition~\ref{prop:cover-rule} then turns it
into a deterministic rule with a list of size at most $d+1$.

\paragraph{From carriers to a map.}
The cubical carrier theorem, Theorem~\ref{thm:carrier}, supplies the
map from a nested family of sets outside $N[C]$, which we call carriers. For
$a\ge1$ and $d=n-a-1\ge2$, it asks for compact semialgebraic sets
\[
 T_S\subseteq Q_S\setminus N[C]\qquad(1\le|S|\le a)
\]
with $T_{S'}\subseteq T_S$ when $S\subseteq S'$, and
\[
 \widetilde H_j(T_S;\mathbb Z)=0\qquad(j\le a-|S|-1).
\]
Reduced homology in degree $-1$ records nonemptiness. The conclusion
is a face-preserving map to $Q^{(d)}$, and hence a list bound $n-a$.
This is the central reduction of the proof: the construction of the carriers
needs only the stated integral homology and nesting properties.

\subsection{Connected regions with coordinate constraints}
Fix the row matrix $A$ and a cube face $Q_S$. We seek vectors
whose correlations with all rows are small while the coordinates in
$S$ have prescribed signs. The clamp sends a vector to the point of $Q$ obtained
by dividing it by a small scale and truncating each coordinate to $[-1,1]$. The first
condition keeps the clamped images of these vectors outside $N[C]$, and the second
puts those images in the correct face.

Let $G=\ln(2m)\ln(2n)$, $c'=\lceil800G\rceil$, and $a=n-c'-3$.
The case $a<1$ is covered by the identity map on $N[C]$.
For $a\ge1$, the construction runs through stages $i=1,\ldots,L$, with exponents
$q_i>1$ decreasing toward one; Section~\ref{app:parameters} fixes every
parameter explicitly. Proposition~\ref{prop:cones} constructs open
cones $\mathcal C_{i,S}$ with connectivity at least
\[
 n-3-ic'-|S|.
\]
The cones shrink as $S$ grows. Their vectors have small correlation with all rows,
and their coordinates in $S$ have the prescribed signs and a lower bound on their
size. The proof uses a smoothed $\ell_{q_i}$ sphere and logarithmic barriers for the
literals in $S$. Lemma~\ref{lem:morse-count} bounds the number of nonnegative
Hessian directions at high critical values, so that Morse theory above a chosen
level attaches only cells of high dimension. This gives the required connectivity.

These cones provide room in which later cell boundaries can be
filled. The quantity $ic'$ is the dimension lost at stage $i$.

\subsection{Filling faces while preserving earlier choices}
A cell of stage $i$ at a face $Q_S$ carries two maps on a ball: a base coordinate
into $Q_S$ and a vector coordinate into $\mathcal C_{i,S}$. The boundary of a new cell
is attached to cells already constructed. The interior of a cell is a sheet over the
cube, and several sheets can pass over one base point with different vector values.
At stage $e$ we adjoin the convex hull of any such values whose stage weights sum to
at most $e$. The dimension losses add over the sheets above one cube point, and this
bounds how many sheets can meet there (Lemma~\ref{lem:fiber-weight}).

Three properties make the construction work. First,
Lemma~\ref{lem:weighted-hull} keeps a convex combination of such vectors in
the cone allowed by its total weight. Second,
Lemma~\ref{lem:base-generic} bounds the dimension of the loci where several sheets
have one base point, and Lemma~\ref{lem:height-generic} separates their vectors. Third,
Lemma~\ref{lem:one-cell} realizes an abstract attachment by an
actual cell without changing the sets over faces already completed.
Its proof separately treats the vector coordinate, the base coordinate, and
the two perturbations needed for coincidence and injectivity.

Faces are processed from smaller to larger dimension, and cells are
ordered by stage, face dimension, and cell dimension. At each stage,
Lemma~\ref{lem:killing} removes the required homology by finitely many
attachments. The explicit rank and norm bounds in
Section~\ref{app:construction} ensure that successive attachments
remain in the same shell and outside $N[C]$.
The result is a family of compact hull sets with the required vanishing of
integral homology (Proposition~\ref{prop:stage-induction}).

\subsection{Projection to compatible carriers}
The vector coordinates are retained during the construction to control its
convex hulls. In the final family, the sheet points above a fixed base point
have total weight at most the last stage index $L$, so all their vectors enter
one convex hull. The fiber of the projection to the base is therefore a
nonempty convex set.

The projection is proper, and its fibers are acyclic. The
Vietoris--Begle theorem transfers the vanishing of cohomology, and
triangulability and universal coefficients recover the vanishing of integral
homology. The projected sets are the carriers $T_S$ of
Theorem~\ref{thm:carrier}. They are nested because the hull sets are nested.
Proposition~\ref{prop:carriers} proves these properties.

The carrier theorem now gives multiplicity at most $c'+3$.
The parameter inequalities imply $c'+3\le805G$ for $a\ge1$. Together with the
case $a<1$, this proves the bound on $\mathrm{LR}(C)$ in Theorem~\ref{thm:main}.

\subsection{From the cover to a learning rule}
Fix the desired error $\varepsilon>0$, and construct the closed cover with
label loss at most $\varepsilon/2$. Compactness gives one radius
$\rho\le\varepsilon/2$, valid for all centers, such that every $\rho$-ball around a
realizable distribution meets no more cover members than the multiplicity of the cover.

Given a nonempty realizable sample, form its empirical distribution
and output the label of the first cover member containing it, in a
fixed order. Return a fixed Boolean function on all other samples.
For a true distribution $\mu$, use as its list the labels of members
meeting the $\rho$-ball about $\mu$. Each label has loss at most
$\varepsilon$, and the output belongs to this list whenever the empirical
distribution is within total variation $\rho$ of $\mu$.
Proposition~\ref{prop:cover-rule} gives the uniform sufficient sample
size
\[
 T\ge\left\lceil\frac{\ln(2^{2n+1}/\delta)}{2\rho^2}\right\rceil.
\]
Thus the multiplicity of the cover becomes exactly the required bound
on the number of accurate outputs.

\section{Lower bounds, exact values, and face separation}\label{sec:structure}
\begin{figure}[htbp]
\centering
\begin{tikzpicture}[node distance=6mm and 7mm,
 box/.style={draw,rectangle,align=center,text width=5.4cm,inner sep=4pt,font=\small},
 arr/.style={-{Stealth[length=2mm]},thick}]
\node[box] (p) {Polytope representation\\and signed join\\Theorems~\ref{res:4.1} and~\ref{res:4.2}};
\node[box,below=of p] (h) {Hanner recursion and lower bounds\\Theorem~\ref{res:4.3}};
\node[box,right=of p] (w) {Face models and anchoring\\Theorem~\ref{thm:anchor-width}};
\node[box,below=of w] (g) {Common fillings and exact graph values\\Lemma~\ref{lem:common-filling}, Theorem~\ref{thm:graph-budget}};
\node[box,below=of g] (pr) {Algebraic products\\Lemma~\ref{lem:algebraic-product}, Corollary~\ref{cor:graph-products}};
\node[box,below=of pr] (gap) {Unbounded index gap\\Theorem~\ref{thm:index-gap}};
\draw[arr] (p)--(h);\draw[arr] (w)--(g);\draw[arr] (g)--(pr);\draw[arr] (pr)--(gap);
\end{tikzpicture}
\caption{Main dependencies of the lower bounds and the exact families. The comparison of
the sign complex with list replicability in Section~\ref{ssec:2} supplies the step from
index to learning on the left and the index calculations on the right.}
\label{fig:structural-dependencies}
\end{figure}
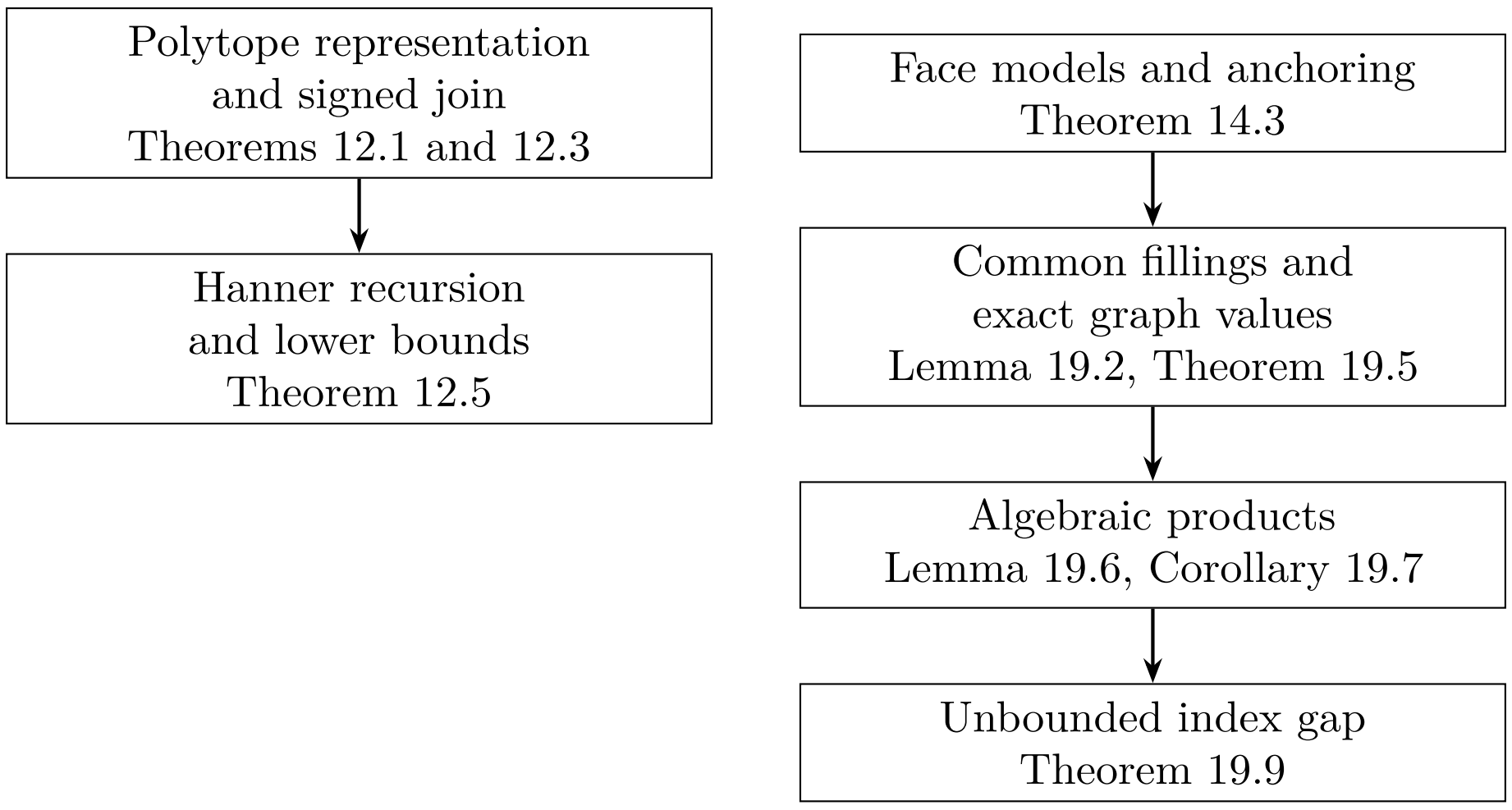

Theorem~\ref{thm:main} bounds the list number of every matrix. This section
describes the constructions that attain its order and that
distinguish list replicability from related invariants. Parts~\ref{part:lower}
and~\ref{part:topology} contain the proofs, and
Figure~\ref{fig:structural-dependencies} shows their main dependencies.

\subsection{Hanner matrices attain the logarithmic product}\label{sec:lower}
We start from a centrally symmetric polytope together with its
vertex--facet incidences. A signed incidence matrix represents its
boundary as a sign complex. The sphere in this complex supplies an
index lower bound, and the inequality
$\operatorname{Ind}\le2\operatorname{lr}-1$ turns it into a lower bound
for list replicability. Products and free sums of Hanner polytopes make the
numbers of rows and columns explicit, which gives
Theorem~\ref{thm:main-lower}. Theorem~\ref{res:4.1} supplies the polytope
representation, Theorem~\ref{res:4.2} proves the signed-join step, and
Theorem~\ref{res:4.3} carries out the full recursion, the rounding estimates, and
the padding to arbitrary square size. These lower bounds remove the
$\log\log N$ factor from~\cite[Proposition 1.16]{BHHLT26}, and together with
Theorem~\ref{thm:main} they determine the order of the largest list number.

\subsection{Exact families and output restrictions}
The exact calculations in Section~\ref{ssec:5} compare list replicability with
related parameters on concrete families. They include anchored cubes, anchored
singletons, Sylvester Hadamard classes, projective-plane classes, and the Fano plane.
The binary-cube and threshold list values appeared first in~\cite{CMY23}, and the fixed-error cube formulation in~\cite{BHH26c}.
Theorems~\ref{res:5.1}--\ref{res:5.6}
prove the anchored value $k+1$ at every fixed error and the restricted-output
separation. They also sharpen two earlier bounds. For the Sylvester Hadamard matrices
$H_{2^m}$, \cite[Theorem 32]{FHV26} gives $m-1\le\operatorname{Ind}\le2m+1$, and we show
$\operatorname{Ind}S(H_{2^m})=\operatorname{coInd}S(H_{2^m})=m$ and
$\operatorname{lr}(H_{2^m},\varepsilon)=m+1$.
For the projective plane of order $q$, \cite[Theorems 1.2 and 3.2]{BHHLT26}
give $\operatorname{lr}\le3$ and $\operatorname{Ind}\le5$, and we show that the
list number is $3$ at every fixed error and that
$\operatorname{Ind}=\operatorname{coInd}=4$ for $q\ge4$.
Proposition~\ref{res:5.8} shows that transposition does not preserve the
list number and that the sign complex does not determine it.

The anchored singleton example illustrates why output freedom matters.
Its list number with unrestricted outputs is two, whereas restricting hypotheses to
the signed class $C\cup(-C)$ forces a list whose size grows with the
number of concepts at small errors. The proof identifies
the distributions at which each restricted output is necessary.
The separation for plain singleton classes appeared first in~\cite{CMY23},
after Theorem 6.

\subsection{Anchoring and face separation}\label{sec:width}
Let $C^+$ denote the class obtained by appending one constant positive
coordinate; we say that $C^+$ is obtained by anchoring $C$. Define the face width
$u(C)$ as the least multiplicity of a finite open
cover of $\operatorname{conv}C$ in which no member meets two opposite
coordinate faces $\{x_j=1\}$ and $\{x_j=-1\}$. Equivalently, $u(C)-1$ is the least
dimension of a finite polyhedral model that preserves all extreme coordinates.
Theorem~\ref{thm:anchor-width} proves
\[
 \operatorname{lr}(C)\le\operatorname{lr}(C^+)=u(C)
 \le\operatorname{lr}(C)+1,
 \qquad
 u\Big(\prod_i C_i\Big)\le1+\sum_i(u(C_i)-1).
\]
Thus a single constant coordinate turns the list number into an exact
geometric quantity. The proof gives both conversions: a model
produces accurate covers through Gibbs weights, and an anchored rule
produces a face-separating cover by a continuous distribution map.
For anchored classes the product inequality for face width reads
$\operatorname{lr}(C_1^+\times C_2^+)\le\operatorname{lr}(C_1^+)+\operatorname{lr}(C_2^+)-1$,
with equality on anchored cubes; compare~\cite[Theorem 1.12]{BHHLT26}.

Two constructive bounds on $u$ complement Theorem~\ref{thm:main}.
A random coordinate-threshold construction gives
\[
 u(C)\le1+7.76(n^2\ln m)^{1/3}\qquad(m\ge2),
\]
proved in Theorem~\ref{thm:volume-bound}. If every row lies within
Hamming distance $r$ of one binary affine subspace of dimension $d$,
then Proposition~\ref{prop:affine-neighborhood} gives
$u(C)\le(r+1)(d+1)$. The first bound uses projected volumes; the
second preserves a coherent choice of affine representatives along
chains of faces.

\subsection{Integral topology and stabilization}
The literal complex of $C$ has as faces the consistent literal sets contained in a
row. A completion of this complex, a complex on the same vertices that contains it
and has no face with two opposite literals, gives another certificate for upper
bounds. If the link of every nonempty face has vanishing integral cohomology in
degrees at least $d\ge2$, Theorem~\ref{thm:literal-upper} gives
$\operatorname{lr}(C)\le d+1$. Including the link of the empty face gives the same
bound for $u(C)$. Theorem~\ref{thm:proper-kernel} also shows that the least support
size of a face-preserving continuous kernel on the original rows equals one
plus the all-field Leray number. In the other direction,
Theorem~\ref{thm:integral-sphere} detects a lower bound from a
shattered sphere whose integral homology class is nonprimitive.
Theorem~\ref{thm:literal-threshold-gap} shows why a further choice
of completion matters: coordinatewise literal thresholds can require
a polynomially large Leray number even for classes whose face width is logarithmic.
These statements keep track of the coordinate constraints that an
unrestricted map of spaces would forget.

For $n\ge5$, Theorem~\ref{thm:codimension-two} gives an exact test
for $\operatorname{lr}(C)\le n-2$: integer masses of total one on
components of missing cube vertices must agree under all one- and
two-coordinate restrictions. Corollary~\ref{cor:facet-test} gives
a short lower-bound certificate from a single facet.

Adding $t$ independent binary coordinates gives a quantitative
stabilization result with a smaller explicit constant. With the integer $D$
defined before Theorem~\ref{thm:stable}, $D<105.09\ln(2m)\ln(2n)$ and
\[
 \operatorname{lr}(C\times\{\pm1\}^t)\le t+D+2
 \quad\text{for }t\ge\max\{0,n-2D-4\}.
\]
The analogous face-width bound holds when $t\ge\max\{0,n-2D-3\}$.
The construction embeds compatible sections once the extra coordinates
provide enough room. Section~\ref{app:stable} proves the connectivity
estimate, the embedding step, and the range conditions on the integers.

\subsection{Orientation classes give exact products and index gaps}\label{sec:orientation}
For a finite connected simple graph $G$ containing a cycle, let $C_G$ be the class
of orientations with an incoming edge at every vertex. The domain
consists of edges. Write $r=|E|-|V|+1$ and let $\tau(G)$ be the minimum
size of a feedback vertex set. Theorem~\ref{thm:graph-budget} gives
\[
 \operatorname{lr}(C_G^+)=u(C_G)=r+\tau(G).
\]
The upper bound protects the vertices outside a feedback
vertex set and retracts coordinate sections along a forest.
The lower bound uses a nonzero cubical cycle whose possible
fillings must satisfy several coordinate constraints simultaneously.
Lemma~\ref{lem:common-filling} states this obstruction precisely.

An algebraic version of the obstruction is additive under products
(Lemma~\ref{lem:algebraic-product}). It gives the exact formula
\[
 u\Big(\prod_iC_{G_i}\Big)=1+\sum_i(r_i+\tau(G_i)-1)
\]
in Corollary~\ref{cor:graph-products}. In particular, let $C_0$ be
the $32$ source-free orientations of $K_4$ on its six edges and put
$D_k=(C_0^k)^+$. Theorem~\ref{thm:index-gap} proves
\[
 |D_k|=32^k,\quad |\operatorname{dom}D_k|=6k+1,\quad
 \operatorname{lr}(D_k)=4k+1,\quad\operatorname{Ind}S(D_k)=3k.
\]
The unanchored product has $\operatorname{lr}(C_0^k)=4k$ by
Proposition~\ref{prop:unanchored-products}. Its upper bound rests
on an exact integral homology calculation for the links of the base class
(Lemma~\ref{lem:k4-links} and Appendix~\ref{sec:computation}).
The additive gap is $\operatorname{lr}(D_k)-(\operatorname{Ind}S(D_k)+1)=k$. Thus the inequality
$\operatorname{Ind}\le2\operatorname{lr}-1$ of~\cite[Theorem 3.1]{BHHLT26}
has no additive converse, which answers negatively the question
of~\cite[Section 9.2]{Zhang26} whether $\operatorname{lr}\le\operatorname{Ind}+O(1)$. Proposition~\ref{prop:cycle-coloring} also bounds
$\operatorname{Ind}(K(C_G)\cap-K(C_G))$ by $r+\chi_c(G)-2$,
where $K(C_G)$ is the literal complex and $\chi_c(G)$ is the cycle-intersection
coloring parameter defined there.

\subsection{Oriented matroids and low index}
Oriented matroids give a further representation parameter, the oriented-matroid
rank $\operatorname{omr}(C)$: the least $r$ such that the rows
and their negatives occur as regions of a rank-$r$ pseudosphere
arrangement. From such an arrangement, Theorem~\ref{thm:om-bound} constructs an
accurate cover of multiplicity at most $r$, and for the full region class the value
is exactly $r$. This refines
$\operatorname{lr}\le\operatorname{signrk}$~\cite[Corollary 2.9]{BHH26a}
and $\operatorname{Ind}+1\le\operatorname{signrk}$~\cite[Corollary 2]{FHV26},
since $\operatorname{omr}\le\operatorname{signrk}$. Theorem~\ref{thm:h8-om} gives a concrete separation:
the order-eight Hadamard matrix has list number and oriented-matroid
rank four, but ordinary sign rank five. Forster's bound~\cite{Forster02} gives
$\operatorname{signrk}(H_8)\ge3$ and $\operatorname{signrk}(H_{16})\ge4$; here
$\operatorname{signrk}(H_8)=5$ and $6\le\operatorname{signrk}(H_{16})\le7$.
Section~\ref{app:om} provides the construction and the exact finite certificates.

The index-one case has a complete description:
Theorem~\ref{thm:index-one} proves that an odd circle map forces
a rank-two sign representation and a common threshold order of the
columns. Proposition~\ref{prop:cyclic-intervals} then determines the
list number, index, and face width of every cyclic-interval class.
At the other end of the dimension range, for nonempty proper antipodal classes on
$n\ge5$ coordinates, Theorem~\ref{thm:corank-paths} shows that an antipodal path among the
missing cube vertices characterizes $\operatorname{Ind}S(C)\le n-3$, while a geodesic
antipodal path characterizes $\operatorname{omr}(C)\le n-2$.
Theorem~\ref{thm:complement-family} uses this distinction to give,
for every $n\ge7$, an explicit class with index $n-3$ and list
number and oriented-matroid rank both $n-1$.

\section{Further upper bounds, Paley matrices, and limits of representations}\label{sec:methods}
The results of this section exploit additional structure of a matrix, or
limit what a specific representation can achieve. Each upper bound starts
from a different certificate for an accurate cover.
Parts~\ref{part:analytic} and~\ref{part:limits} contain the proofs.

\subsection{Entropy tilting transports a fixed-margin cover}
We start with the signed column profile of a realizable distribution, the vector of
its correlations with the rows.
A coordinate power map embeds the profile into a Hilbert sphere
whose image has a controlled finite net. This gives a bounded-multiplicity
cover with a fixed positive accuracy margin. The transport step then
changes the distribution through a support-preserving exponential
tilt. The tilt is a homeomorphism, and its inverse makes the transported
labels accurate at any requested positive error.

Theorem~\ref{res:3.8} combines the tilt identity and cover
transport to give the list bound $2^{31}\ln^2(2m)\ln(2n)$. Lemmas~\ref{res:3.5}--\ref{res:3.7}
establish the invertibility and quantitative loss estimates that
support this transport, and Section~\ref{ssec:3} also contains the margin-cover
lemma and a boosting bound.
Theorem~\ref{res:3.9} shows that $\operatorname{lr}(C,\eta)=\operatorname{lr}(C)$
for $0<\eta<\min(1/2,1/n)$, a partial answer to~\cite[Question 2.4]{BHH26c}.

A two-sided Gibbs model simultaneously records a distribution on rows
and one on signed coordinates. Its coupled relative-entropy identity
controls the error from empirical row sampling. A joint sampling
representation encodes the entire coordinate sample under its exact
law. Theorem~\ref{thm:two-sided-entropy} gives a fixed-scale entropy
bound proportional to $\ln^{3/2}m\ln(2n)$ and the corresponding bound
for $u(C)$. Section~\ref{app:two-sided} states the parameters and
proves the coupling and coding estimates.

\subsection{Closure chains and bounded intersections}\label{sec:static}
A combinatorial construction uses the pattern of common
labeled coordinates among rows. A positive closure is the set of labeled coordinates
common to a nonempty family of rows. Label each positive closure by a
Boolean hypothesis that extends it, with few distinct labels on each
chain. A barycentric cover of the realizable-distribution space
turns a bound on the number of labels along chains into a list bound.
Theorem~\ref{res:6.3} proves, for every total $B$ and $0<\varepsilon<1/2$,
\[
 \operatorname{lr}(C_B,\varepsilon)\le\chi^*(B)
 \le h_+^*(B)\le h(B)\le r_2(B)+1.
\]
The parameter $\chi^*$ is the nonempty-prefix chain-label number, $h_+^*$
the positive-closure height, $h$ the signed-closure height, and
$r_2$ the switching binary rank. Their definitions and the empty-prefix
variant appear in Sections~\ref{ssec:2} and~\ref{ssec:6}.
This refines~\cite[Theorem 1.3]{BHHLT26}, which gives
$\operatorname{lr}(C)\le H(C)=h_+^0(B)$ (Lemma~\ref{res:2.2}): $h_+^*$ drops the
empty prefix, and $\chi^*$ counts distinct labels rather than chain length.

One useful way to construct these labels is to fix a priority order
of hypotheses and use the first one consistent with the current
closure. Let $q(C)$ be the minimum worst-case number of changes along
a closure chain under the requirement that a consistent old label
is retained. Then $\operatorname{lr}(C)\le q(C)+1$.
Let $d$ be the Littlestone dimension of $C$.
Section~\ref{app:static} proves exact additivity of $q$ on products
and a uniform bound $q(C)\le a_{d-1}$ for $d\ge1$, where $a_0=1$ and
$a_{j+1}=a_j(2a_j+1)$, realized by a priority order of robust low-rank
subclasses. This bound is optimal for $d\le2$. With Theorem~\ref{res:5.1} and
the finite-class extremal result of~\cite{CCMY24}, it gives the exact
suprema $2$ and $4$ of list replicability on finite classes with $d=1$
and $d=2$, and $\operatorname{lr}(C)\le22$ at $d=3$. For general $d$ it gives
$\operatorname{lr}(C)\le a_{d-1}+1$.

A doubly exponential bound in $d$ goes back
to Bun, Livni, and Moran~\cite{BLM20}. Blondal, P.~Hatami, Lalov, and
Tretiak~\cite{BHLT26} first proved the bound $(d+2)^d$, and their argument gives values smaller than
$a_{d-1}+1$ from $d=4$ on.
Section~\ref{sec:static-comparison} gives the comparison and derives a
bound of the form $2^{O(d\log d)}$ on $q$ from~\cite[Lemmas 5.5 and 5.7]{BHLT26}.

The results on low dual dimension, the Littlestone dimension of the columns as
functions on the rows, are sharp: dual dimension at most one
gives $q=d\le3$, and dual dimension at most two gives $q\le7$, with
equality on the seven-dimensional cube.

A closure chain splits into rank plateaus, the blocks on which the Littlestone
dimension of the subclass extending the current closure is constant. Adding worst-case
bounds over the plateaus does not determine $q$.
On the exact family of Theorem~\ref{thm:plateau-family}, a suitable label
changes at most $2k$ times along every chain, whereas for every consistent label the
worst-case numbers of changes on the individual rank plateaus sum to at least
$k^2$, with equality for one conservative label. Thus adding local worst cases can
overestimate the total number of changes by a factor that grows with $k$.

When rows differ from a common background on sets with pairwise
intersections of size at most $s$, a two-piece equivariant join gives
$\operatorname{Ind}S(C)\le2s+2$. The inequality of~\cite[Theorem 3.1]{BHHLT26}
applied to Theorem~\ref{res:5.2} gives $2s+3$.
Theorem~\ref{res:6.6} proves the bound and its sharpness. The
more flexible three-piece construction in
Theorem~\ref{res:6.7} separates a residual region that can be
analyzed by additional structure, as for the Paley matrices.

Section~\ref{app:families} adds exact formulas for a cube with
its two constant rows removed, Hamming balls, and their
symmetrizations. It also gives product identities for extremal classes
and for the number of constant runs along closure chains.
These formulas distinguish list replicability from conservative labels
and from the topology of the symmetrization.

Section~\ref{ssec:7.6} shows that covers correct on heavy atoms characterize list
replicability, and Section~\ref{ssec:7.7} bounds the switching rank on classes generated
from finitely many matrices.

\subsection{Paley matrices}
For a Paley matrix $P_p$, inversion mixing relates the residual region
of the three-piece construction to switchable monochromatic squares.
Let $\beta_p$ be their maximum side length and let $E_{q,q}$ be that
residual region. For $p\ge29$ and $q\ge1$,
\[
 \beta_p\le q\quad\Longrightarrow\quad E_{q,q}=\varnothing
 \quad\Longrightarrow\quad\beta_p<4(q+1).
\]
Theorem~\ref{res:8.6} and Corollary~\ref{res:8.7} prove these implications.
Section~\ref{ssec:8} also gives parity-sensitive
spectral inequalities, shattering certificates, and a family showing
that spectral pseudorandomness alone does not force logarithmic index, in
contrast with the bounds $O(\log N)$ for random matrices and the Sylvester Hadamard matrices
in~\cite{FHV26}.
For $P_{13}$, Proposition~\ref{res:8.9} determines the index exactly:
$\operatorname{coInd}S(P_{13})=\operatorname{Ind}S(P_{13})=3$.
The upper certificate consists of an integer Euler-class primitive
and an equivariant collapse, both verified cell by cell by exact computation.

\subsection{Limits of specific representations}
The entropy-tilting argument of Section~\ref{ssec:3} passes through a small net
for a fixed encoding of the realizable distributions, while the proof of
Theorem~\ref{thm:main} does not. Section~\ref{ssec:7} examines what such
encodings can achieve. It establishes quantitative limits for the coordinate
power map, affine statistics, and transport by average advantage,
and it bounds the tilt dimension above by
the fixed-scale covering dimension of the vote image.

Section~\ref{app:vote-geometry} sharpens this comparison for the Sylvester
Hadamard matrices $H_{2^m}$. The actual vote image has sign-compatible cover
dimension exactly $m$, while the full signed hull has dimension at
least $m+\lfloor m/6\rfloor+1$ at the same confidence threshold.
Theorem~\ref{thm:walsh-sign-geometry} proves both assertions.
Theorem~\ref{thm:two-threshold-geometry} and
Proposition~\ref{prop:linear-sign-measurements} quantify two concrete
encodings of the full hull, and
Proposition~\ref{prop:coordinate-cube-bound} bounds coordinate-cube
obstructions by the logarithm of the number of rows.

\section{Summary and open questions}\label{sec:open}
Every finite sign matrix with at least two rows and two columns admits, at every
positive accuracy, a list-replicable learning rule whose
list size is bounded by a constant times the product of the logarithms
of its numbers of rows and columns. The Hanner families attain this
order. The proof preserves the coordinate information needed for
prediction throughout the geometric construction: connected regions
support weighted attachments, compatible carriers yield a cube map,
and a finite accurate cover supplies the output rule.

The exact families separate list replicability from related
parameters. Anchoring identifies the list number with the face width, orientation
classes yield exact product formulas, and the index gap shows the
additional information carried by list replicability beyond the index. Entropy
tilting and closure chains provide complementary
ways to exploit structure beyond the general bound.

Two questions from~\cite[Sections 9.3 and 9.4]{Zhang26} remain open. Does
the chain-label number $\chi^0$ of Section~\ref{ssec:6.2} admit a general
upper bound of order $\log N\log M$ on $N\times M$ sign matrices? It bounds
list replicability from above by Theorem~\ref{res:6.3}, and it can exceed it by
an unbounded amount: on the threshold matrices of Theorem~\ref{res:6.4},
$\chi^0(T_n)=\lfloor\log_2n\rfloor+1$, while $\operatorname{lr}(T_n)=2$ for
$n\ge2$~\cite{CMY23}. What is the growth of the index of the Paley matrices?
Proposition~\ref{res:8.5} gives $\operatorname{Ind}S(P_p)=\Omega(\log p)$, and
the index bound of~\cite{FHMNV26v1}, whose simplicial case also follows from
Theorem~\ref{thm:main} and Theorem~\ref{res:a}, gives $\operatorname{Ind}S(P_p)=O(\log^2p)$.

\clearpage
\section{Preliminaries on sign complexes, covers, and profiles}\label{ssec:2}

A sign matrix gives rise to three objects that recur throughout the paper: its sign complex, the compact space of its realizable distributions, and the profile map from distributions to a cube. This section introduces them and collects the general facts shared by the later sections. The proof of Theorem~\ref{thm:main} in Sections~\ref{app:bridge}--\ref{app:construction} needs from this section only Lemma~\ref{res:2.0}.

Section~\ref{ssec:2.1} identifies the sign complex with the realizable distributions of the signed class and recalls the comparisons among its index and coindex, list replicability, VC dimension, and sign rank; these comparisons are the source of the lower bounds in Sections~\ref{ssec:4} and~\ref{ssec:5}. Section~\ref{ssec:2.2} shows that list replicability and the least multiplicity of an accurate closed cover agree up to an arbitrarily small loss margin, so that upper bounds can be proved by constructing covers. Section~\ref{ssec:2.3} treats transposition and completion, which are used in Sections~\ref{ssec:4}, \ref{ssec:5}, \ref{ssec:3}, and~\ref{ssec:6}. Section~\ref{ssec:2.4} fixes the closure posets and height parameters used in Sections~\ref{ssec:6} and~\ref{app:static}. Section~\ref{ssec:2.5} sets up the profile map and its coordinate powers for Sections~\ref{ssec:3}, \ref{ssec:7}, and~\ref{app:vote-geometry}.

As in Section~\ref{sec:model}, \(\ln\) denotes the natural logarithm and \(\log_2\) the binary logarithm. Maps between topological spaces are continuous, and an equivariant map between \(\mathbb Z_2\)-spaces is a continuous map commuting with the two involutions.

\subsection{Sign complexes and indices}\label{ssec:2.1}

We recall the sign complex of a matrix and its equivariant invariants, and we relate them to list replicability. We use the sign-complex definition of \cite{FHV26}, written with our convention of \(N\) rows and \(M\) columns.

\begin{definition}[Sign complex; {\cite[Section 3.1]{FHV26}}]
Let \(A\in\{-1,1,*\}^{N\times M}\) be a partial sign matrix. For each row \(r\in\{-1,1,*\}^M\), define the simplex
\[
\sigma_r^+
:=
\{i^+:r_i=+1\}\cup\{i^-:r_i=-1\},
\]
and its antipodal copy
\[
\sigma_r^-
:=
\{i^-:r_i=+1\}\cup\{i^+:r_i=-1\}.
\]
The sign complex \(S(A)\) is the simplicial complex on vertex set
\[
\{1^+,1^-,\ldots,M^+,M^-\}
\]
whose simplices are all subsets of \(\sigma_r^+\) or \(\sigma_r^-\) for some row \(r\) of \(A\). The involution exchanging \(i^+\) and \(i^-\) defines a free \(\mathbb Z_2\)-action on \(S(A)\).
\end{definition}

Writing \(A_{ri}\) for the entry \(r_i\) of row \(r\), the two definitions read, for \(s\in\{\pm1\}\),
\begin{equation}\label{eq:signed-row-simplex}
\sigma_r^s=\{i^b:\ A_{ri}\ne*,\ sA_{ri}=b\}.
\end{equation}
Such a set contains at most one of \(i^+\) and \(i^-\), so no simplex of \(S(A)\) contains an antipodal pair. The involution carries \(\sigma_r^s\) onto \(\sigma_r^{-s}\), so it acts simplicially on \(S(A)\). It is free: a fixed point of \(|S(A)|\) would have a nonempty support \(F\) with \(-F=F\), and \(F\) would then contain an antipodal pair.

The realization of the sign complex is a space of distributions. Let \(C_A\) be the class of rows of \(A\), and let \(C_A^\pm=C_A\cup\{-c:c\in C_A\}\). A row realizes a distribution on \([M]\times\{\pm1\}\) exactly when the distribution is supported on the defined labeled points of that row (Section~\ref{sec:model}). Identify the vertex \(i^y\) with the labeled point \((i,y)\). The defined labeled points of the row \(r\) and of its negative are then \(\sigma_r^+\) and \(\sigma_r^-\), so a distribution belongs to \(\Delta_{C_A^\pm}\) if and only if its support is a simplex of \(S(A)\). A point of \(|S(A)|\) is a convex combination of the vertices of a simplex. Thus \(|S(A)|=\Delta_{C_A^\pm}\), with the standard distribution realization of a simplex, and the involution of \(|S(A)|\) becomes the reversal of all labels.

\begin{definition}[Index and coindex; {\cite[Definition 2.9]{BHHLT26}}]
Let \(A\) be a partial sign matrix and let \(C_A\) be its associated partial concept class. Then we define
\[
\operatorname{Ind}_{\mathbb Z_2}(A)
:=
\operatorname{Ind}_{\mathbb Z_2}(C_A)
:=
\operatorname{Ind}_{\mathbb Z_2}(\Delta_{C_A^\pm}),
\]
\[
\operatorname{coInd}_{\mathbb Z_2}(A)
:=
\operatorname{coInd}_{\mathbb Z_2}(C_A)
:=
\operatorname{coInd}_{\mathbb Z_2}(\Delta_{C_A^\pm}).
\]
\end{definition}

We suppress the subscript \(\mathbb Z_2\). The index is the least sphere dimension admitting an equivariant map from the space; the coindex is the greatest sphere dimension admitting an equivariant map to the space. These definitions refer to nonempty spaces, and all sphere lower bounds below concern such spaces. For the empty complex we use the conventions \(\operatorname{Ind}(\varnothing)=\operatorname{coInd}(\varnothing)=-1\). With them, an empty piece contributes the dimension cost \(\operatorname{Ind}(\varnothing)+1=0\) in the piecewise join bounds of Sections~\ref{ssec:6.4}--\ref{ssec:6.5}, so these bounds need no separate treatment of empty pieces. Elsewhere we isolate the empty case explicitly.

The coindex and the index are separated by a cohomological invariant, which enters the exact computations of Sections~\ref{ssec:5.6} and~\ref{ssec:8.6}. For a nonempty finite free $\mathbb Z_2$-complex $X$, let
$w_1(X)\in H^1(X/\mathbb Z_2;\mathbb F_2)$ be the first
Stiefel--Whitney class of its double cover $X\to X/\mathbb Z_2$. The Stiefel--Whitney
height of $X$ is $\operatorname{swh}X=\max\{j\ge0:w_1(X)^j\ne0\}$; the maximum
exists, because $w_1(X)^0=1\ne0$ and the cohomology of $X/\mathbb Z_2$
vanishes above $\dim X$.
Naturality and the cohomology of real projective space give
$\operatorname{coInd}X\le\operatorname{swh}X\le\operatorname{Ind}X$.
In particular, the coindex never exceeds the index.

These inequalities are standard, and we recall the proof. An equivariant map $f:X\to Y$ of free $\mathbb Z_2$-complexes
induces $\bar f:X/\mathbb Z_2\to Y/\mathbb Z_2$. It also identifies the double
cover of $X$ with the pullback under $\bar f$ of the double cover of $Y$:
the map $x\mapsto([x],f(x))$ is bijective on each two-point fiber, since
$f(-x)=-f(x)\ne f(x)$.
Therefore $\bar f^*w_1(Y)=w_1(X)$ by naturality~\cite[\S4]{MS74}.

For the
antipodal sphere $S^d$ put $a_d=w_1(S^d)\in H^1(\mathbb{RP}^d;\mathbb F_2)$.
The ring $H^*(\mathbb{RP}^d;\mathbb F_2)$ is $\mathbb F_2[a]/(a^{d+1})$
with $a$ of degree one~\cite[Section 3.2]{Hatcher02}. For $d\ge1$ the cover
$S^d\to\mathbb{RP}^d$ is connected, hence nontrivial; since $w_1$ vanishes only on
trivial double covers, $a_d\ne0$, and so $a_d=a$. For $d=0$
both $a_0$ and $a$ vanish. In every case $a_d^d\ne0$ and $a_d^{d+1}=0$.

If $f:X\to S^d$ is equivariant, then
$w_1(X)^{d+1}=\bar f^*(a_d^{d+1})=0$, so $\operatorname{swh}X\le d$.
If $g:S^d\to X$ is equivariant, then
$\bar g^*(w_1(X)^d)=a_d^d\ne0$, so $w_1(X)^d\ne0$ and
$\operatorname{swh}X\ge d$. These two implications give the stated inequalities.

We turn to closed covers of simplicial complexes. The following elementary lemma supplies the closed covers of bounded multiplicity used throughout the paper.

\begin{resultenv}{Lemma}{res:2.0}{Maximal barycentric regions}
Let \(K\) be a finite simplicial complex of dimension \(d\) with vertex set \(V\), and for \(v\in V\) let \(\beta_v:|K|\to[0,1]\) be the barycentric coordinate function of \(v\). The sets
\[
D_v=\{x\in|K|:\beta_v(x)\ge\beta_u(x)\text{ for all }u\in V\}
\]
are closed, cover \(|K|\), have multiplicity at most \(d+1\) (every point of \(|K|\) lies in at most \(d+1\) of them), and satisfy \(D_v\subseteq\operatorname{St}(v)\), the closed star of \(v\) (the union of the closed simplices containing \(v\)). We call \(D_v\) the region on which the barycentric coordinate of \(v\) is maximal.
\end{resultenv}

\begin{proof}
Each function \(\beta_u\) is affine on the closed simplices containing \(u\) and vanishes on the others, so it is continuous on \(|K|\). Hence each \(D_v\), a finite intersection of the closed sets \(\{\beta_v-\beta_u\ge0\}\), is closed. The set \(V\) is finite, so every point has a maximal barycentric coordinate, and the sets \(D_v\) cover \(|K|\).

Let \(x\in D_v\), and let \(\sigma\) be the unique simplex containing \(x\) in its relative interior, so that \(\beta_u(x)>0\) exactly for the vertices \(u\) of \(\sigma\). Since \(\sum_u\beta_u(x)=1\) and at most \(\dim\sigma+1\) coordinates are nonzero, the maximal coordinate satisfies \(\beta_v(x)\ge1/(\dim\sigma+1)>0\), so \(v\) is a vertex of \(\sigma\). Consequently \(x\) lies in at most \(\dim\sigma+1\le d+1\) of the sets \(D_v\), and \(x\in\sigma\subseteq\operatorname{St}(v)\).
\end{proof}

Our covers are always formed by the regions \(D_v\), because the closed stars of all vertices may have multiplicity far exceeding \(d+1\): in a star graph the center lies in every closed star. Closed stars of a subdivision are used below only to control oscillations and diameters, through the inclusion \(D_v\subseteq\operatorname{St}(v)\).

Equivariant maps are related to list replicability by the following index bound, which turns the topological lower bounds of this paper into lower bounds for learning.

\begin{resultenv}{Theorem}{res:a}{\cite[Theorem 3.1]{BHHLT26}}
For every (partial) concept class \(C\) and error parameter \(\varepsilon\in(0,1/2)\),
\[
\operatorname{Ind}_{\mathbb Z_2}(C)
\le2\operatorname{lr}(C,\varepsilon)-1.
\]
\end{resultenv}

We also use two comparisons with the VC dimension \(\operatorname{vc}\):
\[
\operatorname{vc}(C)\le\operatorname{lr}(C),
\qquad
\operatorname{VC}(A)-1\le\operatorname{coInd}(S(A)),
\]
where $\operatorname{VC}(A)=\operatorname{vc}(C_A)$, from \cite[Theorem 3]{CMY23} and \cite[Proposition 6]{FHV26}, respectively. The finite-domain bound
\[
\operatorname{lr}(C)\le\operatorname{signrk}(C)
\]
is \cite[Corollary 2.9]{BHH26a}.
The sign rank \(\operatorname{signrk}(C)\) is the least \(r\) admitting vectors \(w_c,v_x\in\mathbb R^r\) with \(c(x)\langle w_c,v_x\rangle>0\) for all \(c\in C\) and \(x\in X\).

Lower bounds on the coindex often come from a direct combinatorial condition on the columns. A set of \(k\) columns of a sign matrix \(A\) is \emph{signed-shattered} if every \(\varepsilon\in\{\pm1\}^k\) occurs as the restriction of some signed row \(\pm A_{r,\cdot}\) to these columns. In that case the boundary \(\partial\Diamond_k\) of the \(k\)-dimensional cross-polytope, which is equivariantly \(S^{k-1}\), is an equivariant subcomplex of \(S(A)\), so \(\operatorname{coInd}(S(A))\ge k-1\).

To see this, let \(i_1,\ldots,i_k\) be the columns. If \(sA_{r,\cdot}\) restricts to \(\varepsilon\), then \(\{i_j^{\varepsilon_j}:j\in[k]\}\subseteq\sigma_r^s\) by \eqref{eq:signed-row-simplex}, so this set is a simplex of \(S(A)\). The facets of \(\partial\Diamond_k\), for \(\Diamond_k=\operatorname{conv}\{\pm e_1,\ldots,\pm e_k\}\), are the sets \(\{\varepsilon_je_j:j\in[k]\}\), so \(\pm e_j\mapsto i_j^\pm\) embeds \(\partial\Diamond_k\) in \(S(A)\) as a subcomplex and commutes with the involutions. Radial projection \(\partial\Diamond_k\to S^{k-1}\) is an equivariant homeomorphism, so we obtain an equivariant map \(S^{k-1}\to|S(A)|\). This criterion recurs in Sections~\ref{ssec:5}, \ref{app:orientations}, \ref{app:om}, \ref{app:low-index}, \ref{ssec:6}, \ref{app:families}, and~\ref{ssec:8}.

\subsection{Closed covers and positive error margins}\label{ssec:2.2}
List replicability is defined through learning rules, whereas our upper bounds are obtained by constructing closed covers of the space of realizable distributions. This subsection compares the list number with the least multiplicity of an accurate closed cover, first for unrestricted outputs and then for outputs in a prescribed set of hypotheses. Both directions of the comparison are used: the upper bounds pass from covers to rules, while exact values such as those of Sections~\ref{ssec:5}, \ref{app:width}, and~\ref{app:literal} also require the passage from rules to covers.

We first record the elementary facts about realizable distributions used in the cover arguments below and in Section~\ref{ssec:3}. Let \(C\) be a nonempty finite class on a domain \(X\) of size \(M\), possibly partial, with at least one defined value, so that \(\Delta_C\ne\varnothing\), and put \(\Omega=X\times\{\pm1\}\); the elements of \(\Omega\) are the labeled points, or \emph{atoms}. The atom set of a concept \(c\) is \(\sigma_c=\{(i,c(i)):c(i)\ne*\}\).

On distributions on \(\Omega\) we use the distance \(\|\mu-\nu\|_1=\sum_{z\in\Omega}|\mu_z-\nu_z|\). For a closed set \(G\) of distributions put \(\operatorname{dist}_1(\mu,G)=\min_{\nu\in G}\|\mu-\nu\|_1\), with \(\operatorname{dist}_1(\mu,\varnothing)=+\infty\). An atom \((i,y)\) is \emph{erroneous} for a Boolean hypothesis \(h\) if \(h(i)\ne y\). The empirical distribution of a sample \(S\) of size \(n\ge1\) is \(\widehat\mu_S\), with \((\widehat\mu_S)_z=|\{t:S_t=z\}|/n\).

\begin{lemma}[Realizable distributions]\label{lem:distribution-facts}
With this notation:
\begin{enumerate}[label=(\alph*),leftmargin=2em]
\item A distribution belongs to \(\Delta_C\) if and only if its support is contained in \(\sigma_c\) for some \(c\in C\). Hence realizability depends only on the support, \(\Delta_C\) is compact, a member of \(\Delta_C\) has at most \(M\) atoms in its support, and every sample of positive probability under \(\mu\in\Delta_C\) has its empirical distribution in \(\Delta_C\).
\item If \(\mu\in\Delta_C\) and \(\operatorname{loss}_\mu(h)>t\ge0\), then \(\mu_z>t/M\) for some atom \(z\) erroneous for \(h\).
\item For every hypothesis \(h\) and all distributions \(\mu,\nu\) on \(\Omega\), \(|\operatorname{loss}_\mu(h)-\operatorname{loss}_\nu(h)|\le\|\mu-\nu\|_1\).
\item If \(S\) consists of \(n\) independent draws from a distribution \(\mu\) on \(\Omega\) and \(\rho>0\), then
\[
\mathbb P\bigl(\|\widehat\mu_S-\mu\|_1>\rho\bigr)\le4M\exp\!\left(-\frac{n\rho^2}{2M^2}\right).
\]
\item If \(G_1,\ldots,G_t\) are closed subsets of \(\Delta_C\) and every point of \(\Delta_C\) lies in at most \(b\) of them, then there is \(d>0\) such that for every \(\mu\in\Delta_C\) at most \(b\) indices \(j\) satisfy \(\operatorname{dist}_1(\mu,G_j)<d\).
\end{enumerate}
\end{lemma}

\begin{proof}
(a) A concept \(c\) realizes \(\mu\) exactly when \(\mu\) has no mass outside \(\sigma_c\). Thus \(\Delta_C\) is the union of the finitely many closed faces of the probability simplex on \(\Omega\) determined by the sets \(\sigma_c\), and it is compact. The set \(\sigma_c\) has at most one atom over each domain point, so at most \(M\) atoms. A sample of positive probability under \(\mu\) consists of atoms in the support of \(\mu\), so its empirical distribution has its support in the same \(\sigma_c\).

(b) By (a) the erroneous atoms of positive mass number at most \(M\), and their total mass is \(\operatorname{loss}_\mu(h)>t\), so one of them has mass above \(t/M\).

(c) The loss is the sum of the masses of the erroneous atoms, so the difference of two losses is at most \(\sum_z|\mu_z-\nu_z|\).

(d) For one atom \(z\), the quantity \((\widehat\mu_S)_z\) is an average of \(n\) independent indicators with mean \(\mu_z\). Hoeffding's inequality gives \(\mathbb P(|(\widehat\mu_S)_z-\mu_z|>\rho/(2M))\le2\exp(-2n\rho^2/(4M^2))\). A union bound over the \(2M\) atoms gives the displayed bound. Outside this event, \(\|\widehat\mu_S-\mu\|_1\le2M\cdot\rho/(2M)=\rho\).

(e) If at most \(b\) of the sets are nonempty, any \(d>0\) works. Otherwise, consider a set \(J\) of \(b+1\) indices with all \(G_j\) \((j\in J)\) nonempty. The function \(\mu\mapsto\max_{j\in J}\operatorname{dist}_1(\mu,G_j)\) is continuous on the compact space \(\Delta_C\). It is positive everywhere: if it vanished at \(\mu\), then \(\mu\) would lie in all \(G_j\), \(j\in J\), because these sets are closed, contrary to the hypothesis. Let \(m_J>0\) be its minimum and let \(d\) be the least \(m_J\) over the finitely many such \(J\). If \(b+1\) indices had distance below \(d\) at some \(\mu\), their sets would be nonempty, and the corresponding maximum would be less than \(m_J\), a contradiction.
\end{proof}

For a finite class, write $\mathrm{LR}(C,\varepsilon)$, as in Section~\ref{sec:model}, for the least
multiplicity of a finite closed cover of $\Delta_C$ by sets, each labeled with a
Boolean hypothesis whose loss is at most $\varepsilon$ throughout the set. The following
theorem shows that this cover number and the list number agree up to a positive margin in the error.
\begin{resultenv}{Theorem}{res:b}{Cover comparison}
For every finite class $C$ and $\varepsilon,\tau>0$,
\[
\mathrm{LR}(C,\varepsilon)\le\operatorname{lr}(C,\varepsilon),\qquad
\operatorname{lr}(C,\varepsilon+\tau)\le\mathrm{LR}(C,\varepsilon).
\]
Consequently $\sup_{\varepsilon>0}\mathrm{LR}(C,\varepsilon)
=\operatorname{lr}(C)$.
\end{resultenv}
\begin{proof}
\proofstep{From a rule to a cover.}
Let $L=\operatorname{lr}(C,\varepsilon)$; we may assume $L<\infty$. Fix a rule $A$ witnessing this
value. Choose $0<\delta<1/(L+1)$ and a sample size $s$ at which its
list guarantee holds with failure probability at most $\delta$: for every
$\mu\in\Delta_C$ there is a list $\mathcal L_\mu$ of at most $L$ hypotheses of
$\mu$-loss at most $\varepsilon$ with $\Pr_{S\sim\mu^s}\{A(S)\in\mathcal L_\mu\}\ge1-\delta$.
Since $\Delta_C\ne\varnothing$ and $\delta<1$, such a list is nonempty, so $L\ge1$.

For every Boolean hypothesis $h$, write
$p_h(\mu)=\Pr[A(S)=h]$, where $S\sim\mu^s$ and the probability includes
any internal randomness. The sample alphabet is finite, so $p_h$ is
a polynomial in $\mu$: sum over the ordered samples and multiply their
probabilities by the conditional probability of output $h$,
\[
 p_h(\mu)=\sum_{S\in\Omega^s}\Pr[A(S)=h\mid S]\prod_{t=1}^s\mu_{S_t}.
\]
Consequently the sets
\[
 G_h=\{\mu\in\Delta_C:p_h(\mu)\ge(1-\delta)/L\}
\]
are closed, and there are finitely many of them, one for each Boolean hypothesis.

Fix $\mu\in\Delta_C$. The accurate list $\mathcal L_\mu$ has at most $L$ members
and carries probability at least $1-\delta$. Some member therefore has
probability at least $(1-\delta)/L$, which proves coverage. This
threshold exceeds $\delta$, because $(L+1)\delta<1$. A hypothesis outside $\mathcal L_\mu$ has output probability at most $\delta$, since the list carries probability at least $1-\delta$. Thus membership $\mu\in G_h$ forces
$h\in\mathcal L_\mu$. It follows that every label $h$ with $\mu\in G_h$ has loss at most
$\varepsilon$ at $\mu$, and that $\mu$ lies in at most $L$ of the sets.

This is the
learner-to-cover construction of~\cite[Theorem 4.1]{BHH26b}; see
also~\cite[Theorem 7]{CMY23}, \cite[Corollary 23]{CCMY24},
and~\cite[Theorem 2.11]{BHHLT26}. For later use, note that if the rule outputs only members of a set $H$ of hypotheses,
then $p_h\equiv0$ for $h\notin H$. Since the threshold is positive,
the same argument then yields $G_h=\varnothing$ for $h\notin H$.

\proofstep{From a cover to a rule, and the supremum.}
Proposition~\ref{prop:cover-rule} of Section~\ref{app:bridge}, applied with $\alpha=\varepsilon$ to a cover of multiplicity $\mathrm{LR}(C,\varepsilon)$, proves the converse with the positive
margin $\tau$, including a uniform sample bound. That proposition is stated for total classes. For a partial class, merge the cover members with equal labels, which does not increase the multiplicity, and apply the second inequality of Proposition~\ref{prop:restricted-cover} with $H=\{\pm1\}^X$. The proof of that inequality uses only Lemma~\ref{lem:distribution-facts}, so it does not use totality and does not depend on the present theorem.

Taking suprema over
positive errors gives the equality. By the first inequality,
$\sup_{\varepsilon>0}\mathrm{LR}(C,\varepsilon)\le\operatorname{lr}(C)$. By the second, with
$\tau=\varepsilon/2$, we have
$\operatorname{lr}(C,\varepsilon)\le\mathrm{LR}(C,\varepsilon/2)\le\sup_{\varepsilon'>0}\mathrm{LR}(C,\varepsilon')$ for every $\varepsilon>0$.
\end{proof}

The comparison also applies to restricted outputs; compare~\cite[Theorem 4.1]{BHH26b}. This form is needed in Section~\ref{ssec:5.4}, which compares outputs confined to the signed class with unrestricted outputs. For a set \(H\subseteq\{\pm1\}^X\) of admissible hypotheses, not necessarily containing \(C\), let \(\operatorname{lr}_H(C,\varepsilon)\) be the least \(L\) admitting an \((\varepsilon,L)\)-list-replicable learner for \(C\) whose outputs always lie in \(H\), so that the lists consist of members of \(H\). Let \(c_H^{\mathrm{cl}}(C,\varepsilon)\) be the least multiplicity of a finite closed cover \(\{G_h\}_{h\in H}\) of \(\Delta_C\) with \(G_h\subseteq\{\mu:\operatorname{loss}_\mu(h)\le\varepsilon\}\).

\begin{proposition}[Restricted-output cover comparison]\label{prop:restricted-cover}
With the notation just introduced,
\[
c_H^{\mathrm{cl}}(C,\varepsilon)
\le\operatorname{lr}_H(C,\varepsilon),
\]
and
\[
\operatorname{lr}_H(C,\varepsilon+\tau)
\le c_H^{\mathrm{cl}}(C,\varepsilon)
\qquad(\forall\tau>0).
\]
For restricted hypothesis sets, exact equivalence at the same error can fail: there are \(C\), \(H\), \(\varepsilon\) with \(c_H^{\mathrm{cl}}(C,\varepsilon)\le3\) and \(\operatorname{lr}_H(C,\varepsilon)=\infty\).
\end{proposition}

\begin{proof}
\proofstep{From a restricted rule to a cover.}
The probability-threshold argument in the first step of the proof of Theorem~\ref{res:b}, which does not use the rest of that proof, gives the
first inequality with labels in $H$: for a learner with outputs in $H$ and list size $L=\operatorname{lr}_H(C,\varepsilon)<\infty$, the sets $G_h$ of that proof are closed, cover $\Delta_C$, have multiplicity at most $L$, satisfy $G_h\subseteq\{\mu:\operatorname{loss}_\mu(h)\le\varepsilon\}$, and are empty for $h\notin H$.

\proofstep{From a cover to a restricted rule.}
We may assume that $c_H^{\mathrm{cl}}$ is finite, since otherwise there is nothing to prove. Then there is a nonempty finite closed
cover $\{G_h\}$ of multiplicity $b=c_H^{\mathrm{cl}}$. We have $H\ne\varnothing$ because $\Delta_C\ne\varnothing$, so we can fix a default
label $h_0\in H$ and an order of $H$. By compactness choose $\rho>0$ so that
$2\rho\le\tau$ and the family of closed $2\rho$-neighborhoods of
the original $G_h$, in the $\ell_1$ metric on $\Delta_C$, still has
multiplicity at most $b$: it suffices that $2\rho<d$ for the constant $d$ of Lemma~\ref{lem:distribution-facts}(e).

For each true distribution $\mu$, use the fixed list
\[
 \mathcal L_\mu=\{h:\operatorname{dist}_1(\mu,G_h)\le2\rho\}.
\]
Its size is at most $b$, and every member has loss at most
$\varepsilon+2\rho\le\varepsilon+\tau$ by the $1$-Lipschitz
dependence of loss on the distribution (Lemma~\ref{lem:distribution-facts}(c)), applied to $\mu$ and a point of $G_h$ at distance at most $2\rho$.

The rule is defined as follows. On an empirical distribution $\widehat\mu$, output the first label
whose original set $G_h$ has distance at most $\rho$ from
$\widehat\mu$; if there is none, and on the empty sample, output $h_0$.
On the event $\|\widehat\mu-\mu\|_1\le\rho$, coverage makes such
a label available: $\mu$ lies in some $G_{h'}$, so $\operatorname{dist}_1(\widehat\mu,G_{h'})\le\rho$. The selected label $h$ has $\operatorname{dist}_1(\widehat\mu,G_h)\le\rho$, hence $\operatorname{dist}_1(\mu,G_h)\le2\rho$ by the triangle inequality, and so it belongs to $\mathcal L_\mu$.
By Lemma~\ref{lem:distribution-facts}(d), the sample size $\max\{1,\lceil(2|X|^2/\rho^2)\ln(4|X|/\delta)\rceil\}$, uniform over $\mu$, makes this event have probability at least $1-\delta$, for each
$\delta>0$. All outputs lie in $H$.

\proofstep{Failure at the same error.}
Take
\[
C=\{+++, -++, +-+, ++-\},
\]
let \(H\) consist of the three single-flip hypotheses, and set \(\varepsilon=1/3\). Consider the realizable distributions \(\mu_w\) carried by the three points with the all-positive labels and weights \(w=(w_1,w_2,w_3)\); the flip of point \(i\) has loss \(w_i\). On the all-positive simplex, some \(w_i\le1/3\). On a simplex realized by a single-flip row, that row itself has zero loss and belongs to \(H\). By Lemma~\ref{lem:distribution-facts}(a), \(\Delta_C\) is the union of these four simplices. Thus the three closed sets \(G_h=\{\mu\in\Delta_C:\operatorname{loss}_\mu(h)\le1/3\}\) cover the entire realizable space, and their multiplicity is at most three.

Nevertheless, a restricted learner at this exact error cannot uniformly select an accurate member near the uniform point. Suppose a rule with outputs in \(H\) is \((1/3,L)\)-list replicable for some finite \(L\), and fix \(\delta<1/2\) and its sample size \(n(\delta)\). Removing from each list the hypotheses outside \(H\) does not change the probability of the list, so all lists consist of members of \(H\) of loss at most \(1/3\). For weights with \(w_j,w_k>1/3\) the only accurate member of \(H\) is the flip of the remaining point \(i\), so the learner must output that flip with probability at least \(1-\delta\).

The output probabilities are polynomials in \(\mu\), as in the proof of Theorem~\ref{res:b}, and are therefore continuous in \(w\) for fixed \(n(\delta)\). Letting \(w_i=1/3-2t\), \(w_j=w_k=1/3+t\) with \(t\downarrow0\), we find that at the uniform point the flip of \(i\) has probability at least \(1-\delta\). This holds for each of the three flips, so \(3(1-\delta)\le1\), which is impossible for \(\delta<1/2\). This continuity argument gives
\[
c_H^{\mathrm{cl}}\le3,
\qquad
\operatorname{lr}_H=\infty.
\]
Thus the margin cannot be deleted from this comparison.
\end{proof}

\subsection{Transposition and completion}\label{ssec:2.3}

Transposition exchanges domain points and concepts, and it can change list replicability (Proposition~\ref{res:5.8}). The sign complex behaves better: we prove that \(S(A)\) and \(S(A^t)\) are equivariantly homotopy equivalent, so the index and the coindex do not distinguish a matrix from its transpose. The equivalence is induced by a pair of order-reversing maps of face posets whose composites are closure operators, and we first record the homotopy that such an operator provides. The subsection ends with the behavior of the invariants under completion of a partial matrix.

Let \(K\) be a finite simplicial complex with a simplicial involution, and write \(-x\) and \(-F\) for the images of a point \(x\) and a face \(F\). Let \(P(K)\) be the poset of nonempty faces, ordered by inclusion, and let \(b_F\) be the barycenter of \(F\) in \(|K|\). The barycentric subdivision of \(K\) has the chains of \(P(K)\) as its simplices, and the affine extension of \(F\mapsto b_F\) is a homeomorphism of its realization onto \(|K|\). Thus every \(x\in|K|\) has a unique expression
\[
x=\sum_{j=0}^l\lambda_jb_{F_j},
\qquad
F_0\subsetneq\cdots\subsetneq F_l,
\quad
\lambda_j>0,
\quad
\sum_j\lambda_j=1.
\]
This homeomorphism is equivariant, because \(-b_F=b_{-F}\).

\begin{lemma}[Straight-line homotopy to a closure]\label{lem:closure-homotopy}
Let \(c:P(K)\to P(K)\) be order-preserving and commute with the involution, with \(F\subseteq c(F)\) for every \(F\). Then \(\bar c(x)=\sum_j\lambda_jb_{c(F_j)}\) defines an equivariant map \(|K|\to|K|\), and \(\bar c\) is equivariantly homotopic to the identity.
\end{lemma}

\begin{proof}
The faces \(c(F_0)\subseteq\cdots\subseteq c(F_l)\) form a chain, possibly with repetitions, so \(\bar c\) is the realization of a simplicial map of the barycentric subdivision. It is equivariant because \(c(-F)=-c(F)\). For a chain as above, all faces \(F_j\) and \(c(F_j)\) are contained in \(c(F_l)\), so \(x\) and \(\bar c(x)\) both lie in the closed simplex \(|c(F_l)|\) of \(K\). Hence
\[
H(x,t)=(1-t)x+t\,\bar c(x)
\]
is a point of that simplex for \(t\in[0,1]\). Regard \(|K|\) as a subset of \(\mathbb R^V\), with \(V\) the vertex set. On the closed simplex of the subdivision given by a chain, \(x\) and \(\bar c(x)\) are affine in the coefficients \(\lambda_j\), and these formulas agree on common faces. Hence \(H\) is continuous on \(|K|\times[0,1]\). The involution is affine on every simplex and carries \(|c(F_l)|\) to \(|c(-F_l)|\), so \(-H(x,t)=H(-x,t)\). Finally \(H(x,0)=x\) and \(H(x,1)=\bar c(x)\).
\end{proof}

For the sign complex, the relevant closure operator comes from the compatibility relation between signed rows and signed columns.

\begin{resultenv}{Theorem}{res:2.1}{Transpose invariance, including partial matrices}
For every partial sign matrix \(A\),
\[
S(A)\simeq_{\mathbb Z_2}S(A^t).
\]
Consequently,
\[
\operatorname{Ind}(A)=\operatorname{Ind}(A^t),
\qquad
\operatorname{coInd}(A)=\operatorname{coInd}(A^t).
\]
\end{resultenv}

\begin{proof}
If \(A\) has no entry \(\pm1\), both complexes are empty and there is nothing to prove, so we assume that \(A\) has such an entry.

\proofstep{The compatibility relation.}
The vertices of \(S(A^t)\) are the signed rows \(r^s\) of \(A\). For a nonempty face \(F\) of \(S(A)\), let \(R_A(F)\) be its compatible signed rows:
\[
R_A(F)
=
\{(r,s):sA_{ri}=b\text{ for every }i^b\in F\}.
\]
An entry \(*\) never satisfies a compatibility equation. Since \(s,b,A_{ri}\in\{\pm1\}\) in such an equation, \(sA_{ri}=b\) is equivalent to \(bA_{ri}=s\). Thus \eqref{eq:signed-row-simplex} for \(A\) and for \(A^t\) reads
\begin{equation}\label{eq:transpose-compatibility}
(r,s)\text{ compatible with }i^b
\iff i^b\in\sigma_r^s(A)
\iff r^s\in\sigma_i^b(A^t).
\end{equation}
Let \(R_{A^t}\) be the same construction for \(A^t\), sending a nonempty face \(G\) of \(S(A^t)\) to the signed columns compatible with all members of \(G\).

\proofstep{The maps of face posets.}
The set \(R_A(F)\) is nonempty because \(F\) is a face: \(F\subseteq\sigma_r^s(A)\) for some \((r,s)\), and that \((r,s)\) is compatible with all of \(F\) by \eqref{eq:transpose-compatibility}. It is a face of \(S(A^t)\): any vertex \(i^b\in F\) witnesses that all its elements lie in the corresponding signed row simplex \(\sigma_i^b(A^t)\) of \(A^t\). The map \(F\mapsto R_A(F)\) reverses inclusion, since a larger \(F\) imposes more equations. It commutes with the involutions, since replacing every \(b\) by \(-b\) in the equations replaces every solution \((r,s)\) by \((r,-s)\).

Applying the construction twice gives
\[
F\subseteq R_{A^t}(R_A(F)),
\]
because each \(i^b\in F\) is compatible with every member of \(R_A(F)\). The resulting composition \(c=R_{A^t}\circ R_A\) is the compatibility closure. It is order-preserving and commutes with the involution, and the same holds for \(c'=R_A\circ R_{A^t}\) on the faces of \(S(A^t)\), with \(G\subseteq c'(G)\).

\proofstep{The equivariant homotopy equivalence.}
An order-reversing map of face posets carries chains to chains, with the order of the members reversed. So \(R_A\) and \(R_{A^t}\) induce simplicial maps of the barycentric subdivisions, and thus maps
\[
\rho:|S(A)|\to|S(A^t)|,
\qquad
\rho\Bigl(\sum_j\lambda_jb_{F_j}\Bigr)=\sum_j\lambda_jb_{R_A(F_j)},
\]
and \(\rho':|S(A^t)|\to|S(A)|\), given by the same formula with \(R_{A^t}\). They are equivariant, because \(R_A\) and \(R_{A^t}\) commute with the involutions. Their composite \(\rho'\rho\) sends \(\sum_j\lambda_jb_{F_j}\) to \(\sum_j\lambda_jb_{c(F_j)}\), because \(\rho'\) is affine on each simplex of the subdivision and so its formula remains valid when the faces \(R_A(F_j)\) repeat. Thus \(\rho'\rho\) is the map \(\bar c\) of Lemma~\ref{lem:closure-homotopy}, and it is equivariantly homotopic to the identity. In the same way \(\rho\rho'=\bar c'\) is equivariantly homotopic to the identity. The two order-reversing maps therefore induce inverse equivariant homotopy equivalences after barycentric subdivision.

\proofstep{The index and the coindex.}
An equivariant map \(|S(A)|\to S^d\) composed with \(\rho'\) is an equivariant map \(|S(A^t)|\to S^d\), and an equivariant map \(S^d\to|S(A)|\) composed with \(\rho\) is an equivariant map \(S^d\to|S(A^t)|\). Together with the same statements for \(A^t\) in place of \(A\), and with \(|S(A)|=\Delta_{C_A^\pm}\) from Section~\ref{ssec:2.1}, this gives the equalities for the index and the coindex.

Partial matrices need no separate treatment, because entries \(*\) never enter the compatibility relation.
\end{proof}

Theorem~\ref{res:2.1} is an equivariant strengthening of the nerve relation in~\cite[Claim 24]{FHV26}.

We turn to completions, which allow lower bounds for a partial matrix to be transferred to total matrices. A completion of a partial matrix enlarges both its realizable-distribution space and its sign complex. Indeed, if \(\widetilde A\) arises from \(A\) by replacing each \(*\) by a sign, then \(\sigma_r^s(A)\subseteq\sigma_r^s(\widetilde A)\) for all \(r,s\). So \(S(A)\) is an equivariant subcomplex of \(S(\widetilde A)\). Restricting an equivariant map out of \(S(\widetilde A)\), and composing an equivariant map into \(S(A)\) with the inclusion, we get \(\operatorname{Ind}(A)\le\operatorname{Ind}(\widetilde A)\) and \(\operatorname{coInd}(A)\le\operatorname{coInd}(\widetilde A)\).

Moreover a distribution realized by a row of \(A\) is realized by the same row of \(\widetilde A\). The corresponding learning monotonicity follows by restricting a learner to the original realizable distributions: an \((\varepsilon,L)\)-list-replicable rule for \(C_{\widetilde A}\) is one for \(C_A\), with the same sample sizes and lists. We use this completion step in Corollary~\ref{res:3.2}, Theorem~\ref{res:4.1}, and Proposition~\ref{res:4.4}.

\subsection{Height parameters and chain-label conventions}\label{ssec:2.4}

Upper bounds for list replicability and for the index can be expressed through the lengths of chains in closure posets, and several such heights occur in the literature and in Section~\ref{ssec:6}. They differ in whether signs are recorded and in whether the empty intersection is counted, so we fix the notation here.

Let \(C\) be a total class with matrix \(B\) and with concept atom sets \(\sigma_c=\{i^{c(i)}:i\in[M]\}\). Let \(P_+^0(B)\) be the set of all intersections \(\bigcap_{c\in S}\sigma_c\) over nonempty subfamilies \(S\subseteq C\), including the possibly empty intersection, ordered by inclusion, and let \(P_+^*(B)=P_+^0(B)\setminus\{\varnothing\}\). The members of \(P_+^0(B)\) are the positive closures \(c_+(F)\) of Section~\ref{ssec:6.2}, and \(P_+^0(B)\) is the closure poset of Section~\ref{app:static}. We distinguish the following heights:

\begin{itemize}[leftmargin=1.6em]
\item \(h(B)\): the length (number of members) of a longest chain in the \emph{signed closure poset} of Section~\ref{ssec:6.1};
\item \(h(A)\) in \cite[Theorem 27]{FHV26}: the height appearing in
\[
\operatorname{Ind}(S(A))\le2h(A)-1;
\]
\item \(H(C)\): the \emph{junction height} of \cite[Theorem 1.3]{BHHLT26}, recalled below;
\item \(h_+^*(B)\): the length of a longest chain in \(P_+^*(B)\);
\item \(h_+^0(B)\): the length of a longest chain in \(P_+^0(B)\).
\end{itemize}

The junction height of \cite{BHHLT26} is a positive-closure height, so the bound of \cite[Theorem 1.3]{BHHLT26} can be compared directly with the bounds of Section~\ref{ssec:6}.

\begin{resultenv}{Lemma}{res:2.2}{Junction height as positive-closure height}
For every finite total class \(C\) with matrix \(B\), \(H(C)=h_+^0(B)\).
\end{resultenv}

\begin{proof}
The authors of \cite{BHHLT26} define the junction of a nonempty subfamily \(S\subseteq C\) as the partial concept \(\bigcap_{c\in S}c\) that takes the common value where all members of \(S\) agree and \(*\) elsewhere. The set \(J(C)\) of these junctions is ordered by \(g\preceq h\) if and only if \(g=g\cap h\), that is, \(h\) extends \(g\), and \(H(C)\) is the length of a longest chain in \(J(C)\).

For a total class the set of defined atoms of \(\bigcap_{c\in S}c\) is exactly \(\bigcap_{c\in S}\sigma_c\), and \(g\preceq h\) if and only if the atom set of \(g\) is contained in the atom set of \(h\). A partial concept is determined by its atom set. Hence \(g\mapsto\{\text{atoms of }g\}\) is an order isomorphism of \((J(C),\preceq)\) onto \((P_+^0(B),\subseteq)\), the nowhere-defined junction corresponding to the empty intersection, and the two longest-chain lengths coincide.
\end{proof}

The chain-label numbers come with the same two conventions: \(\chi^*\) denotes the nonempty-prefix chain-label number, whereas \(\chi^0\) includes the empty-intersection convention (both are defined in Section~\ref{ssec:6.2}). The proof of Theorem~\ref{res:6.3} gives \(\operatorname{lr}\le\chi^*\). The two numbers can differ, as Theorem~\ref{res:6.4}(h) shows on the cube:
\[
\chi^*(\{\pm1\}^k)=k,
\qquad
\chi^0(\{\pm1\}^k)=k+1.
\]

\subsection{Profiles, powers, and potential votes}\label{ssec:2.5}

The profile of a distribution on atoms is the vector of its correlations with the rows, and these correlations determine the losses of all concepts of the class. Sections~\ref{ssec:3}, \ref{ssec:7}, and~\ref{app:vote-geometry} study covers built from this vector through coordinate power maps and exponential tilts, and this subsection fixes the notation they share.

For \(B\in\{\pm1\}^{N\times M}\), let \(q_i\in\{\pm1\}^N\) be column \(i\), and put
\[
Q=\operatorname{conv}(\pm q_i),
\qquad
X_B=Q\cap\partial[-1,1]^N.
\]
For an atom \(z=(i,y)\), write
\[
a_z=yq_i.
\]
The profile map is
\[
\Phi(\nu)=\sum_z\nu_z a_z.
\]
Let \(c_r\) be the concept given by row \(r\). The entry \((a_{(i,y)})_r=yB_{ri}\) equals \(+1\) if \(c_r(i)=y\) and \(-1\) otherwise. Hence coordinate \(r\) is
\[
\Phi(\nu)_r=1-2\operatorname{loss}_\nu(c_r).
\]
The points \(a_z\) are the signed columns, so \(\Phi\) maps the distributions on atoms onto \(Q\). For realizable \(\nu\), some concept has zero loss, so
\[
\Phi(\Delta_{C_B})
=
Y
=
\{v\in Q:\max_rv_r=1\}.
\]
Indeed, if \(c_r\) realizes \(\nu\), then \(\Phi(\nu)_r=1\), while all coordinates of points of \(Q\) are at most \(1\). Conversely, if \(v\in Q\) has \(v_r=1\) and \(v=\Phi(\nu)\), then \(\operatorname{loss}_\nu(c_r)=0\), so \(\nu\) is realizable. The same argument with \(\pm c_r\) in place of \(c_r\), together with \(\partial[-1,1]^N=\{v\in[-1,1]^N:\max_r|v_r|=1\}\), shows that on the antipodally completed class, the image is \(X_B\).

The next definitions pass from a profile to a point of the unit sphere by a coordinate power map. For \(p\ge1\), define, coordinatewise on \(\mathbb R^N\),
\[
\phi_p(t)=\operatorname{sgn}(t)|t|^p,
\qquad
u_p(v)=\frac{\phi_p(v)}{\|\phi_p(v)\|_2}
\quad(v\ne0).
\]
The parameters used in this subsection and in Sections~\ref{ssec:3} and~\ref{ssec:7} are
\[
p_N=\lceil16\ln(2N)\rceil,
\qquad
s_N=2^{16}p_N^2,
\]
\[
\kappa(N,M)
=
\left\lceil
32\bigl(2s_N\ln(2M)+\ln16\bigr)
\right\rceil.
\]
The number \(p_N\) is the exponent of the power map, \(s_N\) is the number of signed columns averaged in the net of Section~\ref{ssec:3.1}, and \(\kappa(N,M)\) is the resulting bound on the target dimension.

A unit vector induces a weighted vote of the rows on the columns. For a unit vector \(w\in\mathbb R^N\), the squared soft vote is
\[
f_w(i)=\sum_r w_r|w_r|c_r(i).
\]
Since \(\sum_r|w_r|^2=1\), it takes values in \([-1,1]\).

The vote of the power image of a profile is a normalized gradient of a potential. With \(p=p_N\), set
\[
\Psi(v)=\sum_r\frac{|v_r|^{2p+1}}{2p+1},
\]
\[
A_r(v)=\operatorname{sgn}(v_r)|v_r|^{2p},
\qquad
Z(v)=\sum_r|v_r|^{2p}.
\]
For \(v\ne0\) we have \(\|\phi_p(v)\|_2^2=Z(v)\), so
\begin{equation}\label{eq:squared-power}
u_p(v)_r\,|u_p(v)_r|=\frac{A_r(v)}{Z(v)}.
\end{equation}
Since \((B^TA(v))_i=\sum_rA_r(v)c_r(i)\), the potential vote is
\[
g_\nu
=
\frac{B^TA(\Phi(\nu))}{Z(\Phi(\nu))}
=
f_{u_p(\Phi(\nu))}.
\]

The derivative of \(t\mapsto|t|^{2p+1}/(2p+1)\) is \(\operatorname{sgn}(t)|t|^{2p}\), so \(\nabla\Psi=A\).
The normalization in the potential vote is therefore the \(\ell_1\) norm of the gradient, and not its \(\ell_2\) norm:
\[
Z(v)=\|\nabla\Psi(v)\|_1.
\]
On \(X_B\), \(Z(v)\ge1\), because some coordinate of \(v\) has absolute value \(1\).

Finally, the exponential tilt \(\Theta_\lambda\) of a distribution \(\nu\) on atoms is defined by
\[
\Theta_\lambda(\nu)_z
\propto
\nu_z
\exp\!\left(
\frac{\langle\nabla\Psi(\Phi(\nu)),a_z\rangle}{\lambda}
\right),
\qquad \lambda>0.
\]
Section~\ref{ssec:3.4} proves that this is a support-preserving homeomorphism.

\part{The logarithmic-product bound}\label{part:upper}
This part contains the complete proof of Theorem~\ref{thm:main} and carries out the
construction outlined in Section~\ref{sec:construction}. Section~\ref{app:bridge}
reduces the theorem to the existence of nested acyclic sets in the faces of a cube.
Section~\ref{app:morse} constructs highly connected cones of vectors by a
critical-point estimate, Section~\ref{app:hulls} develops the tools for building sets
over the faces from these cones, and Section~\ref{app:construction} builds the required
sets and completes the proof. From the preliminaries of Section~\ref{ssec:2} the proof
uses only Lemma~\ref{res:2.0}, so a reader interested in the upper bound alone may
start at Section~\ref{app:bridge}.

\section{From cubical carriers to learning rules}\label{app:bridge}
This section reduces Theorem~\ref{thm:main} to the construction of a family of
sets in the faces of a cube; an outline of the whole argument is in
Section~\ref{sec:construction}. A finite closed cover of the
realizable-distribution space with accurate labels gives a learning rule, at the
cost of an arbitrarily small loss margin. Such a cover of multiplicity at most
$d+1$ exists as soon as a certain region $N[C]$ of the boundary of the cube maps
to the $d$-skeleton by a map preserving all faces, and this criterion is in fact
equivalent to $\operatorname{lr}(C)\le d+1$. Finally, the cubical carrier theorem
shows, by duality and obstruction theory, that the map exists whenever the faces
contain nested compact sets outside $N[C]$ whose reduced homology vanishes in low
degrees. These sets are constructed in
Sections~\ref{app:morse}--\ref{app:construction}.

\subsection{A finite cover with a positive loss margin}
We begin with the step closest to learning. Given a closed cover of $\Delta_C$
with accurate labels, a rule can output the label of the first cover member
containing the empirical distribution of its sample. Compactness provides a
radius at which every ball meets no more cover members than the multiplicity bound
of the cover, and this radius yields a list that depends only on the true distribution. The
loss margin in the following proposition is the price of this radius.
\begin{proposition}[Cover-to-rule conversion]\label{prop:cover-rule}
Let $C\subseteq\{\pm1\}^n$ be finite and nonempty. Suppose that
$R_1,\ldots,R_t$ are closed subsets covering $\Delta_C$, that their
multiplicity is at most $b$, and that each $R_j$ has a Boolean label $h_j$
whose loss is at most $\alpha$ throughout $R_j$. For every $\tau>0$ there
is a deterministic learning rule with list size at most $b$ and error
$\alpha+\tau$. Consequently
\[
 \operatorname{lr}(C,\alpha+\tau)\le \mathrm{LR}(C,\alpha),\qquad
 \operatorname{lr}(C)\le\mathrm{LR}(C),
\]
where $\mathrm{LR}(C)=\sup_{\alpha>0}\mathrm{LR}(C,\alpha)$.
\end{proposition}
\begin{proof}
\proofstep{The metric space.}
We use total variation distance on distributions on the $2n$ labeled points:
\[
 d_{\rm TV}(\mu,\nu)=\max_{E\subseteq[n]\times\{\pm1\}}|\mu(E)-\nu(E)|
 =\tfrac12\|\mu-\nu\|_1.
\]
The distributions realized by a fixed $c\in C$ form the closed face of the
simplex of all distributions spanned by the points $(j,c_j)$, $j\in[n]$.
Thus $\Delta_C$ is a finite union of closed faces and is compact. Each
$R_j$ is a closed subset of $\Delta_C$, hence compact. For
$\mu\in\Delta_C$ and $\rho>0$, write
$\overline B(\mu,\rho)=\{\nu\in\Delta_C:d_{\rm TV}(\mu,\nu)\le\rho\}$.

\proofstep{A uniform radius.}
We claim that there is $\rho>0$ such that for every $\mu\in\Delta_C$ the ball
$\overline B(\mu,\rho)$ meets at most $b$ of the sets $R_j$. Suppose
otherwise. For each integer $k\ge1$ choose a center
$\mu_k\in\Delta_C$, a set $I_k$ of $b+1$ indices, and points
$\nu_{k,j}\in R_j\cap\overline B(\mu_k,1/k)$ for $j\in I_k$. There are
finitely many index sets, so we may pass to a subsequence with $I_k=I$
constant, and then to a subsequence with $\mu_k\to\mu\in\Delta_C$, using
compactness of $\Delta_C$. For $j\in I$,
\[
 d_{\rm TV}(\nu_{k,j},\mu)\le 1/k+d_{\rm TV}(\mu_k,\mu)\to0.
\]
Since $R_j$ is closed, $\mu\in R_j$. Thus $\mu$ belongs to $b+1$ members,
a contradiction. The property persists when $\rho$ decreases, so we may
assume $\rho\le\tau$.

\proofstep{The rule.}
For a nonempty sample $S=((j_1,y_1),\ldots,(j_T,y_T))$, let
$\widehat\mu_S=T^{-1}\sum_{\ell=1}^T\delta_{(j_\ell,y_\ell)}$ be its
empirical distribution.
If $\widehat\mu_S\in\Delta_C$, output the label of the first set containing
$\widehat\mu_S$, in a fixed order of the cover; such a set exists because
the $R_j$ cover $\Delta_C$. On every other sample, including the empty
sample, output the constant $+1$ function. This defines one rule on all
finite samples. It depends on the cover, and not on $\rho$, $\delta$, the
sample size, or an unknown distribution.

\proofstep{The list.}
For a realizable distribution $\mu$, define $\mathcal L_\mu$ to be the set
of labels of cover members meeting $\overline B(\mu,\rho)$.
This list is fixed by $\mu$ and the cover; it does not depend on confidence
or the observed sample. It has at most $b$ members by the choice of $\rho$.
Let $h\in\mathcal L_\mu$, and choose $j$ with $h_j=h$ and $\nu\in R_j$ with
$d_{\rm TV}(\mu,\nu)\le\rho$. The loss of $h$ under a distribution is the
probability of the fixed event $\{(x,y):h(x)\ne y\}$, so
\[
 \operatorname{loss}_\mu(h)
 \le\operatorname{loss}_\nu(h)+d_{\rm TV}(\mu,\nu)
 \le\alpha+\rho\le\alpha+\tau.
\]

\proofstep{The output lies in the list.}
Let $\mu$ be realized by $c\in C$. With probability one every sample point
lies in the support of $\mu$, hence on the graph of $c$. Then $\widehat\mu_S$
is realized by $c$, so $\widehat\mu_S\in\Delta_C$ whenever the sample is
nonempty. If moreover $d_{\rm TV}(\widehat\mu_S,\mu)\le\rho$, then the first
cover member containing $\widehat\mu_S$ meets $\overline B(\mu,\rho)$, and the
rule outputs a member of $\mathcal L_\mu$.

\proofstep{The sample size.}
Total variation on a finite space is the maximum discrepancy over its
subsets. For a fixed event $E$, the number $\widehat\mu_S(E)$ is the average
of $|S|$ independent indicators with mean $\mu(E)$, so Hoeffding's inequality
gives $\Pr\{|\widehat\mu_S(E)-\mu(E)|>\rho\}\le2\exp(-2|S|\rho^2)$. A union
bound over the $2^{2n}$ subsets gives
\[
 \Pr\{d_{\rm TV}(\widehat\mu_S,\mu)>\rho\}
 \le 2^{2n+1}\exp(-2|S|\rho^2).
\]
Thus $|S|\ge\lceil\ln(2^{2n+1}/\delta)/(2\rho^2)\rceil$ suffices,
uniformly in $\mu$, for $0<\delta<1$; this integer is positive, so the sample
is nonempty. For such a sample size, the rule outputs a member of
$\mathcal L_\mu$ with probability at least $1-\delta$. For $\delta\ge1$ there
is nothing to prove, and we take $|S|=1$. Hence the rule has list size at most
$b$ and error $\alpha+\tau$.

\proofstep{The two inequalities.}
The faces of the rows of $C$, labeled by the rows, form a closed cover of
$\Delta_C$ with loss zero and multiplicity at most $|C|$. Thus
$\mathrm{LR}(C,\alpha)$ is a well-defined integer, attained by some cover.
Applying the first part to such a cover gives
$\operatorname{lr}(C,\alpha+\tau)\le\mathrm{LR}(C,\alpha)$. Taking
$\alpha=\tau=\varepsilon/2$ gives
$\operatorname{lr}(C,\varepsilon)\le\mathrm{LR}(C,\varepsilon/2)
\le\mathrm{LR}(C)$ for every $\varepsilon>0$. This proves the supremum claim.
\end{proof}

\subsection{A face-preserving map supplies the cover}\label{app:cube-map}
We now express the existence of accurate covers through maps between subsets of
a cube. A distribution realized by a concept $c$ determines a point of the cube
whose coordinates have the signs of $c$, and a coordinate of large mass places
this point on a facet. An accurate label has to agree with such a coordinate, and
this is the condition that a face-preserving map encodes.

Write $Q=[-1,1]^n$. A literal is a pair $v=(j_v,s_v)$ with
$j_v\in[n]$ and $s_v\in\{\pm1\}$, that is, a labeled point, or atom, in the language of
Section~\ref{ssec:2.2}. A consistent literal set uses each
coordinate at most once. For such a set put
\[
 Q_S=\{y\in Q:y_{j_v}=s_v\text{ for all }v\in S\},\qquad
 m_S=n-|S|.
\]
All interiors and boundaries of faces are relative. Let $Q^{(d)}$ be the
$d$-skeleton of the cube, and put $Q_S^{(d)}=Q_S\cap Q^{(d)}$. Define
\[
 N[C]=\bigcup_{c\in C}\{y\in\partial Q:c_jy_j\ge0\text{ for all }j\}.
\]
For a literal $v$ write $Q_v=Q_{\{v\}}$. If the coordinate of $v$ is not used
by $S$, write $S+v=S\cup\{v\}$.

We record the elementary facts about this notation. The set $Q_S$ is a cube
of dimension $m_S$ whose faces are the $Q_{S'}$ with $S'\supseteq S$
consistent. Its facets are the $Q_{S+v}$, so
$\partial Q_S=\bigcup_vQ_{S+v}$, the union over literals $v$ whose coordinate
is not used by $S$. Two literals with the same coordinate and opposite signs
have disjoint faces. Therefore, if $v\ne w$ and
$Q_{S+v}\cap Q_{S+w}\ne\varnothing$, then $v$ and $w$ use different
coordinates and $Q_{S+v}\cap Q_{S+w}=Q_{S+v+w}$.

A point $y\in Q$ lies in the interior of exactly one face, namely $Q_{S(y)}$
with $S(y)=\{(j,y_j):|y_j|=1\}$, and $y\in Q^{(d)}$ exactly when
$n-|S(y)|\le d$. If $y\in Q_S$, then $S\subseteq S(y)$. Hence $Q_S^{(d)}$ is
the $d$-skeleton of the cube $Q_S$.

For $c\in C$, a point $y\in Q$ with $c_jy_j\ge0$ for all $j$ lies in $\partial Q$
exactly when $c_ky_k=1$ for some $k$. So $N[C]$ is a finite union of closed
boxes, and it is compact and semialgebraic.
A continuous map $f:N[C]\to Q$ is called face-preserving if
$f(N[C]\cap Q_v)\subseteq Q_v$ for every literal $v$. Intersecting over
$v\in S$ gives $f(N[C]\cap Q_S)\subseteq Q_S$ for consistent nonempty $S$.

Next we place the realizable distributions in $\mathbb R^n$.
Let $\Delta^{n-1}$ be the simplex of probability vectors in $\mathbb R^n$,
and let $\odot$ denote the coordinatewise product.
The realizable-distribution space is naturally identified with
\[
 X(C)=\bigcup_{c\in C}\{p\odot c:p\in\Delta^{n-1}\}.
\]
Indeed, a realizable $\mu$ corresponds to the vector $x$ with
$x_j=\mu(j,+1)-\mu(j,-1)$, and realizability ensures that the two
labels never both have positive mass at one coordinate. Explicitly, if $c$
realizes $\mu$ and $p_j=\mu(j,+1)+\mu(j,-1)$, then $\mu(j,-c_j)=0$ and
therefore $x=p\odot c$. Conversely, $\mu(j,s)=\max\{sx_j,0\}$ recovers $\mu$
from $x$, and for $x=p\odot c$ this formula defines a distribution realized
by $c$. Both maps are continuous, so $\Delta_C$ and $X(C)$ are homeomorphic.
Under this identification, $\operatorname{loss}_\mu(h)=\sum_{j:h_jx_j<0}|x_j|$,
because $\operatorname{loss}_\mu(h)=\sum_j\mu(j,-h_j)$ and
$\mu(j,-h_j)=\max\{-h_jx_j,0\}$.

Triangulations without new vertices are used twice below: in the proof of
Lemma~\ref{lem:cube-cover}, where the vertices of the triangulation serve as
Boolean labels, and in the proof of Theorem~\ref{thm:carrier}, where a simplicial
complex is cut along a level set. Both uses rest on the classical pulling
construction. We state it for polyhedral complexes and include a proof, because both
uses need its compatibility with faces in the form stated below. A finite polyhedral
complex is a finite family $\mathcal P$ of convex polytopes in a Euclidean
space such that every nonempty face of a member is a member and any two
members are disjoint or meet in a common face.
\begin{lemma}[Pulling triangulation]\label{lem:pulling}
Let $\mathcal P$ be a finite polyhedral complex with a total order on its
vertices. For $P\in\mathcal P$ let $v_P$ be its first vertex, and define, by
induction on $\dim P$,
\begin{align*}
 \mathcal A(P)&=\bigcup\{\mathcal T(G):G\text{ a proper face of }P,\ v_P\notin G\},\\
 \mathcal T(P)&=\{\{v_P\}\}\cup\mathcal A(P)\cup
 \{\operatorname{conv}(\{v_P\}\cup\sigma):\sigma\in\mathcal A(P)\}.
\end{align*}
Then $\mathcal T=\bigcup_{P\in\mathcal P}\mathcal T(P)$ is a simplicial
complex whose vertices are the vertices of $\mathcal P$ and whose underlying
set is $\bigcup\mathcal P$. For every $P\in\mathcal P$, the family
$\mathcal T(P)$ consists of the simplices of $\mathcal T$ contained in $P$,
and its union is $P$.
\end{lemma}
\begin{proof}
We prove the following two claims by induction on $\dim P$:
\begin{enumerate}[label=(\alph*),leftmargin=2em]
\item $\mathcal T(P)$ is a simplicial complex with underlying set $P$, whose
vertices are the vertices of $P$;
\item $\mathcal T(G)=\{\sigma\in\mathcal T(P):\sigma\subseteq G\}$ for every
face $G$ of $P$.
\end{enumerate}

\proofstep{A consequence of (a) and (b).}
Suppose (a) and (b) hold for members
$P,P'$ of $\mathcal P$ and for their common face $G=P\cap P'$, and let
$\sigma\in\mathcal T(P)$, $\sigma'\in\mathcal T(P')$. If $G=P$, then
$\sigma\cap G=\sigma$; otherwise a supporting hyperplane
cutting $G$ out of $P$ shows that $\sigma\cap G$ is empty or a face of $\sigma$.
By (a) it belongs to $\mathcal T(P)$, and by (b) to $\mathcal T(G)$. The same
holds for $\sigma'\cap G$. Since $\mathcal T(G)$ is a simplicial complex,
$\sigma\cap\sigma'=(\sigma\cap G)\cap(\sigma'\cap G)$ is empty or a common
face of $\sigma$ and $\sigma'$. For disjoint $P,P'$ the two simplices are
disjoint. We call this property (c).

\proofstep{Proof of (a).}
For $\dim P=0$ both claims hold, since $\mathcal T(P)=\{P\}$. Let
$\dim P\ge1$, put $v=v_P$, and assume (a) and (b) for all proper faces of $P$.
Let $A$ be the union of the proper faces of $P$ not containing $v$. By (a) and
(c), $\mathcal A(P)$ is a simplicial complex with underlying set $A$.

Every $y\in P\setminus\{v\}$ has a unique representation
$y=(1-\lambda)v+\lambda y'$ with $0<\lambda\le1$ and $y'\in A$. For existence,
let $y'$ be the last point of $P$ on the ray from $v$ through $y$, and let $G$
be the face of $P$ with $y'$ in its relative interior. If $v\in G$, the ray
would continue beyond $y'$ inside $G$. Hence $v\notin G$ and $y'\in A$. For
uniqueness, let $y'\in G$ with $G$ a face and $v\notin G$. If $P$ contained a
point $y''$ of the ray beyond $y'$, then $y'$ would lie in the open segment
from $v$ to $y''$, and the face $G$ would contain $v$. So $y'$ is necessarily
the last point of $P$ on its ray.

If $\sigma\in\mathcal A(P)$ lies in a proper face $G$ with $v\notin G$, then
$v$ is not in the affine hull of $\sigma$, because $P$ meets the affine hull of
$G$ exactly in $G$. So $\operatorname{conv}(\{v\}\cup\sigma)$ is a simplex.
Its faces are $\{v\}$, the faces $\tau$ of $\sigma$, and the simplices
$\operatorname{conv}(\{v\}\cup\tau)$; all of them belong to $\mathcal T(P)$.
The unique representation shows, for $\sigma,\sigma'\in\mathcal A(P)$, that
\[
 \operatorname{conv}(\{v\}\cup\sigma)\cap\operatorname{conv}(\{v\}\cup\sigma')
 =\operatorname{conv}(\{v\}\cup(\sigma\cap\sigma')),\qquad
 \operatorname{conv}(\{v\}\cup\sigma)\cap\sigma'=\sigma\cap\sigma'.
\]
These are empty or common faces, with the convention
$\operatorname{conv}(\{v\}\cup\varnothing)=\{v\}$, and $v\notin A$. So
$\mathcal T(P)$ is a simplicial complex. Its union is $P$ by the representation.
Its vertices are $v$ and the vertices of $\mathcal A(P)$; every vertex of $P$ other
than $v$ is a proper face not containing $v$. This proves (a).

\proofstep{Proof of (b).}
The case $G=P$ is part of (a), so let $G$ be a proper face of $P$.
Suppose first that $v\notin G$. Then
$\mathcal T(G)\subseteq\mathcal A(P)\subseteq\mathcal T(P)$. Conversely, if
$\sigma\in\mathcal T(P)$ and $\sigma\subseteq G$, then $v\notin\sigma$, so
$\sigma\in\mathcal T(G')$ for a proper face $G'$ with $v\notin G'$. Thus
$\sigma\subseteq G\cap G'$, a common face of $G$ and $G'$. Claim (b) for $G'$
gives $\sigma\in\mathcal T(G\cap G')$, and (b) for $G$ gives
$\mathcal T(G\cap G')\subseteq\mathcal T(G)$.

Suppose now that $v\in G$. Then $v$ is the first vertex of $G$, and each proper
face of $G$ not containing $v$ is a proper face of $P$ not containing $v$.
Hence $\mathcal A(G)\subseteq\mathcal A(P)$ and
$\mathcal T(G)\subseteq\mathcal T(P)$. Conversely, let
$\sigma\in\mathcal T(P)$ with $\sigma\subseteq G$ and $\sigma\ne\{v\}$. Then
$\sigma=\tau$ or $\sigma=\operatorname{conv}(\{v\}\cup\tau)$ with
$\tau\in\mathcal A(P)$ and $\tau\subseteq G$. As in the previous paragraph,
$\tau\in\mathcal T(G\cap G')$ for a proper face $G'$ of $P$ with $v\notin G'$,
and $G\cap G'$ is a proper face of $G$ not containing $v$. So
$\tau\in\mathcal A(G)$ and $\sigma\in\mathcal T(G)$. This proves (b).

\proofstep{The whole complex.}
Property (c) shows that $\mathcal T$ is a simplicial complex, and the statements
on vertices and underlying sets follow from (a). If $\sigma\in\mathcal T(P')$
and $\sigma\subseteq P$, then $\sigma\subseteq G=P\cap P'$; claim (b) for $P'$
gives $\sigma\in\mathcal T(G)$, and (b) for $P$ gives
$\mathcal T(G)\subseteq\mathcal T(P)$.
\end{proof}

Together with Proposition~\ref{prop:cover-rule}, the next two lemmas show that a
face-preserving map $N[C]\to Q^{(d)}$ exists if and only if
$\operatorname{lr}(C)\le d+1$, so the geometric question is equivalent
to the learning question. The two lemmas are in skeleton-map form; compare the
covering-dimension formulation of~\cite[Theorem A]{BHH26b}.
\begin{lemma}[Cubical cover lemma]\label{lem:cube-cover}
If $f:N[C]\to Q^{(d)}$ is continuous and
$f(N[C]\cap Q_v)\subseteq Q_v$ for every literal $v$, then
$\mathrm{LR}(C)\le d+1$.
\end{lemma}
\begin{proof}
\proofstep{A cover of the skeleton.}
Order the vertices of $Q$ and apply Lemma~\ref{lem:pulling} to the complex of
all faces of $Q$; this is the pulling triangulation on every face.
These triangulations agree on common faces and introduce no new vertices.
The simplices contained in faces of dimension at most $d$ form a triangulation
of $Q^{(d)}$ of dimension at most $d$ whose vertices are the Boolean vectors.
For a Boolean cube vertex $h$, let $\beta_h$ be its barycentric coordinate
on the resulting triangulation of $Q^{(d)}$. The closed sets
\[
 D_h=\{y:\beta_h(y)=\max_{h'}\beta_{h'}(y)\}
\]
cover the skeleton and have multiplicity at most $d+1$, by
Lemma~\ref{res:2.0}: a maximizing coordinate is positive, and the supporting
simplex of $y$, the simplex containing $y$ in its relative interior, has at most
$d+1$ vertices.

Moreover, $y\in D_h\cap Q_{(j,s)}$ implies $h_j=s$. Indeed, $\beta_h(y)>0$, so $h$ is a
vertex of the supporting simplex of $y$. The face $Q_{S(y)}$ has dimension at most $d$
and is contained in the facet $Q_{(j,s)}$, so by Lemma~\ref{lem:pulling}, $y$ lies in a
simplex contained in the facet $Q_{(j,s)}$. The supporting simplex is a face of
that simplex. So it lies in the facet, and its vertex $h$ satisfies $h_j=s$.

\proofstep{The clamp.}
Given $\alpha>0$, choose $\zeta$ with $0<\zeta<\min\{1/n,\alpha/n\}$, and define
$\psi(x)_j=\max\{-1,\min\{1,x_j/\zeta\}\}$. Let $x=p\odot c\in X(C)$. Since
$\|x\|_1=1$, one coordinate has magnitude at least $1/n>\zeta$, and that
coordinate of $\psi(x)$ equals $\pm1$. So $\psi(x)\in\partial Q$. Also
$c_j\psi(x)_j=\min\{1,p_j/\zeta\}\ge0$ for every $j$. Hence
$\psi(x)\in N[C]$, and $\psi:X(C)\to N[C]$ is continuous.

\proofstep{Pulling back.}
The sets $R_h=(f\psi)^{-1}(D_h)$ form a finite closed cover of $X(C)$ of
multiplicity at most $d+1$. Label $R_h$ by $h$. Let $x\in R_h$ and suppose
$|x_j|\ge\zeta$. With $s=\operatorname{sgn}(x_j)$ we have $\psi(x)_j=s$, so
$\psi(x)\in N[C]\cap Q_{(j,s)}$. Face preservation gives
$f\psi(x)\in D_h\cap Q_{(j,s)}$, and the previous step gives $h_j=s$. Thus
$h$ agrees with $x$ wherever $|x_j|\ge\zeta$. Its loss is therefore at most
$\sum_{|x_j|<\zeta}|x_j|\le n\zeta<\alpha$. This proves
$\mathrm{LR}(C,\alpha)\le d+1$ for every $\alpha>0$.
\end{proof}

The converse is not needed for the upper bound of Theorem~\ref{thm:main}. It shows
that the reduction to face-preserving maps loses nothing, and it is used in the exact
criterion of Section~\ref{app:codimension}.

\begin{lemma}[Converse cube representation]\label{lem:cube-converse}
If $\operatorname{lr}(C)\le d+1$, there is a continuous
face-preserving map $N[C]\to Q^{(d)}$.
\end{lemma}
\begin{proof}
\proofstep{A closed cover of $N[C]$.}
Put $\varepsilon=1/(3n)$. By definition of the supremum and
Theorem~\ref{res:b}, $\mathrm{LR}(C,\varepsilon)\le
\operatorname{lr}(C,\varepsilon)\le d+1$. Choose a closed accurate cover of
$X(C)$ of multiplicity at most $d+1$ and error $\varepsilon<1/(2n)$, with
members $R_1,\ldots,R_t$ and labels $h_1,\ldots,h_t$.

Every $y\in N[C]$ has $\|y\|_\infty=1$, so $1\le\|y\|_1\le n$.
Radial normalization $\varrho(y)=y/\|y\|_1$ is continuous and sends $N[C]$ to
$X(C)$: if $c_jy_j\ge0$ for all $j$, then $\varrho(y)=(|y|/\|y\|_1)\odot c$, where
$|y|$ is the vector of absolute values.
Discard empty preimages and put $G_i=\varrho^{-1}(R_i)$. These closed sets cover
$N[C]$ and have multiplicity at most $d+1$.

On a cube facet $y_j=s$, the literal has mass at least $1/n$, and
every cover member meeting that facet must therefore have label $h_j=s$.
Indeed, if $y\in G_i\cap Q_{(j,s)}$ and $x=\varrho(y)$, then $sx_j\ge1/n$. If
the label of $G_i$ had $j$th coordinate $-s$, its loss at $x$ would be at least
$1/n>\varepsilon$.

\proofstep{Thickening.}
We thicken the pulled-back cover to an open cover without changing its
multiplicity or these face-separation properties.
Fix a metric on $N[C]$. If $I$ is an index set with
$\bigcap_{i\in I}G_i=\varnothing$, the continuous function
$\max_{i\in I}\operatorname{dist}(\cdot,G_i)$ is positive on the compact set
$N[C]$, so it has a positive minimum. If $(h_i)_j\ne s$, the compact sets $G_i$
and $N[C]\cap Q_{(j,s)}$ are disjoint, so the second is empty or has positive
distance from the first. There are finitely many such conditions. Choose $\varpi>0$
below all these minima and distances, and put
$U_i=\{y\in N[C]:\operatorname{dist}(y,G_i)<\varpi\}$. These open sets cover
$N[C]$. If $\bigcap_{i\in I}U_i\ne\varnothing$, then
$\bigcap_{i\in I}G_i\ne\varnothing$, so $|I|\le d+1$. If $(h_i)_j\ne s$, then
$U_i\cap Q_{(j,s)}=\varnothing$.

\proofstep{The nerve.}
Let $K$ be the nerve of $\{U_i\}$: its simplices are the sets $I$ with
$\bigcap_{i\in I}U_i\ne\varnothing$, so $\dim K\le d$. Put
$\varphi_i(y)=\max\{0,\varpi-\operatorname{dist}(y,G_i)\}$. These functions are
continuous, $\varphi_i(y)>0$ exactly on $U_i$, and $\sum_k\varphi_k>0$ because the
$G_i$ cover $N[C]$. Therefore
$\Phi(y)=\sum_i(\varphi_i(y)/\sum_k\varphi_k(y))e_i$, with $e_i$ the vertices of
$K$, is a continuous map $N[C]\to|K|$. It sends $y$ into the simplex
$I(y)=\{i:y\in U_i\}$. For $y\in N[C]\cap Q_S$, the thickening step shows that
all labels in $I(y)$ satisfy $S$.

\proofstep{A map of the nerve.}
For a simplex $\sigma$ of $K$, let $F(\sigma)$ be the smallest cube face
containing the labels of $\sigma$: it fixes exactly the coordinates on which
all these labels agree. If $\tau\subseteq\sigma$, then
$F(\tau)\subseteq F(\sigma)$, and every $k$-dimensional face of $F(\tau)$ is a
$k$-dimensional face of $F(\sigma)$. We construct $g:|K|\to Q$ with
\[
 g(\sigma)\subseteq F(\sigma)^{(\dim\sigma)}
 \qquad\text{for every simplex }\sigma,
\]
by induction on the dimension $r$ of a nerve simplex.
Map a nerve vertex to its Boolean label in $Q$. Let $r\ge1$, let $\sigma$ be an
$r$-simplex, and suppose $g$ is defined on $\partial\sigma$. Each
$k$-dimensional face $\tau$ of $\sigma$ is mapped into
$F(\tau)^{(k)}\subseteq F(\sigma)^{(k)}$. Thus the boundary is already mapped
into the $(r-1)$-skeleton of the smallest cube face containing all its labels.

We extend over the simplex into the $r$-skeleton of that cube face, as follows.
First extend by coning from the barycenter of $\sigma$ to a point of the convex
set $F(\sigma)$. If $\dim F(\sigma)\le r$, this already maps into
$F(\sigma)^{(r)}=F(\sigma)$. If $\dim F(\sigma)>r$, give $\sigma$ the cell
structure by its faces and $F(\sigma)$ its cube structure. The extension is
cellular on the subcomplex $\partial\sigma$, so the cellular approximation
theorem \cite[Theorem 4.8]{Hatcher02} provides a homotopy, stationary on
$\partial\sigma$, to a cellular map, whose image lies in $F(\sigma)^{(r)}$.

The extensions over different $r$-simplices agree on common faces, because
they are prescribed on boundaries. This produces a map of the whole nerve into
$Q^{(d)}$, since $\dim K\le d$ and $F(\sigma)^{(\dim\sigma)}\subseteq Q^{(d)}$.
It preserves every common fixed coordinate of the labels, since
$g(\sigma)\subseteq F(\sigma)$.

\proofstep{The composition.}
Put $f=g\circ\Phi:N[C]\to Q^{(d)}$. Let $y\in N[C]\cap Q_{(j,s)}$. All labels
of the simplex $I(y)$ have $j$th coordinate $s$, so
$F(I(y))\subseteq Q_{(j,s)}$, and $f(y)\in g(I(y))\subseteq F(I(y))$.
Thus $f$ is face-preserving.
\end{proof}

\subsection{Nested acyclic sets supply the map}\label{app:carrier}
It remains to produce a face-preserving map into a low skeleton. The map is built
over the faces of the cube by descending induction on the size of the literal set,
and the obstructions to extending it over a face live in the relative cohomology
of the part of the face away from a prescribed set $T_S$. Duality in the face
identifies these groups with the reduced homology of $T_S$, so it suffices to have
sets $T_S$ with vanishing reduced homology in low degrees.

All homology below is reduced singular homology with integer coefficients
unless another coefficient group is displayed. We use
$\widetilde H_{-1}(T)=0$ to mean $T\ne\varnothing$, and reduced homology
vanishes in degrees below $-1$.

Semialgebraic sets and maps are understood as in \cite[Chapter 2]{BCR98}.
The proof below and the constructions of
Sections~\ref{app:hulls} and~\ref{app:construction} use the following facts. Images and preimages
of semialgebraic sets under semialgebraic maps are semialgebraic, by the
Tarski--Seidenberg theorem \cite[Section 2.2]{BCR98}. Dimension does not
increase under semialgebraic maps, the dimension of a finite union is the
maximum of the dimensions, and the dimension of a product is the sum of the
dimensions \cite[Section 2.8]{BCR98}. Hardt's theorem \cite[Section 9.3]{BCR98}
gives, for a continuous semialgebraic map, a finite semialgebraic partition of
the target over whose members the map is semialgebraically trivial. Finally, a
compact semialgebraic set with finitely many semialgebraic subsets admits a
simultaneous triangulation \cite[Theorem 9.2.1]{BCR98}: a finite simplicial
complex and a semialgebraic homeomorphism from its underlying space onto the
set, under which each given subset is a union of images of open simplices.
A closed given subset is then the image of a subcomplex, because it contains
the image of the closure of each open simplex it contains.

The main result of this section is the following conversion. From a compatible
family of sets outside $N[C]$, which we call carriers, it produces the face-preserving
map required by Lemma~\ref{lem:cube-cover}.

\begin{theorem}[Cubical carrier theorem]\label{thm:carrier}
Let $a\ge1$ and $d=n-a-1\ge2$. Suppose that for every consistent $S$ with
$1\le |S|\le a$ there is a compact semialgebraic set
$T_S\subseteq Q_S\setminus N[C]$ such that
\[
 S\subseteq S'\Longrightarrow T_{S'}\subseteq T_S,
 \qquad
 \widetilde H_j(T_S;\mathbb Z)=0\quad(j\le a-|S|-1).
\]
Then a continuous face-preserving map $N[C]\to Q^{(d)}$ exists.
In particular, $\mathrm{LR}(C)\le n-a$.
\end{theorem}
\begin{proof}
\proofstep{Triangulation.}
Simultaneously triangulate $Q$, all its faces, $N[C]$, and the finitely
many $T_S$, using semialgebraic triangulation \cite[Theorem 9.2.1]{BCR98}.
This gives a finite simplicial complex $K_0$ and a semialgebraic homeomorphism
$|K_0|\to Q$ under which these compact sets are images of subcomplexes.
Let $K$ be the barycentric subdivision of $K_0$; its vertices are the barycenters
of the simplices of $K_0$, and its simplices are the chains of simplices of
$K_0$. Identify $Q$ with $|K|$ through the homeomorphism.
Barycentric coordinates and straight segments in simplices refer to this
structure. The homeomorphism need not be piecewise linear: the structure of $|K|$
serves to define subsets of $Q$ and homotopies, and all the statements about
them below are topological.

After barycentric subdivision the specified subcomplexes are full. Indeed, the
subdivision $L$ of a subcomplex $L_0$ of $K_0$ has as vertices the barycenters of
the simplices of $L_0$, because a barycenter lies in $|L_0|$ exactly when its
simplex belongs to $L_0$. A simplex of $K$ with all its vertices in $L$ is a
chain of simplices of $L_0$, hence a simplex of $L$.

For a full subcomplex $T$, write $\chi_T$ for the sum of barycentric
coordinates at vertices in $T$, that is, $\chi_T=\sum_{v\in T}\beta_v$, with $\beta_v$
the barycentric coordinate at the vertex $v$. This function is continuous, and it is
affine on each simplex of $K$. For $y\in Q$ let $\sigma_y$ be the supporting simplex of $y$,
the simplex containing $y$ in its interior. Then $\chi_T(y)=1$ exactly when all
vertices of $\sigma_y$ lie in $T$; by fullness this happens exactly when
$y\in T$. Also $\chi_T(y)=0$ when $\sigma_y$ has no vertex in $T$.

\proofstep{Compatible neighborhoods.}
For $1\le |S|\le a$, set
\[
 U_S=\{y\in Q_S:\chi_{T_S}(y)>1/2\},\qquad
 E_S=\{y\in Q_S:\chi_{T_S}(y)\le1/2\};
\]
for $|S|>a$ set $E_S=Q_S$. For $1\le|S|\le a$ the range of the homology
hypothesis contains $j=-1$, so $T_S\ne\varnothing$. Moreover $T_S\subseteq U_S$
and $U_S=Q_S\setminus E_S$. We need the following four properties.

\emph{(N1) $U_S$ deformation retracts onto $T_S$.}
Write $\chi=\chi_{T_S}$.
The normalized $T_S$-part of the barycentric coordinates defines a
deformation retraction $U_S\to T_S$. For $y\in U_S$, let $\varrho(y)$ be the point
with barycentric coordinate $\beta_v(y)/\chi(y)$ at each vertex $v$ of $T_S$ and
coordinate zero at all other vertices. The vertices of $\sigma_y$ in $T_S$ span
a face of $\sigma_y$. Fullness ensures that this part lies in $T_S$, so
$\varrho(y)\in T_S$. Put $\Gamma(y,\lambda)=(1-\lambda)y+\lambda\varrho(y)$, the straight
homotopy in the supporting simplex. Since $Q_S$ is a subcomplex containing $y$,
we have $\sigma_y\subseteq Q_S$. Since $\chi$ is affine on $\sigma_y$ and
$\chi(\varrho(y))=1$,
\[
 \chi(\Gamma(y,\lambda))=(1-\lambda)\chi(y)+\lambda\ge\chi(y)>1/2.
\]
So the homotopy stays inside $U_S$. The barycentric coordinates of
$\Gamma(y,\lambda)$ are continuous in $(y,\lambda)$, and hence $\Gamma$ is continuous.
Finally $\Gamma(y,0)=y$, $\Gamma(y,1)\in T_S$, and $\Gamma(y,\lambda)=y$ for $y\in T_S$,
where $\chi(y)=1$.

\emph{(N2) $N[C]\cap Q_S\subseteq E_S$.}
For $|S|>a$ we have $E_S=Q_S$. For $1\le|S|\le a$, the sets $N[C]$ and $T_S$ are
disjoint subcomplexes. If $y\in N[C]$, then $\sigma_y\subseteq N[C]$, so
$\sigma_y$ has no vertex in $T_S$, and $\chi_{T_S}(y)=0$.

\emph{(N3) If the coordinate of $v$ is not used by $S$, then
$E_S\cap Q_{S+v}\subseteq E_{S+v}$.}
For $|S+v|>a$ this holds because $E_{S+v}=Q_{S+v}$. Otherwise, nestedness gives
$T_{S+v}\subseteq T_S$, so the vertices of $T_{S+v}$ are vertices of $T_S$.
Hence $\chi_{T_{S+v}}\le\chi_{T_S}$, and $\chi_{T_S}(y)\le1/2$ implies
$\chi_{T_{S+v}}(y)\le1/2$.

\emph{(N4) For $1\le|S|\le a$, the pair $(E_S,E_S\cap\partial Q_S)$ is a finite
simplicial pair, and in particular a finite CW pair.}
Realize $K$ with its vertices as a basis of a Euclidean space, so that
$\chi=\chi_{T_S}$ is a linear functional and each simplex is a convex polytope.
Cut each simplex by the affine level $\chi=1/2$. For a simplex $\sigma$ of $K$,
the nonempty pieces $\sigma\cap\{\chi\le1/2\}$, $\sigma\cap\{\chi=1/2\}$ and
$\sigma\cap\{\chi\ge1/2\}$, on either side of the level set and on it, form a finite
polyhedral subdivision: a face of $\sigma\cap\{\chi\le1/2\}$ has the form
$\tau\cap\{\chi\le1/2\}$ or $\tau\cap\{\chi=1/2\}$ for a face $\tau$ of $\sigma$,
likewise for $\{\chi\ge1/2\}$, and a face of $\sigma\cap\{\chi=1/2\}$ has the form
$\tau\cap\{\chi=1/2\}$.
Their intersections are the corresponding pieces of common faces: for instance,
$(\sigma\cap\{\chi\le1/2\})\cap(\sigma'\cap\{\chi\ge1/2\})
=(\sigma\cap\sigma')\cap\{\chi=1/2\}$.

Now $E_S$ is the union of the pieces $\sigma\cap\{\chi\le1/2\}$ with
$\sigma\subseteq Q_S$. Also $E_S\cap\partial Q_S$ is the union of those with
$\sigma\subseteq\partial Q_S$, because a point of the subcomplex $\partial Q_S$
has its supporting simplex in $\partial Q_S$.
A pulling triangulation, Lemma~\ref{lem:pulling}, makes both $E_S$ and its
boundary intersection subcomplexes. This subdivision may depend on $S$.

\proofstep{Relative cohomology.}
Fix $S$ with $1\le|S|\le a$, and write
$\mu=m_S$, $Q'=Q_S$, $E=E_S$, $U=U_S$, and $A'=E\cap\partial Q'$.
Identify the affine hull of $Q'$ with $\mathbb R^\mu$. The sets $E$ and $A'$ are
compact and $A'\subseteq E$.
For every abelian coefficient group $H$, compact-pair
Poincar\'e--Alexander--Lefschetz duality \cite[Theorem VI.8.3]{Bredon93}, \cite[Chapter VIII, Section 7]{Dold80} gives
\[
 \check H^k(E,A';H)
 \cong H_{\mu-k}(\mathbb R^\mu\setminus A',\mathbb R^\mu\setminus E;H).
\]
This applies to the oriented manifold $\mathbb R^\mu$ and arbitrary compact
subsets, so no regularity of $E$ and $A'$ is required.

Excision \cite[Theorem 2.20]{Hatcher02} removes
$Z=\mathbb R^\mu\setminus Q'$ from this pair. Write
$X=\mathbb R^\mu\setminus A'$ and $W=\mathbb R^\mu\setminus E$. Then $Z\subseteq W$.
The closure of $Z$ in $\mathbb R^\mu$ is $Z\cup\partial Q'$, so its closure
in $X$ is $Z\cup(\partial Q'\setminus A')$. This adds only boundary points outside
$E$, since $A'=E\cap\partial Q'$. Hence the closure of $Z$ in $X$ lies in $W$, and
$W$ is open in $X$. So the excision condition holds. The resulting pair is
$(X\setminus Z,W\setminus Z)=(Q'\setminus A',U)$.

The first space is star-shaped about the center of $Q'$: for
$y\in Q'\setminus A'$, the half-open segment from the center to $y$, excluding
$y$, lies in the interior of $Q'$, while $A'\subseteq\partial Q'$. Also
$U\ne\varnothing$ by the degree $-1$ hypothesis. The exact sequence of this
pair in reduced homology therefore gives
$H_q(Q'\setminus A',U;H)\cong\widetilde H_{q-1}(U;H)$ for all $q$. Combined with
the retraction (N1) and $q=\mu-k$, this yields
\begin{equation}\label{eq:carrier-duality}
 H^k(E_S,E_S\cap\partial Q_S;H)
 \cong\widetilde H_{m_S-k-1}(T_S;H).
\end{equation}
The left side of~\eqref{eq:carrier-duality} is singular cohomology. All spaces there
are triangulable by (N4), so the \v Cech groups in the
compact-pair formulation agree with singular cohomology. Indeed, a finite
CW complex is a Euclidean neighborhood retract, and for such a space embedded in
$\mathbb R^\mu$ the direct limit of the cohomology of its neighborhoods is its
singular cohomology \cite[Appendix E, Corollaries E.6 and E.8]{Bredon93}. The
five lemma, applied to the exact sequences of the pairs and their direct limit,
gives the same statement for $(E,A')$.

Universal coefficients \cite[Theorem 3A.3]{Hatcher02} transfer the assumed
vanishing to every $H$:
\[
 \widetilde H_j(T_S;H)\cong\widetilde H_j(T_S;\mathbb Z)\otimes H
 \oplus\operatorname{Tor}(\widetilde H_{j-1}(T_S;\mathbb Z),H),
\]
and both terms vanish for $j\le a-|S|-1$. If $i\ge d$, then
\[
 m_S-i-2\le n-|S|-(n-a-1)-2=a-|S|-1,
\]
and consequently, with $k=i+1$ in~\eqref{eq:carrier-duality},
\begin{equation}\label{eq:carrier-vanishing}
 H^{i+1}(E_S,E_S\cap\partial Q_S;H)=0
 \qquad(i\ge d,\ H\text{ abelian}).
\end{equation}

\proofstep{Extension over faces.}
We construct continuous maps $f_S:E_S\to Q_S^{(d)}$, for all consistent nonempty $S$,
by descending induction on $|S|$, such that
\begin{equation}\label{eq:carrier-compatibility}
 f_S=f_{S+v}\quad\text{on }E_S\cap Q_{S+v}
\end{equation}
for every literal $v$ whose coordinate is not used by $S$. The right-hand side
is defined there by (N3).

For $|S|>a$, the face has dimension $m_S\le n-a-1=d$; so $Q_S^{(d)}=Q_S$ and we use
its identity map. Then~\eqref{eq:carrier-compatibility} holds, since $E_S=Q_S$ and
$E_{S+v}=Q_{S+v}$.

Let $1\le|S|\le a$, and
suppose the maps on all proper faces of $Q_S$ have been constructed, with
$f_{S'}=f_{S'+w}$ on $E_{S'}\cap Q_{S'+w}$ whenever $S'\supsetneq S$ and
$S'+w$ is consistent.
Their restrictions define a map $\gamma_S$ on
$E_S\cap\partial Q_S=\bigcup_v(E_S\cap Q_{S+v})$, equal to $f_{S+v}$ on
$E_S\cap Q_{S+v}$. This is well defined:
let $y\in E_S\cap Q_{S+v}\cap Q_{S+w}$ with $v\ne w$. Then $v$ and $w$ use
different coordinates and $y\in Q_{S+v+w}$. By (N3),
$y\in E_{S+v}\cap Q_{S+v+w}$ and $y\in E_{S+w}\cap Q_{S+v+w}$. So on an
intersection $Q_{S+v+w}$ both restrictions equal $f_{S+v+w}$,
by~\eqref{eq:carrier-compatibility} for $S+v$ and for $S+w$.
The pieces are finitely many closed sets, so $\gamma_S$ is continuous. Its values lie in
$Q_{S+v}^{(d)}\subseteq Q_S^{(d)}$.

We extend $\gamma_S$ by obstruction theory, and we first examine the target. Here
$m_S\ge n-a=d+1$, and the target $Y_S=Q_S^{(d)}$ is $(d-1)$-connected.
This follows by cellular approximation in the contractible cube: a null
homotopy of an $i$-sphere can be moved into its $(i+1)$-skeleton when $i<d$.
To be explicit, let $0\le i<d$ and $g:S^i\to Y_S$. Give $S^i$ a CW structure, and $D^{i+1}$ the
structure adding one cell. By \cite[Theorem 4.8]{Hatcher02}, $g$ is homotopic in
$Y_S$ to a cellular map $g'$ with image in $Q_S^{(i)}$. Coning extends $g'$ to a map
$D^{i+1}\to Q_S$ that is cellular on $S^i$, and the same theorem makes this
extension cellular, relative to $S^i$, with image in $Q_S^{(i+1)}\subseteq Y_S$.
Thus $g'$, and so $g$, is null-homotopic in $Y_S$.
The target is simply connected since $d\ge2$, and therefore it is a connected
abelian CW complex in the sense of \cite[Section 4.1]{Hatcher02}: $\pi_1$ acts
trivially on all homotopy groups.

The obstruction to extending $\gamma_S$ over the cells of dimension $i+1$ lies in
\[
 H^{i+1}(E_S,E_S\cap\partial Q_S;\pi_i(Q_S^{(d)})).
\]
We use this in the form of \cite[Corollary 4.73]{Hatcher02}: if $Y$ is a connected abelian
CW complex and $(W_0,A_0)$ is a CW pair with $H^{i+1}(W_0,A_0;\pi_iY)=0$ for all
$i$, then every map $A_0\to Y$ extends to a map $W_0\to Y$. We apply it to
$Y=Y_S$ and $(W_0,A_0)=(E_S,E_S\cap\partial Q_S)$, a CW pair by (N4).
For $i<d$ the coefficient group is zero, and for $i\ge d$
the vanishing~\eqref{eq:carrier-vanishing} applies. This gives
the extension $f_S$ of $\gamma_S$. If $E_S=\varnothing$, then $f_S$ is the empty map,
and if $E_S\ne\varnothing$ while $E_S\cap\partial Q_S=\varnothing$, a constant map
serves. By construction, \eqref{eq:carrier-compatibility} holds for $S$.

\proofstep{Gluing.}
Since $a\ge1$, every singleton $\{v\}$ is in the range $1\le|S|\le a$, and (N2)
gives $N[C]\cap Q_v\subseteq E_{\{v\}}$. Define $f$ on $N[C]\cap Q_v$ by
$f_{\{v\}}$. These pieces cover $N[C]$, because $N[C]\subseteq\partial Q$.
The maps agree on all intersections: if $y\in N[C]\cap Q_v\cap Q_w$ with $v\ne w$,
then $v,w$ use different coordinates, and
\eqref{eq:carrier-compatibility} gives
$f_{\{v\}}(y)=f_{\{v,w\}}(y)=f_{\{w\}}(y)$.
These are finitely many closed pieces, so they glue continuously. Each piece
maps to its own cube face: $f(N[C]\cap Q_v)\subseteq Q_v^{(d)}\subseteq Q_v\cap Q^{(d)}$.
Lemma~\ref{lem:cube-cover} now proves the claimed cover bound
$\mathrm{LR}(C)\le d+1=n-a$.
\end{proof}

\section{Connected regions from a critical-point estimate}\label{app:morse}
We now construct the regions in which the carriers of
Theorem~\ref{thm:carrier} will be filled. Given the row matrix $A$ of the class and
a consistent literal set $S$, we need vectors whose correlations with all rows are
small relative to their norm, and whose coordinates have the signs prescribed by $S$. The main
result, Proposition~\ref{prop:cones}, provides open cones of such vectors, nested in
$S$, whose connectivity at stage $i$ is at least $n-3-ic'-|S|$, where $c'$ is of order
$\ln(2m)\ln(2n)$. The cones arise from
sublevel sets of a smooth function on a smoothed $\ell_q$ sphere, and their
connectivity follows from a bound on the number of nonnegative Hessian directions
at high critical points.
The trace estimate behind this bound follows the smoothed $\ell_q$ geometry used in the
Morse-theoretic proof of the topological Figiel--Lindenstrauss--Milman theorem \cite{FHMNV26v1}.
The logarithmic barriers below make the construction compatible with
prescribed cube faces.

\subsection{Parameters and elementary estimates}\label{app:parameters}
We fix the parameters of the construction, dispose of the case in which the
dimension $n$ is already small, and record the numerical bounds used in the rest of the proof.
Fix $C\subseteq\{\pm1\}^n$, $|C|=m\ge1$, and let $A$ be its row matrix.
Write $G=\ln(2m)\ln(2n)$. The following parameters are used throughout
Sections~\ref{app:morse}--\ref{app:construction}:
\begin{align*}
 p&=2\lceil\ln(2m)\rceil+2,& u_1&=\frac1{2\ln(2n)},\\
 u_i&=u_1/i,&q_i&=1+u_i,\\
 \tau_i&=\tfrac12 i^{u_1\ln2},&c'&=\lceil800G\rceil,\\
 c_*&=c'+3,&a&=n-c_*,\\
 \kappa_i&=ic'+3,&r_S&=a-|S|-1,\\
 f_i(S)&=m_S-\kappa_i,&z_i(S)&=f_i(S)-1,\\
 L&=\lceil n/c'\rceil+1,&\sigma_0&=\frac{u_1\ln2}{2L^2},\\
 \sigma&=\sigma_0/8,&\delta_1&=\frac1{20n},\\
 \delta_i&=\delta_1 n^{-(1/q_i-1/q_1)}(1+\sigma)^{-2(i-1)},&e_0&=\frac1{200n},\\
 \theta_0&=\min\left\{\frac{\sigma_0}{28n},\frac1{14Ln^3}\right\},&
 \eta&=\min\{1/16,50n\theta_0\},\\
 \theta&=4\eta e_0.
\end{align*}
The construction proceeds in stages $i=1,\ldots,L$. The exponents $q_i$ decrease toward one and the levels $\tau_i$ increase with the
stage $i$. A cone at stage $i$ will consist of vectors $x$ for which $\|Ax\|_p$ is at
most about $\tau_i\|x\|_{q_i}$ and the prescribed coordinates, multiplied by their
prescribed signs, are at least about $\delta_i\|x\|_{q_i}$. The integer $c'$ is the dimension lost per stage, and $a$ is the
parameter of Theorem~\ref{thm:carrier}. The number $e_0$ is the scale of the clamp
map in Section~\ref{app:hulls}. The remaining quantities are small error terms and
integer ranges, whose roles appear where they are used. In this section,
$e$ without an index is Euler's number.
We also fix the shell $\mathcal S=\{x:1/2<\|x\|_1<2\}$, in which the vectors of the construction will lie.

The exponent $p$ is even, and $p\ge4$ since $\ln(2m)>0$. It satisfies
\[
 m^{2/p}\le e,\qquad p-1\le6.4\ln(2m),\qquad n^{u_1}<\sqrt e.
\]
For the first inequality, $p\ge2\ln(2m)+2>2\ln m$. For the second,
$p-1\le2(\ln(2m)+1)+1=2\ln(2m)+3$, and $3\le4.4\ln2\le4.4\ln(2m)$. For the
third, $u_1\ln n=\ln n/(2\ln(2n))<1/2$.

The upper bound of Theorem~\ref{thm:main} is immediate when $n$ is small compared
with $G$. Lemma~\ref{lem:cube-cover}, applied to the identity on $N[C]$, always
gives $\mathrm{LR}(C)\le n$: the inclusion of $N[C]$ in $\partial Q=Q^{(n-1)}$ is
continuous and preserves every face, so the lemma applies with $d=n-1$.
Since $a$ is an integer, $a<1$ means $n\le c_*$.
If $a<1$, this proves
$\mathrm{LR}(C)\le c_*\le805G$ for $n\ge2$. Indeed $c_*<800G+4$, and
$G\ge\ln2\ln4>0.96$ gives $4<5G$. The case $n=1$ follows from
$\mathrm{LR}(C)\le1\le805(\ln2)^2\le805G$.

Henceforth assume $a\ge1$. Then
$n>c_*>700$, $L\le n$, and, for $1\le i\le L$,
\begin{equation}\label{eq:parameter-bounds}
 \tau_i<1/\sqrt2,\quad (1+\sigma)\tau_i\le0.71,\quad
 0.58\delta_1\le\delta_i\le\delta_1,\quad \theta\le\theta_0.
\end{equation}
We verify these statements in turn. Always $G\ge(\ln2)^2>0.48$, so $c'\ge385$ and
$n>c_*\ge388$. Therefore $n\ge2$ and $G>0.96$, which gives $c'\ge769$ and
$c_*\ge772$. Since $n>c'$, we have $L\ge3$. Also $L\le n/c'+2\le n/2+n/2=n$.

The levels $\tau_i$ increase with $i$. Hence
$1/2=\tau_1\le\tau_i\le\tau_L\le\tfrac12 n^{u_1\ln2}<\tfrac12e^{(\ln2)/2}=1/\sqrt2$.
Since $n>700$, we have $u_1<1/(2\ln1400)<0.07$. Thus
$\sigma=u_1\ln2/(16L^2)<0.0031/L^2$. In particular $\sigma<0.02/L^2$, and
$\sigma<0.001$ because $L\ge2$.
Consequently $(1+\sigma)\tau_i<1.001/\sqrt2<0.71$ and
\[
 (1+\sigma)^{2L}\le\exp(2L\sigma)\le\exp(0.04/L)\le\exp(0.02)\le1.04.
\]
Since $q_i$ decreases with $i$,
\[
 0\le1/q_i-1/q_1\le1-1/q_1=\frac{u_1}{1+u_1}\le u_1.
\]
The first inequality gives $\delta_i\le\delta_1$. The last one gives
$\delta_i\ge\delta_1n^{-u_1}(1+\sigma)^{-2L}\ge\delta_1e^{-1/2}/1.04\ge0.58\delta_1$.
Finally $\theta=4\eta/(200n)=\eta/(50n)\le\theta_0$ by the definition of $\eta$.

\subsection{Smooth level sets and walls}
This subsection introduces the manifolds and the functions to which Morse theory
will be applied. The energy $F$ below is the $p$-th power of $\|Ay\|_p$, considered on
a smoothed $\ell_q$ sphere $\Sigma$, and logarithmic barriers keep the prescribed
coordinates, multiplied by their prescribed signs, above the level $\delta$. The smoothing makes $\Sigma$ a smooth
hypersurface and preserves the strict convexity of the $\ell_q$ ball. This convexity
supplies the negative Hessian directions in Lemma~\ref{lem:morse-count}.

Fix a stage $i$ and abbreviate $q=q_i$, $u=u_i$, and $\delta=\delta_i$. Since
$0<u\le u_1<0.07$, we have $1<q<2$.
Choose $s>0$ so small that $ns\le\min\{\sigma/8,1/100\}$; this choice does not
depend on the stage. Set
\[
 r_j(y)=(y_j^2+s^2)^{1/2},\quad
 \Psi(y)=\sum_j r_j(y)^q,\quad
 \Sigma=\{\Psi=1\},\quad B=\{\Psi\le1\},\quad
 F(y)=\sum_{c\in C}\langle c,y\rangle^p.
\]
Thus $F(y)^{1/p}=\|Ay\|_p$, because $p$ is even, and
$\langle y,\nabla F(y)\rangle=pF(y)$, by Euler's identity for the homogeneous
polynomial $F$ of degree $p$.
For $r=(t^2+s^2)^{1/2}$, we have $\frac{d}{dt}(t^2+s^2)^{q/2}=qtr^{q-2}$ and
\begin{align*}
 \frac{d^2}{dt^2}(t^2+s^2)^{q/2}
 &=qr^{q-2}+q(q-2)t^2r^{q-4}
 =qr^{q-4}\bigl((q-1)t^2+s^2\bigr)\\
 &=q(q-1)r^{q-2}+q(2-q)s^2r^{q-4}
 \ge q(q-1)r^{q-2}>0.
\end{align*}
Since $r_j\ge s>0$, the function $\Psi$ is smooth on $\mathbb R^n$. Its Hessian is
diagonal with these second derivatives as entries, so at every point
\[
 \nabla^2\Psi\succeq q(q-1)D_r,
 \qquad D_r=\operatorname{diag}(r_j^{q-2}),
\]
and $\Psi$ is strictly convex.

Consequently $B$ is a symmetric strictly convex body with the origin
in its interior, and $\Sigma$ is smooth. Indeed, $B$ is a sublevel set of a
continuous strictly convex function, hence closed and strictly convex. Also
$\Psi(0)=ns^q\le ns<1$, so the origin is interior. Since $\|y\|_q^q\le\Psi(y)$, the set $B$
is bounded, and $\Psi(-y)=\Psi(y)$. For $y\ne0$ the function $t\mapsto\Psi(ty)$ on
$[0,\infty)$ is strictly convex with zero derivative at $t=0$, so it is
strictly increasing. Thus every ray from the origin meets $\Sigma$ exactly once,
and $\Sigma=\partial B$. Finally $\partial_j\Psi(y)=qy_jr_j^{q-2}$, so $\nabla\Psi$
vanishes at the origin and nowhere else, and $\Sigma$ is a smooth compact
hypersurface.

Let $y\in\Sigma$, so that $\sum_jr_j^q=1$. Then
\[
 r_j\le1,\quad \|y\|_q\le1,\quad\|y\|_1<\sqrt e.
\]
The first bound holds since each term is at most the sum. The second follows from
$|y_j|\le r_j$. For the third, H\"older gives
$\|y\|_1\le n^{1-1/q}\|y\|_q$, and $1-1/q=u/(1+u)<u_1$ gives
$n^{1-1/q}\le n^{u_1}<\sqrt e$.

Let $\nu$ be the gauge of $B$; it is a norm, since $B$ is a symmetric convex body
with the origin in its interior. For $x\ne0$, the point $y=x/\nu(x)$ lies in
$\Sigma$. Minkowski's inequality gives
\begin{equation}\label{eq:gauge}
 (1-n^{1/q}s)\nu(x)\le\|x\|_q\le\nu(x),
 \qquad n^{1/q}s\le\sigma/8.
\end{equation}
Indeed, $\|x\|_q=\nu(x)\|y\|_q$, and the upper bound is $\|y\|_q\le1$. For the lower
bound, $r_j(y)\le|y_j|+s$ gives $1=(\sum_jr_j^q)^{1/q}\le\|y\|_q+n^{1/q}s$. The last
inequality in the display follows from $n^{1/q}s\le ns$.

H\"older's inequality with exponents $p/(p-1)$ and $p$ also gives
\begin{equation}\label{eq:gradient}
 |\partial_j F(y)|\le p\sum_{c\in C}|\langle c,y\rangle|^{p-1}
 \le p m^{1/p}F(y)^{1-1/p}
 \le p\sqrt e\,F(y)^{1-1/p}.
\end{equation}
On $\Sigma$, moreover, $|\langle c,y\rangle|\le\|y\|_1<\sqrt e$, so
$|\partial_jF|$ is at most $pm e^{(p-1)/2}$.
Put $E_0=pm e^{(p-1)/2}+2$ and $B_0=(2n+1)E_0$.

We now impose the coordinate constraints. For a consistent nonempty literal set $S$, define
\[
 w_v(y)=s_vy_{j_v}-\delta,\qquad
 \mathcal X_S=\{y\in\Sigma:w_v(y)>0\ (v\in S)\}.
\]
We refer to the inequalities $w_v>0$ as walls.
The set $\mathcal X_S$ is an open subset of $\Sigma$, hence a smooth manifold of dimension $n-1$.

Each $\mathcal X_S$ is nonempty and contractible. To see this, define $y^0$ by
$y^0_{j_v}=s_v\delta$ on the prescribed coordinates and zero elsewhere.
Each $r_j(y^0)$ is at most $\delta+s<1$, so $r_j(y^0)^q\le\delta+s$. Then
$\Psi(y^0)\le n(\delta+s)\le n\delta_1+ns\le0.06<1$, and $y^0$ is an interior point
of $B$.
Radial projection $y\mapsto(y-y^0)/|y-y^0|$ about $y^0$ therefore maps
$\Sigma$ homeomorphically to the unit sphere: it is a continuous bijection between
compact Hausdorff spaces, because each ray from an interior point of a convex
body meets the boundary exactly once. Since
$s_v(y-y^0)_{j_v}=w_v(y)$, it maps $\mathcal X_S$ onto
$\{\omega:s_v\omega_{j_v}>0\ (v\in S)\}$. This is the sphere section
of an open convex cone contained in a hemisphere, hence contractible. Explicitly,
let $\omega_0$ be the normalization of the vector with coordinates $s_v$ at the
$j_v$ and zero elsewhere. For $\omega$ in this set, $(1-\lambda)\omega+\lambda\omega_0$
has positive $s_v$-signed coordinates and is nonzero, and normalizing it gives
a contraction onto $\omega_0$.

For $0<b<1$ and $\ell\in\mathbb R^n$, consider the smooth function
\[
 F_S(y)=F(y)+\langle\ell,y\rangle
       +b\sum_{v\in S}\ln(1/w_v(y))\qquad(y\in \mathcal X_S).
\]
We have $0<w_v<1$, because $w_v(y)\le|y_{j_v}|-\delta<r_{j_v}(y)\le1$, so every barrier
term is nonnegative. Moreover $F\ge0$ and
$|\langle\ell,y\rangle|\le\|\ell\|_\infty\|y\|_1\le\sqrt e\|\ell\|_\infty$.
Thus $F_S\ge-\sqrt e\|\ell\|_\infty$.
The function is proper and bounded below. To see this, write
$M^t=\{y\in \mathcal X_S:F_S(y)\le t\}$. If $y\in M^t$, then for each $v\in S$
\[
 b\ln(1/w_v(y))\le F_S(y)-F(y)-\langle\ell,y\rangle
 \le t+\sqrt e\|\ell\|_\infty.
\]
So a sublevel of height $t$ is
contained in $\{y\in\Sigma:w_v(y)\ge\exp(-(t+\sqrt e\|\ell\|_\infty)/b)\ (v\in S)\}$,
a compact subset of $\mathcal X_S$, and it is closed there. Therefore each $M^t$ is compact.
On $\mathcal X_{S'}\subseteq \mathcal X_S$ with $S\subseteq S'$ we have
$F_S\le F_{S'}$, since the additional barrier terms are nonnegative.

The next lemma is the main estimate of this section. At a critical point of $F_S$
where the energy $F$ is not small, the Lagrange multiplier of the constraint is
positive, so the curvature of $\Sigma$ contributes a large negative term to the
Hessian. A trace bound then shows that this term dominates in all but few
directions.

\begin{lemma}[Critical-point estimate]\label{lem:morse-count}
Assume $\|\ell\|_\infty\le1$ and $b/\delta\le1$.
At a critical point $z$ of $F_S$, write, with the Lagrange multiplier $\lambda$ of the constraint $\Psi=1$,
\begin{equation}\label{eq:critical}
 \partial_j F(z)+\ell_j-\sum_{v:j_v=j}\mu_vs_v
 =\lambda q z_jr_j^{q-2},\qquad \mu_v=b/w_v(z)>0.
\end{equation}
Then $\mu_v\le B_0$ and
$F_S(z)\le F(z)+\langle\ell,z\rangle+nb\ln(B_0/b)$.
If, in addition,
\[
 F(z)\ge0.49^p,\qquad
 \|\ell\|_\infty+b/\delta\le0.08p\,0.49^p,
\]
then $\lambda q\ge pF(z)/2>0$. Under these additional hypotheses, the Hessian of $F_S$ on $\mathcal X_S$ at $z$
has at most $400\ln(2m)/u+|S|$ nonnegative eigenvalues.
\end{lemma}
\begin{proof}
\proofstep{The critical equation.}
Let $\widehat F_S$ be the function given by the formula for $F_S$ on the open set
$\{y\in\mathbb R^n:w_v(y)>0\ (v\in S)\}$. Since $\nabla\Psi(z)\ne0$, the tangent
space of $\Sigma$ at $z$ is $T=\nabla\Psi(z)^\perp$. Thus $z$ is critical exactly
when $\nabla\widehat F_S(z)=\lambda\nabla\Psi(z)$ for a real number $\lambda$,
which is then unique. We have $\partial_j\Psi(z)=qz_jr_j^{q-2}$ with $r_j=r_j(z)$,
and the derivative of $b\ln(1/w_v)$ in the coordinate $j_v$ is
$-bs_v/w_v=-\mu_vs_v$. This is~\eqref{eq:critical}. Since $S$ is consistent, its
sum has at most one term.

\proofstep{The multiplier bound.}
Let $J_1$ be the prescribed coordinates satisfying $s_vz_{j_v}<2\delta$.
For $j=j_v\in J_1$ we have $0<s_vz_j<2\delta$, since $w_v(z)>0$. So
$r_j\le|z_j|+s<2\delta+s\le1$, and $r_j^q\le r_j\le2\delta+s$.
Their total smoothed mass is
$\sum_{j\in J_1}r_j^q\le n(2\delta+s)\le0.11$, because
$2n\delta\le2n\delta_1=0.1$ by~\eqref{eq:parameter-bounds} and $ns\le0.01$.
For a prescribed coordinate outside $J_1$, $w_v(z)\ge\delta$ and so
$\mu_v\le b/\delta\le1$.
Since $\sum_jr_j^q=1$, the coordinates outside $J_1$ carry smoothed mass at least
$0.89$, and some $j_*\notin J_1$ satisfies $r_{j_*}^q\ge0.89/n$. From $r_{j_*}\le1$
and $q>1$ we obtain $r_{j_*}\ge r_{j_*}^q\ge0.89/n$, and hence
$|z_{j_*}|\ge r_{j_*}-s\ge0.88/n$.

Take absolute values in~\eqref{eq:critical} at $j_*$. The left side is at most
$pme^{(p-1)/2}+\|\ell\|_\infty+1\le E_0$, by the bound on $\partial_jF$ over
$\Sigma$ and the bound $\mu_v\le1$ outside $J_1$. On the right,
$r_{j_*}^{q-2}\ge1$. So $|\lambda|q\cdot0.88/n\le E_0$, which gives
$|\lambda|q\le2nE_0$. Since $|z_j|r_j^{q-2}\le r_j^{q-1}\le1$,
the equation at any prescribed coordinate now yields
\[
 \mu_v\le pme^{(p-1)/2}+\|\ell\|_\infty+|\lambda|q
 \le E_0+2nE_0=B_0.
\]
Therefore $w_v(z)=b/\mu_v\ge b/B_0$ and $0<\ln(1/w_v(z))\le\ln(B_0/b)$. A consistent
set has $|S|\le n$, so the barrier sum at $z$ is at most $nb\ln(B_0/b)$.
This proves the first assertion.

\proofstep{The sign of the multiplier.}
Assume now the additional hypotheses, and write $F=F(z)$.
Multiply~\eqref{eq:critical} by $z_j$ and sum outside $J_1$:
\[
 \sum_{j\notin J_1}z_j\partial_jF(z)
 +\sum_{j\notin J_1}z_j\ell_j
 -\sum_{v:\,j_v\notin J_1}\mu_vs_vz_{j_v}
 =\lambda q\sum_{j\notin J_1}z_j^2r_j^{q-2}.
\]
Euler's identity gives $\sum_jz_j\partial_jF(z)=pF$. On $J_1$ we use $|z_j|<2\delta$,
\eqref{eq:gradient}, $2n\delta\le0.1$, and $F^{-1/p}\le1/0.49$. These show
\[
 \sum_{j\notin J_1}z_j\partial_jF(z)
 \ge pF-2n\delta\,p\sqrt e F^{1-1/p}
 \ge(1-0.1\sqrt e/0.49)pF\ge0.663pF.
\]
The linear and barrier contributions have absolute value at most
$\sqrt e(\|\ell\|_\infty+b/\delta)\le0.08\sqrt e\,p\,0.49^p\le0.133pF$, using
$\|z\|_1<\sqrt e$, the bound $\mu_v\le b/\delta$ outside $J_1$, and
$F\ge0.49^p$.
Thus $\lambda q\sum_{j\notin J_1}z_j^2r_j^{q-2}\ge0.53pF>0$.
The sum is nonnegative and at most $\sum_jr_j^q=1$, because $z_j^2\le r_j^2$. This
proves $\lambda>0$ and $\lambda q\ge0.53pF\ge pF/2$.

\proofstep{The Hessian.}
At the critical point $z$, the Hessian of $F_S$ is the quadratic form
$\xi\mapsto(F_S\circ\gamma)''(0)$ on $T$, where $\gamma$ is any smooth curve in $\mathcal X_S$
with $\gamma(0)=z$ and $\gamma'(0)=\xi$.
Differentiating $F_S\circ\gamma$ and $\Psi\circ\gamma\equiv1$ twice at zero gives
\begin{align*}
 (F_S\circ\gamma)''(0)&=\xi^{\mathsf T}\nabla^2\widehat F_S(z)\xi
 +\langle\nabla\widehat F_S(z),\gamma''(0)\rangle,\\
 0&=\xi^{\mathsf T}\nabla^2\Psi(z)\xi+\langle\nabla\Psi(z),\gamma''(0)\rangle.
\end{align*}
Since $\nabla\widehat F_S(z)=\lambda\nabla\Psi(z)$, subtracting $\lambda$ times the
second line from the first gives
$(F_S\circ\gamma)''(0)=\xi^{\mathsf T}(\nabla^2\widehat F_S(z)-\lambda\nabla^2\Psi(z))\xi$.
Thus on the tangent space $T=\nabla\Psi(z)^\perp$ the Hessian is the restriction of
\[
 \nabla^2F+b\sum_{v\in S}w_v^{-2}e_{j_v}e_{j_v}^{\mathsf T}
 -\lambda\nabla^2\Psi,
\]
evaluated at $z$. The linear term has zero Hessian, and the second derivative of
$-b\ln w_v$ in the coordinate $j_v$ is $bs_v^2/w_v^2=b/w_v^2$.

\proofstep{Reduction to one matrix.}
Let $V\subseteq T$ be the span of the eigenvectors of this form with nonnegative
eigenvalues. The form is nonnegative on $V$, and $\dim V$ is the number to be
bounded. Intersect $V$ with $\{\xi:\xi_{j_v}=0\ (v\in S)\}$, losing at most
$|S|$ dimensions, and call the result $V_0$. The barrier Hessian vanishes on $V_0$.
Since $p$ is even, $W=\operatorname{diag}(\langle c,z\rangle^{p-2})$ is
positive semidefinite and $\nabla^2F(z)=p(p-1)A^{\mathsf T}WA$. Since $\lambda>0$,
$\nabla^2\Psi\succeq q(q-1)D_r$, and $\lambda q\ge pF(z)/2$, we have
\[
 \lambda\nabla^2\Psi(z)\succeq\lambda q(q-1)D_r\succeq\tfrac12pF(z)(q-1)D_r.
\]
Hence, for $\xi\in V_0$,
\[
 0\le\xi^{\mathsf T}(\nabla^2F(z)-\lambda\nabla^2\Psi(z))\xi
 \le p\,\xi^{\mathsf T}M\xi,\qquad
 M=(p-1)A^{\mathsf T}WA-(q-1)F(z)D_r/2.
\]
So $M$ is nonnegative on $V_0$. Then $V_0$ meets the span of the eigenvectors of $M$
with negative eigenvalues only at zero. A subspace on which the Hessian is
nonnegative therefore has, after discarding at most $|S|$ dimensions, dimension
bounded by the number of nonnegative eigenvalues of $M$.

\proofstep{The trace estimate.}
Since $r_j\ge s>0$, we have $D_r\succ0$. Put $\varkappa=(q-1)F(z)/2>0$ and
$P=(p-1)D_r^{-1/2}A^{\mathsf T}WAD_r^{-1/2}\succeq0$. Congruence by $D_r^{-1/2}$ turns $M$ into $P-\varkappa I$. By
Sylvester's law of inertia, the number of nonnegative eigenvalues of $M$ equals
the number of eigenvalues of $P$ that are at least $\varkappa$. As $P$ is positive
semidefinite, this number is at most $\operatorname{tr}P/\varkappa$. Since $c_j^2=1$,
\[
 \operatorname{tr}P
 =(p-1)\sum_jr_j^{2-q}\sum_cc_j^2\langle c,z\rangle^{p-2}
 =(p-1)\Bigl(\sum_jr_j^{2-q}\Bigr)\Bigl(\sum_c\langle c,z\rangle^{p-2}\Bigr).
\]
Therefore the number of nonnegative eigenvalues of $M$ is at most
\[
 \frac{2(p-1)(\sum_jr_j^{2-q})(\sum_c|\langle c,z\rangle|^{p-2})}
      {(q-1)F(z)}
 \le\frac{2(p-1)n^{2-2/q}m^{2/p}F(z)^{-2/p}}u.
\]
This uses $q-1=u$. It also uses $\sum_jr_j^{2-q}\le n^{2-2/q}$, by H\"older with exponents $q/(2-q)$
and $q/(2q-2)$ and by $\sum_jr_j^q=1$. Finally
$\sum_c|\langle c,z\rangle|^{p-2}\le m^{2/p}F(z)^{1-2/p}$ by H\"older with exponents
$p/(p-2)$ and $p/2$, where $p\ge4$.

\proofstep{The constants.}
Using $n^{2-2/q}=(n^{1-1/q})^2<e$, $m^{2/p}\le e$, $F(z)^{-2/p}\le0.49^{-2}$,
and $p-1\le6.4\ln(2m)$, the last display is at most
\[
 \frac{2\cdot6.4\cdot e^2}{0.49^2}\cdot\frac{\ln(2m)}u\le\frac{394\ln(2m)}u
 \le\frac{400\ln(2m)}u.
\]
Adding the $|S|$ discarded dimensions gives the stated bound.
\end{proof}

\subsection{The nested cones}
We can now state and prove the main result of this section. By
Lemma~\ref{lem:morse-count}, every critical point of $F_S$ above a suitable level has large
index. Passing such a critical point therefore does not change the low-dimensional homotopy
type of the sublevel sets, and the sublevel sets inherit high connectivity from
the contractible manifold $\mathcal X_S$.

For an integer $k\ge0$, a space $Y$ is $k$-connected if it is nonempty and, for
$0\le j\le k$, every map $S^j\to Y$ extends to a map $D^{j+1}\to Y$; equivalently,
$Y$ is path connected and $\pi_j(Y,y)=0$ for $1\le j\le k$ and all $y\in Y$.
\begin{proposition}[Nested cones]\label{prop:cones}
For every stage $1\le i\le L$, there are open cones
$\mathcal C_{i,S}\subseteq\mathbb R^n\setminus\{0\}$, for all
consistent nonempty $S$, with the following properties:
\begin{enumerate}[label=(\roman*),leftmargin=2em]
\item $S\subseteq S'\Longrightarrow\mathcal C_{i,S'}\subseteq\mathcal C_{i,S}$.
\item $\mathcal C_{i,S}$ is $(n-3-ic'-|S|)$-connected.
\item If $x\in\mathcal C_{i,S}$, then
$\|Ax\|_p<(1+\sigma)\tau_i\|x\|_{q_i}$ and
$s_vx_{j_v}>\delta_i\|x\|_{q_i}$ for $v\in S$.
\item If $x\ne0$, $\|Ax\|_p\le(1-\sigma)\tau_i\|x\|_{q_i}$,
and $s_vx_{j_v}\ge(1+\sigma)\delta_i\|x\|_{q_i}$ for $v\in S$,
then $x\in\mathcal C_{i,S}$.
\end{enumerate}
Connectivity $-1$ means nonemptiness; a bound below $-1$ imposes no condition.
\end{proposition}
\begin{proof}
\proofstep{Levels and the barrier coefficient.}
Fix a stage $i$; the objects of the preceding subsection are taken at this stage,
and $\nu_i$ denotes the gauge of $B$ at stage $i$.
Put $a_i=((1-\sigma)\tau_i)^p$, $b_i=((1-\sigma/2)\tau_i)^p$,
$b_i'=((1-\sigma/4)\tau_i)^p$, and
\[
 \xi_i=\min\{b_i-a_i,\tau_i^p-b_i',0.495^p-0.49^p\}>0.
\]
Positivity holds because $0<1-\sigma<1-\sigma/2<1-\sigma/4<1$. Moreover
$\tau_i\ge1/2$ and $\sigma<0.001$ give $(1-\sigma/2)\tau_i\ge0.4997$, so
$b_i\ge0.495^p$.
Choose $b>0$ so small that
\[
 b/\delta_i\le\min\{1,0.04p\,0.49^p\},\quad
 nb\ln(B_0/b)\le\xi_i/4,\quad
 nb\ln(2/(\delta_i\sigma))\le\xi_i/4.
\]
This is possible since $t\ln(B_0/t)\to0$ as $t\to0^+$, and it gives
$b\le\delta_i<1$.

\proofstep{A common Morse perturbation.}
Choose a generic $\ell$ with
$\|\ell\|_\infty\le\min\{1,0.04p\,0.49^p\}$ and
$\sqrt e\|\ell\|_\infty\le\xi_i/4$.
Generic linear perturbations make $F_S$ Morse on the submanifold $\mathcal X_S$,
and one choice works for the finitely many $S$.
Indeed, by a standard theorem on Morse functions
\cite[Chapter 1, Section 7, p.~43]{GP74}, for a smooth function
$f_0$ on a submanifold of $\mathbb R^n$, compact or not, the function
$f_0+\langle\ell,\cdot\rangle$ is Morse for almost every $\ell\in\mathbb R^n$. We
use it with $f_0=F+b\sum_{v\in S}\ln(1/w_v)$ on $\mathcal X_S$. A finite union of null sets
is null, and the box above has positive measure.
The hypotheses of Lemma~\ref{lem:morse-count} on $\ell$ and $b$ now hold, including
$\|\ell\|_\infty+b/\delta_i\le0.08p\,0.49^p$.

Lemma~\ref{lem:morse-count} gives $w_v(z)=b/\mu_v\ge b/B_0$ at every critical point
$z$ of $F_S$, and this puts all critical points in the compact subset
$\{y\in\Sigma:w_v(y)\ge b/B_0\ (v\in S)\}$ of $\mathcal X_S$. The critical set is closed
in $\mathcal X_S$, hence compact.
Nondegeneracy makes the critical points isolated \cite[Corollary 2.3]{Milnor63},
so there are finitely many.
Choose a common regular value $t_i\in(b_i,b_i')$ of all the functions $F_S$ and set
\[
 \mathcal A_{i,S}=\{y\in \mathcal X_S:F_S(y)<t_i\},\qquad
 \mathcal C_{i,S}=\{x\ne0:x/\nu_i(x)\in\mathcal A_{i,S}\}.
\]
The set $\mathcal A_{i,S}$ is open in $\Sigma$, the map $x\mapsto x/\nu_i(x)$ is
continuous on $\mathbb R^n\setminus\{0\}$, and it is invariant under positive
scaling. So $\mathcal C_{i,S}$ is an open cone.

\proofstep{Nestedness.}
Let $S\subseteq S'$ and $x\in\mathcal C_{i,S'}$, and put $y=x/\nu_i(x)$. Then
$y\in \mathcal X_{S'}\subseteq \mathcal X_S$, and $F_S(y)\le F_{S'}(y)<t_i$, with the same $b$, $\ell$
and $t_i$ for $S$ and $S'$. This proves (i).

\proofstep{The index at high critical points.}
Fix $S$, and let $z$ be a critical point of $F_S$ with $F_S(z)\ge t_i$. The first
assertion of Lemma~\ref{lem:morse-count} and the choices of $b$ and $\ell$ give
\[
 F(z)\ge F_S(z)-\sqrt e\|\ell\|_\infty-nb\ln(B_0/b)\ge t_i-\xi_i/2>b_i-\xi_i/2
 \ge0.49^p.
\]
The last step uses $b_i\ge0.495^p$ and $\xi_i\le0.495^p-0.49^p$. The Hessian
bound of Lemma~\ref{lem:morse-count} therefore applies. Since
$400\ln(2m)/u_i=400\ln(2m)\cdot2i\ln(2n)=800iG\le ic'$, the Hessian at $z$ has at
most $ic'+|S|$ nonnegative eigenvalues.
Since $\mathcal X_S$ has dimension $n-1$, the Morse index of $z$ is at least
\[
 d_{i,S}=n-1-ic'-|S|.
\]
The connectivity bound in (ii) is $n-3-ic'-|S|=d_{i,S}-2$.
If $d_{i,S}\le0$, this bound is below $-1$, and there is nothing to show.

\proofstep{Nonemptiness.}
Let $d_{i,S}\ge1$. Since $\mathcal X_S\ne\varnothing$ and all $M^t$ are compact, $F_S$
attains its minimum at a point of $\mathcal X_S$. This is a critical point of index zero. If
its value were at least $t_i$, the previous step would give index at least
$d_{i,S}\ge1$. So its value is below $t_i$, and $\mathcal A_{i,S}$ is nonempty. For
$d_{i,S}=1$ this is all that (ii) requires.

\proofstep{Morse theory on the open manifold.}
Assume $d_{i,S}\ge2$. The manifold $\mathcal X_S$ is not compact, but the slabs $F_S^{-1}[t,t']$ are closed subsets of $M^{t'}$,
hence compact, and this is what
the results used here require. We use the following
statements: (M1) and (M2) from \cite[Section 3]{Milnor63}, and (M3) from \cite{Hatcher02}.
\begin{enumerate}[label=(M\arabic*),leftmargin=3em]
\item \cite[Theorem 3.1]{Milnor63}. If $F_S^{-1}[t,t']$ contains no critical point,
then there is a deformation $\varrho_\lambda:M^{t'}\to M^{t'}$, $0\le\lambda\le1$, with
$\varrho_0$ the identity, $\varrho_1(M^{t'})=M^t$, and $\varrho_\lambda=\mathrm{id}$ on $M^t$. On
the slab it has the form $\varrho_\lambda(y)=\Theta_{\lambda(t-F_S(y))}(y)$, where $\Theta$ is
a flow along which $F_S$ grows at unit speed within the slab. Hence
\begin{equation}\label{eq:milnor-track}
 F_S(\varrho_\lambda(y))=(1-\lambda)F_S(y)+\lambda t\le F_S(y)
 \qquad(t\le F_S(y)\le t').
\end{equation}
\item \cite[proof of Theorem 3.2, Assertions 3 and 4, and Remark 3.3]{Milnor63}. Let
$c$ be a critical value, with critical points of indices
$\iota_1,\ldots,\iota_\gamma$ at level $c$. For all sufficiently small $\varepsilon>0$,
the set $M^{c+\varepsilon}$ contains pairwise disjoint closed cells
$\mathsf D_1,\ldots,\mathsf D_\gamma$ with $\dim\mathsf D_l=\iota_l$ and
$\mathsf D_l\cap M^{c-\varepsilon}=\partial\mathsf D_l$, and it deformation retracts
onto $M^{c-\varepsilon}\cup\mathsf D_1\cup\cdots\cup\mathsf D_\gamma$. This deformation
fixes $M^{c-\varepsilon}$ pointwise.
\item A consequence of \cite[Lemma 4.10]{Hatcher02}. Let $Z$ be obtained from a subspace $Z_0$ by
attaching a cell of dimension $\iota>k$, and let $\varphi:D^k\to Z$ satisfy
$\varphi(S^{k-1})\subseteq Z_0$. Then $\varphi$ is homotopic rel $S^{k-1}$ to a map into
$Z_0$. Indeed, identify $D^k$ with the cube $[0,1]^k$. The lemma gives a homotopy rel
$\varphi^{-1}(Z_0)$ to a map $\varphi_1$, a polyhedron $K_0\subseteq D^k$ on which $\varphi_1$
is piecewise linear into the open cell, and a nonempty open subset $U_0$ of the open
cell with $\varphi_1^{-1}(U_0)\subseteq K_0$. Since $k<\iota$, the set $\varphi_1(K_0)$
does not contain $U_0$. So $\varphi_1$ misses a point of the open cell, and the radial
deformation of the punctured cell pushes $\varphi_1$ into $Z_0$, fixing $Z_0$. The
argument uses no CW structure on $Z_0$.
\end{enumerate}

\proofstep{Compression.}
Write $\mathcal A=\mathcal A_{i,S}$. We claim: for $1\le k\le d_{i,S}-1$, every map
$\varphi:D^k\to \mathcal X_S$ with $\varphi(S^{k-1})\subseteq\mathcal A$ is homotopic rel $S^{k-1}$ to
a map into $\mathcal A$.

Let $c_1<\cdots<c_N$ be the critical values of $F_S$ above $t_i$; possibly $N=0$.
Fix levels $T_0=t_i<c_1<T_1<c_2<\cdots<c_N<T_N$. Since $t_i$ is a regular value, every
critical point with value at least $t_i$ lies at some level $c_h$.

The compact set $\varphi(D^k)$ lies in some $M^{T'}$ with $T'\ge T_N$. There are no
critical values in $[T_N,T']$, so (M1) moves $\varphi$ into $M^{T_N}$, fixing $M^{T_N}$
and hence $\varphi|S^{k-1}$.

Next let $1\le h\le N$, and suppose the current map sends $D^k$ into $M^{T_h}$ and
agrees with $\varphi$ on $S^{k-1}$. Choose
$\varepsilon>0$ as in (M2) for $c=c_h$, with
$T_{h-1}<c_h-\varepsilon<c_h+\varepsilon<T_h$. By (M1), the map moves into
$M^{c_h+\varepsilon}$. By (M2), it moves into
$M^{c_h-\varepsilon}\cup\mathsf D_1\cup\cdots\cup\mathsf D_\gamma$. All the dimensions
$\iota_l$ are at least $d_{i,S}>k$ by the index bound, because $c_h>t_i$. The sublevel is
closed, the cells are compact and pairwise disjoint, and
$\mathsf D_l\cap M^{c_h-\varepsilon}=\partial\mathsf D_l$, so this union is obtained from
$M^{c_h-\varepsilon}$ by attaching the cells one at a time. Applying (M3)
once for each cell moves the map into $M^{c_h-\varepsilon}$. A final use of (M1) moves it
into $M^{T_{h-1}}$. All these homotopies are rel $S^{k-1}$, because
$\varphi(S^{k-1})\subseteq\mathcal A\subseteq M^{T_{h-1}}$ and the deformations in (M1) and
(M2) fix the smaller sublevel pointwise. Proceeding from $h=N$ to $h=1$, we obtain a map
$\varphi_2:D^k\to M^{t_i}$, homotopic to $\varphi$ rel $S^{k-1}$.

It remains to pass to the open sublevel. Choose a short interval
$[t_i-\varepsilon_0,t_i]$ without critical values, and let $\varrho_\lambda$ be the
deformation (M1) of $M^{t_i}$ onto $M^{t_i-\varepsilon_0}$. By~\eqref{eq:milnor-track},
it does not increase $F_S$. The closed set $\{x\in D^k:F_S(\varphi_2(x))=t_i\}$ is
disjoint from $S^{k-1}$. Choose a continuous $\zeta:D^k\to[0,1]$ equal to $0$ on
$S^{k-1}$ and equal to $1$ on this set. The homotopy
$(x,\lambda)\mapsto\varrho_{\lambda\zeta(x)}(\varphi_2(x))$ is rel $S^{k-1}$. Its end map
takes values in $\mathcal A$: if $F_S(\varphi_2(x))=t_i$, the value is
$\varrho_1(\varphi_2(x))\in M^{t_i-\varepsilon_0}\subseteq\mathcal A$, and otherwise $F_S$ at
the value is at most $F_S(\varphi_2(x))<t_i$. This proves the claim.

\proofstep{Connectivity.}
Let $0\le j\le d_{i,S}-2$ and let $g:S^j\to\mathcal A$ be a map. Since $\mathcal X_S$ is
contractible, $g$ extends to a map $\varphi:D^{j+1}\to \mathcal X_S$. The claim with $k=j+1$
makes $\varphi$ homotopic rel $S^j$ to a map into $\mathcal A$, and this is an extension
of $g$ over $D^{j+1}$ inside $\mathcal A$. Together with nonemptiness, this shows that
$\mathcal A_{i,S}$ is $(d_{i,S}-2)$-connected.
Radial scaling identifies
$\mathcal C_{i,S}$ with $\mathcal A_{i,S}\times(0,\infty)$, through
$x\mapsto(x/\nu_i(x),\nu_i(x))$, whose inverse is $(y,\lambda)\mapsto\lambda y$. This
proves (ii).

\proofstep{The two quantitative inclusions.}
Let $x\in\mathcal C_{i,S}$ and $y=x/\nu_i(x)\in\mathcal A_{i,S}$. The barrier terms are
nonnegative and $-\langle\ell,y\rangle\le\sqrt e\|\ell\|_\infty\le\xi_i/4$. Thus
\[
 F(y)\le F_S(y)+\xi_i/4<t_i+\xi_i/4<b_i'+\xi_i/4<\tau_i^p,
\]
where the last step uses $\xi_i\le\tau_i^p-b_i'$.
Equation~\eqref{eq:gauge} gives
\[
 \|Ax\|_p=\nu_i(x)F(y)^{1/p}<\tau_i\nu_i(x)
 \le\frac{\tau_i\|x\|_{q_i}}{1-\sigma/8}\le(1+\sigma)\tau_i\|x\|_{q_i},
\]
where $(1+\sigma)(1-\sigma/8)\ge1$ because $\sigma<1$. The wall inequality
$w_v(y)>0$ gives
$s_vx_{j_v}=\nu_i(x)s_vy_{j_v}>\delta_i\nu_i(x)\ge\delta_i\|x\|_{q_i}$.
This proves (iii).

Conversely, let $x$ satisfy the hypotheses in (iv), and put $y=x/\nu_i(x)\in\Sigma$. The
bound $\|x\|_{q_i}\le\nu_i(x)$ gives $F(y)^{1/p}=\|Ax\|_p/\nu_i(x)\le(1-\sigma)\tau_i$,
that is, $F(y)\le a_i$. The bound $\|x\|_{q_i}/\nu_i(x)\ge1-\sigma/8$ gives
\[
 w_v(y)\ge\delta_i((1+\sigma)(1-\sigma/8)-1)
 =\delta_i\sigma(7/8-\sigma/8)
 \ge\delta_i\sigma/2,
\]
as $\sigma<1$. So $y\in \mathcal X_S$ and $\ln(1/w_v(y))\le\ln(2/(\delta_i\sigma))$ for each $v\in S$. Since
$|S|\le n$, the barrier sum is at most $\xi_i/4$. Also
$\langle\ell,y\rangle\le\xi_i/4$. Therefore
$F_S(y)\le a_i+\xi_i/2<a_i+\xi_i\le b_i<t_i$.
This proves (iv), and completes the construction of the cones.
\end{proof}
We summarize the order of the choices. The parameters of Section~\ref{app:parameters} depend on $(m,n)$ alone. The number
$s$ is chosen next, and $b$, $\ell$, and $t_i$ are chosen afterward at each stage. Thus
all the cones are fixed at this point, and the constructions of Sections~\ref{app:hulls} and~\ref{app:construction} do not change them.

\section{Weighted hulls, cell attachments, and general position}\label{app:hulls}
By Theorem~\ref{thm:carrier}, the upper bound reduces to the construction of
nested acyclic sets in the faces of the cube, outside $N[C]$, and Proposition~\ref{prop:cones}
provides highly connected cones with which these sets can be built. The construction
in Section~\ref{app:construction} attaches cells that carry both a point of the cube
and a vector in a cone, and it adjoins the convex hulls of the vectors lying over
one cube point. This section develops the tools for that construction. The central
one is the weighted hull lemma: a convex combination of vectors of $\ell_1$ norm about
one from the cones at stages $i_1,\ldots,i_k$, whose clamped images lie near one cube
point, belongs to the cone at stage $i_1+\cdots+i_k$. The other tools are a recursion that removes homology
through a given degree by attaching cells of bounded dimension, and generic polynomial
perturbations that bound the dimension of the loci where several cells lie over one
cube point.

Throughout this section and Section~\ref{app:construction} we are in the case $a\ge1$ of
Section~\ref{app:parameters}. Thus $n>c_*>700$, $L\le n$, $\sigma<0.001$,
and~\eqref{eq:parameter-bounds} holds for $1\le i\le L$.
A stage is an integer between $1$ and $L$. It indexes the parameters $q_i$, $\tau_i$,
$\delta_i$ of Section~\ref{app:parameters} and the cones $\mathcal C_{i,S}$, and we call
$\tau_i$ the level at stage~$i$. The letter $S$ denotes a consistent nonempty set of
literals, so that the cones $\mathcal C_{i,S}$ are defined. As in
Section~\ref{app:parameters}, the shell is $\mathcal S=\{x:1/2<\|x\|_1<2\}$.
For $x\in\mathcal C_{i,S}$, we call the bound on $\|Ax\|_p$ in
Proposition~\ref{prop:cones}(iii) the energy bound, and the lower bounds on $s_vx_{j_v}$
there the wall inequalities.

Two notational conventions apply in both sections. We write $\mathrm e$ for Euler's
number, because $e$ denotes a stage. The letter $a$ denotes the parameter $a=n-c_*$,
except that as a subscript or a running index, as in $x_a$, $\sum_a$, and
$1\le a\le k$, it runs over the members of a finite list.

\subsection{Levels and comparison of stages}
We first show that the cones increase with the stage, which requires comparing the
levels, the wall parameters, and the norms $\|\cdot\|_{q_i}$ at different stages. The
levels are calibrated for this comparison and for convex combinations. In the weighted
hull lemma below, a combination of $k$ vectors loses a factor $k^{u_w}$, and the first
inequality of the next lemma shows that passing from the stages $i_1,\ldots,i_k$ to
their sum $w$ more than compensates for this factor.
\begin{lemma}[Growth of the levels]\label{lem:levels}
For integers $k\ge2$ and $i_1,\ldots,i_k\ge1$ with $w=\sum_ai_a\le L$,
\[
 \max_a\ln\tau_{i_a}+u_w\ln k\le\ln\tau_w-\sigma_0.
\]
Also $\ln\tau_{i+1}-\ln\tau_i\ge\sigma_0$ for $i<L$.
\end{lemma}
\begin{proof}
\proofstep{Sums of stages.}
The definition $\tau_i=\tfrac12i^{u_1\ln2}$ gives
$\ln\tau_i=-\ln2+u_1\ln2\,\ln i$ for every integer $i\ge1$.
Put $M=\max_ai_a$ and $t=(k-1)/w$. The $k-1$ entries other than a largest
one are each at least one, so $M\le w-k+1$, that is, $M/w\le1-t$. For the same
reason $k\le w$, and this gives $0<t<1$. The power series of the logarithm has positive
coefficients, hence
\[
 \ln\tau_w-\ln\tau_M
 =u_1\ln2\,\ln\frac wM
 \ge u_1\ln2\,[-\ln(1-t)]
 =u_1\ln2\sum_{j\ge1}\frac{t^j}j
 \ge u_1\ln2\,(t+t^2/2).
\]
The inequality $k\le2^{k-1}$, proved by induction on $k$, gives
$\ln k\le(k-1)\ln2$, and therefore
\[
 u_w\ln k=\frac{u_1}w\ln k\le u_1\ln2\,\frac{k-1}w=u_1\ln2\,t.
\]
Subtracting the last two displays and using $t\ge1/w$ and $w\le L$,
\[
 \ln\tau_w-\ln\tau_M-u_w\ln k
 \ge u_1\ln2\,\frac{t^2}2
 \ge\frac{u_1\ln2}{2w^2}
 \ge\frac{u_1\ln2}{2L^2}
 =\sigma_0.
\]
\proofstep{Consecutive stages.}
The function $x\mapsto\ln(1+x)$ is concave on
$[0,1]$, vanishes at $0$, and equals $\ln2\ge\tfrac12$ at $1$. Thus
$\ln(1+x)\ge x/2$ on $[0,1]$, and for $1\le i<L$,
\[
 \ln\tau_{i+1}-\ln\tau_i
 =u_1\ln2\,\ln(1+1/i)
 \ge\frac{u_1\ln2}{2i}
 \ge\frac{u_1\ln2}{2L^2}
 =\sigma_0.
\qedhere
\]
\end{proof}
Three elementary comparisons prepare the proof that the cones increase. For $i<e\le L$,
the wall parameters telescope as
\begin{equation}\label{eq:wall-recursion}
 \delta_i n^{-(1/q_e-1/q_i)}=(1+\sigma)^{2(e-i)}\delta_e.
\end{equation}
Indeed, the definitions of $\delta_i$ and $\delta_e$ give
\[
 \frac{\delta_i}{\delta_e}
 =n^{(1/q_e-1/q_1)-(1/q_i-1/q_1)}(1+\sigma)^{2(e-1)-2(i-1)}
 =n^{1/q_e-1/q_i}(1+\sigma)^{2(e-i)}.
\]
Norm comparison gives
$\|x\|_q\le\|x\|_{q'}\le n^{1/q'-1/q}\|x\|_q$ for $1\le q'\le q$: the first
inequality is the monotonicity of $\ell_q$ norms, and the second is
H\"older's inequality. For $q'=1$ and $q=q_i$ it gives
$\|x\|_{q_i}\le\|x\|_1\le n^{1-1/q_i}\|x\|_{q_i}$. Since
$1-1/q_i=u_i/(1+u_i)\le u_1$ and $n^{u_1}<\sqrt{\mathrm e}$, we obtain
\begin{equation}\label{eq:hull-one-norm}
 \|x\|_{q_i}\le\|x\|_1\le\sqrt{\mathrm e}\,\|x\|_{q_i}.
\end{equation}
Finally, $0<\sigma<0.001$ and $\sigma=\sigma_0/8$ imply
\begin{equation}\label{eq:hull-sigma-log}
 \ln\frac{1+\sigma}{1-\sigma}
 =2\sum_{j\ge0}\frac{\sigma^{2j+1}}{2j+1}
 \le\frac{2\sigma}{1-\sigma^2}
 \le2.1\sigma
 =\frac{2.1}8\,\sigma_0.
\end{equation}
These inequalities and Lemma~\ref{lem:levels} show that the cones increase with the
stage:
\begin{equation}\label{eq:cone-monotone}
 \mathcal C_{i,S}\subseteq\mathcal C_{e,S}\qquad(i<e).
\end{equation}
To prove this, let $x\in\mathcal C_{i,S}$, so that $x\ne0$. We verify the two hypotheses
of Proposition~\ref{prop:cones}(iv) at stage $e$. Summing the second
assertion of Lemma~\ref{lem:levels} over the stages from $i$ to $e$ gives
$\ln\tau_e-\ln\tau_i\ge(e-i)\sigma_0\ge\sigma_0$, and $q_e<q_i$ gives
$\|x\|_{q_i}\le\|x\|_{q_e}$. With these inequalities, the energy bound of
Proposition~\ref{prop:cones}(iii) at stage $i$ becomes
\[
 \|Ax\|_p
 <(1+\sigma)\tau_i\|x\|_{q_i}
 \le(1+\sigma)\exp(-\sigma_0)\tau_e\|x\|_{q_e}
 \le(1-\sigma)\tau_e\|x\|_{q_e},
\]
where the last step is~\eqref{eq:hull-sigma-log}, which gives
$\ln((1+\sigma)/(1-\sigma))<\sigma_0$.
For $v\in S$, Proposition~\ref{prop:cones}(iii), the inequality
$\|x\|_{q_e}\le n^{1/q_e-1/q_i}\|x\|_{q_i}$, and~\eqref{eq:wall-recursion}
improve the wall inequalities to
\[
 s_vx_{j_v}
 >\delta_i\|x\|_{q_i}
 \ge\delta_in^{-(1/q_e-1/q_i)}\|x\|_{q_e}
 =(1+\sigma)^{2(e-i)}\delta_e\|x\|_{q_e}
 \ge(1+\sigma)^2\delta_e\|x\|_{q_e}.
\]
Proposition~\ref{prop:cones}(iv) applies, since $(1+\sigma)^2\ge1+\sigma$.

\subsection{The clamp}
The clamp links the two coordinates of a cell in Section~\ref{app:construction}: the cube
point of a cell stays within $2\eta$ of the clamped image of its vector. We define the
clamp and show that it maps the part of each cone in the shell into the correct face,
and that the relation above keeps the cube point outside $N[C]$.

Define the piecewise-linear clamp
$\phi(x)_j=\max\{-1,\min\{1,x_j/e_0\}\}$. The scalar function
$\operatorname{cl}(t)=\max\{-1,\min\{1,t/e_0\}\}$ is continuous, nondecreasing,
and $(1/e_0)$-Lipschitz. It equals $\operatorname{sgn}t$ for $|t|\ge e_0$ and
$t/e_0$ for $|t|\le e_0$. Thus $\phi$ maps $\mathbb R^n$ into $Q$, and
\begin{equation}\label{eq:clamp-lipschitz}
 \|\phi(x)-\phi(x')\|_\infty\le\|x-x'\|_\infty/e_0.
\end{equation}
On the part of the cones in the shell, the wall inequalities determine the face of the
clamped image, and the energy bound keeps it, together with a neighborhood, away from $N[C]$.
\begin{lemma}[The clamp on the cones]\label{lem:clamp}
If $x\in\mathcal C_{i,S}\cap\mathcal S$, then $\phi(x)\in Q_S$.
Every $y$ with $\|y-\phi(x)\|_\infty\le2\eta$ lies outside $N[C]$.
\end{lemma}
\begin{proof}
\proofstep{The prescribed coordinates.}
The wall inequalities lift the coordinates prescribed by $S$ above the clamp scale.
For $v\in S$, use the wall inequality of Proposition~\ref{prop:cones}(iii), the
lower bound on $\delta_i$ in~\eqref{eq:parameter-bounds}, \eqref{eq:hull-one-norm},
and $\|x\|_1>\tfrac12$ to obtain
\[
 s_vx_{j_v}>\delta_i\|x\|_{q_i}
 \ge\frac{0.58}{20n}\mathrm e^{-1/2}\|x\|_1
 >\frac{0.58}{20n}\cdot0.6\cdot\frac12
 >\frac{0.008}n
 >\frac1{200n}=e_0.
\]
Thus the prescribed coordinates clamp to $s_v$: one has
$\phi(x)_{j_v}=s_v$ for $v\in S$, and with $\phi(x)\in Q$ this says
$\phi(x)\in Q_S$.

\proofstep{Distance from $N[C]$.}
If a point near $\phi(x)$ agreed in sign with a row $c$, the correlation
$\langle c,x\rangle$ would nearly equal $\|x\|_1$, and the energy bound excludes this.
Let $\|y-\phi(x)\|_\infty\le2\eta$, and suppose $c_jy_j\ge0$ for every $j$
for some $c\in C$. This is the case for every $y\in N[C]$, so it suffices
to reach a contradiction.
If $|x_j|\ge e_0$, then $\phi(x)_j=\operatorname{sgn}(x_j)$ and
$2\eta\le\tfrac18<1$, so $y_j\ne0$ has the sign of $x_j$. Then
$c_jy_j\ge0$ forces $c_j=\operatorname{sgn}(x_j)$, that is,
$c_jx_j=|x_j|$. If $|x_j|<e_0$, then
$c_jx_j\ge-|x_j|\ge|x_j|-2e_0$. Summing over all $j$ gives
\[
 \langle c,x\rangle\ge\|x\|_1-2ne_0
 =\|x\|_1-1/100.
\]
On the other hand, $c$ is a row of $A$, so
$\langle c,x\rangle\le\|Ax\|_\infty\le\|Ax\|_p$. Then
Proposition~\ref{prop:cones}(iii), the bound
$(1+\sigma)\tau_i\le0.71$ of~\eqref{eq:parameter-bounds}, and
$\|x\|_{q_i}\le\|x\|_1$ give
$\langle c,x\rangle<0.71\|x\|_1$.
This would force $0.29\|x\|_1<1/100$, contrary to $\|x\|_1>1/2$.
\end{proof}

\subsection{Convex combinations over one cube point}
We now consider several vectors whose clamped images lie within $2\eta$ of one point
$z$. Such vectors nearly share a closed orthant. As a consequence, a convex combination
loses little $\ell_1$ norm, and, for vectors in the shell taken from the cones, it belongs
to the cone at the stage given by the sum of their stages.
\begin{lemma}[Vectors with nearby clamped images]\label{lem:close-hulls}
Suppose $\|z-\phi(x_a)\|_\infty\le2\eta$ for $1\le a\le k$, and
$y=\sum_a\lambda_ax_a$ is a convex combination. Then
$\|z-\phi(y)\|_\infty\le2\eta$ and
\[
 \min_a\|x_a\|_1-n\theta\le\|y\|_1\le\max_a\|x_a\|_1.
\]
At every coordinate where two of the $x_a$ have opposite signs, all
$|x_{a,j}|\le\theta$. Setting these coordinates to zero puts all the
vectors into a common closed orthant.
\end{lemma}
\begin{proof}
\proofstep{The clamp bound.}
The vectors whose clamped image lies within $2\eta$ of $z$ form a convex set.
Indeed, the inverse image under the nondecreasing scalar clamp $\operatorname{cl}$ of an
interval is an interval. Therefore, for each $j$, the set
$I_j=\{t:|z_j-\operatorname{cl}(t)|\le2\eta\}$ is an interval, and
$\{x:\|z-\phi(x)\|_\infty\le2\eta\}=\prod_jI_j$ is convex. It contains
all the $x_a$, hence $y$. This proves the first assertion.

\proofstep{Conflicted coordinates.}
Entries of opposite signs in one coordinate must be small, because their clamped values
are both close to $z_j$.
Call $j$ conflicted if $x_{a,j}>0>x_{b,j}$ for some $a,b$, and let $J$ be
the set of conflicted coordinates. For such $a,b,j$, the scalar clamp keeps
the sign of its argument, so $\phi(x_a)_j>0>\phi(x_b)_j$, and the triangle
inequality gives
\[
 \phi(x_a)_j-\phi(x_b)_j
 \le|\phi(x_a)_j-z_j|+|z_j-\phi(x_b)_j|
 \le4\eta\le\tfrac14<1.
\]
A value $\phi(x_a)_j=1$ or $\phi(x_b)_j=-1$ would make the left side larger
than one, so neither is clamped at an endpoint. Hence
$\phi(x_a)_j=x_{a,j}/e_0$, $\phi(x_b)_j=x_{b,j}/e_0$, and the display
becomes $x_{a,j}+|x_{b,j}|\le4\eta e_0=\theta$.
Every third nonzero entry $x_{c,j}$ conflicts with one of these two, and
the same inequality for that pair gives $|x_{c,j}|\le\theta$.
Zero entries satisfy the bound trivially. This proves the third assertion.

\proofstep{Common orthant and norms.}
Away from the conflicted coordinates the vectors agree in sign, and this gives the last
assertion and the bounds on $\|y\|_1$.
For $j\notin J$, the nonzero entries among $x_{1,j},\ldots,x_{k,j}$ all have
one sign. So the vectors with the coordinates in $J$ set to zero lie in one
closed orthant. Moreover, on the coordinates $j\notin J$ there is no
cancellation: $|y_j|=\sum_a\lambda_a|x_{a,j}|$. Therefore
\[
 \|y\|_1
 \ge\sum_{j\notin J}|y_j|
 =\sum_a\lambda_a\sum_{j\notin J}|x_{a,j}|
 \ge\sum_a\lambda_a\bigl(\|x_a\|_1-|J|\theta\bigr)
 \ge\min_a\|x_a\|_1-n\theta.
\]
The removed mass is at most $n\theta$ since $|J|\le n$. The upper bound
is the triangle inequality $\|y\|_1\le\sum_a\lambda_a\|x_a\|_1$.
\end{proof}

Inside one closed orthant, the $\ell_q$ norm with $q$ close to one is nearly additive, so
a convex combination is not much shorter than the average of its terms. The next lemma
turns this observation into the rule that stages add under convex combination. This rule
is the reason why the cells of Section~\ref{app:construction} carry weights.
\begin{lemma}[Weighted hull lemma]\label{lem:weighted-hull}
Let $k\ge2$, $w=\sum_ai_a\le L$, and
$x_a\in\mathcal C_{i_a,S}\cap\mathcal S$. Suppose
$\|z-\phi(x_a)\|_\infty\le2\eta$ for all $a$.
Every convex combination $y$ of these vectors with $\|y\|_1>1/2$
belongs to $\mathcal C_{w,S}$.
\end{lemma}
\begin{proof}
We verify the two hypotheses of Proposition~\ref{prop:cones}(iv) at stage $w$.
Write $q'=q_w$ and $y=\sum_a\lambda_ax_a$. Since $k\ge2$ and all $i_a\ge1$,
we have $i_a<w$, hence $q'<q_{i_a}$, for every $a$.
Set to zero the coordinates at which two of the vectors have opposite signs, as in
Lemma~\ref{lem:close-hulls}, to obtain
$x_a'$, and put $y'=\sum_a\lambda_ax_a'$.

\proofstep{Hulls in one orthant.}
We first bound the average of the norms $\|x_a\|_{q'}$ by the norm of $y$, which is the
estimate~\eqref{eq:hull-mass} below.
Let $v_1,\ldots,v_k$ lie in one closed orthant. Then
$|\sum_av_{a,j}|=\sum_a|v_{a,j}|$ for each $j$. Superadditivity of $t^{q'}$
on $[0,\infty)$, applied in each coordinate, gives the first inequality below, and
H\"older's inequality for the sum over $a$ gives the second:
\[
 \Big\|\sum_av_a\Big\|_{q'}
 \ge\Big(\sum_a\|v_a\|_{q'}^{q'}\Big)^{1/q'}
 \ge k^{1/q'-1}\sum_a\|v_a\|_{q'}.
\]
Apply this with $v_a=\lambda_ax_a'$. By Lemma~\ref{lem:close-hulls},
$x_a-x_a'$ is supported on at most $n$ coordinates, on which its entries have
absolute value at most $\theta$. So
$\|x_a-x_a'\|_{q'}\le\|x_a-x_a'\|_1\le n\theta$, and also
$\|y-y'\|_{q'}\le\sum_a\lambda_a\|x_a-x_a'\|_{q'}\le n\theta$.
With $\|y'\|_{q'}\le\|y\|_{q'}+n\theta$ and $k^{1-1/q'}\ge1$ for the last step,
\[
 \sum_a\lambda_a\|x_a\|_{q'}
 \le\sum_a\lambda_a\|x_a'\|_{q'}+n\theta
 \le k^{1-1/q'}\|y'\|_{q'}+n\theta
 \le k^{1-1/q'}(\|y\|_{q'}+2n\theta).
\]
By~\eqref{eq:hull-one-norm}, $\|y\|_{q'}\ge\mathrm e^{-1/2}\|y\|_1>0.3$, so that
$2n\theta\le(2/0.3)n\theta\|y\|_{q'}\le7n\theta\|y\|_{q'}$. Also
$1-1/q'=u_w/(1+u_w)\le u_w$. Therefore
\begin{equation}\label{eq:hull-mass}
 \sum_a\lambda_a\|x_a\|_{q'}
 \le k^{u_w}(1+7n\theta)\|y\|_{q'}.
\end{equation}

\proofstep{The energy bound.}
The growth of the levels absorbs the loss in~\eqref{eq:hull-mass}.
Proposition~\ref{prop:cones}(iii) and
$\|x_a\|_{q_{i_a}}\le\|x_a\|_{q'}$ give
$\|Ax_a\|_p<(1+\sigma)\tau_{i_a}\|x_a\|_{q'}$. The triangle inequality for
$\|A\cdot\|_p$, \eqref{eq:hull-mass}, and Lemma~\ref{lem:levels} in the form
$\max_a\tau_{i_a}k^{u_w}\le\exp(-\sigma_0)\tau_w$ now give
\begin{align*}
 \|Ay\|_p
 &<(1+\sigma)\max_a\tau_{i_a}\sum_a\lambda_a\|x_a\|_{q'}\\
 &\le(1+\sigma)(1+7n\theta)\exp(-\sigma_0)\tau_w\|y\|_{q'}
 \le(1-\sigma)\tau_w\|y\|_{q'}.
\end{align*}
For the last step we need
$\ln((1+\sigma)/(1-\sigma))+\ln(1+7n\theta)\le\sigma_0$.
This follows from~\eqref{eq:hull-sigma-log} and
$\ln(1+7n\theta)\le7n\theta\le7n\theta_0\le\sigma_0/4$, where
$\theta\le\theta_0\le\sigma_0/(28n)$. The sum is at most
$(2.1/8+1/4)\sigma_0<\sigma_0$.

\proofstep{The walls.}
The left side of a wall inequality is linear in the vector and its right side is convex,
so the inequalities pass to convex combinations, and~\eqref{eq:wall-recursion} provides
the room needed at stage $w$.
Let $v\in S$. By Proposition~\ref{prop:cones}(iii), then
$\|x_a\|_{q'}\le n^{1/q'-1/q_{i_a}}\|x_a\|_{q_{i_a}}$, then
\eqref{eq:wall-recursion} with $i=i_a<e=w$, and finally the triangle
inequality,
\begin{align*}
 s_vy_{j_v}=\sum_a\lambda_as_vx_{a,j_v}
 &>\sum_a\lambda_a\delta_{i_a}\|x_a\|_{q_{i_a}}
 \ge\sum_a\lambda_a\delta_{i_a}n^{-(1/q'-1/q_{i_a})}\|x_a\|_{q'}\\
 &\ge(1+\sigma)^2\delta_w\sum_a\lambda_a\|x_a\|_{q'}
 \ge(1+\sigma)\delta_w\|y\|_{q'}.
\end{align*}
Since $y\ne0$, Proposition~\ref{prop:cones}(iv) finishes the proof.
\end{proof}

\subsection{Removing homology by bounded-dimensional attachments}
In the induction of Section~\ref{app:stage-induction}, the set constructed over a face
has reduced homology that vanishes in a range of degrees above some integer $z$ and is
free in degree $z$, and the homology through degree $z$ must be removed by cells of
dimension at most $z+1$. The next lemma specifies the order of cell attachments used in
each face. Its formulation reflects two features of that application. The set over a face
has no prescribed cell structure, so the attaching homotopy classes are chosen in the
space constructed so far, one dimension at a time. Moreover, the attaching maps are
semialgebraic representatives of these classes, so the conclusion has to hold for
arbitrary representatives.
\begin{lemma}[Removing homology by attachments]\label{lem:killing}
Let $Y$ be homeomorphic to a finite CW complex, possibly empty. Let
$r\ge z\ge-1$ be integers, with
$\widetilde H_j(Y)=0$ for $z<j\le r$ and with $H_z(Y)$ free abelian
when $z\ge1$. There is a recursion attaching finitely many $i$-cells
for $i=0,1,\ldots,z+1$, after which all reduced homology through degree
$r$ vanishes. At each dimension, the attaching homotopy classes are
chosen in the current space before any cell of that dimension is
attached. Arbitrary representatives of these classes give the conclusion.
For $z=-1$ and nonempty $Y$, no cell is required.
\end{lemma}
\begin{proof}
Homology is singular with integer coefficients. Put $Y_{-1}=Y$, and let $Y_i$
be $Y_{i-1}$ with the $i$-cells of the recursion attached.

\proofstep{One batch of cells.}
We first record how a batch, that is, the attachment of finitely many cells of one
dimension, changes homology and the fundamental group.
Let $X$ be a nonempty space, let $i\ge1$, and let $X'$ be obtained from $X$
by attaching finitely many $i$-cells along arbitrary continuous maps
$\alpha_b:S^{i-1}\to X$, $1\le b\le\beta$. The outer collars of the balls
give a neighborhood of $X$ in $X'$ that deformation retracts onto $X$, and
the characteristic maps induce a homeomorphism of quotient spaces
$\bigsqcup_bD^i/\bigsqcup_bS^{i-1}\to X'/X$. By
\cite[Proposition 2.22]{Hatcher02}, they induce isomorphisms
$\bigoplus_bH_k(D^i,S^{i-1})\cong H_k(X',X)$. Thus:
\begin{enumerate}[label=(\roman*),leftmargin=2em]
\item $H_k(X',X)=0$ for $k\ne i$, and $H_i(X',X)$ is free abelian with one
basis element for each cell;
\item by naturality of the connecting homomorphism,
$\partial:H_i(X',X)\to\widetilde H_{i-1}(X)$ sends the basis element of the
$b$th cell to the class of its attaching map, $(\alpha_b)_*[S^{i-1}]$, where
for $i=1$ the class $[S^0]$ is the difference of the two points of $S^0$.
\end{enumerate}
The reduced exact sequence of the pair therefore splits into isomorphisms
$\widetilde H_k(X)\cong\widetilde H_k(X')$ for $k\notin\{i-1,i\}$ and the
exact sequence
\begin{equation}\label{eq:batch-sequence}
 0\to H_i(X)\to H_i(X')\to H_i(X',X)\xrightarrow{\ \partial\ }
 \widetilde H_{i-1}(X)\to\widetilde H_{i-1}(X')\to0.
\end{equation}
If $\partial$ is surjective, then $\widetilde H_{i-1}(X')=0$. If $\partial$ is
bijective, then also $H_i(X)\cong H_i(X')$. If all homology groups of $X$ are
finitely generated, then~\eqref{eq:batch-sequence} shows that those of $X'$
are, because the attachment has finitely many cells. For fundamental groups we
use \cite[Proposition 1.26]{Hatcher02}: for path connected $X$, attaching
$2$-cells divides $\pi_1(X)$ by the normal closure of the attaching loops, and
attaching cells of dimension at least $3$ does not change $\pi_1(X)$.
We refer to the statements of this step as the batch facts.

\proofstep{Empty $Y$ and $z=-1$.}
These cases need at most one cell.
If $Y$ is empty, first add one point. Then $Y_0$ is a point, all its reduced
homology vanishes, and all subsequent batches are empty. This covers $Y=\varnothing$
for every $z$. From now on $Y\ne\varnothing$, and no $0$-cell is added, so
$Y_0=Y$. For $z=-1$ this finishes the task: the hypothesis gives
$\widetilde H_j(Y)=0$ for $0\le j\le r$, and $\widetilde H_{-1}(Y)=0$ since
$Y\ne\varnothing$.

\proofstep{Dimension one.}
Let $z\ge0$. The $1$-cells connect the path components of $Y$ and change no other
homology. The finite CW complex $Y$ has finitely many path components, and
a homotopy class of maps $S^0\to Y$ is an ordered pair of them. Choose the
classes given by the edges of a spanning tree on the set of components, and
attach arcs along representatives, joining the components in a tree.
The resulting $Y_1$ is path connected. By induction on the number of components,
removing a leaf of the tree, the classes of the attaching maps form a basis of
$\widetilde H_0(Y)$. So $\partial$ is bijective in
\eqref{eq:batch-sequence} with $i=1$, and $\widetilde H_0(Y_1)=0$ and
$H_j(Y_1)\cong H_j(Y)$ for $j\ge1$. For $z=0$ this is sufficient.

\proofstep{The case $z=1$.}
Since $H_1(Y)$ is free, $2$-cells can remove it without creating homology in degree $2$.
The group $H_1(Y_1)\cong H_1(Y)$ is free abelian by hypothesis, and of finite
rank since $Y$ is a finite CW complex. The Hurewicz map from $\pi_1(Y_1)$ to
$H_1(Y_1)$ is surjective, so a basis is represented by loops. Attach $2$-cells
along loops representing this basis. The boundary map
$\partial:H_2(Y_2,Y_1)\to H_1(Y_1)$ sends a basis to a basis, so it is an
isomorphism.
By~\eqref{eq:batch-sequence} with $i=2$, $H_1(Y_2)=0$ and
$H_2(Y_2)\cong H_2(Y_1)$, and there is no change in higher degrees.
Thus $H_j(Y_2)\cong H_j(Y)$ for all $j\ge2$.

\proofstep{Connectivity for $z\ge2$.}
For $z\ge2$, cells of dimensions $2,\ldots,z$ make the space $(z-1)$-connected, so that
the Hurewicz theorem represents its homology in degree $z$ by maps of spheres.
Sliding the ends of an arc inside a path component does not change the homotopy
type of $Y_1$ \cite[Proposition 0.18]{Hatcher02}, and every path component of
$Y$ contains a vertex. So $Y_1$ is homotopy equivalent to a connected finite CW
complex, and $\pi_1(Y_1)$ is finitely generated.
Attach $2$-cells along a finite generating set of
$\pi_1(Y_1)$. By the batch facts, the space $Y_2$ is simply connected.
Let $3\le i\le z$ and suppose that $Y_{i-1}$ is $(i-2)$-connected. By the
Hurewicz theorem \cite[Theorem 4.32]{Hatcher02},
$\pi_{i-1}(Y_{i-1})\cong H_{i-1}(Y_{i-1})$ through the Hurewicz map, and this
group is finitely generated by the batch facts. Attach $i$-cells along a finite
generating set. Then $\partial$ is surjective in~\eqref{eq:batch-sequence}, so
$\widetilde H_{i-1}(Y_i)=0$, and $\widetilde H_k(Y_i)\cong\widetilde H_k(Y_{i-1})=0$
for $k\le i-2$. Also $Y_i$ is simply connected, as $i\ge3$. Applying the Hurewicz
theorem degree by degree, $Y_i$ is $(i-1)$-connected.
Thus $Y_z$ is $(z-1)$-connected.

\proofstep{Relative groups for $z\ge2$.}
The relative homology of $(Y_z,Y)$ shows that these attachments keep the group in degree
$z$ free and leave homology in higher degrees unchanged.
We claim that for $0\le i\le z$, the group $H_k(Y_i,Y)$ vanishes for $k>i$, and
$H_i(Y_i,Y)$ is free of finite rank. For $i=0$ the pair is $(Y,Y)$. For $i\ge1$,
use the exact sequence of the triple $(Y_i,Y_{i-1},Y)$,
\[
 H_k(Y_{i-1},Y)\to H_k(Y_i,Y)\to H_k(Y_i,Y_{i-1}).
\]
For $k>i$ both outer groups vanish, by induction and by the batch facts. For
$k=i$ the left group vanishes by induction, so $H_i(Y_i,Y)$ embeds in the free
group $H_i(Y_i,Y_{i-1})$ of finite rank.

In the exact sequence of the pair $(Y_z,Y)$, the relative groups vanish above
degree $z$. So
\[
 0\to H_z(Y)\to H_z(Y_z)\to H_z(Y_z,Y),
 \qquad
 H_j(Y)\cong H_j(Y_z)\quad(j\ge z+1).
\]
This embeds $H_z(Y)$ into $H_z(Y_z)$ and makes the quotient a subgroup of the
free group $H_z(Y_z,Y)$. The quotient is therefore free, so the extension splits.
The group $H_z(Y)$ is free by hypothesis, and of finite rank since $Y$ is a finite
CW complex. Therefore $H_z(Y_z)$ is free of finite rank.

\proofstep{The last batch for $z\ge2$.}
A basis of the free group $H_z(Y_z)$ determines the last batch.
Attach $(z+1)$-cells along a basis of
$\pi_z(Y_z)\cong H_z(Y_z)$. In~\eqref{eq:batch-sequence} with $i=z+1$, the map
$\partial$ sends a basis to a basis. So it kills degree $z$ without creating
homology in degree $z+1$:
$H_z(Y_{z+1})=0$ and $H_{z+1}(Y_{z+1})\cong H_{z+1}(Y_z)\cong H_{z+1}(Y)$.
In degrees at least $z+2$, homology is that of $Y$, and in degrees below $z$ it
is that of $Y_z$, which vanishes.

\proofstep{Conclusion.}
In all cases with $Y\ne\varnothing$, the preceding steps give
\begin{equation}\label{eq:killing-conclusion}
 \widetilde H_j(Y_{z+1})=0\quad(j\le z),
 \qquad
 \widetilde H_j(Y_{z+1})\cong\widetilde H_j(Y)\quad(j>z).
\end{equation}
By hypothesis the groups in the second family vanish for $z<j\le r$.

\proofstep{Arbitrary representatives.}
Finally, we check that only the homotopy classes of the attaching maps enter the argument.
The classes of dimension $i$ are chosen in $Y_{i-1}$, and the steps above use the
attaching maps of dimension $i$ only through the following data. For $i=1$ they use the
pair of components joined. For $i=2$ they use the homology classes of the loops
if $z=1$, and the normal closure of the loops in $\pi_1(Y_1)$ if $z\ge2$. For $i\ge3$ they
use the homology classes of the attaching maps. Replacing an
attaching map by a homotopic one changes none of these: homotopic maps have the
same homology class, and freely homotopic loops are conjugate after joining them
to the base point, so the normal closure is unchanged. For $i\ge3$ the space
$Y_{i-1}$ is simply connected, so that free homotopy classes of maps
$S^{i-1}\to Y_{i-1}$ are the elements of $\pi_{i-1}(Y_{i-1})$.
The batch facts hold for arbitrary attaching maps, so they give finite generation
of all homology groups of all the $Y_i$ without any cell structure on $Y_i$. Thus
every step depends only on the chosen classes, and the argument applies to all
their representatives.
\end{proof}

\subsection{Independent polynomial perturbations}
When a cell is attached in Section~\ref{app:construction}, its base coordinate, a map to
the cube, should meet the existing cells only in the expected dimension, and its vector
coordinate, a map to $\mathbb R^n$, should separate the points where the base coordinate
still coincides. We achieve both by adding a small generic polynomial,
multiplied by a function that vanishes on the boundary of the ball, so that the boundary
values do not change. The polynomials have degree at most $n$, because the values of such
polynomials at up to $n+1$ distinct points are independent, and the incidence counts below
rest on this independence.

We use the facts on semialgebraic sets, maps, and dimension recalled in
Section~\ref{app:carrier}. In addition, a semialgebraic bijection preserves dimension,
a nonempty open subset of $\mathbb R^N$ has dimension $N$, and the empty set has
dimension $-\infty$. Thus a negative upper bound on a dimension means that the set is
empty.

Each count adds the coefficients of the polynomial as variables, bounds the dimension of
the resulting incidence set, and then reads off the dimension of its fibers over a generic
polynomial. Both bounds come from the following form of Hardt's theorem.
\begin{lemma}[Fiber dimension]\label{lem:hardt-fiber-dimension}
Let $\pi:I\to T$ be a continuous semialgebraic map between semialgebraic sets, with
$T\ne\varnothing$.
\begin{enumerate}[label=(\alph*),leftmargin=2em]
\item If all fibers of $\pi$ have dimension at most $d$, then
$\dim I\le\dim T+d$.
\item There is a semialgebraic subset $E\subseteq T$ with $\dim E<\dim T$ such that
every fiber of $\pi$ over $T\setminus E$ has dimension at most $\dim I-\dim T$.
\end{enumerate}
\end{lemma}
\begin{proof}
By Hardt's semialgebraic triviality theorem \cite[Section 9.3]{BCR98},
$T$ has a finite semialgebraic partition $T=T_1\sqcup\cdots\sqcup T_N$ over each
member of which the projection is trivial: there are a semialgebraic set $F_l$ and a
semialgebraic homeomorphism $\pi^{-1}(T_l)\to T_l\times F_l$ compatible with the
projections to $T_l$. We discard empty members. Then
$\dim\pi^{-1}(T_l)=\dim T_l+\dim F_l$, and each fiber of $\pi$ over $T_l$ is
semialgebraically homeomorphic to $F_l$. For (a),
\[
 \dim I=\max_l\dim\pi^{-1}(T_l)=\max_l(\dim T_l+\dim F_l)\le\dim T+d.
\]
For (b), let $E$ be the union of the members with $\dim T_l<\dim T$. On a
full-dimensional member, $\dim T+\dim F_l=\dim\pi^{-1}(T_l)\le\dim I$, so the
dimension of a nonempty fiber is at most $\dim I-\dim T$. Over such a member the
fiber is empty if $\dim I-\dim T<0$, because a nonempty $F_l$ has $\dim F_l\ge0$.
\end{proof}

The polynomials will be chosen generically in the following sense.
A property of the elements of a finite-dimensional real vector space $\mathcal V$ is
called generic if it holds outside a semialgebraic subset of $\mathcal V$ of dimension
less than $\dim\mathcal V$. Such a subset has empty interior, because a nonempty open
subset of $\mathcal V$ has dimension $\dim\mathcal V$. Finite unions of such subsets
are again of this kind. Thus finitely many generic properties hold simultaneously
on a dense subset of $\mathcal V$, and in particular on elements arbitrarily close
to $0$.

We now fix the objects used in the perturbations.
Let $D=D^f$ be the closed unit ball, with $D^0$ a point, equal to its own
interior, and put
$t(w)=\min\{1,2(1-|w|)\}$, with $t=1$ for $f=0$. Thus $t>0$ on
$\operatorname{int}D$, $t=0$ on $\partial D$, and $t=1$ for $|w|\le\tfrac12$.
For a vector space $V$, let $\mathcal P(V)$ be the polynomials
$\mathbb R^f\to V$ of degree at most $n$, with their coefficients as coordinates. We
write $\|P\|_\infty=\sup_{w\in D}\|P(w)\|_\infty$. On $D$ every monomial has absolute
value at most one, so $\|P\|_\infty$ is at most the sum of the absolute values of the
coefficients of $P$. Therefore each set $\{P:\|P\|_\infty<\rho\}$, $\rho>0$, is a
neighborhood of $0$ in $\mathcal P(V)$, and generic elements can be chosen in it.

At $s\le n+1$ distinct points, polynomial evaluations are independent. In
other words, $P\mapsto(P(w_a))_{a}$ is a surjective linear map $\mathcal P(V)\to V^s$.
To see this, put
\[
 L_a(x)=\prod_{b\ne a}
 \frac{\langle w_a-w_b,x-w_b\rangle}{|w_a-w_b|^2}.
\]
This is a product of $s-1\le n$ affine functions, and it satisfies $L_a(w_b)=\mathbf1_{a=b}$:
the factor with index $b$ vanishes at $w_b$, and every factor equals one at $w_a$.
Thus $P=\sum_aL_ay_a$ realizes arbitrary prescribed values $y_a$.
For $f=0$ only the one-point assertion is needed.

The first perturbation acts on the base coordinate of a new cell. It bounds the dimension
of the self-coincidences of the cell, and of its coincidences with given semialgebraic
families of points, which are described by the maps $v_b$ below.
\begin{lemma}[Generic base coordinate]\label{lem:base-generic}
Let $g_*:D\to\mathbb R^\mu$ be semialgebraic. Let
$v_b:\mathcal L_b\to\mathbb R^\mu$ be finitely many semialgebraic maps.
For $P$ outside a semialgebraic subset of $\mathcal P(\mathbb R^\mu)$
of smaller dimension, $g=g_*+tP$ satisfies, for $1\le s\le n+1$,
\[
 \dim\{(w_1,\ldots,w_s,\ell):g(w_1)=\cdots=g(w_s)=v_b(\ell)\}
 \le s(f-\mu)+\dim\mathcal L_b,
\]
where the $w_a$ are distinct interior points and $\ell\in\mathcal L_b$.
For $2\le s\le n+1$, the locus of distinct interior points satisfying
$g(w_1)=\cdots=g(w_s)$ has dimension at most $sf-(s-1)\mu$.
A negative dimension bound means the locus is empty.
\end{lemma}
\begin{proof}
\proofstep{Coincidences with the maps $v_b$.}
Write $\mathcal P=\mathcal P(\mathbb R^\mu)$. Fix $b$ and $s$ with $1\le s\le n+1$, and
let $Z$ be the set of tuples $(w_1,\ldots,w_s,\ell)$ with pairwise distinct
$w_a\in\operatorname{int}D$ and $\ell\in\mathcal L_b$. If $Z=\varnothing$, which happens
for $\mathcal L_b=\varnothing$ and for $f=0$, $s\ge2$, the first assertion is
vacuous for this pair $(b,s)$. Otherwise $Z$ is semialgebraic of dimension
$sf+\dim\mathcal L_b$.

Include $P$ as a variable: put
\[
 I=\bigl\{(P,w_1,\ldots,w_s,\ell)\in\mathcal P\times Z:\
 g_*(w_a)+t(w_a)P(w_a)=v_b(\ell)\ \ (1\le a\le s)\bigr\},
\]
a semialgebraic set.
At fixed distinct $w_a$ and fixed $\ell$, the equations read
$P(w_a)=(v_b(\ell)-g_*(w_a))/t(w_a)$, because all $t(w_a)>0$.
Since the evaluations are independent, this is a system of $s\mu$ independent
affine conditions on $P$, and its solution set is a nonempty affine subspace of
$\mathcal P$ of codimension $s\mu$. Lemma~\ref{lem:hardt-fiber-dimension}(a), applied to
the projection $I\to Z$, gives
\[
 \dim I\le\dim\mathcal P-s\mu+sf+\dim\mathcal L_b.
\]
Project this incidence set to the polynomial parameter space, $\pi:I\to\mathcal P$.
The fiber of $\pi$ over $P$ is the locus in the statement for $g=g_*+tP$.
Lemma~\ref{lem:hardt-fiber-dimension}(b) gives a semialgebraic subset
$E_{b,s}\subseteq\mathcal P$ of smaller dimension such that for $P\notin E_{b,s}$
this locus has dimension at most
\[
 \dim I-\dim\mathcal P\le s(f-\mu)+\dim\mathcal L_b.
\]
This proves the first generic bound, including emptiness when its right-hand
side is negative.

\proofstep{Self-coincidences.}
For the second assertion, fix $2\le s\le n+1$, and let $Z'$ be the set of tuples of
pairwise distinct interior points, of dimension $sf$. For $f=0$ it is empty and there
is nothing to prove. The incidence set $I'\subseteq\mathcal P\times Z'$ is given by the
equations
\[
 g_*(w_a)+t(w_a)P(w_a)=g_*(w_1)+t(w_1)P(w_1)\qquad(2\le a\le s).
\]
At fixed $(w_a)$ they impose $(s-1)\mu$ independent affine conditions, as the linear
map $P\mapsto(t(w_a)P(w_a)-t(w_1)P(w_1))_{a\ge2}$ is surjective onto
$(\mathbb R^\mu)^{s-1}$: to reach $(y_a)_{a\ge2}$, interpolate the values
$P(w_1)=0$ and $P(w_a)=y_a/t(w_a)$. Thus Lemma~\ref{lem:hardt-fiber-dimension}(a) gives
$\dim I'\le\dim\mathcal P-(s-1)\mu+sf$, and
Lemma~\ref{lem:hardt-fiber-dimension}(b) gives the bound $sf-(s-1)\mu$ outside a semialgebraic
subset $E_s'$ of smaller dimension.

\proofstep{The exceptional set.}
Only finitely many values of $s$ and maps $v_b$ occur, so the union of the exceptional
sets $E_{b,s}$ and $E_s'$ still has smaller dimension. The count is pointwise: it uses
only that $g_*$ and $t$ are semialgebraic and that $t(w_a)>0$. So $g_*$ may collapse sets,
and the points $w_a$ may lie on one radial segment or arbitrarily near $\partial D$.
\end{proof}

The second perturbation acts on the vector coordinate of the new cell. Once the base
coordinate is in general position, the loci to be separated have dimension below $n$, while
the vector coordinate takes values in $\mathbb R^n$, so one evaluation count suffices.
\begin{lemma}[Generic vector coordinate]\label{lem:height-generic}
Let $h_*:D\to\mathbb R^n$ be semialgebraic.
If $\mathcal M$ is semialgebraic of dimension less than $n$, and
$w:\mathcal M\to\operatorname{int}D$ and
$\xi:\mathcal M\to\mathbb R^n$ are semialgebraic, then generic
$R\in\mathcal P(\mathbb R^n)$ satisfies
$(h_*+tR)(w(\cdot))\ne\xi(\cdot)$ throughout $\mathcal M$.
Likewise, on a semialgebraic set of pairs of distinct interior points
of dimension less than $n$, generic $R$ separates the two values of
$h_*+tR$.
\end{lemma}
\begin{proof}
If $\mathcal M=\varnothing$ there is nothing to prove. Otherwise write
$\mathcal P=\mathcal P(\mathbb R^n)$, and let $I$ be the semialgebraic set of
$(R,m)\in\mathcal P\times\mathcal M$ with
$h_*(w(m))+t(w(m))R(w(m))=\xi(m)$.
Each point $m$ of the parameter set imposes $n$ independent affine equations
on $R$, since $t(w(m))>0$ and evaluation at one point is surjective onto $\mathbb R^n$.
Lemma~\ref{lem:hardt-fiber-dimension}(a) gives
$\dim I\le\dim\mathcal M+\dim\mathcal P-n<\dim\mathcal P$. The projection of $I$ to
$\mathcal P$ is therefore a smaller-dimensional semialgebraic set, and every $R$ outside
it satisfies the first assertion.

For the second assertion, let $\mathcal M'$ be the set of pairs. If it is empty, which is
always the case for $f=0$, there is nothing to prove. Otherwise let $I'$ be the set
of $(R,w,w')\in\mathcal P\times\mathcal M'$ with
$h_*(w)+t(w)R(w)=h_*(w')+t(w')R(w')$. Independent two-point evaluation, with
$2\le n+1$, makes $R\mapsto t(w)R(w)-t(w')R(w')$ surjective onto $\mathbb R^n$ for
$w\ne w'$, since $t(w),t(w')>0$. So again $n$ independent affine equations are imposed
at each point of $\mathcal M'$. Lemma~\ref{lem:hardt-fiber-dimension}(a) now gives
$\dim I'\le\dim\mathcal M'+\dim\mathcal P-n<\dim\mathcal P$, and every $R$ outside the
projection of $I'$ to $\mathcal P$ separates the two values on $\mathcal M'$.
\end{proof}
In Lemmas~\ref{lem:base-generic} and~\ref{lem:height-generic} the good choices are generic,
hence dense, so they can be made arbitrarily small in $\|\cdot\|_\infty$.

\section{Compatible carriers and the proof of the upper bound}\label{app:construction}
We now construct the sets required by Theorem~\ref{thm:carrier} and complete the proof of
Theorem~\ref{thm:main}. The clamp maps the part of each cone of
Proposition~\ref{prop:cones} in the shell into the cube and outside $N[C]$, but a continuous
image of a highly connected set need not be
highly connected. We therefore work in $Q\times\mathbb R^n$: the intermediate object
records both a cube point and a vector in one of the cones, and convex hulls are formed
only among vectors with the same cube point. These hulls make the fibers of the final
projection to the cube contractible, so that the projection preserves homology.
Stage weights control which hulls enter at each step: by the weighted hull lemma, a hull of
total weight $e$ lies in a cone at stage $e$, and a general-position count shows that the
hulls of total weight $e$ have dimension small enough for the cones at stage $e$ to fill
the homology they create.
The cones of Proposition~\ref{prop:cones} are fixed before the construction
begins, and nothing below changes them.

\subsection{Cells, ranks, and weighted hulls}
This subsection introduces the objects of the construction and the conditions they satisfy
at every step: cells, the hull sets they generate, and admissible families. We begin with
the faces of the cube.
For $z\in Q$ put $S(z)=\{(j,z_j):|z_j|=1\}$, a consistent set of literals.
The definitions of $Q_S$ and of its relative interior give
\begin{equation}\label{eq:cube-face-facts}
 z\in Q_S\iff S\subseteq S(z),
 \qquad
 z\in\operatorname{int}Q_S\iff S=S(z).
\end{equation}
Thus every point of $Q$ lies in $\operatorname{int}Q_{S'}$ for exactly one
consistent $S'$. For consistent $S$ and $S'$, the set
$\operatorname{int}Q_{S'}$ is contained in $Q_S$ if $S\subseteq S'$, and it is
disjoint from $Q_S$ otherwise.

\begin{definition}[Cell]\label{def:cell}
A cell $B$ has a dimension $f_B\ge0$, a consistent literal set $S_B$
with $1\le|S_B|\le a$, a stage $i_B\ge1$, and maps
\[
 g_B:D^{f_B}\to Q_{S_B},\qquad h_B:D^{f_B}\to\mathbb R^n.
\]
Both maps are continuous and semialgebraic, and
$g_B(\operatorname{int}D^{f_B})\subseteq\operatorname{int}Q_{S_B}$.
We call $g_B$ and $h_B$ the base coordinate and the vector coordinate of $B$, and we say
that $B$ is at $S_B$. The ball $D^0$ is a point; it equals its own interior
and has empty boundary.
We require $f_B\le f_{i_B}(S_B)$ and assign the rank
\begin{equation}\label{eq:cell-rank}
 R(B)=(i_B-1)(n+1)^2+(a-|S_B|)(n+1)+f_B+1.
\end{equation}
\end{definition}
Since $f_{i_B}(S_B)=m_{S_B}-\kappa_{i_B}$, a cell of stage $i$ has codimension at least
$\kappa_i$ in its face. The rank has two uses: the boundary of a cell will be attached to
cells of smaller rank, and the $\ell_1$ norm of the vectors of a cell may deviate from one
by an amount proportional to its rank. We now bound the rank.
Since $0\le f_B\le f_{i_B}(S_B)=n-|S_B|-i_Bc'-3$ and $|S_B|\ge1$, we have
$i_Bc'\le n-4$ and $f_B\le n-1$. As $i_B$ is an integer and
$(n-4)/c'<\lceil n/c'\rceil=L-1$, this gives
\begin{equation}\label{eq:cell-stage-bound}
 i_B\le L-2.
\end{equation}
In particular, the cones $\mathcal C_{i_B,S}$ are defined. Also
$0\le a-|S_B|\le a-1\le n-1$.
The last two digits of $R(B)-1$ in base $n+1$ are therefore $a-|S_B|$ and $f_B$,
and they lie between zero and $n-1$.
Thus the rank compares the triples $(i_B,a-|S_B|,f_B)$ lexicographically, and two cells
have equal rank exactly when their triples agree.
Every possible rank is at most $L(n+1)^2$, since~\eqref{eq:cell-stage-bound} gives
\[
 R(B)\le(L-3)(n+1)^2+(n-1)(n+1)+(n-1)+1\le(L-2)(n+1)^2.
\]
These statements use only the numbers $f_B$, $|S_B|$, $i_B$. So they also apply to the
rank of a prospective stage-$e$, dimension-$f$ cell at $S$, whenever
$0\le f\le f_e(S)$ and $1\le|S|\le a$.

All families of cells below are finite. The interior of a cell can be regarded as a sheet
lying over a cube face. The next definition gives names to the points of these sheets, and
to the convex hulls of the vectors over one cube point.
\begin{definition}[Sheet points and hull sets]\label{def:hull-sets}
For a finite family $\mathcal B$ of cells, a sheet point means a pair
$(B,w)$ with $B\in\mathcal B$ and $w\in\operatorname{int}D^{f_B}$. Its base, vector, and
weight are $g_B(w)$, $h_B(w)$, and $i_B$, and its rank is $R(B)$. When the cell is
understood, we write $g(w)$, $h(w)$, and $i(w)$. The sheet fiber $g^{-1}(z)$ is the set
of all sheet points with base $z$.

For a nonempty finite set $A'$ of sheet points with a common base,
put $\operatorname{wt}(A')=\sum_{w\in A'}i(w)$, and let $\operatorname{conv}h(A')$ be
the convex hull of the vectors of its points.
Let $\mathsf H_e(\mathcal B)$ consist of the pairs $(z,x)$ for which such a
set $A'$ satisfies
\[
 g(A')=\{z\},\qquad \operatorname{wt}(A')\le e,\qquad
 x\in\operatorname{conv}h(A').
\]
We call $A'$ a representation of $(z,x)$, and we call $\mathsf H_e(\mathcal B)$ a weighted hull set,
or briefly a hull set.
Write $\mathsf H_e(\mathcal B)_S=\mathsf H_e(\mathcal B)\cap(Q_S\times\mathbb R^n)$. This is the hull set over
the face $Q_S$.
\end{definition}
We keep the italic letter $H$ for homology.

The base of a sheet point determines the face of its cell. Since
$g_B(\operatorname{int}D^{f_B})\subseteq\operatorname{int}Q_{S_B}$, the
facts~\eqref{eq:cube-face-facts} give
\begin{equation}\label{eq:sheet-face}
 S_B=S(g_B(w))\qquad(w\in\operatorname{int}D^{f_B}).
\end{equation}
Therefore all sheet points with a common base $z\in\operatorname{int}Q_{S'}$ belong to
cells at $S'$, and a cell $B$ has sheet points over $Q_S$ exactly when $S\subseteq S_B$.

The hull sets are compatible with faces and monotone in the family and in the weight
bound. Passing to a facet gives
$\mathsf H_e(\mathcal B)_S\cap(Q_{S+v}\times\mathbb R^n)=\mathsf H_e(\mathcal B)_{S+v}$, because
$Q_{S+v}\subseteq Q_S$. A representation in a family remains a representation in every
larger family and for every larger weight bound, so
\begin{equation}\label{eq:hull-monotone}
 \mathsf H_e(\mathcal B)\subseteq \mathsf H_{e'}(\mathcal B')
 \qquad(\mathcal B\subseteq\mathcal B',\ e\le e').
\end{equation}

The dimension counts of the construction concern the coincidences among sheet points.
For a list of cells $B_1,\ldots,B_k$ at one face $S$, repetitions allowed,
write $P(B_1,\ldots,B_k)$ for the set of tuples of pairwise distinct
sheet points with equal bases, the coincidence set of the list. Distinctness refers to the pairs $(B,w)$,
so different cells have different sheet points even if their parameters
coincide. This is a semialgebraic subset of $\prod_a\operatorname{int}D^{f_{B_a}}$, and
reordering the list permutes its coordinates. If $A'$ is a set of $k\ge1$ sheet points
with a common base in $\operatorname{int}Q_{S'}$, then listing its points gives a tuple in
some $P(B_1,\ldots,B_k)$ with all $B_a$ at $S'$ and
$\operatorname{wt}(A')=\sum_ai_{B_a}$. Conversely, the entries of a tuple in
$P(B_1,\ldots,B_k)$ form such a set.

We can now state the conditions that the construction maintains. A family of cells comes
with its creation order, a total order in which a cell is older than the cells created
after it. For an integer $r$, the family $\mathcal B^{<r}$ consists of the cells of
$\mathcal B$ of rank less than $r$.
\begin{definition}[Admissible family]\label{def:admissible}
A family $\mathcal B$ of cells, equipped with its creation order, is admissible if the
following hold for all its cells $B$ and all lists $B_1,\ldots,B_k$ of its cells at one
face $S$.
\begin{enumerate}[label=(B\arabic*),leftmargin=3em]
\item For every boundary point $w$ of the ball of $B$,
$(g_B,h_B)(w)\in \mathsf H_{i_B}(\mathcal B^{<R(B)})_{S_B}$.
\item $\dim P(B_1,\ldots,B_k)\le m_S-\sum_a\kappa_{i_{B_a}}$,
with emptiness if the right-hand side is negative.
\item At every sheet point of a cell at $S$,
$h(w)\in\mathcal C_{i(w),S}$ and
$\|g(w)-\phi(h(w))\|_\infty\le2\eta$.
\item $(g_B,h_B)$ is injective on the interior of its ball, and its
interior image is disjoint from the fiberwise convex hull of all older
cells: for $w\in\operatorname{int}D^{f_B}$, one has
$h_B(w)\notin\operatorname{conv}h(A')$ for every finite set $A'$ of sheet
points of older cells with base $g_B(w)$.
\item With $\beta=3n\theta$, every sheet point of $B$ satisfies
$1-R(B)\beta\le\|h(w)\|_1\le1+R(B)\beta$.
\end{enumerate}
\end{definition}
Condition (B1) makes every cell an attachment to the hull set of cells of smaller rank.
Condition (B2) says that coincidences behave as in general position, in that the
codimensions $\kappa_{i_{B_a}}$ of the cells add. Condition (B3) places each vector in the
cone at the stage of its cell and ties the base to the clamped vector, and (B4) makes the
attachment injective. Condition (B5) controls the norms, through the norm estimate
\begin{equation}\label{eq:norm-budget}
 R(B)\beta+n\theta
 \le\frac{3L(n+1)^2n}{14Ln^3}+\frac1{14Ln^2}<0.3,
\end{equation}
using $n>700$. Indeed, $R(B)\le L(n+1)^2$ and
$\theta\le\theta_0\le1/(14Ln^3)$ give the first inequality, and the two terms are
$3(n+1)^2/(14n^2)<0.215$ and less than $10^{-6}$. The bound holds with every integer
between $1$ and $L(n+1)^2$ in place of $R(B)$. Lemma~\ref{lem:close-hulls} will
therefore put all the hulls considered below in the shell $\mathcal S$.

We draw three consequences of admissibility. The first is that (B2) limits the number and
the total weight of the sheet points over one cube point, because $\kappa_i=ic'+3$ is large.
\begin{lemma}[Sheet fibers are small]\label{lem:fiber-weight}
For an admissible family and a finite nonempty set $A'$ of sheet points
over $z\in\operatorname{int}Q_{S'}$,
\[
 c'\operatorname{wt}(A')+3|A'|\le m_{S'}\le n-1.
\]
In particular $|A'|\le n$, $\operatorname{wt}(A')\le L-1$, and the
entire sheet fiber over $z$ is finite.
\end{lemma}
\begin{proof}
Every sheet point over $z$ belongs to a cell at $S'$, since its base lies
in that cell's relative interior; see~\eqref{eq:sheet-face}. In particular
$|S'|\ge1$, which gives $m_{S'}\le n-1$. List the points of $A'$ as a tuple
$(w_a)_{a\le k}\in P(B_1,\ldots,B_k)$, with $k=|A'|$. The coincidence set containing
the tuple is nonempty, so the right-hand side in (B2) is not negative. With
$\kappa_i=ic'+3$ this says
\[
 0\le m_{S'}-\sum_a(i_{B_a}c'+3)=m_{S'}-c'\operatorname{wt}(A')-3|A'|,
\]
which is the displayed inequality. Hence $3|A'|\le n-1$, and $|A'|\le n$.
Also $c'\operatorname{wt}(A')\le n-1-3|A'|\le n-4$. The definition of $L$ implies the
weight bound: $\operatorname{wt}(A')<n/c'\le\lceil n/c'\rceil=L-1$.
An infinite fiber would contain finite subsets violating the cardinality
bound, so the fiber is finite, with at most $n$ points.
\end{proof}

Next, the hull sets are compact, even though sheet points are interior points of balls.
This is the role of (B1): a limit of sheet points that reaches the boundary of a ball is
represented by sheet points of cells of smaller rank.
\begin{lemma}[Compactness of hull sets]\label{lem:hull-compact}
For an admissible finite family and every $e\ge1$, $\mathsf H_e(\mathcal B)$
is compact and semialgebraic.
\end{lemma}
\begin{proof}
\proofstep{Semialgebraicity and boundedness.}
We write $\mathsf H_e(\mathcal B)$ as a finite union of images of coincidence sets.
For a list $(B_1,\ldots,B_k)$ of cells at one face, consider the map
\[
 (w,\lambda)\mapsto\Bigl(g(w_1),\sum_a\lambda_ah(w_a)\Bigr)
\]
on $P(B_1,\ldots,B_k)\times\Delta^{k-1}$, where $\Delta^{k-1}$ is the simplex of convex
coefficients. If $\sum_ai_{B_a}\le e$, its image lies in $\mathsf H_e(\mathcal B)$. Conversely,
by Lemma~\ref{lem:fiber-weight} every representation of a point of $\mathsf H_e(\mathcal B)$ has
$k\le n$ points, so listing its points shows that the point lies in such an image with
$k\le n$. The family is finite, so there are finitely many such lists. Thus
$\mathsf H_e(\mathcal B)$ is a finite union of images of semialgebraic sets under semialgebraic
maps. It is consequently semialgebraic, by the Tarski--Seidenberg theorem. It is bounded:
$g$ takes values in $Q$, and $h$ is bounded on each of the finitely many compact balls.

\proofstep{Closedness.}
We pass to the limit in representations, and use (B1) when a limit reaches the boundary of
a ball.
Take a convergent sequence
$(z^{(\nu)},x^{(\nu)})\to(z,x)$ of points of $\mathsf H_e(\mathcal B)$. Pass to
a subsequence with a fixed cell list $(B_1,\ldots,B_k)$. By compactness of the balls
and of the simplex, pass to a further subsequence with convergent coefficients
$\lambda_a^{(\nu)}\to\lambda_a$ and convergent ball parameters
$w_a^{(\nu)}\to b_a\in D^{f_{B_a}}$. By continuity of $g_{B_a}$ and $h_{B_a}$ on the closed
balls,
\[
 g_{B_a}(b_a)=z\quad(1\le a\le k),\qquad
 x=\sum_a\lambda_ah_{B_a}(b_a),\qquad(\lambda_a)\in\Delta^{k-1}.
\]
For each $a$ we now define a nonempty finite set $A_a$ of sheet points with base $z$, with
$\operatorname{wt}(A_a)\le i_{B_a}$, and with $h_{B_a}(b_a)\in\operatorname{conv}h(A_a)$.
At an interior limit, retain the corresponding sheet point:
$A_a=\{(B_a,b_a)\}$.
At a boundary limit, (B1) gives
$(g_{B_a},h_{B_a})(b_a)\in\mathsf H_{i_{B_a}}(\mathcal B^{<R(B_a)})_{S_{B_a}}$, which expresses
this value using a set $A_a$ of smaller-rank sheet points with
the same limiting base and weight at most $i_{B_a}$.

The union $A^*=\bigcup_aA_a$ is a nonempty finite set of sheet points with base $z$. Its
weight is at most the sum of these bounds,
since duplicate points are counted only once:
\[
 \operatorname{wt}(A^*)\le\sum_a\operatorname{wt}(A_a)\le\sum_ai_{B_a}\le e.
\]
Substituting the convex
representations gives the limiting vector $x$ in $\operatorname{conv}h(A^*)$, as
each $\operatorname{conv}h(A_a)$ is contained in the convex set $\operatorname{conv}h(A^*)$.
So $A^*$ is a representation of $(z,x)$, which is therefore still in $\mathsf H_e(\mathcal B)$.
This proves closedness and compactness.
\end{proof}

Finally, points of hull sets inherit the properties of sheet points. For a point
represented by several sheet points, this is where the weighted hull lemma enters the
construction.
\begin{lemma}[Hull points lie in the cones]\label{lem:hull-level}
Let $\mathcal B$ be admissible, $1\le e\le L$, and
$(z,x)\in \mathsf H_e(\mathcal B)_S$. Then
\[
 x\in\mathcal C_{e,S}\cap\mathcal S,\qquad
 \|z-\phi(x)\|_\infty\le2\eta,\qquad z\notin N[C].
\]
If a representation $x\in\operatorname{conv}h(A')$ uses points of ranks
at most $r$ and common base $z$, then
$1-r\beta-n\theta\le\|x\|_1\le1+r\beta$.
\end{lemma}
\begin{proof}
Let $A'$ be a representation of $(z,x)$ with $\operatorname{wt}(A')\le e$.
The common base belongs to the relative interior of a unique face
$Q_{S'}$, with $S\subseteq S'$ by~\eqref{eq:cube-face-facts}, and every point in the
representation comes from a cell at $S'$ by~\eqref{eq:sheet-face}.

\proofstep{Clamp and norm bounds.}
Both bounds pass from the sheet points to their convex combinations.
Condition (B3) gives $\|z-\phi(h(w))\|_\infty\le2\eta$ for $w\in A'$, and
Lemma~\ref{lem:close-hulls} gives the clamp bound for $x$.
If all points of $A'$ have ranks at most $r$, then (B5) gives
$1-r\beta\le\|h(w)\|_1\le1+r\beta$ for $w\in A'$, and Lemma~\ref{lem:close-hulls} gives
$1-r\beta-n\theta\le\|x\|_1\le1+r\beta$. This is the norm bound of the statement.
All ranks are at most $L(n+1)^2$. So taking $r=L(n+1)^2$ and
using~\eqref{eq:norm-budget} gives $\|x\|_1\in(0.7,1.3)$ and
$\|h(w)\|_1\in(0.7,1.3)$ for $w\in A'$. Thus $x\in\mathcal S$, and all $h(w)\in\mathcal S$.

\proofstep{Cone membership.}
One sheet point lies in the cone at its own stage by (B3), and several sheet points are
handled by the weighted hull lemma.
For one sheet point, $A'=\{w\}$, $x=h(w)$, and $i(w)=\operatorname{wt}(A')\le e$. Then
(B3), Proposition~\ref{prop:cones}(i), and~\eqref{eq:cone-monotone} give
$x\in\mathcal C_{i(w),S'}\subseteq\mathcal C_{i(w),S}\subseteq\mathcal C_{e,S}$, the last
inclusion being an equality when $i(w)=e$.
For at least two sheet points, apply Lemma~\ref{lem:weighted-hull} at the face $S'$: the vectors
$h(w)$ lie in $\mathcal C_{i(w),S'}\cap\mathcal S$, their clamps are within $2\eta$ of
$z$, the total weight satisfies $\operatorname{wt}(A')\le e\le L$, and $\|x\|_1>1/2$.
This gives $x\in\mathcal C_{\operatorname{wt}(A'),S'}$, and the same inclusions give
$x\in\mathcal C_{e,S}$.

Finally, Lemma~\ref{lem:clamp}, with $y=z$, proves $z\notin N[C]$.
\end{proof}

\subsection{Realizing one attachment}\label{app:one-cell}
The construction grows by one cell at a time. The next lemma shows that every semialgebraic
attaching map into the hull set of the cells of smaller rank over a face, in a dimension
allowed by the stage, is the
boundary of a cell that keeps the family admissible, and that the hull set over the face
changes exactly by the corresponding attachment. This allows the induction of
Section~\ref{app:stage-induction} to treat the addition of cells as an abstract attachment.
\begin{lemma}[Realizing one attachment]\label{lem:one-cell}
Let $\mathcal B$ be admissible with all stages at most $e$, where
$1\le e\le L$. Fix $S$, $1\le|S|\le a$, and
$0\le f\le f_e(S)$. Let $r$ be the rank of a stage-$e$, dimension-$f$
cell at $S$. Given a continuous semialgebraic map
\[
 \alpha:S^{f-1}\to \mathsf H_e(\mathcal B^{<r})_S,
\]
there is such a cell $B$ with boundary map $\alpha$ for which
$\mathcal B\cup\{B\}$ is admissible. Moreover, writing
$X=\mathsf H_e(\mathcal B)_S$, the induced map is a homeomorphism
\[
 X\cup_\alpha D^f\ \longrightarrow\ \mathsf H_e(\mathcal B\cup\{B\})_S.
\]
For $S''\not\subseteq S$, the set $\mathsf H_e(\mathcal B)_{S''}$ is unchanged, that is,
$\mathsf H_e(\mathcal B\cup\{B\})_{S''}=\mathsf H_e(\mathcal B)_{S''}$.
\end{lemma}
\begin{proof}
The proof fills the vector coordinate using the connectivity of the cone, and then fills
the base coordinate near the clamped vector. Two generic perturbations, first of the base
and then of the vector, give general position. We then verify admissibility and identify
the new hull set.
Write $\mu=m_S$, $\mathcal C=\mathcal C_{e,S}$, $D=D^f$. For $f\ge1$ write
$\alpha=(z_0,x_0)$, with $z_0:S^{f-1}\to Q_S$ and $x_0:S^{f-1}\to\mathbb R^n$. For $f=0$
the sphere $S^{-1}$ is empty and $\alpha$ is the empty map. We use throughout that
$f\le f_e(S)=\mu-\kappa_e$, and that $1\le r\le L(n+1)^2$.

\proofstep{Fill the vector coordinate.}
We extend the vector part of $\alpha$ to a semialgebraic map $h_*:D\to\mathcal C\cap\mathcal S$
with controlled $\ell_1$ norm.
Let $f\ge1$. By~\eqref{eq:hull-monotone}, $\alpha$ takes values in $\mathsf H_e(\mathcal B)_S$,
and each value has a representation by sheet points of ranks at most $r-1$. Thus
Lemma~\ref{lem:hull-level} gives, for every $\vartheta\in S^{f-1}$,
\begin{equation}\label{eq:cell-boundary-data}
 x_0(\vartheta)\in\mathcal C\cap\mathcal S,\qquad
 \|z_0(\vartheta)-\phi(x_0(\vartheta))\|_\infty\le2\eta,\qquad
 1-(r-1)\beta-n\theta\le\|x_0(\vartheta)\|_1\le1+(r-1)\beta.
\end{equation}
The inequality
$f-1\le n-|S|-ec'-4\le n-3-ec'-|S|$ and
Proposition~\ref{prop:cones}(ii) give a continuous extension
$V_0:D\to\mathcal C$ of $x_0$: the map $x_0$ is null-homotopic in $\mathcal C$, which for
$f=1$ means that its two values lie in one path component.
Let $b_-$ and $b_+$ be the minimum and maximum of $\|x_0\|_1$. Then $b_->0$, because
$\mathcal C$ does not contain zero, and $[b_-,b_+]$ lies in the interval of
norms in~\eqref{eq:cell-boundary-data}.
Radially rescale the extension so that
$\|V_0\|_1\in[b_-,b_+]$, leaving its boundary fixed: replace $V_0(w)$ by
$\max\{b_-,\min\{b_+,\|V_0(w)\|_1\}\}\,V_0(w)/\|V_0(w)\|_1$. The cone property
permits this, since $\mathcal C$ is invariant under positive scalars. For $f=0$, the
connectivity
$n-3-ec'-|S|=f_e(S)\ge0$ shows that the cone is nonempty, and we take for $V_0$ any vector
in it, normalized to $\ell_1$-norm one.

The map $V_0$ is only continuous, and we now replace it by a semialgebraic filling.
Since $V_0(D)$ is compact and $\mathcal C$ is open, we can choose $0<\varpi<\theta$ so
that the closed $\varpi$-neighborhood of $V_0(D)$ in the sup norm lies in $\mathcal C$.
By the Weierstrass approximation theorem, choose a polynomial map
$V_1:\mathbb R^f\to\mathbb R^n$ with $\|V_1-V_0\|_\infty\le\varpi$ on $D$.
For $f\ge1$, define a semialgebraic filling by
\[
 h_*(w)=
 \begin{cases}
 x_0(w/|w|),&1/2\le |w|\le1,\\
 (4|w|-1)x_0(w/|w|)+(2-4|w|)V_1(w/|w|),&1/4\le|w|\le1/2,\\
 V_1(4w),&|w|\le1/4.
 \end{cases}
\]
These formulas agree on the seams: at $|w|=1/2$ the middle formula has coefficients
$(1,0)$; at $|w|=1/4$ it has coefficients $(0,1)$, and $4w=w/|w|$. So $h_*$ is continuous.
Each value is within $\varpi$ of a value of $V_0$. For $|w|\ge1/2$ it is the value
$x_0(w/|w|)=V_0(w/|w|)$. For $1/4\le|w|\le1/2$, with $\vartheta=w/|w|$, it is
$V_0(\vartheta)+(2-4|w|)(V_1(\vartheta)-V_0(\vartheta))$. For $|w|\le1/4$ it is within
$\varpi$ of $V_0(4w)$. So the image lies in $\mathcal C$.
For $f=0$ use $h_*=V_1$. The polynomial approximation serves only to make the
filling semialgebraic, so its degree has no restriction. The degree bound $n$ applies only
to the perturbations $\Pi$ and $R$ below.

Since $\|v\|_1\le n\|v\|_\infty$, the same comparison and
$\|V_0\|_1\in[b_-,b_+]$ give, for $f\ge1$ and $w\in D$,
\begin{equation}\label{eq:cell-hstar-norm}
 1-(r-1)\beta-n\theta-n\varpi\le\|h_*(w)\|_1\le1+(r-1)\beta+n\varpi,
\end{equation}
and for $f=0$ they give $|\|h_*\|_1-1|\le n\varpi$. Since $n\varpi<n\theta$ and
$\beta=3n\theta$, in both cases $\|h_*(w)\|_1$ differs from one by less than $r\beta$. By
the norm estimate~\eqref{eq:norm-budget} with $r$ in place of $R(B)$, it lies in $(0.7,1.3)$. Thus
$h_*(D)\subseteq\mathcal C\cap\mathcal S$.

\proofstep{Fill the base coordinate and put it in general position.}
The base coordinate interpolates between the boundary values $z_0$ and a shifted clamp of
$h_*$. The shift moves it away from the boundary of $Q_S$, so that a small generic
perturbation keeps the interior of the ball over the interior of the face.
Let $o_S$ be the center of $Q_S$, with coordinates $s_v$ at $j_v$ for $v\in S$ and zero
elsewhere, and put
$c_S(x)=(1-\eta/4)\phi(x)+(\eta/4)o_S$.
For $x\in\mathcal C\cap\mathcal S$, Lemma~\ref{lem:clamp} gives $\phi(x)\in Q_S$. So
$c_S(x)$ lies in $Q_S$, with every free coordinate
at distance at least $\eta/4$ from its boundary. The difference
$c_S(x)-\phi(x)=(\eta/4)(o_S-\phi(x))$ vanishes on the prescribed coordinates and has
entries of absolute value at most $\eta/4$ on the free ones. Thus
\begin{equation}\label{eq:cell-center-shift}
 \|c_S(x)-\phi(x)\|_\infty\le\eta/4\qquad(x\in\mathcal C\cap\mathcal S).
\end{equation}
Set
\[
 g_*(w)=(1-t(w))z_0(w/|w|)+t(w)c_S(h_*(w)).
\]
For $|w|\le1/2$ the first term is defined to be zero, consistently with $t=1$ there, and
for $f=0$ use $g_*=c_S(h_*)$. By convexity of $Q_S$, this is a continuous semialgebraic
map $D\to Q_S$, and $g_*=z_0$ on $\partial D$.

Let $V_S=\{y\in\mathbb R^n:y_{j_v}=0\ (v\in S)\}$ be the direction space of $Q_S$; we
identify it with $\mathbb R^\mu$ through the free coordinates.
We call a list of cells of $\mathcal B$ at $S$ with $\sum_a\kappa_{i_{B_a}}\le\mu$ an older
list. There are finitely many older lists because each $\kappa_i\ge3$, which gives
$k\le\mu/3$, and $\mathcal B$ is finite.
Apply Lemma~\ref{lem:base-generic}, in this direction space, to $g_*-o_S$ and to
the coincidence sets $P(B_1,\ldots,B_k)$ of all older lists,
using their common base minus $o_S$ as $v_b$. Choose a generic polynomial $\Pi\in\mathcal P(V_S)$ with
$\|\Pi\|_\infty<\eta/8$ on $D$, and put $g=g_*+t\Pi$.
With this choice, the conclusions of Lemma~\ref{lem:base-generic} hold for $g-o_S$. Thus,
for every one of these older lists and every $1\le s\le n+1$,
the set of tuples $(w_1',\ldots,w_s',(w_a)_a)$ with pairwise distinct
$w_b'\in\operatorname{int}D$, with $(w_a)_a\in P(B_1,\ldots,B_k)$, and with
$g(w_1')=\cdots=g(w_s')=g_{B_1}(w_1)$ satisfies
\begin{equation}\label{eq:cell-generic-old}
 \dim\{(w_1',\ldots,w_s',(w_a)_a)\}\le s(f-\mu)+\dim P(B_1,\ldots,B_k),
\end{equation}
and for $2\le s\le n+1$ the set of pairwise distinct $w_1',\ldots,w_s'\in\operatorname{int}D$
with equal values of $g$ satisfies
\begin{equation}\label{eq:cell-generic-self}
 \dim\{(w_1',\ldots,w_s')\}\le sf-(s-1)\mu.
\end{equation}

We next check that $g$ maps the interior of the ball to the interior of the face.
For $t>0$, $g$ is a convex combination of $z_0\in Q_S$ and
$c_S(h_*)+\Pi$. The latter point lies in $\operatorname{int}Q_S$: since $\Pi$ takes values in
$V_S$, its prescribed coordinates are those of $c_S(h_*)$, and its free coordinates have
absolute value at most $1-\eta/4+\eta/8<1$. So the base $g(w)$ is in
$\operatorname{int}Q_S$ for $w\in\operatorname{int}D$, where $t(w)>0$.
On the boundary it equals $z_0$, as $t=0$ there.

\proofstep{Separate the vector coordinate.}
With $g$ fixed, the loci where the new cell meets itself or the older cells over one cube
point have dimension below $n$. A generic perturbation of the vector coordinate therefore
separates the new vectors from each other and from the older convex hulls.
The self-coincidence set
$\mathcal M'=\{(w,w')\in(\operatorname{int}D)^2:w\ne w',g(w)=g(w')\}$ has
dimension at most $2f-\mu\le\mu-2\kappa_e<n$, by~\eqref{eq:cell-generic-self} with $s=2$.
For an older list, consider the semialgebraic set
\[
 \mathcal M=\{(w,(w_a),\lambda):w\in\operatorname{int}D,\ g(w)=g_{B_1}(w_1),\
 (w_a)\in P(B_1,\ldots,B_k),\ \lambda\in\Delta^{k-1}\},
\]
with the maps $(w,(w_a),\lambda)\mapsto w$ and
$\xi(w,(w_a),\lambda)=\sum_a\lambda_ah_{B_a}(w_a)$.
By~\eqref{eq:cell-generic-old} with $s=1$, by (B2) for $\mathcal B$, and by
$f-\mu\le-\kappa_e$ and $\sum_a\kappa_{i_a}\ge3k$ (with $i_a=i_{B_a}$), its dimension is at most
\[
 (f-\mu)+\Bigl(\mu-\sum_a\kappa_{i_a}\Bigr)+k-1
 \le\mu-\kappa_e-2k-1<n.
\]
Lists with $\sum_a\kappa_{i_a}>\mu$ have empty coincidence sets by (B2), so they need no
treatment.
Lemma~\ref{lem:height-generic}, applied to $h_*$, to $\mathcal M'$, and to the finitely many
sets $\mathcal M$, therefore gives a generic polynomial $R\in\mathcal P(\mathbb R^n)$
that separates self-coincident points and avoids every older fiberwise
convex hull, in the sense that $h=h_*+tR$ satisfies
\begin{equation}\label{eq:cell-height-separation}
 h(w)\ne h(w')\ \ \text{on }\mathcal M',\qquad
 h(w)\ne\sum_a\lambda_ah_{B_a}(w_a)\ \ \text{on each }\mathcal M.
\end{equation}
Since $h_*(D)$ is a compact subset of the open cone $\mathcal C$, density of generic
polynomials lets us choose $R$ with $\|R\|_\infty\le\eta e_0/8$ and small
enough that $h=h_*+tR$ still lies in $\mathcal C$.
The pair $(g,h)$ has the prescribed boundary values $\alpha$, because $t=0$ on $\partial D$.

Let $B$ be the cell with $f_B=f$, $S_B=S$, $i_B=e$, $g_B=g$, $h_B=h$, created after all cells
of $\mathcal B$. By construction $g$ and $h$ are continuous and semialgebraic,
$g(\operatorname{int}D)\subseteq\operatorname{int}Q_S$, $f\le f_e(S)$, and $R(B)=r$.
Put $\mathcal B'=\mathcal B\cup\{B\}$.

\proofstep{Admissibility of the old cells.}
All old cells retain their conditions. Hull sets are monotone in the
family by~\eqref{eq:hull-monotone}, and $\mathcal B^{<R(B')}\subseteq(\mathcal B')^{<R(B')}$
for each $B'\in\mathcal B$. So their boundary inclusions in (B1) persist even if the new
cell has smaller rank. Lists of old cells retain (B2), since their coincidence sets
do not involve $B$.
Conditions (B3) and (B5) concern one cell, and (B4) refers
to cells created earlier than that cell, which are unchanged since $B$ is the youngest.

\proofstep{Admissibility of the new cell.}
We check the five conditions in turn.
For (B1), since $R(B)=r$, we have $(\mathcal B')^{<r}=\mathcal B^{<r}$, so (B1) is the
hypothesis on $\alpha$.

To verify (B2), take a list containing $s\ge1$ occurrences of $B$. All its cells are at
$S$. Reorder it as these occurrences followed by
$k\ge0$ older cells $B_1,\ldots,B_k$, with stages $i_a$. We must bound the dimension of
$P(B,\ldots,B,B_1,\ldots,B_k)$ by
$\mu-s\kappa_e-\sum_a\kappa_{i_a}$, and show that this set is empty when the bound is negative.
\begin{enumerate}[label=(\alph*),leftmargin=2em]
\item Let $s\le n+1$, $k\ge1$, and $\sum_a\kappa_{i_a}\le\mu$.
By~\eqref{eq:cell-generic-old}, by (B2) for $\mathcal B$, and by $f-\mu\le-\kappa_e$,
\[
 \dim P(B,\ldots,B,B_1,\ldots,B_k)\le s(f-\mu)+\mu-\sum_a\kappa_{i_a}
 \le\mu-s\kappa_e-\sum_a\kappa_{i_a},
\]
and a negative bound means that the set is empty.
\item Let $k\ge1$ and $\sum_a\kappa_{i_a}>\mu$. The list $B_1,\ldots,B_k$ has negative bound, so
$P(B_1,\ldots,B_k)$ is empty by (B2) for $\mathcal B$. The new set projects into it, so
the new list is empty too, as required by its negative bound.
\item Let $k=0$ and $2\le s\le n+1$. Then~\eqref{eq:cell-generic-self} gives
$\dim P(B,\ldots,B)\le sf-(s-1)\mu=s(f-\mu)+\mu\le\mu-s\kappa_e$, and again a negative
bound means that the set is empty.
\item Let $k=0$ and $s=1$. Then $P(B)=\operatorname{int}D$ has dimension
$f\le\mu-\kappa_e$, and this bound is nonnegative.
\item Let $s>n+1$. A sublist with $n+1$ occurrences of $B$ already has
negative bound $\mu-(n+1)\kappa_e$, so by (c) its set is empty. The set of the given list
projects into it, which again forces emptiness, and the required bound is negative.
\end{enumerate}

For (B3), cone membership was arranged by the choice of $R$.
For the distance from the base to the clamped vector, let $w\in\operatorname{int}D$, and write
$x=h_*(w)$ and $t=t(w)$. Then
$x\in\mathcal C\cap\mathcal S$ and
\begin{align*}
 g(w)-\phi(h(w))
 ={}&(1-t)\bigl(z_0(w/|w|)-\phi(x)\bigr)
   +t\bigl(c_S(x)+\Pi(w)-\phi(x)\bigr)\\
 &+\bigl(\phi(x)-\phi(x+tR(w))\bigr).
\end{align*}
On the collar $|w|\ge1/2$, we have $x=x_0(w/|w|)$, so the first term has sup norm at most
$(1-t)2\eta$ by~\eqref{eq:cell-boundary-data}. For $|w|\le1/2$, and for $f=0$, we have $t=1$,
so the first term disappears. The second term is bounded by $t(\eta/4+\eta/8)$, by~\eqref{eq:cell-center-shift}
and $\|\Pi\|_\infty<\eta/8$. The third is bounded by $t\|R\|_\infty/e_0\le t\eta/8$,
by~\eqref{eq:clamp-lipschitz}. Therefore
\[
 \|g(w)-\phi(h(w))\|_\infty
 \le(1-t)2\eta+t(\eta/4+\eta/8)+t\eta/8\le2\eta.
\]

Condition (B4) is exactly the separation ensured by $R$. If
$w\ne w'$ in $\operatorname{int}D$ and $g(w)=g(w')$, then $(w,w')\in\mathcal M'$ and
$h(w)\ne h(w')$ by~\eqref{eq:cell-height-separation}. Next, let $w\in\operatorname{int}D$, and let
$A'$ be a nonempty finite set of sheet points of older cells with base $g(w)$. Since
$g(w)\in\operatorname{int}Q_S$, all its points belong to cells at $S$
by~\eqref{eq:sheet-face}. Listing them gives a nonempty set $P(B_1,\ldots,B_k)$, so
$\sum_a\kappa_{i_{B_a}}\le\mu$ by (B2) for $\mathcal B$, and this list is an older list.
If we had $h(w)=\sum_a\lambda_ah_{B_a}(w_a)$ with $\lambda\in\Delta^{k-1}$, the point
$(w,(w_a),\lambda)$ would contradict~\eqref{eq:cell-height-separation}.

For (B5), since $\|tR\|_\infty\le\eta e_0/8=\theta/32$, passing from $h_*$ to $h$ changes the
$\ell_1$ norm by at most $n\theta/32$. When $f\ge1$, inequality~\eqref{eq:cell-hstar-norm} collects the
preceding changes of norm. The boundary vectors have representations of ranks at most $r-1$, so
that their norms lie in the interval of~\eqref{eq:cell-boundary-data}. The radial extension
stays in this interval, and passing to $h_*$ changes the norm by at most $n\varpi<n\theta$. Therefore
\[
 \|h\|_1\in
 [1-(r-1)\beta-(2+1/32)n\theta,
  1+(r-1)\beta+(1+1/32)n\theta]
 \subseteq[1-r\beta,1+r\beta],
\]
because $\beta=3n\theta$.
For $f=0$, the starting norm is one and the two changes amount to less than
$2n\theta<\beta\le r\beta$. The new cell therefore has all five properties.

\proofstep{The new hull set is the attachment.}
A new sheet point has the full weight $e$, so it can occur in a representation only by
itself. This gives the new hull set.
Let $(z,x)\in \mathsf H_e(\mathcal B')_S$ with a representation $A'$. If $A'$ contains no sheet point
of $B$, then $(z,x)\in X$. If $A'$ includes a new sheet point, of weight $e$, then, since all
weights are at least one and $\operatorname{wt}(A')\le e$, it contains only that point.
Conversely, $\{(B,w)\}$ has weight $e$ and base in $Q_S$, and $X\subseteq \mathsf H_e(\mathcal B')_S$
by~\eqref{eq:hull-monotone}. Therefore the new set is
\[
 \mathsf H_e(\mathcal B')_S=X\cup(g,h)(\operatorname{int}D).
\]
Let $\Phi:X\cup_\alpha D\to \mathsf H_e(\mathcal B')_S$ be the identity on $X$ and $(g,h)$ on $D$. It
is well defined and continuous, because $(g,h)=\alpha$ on $\partial D$ and $\alpha$ takes
values in $\mathsf H_e(\mathcal B^{<r})_S\subseteq X$. It is surjective by the last display and
$(g,h)(\partial D)\subseteq X$.
Condition (B4) makes it injective,
both on the new interior and relative to $X$: a point of $X$ over $g(w)$, with
$w\in\operatorname{int}D$, has its vector in $\operatorname{conv}h(A')$ for a set $A'$ of
sheet points of $\mathcal B$ with base $g(w)$, and all cells of $\mathcal B$ are older than
$B$.
By Lemma~\ref{lem:hull-compact}, $X$ is compact, so $X\cup_\alpha D$ is compact.
Thus $\Phi$ is a continuous bijection from a compact space onto a Hausdorff
space and is hence a homeomorphism.

Finally, if $S''\not\subseteq S$, then
$Q_{S''}\cap\operatorname{int}Q_S=\varnothing$ by~\eqref{eq:cube-face-facts}; no new interior point
lies over that face. So no representation of a point of $\mathsf H_e(\mathcal B')_{S''}$ contains a
new sheet point, and its weighted hull set is unchanged.
\end{proof}

\subsection{The stage induction}\label{app:stage-induction}
We now build the families stage by stage. The invariant of the induction is the vanishing
of the reduced homology of the hull set over each face, in the range of degrees required
by Theorem~\ref{thm:carrier}. For families $\mathcal B_\ell$, $0\le\ell\le L$, put
$W^\ell=\mathsf H_\ell(\mathcal B_\ell)$ for $\ell\ge1$,
and $W^0=\varnothing$, and write $W^\ell_S=W^\ell\cap(Q_S\times\mathbb R^n)$.
\begin{proposition}[The stage induction]\label{prop:stage-induction}
Starting with $\mathcal B_0=\varnothing$, there are admissible finite
families $\mathcal B_\ell$ for $0\le\ell\le L$, with
$\mathcal B_\ell\subseteq\mathcal B_{\ell+1}$ for $\ell<L$ and with cells of stage at
most $\ell$ in $\mathcal B_\ell$, such that
\begin{equation}\label{eq:stage-invariant}
 \widetilde H_j(W^\ell_S)=0\qquad
 (j\le r_S,\ 1\le|S|\le a),\qquad \ell\ge1.
\end{equation}
Moreover, $\widetilde H_j(W^\ell_S)=0$ for all $j$, $\ell\ge1$, and $1\le|S|\le a$.
\end{proposition}
As in Section~\ref{app:carrier}, homology is singular with integer coefficients unless
another coefficient group is displayed, and $\widetilde H_{-1}(T)=0$ means $T\ne\varnothing$.
For $|S|\le a$ we have $r_S\ge-1$, so~\eqref{eq:stage-invariant} includes
$W^\ell_S\ne\varnothing$. The induction itself uses only~\eqref{eq:stage-invariant}, which is
also the range required by Theorem~\ref{thm:carrier}. The last assertion of the proposition,
which follows from the same construction, shows that the hull sets over the faces are acyclic.

Stage $e=\ell+1$ works as follows. The faces are processed in descending order of $|S|$, so that
the hull sets over the boundary of a face are complete before the face itself is processed.
Over a face $S$, the current hull set differs from the hull set of the preceding stage only by a
set of dimension at most an integer $z(S)$, defined in~\eqref{eq:z-identity}. Therefore its reduced homology
through degree $r_S$ is concentrated in degrees at most $z(S)$, and it is free in degree
$z(S)$. Lemma~\ref{lem:killing} removes this homology with cells of dimension at most
$z(S)+1=f_e(S)$, and Lemma~\ref{lem:one-cell} realizes these cells.
\begin{proof}
\proofstep{Setting up a stage.}
We construct the families by induction on $\ell$, and we first fix the notation for one stage.
The empty family is admissible, since all conditions are vacuous.
Fix $0\le\ell<L$. Suppose that $\mathcal B_\ell$ has been constructed, and
that~\eqref{eq:stage-invariant} holds at $\ell$ when $\ell\ge1$.
At stage $e=\ell+1$, process the faces with $1\le|S|\le a$ in descending order of $|S|$, with
an arbitrary order among faces of equal size. The current family $\mathcal B$ starts as $\mathcal B_\ell$ and
grows by stage-$e$ cells added through Lemma~\ref{lem:one-cell}, so it remains admissible
and finite, and its stages remain at most $e$. A set of sheet points of weight at most
$\ell$ contains no sheet point of a stage-$e$ cell. Hence, during the whole stage,
\begin{equation}\label{eq:stage-old-level}
 \mathsf H_\ell(\mathcal B)=\mathsf H_\ell(\mathcal B_\ell)=W^\ell\qquad(\ell\ge1).
\end{equation}
When the processing of a face $S$ begins, let
$X_S=\mathsf H_e(\mathcal B)_S$, where $\mathcal B$ is the family at that moment.
By~\eqref{eq:hull-monotone} and~\eqref{eq:stage-old-level}, $W^\ell_S\subseteq X_S$.
No new stage-$e$ cell has yet been placed at $S$.
Set
\begin{equation}\label{eq:z-identity}
 z(S)=m_S-ec'-4=r_S-\ell c'\le r_S.
\end{equation}
The equality holds because
$r_S=a-|S|-1=n-c'-3-|S|-1=m_S-c'-4$. Also $z(S)=f_e(S)-1=z_e(S)$.

\proofstep{The new part has small dimension.}
We show that $X_S$ differs from $W^\ell_S$ only by a set of dimension at most $z(S)$.
Let $(z,x)\in X_S\setminus W^\ell_S$, with a representation $A'$. If
$\operatorname{wt}(A')\le\ell$, then $\ell\ge1$ and all stages in $A'$ are at most $\ell$,
so that $A'$ consists of sheet points of $\mathcal B_\ell$. This would put $(z,x)$ in
$W^\ell_S$. So a point in $X_S\setminus W^\ell_S$ has a representation of weight
exactly $e$ over some $\operatorname{int}Q_{S'}$ with $S\subseteq S'$, by~\eqref{eq:cube-face-facts},
and all its points come from cells at $S'$ by~\eqref{eq:sheet-face}.

If the representation is a single point, it comes from a stage-$e$
cell $B$ at $S'$, and $S'\ne S$ since no such cell exists at $S$. Thus $S'$ is a proper
face, $S'\supsetneq S$. All these points lie in the image of $\operatorname{int}D^{f_B}$
under $(g_B,h_B)$, so their dimension is at most
$f_B\le f_e(S')\le f_e(S)-1=z(S)$.

If it uses $k\ge2$ points, with stages $i_1,\ldots,i_k$ summing to $e$, then every
$i_a\le e-1=\ell$, so all are sheet points of cells in $\mathcal B_\ell$. The point
$(z,x)$ lies in the image of $P(B_1,\ldots,B_k)\times\Delta^{k-1}$ under a semialgebraic map,
and (B2) bounds the dimension, including the convex coefficients, by
\[
 m_{S'}-\sum_{a=1}^k(i_ac'+3)+k-1
 =m_{S'}-ec'-2k-1\le z(S)-1.
\]
The last inequality uses $m_{S'}\le m_S$ and $k\ge2$. When $P(B_1,\ldots,B_k)\ne\varnothing$, (B2) gives also
$m_{S'}-ec'-3k\ge0$, so $z(S)\ge m_{S'}-ec'-4\ge3k-4\ge2$. Thus this case does not occur
for $z(S)<2$.
There are finitely many such lists, by Lemma~\ref{lem:fiber-weight} and finiteness of
$\mathcal B$. Hence the semialgebraic set $X_S\setminus W^\ell_S$ satisfies
\begin{equation}\label{eq:new-dimension}
 \dim(X_S\setminus W^\ell_S)\le z(S),
\end{equation}
and it is empty when $z(S)<0$: then the first case would give $0\le f_B\le z(S)<0$, and the
second case does not occur.
For $\ell=0$, only the first case occurs, as $e=1$ is not a sum of two positive integers.

\proofstep{Homology of the current set.}
The dimension bound confines the homology of $X_S$ through degree $r_S$ to degrees at most
$z(S)$, and it makes the group in degree $z(S)$ free.
By Lemma~\ref{lem:hull-compact}, $(X_S,W^\ell_S)$ is a compact semialgebraic pair, so it is
triangulable \cite[Section 9.2]{BCR98}: there are a finite simplicial complex $K$ with a
subcomplex $K_0$ and a semialgebraic homeomorphism $\Theta:|K|\to X_S$ with
$\Theta(|K_0|)=W^\ell_S$. For a simplex $\tau$ of $K$ not in $K_0$, the open simplex of $\tau$
is disjoint from $|K_0|$, so $\Theta$ maps it into $X_S\setminus W^\ell_S$. Since $\Theta$
preserves dimension, \eqref{eq:new-dimension} gives $\dim\tau\le z(S)$.
So the relative chains of $(K,K_0)$ vanish above degree $z(S)$. Singular relative
homology is computed by these chains \cite[Theorem 2.27]{Hatcher02}. Thus
\begin{equation}\label{eq:stage-relative-top}
 H_j(X_S,W^\ell_S)=0\quad(j>z(S)),
\end{equation}
and the top relative homology $H_{z(S)}(X_S,W^\ell_S)$ is free, being the group of relative
cycles, a subgroup of the top free chain group.

For $\ell\ge1$, the set $W^\ell_S$ is nonempty by~\eqref{eq:stage-invariant} at $\ell$, and the
reduced exact sequence
\[
 \widetilde H_j(W^\ell_S)\to\widetilde H_j(X_S)\to H_j(X_S,W^\ell_S)
\]
gives, using~\eqref{eq:stage-invariant} at $\ell$ on the left and~\eqref{eq:stage-relative-top} on
the right,
\[
 \widetilde H_j(X_S)=0\quad(z(S)<j\le r_S).
\]
When $z(S)\ge1$, the left term vanishes at $j=z(S)$, as $z(S)\le r_S$. Therefore
$H_{z(S)}(X_S)$ embeds in the free top relative
homology group, and is therefore free. For $\ell=0$, these conclusions
follow directly from $\dim X_S\le z(S)=r_S$: in this case $K_0$ is empty, the range
$z(S)<j\le r_S$ is empty, and $H_{z(S)}(X_S)$ is the group of top-dimensional cycles.

\proofstep{Attaching cells over $S$.}
We now remove the homology in degrees at most $z(S)$, starting with the cases where
$z(S)<0$, in which at most one cell is needed.
If $z(S)\le-2$, then $\ell\ge1$, because $\ell=0$ gives $z(S)=r_S\ge-1$. Then
$X_S=W^\ell_S$ by~\eqref{eq:new-dimension}, and $f_e(S)=z(S)+1<0$, so
no new cell is needed or permitted. If $z(S)=-1$ and $\ell\ge1$, again
$X_S=W^\ell_S$, and no new cell is needed. In both cases the set over $S$ remains
$W^\ell_S$, and its reduced homology vanishes through degree $r_S$ by~\eqref{eq:stage-invariant}
at $\ell$.

For $\ell=0$ and $z(S)=-1$, we have $r_S=-1$, so $|S|=a$, and $X_S=\varnothing$: a sheet point
over $Q_S$ would belong to a cell at some $S'\supseteq S$, while no cell has $|S'|>a$
and no cell at $S$ exists yet.
Lemma~\ref{lem:one-cell} with $f=0=f_1(S)$ adds one $0$-cell. The set over $S$ is then a
single point, which is all that $r_S=-1$ requires.

If $z(S)\ge0$, apply Lemma~\ref{lem:killing} to $Y=X_S$ with parameters
$z(S)$ and $r_S$. Its hypotheses have been verified. The set $Y$ is compact and
semialgebraic, hence homeomorphic to a finite simplicial complex, and $r_S\ge z(S)\ge0$
by~\eqref{eq:z-identity}. The vanishing for $z(S)<j\le r_S$ and the freeness of
$H_{z(S)}(Y)$ for $z(S)\ge1$ were proved in the previous step. Moreover $Y\ne\varnothing$:
for $\ell\ge1$ it contains $W^\ell_S$, and for $\ell=0$ we have $|S|<a$, so it contains the
$0$-cell added at a face $S'\supseteq S$ with $|S'|=a$. So no $0$-cell is added.

We run the recursion on the weighted hull sets themselves: for $i\ge1$, let $\mathcal B^{(i-1)}$ be
the current family when the batch of dimension $i$ starts, and let
$Y_{i-1}=\mathsf H_e(\mathcal B^{(i-1)})_S$, with $\mathcal B^{(0)}$ the family at the start of the
processing of $S$.
Before adding cells of dimension $i$, triangulate the current compact
semialgebraic set $Y_{i-1}$, by Lemma~\ref{lem:hull-compact} and
\cite[Section 9.2]{BCR98}. Simplicial approximation represents each
of the finitely many required homotopy classes by a semialgebraic map
from $S^{i-1}$ into $Y_{i-1}$. To see this, take a simplex $\Delta^i\subset\mathbb R^i$ with
barycenter at the origin, so that radial projection is a semialgebraic homeomorphism
$\partial\Delta^i\to S^{i-1}$. A continuous map from $\partial\Delta^i$ to the triangulating
complex is homotopic to a map that is simplicial for an iterated barycentric subdivision of
$\partial\Delta^i$ \cite[Theorem 2C.1]{Hatcher02}. A simplicial map is piecewise affine,
so its composition with the semialgebraic triangulation map is semialgebraic.
Choose all these maps before attaching
any cell of dimension $i$, as allowed by Lemma~\ref{lem:killing}.
Add them one at a time by Lemma~\ref{lem:one-cell}. Their dimension is
at most $z(S)+1=f_e(S)$. The current family is admissible with stages at most $e$,
$1\le e\le L$, and $1\le|S|\le a$.

\proofstep{The rank hypothesis.}
Lemma~\ref{lem:one-cell} requires each attaching map to land in the hull set of the cells of
rank smaller than that of the new cell. A cell has sheet points over
$Q_S$ precisely when $S\subseteq S_B$.
Let $r$ be the rank of a stage-$e$, dimension-$i$ cell at $S$.
During the dimension-$i$ batch at $S$, a cell with sheet points over $Q_S$ is either at an
earlier stage, at stage $e$ on a proper face of $S$, or at stage $e$
and $S$ with dimension at most $i$, because cells at $S$ are created in increasing dimension.
The first two kinds and the cells of dimension less than $i$ have rank
strictly smaller than $r$, by~\eqref{eq:cell-rank}: the first digit, the digit
$a-|S_B|<a-|S|$, or the last digit is smaller, with the preceding digits equal.
The attaching maps land in $Y_{i-1}$, which contains no dimension-$i$
cell from this batch: every representation of a point of $Y_{i-1}$ consists of sheet points
over $Q_S$ of cells of $\mathcal B^{(i-1)}$, and these cells are of the kinds of rank less
than $r$. Thus $Y_{i-1}\subseteq\mathsf H_e((\mathcal B^{(i-1)})^{<r})_S$. The attaching maps therefore
land in $\mathsf H_e(\mathcal B^{<r})_S$ for the current family $\mathcal B$,
by~\eqref{eq:hull-monotone}, as required. The other cells of $\mathcal B$, including
those of stage $e$ at other faces of size $|S|$, have no sheet points over $Q_S$ and play no
role, whatever their rank.

\proofstep{The additions are the abstract attachments.}
The homeomorphism in Lemma~\ref{lem:one-cell} is the identity on the set before the
attachment, and all attaching maps of the batch land in $Y_{i-1}$. Therefore the set over $Q_S$
after the batch is $Y_{i-1}$ with these $i$-cells attached simultaneously. This identifies every
addition of a cell with the abstract attachment in Lemma~\ref{lem:killing}.
Thus all reduced homology through degree $r_S$ vanishes after this face
is processed.

\proofstep{Completed faces and the end of the stage.}
Subsequent additions do not disturb a completed face.
Any subsequently processed $S''$ has $|S''|\le|S|$ and $S''\ne S$,
so $S\not\subseteq S''$. The last clause of Lemma~\ref{lem:one-cell}, with $S''$ as the face
of the new cell and $S$ as the face of the unchanged set, ensures that these additions leave
the completed set over $S$ unchanged.
There are finitely many faces and finitely many attachments at each
face, so the construction is finite. Let $\mathcal B_e$ be the family at the end of the
stage. For every $S$, the set $W^e_S=\mathsf H_e(\mathcal B_e)_S$ is the set reached when $S$ was
completed, and this proves~\eqref{eq:stage-invariant} at $e$. No cell of stage $L-1$ or $L$ is
ever added, by~\eqref{eq:cell-stage-bound}; these stages fall in the case $z(S)\le-2$.

\proofstep{Acyclicity in every degree.}
It remains to prove the last assertion of the proposition: the construction gives
acyclicity in every degree, $\widetilde H_j(W^\ell_S)=0$ for all $j$,
$\ell\ge1$, and $1\le|S|\le a$. We use induction on $\ell$, with the notation of this
proof.
At the first stage, $\dim X_S\le r_S$, and~\eqref{eq:stage-relative-top} says that
$H_j(X_S)=0$ for $j>z(S)=r_S$. At each stage $e\ge2$, the reduced exact sequence and
acyclicity of the preceding stage give
$\widetilde H_j(X_S)\cong H_j(X_S,W^\ell_S)$ for every $j$, and these groups vanish above
$z(S)$ by~\eqref{eq:stage-relative-top}. In both situations with $z(S)\ge0$, the
conclusion~\eqref{eq:killing-conclusion} of Lemma~\ref{lem:killing} shows that the recursion
removes the homology through degree $z(S)$ without creating any
in higher degrees. The cases $z(S)<0$ are the unchanged space or
the single point described above.
\end{proof}

\subsection{Projection to the cube}
The hull sets of the last stage have the required homology, but they lie in
$Q\times\mathbb R^n$. We now project them to the cube. At the last stage the weight bound
excludes no subset of a sheet fiber, so the fiber of the projection over a cube point is the
convex hull of all vectors over that point. The Vietoris--Begle theorem therefore transfers
the homology to the projected sets, and these are the carriers required by
Theorem~\ref{thm:carrier}.

Put $W=W^L$, $W_S=W^L_S$, and $T_S=\operatorname{pr}_1(W_S)$ for $1\le|S|\le a$, where
$\operatorname{pr}_1:Q\times\mathbb R^n\to Q$ is the projection. Sheet points and
sheet fibers refer to the family $\mathcal B_L$.
\begin{proposition}[The carriers]\label{prop:carriers}
For $1\le|S|\le a$, the set $T_S$ is nonempty, compact, semialgebraic, and contained in
$Q_S\setminus N[C]$. If $S\subseteq S'$, then $T_{S'}\subseteq T_S$. Moreover,
\[
 \widetilde H_j(T_S;\mathbb Z)=0\qquad(j\le a-|S|-1),
\]
and in fact $\widetilde H_j(T_S;\mathbb Z)=0$ for all $j$.
\end{proposition}
\begin{proof}
\proofstep{Fibers.}
We first identify the fibers of the projection $W_S\to T_S$.
Every nonempty subset of a sheet fiber has weight at most $L-1$ by
Lemma~\ref{lem:fiber-weight}. Thus, over every $z\in T_S$, the fiber
of $W_S\to T_S$ is exactly
\[
 \operatorname{conv}h(g^{-1}(z)).
\]
To see this, let $z\in T_S$. Some $(z,x)\in W_S$ has a representation $A'\subseteq g^{-1}(z)$,
so $g^{-1}(z)$ is nonempty, and $x\in\operatorname{conv}h(A')\subseteq\operatorname{conv}h(g^{-1}(z))$.
Conversely, $g^{-1}(z)$ is finite and of weight at most $L-1\le L$, so it is itself a
representation of $(z,x')$ for every $x'\in\operatorname{conv}h(g^{-1}(z))$, and $(z,x')\in W_S$
as $z\in Q_S$.
The fiber is a nonempty compact convex set, in fact a polytope. The formula applies also on
face boundaries: a point of $T_S$ lies in $\operatorname{int}Q_{S'}$ for
exactly one $S'\supseteq S$, and Lemma~\ref{lem:fiber-weight} applies over
that face.

\proofstep{The properties that do not involve homology.}
The sets $T_S$ are compact and semialgebraic, as projections of the compact semialgebraic sets
$W_S$ of Lemma~\ref{lem:hull-compact}. They are nested under face
inclusion: if $S\subseteq S'$, then $Q_{S'}\subseteq Q_S$, so $W_{S'}\subseteq W_S$ and
$T_{S'}\subseteq T_S$. They are contained in $Q_S$, and
disjoint from $N[C]$ by Lemma~\ref{lem:hull-level} with $e=L$.
Since $r_S\ge-1$, \eqref{eq:stage-invariant} at $\ell=L$ gives $W_S\ne\varnothing$, and so
$T_S\ne\varnothing$.

\proofstep{Vietoris--Begle.}
Since the fibers are contractible, the projection induces an isomorphism on cohomology.
Let $p:W_S\to T_S$ be the projection. The Vietoris--Begle mapping theorem
\cite[Chapter 6, Section 9, Theorem 15]{Spanier66} states the following.
A closed continuous surjection between paracompact Hausdorff spaces
whose fibers have vanishing reduced Alexander--Spanier cohomology with coefficients in an
abelian group $\Gamma$ in all degrees below $N$ induces an isomorphism on Alexander--Spanier
cohomology with coefficients in $\Gamma$ in all degrees below $N$.

The theorem applies to the continuous surjection $p$: both spaces are compact
metric spaces, so $p$ is a closed map, and all fibers are acyclic, being contractible
polytopes. So $N$ is arbitrary. It gives an isomorphism on Alexander--Spanier
(equivalently \v Cech) cohomology with every coefficient group, in every degree.
Triangulability identifies these groups naturally with singular cohomology: $W_S$ and $T_S$ are
triangulable by \cite[Section 9.2]{BCR98}, and compact triangulable spaces are paracompact and
locally contractible, so that Alexander--Spanier and singular cohomology are naturally
isomorphic on them \cite[Chapter 6, Section 9]{Spanier66}. Naturality transfers the
statement to the map induced by $p$. Thus $p^*$ is an isomorphism on singular cohomology with
every coefficient group, and, as both spaces are nonempty, also on reduced singular cohomology.

\proofstep{Integer homology.}
Theorem~\ref{thm:carrier} asks for vanishing in integer homology, so we pass from cohomology
with all coefficient groups to integer homology through the mapping cone.
Take the mapping cone $\mathsf C$ of the
projection $p$, the quotient of its mapping cylinder $M_p$ by the copy of $W_S$. This copy has a
collar in $M_p$, so the reduced
groups of $\mathsf C$ are the relative groups of $(M_p,W_S)$, and $M_p$ deformation retracts onto
$T_S$, with the inclusion of $W_S$ corresponding to $p$. For every abelian group $\Gamma$ this
gives the exact sequence
\[
 \widetilde H^{j-1}(T_S;\Gamma)\xrightarrow{\ p^*\ }\widetilde H^{j-1}(W_S;\Gamma)
 \to\widetilde H^j(\mathsf C;\Gamma)
 \to\widetilde H^j(T_S;\Gamma)\xrightarrow{\ p^*\ }\widetilde H^j(W_S;\Gamma).
\]
Therefore the reduced cohomology of $\mathsf C$ vanishes with every
coefficient group. If $\mathsf C$ had a lowest nonzero reduced integral homology group $A$
in degree $j$, the universal coefficient theorem \cite[Theorem 3.2]{Hatcher02} with coefficients
$A$ would give, since $\widetilde H_{j-1}(\mathsf C;\mathbb Z)=0$,
\[
 \widetilde H^j(\mathsf C;A)\cong\operatorname{Hom}(A,A)\ne0,
\]
because $\operatorname{Hom}(A,A)$ contains the identity of $A$. This contradicts the vanishing
of the reduced cohomology of $\mathsf C$. Hence $\widetilde H_*(\mathsf C;\mathbb Z)=0$, and
the homology exact sequence of $(M_p,W_S)$ shows that the projection is
an integral homology isomorphism. By~\eqref{eq:stage-invariant} at $\ell=L$ and
$T_S\ne\varnothing$,
\[
 \widetilde H_j(T_S;\mathbb Z)=0\qquad(j\le a-|S|-1).
\]
The same isomorphism transfers acyclicity in every degree from $W_S$ to $T_S$, by the last
assertion of Proposition~\ref{prop:stage-induction}.
\end{proof}

\subsection{Proof of the logarithmic-product bound}
The carriers give the bound on $\mathrm{LR}(C)$ for $a\ge1$ through
Theorem~\ref{thm:carrier}. Together with the case $a<1$ of Section~\ref{app:parameters} and
the conversion to learning rules of Proposition~\ref{prop:cover-rule}, this completes the
proof of Theorem~\ref{thm:main}, including the derivation of its two constants.
\begin{proof}[Proof of Theorem~\ref{thm:main}]
Fix a nonempty finite class $C\subseteq\{\pm1\}^n$, $n\ge1$, and put $m=|C|$ and
$G=\ln(2m)\ln(2n)$, as in Section~\ref{app:parameters}.

\proofstep{The case $a\ge1$.}
All hypotheses of Theorem~\ref{thm:carrier} hold, by Proposition~\ref{prop:carriers}. Indeed,
$a\ge1$ and
$d=n-a-1=c_*-1=c'+2\ge2$. The sets $T_S\subseteq Q_S\setminus N[C]$ are compact,
semialgebraic, and nested, and their homology vanishes through degree $a-|S|-1$, including
nonemptiness in degree $-1$. Theorem~\ref{thm:carrier} therefore gives
\[
 \mathrm{LR}(C)\le n-a=c_*=c'+3\le800G+4\le805G.
\]
The inequality $c'+3\le800G+4$ is $\lceil800G\rceil\le800G+1$.
The last inequality uses $G>4/5$, which holds since $n>700$ in this case:
$G\ge\ln2\,\ln1402>5$.

\proofstep{The first form of the bound.}
Section~\ref{app:parameters} proves $\mathrm{LR}(C)\le805G$ in the case $a<1$, which includes
$n=1$. So $\mathrm{LR}(C)\le805\ln(2m)\ln(2n)$ for every nonempty finite class
$C\subseteq\{\pm1\}^n$, $n\ge1$. Proposition~\ref{prop:cover-rule} proves the same
bound for $\operatorname{lr}(C)$, because it gives $\operatorname{lr}(C)\le\mathrm{LR}(C)$.

\proofstep{The matrix form.}
Let $B$ be an $N\times M$ sign matrix with $N,M\ge2$, and let $C_B\subseteq\{\pm1\}^M$ be its row class.
Repeated rows count once, so $|C_B|\le N$; repeated columns are allowed, since the bound just
proved makes no assumption on the columns of $C$. Also, for $k\ge2$,
$\ln(2k)=\ln2\,(1+\log_2k)\le2\ln2\,\log_2 k$.
Consequently
\[
 \operatorname{lr}(B)
 \le805\ln(2N)\ln(2M)
 \le3220(\ln2)^2\log_2N\log_2M
 <1548\log_2N\log_2M,
\]
where the last step uses $(\ln2)^2<0.4805$ and $3220\cdot0.4805<1548$. The same chain bounds
$\mathrm{LR}(C_B)$. The restriction $N,M\ge2$ is needed in this form, since its right side vanishes
for $N=1$ or $M=1$, while the form $805\ln(2m)\ln(2n)$ covers those cases too.
This proves both forms of Theorem~\ref{thm:main}.
\end{proof}

\part{Lower bounds and exact values}\label{part:lower}
This part shows that the order of Theorem~\ref{thm:main} is attained for square
matrices and along explicit rectangular families, and it computes the list number
exactly on basic families. Section~\ref{ssec:4} proves Theorem~\ref{thm:main-lower}
through sign complexes of Hanner polytopes. Section~\ref{ssec:5} determines the values
on cubes, Sylvester Hadamard matrices, and projective planes, gives the sharp bound for classes
with bounded error intersections, and shows that restricting the outputs of a learning
rule can increase the list number without bound.

\section{Hanner matrices}\label{ssec:4}

Theorem~\ref{thm:main} bounds the list replicability of an \(N\times N\) sign matrix by a constant multiple of \(\log_2^2N\). This section shows that this order is attained: for every \(N\ge2\) there is an \(N\times N\) sign matrix whose list replicability is at least \(\frac1{72}\log_2^2N\) at every fixed error in \((0,1/2)\), and explicit families give the same lower bound, with an absolute constant, at unequal logarithmic aspect ratios. This removes the \(\log\log N\) factor from the lower bound \(\Omega(\log^2N/\log\log N)\) of~\cite[Proposition 1.16]{BHHLT26}, which originates in~\cite{CMW25}. The matrices are the signed vertex--facet matrices of Hanner polytopes, the polytopes by which Figiel, Lindenstrauss, and Milman~\cite[Example 3.7 and Remark 2]{FLM77} show the asymptotic sharpness of their inequality. The boundary of such a polytope is an antipodal sphere that maps equivariantly into the sign complex of its matrix, so the index is large, and Theorem~\ref{res:a} turns this into a lower bound for list replicability. For the index of free simplicial complexes this sharpness appeared first in~\cite[Remark 9]{FHMNV26v1}.

\subsection{Centrally symmetric polytopes as sign complexes}\label{ssec:4.1}

We first show that the boundary of a centrally symmetric polytope is, up to equivariant homotopy, the sign complex of a partial matrix. This turns the dimension of the polytope into a lower bound for the index and for list replicability.

Let \(P\subset\mathbb R^d\), \(d\ge1\), be a \(d\)-dimensional polytope symmetric about the origin. It has the origin in its interior: if \(x\) is an interior point, so is \(-x\), and hence so is their midpoint. Every facet \(F\) has a unique normalized normal \(a\), given by \(F=P\cap\{\langle a,x\rangle=1\}\) and \(\langle a,x\rangle\le1\) on \(P\). The facet \(-F\) has normal \(-a\), and \(P\) is the set of points with \(\langle a,x\rangle\le1\) for all normals \(a\).

Choose one vertex \(v_i\) from each antipodal pair of vertices of \(P\) and one facet from each antipodal pair of facets. The partial vertex--facet matrix of \(P\) has a row for each chosen facet and a column for each chosen vertex, with entries
\[
A_P[F,i]
=
\begin{cases}
+1,&v_i\in F,\\
-1,&-v_i\in F,\\
*,&\text{otherwise}.
\end{cases}
\]
This is well defined: if \(a\) is the normal of \(F\), then \(v_i,-v_i\in F\) would give \(\langle a,v_i\rangle=1=-\langle a,v_i\rangle\).

For simplicial \(P\) the identification \(S(A_P)\cong\partial P\) is~\cite[Lemma 15]{FHV26}, and Section~7 of~\cite{FHV26} records \(\operatorname{coInd}\partial P=\operatorname{Ind}\partial P=d-1\). The Hanner polytopes used below are in general not simplicial, and for a general polytope the matrix \(A_P\) has undefined entries. Theorem~\ref{res:4.1} therefore adds the equivalence for nonsimplicial \(P\), the completion statement, and the list lower bound.

\begin{resultenv}{Theorem}{res:4.1}{Sign complexes of symmetric polytopes}
Let \(P\subset\mathbb R^d\), \(d\ge1\), be a $d$-dimensional polytope symmetric about the origin. The complex \(S(A_P)\) is nonempty,
\[
S(A_P)\simeq_{\mathbb Z_2}\partial P,
\]
and hence
\[
\operatorname{coInd}(S(A_P))
=
\operatorname{Ind}(S(A_P))
=
d-1.
\]
Every completion \(\widetilde A\) of \(A_P\) satisfies
\[
\operatorname{coInd}(S(\widetilde A))\ge d-1,
\]
and, for every \(\varepsilon\in(0,1/2)\),
\[
\operatorname{lr}(\widetilde A,\varepsilon)\ge d/2.
\]
For Hanner polytopes, \(A_P\) is total.
\end{resultenv}

\begin{proof}
\proofstep{The complex.} Identify the vertex \(i^b\) of \(S(A_P)\) with the vertex \(bv_i\) of \(P\). This is a bijection onto the vertex set \(\operatorname{vert}P\) that carries the involution to \(x\mapsto-x\). By the definition of \(A_P\), the simplex \(\sigma_F^+\) of a chosen facet \(F\) is \(\operatorname{vert}F\), and \(\sigma_F^-\) is \(\operatorname{vert}(-F)\). So the simplices of \(S(A_P)\) are exactly the nonempty subsets of vertex sets of facets of \(P\). We regard \(|S(A_P)|\) as a subset of \(\mathbb R^{\operatorname{vert}P}\), with basis \((e_v)\); the involution exchanges the coordinates of \(v\) and \(-v\).

\proofstep{The map to the boundary.} Map each signed vertex of \(S(A_P)\) to the corresponding vertex of \(P\), and extend affinely: \(f(\sum_vx_ve_v)=\sum_vx_vv\). A simplex of the sign complex is carried into a facet of \(P\), so affine extension defines a continuous equivariant map to \(\partial P\).

\proofstep{The map back.} To construct a map in the other direction, use the barycentric subdivision of \(\partial P\). For a nonempty proper face \(G\) of \(P\), let \(b_G\) be the average of \(\operatorname{vert}G\); it lies in the relative interior of \(G\), and \(b_{-G}=-b_G\). Since \(G\) is convex and compact, each point of \(G\setminus\{b_G\}\) lies on a unique segment from \(b_G\) to the relative boundary of \(G\), which is the union of the proper faces of \(G\). By induction on \(\dim G\), the simplices \(\operatorname{conv}\{b_{G_0},\ldots,b_{G_k}\}\), for chains \(G_0\subsetneq\cdots\subsetneq G_k\) of nonempty proper faces, form a triangulation of \(\partial P\). A point in the relative interior of such a simplex lies in the relative interior of \(G_k\).

Send the subdivision vertex \(b_G\) to the barycenter
\[
g(b_G)=\frac1{|\operatorname{vert}G|}\sum_{v\in\operatorname{vert}G}e_v
\]
of the signed vertex set of \(G\). The set \(\operatorname{vert}G\) is a simplex of \(S(A_P)\), because \(G\) lies in a facet. Along a face chain, all these supports lie in a common facet simplex, namely \(\operatorname{vert}F\) for a facet \(F\supseteq G_k\). So affine extension on the simplices of the subdivision yields a continuous map \(g:\partial P\to|S(A_P)|\). It is equivariant because \(g(b_{-G})\) is obtained from \(g(b_G)\) by the involution.

\proofstep{The two composites.} We show that \(f\) and \(g\) are inverse to each other up to equivariant homotopy. First, \(f\circ g\) is affine on each simplex of the subdivision and fixes its vertices, since \(f(g(b_G))=b_G\); so \(f\circ g=\operatorname{id}\).

Next let \(x\in|S(A_P)|\) lie in the simplex \(\operatorname{vert}F\) for a facet \(F\). Then \(f(x)\in F\). Let \(G_0\subsetneq\cdots\subsetneq G_k\) be the chain whose simplex contains \(f(x)\) in its relative interior. Then \(f(x)\) lies in the relative interior of \(G_k\), and the face \(F\) contains this point, so \(G_k\subseteq F\). Hence \(g(f(x))\) also lies in the closed simplex \(\operatorname{vert}F\), and
\[
\Gamma(x,\lambda)=(1-\lambda)x+\lambda g(f(x))
\]
remains in that simplex for \(\lambda\in[0,1]\). Since \(\Gamma\) is given by a single formula in the coordinates of \(\mathbb R^{\operatorname{vert}P}\), it is a continuous map \(|S(A_P)|\times[0,1]\to|S(A_P)|\), and it commutes with the involution. Thus the maps are equivariant homotopy inverses, for every \(P\), simplicial or not.

\proofstep{The index.} Radial projection identifies \(\partial P\) equivariantly with \(S^{d-1}\): since the origin is an interior point of the compact convex set \(P\), every ray from it meets \(\partial P\) exactly once. Composing with \(f\) and \(g\) gives equivariant maps \(S^{d-1}\to S(A_P)\) and \(S(A_P)\to S^{d-1}\), so \(\operatorname{coInd}(S(A_P))\ge d-1\ge\operatorname{Ind}(S(A_P))\). Since \(\operatorname{coInd}\le\operatorname{Ind}\) (Section~\ref{ssec:2.1}), all three values coincide.

\proofstep{Completions.} The sign complex of a completion contains the original complex equivariantly: a completion \(\widetilde A\) replaces the entries \(*\) by signs, so each signed row simplex can only grow, and the involution is unchanged. Composing with \(S^{d-1}\to S(A_P)\) gives \(\operatorname{coInd}(S(\widetilde A))\ge d-1\). Finally, Theorem~\ref{res:a}, applied to the total class of \(\widetilde A\) and together with \(\operatorname{coInd}\le\operatorname{Ind}\), gives
\[
d-1\le2\operatorname{lr}(\widetilde A,\varepsilon)-1,
\]
which is the stated lower bound. The totality of \(A_P\) for Hanner polytopes is proved in Section~\ref{ssec:4.2}; see \eqref{eq:res-4.1}.
\end{proof}

\subsection{Integer Hanner recursion}\label{ssec:4.2}

Hanner polytopes are built from a segment by two dual operations, products and free sums. We compute their vertices and facet normals exactly. This gives the numbers of rows and columns of their matrices and shows that these matrices are total.

For polytopes \(P\subset\mathbb R^{d_1}\), \(Q\subset\mathbb R^{d_2}\) as in Section~\ref{ssec:4.1}, the free sum is \(P\oplus Q=\operatorname{conv}((P\times\{0\})\cup(\{0\}\times Q))\). The Hanner polytopes are the polytopes obtained from the segment \([-1,1]\) by finitely many products and free sums of polytopes already obtained.

Let \(V(P)\) denote the vertices and \(W(P)\) the normalized facet normals. Starting from the segment with
\[
V=W=\{-1,1\},
\]
the recursion is
\[
V(P\times Q)=\{(v,w)\},
\]
\[
W(P\times Q)=\{(a,0)\}\cup\{(0,b)\},
\]
and
\[
V(P\oplus Q)=\{(v,0)\}\cup\{(0,w)\},
\]
\[
W(P\oplus Q)=\{(a,b)\}.
\]
In these formulas \(v\in V(P)\), \(w\in V(Q)\), \(a\in W(P)\), \(b\in W(Q)\). The next lemma verifies them.

\begin{lemma}[Products and free sums]\label{lem:hanner-product-sum}
For such \(P,Q\), the sets \(P\times Q\) and \(P\oplus Q\) are \((d_1+d_2)\)-dimensional polytopes symmetric about the origin, and their vertices and normalized facet normals are given by the displayed recursion. In particular,
\begin{align*}
|V(P\times Q)|&=|V(P)|\,|V(Q)|, & |W(P\times Q)|&=|W(P)|+|W(Q)|,\\
|V(P\oplus Q)|&=|V(P)|+|V(Q)|, & |W(P\oplus Q)|&=|W(P)|\,|W(Q)|.
\end{align*}
\end{lemma}

\begin{proof}
Both polytopes are symmetric about the origin, and both contain a neighborhood of the origin in \(\mathbb R^{d_1+d_2}\): the product contains a product of small balls, and the free sum contains the convex hull of small balls in \(\mathbb R^{d_1}\times\{0\}\) and \(\{0\}\times\mathbb R^{d_2}\).

\proofstep{Products.} A point \((x,y)\) lies in \(P\times Q\) if and only if \(\langle(a,0),(x,y)\rangle\le1\) and \(\langle(0,b),(x,y)\rangle\le1\) for all \(a\in W(P)\), \(b\in W(Q)\). Equality in the first inequality defines \(F_a\times Q\), where \(F_a\) is the facet of \(a\), and equality in the second defines \(P\times G_b\), where \(G_b\) is the facet of \(b\). Both have dimension \(d_1+d_2-1\). In a finite description by inequalities, every facet is cut out by one of the inequalities, so these are exactly the facets and normals. A point \((x,y)\) is extreme if and only if \(x\) and \(y\) are extreme: a segment through \(x\) in \(P\) gives a segment through \((x,y)\), and a segment through \((x,y)\) projects to segments through \(x\) and through \(y\), at least one of them nondegenerate.

\proofstep{Free sums: the facets.} Let \(g_P(x)=\max_{a\in W(P)}\langle a,x\rangle\), and define \(g_Q\) in the same way. Since \(W(P)=-W(P)\), we have \(g_P\ge0\); moreover \(x\in P\) if and only if \(g_P(x)\le1\), and \(g_P(x)=0\) only for \(x=0\), because \(P\) is bounded. Then
\[
P\oplus Q=\{(x,y):g_P(x)+g_Q(y)\le1\}
=\{(x,y):\langle(a,b),(x,y)\rangle\le1\ \ \forall a,b\}.
\]
For the first equality, note that \(P\) and \(Q\) are convex, so every point of \(P\oplus Q\) has the form \((\lambda x',(1-\lambda)y')\) with \(x'\in P\), \(y'\in Q\), and such a point has \(g_P+g_Q\le\lambda+(1-\lambda)\). Conversely, if \(\lambda=g_P(x)\) and \(\mu=g_Q(y)\) satisfy \(\lambda+\mu\le1\), then \((x,y)=\lambda(x/\lambda,0)+\mu(0,y/\mu)+(1-\lambda-\mu)(0,0)\), where a term with vanishing coefficient is omitted. The second equality is the definition of the maxima.

The face where \(\langle(a,b),\cdot\rangle=1\) contains \(F_a\times\{0\}\) and \(\{0\}\times G_b\). Their affine hulls are flats of dimensions \(d_1-1\) and \(d_2-1\) in complementary subspaces, and neither contains the origin. So the affine hull of their union has dimension \((d_1-1)+(d_2-1)+1=d_1+d_2-1\). Thus every inequality defines a facet, and these are all the facets and normals.

\proofstep{Free sums: the vertices.} The polytope \(P\oplus Q\) is the convex hull of the points \((v,0)\) and \((0,w)\), so its vertices are among them. Each \((v,0)\) is a vertex: choose a linear functional \(\ell\) with \(\ell(v)=1\) and \(\ell<1\) on \(P\setminus\{v\}\). On \((\lambda x',(1-\lambda)y')\), the functional \((\ell,0)\) has the value \(\lambda\ell(x')\), which equals \(1\) only for \(\lambda=1\), \(x'=v\). Exchanging the roles of \(P\) and \(Q\) gives the same for \((0,w)\).

The counts follow, because the displayed sets have no repeated elements.
\end{proof}

The recursion keeps all coordinates integral: by induction, they lie in \(\{-1,0,1\}\). Moreover, every vertex \(v_i\) and every normal \(a_j\) of a Hanner polytope satisfy
\[
\langle a_j,v_i\rangle\in\{-1,1\}.
\refstepcounter{equation}\tag{\theequation}\label{eq:res-4.1}
\]
This holds initially and is preserved by both operations, because \(\langle(a,0),(v,w)\rangle=\langle a,v\rangle\), \(\langle(0,b),(v,w)\rangle=\langle b,w\rangle\), \(\langle(a,b),(v,0)\rangle=\langle a,v\rangle\), and \(\langle(a,b),(0,w)\rangle=\langle b,w\rangle\). For a Hanner polytope, \(\langle a_j,v_i\rangle=1\) means \(v_i\in F_j\), and \(\langle a_j,v_i\rangle=-1\) means \(-v_i\in F_j\). Thus
\[
A_P[j,i]=\langle a_j,v_i\rangle
\]
is total.

The recursion can be carried out on the matrices themselves, without reference to the polytopes. For matrices \(A\) of size \(f_P\times n_P\) and \(B\) of size \(f_Q\times n_Q\), we define two matrices as follows.

The matrix \(\operatorname{Prod}(A,B)\) has size
\[
(f_P+f_Q)\times(2n_Pn_Q).
\]
Its columns are indexed by \((i,j,s)\), \(s\in\{\pm1\}\), and its two row groups are
\[
\operatorname{Prod}_{r,(i,j,s)}=A_{ri},
\qquad
\operatorname{Prod}_{u,(i,j,s)}=sB_{uj}.
\]

The matrix \(J(A,B)\), which we call the \emph{signed join}, has size
\[
(2f_Pf_Q)\times(n_P+n_Q),
\]
with rows
\[
J(A,B)_{(r,u,s),\cdot}
=
(A_{r,\cdot},sB_{u,\cdot}).
\]

These are the matrices of the product and of the free sum. For Hanner polytopes \(P,Q\) and \(A=A_P\), \(B=A_Q\), with vertex representatives \(v_i,w_j\) and normals \(a_r,b_u\), the pairs \((v_i,sw_j)\) represent the antipodal pairs of vertices of \(P\times Q\), and \((a_r,0),(0,b_u)\) represent its facets. Then \(\operatorname{Prod}(A,B)=A_{P\times Q}\) by \(\langle(a_r,0),(v_i,sw_j)\rangle=A_{ri}\) and \(\langle(0,b_u),(v_i,sw_j)\rangle=sB_{uj}\). Likewise, \((v_i,0),(0,w_j)\) represent the vertices of \(P\oplus Q\), and \((a_r,sb_u)\) represent its facets, which gives \(J(A,B)=A_{P\oplus Q}\). The initial matrix, that of the segment, is \([+]\).

The square matrices of Theorem~\ref{res:4.3}(iii) come from the most balanced use of the two operations. For the balanced recursion
\[
\mathcal H_{4d}=(\mathcal H_d\times \mathcal H_d)\oplus(\mathcal H_d\times \mathcal H_d),
\qquad
\mathcal H_1=[-1,1],
\]
the exact counts are
\[
\log_2|V|=2\sqrt d-1,
\qquad
\log_2|F|=3\sqrt d-2.
\]
By Lemma~\ref{lem:hanner-product-sum}, the numbers \(x=\log_2|V|\), \(y=\log_2|F|\) of \(\mathcal H_d\) become \(2x+1\) and \(2(y+1)\) for \(\mathcal H_{4d}\); and with \(\sqrt{4d}=2\sqrt d\), the displayed formulas satisfy these relations and the initial values \(x=y=1\) at \(d=1\).
These exact counts appear in the proof of~\cite[Example 3.7]{FLM77}.
Polytopes with a general exponent, as in Theorem~\ref{res:4.3}(ii), are also studied
in~\cite[Proposition 4.1]{Milo25b}.

Writing \(d=t^2\), the matrix \(A_{\mathcal H_d}\) has
\[
N=2^{3(t-1)}
\quad\text{facet pairs},\qquad
M=2^{2(t-1)}
\quad\text{vertex pairs}.
\]
The matrices of even index in Theorem~\ref{res:4.3}(i) have the transposed shape.

\subsection{Signed joins and the quantitative lower bound}\label{ssec:4.3}

The signed join is the free sum read directly on matrices. Under this operation the coindex grows at least as for a join of spheres and the numbers of columns add, while the numbers of rows multiply. We compute its effect on the sign complex, the coindex, and the rank, and then combine it with transposition to obtain the lower bounds.

Recall that for total matrices \(A,B\),
\[
J(A,B)_{(r,u,s),\cdot}
=
(A_{r,\cdot},sB_{u,\cdot}),
\qquad
s\in\{\pm1\}.
\]
If \(A\) has size \(N_1\times M_1\) and \(B\) has size \(N_2\times M_2\), then \(J(A,B)\) has size \(2N_1N_2\times(M_1+M_2)\).

\begin{resultenv}{Theorem}{res:4.2}{Signed join}
For total matrices \(A,B\),
\[
S(J(A,B))=S(A)*S(B),
\]
\[
\operatorname{coInd}(S(J(A,B)))
\ge
\operatorname{coInd}(S(A))
+
\operatorname{coInd}(S(B))
+1,
\]
and
\[
\operatorname{rank}J
=
\operatorname{rank}A+\operatorname{rank}B.
\]
\end{resultenv}

\begin{proof}
\proofstep{The complex.} The sign \(s\), together with antipodal completion, supplies all combinations of signed row simplices in the two blocks. Indeed, the signed rows of \(J(A,B)\) are \(\kappa(A_{r,\cdot},sB_{u,\cdot})=(\kappa A_{r,\cdot},\kappa sB_{u,\cdot})\), and the pair \((\kappa,\kappa s)\) runs through all of \(\{\pm1\}^2\). So the facets of \(S(J(A,B))\) are exactly the disjoint unions of a facet of \(S(A)\) and a facet of \(S(B)\). As both complexes are nonempty, every union of a simplex of \(S(A)\), or the empty set, and a simplex of \(S(B)\), or the empty set, lies in such a facet. This is precisely the simplicial join.

\proofstep{The coindex.} Joining the equivariant sphere maps gives the coindex inequality. Let \(\phi_1:S^{a_1}\to|S(A)|\) and \(\phi_2:S^{a_2}\to|S(B)|\) be odd maps with \(a_1,a_2\) the two coindices. Regard both realizations in the coordinates of the vertices, and for \((y_1,y_2)\in S^{a_1+a_2+1}\subset\mathbb R^{a_1+1}\oplus\mathbb R^{a_2+1}\) put
\[
\Phi(y_1,y_2)=\|y_1\|^2\phi_1(y_1/\|y_1\|)+\|y_2\|^2\phi_2(y_2/\|y_2\|),
\]
with a term omitted when its coefficient vanishes. The coefficients have sum one, so the value lies in \(|S(A)*S(B)|\). The map \(\Phi\) is continuous, because \(\phi_1,\phi_2\) are bounded, and it commutes with the antipodal maps and the involution.

\proofstep{The rank.} Adding and subtracting the two rows with \(s=+1\) and \(s=-1\) isolates the row spans in the two disjoint column blocks: they give \(2(A_{r,\cdot},0)\) and \(2(0,B_{u,\cdot})\). Conversely, every row of \(J(A,B)\) lies in the sum of these two spans. Their sum is direct, giving the displayed equality.
\end{proof}

To see what Theorem~\ref{res:4.3} below gains, we first use the same join map to recover the lower bound that it improves. For classes \(C_1,C_2\) on disjoint domains, the ordinary join \(C_1*C_2\) is the class of all pairs \((c_1,c_2)\) with \(c_j\in C_j\). A set of labeled points is realized by a member of \(C_1^\pm*C_2^\pm\) exactly when its parts over the two domains are realized in \(C_1^\pm\) and in \(C_2^\pm\). Hence
\[
\Delta_{C_1^\pm*C_2^\pm}
\cong
\Delta_{C_1^\pm}*\Delta_{C_2^\pm} ,
\]
and the join map of the proof of Theorem~\ref{res:4.2} gives \(\operatorname{coInd}(C_1^\pm*C_2^\pm)\ge\operatorname{coInd}(C_1)+\operatorname{coInd}(C_2)+1\).

We apply this to symmetric blocks. The inequality need not hold for \(C_1*C_2\) when the blocks are not symmetric: for \(C_1=C_2=\{+\}\) on singleton domains, each \(C_j\) has coindex \(0\), while \(C_1*C_2=\{(+,+)\}\) has a sign complex consisting of two disjoint edges, which receives no odd map from the connected space \(S^1\), so its coindex is \(0\).

The lower bound of \cite[Proposition 1.16]{BHHLT26} uses the \(m\times2^m\) matrix \(U_m\) with a column for every sign pattern, and taking \(U_m^\pm\) as the blocks gives it in the following form. Writing
$\Delta_{\mathrm{ant}}(U_m)=\Delta U_m\cap(-\Delta U_m)$, the parameter
$\operatorname{sd}(U_m)$ of~\cite{CMW25} is its equivariant coindex.
The inclusion $\Delta_{\mathrm{ant}}(U_m)\subseteq\Delta U_m^\pm$
is equivariant, so their bound $m-2$ transfers to
$\operatorname{coInd}(\Delta U_m^\pm)$. The signed rows of the ordinary join of symmetric classes are the pairs of their rows, so the join of \(k\) copies of \(U_m^\pm\) has the sign complex \((S(U_m^\pm))^{*k}\). This join has \((2m)^k\) rows, \(k2^m\) columns, and, by the join map of the proof of Theorem~\ref{res:4.2}, coindex at least \(k(m-2)+k-1\). For \(k=\lfloor m/\log_2(2m)\rfloor\) there are at most \(2^m\) rows, so repeating rows gives a square matrix of size \(N=k2^m\) with coindex \(\Omega(m^2/\log m)=\Omega(\log^2N/\log\log N)\).

In this estimate each block contributes a factor \(2m\) to the number of rows, and this restricts the number of blocks to \(k=\lfloor m/\log_2(2m)\rfloor\). This is the source of the \(\log\log N\) factor. Theorem~\ref{res:4.3} avoids it by starting from the segment and alternating the signed join with transposition: two consecutive steps multiply the dimension by four and roughly double both \(\log N\) and \(\log M\), so the product of the two logarithms keeps pace with the index.

To reach every size and to compare rectangular and square matrices, we also use the fact that list numbers are unchanged by repetition of rows or columns.

\begin{lemma}[Repetition]\label{lem:hanner-repetition}
If \(B'\) is obtained from a total matrix \(B\) by repeating rows and columns, then \(\operatorname{lr}(B',\varepsilon)=\operatorname{lr}(B,\varepsilon)\) for every \(\varepsilon>0\), and \(\operatorname{rank}B'=\operatorname{rank}B\).
\end{lemma}

\begin{proof}
Repeated rows do not change the row class. For columns, let \(\pi:[M']\to[M]\) be the surjection with \(B'_{r,j}=B_{r,\pi(j)}\), and fix a section \(\iota:[M]\to[M']\), \(\pi\circ\iota=\operatorname{id}\).

Let \(\mathcal A\) be an \((\varepsilon,L)\)-list-replicable rule for \(C_B\). On a sample from the domain \([M']\), apply \(\pi\) to the points, run \(\mathcal A\), and compose its output \(h\) with \(\pi\). If \(\mu'\) is realized by \(c\circ\pi\), the pushforward \(\pi_*\mu'\) is realized by \(c\), and \(\operatorname{loss}_{\mu'}(h\circ\pi)=\operatorname{loss}_{\pi_*\mu'}(h)\). So the list \(\{h\circ\pi:h\in\mathcal L_{\pi_*\mu'}\}\) serves \(\mu'\), with the same sample sizes, and \(\operatorname{lr}(B',\varepsilon)\le\operatorname{lr}(B,\varepsilon)\).

Conversely, let \(\mathcal A'\) be a rule for \(C_{B'}\). On a sample from \([M]\), apply \(\iota\), run \(\mathcal A'\), and compose its output \(h'\) with \(\iota\). If \(\mu\) is realized by \(c\), then \(\iota_*\mu\) is realized by \(c\circ\pi\), and \(\operatorname{loss}_\mu(h'\circ\iota)=\operatorname{loss}_{\iota_*\mu}(h')\). This gives the reverse inequality. Rank is unchanged because the rows and columns added are copies.
\end{proof}

We can now state the lower bounds. Part (i) of the next theorem describes the matrices by the signed join and transposition alone, part (ii) varies the logarithmic aspect ratio, part (iii) gives the square case, and part (iv) shows that for these matrices the list number is determined up to a factor of two. Parts (ii), (iii), and (iv) prove Theorem~\ref{thm:main-lower}. In part (i), \(A_0=[+]\), \(N_t\times M_t\) denotes the size of \(A_t\), and \(^{\mathsf T}\) is transposition.

\begin{resultenv}{Theorem}{res:4.3}{Hanner lower bounds}

(i) The recursive family
\[
A_{t+1}=J(A_t,A_t)^{\mathsf T}
\]
satisfies, for the sign complex and the rank of \(A_t\),
\[
\operatorname{Ind}
=
\operatorname{coInd}
=
2^t-1
=
\operatorname{rank}-1,
\]
\[
N_{2k}=2^{2(2^k-1)},
\qquad
M_{2k}=2^{3(2^k-1)}.
\]

(ii) For every fixed \(a\in(0,1)\) and every integer \(t\ge\lceil4/\min(a,1-a)^2\rceil\), the explicit \(N_t\times M_t\) matrix \(B_t=B_t(a)\) constructed in the proof satisfies, with \(d=2^t\), for every \(\varepsilon\in(0,1/2)\),
\[
\operatorname{lr}(B_t,\varepsilon)\ge d/2
\ge
\frac1{16}\log_2N_t\log_2M_t,
\]
\[
\log M_t=\Theta(d^a),
\qquad
\log N_t=\Theta(d^{1-a}).
\]

(iii) For every \(N\ge2\) and every \(\varepsilon\in(0,1/2)\), there exists an \(N\times N\) total matrix \(B\) with
\[
\operatorname{lr}(B,\varepsilon)\ge\frac1{72}\log_2^2N.
\]
When \(N=2^{3\cdot2^k-3}\), the constant \(1/18\) is available.

(iv) For the matrix of every Hanner polytope of dimension $d$, which includes the
matrices in (i) and (ii), and for its repetitions, which include the matrices in (iii),
\[
\operatorname{lr}\le\operatorname{rank}=d.
\]
\end{resultenv}

\begin{proof}
\proofstep{Part (i): the index.} Begin with the \(1\times1\) segment matrix \(A_0=[+]\), for which \(S(A_0)=S^0\) and \(\operatorname{rank}A_0=1\). By Theorem~\ref{res:4.2}, the signed join doubles the rank and raises the coindex as a join of spheres does. Transposition preserves both indices (Theorem~\ref{res:2.1}), and it preserves rank. Thus \(\operatorname{rank}A_t=2^t\), and if \(\operatorname{coInd}S(A_t)\ge2^t-1\), then
\[
\operatorname{coInd}S(A_{t+1})=\operatorname{coInd}S(J(A_t,A_t))\ge2(2^t-1)+1=2^{t+1}-1.
\]

The rank bounds the index from above, so the two bounds meet at each step. Send \(i^b\) to \(b\) times column \(i\) of \(A_t\) and extend affinely on simplices. On a simplex contained in \(\sigma_r^s\), coordinate \(r\) of the image of every vertex equals \(s\) by \eqref{eq:signed-row-simplex}, so the image of the simplex avoids zero. After normalization this gives an odd map from \(|S(A_t)|\) to the unit sphere of the column span, so \(\operatorname{Ind}S(A_t)\le\operatorname{rank}A_t-1\). This is the profile-map argument in the proof of Theorem~\ref{res:3.1}; the assumption \(N,M\ge2\) of that theorem plays no role in it. With \(\operatorname{coInd}\le\operatorname{Ind}\) (Section~\ref{ssec:2.1}) we get \(2^t-1\le\operatorname{coInd}\le\operatorname{Ind}\le2^t-1\).

\proofstep{Part (i): the sizes.} The size recurrences obtained directly from \(J\) and transposition are \(N_{t+1}=2M_t\) and \(M_{t+1}=2N_t^2\), with \(N_0=M_0=1\). Two steps give \(N_{t+2}=4N_t^2\) and \(M_{t+2}=8M_t^2\). If \(N_{2k}=2^{2(2^k-1)}\) and \(M_{2k}=2^{3(2^k-1)}\), then
\[
N_{2k+2}=2^{2+4(2^k-1)}=2^{2(2^{k+1}-1)},
\qquad
M_{2k+2}=2^{3+6(2^k-1)}=2^{3(2^{k+1}-1)},
\]
which yields the displayed even-index counts.

\proofstep{Part (ii): the polytopes.} Fix \(a\in(0,1)\) and put
\[
\alpha=\min(a,1-a),
\qquad
t_0=\lceil4/\alpha^2\rceil\ge16.
\]
For every integer \(t\ge t_0\) let
\[
p=\lfloor at\rfloor,
\qquad
q=t-p,
\qquad
d=2^t=2^{p+q},
\]
and let \(P_t\) be the Hanner polytope obtained from the segment by the operation word
\[
(PS)^qP^{p-q}\quad(p\ge q),
\]
or
\[
(SP)^pS^{q-p}\quad(q>p).
\]
The word is read from left to right, applying its first letter first.
The letter $P$ replaces the current polytope $Q$ by $Q\times Q$;
the letter $S$ replaces it by $Q\oplus Q$.
Every letter doubles the dimension, and the word has \(2q+(p-q)=t\), respectively \(2p+(q-p)=t\), letters. So \(\dim P_t=2^t=d\).

Writing
\[
x=\log_2|V|,
\qquad
y=\log_2|F|,
\]
the exact formulas are
\[
x=(2^{q+1}-1)2^{p-q},
\qquad
y=3\cdot2^q-2+p-q
\quad(p\ge q),
\refstepcounter{equation}\tag{\theequation}\label{eq:res-4.2}
\]
and
\[
x=3\cdot2^p-2+q-p,
\qquad
y=(2^{p+1}-1)2^{q-p}
\quad(q>p).
\refstepcounter{equation}\tag{\theequation}\label{eq:res-4.3}
\]
Indeed, by Lemma~\ref{lem:hanner-product-sum} the letters act by
\[
P:(x,y)\mapsto(2x,y+1),
\qquad
S:(x,y)\mapsto(x+1,2y),
\]
starting from \((1,1)\). If \(p\ge q\), the block \(PS\) acts by \((x,y)\mapsto(2x+1,2y+2)\), that is, it doubles \(x+1\) and \(y+2\). After \(q\) blocks, \(x+1=2^{q+1}\) and \(y+2=3\cdot2^q\). Then \(P^{p-q}\) multiplies \(x\) by \(2^{p-q}\) and adds \(p-q\) to \(y\). If \(q>p\), the block \(SP\) acts by \((x,y)\mapsto(2x+2,2y+1)\), that is, it doubles \(x+2\) and \(y+1\). After \(p\) blocks, \(x+2=3\cdot2^p\) and \(y+1=2^{p+1}\). Then \(S^{q-p}\) adds \(q-p\) to \(x\) and multiplies \(y\) by \(2^{q-p}\).

Set
\[
B_t=A_{P_t},
\qquad
N_t=2^{y-1},
\qquad
M_t=2^{x-1}.
\refstepcounter{equation}\tag{\theequation}\label{eq:res-4.4}
\]
The matrix \(B_t\) is total by \eqref{eq:res-4.1}, and it has one row for each of the \(2^{y-1}\) facet pairs and one column for each of the \(2^{x-1}\) vertex pairs.

\proofstep{Part (ii): the estimates.} For the quantitative product bound we use
\[
|p-q|\le t\le2^{\min(p,q)}.
\]
The first inequality holds because \(p,q\ge0\) and \(p+q=t\). For the second, \(p\ge at-1\) and \(q\ge(1-a)t\) give \(\min(p,q)\ge\alpha t-1\). Put \(u=\sqrt t\ge2/\alpha\ge4\). Then \(\alpha u^2\ge2u\), hence \(\alpha u^2-u-1\ge u-1\ge1\), and therefore \(\alpha t-1\ge\sqrt t\). Moreover \(\sqrt t\ge\log_2t\) for \(t\ge16\): equality holds at \(t=16\), and the derivative \((\sqrt t\ln2-2)/(2t\ln2)\) of \(\sqrt t-\log_2t\) is positive for \(t\ge16\). Hence \(\min(p,q)\ge\log_2t\), that is, \(t\le2^{\min(p,q)}\). Also \(\min(p,q)\ge\sqrt t\ge4\).

Substitution in \eqref{eq:res-4.2}--\eqref{eq:res-4.3} gives the following. If \(p\ge q\), then \(x=2^{p+1}-2^{p-q}\) and \(y\le3\cdot2^q+|p-q|\le3\cdot2^q+t\), so
\[
2^p\le x\le2\cdot2^p,
\qquad
2^q\le y\le4\cdot2^q,
\]
where \(t\le2^q\) was used and \(y\ge3\cdot2^q-2\ge2^q\). If \(q>p\), then \(y=2^{q+1}-2^{q-p}\) and \(x\le3\cdot2^p+|p-q|\le3\cdot2^p+t\), so
\[
2^p\le x\le4\cdot2^p,
\qquad
2^q\le y\le2\cdot2^q,
\]
where \(t\le2^p\) was used and \(x\ge3\cdot2^p-2\ge2^p\). In both cases
\[
xy\le8\cdot2^{p+q}=8d.
\]
Since \(p\le at<p+1\) and \((1-a)t\le q<(1-a)t+1\), we have \(d^a/2<2^p\le d^a\) and \(d^{1-a}\le2^q<2d^{1-a}\). With \(x,y\ge2\), hence \(x/2\le x-1\le x\) and \(y/2\le y-1\le y\), this gives
\[
\frac{d^a}4\le x-1\le4d^a,
\qquad
\frac{d^{1-a}}2\le y-1\le8d^{1-a},
\]
that is, by \eqref{eq:res-4.4},
\[
\log M_t=\Theta(d^a),
\qquad
\log N_t=\Theta(d^{1-a}).
\]
Theorem~\ref{res:4.1} and Theorem~\ref{res:a} give
\[
d-1
=
\operatorname{coInd}(S(B_t))
=
\operatorname{Ind}(S(B_t))
\le2\operatorname{lr}(B_t,\varepsilon)-1.
\]
Consequently,
\[
\operatorname{lr}(B_t,\varepsilon)\ge d/2
\ge
\frac1{16}xy
\ge
\frac1{16}(x-1)(y-1)
=
\frac1{16}\log_2N_t\log_2M_t.
\]

\proofstep{Part (iii): square matrices.} Use the balanced matrices with
\[
d=4^k,
\qquad
n_k=2^{3\cdot2^k-3}.
\]
The matrix \(A_{\mathcal H_{4^k}}\) has \(n_k\) rows and \(2^{2(2^k-1)}\le n_k\) columns (Section~\ref{ssec:4.2}, with \(t=2^k\)). Repeat columns (rows, in the transposed orientation of (i)) to obtain an \(n_k\times n_k\) matrix. By Lemma~\ref{lem:hanner-repetition}, repetition preserves list replicability.

The balanced Hanner matrices are total by \eqref{eq:res-4.1}, so the lower bound \(\operatorname{lr}\ge d/2\) of Theorem~\ref{res:4.1} applies at these sizes, for every \(\varepsilon\in(0,1/2)\). Together with \(\log_2n_k=3\cdot2^k-3<3\cdot2^k\) it gives \(\operatorname{lr}\ge d/2=4^k/2>\log_2^2n_k/18\), the stated constant \(1/18\). The condition \(N=n_k\ge2\) means \(k\ge1\).

For an arbitrary \(N\ge2\), choose the largest \(k\ge0\) with \(n_k\le N\) and repeat rows and columns to size \(N\). Since \(\log_2N<\log_2n_{k+1}=3\cdot2^{k+1}-3<6\cdot2^k\), we get \(\log_2^2N<36\cdot4^k=72\cdot(d/2)\), hence \(\operatorname{lr}(B,\varepsilon)\ge d/2>\log_2^2N/72\). For \(k=0\) the matrix has all entries positive, and the inequality reduces to the bound \(\operatorname{lr}\ge1\). The matrix \(B\) does not depend on \(\varepsilon\).

\proofstep{Part (iv): the upper bound.} The integer vertex--normal factorization has rank \(d\): by \eqref{eq:res-4.1}, \(A_P=WX^T\), where the rows of \(W\) are the chosen normals \(a_j\) and the rows of \(X\) are the chosen vertices \(v_i\). The vertices of the full-dimensional polytope span \(\mathbb R^d\), and the normals span \(\mathbb R^d\) because \(P\) is bounded, so \(X^T\) is surjective onto \(\mathbb R^d\) and \(W\) is injective on it. The same factorization, with \(A_P[j,i]\langle a_j,v_i\rangle=1\), shows \(\operatorname{signrk}\le d\). Then \(\operatorname{lr}\le\operatorname{signrk}\le\operatorname{rank}\) (Section~\ref{ssec:2.1}) gives (iv) for \(A_P\). The matrices \(B_t\) of (ii) and the balanced matrices of (iii) are of this form, and Lemma~\ref{lem:hanner-repetition} covers their repetitions.

The matrices of (i) are of this form as well, with \(d=2^t\). By Section~\ref{ssec:4.2}, \(J(A_Q,A_Q)=A_{Q\oplus Q}\), and the transpose of \(A_Q\) is the matrix of the polar polytope \(Q^\circ\), because polarity exchanges vertices and normalized facet normals. Put \(Q_0=[-1,1]\) and \(Q_{t+1}=(Q_t\oplus Q_t)^\circ\). The polar of a convex hull is the intersection of the polars, so \(Q_{t+1}=Q_t^\circ\times Q_t^\circ\); and \(Q_{t+1}^\circ=Q_t\oplus Q_t\). By induction \(Q_t\) and \(Q_t^\circ\) are Hanner polytopes, starting from \(Q_0^\circ=Q_0\), and \(A_t=A_{Q_t}\) for suitable representatives.
\end{proof}

Theorem~\ref{res:4.3}(ii) keeps \(a\) fixed in \((0,1)\), and some restriction on the aspect ratio is necessary. A lower bound \(c\log N\log M\) cannot hold for every pair \((N,M)\): for \(N=2\), Theorem~\ref{thm:baseline} gives \(\operatorname{lr}\le1+\lfloor2/2\rfloor=2\) while \(\log M\) is unbounded.

\subsection{The Figiel--Lindenstrauss--Milman inequality and a question of Frick, Hosseini, and Vasileuski}\label{ssec:4.4}

So far we have used polytopes to obtain lower bounds for matrices. In this subsection we read the relation in the other direction and apply Theorem~\ref{thm:main} to polytopes and to free simplicial complexes.

The Figiel--Lindenstrauss--Milman inequality \cite{FLM77} bounds the dimension of a centrally symmetric polytope by a constant multiple of the product of the logarithms of its numbers of vertices and facets. It has a very short proof \cite{Milo25a}. Its asymptotic sharpness is established by the Hanner construction in~\cite[Example 3.7 and Remark 2]{FLM77} and studied further in~\cite{Milo25b}, and the face numbers of Hanner polytopes are the subject of \cite{Milo26}.

Question 37 of~\cite{FHV26} asks for a logarithmic-product index bound
for finite strongly regular free $\mathbb Z_2$-CW complexes. This is
proved in~\cite{FHMNV26v1,FHMNV26}. Since $\operatorname{Ind}\le2\operatorname{lr}-1$,
Theorem~\ref{thm:main} also bounds the index, and the proposition below
deduces the simplicial case from it. Theorem~\ref{thm:index-gap} shows that
$\operatorname{lr}$ can exceed $\operatorname{Ind}+1$ by an unbounded amount.

We make no attempt to optimize the constant in the proposition; its point is that both statements follow from the learning bound. Theorem~\ref{thm:main} supplies the constant $K=1548$ in the form $\operatorname{lr}(B)\le K\log_2(2N)\log_2(2M)$ used below. For an \(N\times M\) sign matrix \(B\) with \(N,M\ge1\), repeat every row and every column once. The resulting \(2N\times2M\) matrix has the same list numbers (Lemma~\ref{lem:hanner-repetition}), and the second bound of Theorem~\ref{thm:main} applies to it, since \(2N,2M\ge2\). This form therefore also covers matrices with one row or one column, as is needed for complexes with one vertex pair or one facet pair.

\begin{resultenv}{Proposition}{res:4.4}{Geometric consequences}
With $K=1548$, every $d$-dimensional polytope \(P\subset\mathbb R^d\) symmetric about the origin, with \(2n\) vertices and \(2f\) facets, satisfies
\[
d\le2K\log_2(2n)\log_2(2f).
\]
Every nonempty finite free \(\mathbb Z_2\)-simplicial complex \(\mathcal X\), with \(n\) vertex pairs and \(f\) maximal-face pairs, satisfies
\[
\operatorname{Ind}(\mathcal X)
\le
2K\log_2(2n)\log_2(2f)-1.
\]
The latter assertion is equivalent to the statement that all total matrices satisfy
\[
\operatorname{Ind}=O(\log(2N)\log(2M)).
\]

\end{resultenv}

\begin{proof}
\proofstep{Polytopes.} Form the \(f\times n\) partial matrix \(A_P\), complete it to a total matrix \(\widetilde A\) of the same size, fix \(\varepsilon\in(0,1/2)\), and use Theorem~\ref{res:4.1} followed by Theorem~\ref{res:a} and the form of Theorem~\ref{thm:main} recorded above, together with \(\operatorname{lr}(\widetilde A,\varepsilon)\le\operatorname{lr}(\widetilde A)\):
\[
d-1\le\operatorname{Ind}(S(\widetilde A))\le2\operatorname{lr}(\widetilde A,\varepsilon)-1\le2K\log_2(2f)\log_2(2n)-1.
\]

\proofstep{Free simplicial complexes.} For a finite free simplicial complex \(\mathcal X\), \cite[Lemma 15]{FHV26} realizes \(\mathcal X\) as the sign complex of an \(f\times n\) partial matrix: its proof takes one row for each pair of facets and one column for each pair of vertices. Complete that matrix and apply the same inequalities to the equivariant subcomplex \(\mathcal X\) of the completed sign complex; restricting an odd map shows \(\operatorname{Ind}(\mathcal X)\le\operatorname{Ind}(S(\widetilde A))\).

\proofstep{The equivalence with total matrices.} Sign complexes of total matrices are among those finite free simplicial complexes, with the corresponding vertex-pair and maximal-face-pair counts: an \(N\times M\) total matrix has \(n=M\) and \(f\le N\) by Section~\ref{ssec:6.1}, because the facets of its sign complex are the sets \(\sigma_r^s\). So the bound for \(\mathcal X\) gives the stated bound for total matrices. In the other direction, if all total matrices satisfy \(\operatorname{Ind}\le K'\log_2(2N)\log_2(2M)\), the completion argument gives \(\operatorname{Ind}(\mathcal X)\le K'\log_2(2f)\log_2(2n)\). Both directions preserve the constant, and this gives the topological equivalence.
\end{proof}

For strongly regular CW complexes, and for the constant $4$, see~\cite{FHMNV26}.

\section{Exact values and output restrictions}\label{ssec:5}

Theorem~\ref{thm:main} and the matrices of Section~\ref{ssec:4} determine, up to a constant, the largest list replicability among square matrices of a given size. They do not say what the value is for a particular matrix. This section computes it for cubes, Sylvester Hadamard matrices, and projective planes, it records the value for thresholds, and it gives a sharp bound for classes with bounded error intersections. For Sylvester Hadamard matrices and projective planes the computation sharpens published bounds of~\cite{FHV26} and~\cite{BHHLT26}. The values show that list replicability can lie far below the real rank, the sign rank, and the switching binary rank, that contractibility of the distribution space does not bound it, and that it depends on more than the sign complex. For the anchored singleton classes, restricting the outputs to the signed class raises the list number at small errors from two to the number of concepts.

\subsection{Cubes and anchored cubes}\label{ssec:5.1}

The binary cube and its anchored version are the basic examples of this section: the lower bounds for Sylvester Hadamard matrices, projective planes, and thresholds are obtained by restricting to a set of columns that carries an anchored cube. We determine both values at every fixed error in \((0,1/2)\). The upper bounds come from sign representations, and the lower bounds from a degree argument.

The bound \(\operatorname{lr}(C)\le\operatorname{signrk}(C)\) of~\cite[Corollary 2.9]{BHH26a} (Section~\ref{ssec:2.1}) has the following explicit form, in which every label of the cover is the sign of a linear functional on the representing vectors.

\begin{resultenv}{Lemma}{res:5.0}{A sign representation gives an explicit margin cover}
Suppose there are vectors \(v_x\in\mathbb R^r\) (\(x\in X\)) and \(w_c\in\mathbb R^r\) (\(c\in C\)) with
\[
c(x)\langle w_c,v_x\rangle\ge1
\qquad(c\in C,\ x\in X).
\]
Then for every \(\varepsilon\in(0,1/2)\) there is a finite closed cover of \(\Delta_C\), indexed by hypotheses, of multiplicity at most \(r\), such that each hypothesis \(h\) satisfies \(\operatorname{loss}_\mu(h)\le\varepsilon/6\) throughout its member. Consequently \(\operatorname{lr}(C,\varepsilon)\le r\).
\end{resultenv}

\begin{proof}
\proofstep{A continuous unit vector.} Put \(R=\max_c\|w_c\|_2\), \(V=\max_x\|v_x\|_2\) and \(\lambda=\varepsilon/(8R^2)\); the hypothesis forces every \(w_c\) and \(v_x\) to be nonzero, so \(R,V>0\). For \(\mu\in\Delta_C\) let \(w(\mu)\) be the unique minimizer of the strictly convex function
\[
w\longmapsto\lambda\|w\|_2^2+\mathbb E_{(x,y)\sim\mu}\bigl(1-y\langle w,v_x\rangle\bigr)_+.
\]
A minimizer exists because this function is continuous and bounded below by \(\lambda\|w\|_2^2\), and it is strictly convex as the sum of a strictly convex function and a convex one.

Comparing with \(w_c\) for a concept \(c\) realizing \(\mu\) gives \(\|w(\mu)\|_2\le R\) and \(\mathbb E_\mu(1-y\langle w(\mu),v_x\rangle)_+\le\lambda R^2=\varepsilon/8\). Indeed, \(\mu\)-almost surely \(y=c(x)\), so the hinge term of \(w_c\) vanishes and the objective at \(w_c\) is \(\lambda\|w_c\|_2^2\le\lambda R^2\), and both terms of the objective at \(w(\mu)\) are nonnegative and bounded by this value. The objective equals \(1\) at \(w=0\) and is at most \(\varepsilon/8<1\) at \(w_c\), so \(w(\mu)\ne0\).

The minimizer depends continuously on \(\mu\): along a convergent sequence of distributions the minimizers are bounded, every convergent subsequence has a limit which, by continuity of the objective in \((w,\mu)\), minimizes the limit problem, and uniqueness identifies that limit. Hence
\[
u(\mu)=w(\mu)/\|w(\mu)\|_2
\]
is a continuous map \(\Delta_C\to S^{r-1}\).

\proofstep{A cover of the sphere.} Take a finite triangulation of \(S^{r-1}\) whose simplices have diameter at most \(\rho=(4RV)^{-1}\). It can be obtained from the boundary of the cross-polytope by repeated barycentric subdivision followed by radial projection: each subdivision multiplies the mesh by at most \((r-1)/r\), and radial projection is Lipschitz on that boundary, where \(\|x\|_2\ge r^{-1/2}\). For \(r=1\) the sphere is two points.

For a vertex \(a\) let \(b_a(u)\) be its barycentric coordinate and \(D_a=\{u:b_a(u)=\max_bb_b(u)\}\). These closed sets cover the sphere, and every point lies in at most \(r\) of them, because a point of \(S^{r-1}\) has at most \(r\) positive barycentric coordinates. Moreover \(u\in D_a\) implies \(\|u-a\|_2\le\rho\), since by Lemma~\ref{res:2.0} the set \(D_a\) lies in the closed star of \(a\), so \(u\) and \(a\) lie in a common simplex.

\proofstep{The labels.} Assign to \(a\) the hypothesis \(h_a(x)=\operatorname{sign}\langle a,v_x\rangle\), with \(\operatorname{sign}0=+1\). If \(h_a(x)\ne y\), then \(y\langle a,v_x\rangle\le0\): for \(y=+1\) the inner product is negative, and for \(y=-1\) it is nonnegative. So, for \(u(\mu)\in D_a\),
\[
y\langle w(\mu),v_x\rangle
=\|w(\mu)\|_2\,y\langle u(\mu),v_x\rangle
\le\|w(\mu)\|_2\bigl(y\langle a,v_x\rangle+\rho V\bigr)
\le RV\rho=\tfrac14 .
\]
The first inequality is the Cauchy--Schwarz bound for \(\langle u(\mu)-a,v_x\rangle\), and the second uses \(y\langle a,v_x\rangle\le0\) and \(\|w(\mu)\|_2\le R\).
Thus every error of \(h_a\) has hinge value at least \(3/4\), and, by Markov's inequality for the nonnegative hinge value,
\[
\operatorname{loss}_\mu(h_a)
\le\tfrac43\,\mathbb E_\mu\bigl(1-y\langle w(\mu),v_x\rangle\bigr)_+
\le\tfrac43\cdot\frac\varepsilon8=\varepsilon/6 .
\]
The sets \(u^{-1}(D_a)\) are closed and cover \(\Delta_C\); merging those carrying the same hypothesis does not increase the multiplicity. Since \(\varepsilon/6\le\varepsilon/2\), Proposition~\ref{prop:cover-rule} with \(\alpha=\tau=\varepsilon/2\) gives \(\operatorname{lr}(C,\varepsilon)\le r\).
\end{proof}

The same representation bounds the index: \(\operatorname{Ind}(S(C))\le r-1\). To see this, send the vertex \((x,y)\) to \(yv_x\), extend affinely on faces, and normalize. A face lies in the simplex of a signed row \(sc\), \(s=\pm1\), so its vertices are \((x,sc(x))\), and \(\langle sw_c,sc(x)v_x\rangle=c(x)\langle w_c,v_x\rangle\ge1\). The inner product with \(sw_c\) is therefore at least one on the whole face. Thus the affine extension avoids zero, and the normalized map is odd. Any strict sign representation can be rescaled so that all these products are at least one, since there are finitely many.

The lower bounds turn a closed cover into a continuous map and then apply degree theory. The first of the following two lemmas replaces a closed cover by a partition of unity with the same number of active labels, at a slightly larger error. The second is the degree statement that we apply to the resulting map.

\begin{lemma}[Thickening a closed cover]\label{lem:thicken-cover}
Let \(\{G_h\}\) be a finite closed cover of \(\Delta_C\), indexed by hypotheses, of multiplicity at most \(L\) and with \(\operatorname{loss}_\mu(h)\le\varepsilon\) on \(G_h\), and let \(\eta>\varepsilon\). There are continuous functions \(\varphi_h:\Delta_C\to[0,1]\) with \(\sum_h\varphi_h=1\) such that at most \(L\) of the values \(\varphi_h(\mu)\) are positive at each \(\mu\), and \(\varphi_h(\mu)>0\) implies \(\operatorname{loss}_\mu(h)<\eta\).
\end{lemma}
\begin{proof}
Use the \(\ell_1\) distance on \(\Delta_C\) and delete empty members. There is \(\rho>0\) such that any \(L+1\) members have open \(\rho\)-neighborhoods with empty common intersection. Otherwise, since there are finitely many members, some \(L+1\) fixed members and points \(\mu_k\) satisfy \(\operatorname{dist}(\mu_k,G)<1/k\) for each of these members \(G\); a limit point of \((\mu_k)\) lies in all \(L+1\) closed sets, contrary to the multiplicity bound.

Decrease \(\rho\) so that \(\rho<\eta-\varepsilon\), and let \(U_h\) be the open \(\rho\)-neighborhood of \(G_h\) in \(\Delta_C\). The loss of \(h\) is the probability of one event, so it is \(1\)-Lipschitz in the \(\ell_1\) distance, and \(\operatorname{loss}_\mu(h)<\varepsilon+\rho<\eta\) on \(U_h\). Finally put \(\varphi_h=\operatorname{dist}(\cdot,\Delta_C\setminus U_h)/\sum_{h'}\operatorname{dist}(\cdot,\Delta_C\setminus U_{h'})\), with the distance to the empty set taken to be one; the denominator is positive because the sets \(U_h\) cover \(\Delta_C\).
\end{proof}

\begin{lemma}[Positive boundary pairing]\label{lem:boundary-pairing}
Let \(k\ge1\), let \(B=\{x\in\mathbb R^k:\|x\|_1\le1\}\), and let \(G:\partial B\to\mathbb R^k\) be continuous with \(\langle G(x),x\rangle>0\) on \(\partial B\). Then \(G/\|G\|_2\) is surjective onto \(S^{k-1}\), and \(G\) has no continuous extension \(B\to\mathbb R^k\setminus\{0\}\).
\end{lemma}
\begin{proof}
For \(k=1\) we have \(G(1)>0>G(-1)\), and both assertions hold by the intermediate value theorem. Let \(k\ge2\). For \(s\in[0,1]\) the vector \((1-s)x+sG(x)\) has positive inner product with \(x\), so it is nonzero, and after normalization this is a homotopy from the radial homeomorphism \(\partial B\to S^{k-1}\) to \(G/\|G\|_2\). Thus \(G/\|G\|_2\) has degree one, after identifying the domain with \(S^{k-1}\) by that homeomorphism. A map of nonzero degree is surjective and is not null-homotopic \cite[Section 2.2]{Hatcher02}, while a nonvanishing extension over the ball would make it null-homotopic.
\end{proof}

We now turn to the values themselves. The value of the binary cube in the supremum sense is \cite[Lemma 4]{CMY23}; the identity at every error in the range \((0,1/2)\) appeared first in \cite[Theorem A]{BHH26c} and the remark following it.

The anchored cube \(\{+\}^a\times\{\pm1\}^k\) is an extremal class in the sense of \cite{BHH26b}: its shattered sets are exactly the sets of free coordinates, and each of them is strongly shattered. Since it is not a full cube, its list replicability number in the supremum sense, \(\operatorname{lr}(\{+\}^a\times\{\pm1\}^k)=k+1\), follows from \cite[Corollary 5.5]{BHH26b} (Theorem B there). Theorem~\ref{res:5.1} gives the anchored value \(k+1\) for every fixed \(\varepsilon\in(0,1/2)\), with a direct degree proof. This strengthens the supremum value of~\cite{BHH26b} and, on this class, the lower bound \(k\) of~\cite[Theorem A]{BHH26c}.

\begin{resultenv}{Theorem}{res:5.1}{Cube and anchored-cube values}
For every \(\varepsilon\in(0,1/2)\),
\[
\operatorname{lr}(\{\pm1\}^k,\varepsilon)=k
\qquad(k\ge1),
\]
and
\[
\operatorname{lr}(\{+\}^a\times\{\pm1\}^k,\varepsilon)=k+1
\qquad(a\ge1,\ k\ge0).
\]
For the anchored class,
\[
\operatorname{Ind}=k,
\qquad
\operatorname{vc}=k,
\qquad
\operatorname{signrk}=k+1.
\]
\end{resultenv}

\begin{proof}
The two lower bounds use the same pairing argument, applied to the signed masses of the free coordinates.

\proofstep{The lower bound for the cube.}
Identify the realizable distributions of the cube with the \(\ell_1\) sphere \(\partial B\) of signed masses of the \(k\) coordinates: a realizable distribution has mass at one of \((i,+),(i,-)\) at most, and it corresponds to \(x\) with \(x_i=\mu(i,+)-\mu(i,-)\). For a hypothesis \(h\in\{\pm1\}^k\),
\[
\langle h,x\rangle
=\sum_{i:h_i=\operatorname{sign}x_i}|x_i|-\sum_{i:h_i\ne\operatorname{sign}x_i}|x_i|
=1-2\operatorname{loss}_x(h).
\]

For \(k=1\) the lower bound \(\operatorname{lr}\ge1\) needs no argument, so let \(k\ge2\). Suppose \(\operatorname{lr}(\{\pm1\}^k,\varepsilon)\le L\) with \(\varepsilon<1/2\), and fix an intermediate error parameter \(\eta\) with \(\varepsilon<\eta<1/2\). The learner-to-closed-cover direction of Theorem~\ref{res:b} gives a finite closed cover of the sphere, indexed by hypotheses, of multiplicity at most \(L\) and with loss at most \(\varepsilon\) on each member. Lemma~\ref{lem:thicken-cover} gives a partition of unity \(\{\varphi_h\}\) with at most \(L\) active labels at each point and loss below \(\eta\) for active labels. Put
\[
F(x)=\sum_h\varphi_h(x)h.
\]
The inequalities \(\operatorname{loss}_x(h)<\eta\) for the active hypotheses imply
\[
\langle F(x),x\rangle=\sum_h\varphi_h(x)\bigl(1-2\operatorname{loss}_x(h)\bigr)>1-2\eta>0.
\]
Thus \(F/\|F\|\) is homotopic to the identity and surjective, by Lemma~\ref{lem:boundary-pairing}. If fewer than \(k\) hypotheses were active at every point, its image would lie in a finite union of proper linear subspaces, because \(F(x)\) lies in the span of the active hypotheses. It would omit a point of the sphere, a contradiction. Hence \(L\ge k\).

\proofstep{The lower bound for the anchored class.}
Let \(C=\{+\}^a\times\{\pm1\}^k\). The case \(k=0\) is a single row, so let \(k\ge1\). We use the \(\ell_1\) ball \(B\) in the \(k\) free coordinates and represent a point \(x\) by the distribution \(\iota(x)\) with mass \(|x_i|\) at \((i,\operatorname{sign}x_i)\) and mass \(1-\|x\|_1\) at the positive atom of the first anchor column. It is realized by a row, and \(\iota:B\to\Delta_C\) is continuous. Let \(\pi(h)\in\{\pm1\}^k\) be the restriction of a hypothesis to the free coordinates.

Suppose \(\operatorname{lr}(C,\varepsilon)\le L\), and let \(\{\varphi_h\}\) be as in Lemma~\ref{lem:thicken-cover} for a closed cover from Theorem~\ref{res:b} and an intermediate error \(\eta\in(\varepsilon,1/2)\). Put \(F(x)=\sum_h\varphi_h(\iota(x))\pi(h)\). On the boundary \(\iota(x)\) has no anchor mass, so \(\langle\pi(h),x\rangle=1-2\operatorname{loss}_{\iota(x)}(h)\), and the same computation as for the cube gives, for \(x\in\partial B\),
\[
\langle F(x),x\rangle>0.
\]

We use the following consequence of Lemma~\ref{lem:boundary-pairing}: a continuous map \(F\) on the \(\ell_1\) ball with \(\langle F(x),x\rangle>0\) on the boundary has an image containing a neighborhood of zero. To see this, perturb the target by a sufficiently small vector: let \(m>0\) be the minimum of \(\langle F(x),x\rangle\) on \(\partial B\) and let \(\|y\|_\infty<m\). The boundary pairing remains positive, since \(\langle F(x)-y,x\rangle\ge m-\|y\|_\infty\|x\|_1>0\), so the normalized boundary map remains homotopic to the identity. If the perturbed target \(y\) were omitted from the ball image, that boundary map would extend over the ball and have degree zero, a contradiction with Lemma~\ref{lem:boundary-pairing} for \(G=F-y\).

If the multiplicity were at most \(k\), the image would lie in the union of convex hulls of at most \(k\) cube vertices, because \(F(x)\) is a convex combination of the active \(\pi(h)\). Each such hull has empty \(k\)-dimensional interior, and there are only finitely many of them. This contradicts the neighborhood conclusion. Hence \(L\ge k+1\).

\proofstep{The upper bounds.}
Apply Lemma~\ref{res:5.0}. For the cube take \(v_x=e_x\) and \(w_c=c\in\mathbb R^k\), so that \(c(x)\langle w_c,v_x\rangle=1\) and \(\operatorname{lr}(\{\pm1\}^k,\varepsilon)\le k\). For the anchored class take \(v_0=e_0\) for the anchor columns, \(v_i=e_i\) for the free coordinates, and \(w_c=(1,c)\in\mathbb R^{k+1}\), giving \(\operatorname{lr}\le k+1\). The case \(k=0\) is the single unanimous concept.

\proofstep{The index and coindex of the anchored class.}
The signed rows of the anchored class are \(\pm(1,c)\) with \(c\) arbitrary, where \(1\) is the all-positive vector on the anchor columns. So its sign complex is the join of \(S(\{\pm1\}^k)=\partial\Diamond_k\) with two disjoint anchor simplices, one for each sign of the anchor vertices.

Contract each anchor simplex equivariantly to a point. This gives a \(\mathbb Z_2\)-homotopy equivalence with the suspension of \(\partial\Diamond_k\cong S^{k-1}\), with the antipodal action; for \(a=1\) the join is literally that suspension. The suspension is \(S^k\) with the antipodal action, and index and coindex are invariant under equivariant homotopy equivalence. Hence \(\operatorname{Ind}=\operatorname{coInd}=k\).

The same values also follow directly. One anchor column and the \(k\) free columns are signed-shattered, since \(s(1,sc)\) restricts to \((s,c)\), so \(\operatorname{coInd}\ge k\) by Section~\ref{ssec:2.1}. The vectors above give \(\operatorname{Ind}\le k\) by the odd map after Lemma~\ref{res:5.0}.

\proofstep{The VC dimension and the sign rank of the anchored class.}
The \(k\) free columns are shattered and no set containing an anchor column is, so \(\operatorname{vc}=k\). For the sign rank, \(\operatorname{signrk}\le\operatorname{rank}=k+1\), since the row space is spanned by the free coordinate vectors and the vector with ones on the anchor columns and zeros elsewhere. Conversely, \(\operatorname{signrk}\ge\operatorname{Ind}+1=k+1\) by the odd map after Lemma~\ref{res:5.0}.
\end{proof}

The model of~\cite{BHH26b}, whose extremal-class theorem is quoted before Theorem~\ref{res:5.1}, allows one rule for each pair \((\varepsilon,\delta)\). Its supremum value agrees with \(\operatorname{lr}\) as defined here, because the proof of Theorem~\ref{res:b} uses a single sufficiently small \(\delta\) and Proposition~\ref{prop:cover-rule} returns one rule for each error.

Lemma~\ref{res:5.0} also gives the elementary bound \(\operatorname{lr}(B,\varepsilon)\le\min(N,M)\), and the anchored cube shows that it cannot be improved uniformly over rectangular matrices:
\[
\sup_{\min(N,M)=n}\operatorname{lr}(B,\varepsilon)=n,
\]
attained by the anchored cube with
\[
N=2^{n-1},
\qquad
M=n.
\]
This class has \(a=1\), \(k=n-1\), and \(2^{n-1}\ge n\), so Theorem~\ref{res:5.1} gives the value \(n\). Conversely Lemma~\ref{res:5.0} bounds every such matrix by \(n\): if \(M=n\) take \(v_x=e_x\) and \(w_c=c\), and if \(N=n\) take \(w_c=e_c\) and \(v_x=(c(x))_{c\in C}\).
Thus no bound of order \(o(\min(N,M))\) holds uniformly over all aspect ratios. In this family the number of rows is exponential in the number of columns.

\subsection{Bounded error intersections}\label{ssec:5.2}

In a projective plane two lines meet in one point, so two observed positive points identify the line. We isolate the property behind this: the rows differ from a fixed background on sets with small pairwise intersections. The resulting bound depends only on the size of these intersections, and not on the numbers of rows and columns. It gives the upper bounds for the anchored singleton classes and the projective planes below.

The case \(s=1\) for projective planes is~\cite[Theorem 3.2]{BHHLT26}, whose two-positive-point identification we follow. Theorem~\ref{res:5.2} treats every \(s\) and every background, adds a coupling through a random threshold, and shows that the bound \(s+2\) is attained.

\begin{resultenv}{Theorem}{res:5.2}{Bounded error intersections}
Let \(C\) be a finite total class, let \(g\in\{\pm1\}^X\), and put
\[
D_c=\{x:c(x)\ne g(x)\}.
\]
If
\[
|D_c\cap D_d|\le s
\]
for every two distinct concepts, where \(s\in\mathbb Z_{\ge0}\), then, for every \(\varepsilon\in(0,1/2)\),
\[
\operatorname{lr}(C,\varepsilon)\le s+2.
\]
Among bounds depending only on \(s\), this bound is the best possible.
\end{resultenv}

\begin{proof}
Fix \(\varepsilon\in(0,1/2)\), and suppose first that \(s\ge1\). Put
\[
\tau=\frac{\varepsilon}{4s}.
\]
For a sample \(S=((x_1,y_1),\ldots,(x_t,y_t))\) let
\[
\widehat q_x=\frac1t\bigl|\{j:x_j=x,\ y_j\ne g(x)\}\bigr|,
\qquad
P_S=\{x:\widehat q_x>0\}
\]
be the empirical frequency of errors of the background at \(x\) and the set of points at which an error of the background is observed.

\proofstep{The learning rule.} (1) If \(|P_S|\ge s+1\), output the unique concept \(c\) with \(P_S\subseteq D_c\). It is unique because two distinct concepts containing these \(s+1\) points in their error sets would violate the intersection hypothesis, and for a realizable sample it exists. If it does not exist, output \(g\). (2) Otherwise draw
\[
U\sim\operatorname{Unif}[\tau,2\tau]
\]
independently of the sample, put \(H_S(U)=\{x:\widehat q_x\ge U\}\), with the non-strict inequality, and output the hypothesis that flips \(g\) on \(H_S(U)\) and agrees with \(g\) elsewhere.

\proofstep{The lists.} Fix a realizable \(\mu\), a concept \(c_\mu\) realizing it, and
\[
q_x=\mathbb P_\mu[x,\ y\ne g(x)],
\qquad
Q_\mu=\{x:q_x\ge\tau/2\}.
\]
There are two cases. In Case A, \(|Q_\mu|\ge s+1\), and the list is \(\{c_\mu\}\). In Case B, \(|Q_\mu|\le s\). The ideal threshold sets
\[
H_\mu(u)=\{x:q_x\ge u\},
\qquad u\in[\tau,2\tau],
\]
are subsets of \(Q_\mu\), because \(u\ge\tau>\tau/2\). They are nested in \(u\), so at most \(s+1\) distinct sets occur, rather than \(2^s\). The list consists of \(c_\mu\) together with those hypotheses obtained by flipping \(g\) on an ideal threshold set whose true loss is at most \(\varepsilon\). It has at most \(s+2\) members and depends only on \(\mu\), \(\varepsilon\) and \(s\), not on the sample or on \(U\). All its members have loss at most \(\varepsilon\), since \(c_\mu\) has loss zero.

\proofstep{Probability estimates.} Let \(M=|X|\), \(0<\delta<1\),
\[
K=(M+1)^s,
\qquad
\eta=\frac{\delta\tau}{8M},
\qquad
t\ge\left\lceil\max\left\{\frac M{\delta\eta^2},\frac{4K}{\delta\varepsilon^2},\frac{8(s+1)}{\delta\tau}\right\}\right\rceil.
\]
The estimates below use only variance bounds, Chebyshev's inequality, and union bounds.

(i) Each \(\widehat q_x\) has variance at most \(1/(4t)\), being an average of \(t\) independent Bernoulli variables, so
\[
\mathbb P[\max_x|\widehat q_x-q_x|>\eta]\le M\cdot\frac1{4t\eta^2}\le\frac\delta4,
\]
by \(t\ge M/(\delta\eta^2)\).

(ii) The probability that \(U\) lies within \(\eta\) of some \(q_x\) is at most \(M\cdot2\eta/\tau=2M\eta/\tau=\delta/4\), since \(U\) has density \(1/\tau\). Outside these two events, \(H_S(U)=H_\mu(U)\); this couples the empirical and the ideal threshold sets through the common random threshold \(U\), for arbitrary values \(q_x\), which need not be distinct. Indeed, if \(q_x\ge U\), then \(q_x>U+\eta\) and \(\widehat q_x\ge q_x-\eta>U\); if \(q_x<U\), then \(q_x<U-\eta\) and \(\widehat q_x\le q_x+\eta<U\).

(iii) A hypothesis output in step (2) flips at most \(s\) points, since \(U>0\) gives \(H_S(U)\subseteq P_S\) and \(|P_S|\le s\) there. For a realizable sample its empirical loss is \(\sum_{x\in P_S\setminus H_S(U)}\widehat q_x\le|P_S|\,U\le2s\tau=\varepsilon/2\). To see the formula, take a sample point at \(x\). If \(x\notin H_S(U)\), the output equals \(g\) at \(x\), and the frequency of errors at \(x\) is \(\widehat q_x\), which is less than \(U\). If \(x\in H_S(U)\), the output equals \(-g\) at \(x\); some sample point at \(x\) has label \(-g(x)\), and a realizable sample has one label at each point, so there are no errors at \(x\).

There are at most \((M+1)^s=K\) hypotheses flipping at most \(s\) points, and Chebyshev's inequality with a union bound gives
\[
\mathbb P[\exists h:\operatorname{loss}_\mu(h)-\widehat{\operatorname{loss}}_S(h)>\varepsilon/2]\le K\cdot\frac{1/(4t)}{(\varepsilon/2)^2}=\frac K{t\varepsilon^2}\le\frac\delta4,
\]
by \(t\ge4K/(\delta\varepsilon^2)\); hence the step-(2) output has true loss at most \(\varepsilon\) outside an event of probability \(\delta/4\).

(iv) In Case A, the errors of the background have mass at least \(\tau/2\) at each of \(s+1\) fixed points of \(Q_\mu\); the probability that one of them is missing from \(P_S\) is at most
\[
(s+1)(1-\tau/2)^t\le(s+1)e^{-t\tau/2}\le\frac{2(s+1)}{t\tau}\le\frac\delta4,
\]
using \(e^{-z}\le1/z\) and \(t\ge8(s+1)/(\delta\tau)\), and when all are present step (1) outputs \(c_\mu\), because \(P_S\subseteq D_{c_\mu}\) for a sample from \(\mu\).

In Case B the output is either the identified concept \(c_\mu\) or, outside the events of (i)--(iii), the flip of \(g\) on the ideal set \(H_\mu(U)\) with true loss at most \(\varepsilon\), which is on the list; the failure probability is at most \(3\delta/4\). In Case A it is at most \(\delta/4\). Thus all quantifiers of the definition of list replicability hold with lists of size at most \(s+2\); for \(\delta\ge1\) any sample size serves. The rule itself does not use \(\delta\), which only determines the sample size, and this size does not depend on \(\mu\).

\proofstep{The case \(s=0\).} The definition of \(\tau\) above requires \(s\ge1\), so we argue directly. The error sets \(D_c\) are pairwise disjoint. Output \(g\) unless an error of the background is observed, in which case output the unique concept whose error set contains the point of the first such error; if there is no such concept, output \(g\). On a realizable sample from \(\mu\) that concept is \(c_\mu\). For a realizable \(\mu\), the list is \(\{g,c_\mu\}\) if \(\operatorname{loss}_\mu(g)\le\varepsilon\) and \(\{c_\mu\}\) otherwise. In the latter case a sample of size \(t\ge\ln(1/\delta)/\varepsilon\) contains an error of the background with probability at least \(1-(1-\varepsilon)^t\ge1-e^{-t\varepsilon}\ge1-\delta\). This gives the bound \(s+2=2\).

\proofstep{Sharpness.} Take a negative anchor and the full class on \(s+1\) free coordinates, with the all-negative background \(g\equiv-1\). The positive supports of two distinct concepts are distinct subsets of the \(s+1\) free coordinates and intersect in at most \(s\) points. Theorem~\ref{res:5.1}, applied after negating the anchor column, gives list replicability \(s+2\). Negating a column does not change the list number, since it conjugates rules, distributions and hypotheses by the same flip.
\end{proof}

Two consequences follow. For positive supports with pairwise intersections at most one, the all-negative background gives
\[
\operatorname{lr}\le3.
\]
For an arbitrary matrix, a random background gives
\[
\operatorname{lr}(B)
\le
\left\lfloor
M/2+\sqrt{M\ln N}
\right\rfloor+2,
\]
which is close to half of the elementary bound \(M\) when \(\ln N\) is small compared with \(M\). To prove the second bound let \(B\) have \(N\ge2\) distinct rows and \(M\) columns, and take \(g\) uniform. For distinct rows \(c,d\), the set \(D_c\cap D_d\) consists of those coordinates where \(c\) and \(d\) agree and \(g\) differs from both, so \(|D_c\cap D_d|\) is binomial with \(a\le M\) trials and parameter \(1/2\). Hoeffding's inequality with \(u=\sqrt{M\ln N}\) gives
\[
\mathbb P\bigl[|D_c\cap D_d|\ge M/2+u\bigr]\le e^{-2u^2/M}=N^{-2},
\]
and the union bound over fewer than \(N^2/2\) pairs gives a background with \(|D_c\cap D_d|<M/2+u\) for all pairs. Theorem~\ref{res:5.2} applies with \(s=\lfloor M/2+u\rfloor\). For \(N=1\) the list number is one.

\subsection{Anchoring}\label{ssec:5.3}

A unanimous column carries no information about the concept, yet by Theorem~\ref{res:5.1} it raises the list number of the cube from \(k\) to \(k+1\). We show that this is the largest possible effect: for the supremum value, and at a fixed error after halving it, any number of unanimous columns raises the list number by at most one. This inequality is an ingredient of Theorem~\ref{thm:anchor-width}, which identifies the list number of an anchored class with a geometric invariant.

For a class \(C\), let \(C^{\mathrm{anc}}\) be obtained by adding any number of unanimous columns.

\begin{resultenv}{Theorem}{res:5.3}{Anchoring}

(i) In the supremum sense,
\[
\operatorname{lr}(C^{\mathrm{anc}})
\le
\operatorname{lr}(C)+1.
\]

(ii) For every \(\varepsilon\in(0,1/2)\) and every
\[
\varepsilon'\in(0,\varepsilon/2],
\]
\[
\operatorname{lr}(C^{\mathrm{anc}},\varepsilon)
\le
\operatorname{lr}(C,\varepsilon')+1.
\]

(iii) At the level of finite closed covers, for every $\varepsilon>0$,
\[
\mathrm{LR}(C^{\mathrm{anc}},\varepsilon)
\le \mathrm{LR}(C,\varepsilon)+1.
\]

The restriction inequality
\[
\operatorname{lr}(C,\varepsilon)
\le
\operatorname{lr}(C^{\mathrm{anc}},\varepsilon)
\]
holds as well.
\end{resultenv}

\begin{proof}
\proofstep{The anchored cover.} Write a realizable anchored distribution as
\[
\mu_t=t\delta_{\mathrm{anchor}}+(1-t)\nu;
\]
for several anchor columns, \(t\) is their total mass and \(\delta_{\mathrm{anchor}}\) denotes the corresponding normalized anchor distribution. For \(t<1\), the distribution \(\nu\) is the normalized restriction of \(\mu_t\) to the original columns, it lies in \(\Delta_C\), and it depends continuously on \(\mu_t\). All anchor labels of a realizable distribution equal the common values of the unanimous columns.

Take a closed \(\varepsilon'\)-loss cover \(\{F_h\}\) of \(\Delta_C\), that is, a finite closed cover with loss at most \(\varepsilon'\) on \(F_h\). Extend each \(h\) correctly on the anchors, and use
\[
G_h
=
\{\mu_t:t\le1-\varepsilon'/2,\ \nu\in F_h\},
\]
together with an anchor hypothesis, namely a fixed row of \(C^{\mathrm{anc}}\), on
\[
G_0=\{\mu_t:t\ge1-\varepsilon'\}.
\]
The first sets are closed because their parameter stays away from the cone apex: on the closed set \(\{t\le1-\varepsilon'/2\}\) the map \(\mu_t\mapsto\nu\) is continuous, as \(1-t\ge\varepsilon'/2\), and \(G_h\) is the preimage of \(F_h\). Together with \(G_0\), they cover the anchored space, since a point with \(t>1-\varepsilon'/2\) lies in \(G_0\). Their multiplicity is at most the original multiplicity plus one, since a point lies in \(G_h\) only if its \(\nu\) lies in \(F_h\). If the anchor hypothesis coincides with an extended hypothesis, we merge the two sets, which does not increase the multiplicity.

\proofstep{The losses and parts (ii) and (i).} On \(G_h\), the extended hypothesis has loss at most
\[
(1-t)\varepsilon'\le\varepsilon'.
\]
On \(G_0\), the anchor hypothesis has loss at most
\[
1-t\le\varepsilon'.
\]
When \(\varepsilon'\le\varepsilon/2\), Proposition~\ref{prop:cover-rule} with \(\alpha=\tau=\varepsilon/2\) proves (ii): by Theorem~\ref{res:b} there is such a cover \(\{F_h\}\) of multiplicity at most \(\operatorname{lr}(C,\varepsilon')\). In particular, taking \(\varepsilon'=\varepsilon/2\) makes the thresholds explicitly
\[
t\le1-\varepsilon/4,
\qquad
t\ge1-\varepsilon/2.
\]
Taking the supremum proves (i): the number \(\operatorname{lr}(C^{\mathrm{anc}},\varepsilon)\) is nonincreasing in \(\varepsilon\), so its supremum is taken over \(\varepsilon\in(0,1/2)\), and each value is at most \(\operatorname{lr}(C,\varepsilon/2)+1\le\operatorname{lr}(C)+1\).

\proofstep{Part (iii) and the restriction inequality.} Using $\varepsilon'=\varepsilon$ in the same construction proves the closed-cover statement (iii): the anchored cover has loss at most \(\varepsilon\) and multiplicity at most the original one plus one. Finally, restriction to the original domain proves the lower inequality: a distribution realizable by \(C\) is realizable by \(C^{\mathrm{anc}}\), and restricting the outputs to the original columns does not change their losses on such distributions.
\end{proof}

By part (i), the effect of anchoring does not accumulate with the number of unanimous columns. Anchored cubes also show that contractibility of \(\Delta_C\) does not control list replicability: with an anchor column, \(\Delta_C\) is a cone with apex \(\delta_{\mathrm{anchor}}\), while the anchored cube has list number \(k+1\).

\subsection{An unbounded restricted-output separation}\label{ssec:5.4}

The learning rules above may output hypotheses outside the class. It is natural to ask how much this freedom gains, since the most direct rule outputs a concept consistent with the sample. The anchored singleton classes answer this question: their list number is two, while at small errors a rule with outputs in the signed class needs lists as large as the class.

Define \(B_N\), of size \(N\times(N+1)\), by
\[
c_i(0)=+,
\qquad
c_i(j)=+\ \Longleftrightarrow\ j=i.
\]
We write $C_N$ for the row class of $B_N$. Later sections use the same family with the size parameter named \(n\), and in particular its \(4\times5\) member \(B_4\).

The proper/improper separation for singleton classes appeared first
after Theorem~6 of~\cite{CMY23} (arXiv v2): for $s\ge3$
singletons the values are $s-1$ and $2$. Theorem~\ref{res:5.4} allows the larger output set $\pm C_N$ and gives the exact anchored
value $N$ throughout $\varepsilon<1/(N+1)$. The comparison of unrestricted rules
with rules that output linear classifiers in
\cite[Theorems C and D]{BHH26c} gives a bounded ratio.

\begin{resultenv}{Theorem}{res:5.4}{Anchored singletons}
For \(N\ge3\),
\[
\operatorname{lr}(B_N,\varepsilon)=2
\qquad(0<\varepsilon<1/2),
\]
\[
\operatorname{Ind}=\operatorname{coInd}=2.
\]
If the outputs are restricted to \(\pm C_N\), in the notation of Section~\ref{ssec:2.2}, then
\[
\operatorname{lr}_{\pm C_N}(C_N,\varepsilon)=N
\qquad(0<\varepsilon<1/(N+1)).
\]
\end{resultenv}

\begin{proof}
\proofstep{The unrestricted value and the index.} Relative to the background that is positive on the anchor and negative elsewhere, the error sets are disjoint singletons. Theorem~\ref{res:5.2} gives the upper bound \(2\). Theorem~\ref{res:6.6} with \(s=0\) gives index at most \(2\).

For the lower bound on the coindex, transpose and restrict to three concept coordinates. The anchor row and the singleton rows supply, after antipodal completion, the boundary of the three-dimensional cross-polytope: on the coordinates \(c_1,c_2,c_3\), the row \(0\) of the transpose is \((+,+,+)\) and the rows \(1,2,3\) are the three vectors with one positive entry, and with their negatives these are all eight patterns. So these three columns of the transpose are signed-shattered, and the transpose has coindex at least \(2\) (Section~\ref{ssec:2.1}). Theorem~\ref{res:2.1} transfers this to \(B_N\). Since \(\operatorname{coInd}\le\operatorname{Ind}\), the index and coindex both equal \(2\). Theorem~\ref{res:a} then forces list replicability at least \(2\), since \(2\le2\operatorname{lr}-1\).

\proofstep{The restricted lower bound: a map on a cone.} Use the cone of distributions
\[
\mu_{t,p}
=
t\delta_{(0,+)}
+
(1-t)\sum_jp_j\delta_{(j,-)},
\]
where \(p\) lies on the boundary of the probability simplex \(\Sigma\subset\mathbb R^N\). Such a distribution is realizable because at least one coordinate of \(p\) vanishes: if \(p_i=0\), then \(c_i\) has loss zero.

Restrict to the cone segment with
\[
t\ge1/(N+1).
\]
Suppose \(\operatorname{lr}_{\pm C_N}(C_N,\varepsilon)\le N-1\) for some \(\varepsilon<1/(N+1)\), and fix \(\eta\) with \(\varepsilon<\eta<1/(N+1)\). The learner-to-closed-cover direction (Proposition~\ref{prop:restricted-cover}) gives a finite closed cover of the realizable space, indexed by hypotheses in \(\pm C_N\), of multiplicity at most \(N-1\) and with loss at most \(\varepsilon\) on each member. Lemma~\ref{lem:thicken-cover} gives a partition of unity \(\{\varphi_h\}\) with at most \(N-1\) active labels at each point and loss below \(\eta\) for active labels.

On the cone segment every negative concept errs on the anchor, whose mass is at least \(1/(N+1)>\eta\), so \(\varphi_h=0\) there for every negative concept \(h\). Hence \(\sum_i\varphi_{c_i}=1\) on the segment. Since at most \(N-1\) labels are active at any point, \(f=(\varphi_{c_1},\ldots,\varphi_{c_N})\) is a continuous map from this cone segment into the boundary of the simplex of positive concept labels.

\proofstep{The restricted lower bound: a degree contradiction.} On the base of the segment, the concept \(c_i\) errs exactly at the atom \((i,-)\), so
\[
\operatorname{loss}_{\mu_{1/(N+1),p}}(c_i)
=
\frac{N}{N+1}p_i.
\]
An active label has loss less than \(\eta<1/(N+1)\), hence \(p_i<1/N\). The resulting boundary self-map \(f_0(p)=f(\mu_{1/(N+1),p})\) of \(\partial\Sigma\) has no fixed point: at a fixed point, every positive coordinate would be \(<1/N\), which is incompatible with their sum being one.

The absence of fixed points gives nonzero degree, whereas extension over the cone gives degree zero: \(\partial\Sigma\) is a sphere of dimension \(N-2\ge1\), a self-map of a sphere without fixed points has degree \((-1)^{N-1}\) \cite[Section 2.2]{Hatcher02}, and \((p,t)\mapsto f(\mu_{t,p})\), \(t\in[1/(N+1),1]\), is a homotopy from \(f_0\) to a constant map, because \(\mu_{1,p}=\delta_{(0,+)}\) does not depend on \(p\). This contradiction proves the lower bound \(N\).

\proofstep{The restricted upper bound.} Fix an order of the \(N\) concepts and let the learner output the first concept consistent with the sample; a realizable sample has a consistent concept, and if none exists the learner outputs the first concept. For a realizable \(\mu\) put \(L_\mu=\{c\in C_N:\operatorname{loss}_\mu(c)\le\varepsilon\}\); it contains the realizing concept, has at most \(N\) members, and all of them are accurate. A concept with \(\operatorname{loss}_\mu(c)>\varepsilon\) is consistent with \(n\) independent samples with probability at most \((1-\varepsilon)^n\le e^{-n\varepsilon}\), so by a union bound the output lies outside \(L_\mu\) with probability at most \(Ne^{-n\varepsilon}\le\delta\) once \(n\ge\ln(N/\delta)/\varepsilon\). The outputs lie in \(C_N\subseteq\pm C_N\), so \(\operatorname{lr}_{\pm C_N}(C_N,\varepsilon)\le N\).
\end{proof}

At the error
\[
\varepsilon_N=\frac1{4N},
\]
Theorem~\ref{res:5.4} rules out a uniform polylogarithmic conversion from unrestricted to restricted-output list replicability, even after doubling the target error: \(\operatorname{lr}(B_N,\varepsilon_N)=2\), and \(2\varepsilon_N=1/(2N)<1/(N+1)\) gives \(\operatorname{lr}_{\pm C_N}(C_N,2\varepsilon_N)=N\).

\subsection{Sylvester Hadamard matrices}\label{ssec:5.5}

The Sylvester Hadamard matrices have full real rank, so they test how far list replicability can lie below rank. Their signed rows form an affine family over \(\mathbb F_2\), and this keeps the chains of closures in the sign complex short. We determine the index and the list number exactly.

Let
\[
H_{2^m}(r,j)=(-1)^{r\cdot j},
\qquad
r,j\in\mathbb F_2^m.
\]
The bound of~\cite[Theorem 32]{FHV26} gives \(m-1\le\operatorname{Ind}S(H_{2^m})\le2m+1\). The next theorem sharpens it to an equality, through the affine-chain argument of~\cite[Lemma 31]{FHV26}, and it gives the list number at every fixed error.

\begin{resultenv}{Theorem}{res:5.5}{Exact values for Sylvester Hadamard matrices}
\[
\operatorname{Ind}S(H_{2^m})
=
\operatorname{coInd}S(H_{2^m})
=
m,
\]
and, for every \(\varepsilon\in(0,1/2)\),
\[
\operatorname{lr}(H_{2^m},\varepsilon)=m+1.
\]
\end{resultenv}

\begin{proof}
\proofstep{The coindex.} The signed rows shatter the columns
\[
0,e_1,\ldots,e_m,
\]
because \(s(-1)^{r\cdot j}\) has the value \(s\) at \(0\) and \(s(-1)^{r_i}\) at \(e_i\), and \((s,r)\) is arbitrary. This gives the coindex lower bound \(m\) (Section~\ref{ssec:2.1}).

\proofstep{The index.} A face specifies affine equations for a compatible signed row parameter in \(\mathbb F_2^{m+1}\): the signed row \((-1)^\sigma(-1)^{r\cdot j}\) is compatible with the vertex \(j^{(-1)^\beta}\) exactly when \(\sigma+r\cdot j=\beta\), and the linear part \((1,j)\) is nonzero. The compatible solution set is a proper affine subspace, is reversed under face inclusion, and its antipodal partner is its translate by \(e_0\). A closed face is determined by its compatible set, being the intersection of the facets of the compatible signed rows, so a strict chain of closed faces gives a strictly decreasing chain of affine subspaces, of dimensions between \(m\) and \(0\). Closure chains therefore have length at most \(m+1\), and Theorem~\ref{res:6.1} gives index at most \(m\). With \(\operatorname{coInd}\le\operatorname{Ind}\) both equal \(m\).

\proofstep{The list lower bound.} Restriction to columns \(0,e_1,\ldots,e_m\) is the anchored cube with \(m\) free coordinates, and a rule for a class restricts to a rule for the restriction of the class to these columns. The anchored-class identity of Theorem~\ref{res:5.1} gives the lower bound \(m+1\).

\proofstep{The list upper bound.} The upper bound is
\[
\operatorname{lr}\le H(C_{H_{2^m}})=h_+^0=m+1
\]
by \cite[Theorem 1.3]{BHHLT26} and Lemma~\ref{res:2.2} (alternatively by Theorem~\ref{res:6.3}, since \(\chi^*\le h_+^*\)); these heights are defined in Section~\ref{ssec:2.4}. To verify the value of the height, note that the rows compatible with a nonempty set of signed atoms form an affine subspace of the parameter space \(\mathbb F_2^{m}\), after fixing the sign by the column \(0\). Along a strict chain of positive closures these affine subspaces strictly shrink, so a chain has at most \(m+1\) members. The columns \(0,e_1,\ldots,e_m\) realize a chain of length \(m+1\), namely the closures of \(\{(0,+)\}\cup\{(e_l,+):l\le i\}\) for \(i=0,\ldots,m\). The unanimous column \(0\) makes every intersection of concept atom sets nonempty, so the two positive-closure conventions agree and \(h_+^*=h_+^0=m+1\).
\end{proof}

Thus the matrices \(H_{2^m}\) satisfy
\[
\operatorname{lr}=\operatorname{Ind}+1.
\]
The value \(m+1\) lies exponentially below the real rank of \(H_{2^m}\), which is \(2^m\), since \(H_{2^m}H_{2^m}^{\mathsf T}=2^mI\).

\subsection{Projective planes and the Fano plane}\label{ssec:5.6}

Projective planes are the model case \(s=1\) of Theorem~\ref{res:5.2}, and their sign rank and binary rank grow with the order of the plane. Theorem~\ref{res:5.6} sharpens~\cite[Theorems 1.2 and 3.2]{BHHLT26}, which give \(\operatorname{lr}\le3\) and \(\operatorname{Ind}\le5\): the list number equals three at every fixed error, and the index equals four for \(q\ge4\). Sign rank is therefore already unbounded on total classes of index four.

Let
\[
\Pi_q=2C-J
\]
where $C$ is the $0/1$ point--line incidence matrix and $J$ is the all-ones matrix. Thus $\Pi_q$ is the point--line sign matrix of the projective plane \(\operatorname{PG}(2,q)\) over the field with \(q\) elements, \(q\) a prime power. Its rows are indexed by lines and its columns by points. The switching binary rank \(r_2\) in the next theorem is defined in Theorem~\ref{res:6.1}.

\begin{resultenv}{Theorem}{res:5.6}{Projective-plane parameters}
For every prime power \(q\ge4\),
\[
\operatorname{Ind}(\Pi_q)
=
\operatorname{coInd}(\Pi_q)
=
4.
\]
For \(q=2\) (the Fano plane), \(\operatorname{Ind}(\Pi_2)=3\) by Proposition~\ref{res:5.7}; for \(q=3\), the index lies in \([3,4]\) and the coindex is at least \(3\).

For the projective-plane class,
\[
\operatorname{lr}(\Pi_q,\varepsilon)=3\qquad\text{for every }\varepsilon\in(0,1/2).
\]
The switching binary rank is
\[
r_2(\Pi_q)=q^2+q
\qquad(q\text{ odd}),
\]
and
\[
r_2(\Pi_q)=3^a
\qquad(q=2^a).
\]
In the \(q\ge4\) range,
\[
\operatorname{signrk}(\Pi_q)
\in
\left[
\max\left\{5,\left\lceil\frac{q+1}{\sqrt q}\right\rceil\right\},
2q+3
\right].
\]
\end{resultenv}

\begin{proof}
\proofstep{The upper bounds on the list number and the index.} Two lines meet in one point. Thus the error-intersection parameter is \(s=1\), for the all-negative background, whose error set for a line is the line itself. Theorems~\ref{res:5.2} and~\ref{res:6.6} give
\[
\operatorname{lr}\le3,
\qquad
\operatorname{Ind}\le4.
\]

\proofstep{The coindex for \(q\ge4\).} Take five points of a conic of \(\operatorname{PG}(2,q)\), which has \(q+1\) points, for instance \([1:t:t^2]\), \(t\in\mathbb F_q\), and \([0:0:1]\); no three of them are collinear, by the Vandermonde determinant. Every pair of them lies on a line meeting the five points exactly in that pair. Through each of them pass \(q+1-4\ge1\) lines meeting the five points only there, since \(q+1\) lines pass through a point and four of them contain a second one of the five points. Finally \(q^2+q+1-10-5(q-3)=q^2-4q+6=(q-2)^2+2>0\) lines miss all five.

Hence every sign pattern on these five columns with at most two positive entries is realized by a line, and every other pattern by a negated line, since such a pattern has at most two negative entries. So the five columns are signed-shattered, \(\partial\Diamond_5\subseteq S(\Pi_q)\), and the coindex is at least \(4\) (Section~\ref{ssec:2.1}). With \(\operatorname{coInd}\le\operatorname{Ind}\le4\) this proves the first display.

\proofstep{The cases \(q=3\) and \(q=2\).} For \(q=3\), the four points of a conic form an arc: each lies on \(q+1-3=1\) tangent, every pair lies on a secant, and \(q^2+q+1-6-4(q-2)=q^2-3q+3=3\) lines miss all four. Patterns with at most two positive entries are realized by lines and the others by negated lines. Thus these four columns are signed-shattered, so \(\operatorname{coInd}(\Pi_3)\ge3\); together with the upper bound \(4\) this gives the interval \([3,4]\). For \(q=2\), Proposition~\ref{res:5.7} gives the value \(3\).

\proofstep{The list lower bound.} Use the four lines
\[
z=0,\quad y=0,\quad x=0,\quad x+y+z=0.
\]
They give all four patterns on
\[
[1:0:0],\quad[0:1:0],
\]
namely \((+,+)\), \((+,-)\), \((-,+)\), \((-,-)\), and are all negative on
\[
[1:1:t],
\qquad
t\notin\{0,-2\},
\]
since the first line contains this point exactly when \(t=0\), the last exactly when \(t=-2\), and the other two never. Such a \(t\) exists: for \(q=2\) we have \(-2=0\), and for \(q\ge3\) the field has more than two elements. This restriction is an anchored cube with two free coordinates and a negative anchor; a rule for \(\Pi_q\) restricts to a rule for these four rows on these three columns, and negating that column does not change list replicability, so Theorem~\ref{res:5.1} gives \(\operatorname{lr}\ge3\).

\proofstep{The switching binary rank.} Put $n=q^2+q+1$. Each line has \(q+1\) points and two lines meet in one point, so \(CC^T=qI+J\) over the integers. For odd \(q\), use the identities
\[
CC^T\equiv I+J,
\qquad
C\mathbf1=C^T\mathbf1=0
\]
over \(\mathbb F_2\). For even \(q=2^a\), the classical incidence-rank identity is
\[
\operatorname{rank}_2C=3^a+1;
\]
this is the incidence-rank formula of~\cite{Hamada68,GM66}.

We first reduce both cases to the rank of a bilinear form. Since $n$ is odd,
$\mathbb F_2^n=\langle\mathbf1\rangle\oplus W$, where $W=\mathbf1^\perp$.
Switching a pivot row and column replaces the binary matrix by its
bilinear form on the basis $e_i+e_0$ of $W$, on both the row and
column sides: \(\Pi_q=(-1)^A\) with \(A=J+C\) over \(\mathbb F_2\), and with pivot \(0\) the matrix \(D\) of Theorem~\ref{res:6.1} has entries \(D_{ri}=(e_r+e_0)^TA(e_i+e_0)\).
The all-ones matrix vanishes on these subspaces, because \(u^TJv=(u^T\mathbf1)(\mathbf1^Tv)=0\) for \(u,v\in W\), so
$r_2(\Pi_q)$ is the rank of the restriction of $C$ to $W$.

When $q$ is odd, $CC^T=I+J$ has rank $n-1$: its kernel is $\langle\mathbf1\rangle$, because \((I+J)x=x+(\mathbf1^Tx)\mathbf1\) and \(n\) is odd. Also $C\mathbf1=C^T\mathbf1=0$;
therefore $\operatorname{rank}C=n-1$, with kernel exactly
$\langle\mathbf1\rangle$. For \(x=\alpha\mathbf1+u\), \(y=\beta\mathbf1+v\) with \(u,v\in W\), one has \(x^TCy=u^TCv\). Thus the restriction of $C$ to $W$ has rank $n-1$.

When $q$ is even, $C\mathbf1=C^T\mathbf1=\mathbf1$. Relative to
the decompositions above, $C$ is the direct sum of the identity
on $\langle\mathbf1\rangle$ and its restriction to $W$, since \(x^TCy=\alpha\beta+u^TCv\) for the same \(x,y\). The latter
therefore has rank $\operatorname{rank}C-1=3^a$, as claimed.

\proofstep{The sign rank.} Number the points \(1,\ldots,n\). For a line \(L\) put
\[
P_L(t)=1-2\prod_{j\in L}(t-j)^2,
\]
a polynomial of degree \(2(q+1)=2q+2\) with \(P_L(j)=1\) for \(j\in L\) and \(P_L(j)\le-1\) for \(j\notin L\), since the product is then a positive integer. Pairing its coefficient vector with the moment curve \((1,t,\ldots,t^{2q+2})\) therefore gives a sign representation of dimension \(2q+3\), so \(\operatorname{signrk}(\Pi_q)\le2q+3\), which is also the sign-change estimate of~\cite[Lemma 18]{AMY16}.

In the lower bound, \(\lceil(q+1)/\sqrt q\rceil\) is the \cite{AMY16} bound as used through \cite[Theorem 1.15]{BHHLT26}, which states \(\operatorname{signrk}\ge(q^2-1)/(\sqrt q(q-1))=(q+1)/\sqrt q\). The term \(5\) is \(\operatorname{Ind}+1\): a sign representation of dimension \(\operatorname{signrk}\) gives an odd map to \(S^{\operatorname{signrk}-1}\), as after Lemma~\ref{res:5.0}, so \(\operatorname{Ind}\le\operatorname{signrk}-1\), and \(\operatorname{Ind}=4\) for \(q\ge4\).
\end{proof}

For \(q=2\) the shattering argument used above is not available. The Fano matrix has no four signed-shattered columns, since it has fourteen signed rows and four columns carry sixteen patterns, so its lower bound requires a different argument. In the labeling below, the line \(x\) contains the point \(y\) exactly when \(x\cdot y=0\), so the matrix \(F\) of the next proposition is \(\Pi_2\).

\begin{resultenv}{Proposition}{res:5.7}{The Fano plane}
Let
\[
F_{xy}=(-1)^{x\cdot y},
\qquad
x,y\in\mathbb F_2^3\setminus\{0\}.
\]
Then
\[
\operatorname{Ind}S(F)=3=\operatorname{signrk}(F)-1.
\]
\end{resultenv}

\begin{proof}
\proofstep{A sign representation of dimension four.} An integer \(i\in\{1,\ldots,7\}\) denotes the vector of \(\mathbb F_2^3\) with binary expansion \(i\). The rank-four integer sign realization is as follows. In the column order
\[
(1,2,4,3,6,7,5),
\]
its signed moment-curve columns are
\[
\begin{aligned}
&(1,1,1,1),\quad(-1,-2,-4,-8),\quad(1,3,9,27),\\
&(-1,-4,-16,-64),\quad(1,5,25,125),\\
&(-1,-6,-36,-216),\quad(1,7,49,343).
\end{aligned}
\]
The corresponding row vectors are
\[
\begin{aligned}
&(-65,36,-4,0),\ (99,-40,4,0),\ (-21,20,-4,0),\\
&(429,-430,108,-8),\ (-819,542,-116,8),\\
&(385,-334,92,-8),\ (-135,174,-68,8).
\end{aligned}
\]
The rows are ordered by $x=1,2,\ldots,7$, with binary expansions in $\mathbb F_2^3$; the columns have the order displayed above. The inner products are
\[
\begin{pmatrix}
-33&9&7&-15&15&-7&-9\\
63&-35&15&-3&-1&-3&15\\
-5&-3&3&5&-21&45&-77\\
99&63&-105&75&-21&-9&-33\\
-385&135&-21&-5&-9&15&35\\
135&-21&-5&-9&15&35&-189\\
-21&-5&-9&15&35&-189&495
\end{pmatrix},
\]
and their signs are the entries of \(F\). This gives \(\operatorname{signrk}(F)\le4\), and the signed realization gives an odd map to \(S^3\), by the remark after Lemma~\ref{res:5.0}.

\proofstep{The first Stiefel--Whitney class of the quotient.} For the lower bound, work in the quotient \(S(F)/\mathbb Z_2\). Let \(w_1\in H^1(S(F)/\mathbb Z_2;\mathbb F_2)\) denote the Stiefel--Whitney class of the double cover \(S(F)\to S(F)/\mathbb Z_2\). On an edge \([i^s,j^t]\) with \(i<j\) it is represented by the sign-change indicator \((1-st)/2\bmod2\), and its cup powers \(w_1^k\) are computed by the Alexander--Whitney formula in column order. A simplex of the quotient is written with increasing column numbers and with a positive first sign, a vertex \(i^s\) being written \(si\); a face and its antipode give the same simplex.

The indicator represents \(w_1\): the lift starting at \(i^+\) of the quotient edge \([i^s,j^t]\) ends at \(j^{st}\), which lies on the vertex section \(i\mapsto i^+\) exactly when \(st=+1\). If \(w_1^k\ne0\) then \(\operatorname{Ind}\ge k\): the Stiefel--Whitney height \(\operatorname{swh}(X)=\max\{k:w_1^k\ne0\}\) of Section~\ref{ssec:2.1} satisfies \(\operatorname{coInd}(X)\le\operatorname{swh}(X)\le\operatorname{Ind}(X)\) (\cite{FHV26}).

It remains to show that \(w_1^3\ne0\). The chain complex of the quotient has
\[
\dim C_2=140,
\qquad
\dim C_3=224,
\qquad
\operatorname{rank}\partial_3=104,
\]
equivalently, the coboundary \(C^2\to C^3\) has rank \(104\), and rank \(105\) after adjoining \(w_1^3\); the boundary \(\partial_2\colon C_2\to C_1\) has rank \(35\). Since adjoining \(w_1^3\) raises the rank, it is not a coboundary. We now give an explicit cycle that detects this class and can be checked by hand.

\proofstep{A cycle detecting \(w_1^3\).} The following twelve tetrahedra form the mod-two cycle \(z\):
\[
\begin{aligned}
&(1,2,3,-4),\ (1,2,3,-5),\\
&(1,2,-3,4),\ (1,2,-3,5),\\
&(1,2,4,-5),\ (1,2,-4,5),\\
&(1,-2,3,-4),\ (1,-2,3,-5),\\
&(1,-2,-3,4),\ (1,-2,-3,5),\\
&(1,-2,4,-5),\ (1,-2,-4,5).
\end{aligned}
\]
They are simplices of \(S(F)\). With \(F_x\) the row of \(x\), the following signed rows, written on the columns \(1,\ldots,5\), contain the tetrahedra listed next to them:
\[
\begin{array}{ll}
F_4=(+,+,+,-,-):&(1,2,3,-4),\ (1,2,3,-5),\\
-F_7=(+,+,-,+,-):&(1,2,-3,4),\ (1,2,4,-5),\\
-F_3=(+,+,-,-,+):&(1,2,-3,5),\ (1,2,-4,5),\\
-F_1=(+,-,+,-,+):&(1,-2,3,-4),\ (1,-2,-4,5),\\
-F_5=(+,-,+,+,-):&(1,-2,3,-5),\ (1,-2,4,-5),\\
F_2=(+,-,-,+,+):&(1,-2,-3,4),\ (1,-2,-3,5).
\end{array}
\]

We claim that
\[
\partial z=0,
\qquad
\langle w_1^3,z\rangle=1.
\]
Both equalities can be read off the list. Let
\[
\gamma=\{(3,-4),(3,-5),(-3,4),(-3,5),(4,-5),(-4,5)\}.
\]
Each vertex \(\pm3,\pm4,\pm5\) lies on exactly two of them, and the antipode of an edge of \(\gamma\) lies in \(\gamma\). The twelve tetrahedra are \((1,s2,e)\) with \(s=\pm1\) and \(e\in\gamma\).

For the boundary, deleting the vertex \(1\) gives \((s2,e)\); for \(s=-1\) this is the simplex \((2,-e)\) of the quotient, so each simplex \((2,e)\) arises twice. Deleting \(s2\) gives \((1,e)\), which arises for both values of \(s\). Deleting a vertex of \(e\) gives \((1,s2,v)\), which arises from the two edges of \(\gamma\) at \(v\). So \(\partial z=0\) modulo two.

For the pairing, by the Alexander--Whitney formula, \(w_1^3\) takes on \((v_0,v_1,v_2,v_3)\) the product of the sign-change indicators of the three consecutive edges. Every edge of \(\gamma\) changes sign, the edge \((1,s2)\) changes sign exactly for \(s=-1\), and then \((-2,u)\) changes sign exactly for a positive \(u\). So the value one occurs exactly on \((1,-2,3,-4)\), \((1,-2,3,-5)\), \((1,-2,4,-5)\), and the pairing is \(3\equiv1\).

\proofstep{Conclusion.} Hence \(w_1^3\ne0\), so \(\operatorname{Ind}\ge3\). Together with the upper bound, this proves \(\operatorname{Ind}S(F)=3\). A sign representation of dimension \(\operatorname{signrk}(F)\) gives \(\operatorname{Ind}\le\operatorname{signrk}(F)-1\), so \(\operatorname{signrk}(F)=4\). This proves the proposition.
\end{proof}

The integer realization, the boundary calculation, and the cup-product pairing have also been verified by exact computation, together with \(\dim C_2\), \(\dim C_3\) and \(\operatorname{rank}\partial_3\), obtained by Gaussian elimination over \(\mathbb F_2\), and the fact that the cochain of \(w_1^3\) is not a coboundary (Appendix~\ref{sec:computation}).

\subsection{Thresholds, single flips, and transposition}\label{ssec:5.7}

We close this section by comparing list replicability with the index on small classes. Classes with the same list number can have different indices, as thresholds and anchored singletons show. The index is invariant under transposition by Theorem~\ref{res:2.1}, and it depends only on the sign complex. Small matrices show that list replicability has neither of these properties.

For the finite nontrivial threshold classes \(T_n\), \(n\ge2\) (Theorem~\ref{res:6.4}),
\[
\operatorname{lr}=2=\operatorname{Ind}+1,
\]
with a direct proof following Theorem~\ref{res:6.4}. The threshold learning result appeared first in~\cite{CMY23}.

The anchored singleton class $B_N$, $N\ge3$, has
\[
\operatorname{lr}=2=\operatorname{Ind}.
\]
Its \(4\times5\) member \(B_4\) also separates the tilt dimension from the covering dimension of the vote image in Section~\ref{ssec:7.8}.

The next proposition compares three small matrices: \(A\) and its transpose have different list numbers, and so do \(A^t\) and a matrix \(B'\) with the same sign complex. It also bounds the ratio between the list numbers of \(C\) and of the symmetric class \(C^\pm\), which have the same sign complex.

\begin{resultenv}{Proposition}{res:5.8}{The sign complex does not determine list replicability}
Let
\[
A=
\begin{bmatrix}
+&+&+&+\\
+&+&-&-\\
+&-&+&-
\end{bmatrix}.
\]
For every \(\varepsilon\in(0,1/2)\),
\[
\operatorname{lr}(A,\varepsilon)=2,
\qquad
\operatorname{lr}(A^t,\varepsilon)=3.
\]
Negating the last row of \(A^t\) gives \(B'\) with
\[
\operatorname{lr}(B',\varepsilon)=2,
\qquad
S(B')=S(A^t)=\partial\Diamond_3.
\]
Moreover, for every finite class \(C\),
\[
\operatorname{lr}(C)
\le
\operatorname{lr}(C^\pm)
\le
2\operatorname{lr}(C).
\]
\end{resultenv}

\begin{proof}
\proofstep{The matrices \(A\) and \(A^t\).} The transpose is the anchored cube with two free coordinates, with rows \((+,+,+)\), \((+,+,-)\), \((+,-,+)\), \((+,-,-)\), so Theorem~\ref{res:5.1} gives its value \(3\). These rows and their negatives are all eight sign vectors, so \(S(A^t)=\partial\Diamond_3\). Theorem~\ref{thm:baseline} with three concepts gives the upper bound \(2\) for \(A\), and the index lower bound gives equality: Theorem~\ref{res:2.1} with \(S(A^t)=\partial\Diamond_3\) gives \(\operatorname{Ind}(A)=2\), and Theorem~\ref{res:a} gives \(\operatorname{lr}\ge2\).

\proofstep{The matrix \(B'\).} After the indicated row negation, the class consists of the all-positive concept and three disjoint single flips, so Theorem~\ref{res:5.2} gives \(2\), with \(g\equiv+1\) and \(s=0\). Its unchanged sign complex again supplies the lower bound, since negating a row does not change the set of signed rows.

\proofstep{The inequalities for \(C^\pm\).} The final inequalities follow from restriction, since a rule for \(C^\pm\) is a rule for \(C\), the identity $\operatorname{lr}(-C)=\operatorname{lr}(C)$ obtained by negating labels and outputs, and the following
union inequality, applied to $C^\pm=C\cup(-C)$:
\begin{equation}\label{eq:union-list}
\operatorname{lr}(C_1\cup C_2)\le\operatorname{lr}(C_1)+\operatorname{lr}(C_2).
\end{equation}
To prove it with the required error margin, fix $\varepsilon>0$ and use
Theorem~\ref{res:b} to obtain closed covers of $\Delta_{C_i}$ at
error $\varepsilon/2$ and multiplicity at most $\operatorname{lr}(C_i)$.
Each $\Delta_{C_i}$ is closed in
$\Delta(C_1\cup C_2)=\Delta_{C_1}\cup\Delta_{C_2}$. Their combined
covers therefore form a closed cover of the union, of multiplicity
at most the sum. Combining sets with the same hypothesis label
preserves this bound. Proposition~\ref{prop:cover-rule} converts it
to error $\varepsilon$, with $\alpha=\tau=\varepsilon/2$; take the supremum over $\varepsilon>0$.
This is the supremum-error union bound also used in Section~\ref{ssec:8}; compare~\cite[Corollary 1.13]{BHHLT26}.
\end{proof}

The matrix \(A\) also illustrates a limitation of the profile map, which returns in Section~\ref{ssec:7}. The profile of a distribution records the loss of every concept, by the identity \(\Phi(\nu)_r=1-2\operatorname{loss}_\nu(c_r)\) of Section~\ref{ssec:2.5}, but it does not record the loss of a hypothesis outside the class. The improper hypothesis
\[
h=\operatorname{sign}(A^T\mathbf1)
\]
has losses \(1/2\) and \(0\) on the two distributions
\[
\mu=(1/2,0,0,1/2),
\qquad
\nu=(0,1/2,1/2,0),
\]
although both profiles are \((1,0,0)\). These distributions are on the atoms \((i,+)\), \(i=1,\ldots,4\), which the first row realizes. One has \(A^T\mathbf1=(3,1,1,-1)\), so \(h=(+,+,+,-)\), and the profiles of Section~\ref{ssec:2.5} are the half sums of the columns \(1,4\) and of the columns \(2,3\) of \(A\). Thus the profile does not in general determine improper-hypothesis loss.

\part{Face separation, topological criteria, and the index gap}\label{part:topology}
This part asks which geometric and topological data determine the list number.
Section~\ref{app:width} proves that the list number of an anchored class equals the
face width, a geometric invariant. Sections~\ref{app:literal}--\ref{app:stable} give
criteria in terms of literal complexes, a complete test in codimension two, a
limitation of coordinate thresholds, and a stabilized bound with a smaller
constant. Section~\ref{app:orientations} computes the face width of orientation
classes and proves Theorem~\ref{thm:index-gap}.
Sections~\ref{app:om} and~\ref{app:low-index} compare the list number with
oriented-matroid rank and with the index at the two ends of the dimension range.

\section{Face separation and anchoring}\label{app:width}
This section introduces the invariant against which the rest of this part is
measured. A rule of small error must predict correctly every coordinate that
carries non-negligible mass, and on the polytope $\operatorname{conv}C$ these
constraints appear as the coordinate faces. The face width $u(C)$ is one plus the
least dimension of a polyhedron through which $\operatorname{conv}C$ can be mapped
so that every coordinate face is retained. Theorem~\ref{thm:anchor-width} shows
that appending a single constant coordinate makes list replicability equal to
$u(C)$, and that without this coordinate the two numbers differ by at most one.
Whereas Theorem~\ref{thm:main} bounds list replicability by the size of the matrix
alone, this anchoring identity gives the exact value for an anchored class as a
geometric quantity, and subsequent sections of this part use it to pass between list
replicability and face width. The section closes with upper bounds on
$u(C)$ obtained by constructing models directly.

\subsection{A face-separating model}
We define the face width through maps and show in Lemma~\ref{lem:cover-model}
that it has an equivalent form in terms of open covers. The map form is the
convenient one for constructing upper bounds, and the cover form is the one
that connects to learning rules.

Let $P=\operatorname{conv}C\subseteq[-1,1]^n$. Define $u(C)$ to be one
plus the least dimension of a finite polyhedron $K$ for which there are
continuous maps $f:P\to K$ and $g:K\to[-1,1]^n$ satisfying
\[
 x_j=s\in\{\pm1\}\quad\Longrightarrow\quad(gf(x))_j=s.
\]
We call $u(C)$ the \emph{face width}, and a triple $(f,g,K)$ with this
property a \emph{model} of dimension $\dim K$. The sets
$P\cap\{x_j=s\}$ are the \emph{coordinate faces} of $P$; the two faces of
one coordinate are \emph{opposite}. A coordinate face may be empty.
The same definition applies to
polytopes in $[0,1]^n$ after the coordinate change $x\mapsto(x+1)/2$,
which preserves coordinate faces and model dimension.

A convex combination of rows with positive weights has $x_j=s$ exactly when
all these rows have the value $s$ at $j$. The condition therefore asks that a
coordinate on which all active rows agree can be recovered from the image of
the point in $K$. The identity map of $P$, with $g$ the inclusion, is a model,
so $u(C)\le\dim P+1\le n+1$.

\begin{lemma}[Covers and models]\label{lem:cover-model}
The number $u(C)$ equals the least multiplicity of a finite open cover of
$P$ none of whose members meets two opposite coordinate faces.
\end{lemma}
\begin{proof}
\proofstep{A cover gives a model.}
Let $\{U_i\}_{i\in I}$ be such a cover of multiplicity $\kappa$, with nonempty
members. Choose a partition of unity $(\varphi_i)$ with
$\{\varphi_i>0\}\subseteq U_i$, and let $N$ be the nerve of the cover: its
simplices are the sets $\sigma\subseteq I$ with
$\bigcap_{i\in\sigma}U_i\ne\varnothing$. Then $\dim N\le\kappa-1$.
The formula $f(x)=\sum_i\varphi_i(x)e_i$ defines a continuous map
$P\to|N|$, because all indices with $\varphi_i(x)>0$ have the point $x$ in
common. Give the vertex $i$ the label $\ell_i\in[-1,1]^n$ with
$(\ell_i)_j=s$ if $U_i$ meets the face $\{x_j=s\}$, and $(\ell_i)_j=0$ if
$U_i$ meets neither face of coordinate $j$. This is well defined because
$U_i$ misses at least one of the two opposite faces.
Let $g:|N|\to[-1,1]^n$ be the affine interpolation of the labels on each
simplex. If $x_j=s$ and $\varphi_i(x)>0$, then $U_i$ meets $\{x_j=s\}$, so
$(\ell_i)_j=s$. Therefore
$(gf(x))_j=\sum_i\varphi_i(x)(\ell_i)_j=s$, and $u(C)\le\dim N+1\le\kappa$.

\proofstep{A model gives a cover.}
Let $(f,g,K)$ be a model with $\dim K=u(C)-1$, and realize $K$ as a finite
simplicial complex in a Euclidean space. The sets
$A_j^s=f(P\cap\{x_j=s\})$ are compact. On $A_j^s$ the $j$th coordinate of
$g$ equals $s$, hence $A_j^+\cap A_j^-=\varnothing$. Let $\rho>0$ be smaller
than the distance between $A_j^+$ and $A_j^-$ for every $j$ for which both
are nonempty, and subdivide $K$ barycentrically until every simplex has
diameter less than $\rho/2$. The open stars of the vertices form an open
cover of $K$. A point lies in the open star of a vertex $w$ exactly when $w$
is a vertex of the simplex carrying the point in its relative interior, so
this cover has multiplicity at most $\dim K+1$. Every open star has diameter
less than $\rho$, so it misses $A_j^+$ or $A_j^-$ for each $j$. Pulling back
under $f$ and discarding empty members gives a finite open cover of $P$ of
multiplicity at most $\dim K+1=u(C)$ in which no member meets two opposite
coordinate faces.
\end{proof}

Learning rules produce closed covers through Theorem~\ref{res:b}, while
Lemma~\ref{lem:cover-model} requires open ones. The next lemma is the
compactness step used whenever a closed cover must be replaced by an open one.
\begin{lemma}[Thickening with the same intersection pattern]\label{lem:thickening}
Let $X$ be a compact metric space, and let $\mathcal F$ and $\mathcal W$ be
finite families of nonempty closed subsets of $X$. There is $\rho>0$ such
that the open sets $U_F=\{x\in X:\operatorname{dist}(x,F)<\rho\}$,
$F\in\mathcal F$, satisfy the following. If
$\bigcap_{F\in\mathcal G}U_F\ne\varnothing$ for a subfamily
$\mathcal G\subseteq\mathcal F$, then
$\bigcap_{F\in\mathcal G}F\ne\varnothing$; and if $F\cap W=\varnothing$ with
$W\in\mathcal W$, then $U_F\cap W=\varnothing$. In particular
$\{U_F\}$ has at most the multiplicity of $\mathcal F$.
\end{lemma}
\begin{proof}
For a subfamily $\mathcal G$ with empty intersection, the continuous function
$x\mapsto\max_{F\in\mathcal G}\operatorname{dist}(x,F)$ has no zero, since each
$F$ is closed. It therefore has a positive minimum $r_{\mathcal G}$ on $X$.
Disjoint members $F\in\mathcal F$, $W\in\mathcal W$ have
$\operatorname{dist}(F,W)>0$ by compactness. Choose $\rho$ smaller than these
finitely many positive numbers. A point of $\bigcap_{F\in\mathcal G}U_F$ with
$\bigcap_{F\in\mathcal G}F=\varnothing$ would give
$r_{\mathcal G}\le\max_{F\in\mathcal G}\operatorname{dist}(x,F)<\rho$, and a point of
$U_F\cap W$ would give $\operatorname{dist}(F,W)<\rho$. Both contradict the
choice of $\rho$.
\end{proof}

The main result of this section compares $\operatorname{lr}$ and $u$, with and
without a constant coordinate, and records the behavior of $u$ on products.
\begin{theorem}[Anchoring identity]\label{thm:anchor-width}
Let $C^+$ be obtained by appending a constant positive coordinate. Then
\[
 \operatorname{lr}(C)\le\operatorname{lr}(C^+)=u(C)
 \le\operatorname{lr}(C)+1.
\]
In addition,
\[
 u(C_1\times\cdots\times C_k)
 \le1+\sum_{a=1}^k(u(C_a)-1).
\]
\end{theorem}
The two directions of the identity $\operatorname{lr}(C^+)=u(C)$ use different
constructions. For the upper bound, Gibbs weights on the rows turn a realizable
distribution into a point of $\operatorname{conv}C$ close to every coordinate
face prescribed by a label of large mass, and a model then yields an accurate
cover. For the lower bound, we map $\operatorname{conv}C$ continuously into the
realizable distributions by putting mass on the coordinates that are close to a
Boolean value. The anchor absorbs the rest of the mass, so this map is defined
at every point, and this is where the constant coordinate is needed. The
constant coordinate can change the value: Theorem~\ref{res:5.1} gives list number
$k$ for the cube $\{\pm1\}^k$ and $k+1$ for the anchored cube, so the last
inequality of the theorem is attained.

The identity is used throughout this part in the following way. A model of $C$
of dimension $d$ gives $\operatorname{lr}(C^+)\le d+1$, and an obstruction to all
models of dimension at most $d$ gives $\operatorname{lr}(C^+)\ge d+2$. The exact
formulas of Section~\ref{app:orientations} are obtained in this way.

\begin{proof}
\proofstep{From a model to a learning rule: Gibbs weights.}
We show $\operatorname{lr}(C)\le u(C)$ for an arbitrary finite class $C$.
Fix a model $(f,g,K)$ of dimension $d=u(C)-1$ and a target error
$\varepsilon>0$. Set $\delta=\varepsilon/(2n)$.
Identify $\Delta_C$ with $X(C)$ as in Section~\ref{app:cube-map}.
For a realizable signed mass vector $x=p\odot c$, with $c\in C$ and
$p\in\Delta^{n-1}$, define Gibbs weights
$q_b(c')\propto\exp(b\langle c',x\rangle)$ on the rows $c'\in C$, and let
$v_b(x)=\sum_{c'}q_b(c')c'\in P$. The map $v_b:X(C)\to P$ is continuous.

Put $Z_b(x)=\sum_{c'}\exp(b\langle c',x\rangle)$. Multiplying
$\ln q_b(c')=b\langle c',x\rangle-\ln Z_b(x)$ by $q_b(c')$ and summing gives
the Gibbs variational identity
\[
 \ln Z_b(x)=b\langle v_b(x),x\rangle+H(q_b),
\]
where $H$ is Shannon entropy, so $H(q_b)\le\ln m$.
The realizing row has inner product
$\langle c,x\rangle=\sum_jp_jc_j^2=1$, so $Z_b(x)\ge e^b$. Hence
\[
 1-\langle v_b(x),x\rangle\le\ln m/b.
\]
Since $\sum_jp_j=1$, the left side equals
$\sum_jp_j(1-c_jv_{b,j})$, and every summand $p_j(1-c_jv_{b,j})$ is
nonnegative because $v_b(x)\in[-1,1]^n$. So on a coordinate
with $p_j\ge\delta$,
\[
 1-c_jv_{b,j}\le\frac{\ln m}{b\delta},
\]
uniformly in $x$.

\proofstep{A uniform margin for the model.}
There is $\eta>0$ such that every $y\in P$ and every literal $(j,s)$
with $sy_j\ge1-\eta$ satisfy $s(gf(y))_j>3/4$. Otherwise, since there are
finitely many literals, some literal $(j,s)$ admits a sequence $y_k\in P$ with
$sy_{k,j}\ge1-1/k$ and $s(gf(y_k))_j\le3/4$. A limit point $y\in P$ has
$y_j=s$ and $s(gf(y))_j\le3/4$, contrary to the model property.
Choose $b>0$ with $b\ge\ln m/(\delta\eta)$. For $x=p\odot c$ and every
coordinate with $p_j\ge\delta$ (a \emph{heavy} coordinate) we get
$c_jv_{b,j}(x)\ge1-\eta$, therefore
$c_j(gf(v_b(x)))_j>3/4$.

\proofstep{An accurate closed cover.}
By uniform continuity of $g$, triangulate $K$ so finely that $g$ oscillates
by less than $1/4$ in sup norm on each simplex. Give each vertex $w$ of this
triangulation the Boolean label $h_w$ with $(h_w)_j=+1$ if $g(w)_j\ge0$ and
$(h_w)_j=-1$ otherwise. Let $D_w$ be the region on which the barycentric
coordinate of $w$ is maximal, as in Lemma~\ref{res:2.0}, and pull back:
$R_w=(f\circ v_b)^{-1}(D_w)\subseteq X(C)$.
These sets are closed, cover $X(C)$, and have multiplicity at most $d+1$ by
Lemma~\ref{res:2.0}.
Let $x=p\odot c\in R_w$ and $y=f(v_b(x))\in D_w$. By Lemma~\ref{res:2.0}, $y$
lies in a closed simplex containing $w$, so
$|g(w)_j-g(y)_j|<1/4$ for every $j$. On a heavy coordinate this gives
$c_jg(w)_j>1/2$, so $(h_w)_j=c_j$. Thus $h_w$ errs only on coordinates with
$p_j<\delta$, and
\[
 \operatorname{loss}_\mu(h_w)=\sum_{j:(h_w)_jx_j<0}|x_j|\le n\delta=\varepsilon/2 .
\]
Therefore $\mathrm{LR}(C,\varepsilon/2)\le d+1$, and
Proposition~\ref{prop:cover-rule} with $\alpha=\tau=\varepsilon/2$ gives
$\operatorname{lr}(C,\varepsilon)\le d+1$. Since $\varepsilon>0$ was arbitrary,
$\operatorname{lr}(C)\le u(C)$.

\proofstep{The anchored class.}
Appending a constant coordinate does not increase the face width.
Indeed $\operatorname{conv}C^+=P\times\{1\}$; the maps
$(x,1)\mapsto f(x)$ and $y\mapsto(g(y),1)$ form a model of $C^+$ on the same
$K$, because the face $x_{n+1}=-1$ is empty and $x_{n+1}=1$ is preserved. So
$u(C^+)\le u(C)$. The inequality just proved, applied to the class $C^+$,
gives $\operatorname{lr}(C^+)\le u(C^+)\le u(C)$.

\proofstep{From an anchored rule to a model: a map into realizable distributions.}
Write $L=\operatorname{lr}(C^+)$, which is finite by the preceding inequality.
It is convenient to write $P$ in $[0,1]^n$ for this step; the literal $(j,+1)$
then corresponds to the Boolean value $1$ of coordinate $j$, and $(j,-1)$ to
the value $0$. Choose $0<\delta_0<\min\{1/2,1/n\}$. For $z\in P$, put mass
\[
 w_j(z)=\frac1n\max\{0,1-\operatorname{dist}(z_j,\{0,1\})/\delta_0\}
\]
on the closer Boolean value of coordinate $j$, and put the rest of the
mass, $1-\sum_jw_j(z)\ge0$, on the positive anchor.
Let $\Phi(z)\in\mathbb R^{n+1}$ be the signed mass vector of this distribution.
Its coordinates are
\[
 \Phi_j(z)=\frac1n\Bigl(\max\{0,1-(1-z_j)/\delta_0\}
                       -\max\{0,1-z_j/\delta_0\}\Bigr)\quad(j\le n)
\]
and $\Phi_{n+1}(z)=1-\sum_{j\le n}|\Phi_j(z)|$. Since $\delta_0<1/2$, at most
one of the two terms in $\Phi_j(z)$ is nonzero, and at a midpoint
$z_j=1/2$ both vanish, so the mass is zero there. These formulas define a
continuous map $\Phi:P\to\mathbb R^{n+1}$.

\proofstep{The map lands in $X(C^+)$.}
Represent $z$ as an average of rows and draw a random
row $c$ from this representation, so $(c_j+1)/2$ is a Boolean variable of mean
$z_j$. Let $W$ be the set of coordinates with $w_j(z)>0$, and for $j\in W$ let
$b_j$ be the closer Boolean value. Then $|z_j-b_j|<\delta_0$, and the
literal $(j,b_j)$ is violated by the random row with probability exactly
$|z_j-b_j|<\delta_0$. If $W\ne\varnothing$, a union bound over the at most $n$
literals gives failure probability less than $n\delta_0<1$. So one row
satisfies all literals of positive mass; if $W=\varnothing$, every row does.
Together with the anchor this row
realizes the distribution, that is, $\Phi(z)\in X(C^+)$.

\proofstep{A face-separating cover.}
On the coordinate face $z_j=s$, with $s\in\{0,1\}$, the distance vanishes and
the corresponding literal has mass $1/n$.
Fix $0<\varepsilon<1/n$. Theorem~\ref{res:b} gives
$\mathrm{LR}(C^+,\varepsilon)\le\operatorname{lr}(C^+,\varepsilon)\le L$, so
$X(C^+)$ has a finite closed cover $\{R_i\}$ of multiplicity at most $L$, with
labels $h_i$ of loss at most $\varepsilon$ on $R_i$.
The nonempty sets $F_i=\Phi^{-1}(R_i)$ form a finite closed cover of $P$ of
multiplicity at most $L$. If $F_i$ meets the face $z_j=s$, then $h_i$ agrees
with the corresponding literal, since otherwise its loss at that point would be
at least $1/n>\varepsilon$. Therefore no $F_i$ meets two opposite coordinate
faces.

Apply Lemma~\ref{lem:thickening} with $X=P$, $\mathcal F=\{F_i\}$, and
$\mathcal W$ the family of nonempty coordinate faces. The enlarged members
$U_{F_i}$ form a finite open cover of $P$ of multiplicity at most $L$, and each
of them misses every face that $F_i$ misses, so none meets two opposite
coordinate faces. Lemma~\ref{lem:cover-model} yields
$u(C)\le\operatorname{lr}(C^+)$. Together with the inequality
$\operatorname{lr}(C^+)\le u(C)$ proved above, this gives
$\operatorname{lr}(C^+)=u(C)$.

\proofstep{The two outer inequalities.}
Restriction of a rule gives the first inequality in the theorem: a rule for
$C^+$ receives samples from any $\mu\in\Delta_C\subseteq\Delta_{C^+}$, and
deleting the anchor from its outputs and from its lists does not increase the
list sizes and does not change the losses.
Theorem~\ref{res:5.3}(i), with the single unanimous column of $C^+$, gives
$\operatorname{lr}(C^+)\le\operatorname{lr}(C)+1$, which is the last inequality.

\proofstep{Products.}
Let $P_a=\operatorname{conv}C_a$ and $C=\prod_aC_a$. Then
$\operatorname{conv}C=\prod_aP_a$: the product is convex and contains $C$, and a
point $(x_1,\ldots,x_k)$ with $x_a=\sum_c\lambda^a_cc$ is the mean of the
product distribution $\bigotimes_a\lambda^a$ on $C$.
Take models $(f_a,g_a,K_a)$ of dimensions $u(C_a)-1$, and put $f=\prod_af_a$,
$g=\prod_ag_a$, $K=\prod_aK_a$. The product is a finite polyhedron of dimension
$\sum_a\dim K_a$, since a product of simplices is a convex polytope.
Every coordinate of $C$ belongs to exactly one factor, and on that factor the
model $(f_a,g_a,K_a)$ preserves its faces; hence $(f,g,K)$ preserves every
coordinate face. This proves the product bound.
\end{proof}

\subsection{Random thresholds and projected volume}
A model can also be built with no reference to learning. We now give the
simplest such construction, a map acting on each coordinate separately. A ramp
around a threshold $t_j$ pushes the coordinate $x_j$ to one of two values unless
$x_j$ is close to $t_j$, so the image of a set lies in a low skeleton as soon as
no point of the set is close to the thresholds in many coordinates. Whether such
thresholds exist is a question about the volumes of coordinate projections. We
first state the estimate for an arbitrary compact set and at an arbitrary scale.
\begin{proposition}[Grid avoidance from projected volumes]\label{prop:coordinate-volume}
Let $X\subseteq\mathbb R^n$ be compact, $1\le k\le n$, and $L_0>0$.
If
\[
 \sum_{|J|=k}\operatorname{vol}_k(\pi_JX)<L_0^k,
\]
then $X$ maps continuously to a $(k-1)$-dimensional polyhedron with
every fiber of $\ell_\infty$-diameter less than $L_0$.
\end{proposition}
\begin{proof}
\proofstep{A grid avoiding $X$.}
We may assume $X\ne\varnothing$.
Choose $t$ uniformly modulo $L_0$ in each coordinate. For fixed $J$ with
$|J|=k$, let $B_J$ be the event that some $x\in X$ has
$x_j\in t_j+L_0\mathbb Z$ for all $j\in J$. This says that $t_J$ belongs to the
compact image of $\pi_JX$ in the torus $(\mathbb R/L_0\mathbb Z)^k$. The quotient
map does not increase Lebesgue measure, and the torus has volume $L_0^k$, so
\[
 \Pr(B_J)\le\operatorname{vol}_k(\pi_JX)/L_0^k.
\]
The strict sum bound gives a choice of $t$ outside every $B_J$, for which no
point of $X$ has $k$ coordinates on grid hyperplanes $x_j\in t_j+L_0\mathbb Z$.

\proofstep{The staircase map.}
Let $\gamma_j(x)=\operatorname{dist}(x_j,t_j+L_0\mathbb Z)\le L_0/2$.
The $k$th smallest of the numbers $\gamma_1(x),\ldots,\gamma_n(x)$ is a
continuous function of $x$, and it is positive on the compact set $X$, so it
has a positive minimum $\eta\le L_0/2$. Fix $0<w<\eta/2$. Around each grid value
$t_j+L_0\ell$ of coordinate $j$ take the ramp
$[t_j+L_0\ell-w,\,t_j+L_0\ell+w]$; since $2w<L_0$, the ramps of one coordinate
are pairwise disjoint.
In each coordinate use the continuous nondecreasing staircase
$\sigma_j:\mathbb R\to\mathbb R$ that equals the integer $\ell$ on the flat interval
$[t_j+L_0(\ell-1)+w,\,t_j+L_0\ell-w]$ and is linear from $\ell$ to $\ell+1$
across the ramp about $t_j+L_0\ell$.
The value $\sigma_j(x_j)$ is noninteger only if $x_j$ is interior to a ramp,
which forces $\gamma_j(x)<w<\eta$. By the choice of $\eta$ this happens for at
most $k-1$ coordinates at every point of $X$. The map
$\sigma=(\sigma_j)_j$ therefore has at most $k-1$ noninteger coordinates at
every point, hence lands in the $(k-1)$-skeleton of the integer cubical grid.
Its compact image meets finitely many grid cells; since $k\le n$, each lies in
a closed $(k-1)$-cell of the grid. The union of these $(k-1)$-cells is the
required polyhedron.

\proofstep{The fibers.}
Let $\sigma(x)=\sigma(x')$ and fix $j$. If the common value of
$\sigma_j$ is noninteger, then both points lie on one ramp, where $\sigma_j$ is
strictly increasing, so $x_j=x'_j$. If it is an integer, both coordinates lie in
one flat interval, of length $L_0-2w$. Thus $\|x-x'\|_\infty\le L_0-2w<L_0$ on
every fiber.
\end{proof}

For a polytope in $[0,1]^n$ the relevant scale is $L_0=1$: opposite coordinate
faces are at $\ell_\infty$-distance one, so a fiber of $\ell_\infty$-diameter
less than one cannot meet both of them. At this scale one threshold in each
coordinate suffices, the staircase fixes the coordinate faces, and the map is
itself a model; the proof below carries out this special case directly. Bounding
the projected volumes of $\operatorname{conv}C$ by an
empirical-mean argument gives the following bound on the face width.
\begin{theorem}[Projected-volume bound]\label{thm:volume-bound}
For every $C\subseteq\{\pm1\}^n$ with $m=|C|\ge2$,
\[
 \operatorname{lr}(C)\le u(C)\le1+7.76(n^2\ln m)^{1/3}.
\]
\end{theorem}
The proof is elementary and uses no topology beyond the definition of a model.
The bound lies below the bound $u(C)\le n+1$ of the identity model when
$\ln m<n/7.76^3$. It lies below the bound
$u(C)=\operatorname{lr}(C^+)\le805\ln(2m)\ln(2n+2)$ of
Theorems~\ref{thm:main} and~\ref{thm:anchor-width} when
$n\ge2$ and $\ln m\ge n/(1000\ln^{3/2}n)$. When both conditions hold, that is, for
classes of size close to exponential in $n$, it improves on both.

\begin{proof}
\proofstep{The parameter $k$.}
Apply $x\mapsto(x+1)/2$ and work in $[0,1]^n$; this leaves $u(C)$
unchanged. Set $k=\lfloor7.76(n^2\ln m)^{1/3}\rfloor+1$, so that
$7.76(n^2\ln m)^{1/3}<k\le1+7.76(n^2\ln m)^{1/3}$.
If $k>n$, the identity model $K=P$ of $P$, with $g$ the inclusion, gives
$u(C)\le\dim P+1\le n+1\le k$.
Assume $k\le n$. Since $m\le2^n$, we have $\ln m\le n\ln2$, so
$(n^2\ln m)^{1/3}\ge\ln m/(\ln2)^{2/3}$; as $7.76>2(\ln2)^{2/3}$, the chosen $k$
satisfies $k\ge2\ln m$.

\proofstep{Volume of one projection.}
Let $J\subseteq[n]$ with $|J|=k$. Then $P'=\pi_JP$ is the convex hull of at most
$m$ points of $[0,1]^k$. Let $s=\lceil k/(2\ln m)\rceil\ge1$.
Fix $y\in P'$ and a representing distribution on these points. The empirical
mean of $s$ independent vertices drawn from it has mean squared Euclidean error
$s^{-1}\sum_{i\le k}\operatorname{Var}_i\le k/(4s)$, because each coordinate is a
$[0,1]$-valued variable and has variance at most $1/4$.
There is thus one of the at most $m^s$ empirical means of ordered $s$-tuples
within $\sqrt{k/(4s)}$ of $y$, so $P'$ is covered by at most $m^s$ balls of this
radius. Let $\omega_k=\pi^{k/2}/\Gamma(k/2+1)$ be the volume of the Euclidean
unit ball. Stirling's lower bound $\Gamma(x+1)\ge\sqrt{2\pi x}\,(x/e)^x$, with $x=k/2\ge1/2$, gives
$\omega_k\le(2\pi e/k)^{k/2}$. Now $s\ge k/(2\ln m)$ means
$1/(2s)\le\ln m/k$, and
$s\ln m\le(k/(2\ln m)+1)\ln m=k/2+\ln m\le k$ means $m^s\le e^k$. Hence
\begin{align*}
 \operatorname{vol}_k(P')
 &\le m^s\omega_k(k/(4s))^{k/2}
  \le e^k\Bigl(\frac{2\pi e}{k}\cdot\frac{k}{4s}\Bigr)^{k/2}
  =e^k\Bigl(\pi e\cdot\frac1{2s}\Bigr)^{k/2}\\
 &\le e^k\Bigl(\frac{\pi e\ln m}{k}\Bigr)^{k/2}
  =(\pi e^3\ln m/k)^{k/2}.
\end{align*}

\proofstep{Summation.}
There are $\binom nk\le(en/k)^k$ sets $J$, so
\[
 \sum_{|J|=k}\operatorname{vol}_k(\pi_JP)
 \le\Bigl(\frac{en}{k}\Bigr)^k\Bigl(\frac{\pi e^3\ln m}{k}\Bigr)^{k/2}
 =\left(\frac{e\sqrt{\pi e^3}\,n\sqrt{\ln m}}{k^{3/2}}\right)^k<1.
\]
The last inequality is equivalent to
$k>(e\sqrt{\pi e^3})^{2/3}(n^2\ln m)^{1/3}$, and it holds since
$(e\sqrt{\pi e^3})^{2/3}=e^{5/3}\pi^{1/3}<7.76$.

\proofstep{The model.}
The set of $t\in[0,1]^n$ with $t_J\in\pi_JP$ for some $|J|=k$ has Lebesgue measure
at most the displayed sum, hence less than one.
Choose $t\in(0,1)^n$ with $t_J\notin\pi_JP$ for all $|J|=k$; thus no point of $P$
agrees with $t$ on $k$ coordinates.
The $k$th smallest of the numbers $|z_j-t_j|$ is continuous and positive on the
compact set $P$; let $\eta>0$ be its minimum, and choose
$0<\rho<\min\{\eta,\min_jt_j,\min_j(1-t_j)\}$.
The map $G:P\to[0,1]^n$ with coordinates
\[
 G_j(z)=\max\{0,\min\{1,(z_j-t_j+\rho)/(2\rho)\}\}
\]
has the following properties. If $z_j=0$, then $(-t_j+\rho)/(2\rho)<0$, so
$G_j(z)=0$. If $z_j=1$, then $(1-t_j+\rho)/(2\rho)>1$, so $G_j(z)=1$. And
$G_j(z)\notin\{0,1\}$ forces $|z_j-t_j|<\rho<\eta$, which holds for at most $k-1$
coordinates at any point of $P$. So $G$ lands in the $(k-1)$-skeleton of the cube
and fixes its coordinate faces.
Since $k\le n$, this skeleton is a finite polyhedron of dimension $k-1$, and with
$g$ the inclusion, $G$ is a model of this dimension. So $u(C)\le k$.

In both cases $u(C)\le k\le1+7.76(n^2\ln m)^{1/3}$.
Theorem~\ref{thm:anchor-width} gives
$\operatorname{lr}(C)\le\operatorname{lr}(C^+)=u(C)$.
\end{proof}

\subsection{Completions and bounded Hamming perturbations}
The model of Theorem~\ref{thm:volume-bound} does not use the combinatorics of the
class. The two constructions below choose the target of the model according to
this combinatorics, and they build the map simplex by simplex over a
triangulation of the polytope. A simplex is sent into a set determined by the
literals that its vertices have in common. A face of a simplex has at least as
many common literals, and this makes the sets nested in the way that an
inductive extension requires.

\paragraph{Cubical completions.}\label{par:carrier-extension}
Let $H\subseteq\{\pm1\}^n$.
Write $\operatorname{Cub}(H)$ for the union of closed cube faces all
of whose vertices lie in $H$. A \emph{coordinate section} of
$\operatorname{Cub}(H)$ is a set $\operatorname{Cub}(H)\cap Q_S$ for a consistent
literal set $S$, in the notation of Section~\ref{app:cube-map}, with $S=\varnothing$
allowed. Call $H$ an \emph{ample completion} of
$C$ when $C\subseteq H$ and every nonempty coordinate section of
$\operatorname{Cub}(H)$ is contractible. The following face-preserving carrier
extension shows that $\operatorname{Cub}(H)$ is then the target of a model.
\begin{proposition}[Carrier extension]\label{prop:carrier-extension}
If $H$ is an ample completion of $C$, then
\[
 u(C)\le\dim\operatorname{Cub}(H)+1.
\]
\end{proposition}
\begin{proof}
Take a pulling triangulation of $\operatorname{conv}H$ (Lemma~\ref{lem:pulling}) with
vertices in $H$, compatible with its coordinate faces. For a simplex $\sigma$,
let $S(\sigma)$ be the set of literals satisfied by all its vertices, so that
$Q_{S(\sigma)}$ is its smallest coordinate face. The section
$\operatorname{Cub}(H)\cap Q_{S(\sigma)}$ contains the vertices of $\sigma$, which
are zero-dimensional cube faces with vertices in $H$, and is therefore nonempty
and contractible. If $\tau$ is a face of $\sigma$, then
$S(\tau)\supseteq S(\sigma)$, so sections are nested under passage to faces.

Starting at the vertices, where the map is the identity, extend over successive
simplices: given a map on $\partial\sigma$ with
each face $\tau$ mapped into the section of $\tau$, the image of $\partial\sigma$
lies in the contractible section of $\sigma$, so the map extends over $\sigma$
inside this section. This defines a continuous map
$G:\operatorname{conv}H\to\operatorname{Cub}(H)$. If $x_j=s$ and $\sigma$ carries $x$
in its relative interior, then $x$ is a positive combination of the vertices of
$\sigma$, so $(j,s)\in S(\sigma)$ and $G(x)_j=s$.
Restriction to $\operatorname{conv}C$, with the inclusion of the finite polyhedron
$\operatorname{Cub}(H)$ as $g$, gives the required model.
\end{proof}
Theorem~\ref{thm:graph-budget} obtains its upper bound for orientation classes
from this proposition, and Proposition~\ref{prop:cyclic-intervals} uses the same
inductive extension with another target whose sections are contractible.

\paragraph{Hamming neighborhoods of affine subspaces.}
The second construction concerns classes close to a binary linear code. If $C$
lies in a binary affine subspace of dimension $d$, each simplex of a triangulation
of $\operatorname{conv}C$ can be labeled by a point of the affine span of its
vertices, and along a chain of simplices these spans form a flag of length at most
$d+1$. We allow each row to differ from the subspace in up to $r$ coordinates. The
error supports of all rows together may be large, but along one chain at most
$d+1$ rows are needed to span, so only $r(d+1)$ coordinates have to be treated
separately.
\begin{proposition}[Hamming neighborhoods of affine subspaces]\label{prop:affine-neighborhood}
Suppose every row of $C$ is at Hamming distance at most $r$ from one binary
affine subspace of dimension $d$. Then
\[
 u(C)\le(r+1)(d+1).
\]
\end{proposition}
The bound is attained at both extremes, because $\operatorname{lr}\le u$. For
$r=0$, the rows of the Sylvester Hadamard matrix $H_{2^m}$ form a linear subspace of dimension
$m$, and its list number is $m+1$ by Theorem~\ref{res:5.5}. For $d=0$, the
Hamming ball $E_{M,k}$ of Theorem~\ref{thm:hamming-balls} lies within distance $k$
of its all-negative row, and its list number is $k+1$.

\begin{proof}
\proofstep{The labels.}
Work in $[0,1]^n$ and identify $\{0,1\}^n$ with $\mathbb F_2^n$; let
$\mathcal A\subseteq\mathbb F_2^n$ be the given affine subspace.
Choose, for each row $c$, a point $\pi(c)\in\mathcal A$ at distance at
most $r$. Take a pulling triangulation of $P=\operatorname{conv}C$ using
its Boolean vertices, followed by barycentric subdivision.
For each binary affine subspace $A$ arising as the span
$\operatorname{aff}_{\mathbb F_2}\pi(F)$ of the images of the vertices of a nonempty
face $F$ of the triangulation,
fix one point $\rho(A)\in A$, depending only on $A$. For a nonempty
face $F$, start from $\rho(\operatorname{aff}_{\mathbb F_2}\pi(F))$.
Correct this representative on
all coordinates unanimous on the vertices of $F$, that is, set each such
coordinate to the common value, and use the corrected
Boolean vector $\ell(F)$ as the label of the barycenter of $F$.
Interpolate affinely over every barycentric simplex. The label depends only on
$F$, so these affine maps agree on intersections and define a global
continuous piecewise-linear map $G:P\to[0,1]^n$.

\proofstep{Extreme coordinates are preserved.}
Let $z_j=s\in\{0,1\}$, and let the
barycentric simplex of a chain $F_1\subsetneq\cdots\subsetneq F_t$ carry $z$ in its
relative interior. Then $z$ lies in the relative interior of $F_t$, so all vertices
of $F_t$ have $j$th coordinate $s$. Coordinate $j$ is thus unanimous with value $s$
on each $F_i$, all labels on the simplex have the prescribed bit, and $G(z)_j=s$.

\proofstep{Few error coordinates along a chain.}
Consider a chain $F_1\subsetneq\cdots\subsetneq F_t$.
The sets $\pi(F_i)$ increase with $i$. Choose affinely independent subsets
$B_i\subseteq\pi(F_i)$ over $\mathbb F_2$, spanning
$\operatorname{aff}_{\mathbb F_2}\pi(F_i)$ and nested by retaining a basis
and extending it as necessary. For $b\in B_t$, let $i$ be the first index with
$b\in B_i$, and choose a row $c_b$ among the vertices of $F_i$ with
$\pi(c_b)=b$; then $c_b\in F_{i'}$ whenever $b\in B_{i'}$. The final basis $B_t$
lies in $\mathcal A$, so it uses at most $d+1$ rows.
Let $E$ be the union of the error supports of these rows, the sets of
coordinates where $c_b$ and $b$ differ; then $|E|\le r(d+1)$.

\proofstep{The labels along a chain lie in a low-dimensional space.}
Fix $i$ and $j\notin E$, and suppose coordinate $j$ is unanimous on $F_i$. Each
$c_b$ with $b\in B_i$ lies in $F_i$ and agrees with $b$ at $j$, so coordinate $j$ is
constant on $B_i$. Every point of the span of $B_i$ is the sum of an odd number of
points of $B_i$, so the coordinate is constant on the entire binary
affine span, with the same value. No correction is needed there.
Therefore $\ell(F_i)$ agrees with $\rho(\operatorname{aff}_{\mathbb F_2}\pi(F_i))$
outside $E$.

There are at most $d+1$
distinct affine spans along the chain, since these spans are nested affine subspaces
of $\mathcal A$ and each strict inclusion
increases affine dimension. Their fixed representative vectors, of which there are
at most $d+1$, have
real affine dimension at most $d$.
All corrected labels therefore lie in the sum of this real affine space
and the coordinate subspace $\mathbb R^E$, of dimension at most
$d+r(d+1)$.

\proofstep{The model.}
The image of $G$ is thus a finite union of polytopes of dimension at most
$d+r(d+1)$.
Triangulating their common polyhedral arrangement gives a finite
polyhedron of the same dimension. Taking $f=G$ and $g$ the inclusion
proves $u(C)\le d+r(d+1)+1=(r+1)(d+1)$.
\end{proof}

\section{Literal complexes and integral obstructions}\label{app:literal}
Face width is defined through maps of a polytope. This section asks how much of
list replicability can be read directly from a finite simplicial complex, the
complex of consistent literal sets contained in a row. Each literal is a
constraint that a rule has to respect, so this complex keeps the coordinate
information that an unrestricted map of spaces would forget. Vanishing of
integral cohomology on links bounds list replicability and face width from above,
and the integral homology class of a shattered sphere bounds list replicability
from below. For continuous kernels that output distributions on the rows of the
class, the answer is exact and is given by the Leray number. A rule may output
functions outside the class, so the upper bound allows the complex to be enlarged;
we also show that the simplest way of enlarging it, one threshold on
$\operatorname{conv}C$, does not always produce a small Leray number.

\subsection{Integral link bounds}
We begin with the upper bound. Because a rule may output Boolean functions that
are not rows, the complex in the hypothesis is allowed to contain literal sets
realized by no row.

Write $K(C)$ for the complex of consistent literal sets contained in a
row of $C$. A completion of $K(C)$ is a complex $K'\supseteq K(C)$ on the same
literal vertices that contains no pair $(j,+),(j,-)$ in a face.
For a face $S$, the link is
$\operatorname{lk}_{K'}S=\{T:T\cap S=\varnothing,\ T\cup S\in K'\}$; the link of
a maximal face is the complex $\{\varnothing\}$, whose reduced homology and
cohomology are concentrated in degree $-1$.
Define $L_{\mathbb Z}(K')$ to be the least nonnegative integer $d$
such that
\[
 \widetilde H^i(\operatorname{lk}_{K'}S;\mathbb Z)=0
 \quad\text{for every face }S\text{ and every }i\ge d.
\]
The empty face is included. Let $L^\circ_{\mathbb Z}$ denote the
same quantity with only nonempty $S$. All complexes in this section are finite.

Obstruction theory needs vanishing with arbitrary coefficients. The first lemma
shows that integral cohomological vanishing provides it, and that the same
condition can be tested over fields.
\begin{lemma}[Change of coefficients]\label{lem:leray-coefficients}
Let $Y$ be a finite simplicial complex and $d\ge1$. The following are
equivalent:
\textup{(a)} $\widetilde H^i(Y;\mathbb Z)=0$ for all $i\ge d$;
\textup{(b)} $\widetilde H_i(Y;\mathbb Z)=0$ for all $i\ge d$ and
$\widetilde H_{d-1}(Y;\mathbb Z)$ is free;
\textup{(c)} $\widetilde H_i(Y;\mathbb F)=0$ for all $i\ge d$ and every field
$\mathbb F$.
They imply $\widetilde H^i(Y;G)=0$ for all $i\ge d$ and every abelian group $G$.
\end{lemma}
\begin{proof}
The augmented chain complex of $Y$ consists of finitely generated free abelian
groups, so the universal coefficient theorems give
\begin{align*}
 \widetilde H^i(Y;G)&\cong\operatorname{Hom}(\widetilde H_i(Y;\mathbb Z),G)
   \oplus\operatorname{Ext}(\widetilde H_{i-1}(Y;\mathbb Z),G),\\
 \widetilde H_i(Y;\mathbb F)&\cong\widetilde H_i(Y;\mathbb Z)\otimes\mathbb F
   \oplus\operatorname{Tor}(\widetilde H_{i-1}(Y;\mathbb Z),\mathbb F),
\end{align*}
with finitely generated homology groups.
If (a) holds, then for $i\ge d$ the vanishing of
$\operatorname{Hom}(\widetilde H_i,\mathbb Z)$ makes $\widetilde H_i$ finite, and
the vanishing of $\operatorname{Ext}(\widetilde H_{i-1},\mathbb Z)$ makes
$\widetilde H_{i-1}$ torsion-free. So $\widetilde H_i=0$ for $i\ge d$, and
$\widetilde H_{d-1}$ is free, which is (b).
If (b) holds, then for $i\ge d$ both summands of the first display vanish for
every $G$, because $\widetilde H_i=0$ and $\widetilde H_{i-1}$ is free; both
summands of the second display vanish for every $\mathbb F$. This gives (a), (c),
and the final assertion.
If (c) holds, then for $i\ge d$ the groups $\widetilde H_i\otimes\mathbb Q$ and
$\widetilde H_i\otimes\mathbb F_p$, for all primes $p$, vanish, which forces the
finitely generated group $\widetilde H_i$ to vanish; and
$\operatorname{Tor}(\widetilde H_{d-1},\mathbb F_p)=0$ for all $p$ makes
$\widetilde H_{d-1}$ torsion-free. This is (b).
\end{proof}

The second lemma is the extension step that will be applied on each face of the
cube.
\begin{lemma}[Extension over a contractible pair]\label{lem:contractible-extension}
Let $d\ge2$, let $(A,B)$ be a finite CW pair with $A$ contractible, and let $Y$ be
a simply connected and $(d-1)$-connected CW complex. If
$\widetilde H^i(B;G)=0$ for all $i\ge d$ and all abelian groups $G$, then every map
$B\to Y$ extends over $A$.
\end{lemma}
\begin{proof}
If $B=\varnothing$, map $A$ to a point of $Y$. Otherwise the exact sequence of the
pair and contractibility of $A$ give
$H^{i+1}(A,B;G)\cong\widetilde H^i(B;G)$ for every $i$ and $G$. The group
$H^{i+1}(A,B;\pi_i(Y))$ thus vanishes for $i\ge d$ by hypothesis, and for $i<d$
because $\pi_i(Y)=0$. Since $Y$ is simply connected, obstruction theory for CW
pairs gives the extension \cite[Section 4.3, Corollary 4.73]{Hatcher02}.
\end{proof}

The two conditions in the next theorem differ only in whether the empty face is
counted. This reflects the difference between $\operatorname{lr}(C)$ and
$u(C)=\operatorname{lr}(C^+)$: passing to the anchored class turns the empty face
into the anchor literal, whose link is the complex itself.
\begin{theorem}[Integral link bounds]\label{thm:literal-upper}
For every completion $K'$ of $K(C)$ and every integer $d\ge2$,
\[
 L^\circ_{\mathbb Z}(K')\le d\ \Longrightarrow\
 \operatorname{lr}(C)\le d+1,
 \qquad
 L_{\mathbb Z}(K')\le d\ \Longrightarrow\ u(C)\le d+1.
\]
\end{theorem}
The bound can be sharp with $K'=K(C)$: for the source-free orientations of $K_4$,
Proposition~\ref{prop:unanchored-products} applies the first implication with
$d=3$ and finds the exact value $4$. The proof realizes a completion radially in
the boundary of the cube, where each nonempty face $S$ becomes a cone over its
link. A face-preserving map to a low skeleton is then built by descending
induction on $|S|$, and the cohomological vanishing supplies each extension.

\begin{proof}
\proofstep{The radial realization.}
For $y\in\partial Q$, let $\operatorname{sgn}(y)$ be the set of literals
$(j,\operatorname{sign}y_j)$ with $y_j\ne0$; it is nonempty and consistent.
The radial realization of $K'$ in the boundary of the cube is the set $N'$ of points
whose nonzero signed coordinates form a face of $K'$, that is, with
$\operatorname{sgn}(y)\in K'$.
Cut $Q$ along the hyperplanes $y_j=0$, so that each coordinate interval has the
cells $\{-1\},(-1,0),\{0\},(0,1),\{1\}$ and $Q$ the product cell structure. The
set $\operatorname{sgn}(y)$ is constant on open cells and can only shrink at
limit points. Since $K'$ is closed under taking subsets, $N'$ is a closed
subcomplex.
If $y\in N[C]$ and $c$ is a row with $c_jy_j\ge0$ for all $j$, then
$\operatorname{sgn}(y)$ is a subset of the literal set of $c$. Thus
$N[C]\subseteq N'$. For a nonempty face $S\in K'$, let
$A_S=N'\cap Q_S$ and $B_S=A_S\cap\partial Q_S$; these are subcomplexes.

\proofstep{Each $A_S$ is a cone over the link of $S$.}
Let $b_S$ have the prescribed signs on $S$ and zero elsewhere. Then $b_S\in A_S$.
For $y\in A_S\setminus\{b_S\}$, let $y_{\rm f}\ne0$ be the part of $y$ on the
coordinates free in $S$. Multiplying $y_{\rm f}$ by a positive number does not
change $\operatorname{sgn}$. Hence the segment from $b_S$ through $y$ lies in $A_S$
until the free part has sup norm one, and it ends at a unique point of $B_S$.
So $A_S$ is the cone from $b_S$ over $B_S$, which is the single point $b_S$ if
$B_S=\varnothing$.

Radial normalization in the free coordinates,
$y\mapsto y_{\rm f}/\|y_{\rm f}\|_1$, with inverse
$x\mapsto x/\|x\|_\infty$ on the free coordinates and the signs of $S$ on the
others, identifies $B_S$
with the realization of $\operatorname{lk}_{K'}S$. The absence of
antipodal pairs makes the link a subcomplex of the boundary complex of the
cross-polytope on the free coordinates, whose realization is
$\{x:\|x\|_1=1,\ \operatorname{sgn}(x)\in\operatorname{lk}_{K'}S\}$.

\proofstep{The structure of $B_S$.}
We have $B_S=\bigcup_vA_{S+v}$, the union over literals $v$ with $S+v\in K'$: a
point of $B_S$ has some free coordinate at a value $s\in\{\pm1\}$, and then the
corresponding literal $v$ has $S+v\subseteq\operatorname{sgn}(y)$. Further,
$A_{S+v}\cap A_{S+w}=A_{S+v+w}$ if $S+v+w\in K'$, and the intersection is empty
otherwise, since a point of the intersection has $S+v+w\subseteq\operatorname{sgn}(y)$.

\proofstep{Construction on faces.}
Assume $L^\circ_{\mathbb Z}(K')\le d$.
Construct $f_S:A_S\to Q_S^{(d)}$ for nonempty faces $S\in K'$, in descending order
of $|S|$, with $f_S|_{A_{S+v}}=f_{S+v}$ whenever $S+v\in K'$.
On faces of dimension at most $d$, that is $n-|S|\le d$, use the inclusion
$A_S\subseteq Q_S=Q_S^{(d)}$; these choices are compatible.
For a face with $n-|S|>d$, the maps already constructed agree on the intersections
$A_{S+v}\cap A_{S+w}=A_{S+v+w}$, where both restrict to $f_{S+v+w}$. They therefore
define a continuous map $B_S\to Q_S^{(d)}$.
The target is $(d-1)$-connected, by cellular approximation in the contractible cube
as in the proof of Theorem~\ref{thm:carrier}, and it is simply
connected because $d\ge2$.
By hypothesis, $\widetilde H^i(\operatorname{lk}_{K'}S;\mathbb Z)=0$ for $i\ge d$.
Lemma~\ref{lem:leray-coefficients} gives
$\widetilde H^i(B_S;G)=0$ for $i\ge d$ and every abelian group $G$, since $B_S$ is
homeomorphic to the realization of the link. Lemma~\ref{lem:contractible-extension}
extends the map over the cone $A_S$.

\proofstep{Gluing over the facets.}
Every point of $N'$ has a coordinate $\pm1$, so $N'=\bigcup_vA_{\{v\}}$ over the
vertices $v$ of $K'$, and $N'\cap Q_v=\varnothing$ for other literals $v$. The maps
$f_{\{v\}}$ agree on intersections, so they glue over the coordinate facets to a
continuous map $f:N'\to Q^{(d)}$ with $f(N'\cap Q_v)\subseteq Q_v$ for every literal
$v$. Restrict to $N[C]$. Lemma~\ref{lem:cube-cover} gives
$\mathrm{LR}(C)\le d+1$, and Proposition~\ref{prop:cover-rule} proves the first
implication.

\proofstep{The anchored class.}
For the second implication, append a positive anchor $a$ and use the complex $K'*\{a\}$. The
faces of $K(C^+)$ are the sets $S$ and $S+a$ with $S\in K(C)$, so $K'*\{a\}$ is a
completion of $K(C^+)$; the literal opposite to $a$ is not a vertex.
The link of a face $S+a$ containing $a$ is the
link $\operatorname{lk}_{K'}S$ in $K'$, including its empty-face link $K'$ for
$S=\varnothing$. The link of a
nonempty face $S$ not containing $a$ is the cone
$\operatorname{lk}_{K'}(S)*\{a\}$, whose reduced cohomology vanishes in all degrees.
Consequently
$L^\circ_{\mathbb Z}(K'*\{a\})\le L_{\mathbb Z}(K')$.
If $L_{\mathbb Z}(K')\le d$, the first implication applied to $C^+$ gives
$\operatorname{lr}(C^+)\le d+1$, and Theorem~\ref{thm:anchor-width} identifies
$\operatorname{lr}(C^+)$ with $u(C)$.
\end{proof}

\subsection{Proper probability kernels and the Leray number}
We now consider the case in which no completion is allowed, because the output
itself is restricted to the rows of the class. The continuous form of this
restriction is a map from distributions on the rows to distributions supported on
few rows that stays in every face determined by a literal. For this problem the
topology of $K(C)$ gives an exact answer.

Let $\Delta_{\mathrm r}(C)$ be the simplex of probability distributions on the
rows; it should not be confused with the space $\Delta_C$ of realizable
distributions on labeled points. A proper continuous kernel of support size $b$ is a continuous map
$F:\Delta_{\mathrm r}(C)\to\Delta_{\mathrm r}(C)^{(b-1)}$ that preserves each face
$\Delta_{\mathrm r}(C_v)$, where $C_v$ consists of the rows satisfying literal
$v$. Denote the least support size by $p(C)$. This is a constraint on
continuous row distributions, rather than the unrestricted Boolean
output allowed in $\operatorname{lr}(C)$. Composing a proper kernel with the
barycentric coordinates of a triangulation of $\operatorname{conv}C$ whose vertices
are the rows yields a model,
as the last step in the proof of Lemma~\ref{lem:hub-game} shows, so
$\operatorname{lr}(C)\le u(C)\le p(C)$.

Let $\mathcal L(K)$ be the least nonnegative integer $d$ for which
every induced subcomplex has zero reduced homology in degrees at
least $d$, over every field. We call $\mathcal L(K)$ the \emph{Leray number} of $K$.
For a set $T$ of literals, $C_T$ denotes the set of rows satisfying all of them, so
$C_T\ne\varnothing$ exactly when $T\in K(C)$.

\begin{theorem}[Proper kernels and the Leray number]\label{thm:proper-kernel}
For every nonempty finite class, $p(C)=1+\mathcal L(K(C))$.
\end{theorem}
\begin{proof}
\proofstep{The nerve lemma.}
Write $K=K(C)$ and $\mathcal L=\mathcal L(K)$. The proof uses the nerve lemma for a
finite simplicial complex covered by subcomplexes whose nonempty finite intersections
are contractible: the complex is homotopy equivalent to the nerve of the cover. This is
\cite[Corollary 4G.3]{Hatcher02} after replacing each subcomplex $W$ by the open star
of its barycentric subdivision in the barycentric subdivision of the ambient complex.
A chain of simplices lies in this star exactly when its least member lies in $W$. So
these open sets have the same intersection pattern, an intersection of stars being the
star of the intersection of the subcomplexes, and each star retracts onto its subcomplex.

\proofstep{Lower bound.}
For a set $V$ of vertices of $K$, put
$U_V=\bigcup_{v\in V}\Delta_{\mathrm r}(C_v)$.
For nonempty $T\subseteq V$,
$\bigcap_{v\in T}\Delta_{\mathrm r}(C_v)=\Delta_{\mathrm r}(C_T)$, which is a simplex when $T\in K$ and empty
otherwise. All nonempty intersections are simplices and the nerve of this cover is the
induced subcomplex $K[V]$, so the nerve lemma gives
$U_V\simeq K[V]$.

Suppose $\widetilde H_j(U_V;\mathbb F)\ne0$ with $j\ge0$ and that a kernel $F$ has
support size at most $j+1$. Write $S=\Delta_{\mathrm r}(C)^{(j)}$ and $A=U_V\cap S$; then
$F(\Delta_{\mathrm r}(C_v))\subseteq\Delta_{\mathrm r}(C_v)\cap S$, so $F$ restricts to a map
$F':U_V\to A$. For $x\in\Delta_{\mathrm r}(C_v)$ with $v\in V$, the straight-line homotopy from
$x$ to $F(x)$ remains in the convex set $\Delta_{\mathrm r}(C_v)$. Hence, if $\iota:A\to U_V$ is
the inclusion, $\iota\circ F'$ is homotopic to the identity of $U_V$, and
$F'$ induces an injection
$\widetilde H_j(U_V;\mathbb F)\to\widetilde H_j(A;\mathbb F)$.
A nonzero class of degree $j$ therefore has nonzero image in
$\widetilde H_j(A;\mathbb F)$. Inclusion $A\to S$ is injective in
that degree: both are simplicial complexes of dimension at most $j$, so they have no
$(j+1)$-chains, and their reduced homology in degree $j$ is the group of reduced
$j$-cycles.
But the composite $U_V\to A\to S$ equals $F$ restricted to $U_V$, which factors through
the contractible simplex $\Delta_{\mathrm r}(C)$, so its reduced homology map is zero. This is a
contradiction, in reduced degree zero as well.

Thus any witness $\widetilde H_j(K[V];\mathbb F)\ne0$, $j\ge0$, gives support size at
least $j+2$. If $\mathcal L\ge1$, then minimality of $\mathcal L$ provides an induced
subcomplex and a field with nonzero reduced homology in some degree at least
$\mathcal L-1$. By the definition of $\mathcal L$ this degree is less than $\mathcal L$,
so it equals $\mathcal L-1$, and this witness proves $p(C)\ge\mathcal L+1$. When
$\mathcal L=0$ the bound
is just $p(C)\ge1$, since the skeleton $\Delta_{\mathrm r}(C)^{(-1)}$ is empty.

\proofstep{Vanishing on links when $d=\mathcal L\ge2$.}
We claim that every link, and every induced subcomplex of a link, has homology
zero in degrees at least $d$ over all fields. Let $K_0$ be a complex all of whose
induced subcomplexes have this property, let $v$ be one of its vertices, and let $W$
be a nonempty subset of the vertices of
$\operatorname{lk}_{K_0}v$. We have
\[
 K_0[W+v]=K_0[W]\cup\bigl(v*(\operatorname{lk}_{K_0}v)[W]\bigr),\qquad
 K_0[W]\cap\bigl(v*(\operatorname{lk}_{K_0}v)[W]\bigr)=(\operatorname{lk}_{K_0}v)[W].
\]
The Mayer--Vietoris sequence of this union, in which the cone has no reduced homology,
contains
\[
 \widetilde H_{i+1}(K_0[W+v];\mathbb F)\to\widetilde H_i((\operatorname{lk}_{K_0}v)[W];\mathbb F)
 \to\widetilde H_i(K_0[W];\mathbb F).
\]
For $i\ge d$ both outer groups vanish, so the middle one does. This gives the property for
$\operatorname{lk}_{K_0}v$. For a face $\{v_1,\ldots,v_r\}$, the link is obtained by
taking the link of $v_1$, then the link of $v_2$ in it, and so on; iteration starting
from $K_0=K$ gives the claim. Lemma~\ref{lem:leray-coefficients} then yields, for each
face $S$ of $K$ including $\varnothing$,
\begin{equation}\label{eq:link-all-coefficients}
 \widetilde H^i(\operatorname{lk}_KS;G)=0\qquad(i\ge d,\ G\text{ abelian}).
\end{equation}

\proofstep{The family of faces of $\Delta_{\mathrm r}(C)$.}
Let $\mathcal R$ be the family of nonempty row sets of the form $C_S$ with $S\in K$,
including $C_\varnothing=C$. It is closed under nonempty intersections, since
$C_S\cap C_{S'}=C_{S\cup S'}$. For $R\in\mathcal R$, let $\widehat S(R)$ be the set of
all literals satisfied by every row of $R$. Then $\widehat S(R)\in K$ and
$C_{\widehat S(R)}=R$.
Put $B_R=\bigcup\{\Delta_{\mathrm r}(R'):R'\in\mathcal R,\ R'\subsetneq R\}$. If $R'=C_{S'}$ is a
proper subset of $R$, then $S'$ contains a literal $v\notin\widehat S(R)$ and
$R'\subseteq C_{\widehat S(R)+v}\subsetneq R$. So
\[
 B_R=\bigcup_v\Delta_{\mathrm r}(C_{\widehat S(R)+v}),
\]
the union over the vertices $v$ of $\operatorname{lk}_K\widehat S(R)$. For a nonempty set
$T$ of such vertices, the intersection of the corresponding simplices is
$\Delta_{\mathrm r}(C_{\widehat S(R)\cup T})$, nonempty exactly when
$T\in\operatorname{lk}_K\widehat S(R)$. The nerve lemma gives
\begin{equation}\label{eq:BR-link}
 B_R\simeq|\operatorname{lk}_K\widehat S(R)|.
\end{equation}

\proofstep{The inductive scheme for the upper bound.}
With $d=\mathcal L\ge1$, we construct continuous maps
$f_R:\Delta_{\mathrm r}(R)\to\Delta_{\mathrm r}(R)^{(d)}$, by induction on $|R|$, such that
\begin{equation}\label{eq:fR-compatible}
 f_R|_{\Delta_{\mathrm r}(R')}=f_{R'}\qquad(R'\in\mathcal R,\ R'\subseteq R).
\end{equation}
This suffices. For a literal $v$ with $C_v\ne\varnothing$ we have $C_v\in\mathcal R$,
and $F=f_C$ then maps $\Delta_{\mathrm r}(C_v)$ into itself; for other literals $\Delta_{\mathrm r}(C_v)$ is
empty. So $F$ is a proper kernel of support size $d+1$.
If $\dim\Delta_{\mathrm r}(R)\le d$, let $f_R$ be the identity; this is compatible since subsets of
$R$ have the same property. For larger $R$, the maps $f_{R'}$, $R'\subsetneq R$, already
agree on intersections, because
$\Delta_{\mathrm r}(R')\cap\Delta_{\mathrm r}(R'')=\Delta_{\mathrm r}(R'\cap R'')$ and \eqref{eq:fR-compatible} holds for
smaller sets. They glue to a continuous map
$\phi_R:B_R\to\Delta_{\mathrm r}(R)^{(d)}$, and we must extend $\phi_R$ over $\Delta_{\mathrm r}(R)$.
The extension is obtained differently according to the value of $d$.

\proofstep{Upper bound when $d=\mathcal L\ge2$.}
The target $\Delta_{\mathrm r}(R)^{(d)}$ is $(d-1)$-connected by cellular approximation in the
simplex, and it is simply connected since $d\ge2$. By \eqref{eq:BR-link} and
\eqref{eq:link-all-coefficients}, $\widetilde H^i(B_R;G)=0$ for $i\ge d$ and all $G$.
Lemma~\ref{lem:contractible-extension}, for the pair $(\Delta_{\mathrm r}(R),B_R)$, gives the
extension. The construction at $R=C$ gives the desired kernel.

\proofstep{The two low-dimensional cases.}
If $d=\mathcal L=0$, every nonempty induced subcomplex is connected, so every
pair of vertices is an edge. Two antipodal literals never form an edge, hence every
column is constant. Since rows are distinct, $C$ has one row and $\Delta_{\mathrm r}(C)$ is a point;
the identity is a kernel of support size one.

If $d=\mathcal L=1$, a minimal nonface of size $s\ge3$ would induce the boundary of a
simplex, with nonzero homology in degree $s-2\ge1$. Thus $K$ is a
clique complex. Its graph is chordal, since an induced cycle of
length at least four would induce itself in the clique complex and give nonzero first
homology. An induced subgraph of a chordal graph is chordal.
Each connected component of the clique complex of a chordal graph is contractible, by
induction on the number of vertices. By Dirac's theorem on chordal graphs there is a
simplicial vertex $x$, whose neighbors form a clique. If $x$ has no neighbor, it is a
component by itself. Otherwise the faces containing $x$ all lie in the simplex on $x$ and
its neighbors, which retracts onto the neighbor face; so the complex is homotopy
equivalent to the complex with $x$ removed, with the same components.
The same holds for every link, because a link in a clique complex is the clique complex of
the induced subgraph on the common neighbors of the face.
Consequently, by \eqref{eq:BR-link}, each component of $B_R$ is contractible.

Now run the induction above with $d=1$. For $|R|\ge3$, the map $\phi_R$ from
$B_R$ to the connected graph $\Delta_{\mathrm r}(R)^{(1)}$ is null homotopic on each
component. Join these componentwise null homotopies to a common point
along paths in the graph. The map is null homotopic on $B_R$, hence
extends over the simplex $\Delta_{\mathrm r}(R)$ by the homotopy extension
property, starting from the constant map. The construction gives support size two.
\end{proof}
The lower bound should be compared with~\cite[Theorem 6]{CMY23}, which bounds the
proper list number below by
the hollow star number minus one. A hollow star of size $s$ is a minimal nonface of
$K(C)$, and in the present setting its induced boundary gives $p(C)\ge s$. The
restriction to the rows of the class is a strong one, as in the proper/improper
separation for singleton classes, which appeared first in~\cite{CMY23}, after Theorem 6.
For a class of $s$ rows, each differing from one fixed vector in a single coordinate,
Theorem~\ref{thm:proper-kernel} gives $p(C)\ge s$, while $u(C)\le2$ by
Proposition~\ref{prop:affine-neighborhood}.

\subsection{A limitation of uniform thresholds}
Theorem~\ref{thm:literal-upper} leaves the completion free, and the Leray number
gives a convenient way to verify its hypothesis. The argument in the proof of
Theorem~\ref{thm:proper-kernel} that passes from induced subcomplexes to links
applies to every finite complex, so $\mathcal L(K')\le d$ with $d\ge2$ implies
$L_{\mathbb Z}(K')\le d$, and hence $u(C)\le d+1$. The most direct candidate for a
completion of small Leray number comes from the polytope: relax each literal to one
common threshold on $\operatorname{conv}C$.

For $0<t\le1$, define the soft literal complex
\[
 K_t(C)=\{S:\exists z\in\operatorname{conv}C\text{ with }
                   sz_j\ge t\text{ for every }(j,s)\in S\}.
\]
It is a completion. It is closed under taking subsets, and it contains $K(C)$ because
a row $z=c$ has $sz_j=1\ge t$ on its literals. It has no antipodal pair, since $sz_j\ge t>0$
and $-sz_j\ge t$ are incompatible. A literal $(j,s)$ is a vertex exactly when some row has
$c_j=s$, because otherwise $sz_j=-1<t$ on all of $\operatorname{conv}C$.

One common threshold does not suffice. The next class is built so that each value of
$t$ meets a block of the class in which the all-positive literal set just fails to be
a face. By Theorems~\ref{thm:main} and~\ref{thm:anchor-width}, the class below has
$\operatorname{lr}(C_q)\le u(C_q)=O(\ln^2q)$, whereas its soft literal complexes have
Leray number at least $q-1$ at every threshold. The class is explicit.
Section~\ref{app:thresholds} allows a separate threshold for every literal, at the price
of a random class and a smaller exponent of the number of columns.
\begin{theorem}[Failure of a uniform threshold]\label{thm:uniform-threshold}
For every prime power $q\ge4$ there is a class $C_q$ with at most
$q^4$ rows and $2q^2$ columns such that, for every $0<t\le1$,
$K_t(C_q)$ has a minimal nonface of size at least $q$.
In particular, its Leray number, defined by homology of induced
subcomplexes over any field, is at least $q-1$.
\end{theorem}
\begin{proof}
\proofstep{Construction and size.}
For $k=q$ use the two-transitive affine group on $\mathbb F_q$, of order $q(q-1)$;
for $k=q+1$ use the projective linear group on
$\mathbb P^1(\mathbb F_q)$, which is sharply three-transitive of order $q(q^2-1)$.
Write $G$ for the group in question; in both cases it is two-transitive.
For each
$1\le r\le\lfloor k/2\rfloor+1$, introduce a disjoint block $E_{k,r}$
of $k$ columns, identified with the permuted set, and fix an $r$-subset $A$ of it. For
each group element
$g$, add the row negative on $gA$, positive on $E_{k,r}\setminus gA$,
and negative outside this block. Remove duplicate rows.
The number of columns is
$(\lfloor q/2\rfloor+1)q+(\lfloor(q+1)/2\rfloor+1)(q+1)$. This equals
$q^2+5q/2+1$ for even $q$ and $q^2+5q/2+3/2$ for odd $q$; it is at most $2q^2$ for
$q\ge4$ because $q^2-5q/2-3/2\ge0$ for $q\ge4$. The number of rows is at most
\[
 (\lfloor q/2\rfloor+1)q(q-1)
 +(\lfloor(q+1)/2\rfloor+1)q(q^2-1)\le q^4
\]
for $q\ge4$. For even $q$ the left side is $q(q-1)(q+2)^2/2$, and
$2q^4-q(q-1)(q+2)^2=q(q-2)^2(q+1)\ge0$. For odd $q\ge5$ it is $q(q^2-1)(q+4)/2$, and
$2q^4-q(q^2-1)(q+4)=q\bigl(q^2(q-4)+q+4\bigr)\ge0$.

\proofstep{The window of one block.}
Put $\rho=(1-t)/2\in[0,1/2)$. For $z=\sum_c\lambda_cc$ with a distribution $\lambda$ on
the rows, $z_j=1-2\Pr_\lambda(c_j=-1)$, so $z_j\ge t$ is the same as
$\Pr_\lambda(c_j=-1)\le\rho$.
In a block $E$ of size $k$ and parameter
$r$, every row has at least $r$ negative entries: exactly $r$ for rows of this block, and
$k$ for all others. Under any $\lambda$, the negative frequencies on $E$ therefore sum to
at least $r$, and one of them is at least $r/k$. Thus the all-positive
literal set on $E$ is not a face if $\rho<r/k$.
For a specified $u$ in the block, average the rows with $u\in gA$, using the uniform
distribution on these group elements; duplicate rows do not affect the average.
By two-transitivity, $\#\{g:u\in gA\}=r|G|/k$ and
$\#\{g:u,w\in gA\}=r(r-1)|G|/(k(k-1))$ for $w\ne u$. This makes the
negative frequency at every other point
equal to $(r-1)/(k-1)$. The all-positive set with $u$ deleted is
therefore a face whenever $\rho\ge(r-1)/(k-1)$. Every proper subset of the all-positive
set lies in such a set.
The positive literals of the entire block form a minimal nonface throughout the
interval
\[
 \frac{r-1}{k-1}\le\rho<\frac rk.
\]

\proofstep{Choice of the block.}
Choose $r=\lfloor\rho k\rfloor+1$. Then $r/k>\rho$; also $\rho k<k/2$ gives
$1\le r\le\lfloor k/2\rfloor+1$, so the block exists. The left inequality of the interval
says $\lfloor\rho k\rfloor\le\rho k-\rho$, that is
$\{\rho k\}\ge\rho$. If it fails for $k=q$, then
$\{\rho q\}<\rho$, so $\{\rho q\}+\rho<2\rho<1$ and
$\{\rho(q+1)\}=\{\rho q\}+\rho\ge\rho$.
It therefore holds for either $q$ or $q+1$, and the minimal nonface has size
$k\ge q$.
The induced complex on the positive literals of this block is the boundary of a simplex,
whose reduced homology in degree $k-2$ is nonzero over every field. So the Leray number
exceeds $k-2$, which proves the
last assertion directly from the induced-subcomplex definition.
\end{proof}

\subsection{A nonprimitive shattered sphere}
We turn to lower bounds. A set of $k$ coordinates shattered by $C$ gives a sphere of
dimension $k-1$ in the literal complex, and the bound
$\operatorname{lr}\ge\operatorname{vc}$ of~\cite[Theorem 3]{CMY23} gives
$\operatorname{lr}(C)\ge k$ for such a set. We show that the integral homology class
of this sphere can raise the bound by one.

If $J=\{j_1,\ldots,j_k\}$ is shattered by $C$, the complex $K(C)$
contains the cross-polytope sphere with fundamental cycle
\[
 z_J=\sum_{s\in\{\pm1\}^J}\Bigl(\prod_{j\in J}s_j\Bigr)
       [(j_1,s_{j_1}),\ldots,(j_k,s_{j_k})].
\]
It is a cycle: deleting the $i$th vertex from the two terms whose sign vectors differ
exactly at $j_i$ gives the same oriented simplex with opposite coefficients. For $k=1$ it
is the augmented cycle $[(j_1,+)]-[(j_1,-)]$, and an integral $0$-cocycle is a function
constant on connected components of $K(C)$.

\begin{theorem}[Nonprimitive shattered spheres]\label{thm:integral-sphere}
Let $J=\{j_1,\ldots,j_k\}$ be shattered by $C$.
If no integral $(k-1)$-cocycle on $K(C)$ evaluates to $1$ or $-1$
on $z_J$, then $\operatorname{lr}(C)\ge k+1$ and
$\operatorname{signrk}(C)\ge k+1$.
For $k\ge2$ the hypothesis says that $[z_J]$ is not primitive in
the torsion-free quotient of $H_{k-1}(K(C);\mathbb Z)$, with zero
also counted as nonprimitive.
\end{theorem}
Theorem~\ref{thm:integral-sphere} contains the lower bound
of~\cite[Lemma 5.2 and Theorem A]{BHH26b} as a special case: for an extremal
class $E\ne\{\pm1\}^X$, a maximum shattered set is strongly shattered, so
$z_J$ bounds a cone. It also improves the bound
$\operatorname{lr}\ge\operatorname{vc}$ of~\cite[Theorem 3]{CMY23} by one
whenever the cycle $z_J$ of a maximum shattered set is not primitive.

The proof compares two maps of pairs to the cube $(I^k,\partial I^k)$. A cover of
multiplicity at most $k$ yields such a map that factors through a complex of dimension
at most $k-1$, so it kills the top class. This map is homotopic to a coordinate clamp,
which has degree one on the shattered sphere, and the comparison produces a cocycle
that evaluates to $\pm1$ on $z_J$.

\begin{proof}
\proofstep{The reformulation for $k\ge2$.}
Evaluation on homology classes identifies the cocycles modulo coboundaries with
homomorphisms $H_{k-1}(K(C);\mathbb Z)\to\mathbb Z$ up to the Ext summand, and it is
onto $\operatorname{Hom}(H_{k-1}(K(C);\mathbb Z),\mathbb Z)$ by the universal coefficient
theorem. A homomorphism to $\mathbb Z$ factors through the torsion-free quotient
$\Gamma$, a finitely generated free abelian group. An element of $\Gamma$ takes the value
$\pm1$ under some homomorphism to $\mathbb Z$ exactly when it is primitive. Indeed, a
primitive element belongs to a basis of $\Gamma$, so some homomorphism takes the value
$1$ on it; a nonzero
element is a multiple $e\gamma$ with $\gamma$ primitive, and all its values are multiples of
$e$. This proves the second sentence.

\proofstep{A cover and its nerve.}
Suppose $\operatorname{lr}(C)\le k$. The point $x\in X(C)$ with barycentric coordinates
$|x_j|$ at the literals $(j,\operatorname{sign}x_j)$ identifies $X(C)$ with $|K(C)|$.
By Theorem~\ref{res:b}, for each error $\alpha>0$ we have
$\mathrm{LR}(C,\alpha)\le\operatorname{lr}(C,\alpha)\le k$.
Take an accurate closed cover
of $X(C)=|K(C)|$ of multiplicity at most $k$ and error $\alpha<1/(2n)$. Put
$W_j^s=\{x\in X(C):sx_j\ge1/n\}$. A label $h$ of a member meeting $W_j^s$ must
have sign $s$ on coordinate $j$, since otherwise its loss at that point is at least
$1/n>\alpha$. Apply Lemma~\ref{lem:thickening} to the nonempty members and the nonempty
sets $W_j^s$. This thickens the cover to an open cover
with the same multiplicity bound and with the following property: a member containing a
point of $W_j^s$ has a label with sign $s$ on coordinate $j$.
Its nerve is a finite complex $Y$ of
dimension at most $k-1$, with a partition-of-unity map $\varphi:X(C)\to Y$, for which
$\varphi(x)$ lies in the simplex spanned by members containing $x$.

For $i\le k$ and $s=\pm1$, let $B_i^s$ be the full subcomplex of $Y$ spanned by the
vertices whose labels have sign $s$ at $j_i$.
The two subcomplexes $B_i^+$ and $B_i^-$ are disjoint, since they have disjoint vertex
sets. Let $G_i:Y\to[0,1]$ be the sum of the barycentric coordinates at the vertices of
$B_i^+$. It equals zero on $B_i^-$ and one
on $B_i^+$. Put $B=\bigcup_{i,s}B_i^s$ and $G=(G_i)_i:(Y,B)\to(I^k,\partial I^k)$.

\proofstep{Comparison with the coordinate clamp.}
Let $Z=\bigcup_{i,s}W_{j_i}^s$.
If $x\in W_{j_i}^s$, then $\varphi(x)\in B_i^s$, so
$G_i\varphi(x)=(1+s)/2$. The coordinate clamp
$g_i(x)=\max\{0,\min\{1,(nx_{j_i}+1)/2\}\}$ also has
$g_i(x)=(1+s)/2$ there.
Thus both $G\varphi$ and $g=(g_i)_i$ map
$(X(C),Z)$ into $(I^k,\partial I^k)$.
Their straight-line homotopy is a homotopy of pairs, because on
each set in the union defining $Z$ the corresponding coordinate is
fixed to the same endpoint. If $\omega$ generates
$H^k(I^k,\partial I^k;\mathbb Z)$, then
$g^*\omega=(G\varphi)^*\omega=0$: this pullback factors through
$H^k(Y,B;\mathbb Z)=0$, by the dimension of $Y$.

\proofstep{Evaluation on the sphere.}
Let $k\ge2$. The boundary generator $\omega'$ of $H^{k-1}(\partial I^k;\mathbb Z)$ has
$\delta\omega'=\omega$, up to sign. Naturality gives $\delta(g|_Z)^*\omega'=g^*\omega=0$
in $H^k(X(C),Z;\mathbb Z)$. By the exact sequence of the pair, $(g|_Z)^*\omega'$ is the
restriction of a class $\beta\in H^{k-1}(X(C);\mathbb Z)$.
The shattered sphere $\Sigma_J=\{x\in X(C):\operatorname{supp}x\subseteq J\}$ lies in
$Z$, since one of its $k\le n$ coordinates has magnitude at least $1/k\ge1/n$.
On that sphere the clamp has degree one up to orientation. Indeed, for
$x\in\Sigma_J$ the vector $g(x)-(\tfrac12,\ldots,\tfrac12)$ has the same coordinate signs
as $(x_{j_1},\ldots,x_{j_k})\ne0$, so the straight-line homotopy between these two vectors
stays in $\mathbb R^k\setminus\{0\}$. The second vector is the inclusion of the
cross-polytope boundary, which radial projection maps homeomorphically onto the boundary
of the cube. Since $z_J$ is the fundamental cycle of $\Sigma_J$, a cocycle representing
$\beta$ evaluates to $\pm1$
on $z_J$, a contradiction.

For $k=1$ the same argument uses the unreduced class $\omega'\in H^0(\partial I;\mathbb Z)$
equal to $1$ at the point $1$ and $0$ at the point $0$, for which $\delta\omega'$ generates
$H^1(I,\partial I;\mathbb Z)$, and the
ordinary degree-zero exact sequence. The sphere $\Sigma_J$ is now the pair of vertices $(j_1,\pm)$,
which lies in $Z$ because $1\ge1/n$. The extension $\beta$ is a locally constant integer
function on $X(C)$, that is, a $0$-cocycle, with
$\beta(j_1,+)-\beta(j_1,-)=1$, again a contradiction.

Finally $\operatorname{lr}(C)\le\operatorname{signrk}(C)$ gives the
sign-rank bound.
\end{proof}

\section{An exact criterion in codimension two}\label{app:codimension}
For list replicability itself, Section~\ref{app:literal} gives sufficient conditions
and lower bounds. Near the top of the possible range of values the question has a
complete combinatorial answer. Every class on $n$ coordinates has
$\operatorname{lr}(C)\le n$~\cite[Lemma 4]{CMY23}, and by
Lemmas~\ref{lem:cube-cover} and~\ref{lem:cube-converse} the inequality
$\operatorname{lr}(C)\le n-2$ asks for a face-preserving map from $N[C]$ to
$Q^{(n-3)}$, a skeleton of codimension two in $\partial Q$. In this codimension all
obstructions to the map are integer degrees, and they are governed by the connected
components of the cube vertices missing from $C$. The criterion turns the list-size
question into an integer linear system whose variables describe these components. It
complements the shattered-sphere test in Theorem~\ref{thm:integral-sphere}, and its
corollary, the facet test, supplies the lower bound for the explicit family in
Theorem~\ref{thm:complement-family}.
Compare~\cite[Question 1.14]{BHH26b}, which asks for a combinatorial
characterization of relative simplicial covering dimension; the criterion
here gives such a test at this one level.

For a consistent literal set $S$, let $G_S$ be the graph on
$(\{\pm1\}^n\setminus C)\cap Q_S$, with edges joining vertices
that differ in one free coordinate. Vertices outside $C$ are called
\emph{missing}. Write $\operatorname{Comp}(S)$
for the set of connected components of $G_S$. If $S\subseteq S'$, then $G_{S'}$ is a
subgraph of $G_S$, so each component
of $G_{S'}$ belongs to one component of $G_S$.
An integer component measure is a function on this finite component
set with values in $\mathbb Z$; its values may be negative.
Its total mass is the sum of its values. A measure $\nu_{S'}$ pushes forward
to the measure on $\operatorname{Comp}(S)$ whose value at $T$ is
$\sum_{T'\subseteq T}\nu_{S'}(T')$, the sum over $T'\in\operatorname{Comp}(S')$ contained
in $T$. We write $\nu_v$ and $\nu_{v,w}$ for $\nu_{\{v\}}$ and $\nu_{\{v,w\}}$.

The criterion asks for a unit of integer mass on the missing components of every
facet and every two-literal face of the cube, consistent under restriction.
\begin{theorem}[Codimension-two criterion]\label{thm:codimension-two}
For $n\ge5$, $\operatorname{lr}(C)\le n-2$ if and only if there
are integer component measures $\nu_S$, for all consistent $S$ with
$1\le|S|\le2$, such that every measure has mass one and
$\nu_{\{v,w\}}$ pushes forward to both $\nu_{\{v\}}$ and $\nu_{\{w\}}$.
The necessity also holds for $n=4$.
\end{theorem}
The measures are the degrees of a face-preserving map. On a two-literal face $Q_S$
the target $Q_S^{(n-3)}$ is the sphere $\partial Q_S$, and a missing component
determines a cycle in the domain whose degree under the map is an integer. Necessity
computes these degrees and shows that they satisfy the two conditions. Sufficiency
reverses this: the measures prescribe degrees on small balls inside the missing
blocks, and the compatibility conditions are what is needed to extend the map over the
facets, after the balls in one component are joined by a tree.

\begin{proof}
\proofstep{Orthant blocks.}
Write $A_S=N[C]\cap Q_S$ and
$Y_S=Q_S^{(n-3)}$. For a cube vertex $p\in Q_S$, let
\[
 \beta_S(p)=\{y\in Q_S:p_jy_j\ge0\text{ for every free coordinate }j\}
\]
be the closed orthant block in $Q_S$ pointing toward
$p$ in all free coordinates. These blocks subdivide $Q_S$.
Adjacent blocks, for vertices $p,p'$ differing in one coordinate $j$, share the wall
$\beta_S(p)\cap\{y_j=0\}$.

\proofstep{Two facts about $A_S$.}
Let $S\ne\varnothing$. If $y\in Q_S$ and a row $c$ has $c_jy_j\ge0$ for all $j$, then
$c_j=s$ on every literal $(j,s)\in S$, so $c\in Q_S$ and $y\in\beta_S(c)$; the converse is
immediate since $Q_S\subseteq\partial Q$. Hence
\begin{equation}\label{eq:AS-blocks}
 A_S=\bigcup_{c\in C\cap Q_S}\beta_S(c).
\end{equation}
So a wall lies in $A_S$ if one of its two adjacent vertices belongs to $C$.
Moreover, a row $c$ with $y\in\beta_S(c)$ has $c_j=\operatorname{sign}y_j$ at each free $j$ with
$y_j\ne0$. This gives the following fact, which we call (M). A point $y\in Q_S$ whose free
coordinates are all nonzero with the signs of a missing vertex $p$ is outside $A_S$; and a
point with exactly one free coordinate $y_j=0$, the others nonzero with the common signs of
two adjacent missing vertices $p,p'$, is outside $A_S$.

\proofstep{Orientations and chains.}
We orient $Q_S$ for each $S$ with $1\le|S|\le2$, and blocks by restriction, so that
$[Q_S]=\sum_p[\beta_S(p)]$. Chains are cellular for the cell structure of $Q$ cut along
all hyperplanes $y_j=0$; the sets $A_S$, $\partial Q_S$ and $Y_S$ are subcomplexes.
For $|S|=2$, the sphere $\partial Q_S$ carries the fundamental cycle
$[\partial Q_S]=\partial[Q_S]$. Homology is integral.

\proofstep{Necessity: extension to low faces.}
Assume $\operatorname{lr}(C)\le n-2$, with $n\ge4$.
Lemma~\ref{lem:cube-converse}, with $d=n-3$, gives a face-preserving
$f:N[C]\to Q^{(n-3)}$; thus $f(A_S)\subseteq Y_S$ for every nonempty $S$.
For all $S$ with $|S|\ge3$, construct maps $F_S:Q_S\to Q_S$ with $F_S|_{A_S}=f|_{A_S}$ and
$F_S|_{Q_{S+w}}=F_{S+w}$, in increasing face dimension. The maps already defined
on $A_S$ and on $\partial Q_S$ agree on
$A_S\cap\partial Q_S=\bigcup_wA_{S+w}$. At each step
the target is the whole convex face, so the Tietze extension theorem applied to each
free coordinate extends this map from the closed set $A_S\cup\partial Q_S$ to $Q_S$.

For $|S|=2$ we have $Y_S=\partial Q_S\cong S^{n-3}$, and $f$ together with the maps
$F_{S+w}$ defines
\[
 h_S:A_S\cup\partial Q_S\to\partial Q_S.
\]
The map on the boundary of a two-literal face preserves all faces of $Q_S$, so it is
homotopic to the identity by straight-line interpolation within each face. It therefore
has degree one.
For a literal $v$, every cell of $Y_v$ lies in some $Q_S$ with $|S|=3$, $v\in S$, and $f$
together with these $F_S$ defines
\[
 h_v:A_v\cup Y_v\to Y_v,
\]
which restricts to $h_{\{v,w\}}$ on $A_{v,w}\cup\partial Q_{v,w}$.

\proofstep{Necessity: the measures on two-literal faces.}
For a component $T\in\operatorname{Comp}(S)$, $|S|=2$, put $b_T=\sum_{p\in T}[\beta_S(p)]$ and
$z_T=\partial b_T$. Walls between two blocks of $T$ cancel. A wall between
a block of $T$ and a block of a vertex $p'\notin T$ has $p'\in C$, since a missing
neighbor would belong to $T$; such a wall lies in $A_S$. All other cells of $z_T$ lie in
$\partial Q_S$. So $z_T$ is a cycle in $A_S\cup\partial Q_S$.
Its image under $h_S$ is a class in
$H_{n-3}(\partial Q_S)=\mathbb Z[\partial Q_S]$. Define $\nu_S(T)$ to be its degree:
\[
 (h_S)_*[z_T]=\nu_S(T)[\partial Q_S].
\]
Reversing the orientation of $Q_S$ changes both $z_T$ and $[\partial Q_S]$ by a sign, so
$\nu_S(T)$ does not depend on the orientation.
Summing all blocks of $Q_S$,
\[
 [\partial Q_S]=\sum_{T\in\operatorname{Comp}(S)}z_T
   +\sum_{c\in C\cap Q_S}\partial[\beta_S(c)].
\]
The blocks whose vertices belong to $C$ lie in $A_S$ by \eqref{eq:AS-blocks}, so their
boundaries are null-homologous in $A_S\cup\partial Q_S$ and contribute zero. Applying $h_S$,
of degree one on $\partial Q_S$, gives $\sum_T\nu_S(T)=1$.

\proofstep{Necessity: the measures on facets.}
Now fix a single literal $v$ and a missing component $T\in\operatorname{Comp}(v)$. Let $w$
range over the literals on the coordinates other than that of $v$, and let $\varepsilon_w$ be
the boundary incidence signs of $Q_v$, so $\partial[Q_v]=\sum_w\varepsilon_w[Q_{v,w}]$. An
outer face $\beta_v(p)\cap Q_w=\beta_{v,w}(p)$ occurs in the boundary of a single block, so
it has coefficient $\varepsilon_w$ in $\partial[\beta_v(p)]$. The set $T\cap Q_w$ is a union
of components of $G_{v,w}$, since an edge of $G_{v,w}$ is an edge of $G_v$. Therefore the block
sum $b_T=\sum_{p\in T}[\beta_v(p)]$ satisfies
\[
 \partial b_T=a_T+\sum_w\varepsilon_w\sum_{T'\subseteq T}b_{T'},
\]
where $T'$ runs over $\operatorname{Comp}(v,w)$ and $a_T$ is a chain in $A_v$ consisting of the
walls leaving $T$, as in the previous step. Taking the boundary,
\[
 \sum_w\varepsilon_w\sum_{T'\subseteq T}z_{T'}=-\partial a_T.
\]
The right side is a boundary in $A_v\cup Y_v$. Let
$\mu_w(T)=\sum_{T'\subseteq T}\nu_{v,w}(T')$.
Applying $h_v$, and using the definition of $\nu_{v,w}$, gives the relation
\[
 \sum_w\varepsilon_w\mu_w(T)\,\partial[Q_{v,w}]=0
       \quad\text{in }H_{n-3}(Y_v;\mathbb Z).
\]
Since $Y_v$ has dimension $n-3$, its top homology is its group of
cellular cycles for the cube structure, so the relation holds at the level of chains.
It says that the cellular $(n-2)$-chain
$\sum_w\varepsilon_w\mu_w(T)[Q_{v,w}]$ on the sphere $\partial Q_v$
has zero boundary. Such a chain is a multiple of the fundamental
cycle $\sum_w\varepsilon_w[Q_{v,w}]$ of the sphere. Thus $\mu_w(T)$ is independent of $w$.
Define this common integer to be $\nu_v(T)$. By construction $\nu_{v,w}$ pushes forward to
$\nu_v$ for every $w$. Each element of $\operatorname{Comp}(v,w)$ lies in a unique $T$, so
$\sum_T\nu_v(T)=\sum_{T'}\nu_{v,w}(T')=1$. The measure $\nu_{v,w}$ depends on the unordered pair;
applying this step to the literal $w$, with $v$ among its partners, shows that
$\nu_{v,w}$ pushes forward to $\nu_w$ as well.

\proofstep{Sufficiency: maps on two-literal faces.}
Let measures as in the statement be given, and let $n\ge5$.
Fix $0<\theta<1/4$. Let $|S|=2$, so $\dim Q_S=n-2\ge3$.
For every component $T$ with $\nu_S(T)\ne0$, choose a vertex $p_T\in T$, and let $D_T$ be
the closed cube of sup-radius $\theta$, in $Q_S$, about the center of the missing orthant
block $\beta_S(p_T)$, the point with free coordinates $p_{T,j}/2$. This is a small closed
ball in the interior of the block, and it misses $A_S$, by fact (M).
Let $W_S$ be $Q_S$ with the interiors of these balls removed, a connected compact
oriented manifold with boundary class $[\partial Q_S]-\sum_T[\partial D_T]$, where $\partial D_T$
has the boundary orientation of $D_T$; thus $A_S\subseteq W_S$.
Map the outer boundary of $Q_S$ by the identity, and map the boundary
of each small ball to $\partial Q_S$ with degree $\nu_S(T)$, with
the outward orientation convention for the removed balls.
These maps extend over the punctured face to a map $h_S:W_S\to\partial Q_S$: the target is
$S^{n-3}$ and the only obstruction, in
$H^{n-2}(W_S,\partial W_S;\pi_{n-3}(S^{n-3}))\cong\mathbb Z$, is its total boundary degree,
which is $1-\sum_T\nu_S(T)=0$.
This is the Hopf extension theorem \cite[Chapter 3, Section 6]{GP74}.
Restrict to $A_S$. All these maps agree on faces with at least
three fixed coordinates, where they are the identity.

\proofstep{Sufficiency: the trees in a facet.}
Fix $v$, so that $\dim Q_v=n-1\ge4$. Let $w$ range over the literals on other
coordinates. Call the cubes $D_{T'}$, for
$T'\in\operatorname{Comp}(v,w)$ with $\nu_{v,w}(T')\ne0$, the leaf balls. Each lies in the
relative interior of the outer face $\beta_{v,w}(p_{T'})$ of the block $\beta_v(p_{T'})$, and
$T'$ belongs to the unique $T\in\operatorname{Comp}(v)$ containing $p_{T'}$. Let
$\mathcal T\subseteq\operatorname{Comp}(v)$ be the set of components containing a leaf ball, and
$b=|\mathcal T|$; we have $b\ge1$ since the measures have mass one.

The degrees on the leaf balls of one component cancel only in total, by the
compatibility of the measures. We therefore join these balls inside the missing region
and remove a neighborhood of the result. In each component $T\in\mathcal T$, connect the chosen balls
on the faces $Q_{v,w}$ by a finite embedded tree $\tau_T$ lying outside $A_v$. For $p\in T$, let
$o_p$ be the center of the block $\beta_v(p)$. Choose a spanning tree of the connected graph
$T$. For each of its edges $pp'$, take the straight segment from $o_p$ to $o_{p'}$. For each
leaf ball $D_{T'}$ with $p=p_{T'}$, take the straight segment from $o_p$ to its center, which
is an inward normal of the outer face. All these segments are
parallel to coordinate axes, and at each $o_p$ they have distinct directions, so they meet
only at common endpoints; the result is an embedded tree. By fact (M) it misses
$A_v$: along a segment between adjacent $o_p,o_{p'}$ only one coordinate varies, and it
passes through zero between two missing sign patterns. The tree meets $\partial Q_v$
only in the centers of its leaf balls.

Let $N_T$ be the set of points of $Q_v$ at sup-distance at most $\theta$ from $\tau_T$, a regular
neighborhood of the tree. It is
the union of the cubes of radius $\theta$ about the points $o_p$ and of boxes joining these
cubes along the segments. A box attached at $o_p$ lies beyond one facet of the cube about $o_p$,
and boxes without a common end vertex are at positive distance because $2\theta<1$. Building
$N_T$ along the tree, each new cube or box therefore meets the part already built exactly in one
of its facets, so $N_T$ is a PL ball. Since $1/2+\theta<1$, it meets $\partial Q_v$ exactly in
the leaf balls of $T$.
Points on different trees have sup-distance at least $1/2$. Indeed, they lie in blocks of
vertices $p,p''$ of different components, and these differ in at least two coordinates, since
adjacent missing vertices lie in one component. A point $y$ of $\tau_T$ in the block of $p$ has at
most one free coordinate different from $p_i/2$, so $y_i=p_i/2$ at some coordinate $i$ where
$p_i\ne p''_i$, while $p''_iy'_i\ge0$ for the second point $y'$. As $\theta<1/4$, the
neighborhoods $N_T$ are disjoint for different $T$. They also miss $A_v$, by fact (M): within
sup-distance $\theta<1/2$ of a segment, all coordinates but the varying one keep their nonzero
signs, and on a leaf segment all coordinates keep the signs of $p_{T'}$.

\proofstep{Sufficiency: the complement of the trees.}
Let $N=\bigsqcup_TN_T$, let $\Lambda$ be the closure of $Q_v\setminus N$, and let
$F_T=N_T\cap\Lambda$, the frontier of $N_T$ in $Q_v$, with $F=\bigsqcup_TF_T$. Let $P$ be
the outer boundary $\partial Q_v$ with the leaf-ball interiors removed. Then $\Lambda$ is a
compact $(n-1)$-manifold with $A_v\subseteq\Lambda$ and
$\partial\Lambda=P\cup F$, $P\cap F=\partial P=\partial F$.
The maps just defined give $g:P\to Y_v$: on
$W_{\{v,w\}}$ it is $h_{\{v,w\}}$, with values in $\partial Q_{v,w}\subseteq Y_v$, and it is the
identity on the faces $Q_{v,w}\cap Q_{v,w'}$. We must extend $g$ over $\Lambda$.

\proofstep{Sufficiency: the topology of $(\Lambda,P)$.}
The boundary $\partial N_T$ is a sphere $S^{n-2}$, and $F_T$ is this sphere with finitely many
disjoint open balls removed. Since $n-2\ge3$, each $F_T$ is connected and simply
connected, and so is $P$. The Mayer--Vietoris sequence of $Q_v=\Lambda\cup N$, with
$\Lambda\cap N=F$, shows that $\Lambda$ is connected, because $F_T$ and $N_T$ are connected.
Attaching the balls $N_T$ to $\Lambda$ along the simply connected sets $F_T$ does not change the
fundamental group, by the Seifert--van Kampen theorem. As $Q_v$ is contractible,
$\Lambda$ is simply connected.

For every $i$, Lefschetz duality \cite[Theorem 3.43]{Hatcher02}, for the
decomposition $\partial\Lambda=P\cup F$, excision and contractibility of $Q_v$ give
\[
 H_i(\Lambda,P)\cong H^{n-1-i}(\Lambda,F)\cong H^{n-1-i}(Q_v,N)
 \cong\widetilde H^{n-2-i}(N).
\]
As $N$ is a disjoint union of $b$ balls,
\[
 H_i(\Lambda,P;\mathbb Z)=
 \begin{cases}\mathbb Z^{b-1},&i=n-2,\\0,&i\ne n-2.\end{cases}
\]
Since $P$ and $\Lambda$ are simply connected, the relative Hurewicz theorem
\cite[Theorem 4.32]{Hatcher02} shows that $(\Lambda,P)$ is $(n-3)$-connected and
represents a basis of $H_{n-2}(\Lambda,P)$ by maps
$\Xi_a:(D^{n-2},S^{n-3})\to(\Lambda,P)$, $1\le a\le b-1$. A homotopy of $\Xi_a|_{S^{n-3}}$ in $P$
extends over the disk without changing the relative class, so the maps
$\phi_a=\Xi_a|_{S^{n-3}}$ may be taken cellular for a triangulation of $P$. Attaching relative
$(n-2)$-cells to $P$ along the $\phi_a$ gives a CW complex $P'$ and a map $P'\to\Lambda$ extending the
inclusion of $P$, which is a homology equivalence relative to $P$, hence a homology
equivalence by the five lemma. Both spaces are simply connected, because $n-2\ge3$, so it
is a homotopy equivalence \cite[Corollary 4.33]{Hatcher02}, and also a homotopy equivalence
relative to $P$ \cite[Proposition 0.19]{Hatcher02}. There is thus a map
$r:\Lambda\to P'$ with $r|_P=\mathrm{id}$, and it suffices to extend $g$ over these cells.

\proofstep{Sufficiency: vanishing of the obstruction.}
The target $Y_v$ is the $(n-3)$-skeleton of a contractible cube, so it is
$(n-4)$-connected, and $n-3\ge2$ makes it simply connected. By the Hurewicz theorem,
$\pi_{n-3}(Y_v)\cong H_{n-3}(Y_v)$, and this group contains the only obstruction, since
the cells to be filled have dimension $n-2$.
So $g$ extends over the cell attached by $\phi_a$ exactly when the homomorphism
\[
 \Theta=g_*\circ\partial:H_{n-2}(\Lambda,P)\to H_{n-3}(P)\to H_{n-3}(Y_v)
\]
vanishes on the class of $\Xi_a$. It suffices to show $\Theta=0$.

The group $H_{n-2}(\Lambda,P;\mathbb Z)$ is generated by the frontier classes
of the tree neighborhoods. Indeed, in the exact sequence of the triple
$(\Lambda,\partial\Lambda,P)$, the map $H_{n-2}(\partial\Lambda,P)\to H_{n-2}(\Lambda,P)$ is
onto, since $H_{n-2}(\Lambda,\partial\Lambda)\cong H^1(\Lambda)=0$ by duality and simple
connectivity. By excision,
$H_{n-2}(\partial\Lambda,P)\cong\bigoplus_TH_{n-2}(F_T,\partial F_T)$, and each summand is generated
by the fundamental class $[F_T]$ of the connected oriented manifold $F_T$, oriented as part of
$\partial N_T$.
The obstruction is the homomorphism sending a
relative class to the image under $g_*$ of its boundary, so it suffices to
evaluate it on these generators.
Compare orientations on a leaf ball $D_{T'}\subseteq Q_{v,w}$: the outward normals of $N_T$ and of
$Q_v$ agree there, so the orientation induced from $\partial N_T$ is the one induced from
$\partial Q_v$, which is $\varepsilon_w$ times the orientation of $Q_{v,w}$. Since
$[\partial N_T]=[F_T]+\sum_{T'}\varepsilon_w[D_{T'}]$ is a cycle,
$\partial[F_T]=-\sum_{T'}\varepsilon_w[\partial D_{T'}]$, with $[\partial D_{T'}]$ oriented as in the
step on two-literal faces.
The obstruction from the tree in component $T$ is, up to this overall sign,
\[
 \sum_w\varepsilon_w
   \sum_{T'\subseteq T\cap Q_{v,w}}\nu_{v,w}(T')
      [\partial Q_{v,w}]
 =\nu_v(T)\sum_w\varepsilon_w[\partial Q_{v,w}]=0.
\]
Components $T'$ with $\nu_{v,w}(T')=0$ have no ball and contribute zero on the left.
The first equality is exactly the measure compatibility; the last
is the boundary-of-boundary identity $\partial\partial[Q_v]=0$.
So $g$ extends over $P'$, and composing with $r$ gives an extension $h_v:\Lambda\to Y_v$ of $g$.

\proofstep{Sufficiency: gluing.}
The extensions restricted to $A_v\subseteq\Lambda$ agree on intersections of facets: if $v,v'$ are
consistent, then $A_v\cap A_{v'}=A_{v,v'}\subseteq W_{\{v,v'\}}$, and both restrict to
$h_{\{v,v'\}}$; if they are opposite, the intersection is empty.
They glue to a continuous map $N[C]\to Q^{(n-3)}$, which sends $N[C]\cap Q_v=A_v$ into $Y_v\subseteq Q_v$.
Lemma~\ref{lem:cube-cover} gives $\mathrm{LR}(C)\le n-2$, and
Proposition~\ref{prop:cover-rule} completes the proof.
\end{proof}

The compatibility condition has an immediate consequence on a single facet. A
component that misses one side of a coordinate receives no mass from the face on that
side, so it cannot carry mass in the facet.
\begin{corollary}[Facet test]\label{cor:facet-test}
If, for some literal $v$, every component of $G_v$ is contained in
a proper coordinate facet of $Q_v$, then
$\operatorname{lr}(C)\ge n-1$ for $n\ge4$.
\end{corollary}
\begin{proof}
Suppose $\operatorname{lr}(C)\le n-2$, and take the measures from
Theorem~\ref{thm:codimension-two}. Let $T\in\operatorname{Comp}(v)$ be contained in $Q_w$,
for a literal $w$ on a coordinate other than that of $v$.
Then $T$ misses $Q_{-w}$, where $-w$ is the opposite literal. No component of $G_{v,-w}$ lies in
$T$, so compatibility with $\nu_{v,-w}$ would
make the mass $\nu_v(T)$ equal to the empty sum from that face, hence zero.
All masses of $\nu_v$ would vanish, contrary to total mass one.
\end{proof}

For a class with a constant coordinate, the criterion reduces to a connectivity
statement in a single facet. Since $\operatorname{lr}(C^+)=u(C)$ by
Theorem~\ref{thm:anchor-width}, this is an exact test for face width at the top of its
range.
\begin{corollary}[Classes with a constant coordinate]\label{cor:half-class}
Suppose every row of $C$ has a fixed literal $v$, and retain that
coordinate as part of the domain. For $n\ge5$,
$\operatorname{lr}(C)\le n-2$ if and only if some component of
$G_v$ meets both sides of every other coordinate.
\end{corollary}
\begin{proof}
Necessity follows from the facet test: if no such component exists, then every component of
$G_v$ misses one side of some other coordinate, so it lies in a proper coordinate facet of $Q_v$.

For sufficiency, let $T_0\in\operatorname{Comp}(v)$ meet all sides, and give it mass one, so
$\nu_v=\delta_{T_0}$. For each literal $w$ on another coordinate, the set $T_0\cap Q_w$ is
a nonempty union of components of $G_{v,w}$; choose a
component of $G_{v,w}$ in it and give that component mass one.

Every other single or double literal graph is connected, because
the whole opposite-anchor face $Q_{-v}$ is missing and every missing vertex
with anchor $v$ is adjacent to it. If $S$ contains $-v$, then $G_S$ is the graph of a
whole cube face. If $S$ has one or two literals, all on coordinates other than that of $v$, then
$G_S$ contains all vertices of the nonempty subcube $Q_S\cap Q_{-v}$, and a missing vertex of
$Q_S\cap Q_v$ is adjacent to it by flipping the coordinate of $v$.
Give the sole component of each of these graphs mass one.

These are compatible measures. The measure $\nu_{v,w}$ pushes forward to $\nu_v$ by the
choice of its component in $T_0$. Every other push-forward is between measures of mass one on
a one-point target, where the condition is automatic.
So Theorem~\ref{thm:codimension-two}
applies.
\end{proof}

\section{Limits of coordinate-threshold completions}\label{app:thresholds}
Section~\ref{app:literal} shows that a completion of small Leray number bounds the
face width, and that one common threshold on $\operatorname{conv}C$ does not always
produce such a completion. This section allows a separate threshold for every literal,
which gives all completions defined by one inequality on the coordinate of each
literal. If one of
these completions always had small Leray number, Theorem~\ref{thm:literal-upper} would
bound the face width from the polytope alone, with no further construction.
Theorem~\ref{thm:literal-threshold-gap} shows that this is not the case: there are
classes of face width $O(\ln n)$ for which every such completion has Leray number
polynomial in $n$. This result explains why the carrier construction of
Section~\ref{app:construction}, outlined in Section~\ref{sec:construction}, does not read
its carriers off the coordinates of $\operatorname{conv}C$ and instead builds compatible
geometric fillings, cell by cell across the faces of the cube. The logarithmic upper
bound for the classes below comes from another kind of completion, a class
$H\supseteq C$ that can be learned in a short game.

For $0\le r_j<p_j\le1$ and disjoint $J_+,J_-\subseteq[n]$, write $+J_+\cup-J_-$ for
the literal set that is positive on $J_+$ and negative on $J_-$, and let $K_{r,p}(C)$ contain
$+J_+\cup-J_-$ exactly when some probability distribution $q$ on $C$
satisfies
\[
 q\{c:c_j=-1\}\le r_j\quad(j\in J_+),\qquad
 q\{c:c_j=-1\}\ge p_j\quad(j\in J_-).
\]
These are precisely the coordinatewise convex threshold completions
that contain $K(C)$ and exclude opposite literals in a common face.
Indeed, for $z=\sum_cq(c)c$ we have $z_j=1-2q\{c:c_j=-1\}$. A condition
$z_j\ge a_j$ on the literal $+j$ and a condition $z_j\le b_j$ on the
literal $-j$ read $q\{c_j=-1\}\le(1-a_j)/2$ and $q\{c_j=-1\}\ge(1-b_j)/2$.
Rows satisfy their own literals for all columns exactly when $a_j\le1$ and
$b_j\ge-1$, that is, $r_j=(1-a_j)/2\ge0$ and $p_j=(1-b_j)/2\le1$, and no $z_j$
satisfies both conditions exactly when $a_j>b_j$, that is, $r_j<p_j$.

The Leray number $\mathcal L$ is the all-field induced-subcomplex
parameter defined before Theorem~\ref{thm:proper-kernel}. If an induced
subcomplex has nonzero reduced homology in degree $i\ge0$ over some field,
then $\mathcal L\ge i+1$. In particular, a minimal nonface $Y$ of a complex with
$|Y|\ge2$ induces the boundary of a simplex, with nonzero homology in degree
$|Y|-2$, so $\mathcal L\ge|Y|-1$.

For a class of polynomial size, Theorems~\ref{thm:main} and~\ref{thm:anchor-width} give
$u(C)=O(\ln^2n)$. The next theorem gives such classes for which no threshold completion
has polylogarithmic Leray number.
\begin{theorem}[Threshold completions and face width]\label{thm:literal-threshold-gap}
For every sufficiently large integer $n$, there is a class $C$ on
$n$ coordinates with $|C|\le n^{12}$ such that
\[
 u(C)=O(\ln n),\qquad
 \inf_{r,p}\mathcal L(K_{r,p}(C))
 \ge c\,\frac{n^{4/15}}{(\ln n)^{2/3}}
\]
for an absolute constant $c>0$.
\end{theorem}

The upper bound on $u(C)$ rests on an elementary completion principle, phrased as
a game. A position is a consistent literal
set $S$ contained in at least one row of a fixed class $H$. A prediction
is a row $h\in H$ extending $S$. A move adds a literal on which $h$
is wrong, provided the new position is still contained in a row of $H$.
Let $M(H)$ be the smallest worst-case number of moves of a strategy.
In the language of exact learning, a prediction is a proper and consistent
equivalence query and a move is a counterexample. If such a learner needs few
counterexamples on a class containing $C$, the face width of $C$ is small.

\begin{lemma}[Short games bound the face width]\label{lem:hub-game}
If $C\subseteq H$ and $M(H)\le b$, where $b\ge1$, then
$u(C)\le b+1$.
\end{lemma}
\begin{proof}
\proofstep{A deletion inequality.}
For any finite complex $K$ and vertex $v$,
\[
 \mathcal L(K)\le
 \max\{\mathcal L(K-v),1+\mathcal L(\operatorname{lk}_Kv)\};
\]
compare the Stanley--Reisner form in~\cite[Lemma 2.10]{DHS13}.
To prove it, let $W$ be a vertex set and fix a field. If $v\notin W$, then $K[W]$ is an
induced subcomplex of $K-v$. If $v\in W$, then
$K[W]=K[W\setminus\{v\}]\cup(v*\Lambda)$ with
\[
 K[W\setminus\{v\}]\cap(v*\Lambda)=\Lambda,\qquad
 \Lambda=\operatorname{lk}_{K[W]}v=(\operatorname{lk}_Kv)[W\setminus\{v\}].
\]
The cone $v*\Lambda$ is acyclic, and the augmented chain complexes give the
Mayer--Vietoris sequence
\[
 \widetilde H_i(K[W\setminus\{v\}])\to\widetilde H_i(K[W])\to\widetilde H_{i-1}(\Lambda).
\]
For $i\ge\max\{\mathcal L(K-v),1+\mathcal L(\operatorname{lk}_Kv)\}$ we have $i\ge1$, and both
outer groups vanish, the first since $K[W\setminus\{v\}]$ is induced in $K-v$, the second since
$\Lambda$ is induced in the link and $i-1\ge\mathcal L(\operatorname{lk}_Kv)\ge0$.
So $\widetilde H_i(K[W])=0$ in both cases, which proves the inequality.
We also use that $\mathcal L$ does not increase on passing to an induced subcomplex,
because an induced subcomplex of an induced subcomplex is induced.

\proofstep{Induction on positions.}
For a position $S$ let $H_S$ be the set of rows of $H$ extending $S$, let
$K_S=K(H_S)$ be its literal complex, and let $M_S$ be the smallest
worst-case number of moves from $S$. Thus $M_\varnothing=M(H)$. We prove
$\mathcal L(K_S)\le M_S$ by downward induction on $|S|$; moves enlarge $S$, so positions with no
move form the base and are included in the following argument.

Let $h$ be the prediction of an optimal strategy at $S$. The literals of $h$ form a face of
$K_S$. Let $v_1,\ldots,v_t$ be the vertices of $K_S$ outside $h$. Each $v_i$ lies in a row of
$H_S$ and $h$ is wrong on it, so it is a legal move, and hence $M_{S+v_i}\le M_S-1$. Put
$K^{(0)}=K_S$ and $K^{(i)}=K^{(i-1)}-v_i$. The complex $K^{(t)}$ is the simplex on the literals of $h$,
so $\mathcal L(K^{(t)})=0$. A face of $K_S$ containing $v_i$ is a subset of a row of $H_{S+v_i}$.
Therefore $\operatorname{lk}_{K_S}v_i$ is the induced subcomplex of $K_{S+v_i}$ on the vertices other
than $v_i$, and $\operatorname{lk}_{K^{(i-1)}}v_i$ is an induced subcomplex of this. Thus
$\mathcal L(\operatorname{lk}_{K^{(i-1)}}v_i)\le\mathcal L(K_{S+v_i})\le M_{S+v_i}\le M_S-1$
by the induction hypothesis. The deletion inequality gives
\[
 \mathcal L(K^{(i-1)})\le\max\{\mathcal L(K^{(i)}),M_S\}\qquad(1\le i\le t),
\]
and descending from $i=t$ to $i=1$ gives $\mathcal L(K_S)\le M_S$. If $t=0$, then
$K_S$ is the simplex on $h$. At $S=\varnothing$ we get $\mathcal L(K(H))\le M(H)\le b$.

\proofstep{From the kernel to a model.}
Theorem~\ref{thm:proper-kernel} gives $p(H)=1+\mathcal L(K(H))\le b+1$: there is a
continuous map $F:\Delta_{\mathrm r}(H)\to\Delta_{\mathrm r}(H)^{(b)}$ that preserves each face $\Delta_{\mathrm r}(H_v)$.
The elements of $H$ are cube vertices, so they are the vertices of
$\operatorname{conv}H$. Choose a pulling triangulation of $\operatorname{conv}H$
(Lemma~\ref{lem:pulling}), whose vertex set is $H$. Let
$\sigma:\operatorname{conv}H\to\Delta_{\mathrm r}(H)$ send a point to its barycentric coordinates in its
supporting simplex; it is continuous. If $x_j=s\in\{\pm1\}$, then $x$ is a positive combination
of the vertices of its supporting simplex, so these vertices all have the value $s$ at $j$
and the supporting simplex lies in $\{x_j=s\}$; this is the fact recorded
in~\eqref{eq:fixed-at-interior} of Section~\ref{app:orientations}. So
$\sigma(x)\in\Delta_{\mathrm r}(H_{(j,s)})$.
Let $f=F\circ\sigma$ and let $g:\Delta_{\mathrm r}(H)^{(b)}\to[-1,1]^n$ be the affine map sending a row
to itself. If $x_j=s$, then $f(x)\in\Delta_{\mathrm r}(H_{(j,s)})$, so all active rows have value $s$ at $j$,
and $(gf(x))_j=s$. Thus $(f,g,\Delta_{\mathrm r}(H)^{(b)})$ is a model for $H$ of dimension at most $b$.
Restricting $f$ to $\operatorname{conv}C\subseteq\operatorname{conv}H$ gives a model for $C$.
\end{proof}

We now prove the theorem. The class is a union of random blocks, with one family of
blocks at each of a sequence of density levels. A threshold vector with many large
entries has many entries just above some level, and a block of that level then gives a
set of negative literals that is infeasible as a whole although all its small subsets
are feasible. This produces a large minimal nonface. If instead many entries are small,
the rows with one negative entry make all small literal sets on those coordinates
feasible. The face width is small for a different reason: distinct block supports meet in few
coordinates, so a few literals identify the block, and within a block a few literals
identify the row up to a short list.

\begin{proof}[Proof of Theorem~\ref{thm:literal-threshold-gap}]
\proofstep{Parameters and density levels.}
Put $w=\lceil n^{2/5}\rceil$, $m_0=n^{10}$, and
\[
 \varepsilon=\left(\frac{8192\ln(4m_0)}w\right)^{1/3},\quad
 \kappa=20\varepsilon^2,\quad
 s=\left\lfloor\frac{\ln m_0}{512\varepsilon^2}\right\rfloor.
\]
All inequalities below hold for sufficiently large $n$. Since $\ln(4m_0)=10\ln n+\ln4$ and
$w=n^{2/5+o(1)}$, we have
\begin{equation}\label{eq:threshold-scales}
 \varepsilon=n^{-2/15+o(1)},\quad \varepsilon^3w=8192\ln(4m_0),\quad
 \varepsilon w=n^{4/15+o(1)},\quad s=n^{4/15+o(1)},\quad s\le w.
\end{equation}
Take levels starting at $\ell_0=16\varepsilon^2$ and recursively set
$\ell_{i+1}=\ell_i+e_i$, where
$e_i=\varepsilon\sqrt{\ell_i(1-\ell_i)}$.
Retain these levels while $\ell_i\le1-25\varepsilon^2$, and add the
special top level $1-50\varepsilon^2$.

Let $\ell\in[16\varepsilon^2,1-25\varepsilon^2]$ and $e=\varepsilon\sqrt{\ell(1-\ell)}$. We use
four facts.
\begin{enumerate}[label=(L\arabic*),leftmargin=2.5em]
\item $e\le\ell/4$, since $e/\ell=\varepsilon\sqrt{(1-\ell)/\ell}\le\varepsilon/\sqrt\ell\le1/4$.
\item $\ell+5e\le1$, since $1-\ell\ge25\varepsilon^2$ gives $\sqrt{1-\ell}\ge5\varepsilon\ge5\varepsilon\sqrt\ell$,
that is, $1-\ell\ge5e$.
\item $e^2\ge15\varepsilon^4$. The concave function $\ell(1-\ell)$ takes its minimum at an endpoint of
the interval, where it equals $16\varepsilon^2(1-16\varepsilon^2)$ or
$25\varepsilon^2(1-25\varepsilon^2)$; both are at least $15\varepsilon^2$.
\item For $t\in[\ell,\ell+e]$ we have $t(1-t)\le(5/4)\ell(1-\ell)$, since $t\le(5/4)\ell$ by (L1) and
$1-t\le1-\ell$.
\end{enumerate}

There are at most $g=10/\varepsilon$ levels.
Indeed $\arcsin\sqrt\ell$ increases by at least $\varepsilon/3$
at each ordinary step: its derivative is $1/(2\sqrt{t(1-t)})$, so by (L4)
\[
 \arcsin\sqrt{\ell+e}-\arcsin\sqrt\ell
 \ge\frac{e}{2\sqrt{(5/4)\ell(1-\ell)}}=\frac{\varepsilon}{\sqrt5}\ge\frac\varepsilon3.
\]
As $\arcsin\sqrt\ell\in[0,\pi/2]$, the number of ordinary levels is at most $3\pi/(2\varepsilon)+1$.
Adding the top level gives at most $10/\varepsilon$ levels.

The intervals
$(\ell_i+e_i,\ell_i+3e_i]$, together with
$(1-25\varepsilon^2,1]$, cover $[\kappa,1]$. Let $\ell_L$ be the last ordinary level and put
$a_i=\ell_i+e_i=\ell_{i+1}$. Then
$a_0=16\varepsilon^2+4\varepsilon^2\sqrt{1-16\varepsilon^2}<\kappa$, and $a_L>1-25\varepsilon^2$ because
$\ell_L$ is the last ordinary level. Let $t\in[\kappa,1-25\varepsilon^2]$, and let $i$ be the largest
index with $a_i<t$; then $i<L$ and $t\le a_{i+1}$. By (L4), $e_{i+1}\le\sqrt{5/4}\,e_i\le2e_i$, so
$a_{i+1}=\ell_i+e_i+e_{i+1}\le\ell_i+3e_i$. Hence $t\in(\ell_i+e_i,\ell_i+3e_i]$.

\proofstep{Random blocks.}
At every level choose $4n$ independent uniform $w$-subsets $A_b$
of $[n]$. On each block choose $m_0$ independent rows, positive
outside $A_b$, and negative at each coordinate of $A_b$ independently
with probability $\ell_b$. Add the all-positive row and the $n$
single-negative rows. The number of rows is at most
$4ngm_0+n+1\le n^{12}$, because $g=n^{2/15+o(1)}$ by~\eqref{eq:threshold-scales}.
Repeated rows can be deleted; probability
distributions on the row multiset push forward to the resulting class.
We write $e_b$ for the number $e$ at an ordinary level $\ell_b$. For the top level
$\ell_b=1-50\varepsilon^2>3/4$.

We verify that with positive probability all the following hold.
\begin{enumerate}[label=(\roman*),leftmargin=2em]
\item Distinct block supports intersect in at most $16$ coordinates.
\item In an ordinary block, for every $T_0\subseteq A_b$ with
$|T_0|\le s$, a distribution on its rows has negative frequency at
least $\ell_b+3e_b$ at every coordinate of $T_0$.
In a top-level block, some row is negative on all of $T_0$.
\item For every level and every $J\subseteq[n]$ of size at least
$n/(2g)$, some block at that level has $T=J\cap A_b$ of size at
least $y_0=\varepsilon w/100$, and every row of this block has fewer
than $(\ell_b+e_b)|T|$ negative entries on $T$.
At the top level replace this fraction by $1-25\varepsilon^2$.
\item Put $r_0=\lceil4\log_{4/3}m_0\rceil$,
$r_1=\lceil4\log_4m_0\rceil$, and $V=\lceil5\ln w\rceil+1$.
In a block of level at most $3/4$, every $r_0$-subset of $A_b$ is contained
in the negative set of at most $V$ rows. In a block of level greater
than $3/4$, every $r_1$-subset of $A_b$ is contained in the positive set of
at most $V$ rows.
\end{enumerate}

\proofstep{Probability estimates.}
For (i), fix distinct blocks $b,b'$ and the set $A_b$. A fixed $17$-set is contained
in $A_{b'}$ with probability $\prod_{i<17}\frac{w-i}{n-i}\le(w/(n-16))^{17}$. The union bound over
$17$-subsets of $A_b$ gives, for a fixed pair,
probability at most $\binom w{17}(w/(n-16))^{17}=O(w^{34}n^{-17})=O(n^{-17/5})$.
There are at most $(4ng)^2=O(n^2/\varepsilon^2)=n^{34/15+o(1)}$ block pairs, so the
total probability is at most $n^{-17/15+o(1)}$, which tends to zero.

For (ii), fix an ordinary block with level $\ell$ and $e=e_b$, and condition on $A_b$. It suffices
to treat a set $T_1\subseteq A_b$ of size exactly $s$, because every $T_0$ can be padded to such a
set by \eqref{eq:threshold-scales}. Call a row heavy when it has at
least $k=\lceil(\ell+4e)s\rceil$ negative entries on $T_1$. The number of negative entries of a row on
$T_1$ has the distribution $\operatorname{Bin}(s,\ell)$. By (L3) and the definition of $s$,
$es\ge\sqrt{15}\,\varepsilon^2s\to\infty$ uniformly over the retained levels, so $k\le(\ell+4e)s+1\le(\ell+5e)s$; by (L2), $k\le s$.
The value $k$ is a mode of $\operatorname{Bin}(s,k/s)$, because the ratio of the probabilities at $i+1$ and $i$ is
$(s-i)k/((i+1)(s-k))$, which is at least one exactly for $i<k$ (for $k=s$ the distribution is a point mass). So
$\Pr[\operatorname{Bin}(s,k/s)=k]\ge1/(s+1)$, and the quotient of the two binomial probabilities of the
value $k$ is $\exp(-sD(k/s\Vert\ell))$. This gives the type bound
\[
 \Pr[\operatorname{Bin}(s,\ell)=k]\ge\frac{\exp(-sD(k/s\Vert\ell))}{s+1},
\]
where $D$ is the binary relative entropy. Since $D(x\Vert\ell)$ increases for $x\ge\ell$, and
$\ln t\le t-1$ gives $D(a\Vert b)\le(a-b)^2/(b(1-b))$, we have
\[
 D(k/s\Vert\ell)\le D(\ell+5e\Vert\ell)\le\frac{25e^2}{\ell(1-\ell)}=25\varepsilon^2.
\]
Thus a row is heavy with probability at least $\exp(-25\varepsilon^2s)/(s+1)$. Since
$25\varepsilon^2s\le(25/512)\ln m_0$ and $s+1\le n^{0.27}=m_0^{0.027}$, this probability is at least
$m_0^{-1/10}$. The number $N$ of heavy rows is binomial with mean at least
$m_0^{9/10}=n^9$. The Chernoff bound gives $N\ge n^9/2\ge n^{8.5}$, except with
probability at most $\exp(-n^9/8)$.

The heaviness of a row depends only on its number of negative entries on $T_1$, and the entries are
independent and identically distributed, so conditional on being heavy the pattern on $T_1$ is exchangeable. Each
coordinate of $T_1$ therefore has the same conditional negative probability, namely
$E[\#\text{negatives}\mid\text{heavy}]/s\ge k/s\ge\ell+4e$. Conditional on which rows are heavy, their patterns are
independent with this law. Hoeffding's inequality makes the empirical frequency at every
coordinate at least $\ell+3e$, with exceptional probability at most
$s\exp(-2n^{8.5}e^2)$. By (L3) and~\eqref{eq:threshold-scales},
$2n^{8.5}e^2\ge30n^{8.5}\varepsilon^4\ge n^{7.9}$. The number of pairs of a block and a set $T_1$ is at most
$4ng\,w^s=\exp(n^{4/15+o(1)})$, since $s\ln w=n^{4/15+o(1)}$. So the union over all choices of $T_1$ and
all ordinary blocks tends to zero. The uniform distribution on the heavy rows
then works for every $T_0\subseteq T_1$.

For the top level, $\ln(1-x)\ge-2x$ for $0\le x\le1/2$ gives
\[
 (1-50\varepsilon^2)^s\ge\exp(-100\varepsilon^2s)\ge\exp\bigl(-\tfrac{100}{512}\ln m_0\bigr)\ge m_0^{-1/4}.
\]
Thus the probability of no row containing the negative literals of a fixed $T_1$ is at most
$(1-m_0^{-1/4})^{m_0}\le\exp(-m_0^{3/4})=\exp(-n^{7.5})$, which permits the same union.

For (iii), fix a level, a set $J$ with $|J|\ge n/(2g)$, and one block at the level. The intersection
$X=|J\cap A_b|$ is hypergeometric with mean $\mu=w|J|/n\ge w/(2g)=\varepsilon w/20$ and variance at most
$\mu$. By Chebyshev's inequality,
\[
 \Pr[X<y_0]\le\Pr[X<\mu/5]\le\frac{\mu}{(4\mu/5)^2}=\frac{25}{16\mu}\le\frac14,
\]
because $\mu\to\infty$ by~\eqref{eq:threshold-scales}.
For an ordinary level the binary divergence satisfies
$D(\ell+e\Vert\ell)\ge\varepsilon^2/8$.
Indeed, integrating the second derivative
of $D(t\Vert\ell)$ over this entire interval and using (L4) gives
\[
 D(\ell+e\Vert\ell)
 =\int_\ell^{\ell+e}\frac{\ell+e-t}{t(1-t)}\,dt
 \ge\frac{2e^2}{5\ell(1-\ell)}=\frac25\varepsilon^2.
\]
The same lower bound holds for the top-level cutoff, whose gap
is $25\varepsilon^2$ and whose negative-complement probability is
$50\varepsilon^2$. By the complement symmetry of $D$ and the same integral formula, with $x=25\varepsilon^2$,
\[
 D(1-x\Vert1-2x)=D(x\Vert2x)
 =\int_x^{2x}\frac{t-x}{t(1-t)}\,dt
 \ge\int_x^{2x}\Bigl(1-\frac xt\Bigr)dt=x(1-\ln2)\ge\frac{\varepsilon^2}8.
\]
Condition on $A_b$ with $|T|\ge y_0$. The number of negative entries of one row on $T$ is
$\operatorname{Bin}(|T|,\ell_b)$, so by the relative-entropy form of the Chernoff bound it reaches the cutoff
with probability at most $\exp(-|T|\varepsilon^2/8)$. A union bound over the $m_0$ rows bounds the
failure probability by
\[
 m_0\exp(-\varepsilon^2 y_0/8)
 =m_0\exp(-\varepsilon^3w/800)=m_0(4m_0)^{-10.24}<1/4.
\]
One block is therefore good with probability at least $(3/4)(3/4)\ge1/2$.
The $4n$ independent blocks fail simultaneously with probability
at most $2^{-4n}$. A union over all $2^n$ choices of $J$ and all
levels gives at most $g2^{-3n}$, which tends to zero.

For (iv), consider a block of level at most $3/4$. A fixed row contains a prescribed negative $r_0$-set
with probability at most $(3/4)^{r_0}\le m_0^{-4}$. If some $r_0$-set lies in the negative sets of at least
$V$ rows, then it does so for a selected $V$-tuple of rows, and these rows are independent.
A union over the at most $m_0^V$ selected row tuples and $w^{r_0}$
coordinate sets gives probability at most $w^{r_0}m_0^{-3V}$.
Using $r_0\le4\ln m_0/\ln(4/3)+1$ and $V\ge5\ln w$, its logarithm is at most
\[
 \Bigl[\frac4{\ln(4/3)}-15\Bigr]\ln m_0\ln w+\ln w.
\]
Since $4/\ln(4/3)<14$, this tends to minus
infinity like a negative multiple of $(\ln n)^2$, faster than the logarithm of the number of blocks grows.
For positive sets in high-level blocks use
$(1-\ell_b)^{r_1}\le(1/4)^{r_1}\le m_0^{-4}$ instead, together with $r_1\le r_0$; the
same estimate applies. The failure probabilities of the four properties sum to less than one,
so all four properties hold simultaneously for some realization.
Fix such a realization from now on.

\proofstep{Every threshold choice has a large induced obstruction.}
Fix $0\le r_j<p_j\le1$. We distinguish two cases.

Suppose first that $p_j<\kappa$ on at least $n/2$ coordinates $J$. Every consistent
literal set on $J$ of size at most $\lfloor1/\kappa\rfloor$ is a
face. To witness a set $+J_1\cup-J_2$, use the uniform distribution on the single-negative
rows named by its negative literals $J_2$; the negative frequency on $J_2$ is
$1/|J_2|\ge\kappa>p_j$. If there are none, use
the all-positive row. The positive literals then have frequency zero, and $0\le r_j$.
Since $r_j<p_j$, no face contains two opposite literals.
If every consistent set is present, the induced complex on the literals of $J$ is the full
cross-polytope sphere, with nonzero homology in degree $|J|-1$, so $\mathcal L\ge|J|\ge\lfloor1/\kappa\rfloor$.
Otherwise a smallest consistent nonface has
size greater than $\lfloor1/\kappa\rfloor$ and is a minimal nonface. In either case the Leray number is at least
$\lfloor1/\kappa\rfloor$.

In the second case at least $n/2$ coordinates have $p_j\ge\kappa$.
The covering intervals and the bound on the number of levels give a level and a set $J$ of size at least
$n/(2g)$ whose thresholds lie in its interval. Choose the block from
(iii) and put $T=J\cap A_b$, viewed as negative literals.
At an ordinary level, $p_j>\ell_b+e_b$ on $J$; at the top level, $p_j>1-25\varepsilon^2$. So every row of this
block has fewer negatives in $T$ than
$\sum_{j\in T}p_j$. Rows of any other block have at most $16$
negatives there, by (i), and the additional rows have at most one.
For large $n$, $16<(\ell_b+e_b)y_0$, including the lowest level, since
$(\ell_b+e_b)y_0\ge16\varepsilon^2\cdot\varepsilon w/100=(16/100)8192\ln(4m_0)$; the same holds with the
top-level fraction. Thus every row has fewer than $\sum_{j\in T}p_j$ negatives in $T$. For a
distribution $q$ on the rows, $\sum_{j\in T}q\{c_j=-1\}$ is the $q$-expectation of the number of negatives in
$T$, so it is less than $\sum_{j\in T}p_j$.
Consequently no distribution on all rows satisfies the constraints
on $T$, and $-T$ is a nonface. Every subset of $T$ of size at most $s$ does satisfy them,
by (ii) and the upper endpoint of the level interval: at an ordinary level the frequency is at
least $\ell_b+3e_b\ge p_j$, and at the top level a row gives frequency $1\ge p_j$.
Hence a minimal nonface contained in $-T$ has size greater than $s$, giving
$\mathcal L\ge s$.

Combining the two cases, and using $\ln m_0/512\ge1/20$, gives
$\mathcal L\ge\lfloor1/(20\varepsilon^2)\rfloor$. Finally,
\[
 \frac1{20\varepsilon^2}=\frac1{20}\Bigl(\frac{w}{8192\ln(4m_0)}\Bigr)^{2/3}
 \ge\frac{n^{4/15}}{20(8192\cdot11\ln n)^{2/3}},
\]
which gives the asserted rate with an absolute constant, uniformly in $r,p$.

\proofstep{A completion with a short game.}
Let $H$ contain $C$, every row with at most $r_0$ negatives, and,
for each block of level greater than $3/4$, every row negative
exactly on $A_b\setminus Q$, where $Q\subseteq A_b$ and $|Q|\le r_1$. We call the last
rows completion rows.
Consider a position $-P\cup+J$, where $P,J\subseteq[n]$ are disjoint.

If $|P|\le r_0$, predict the row negative
exactly on $P$. It belongs to $H$ and extends the position. The literal $+j$ with $j\in P$ and the
literal $-j$ with $j\in J$ are inconsistent with the position, so every counterexample is a
literal $-j$ with $j\notin P\cup J$. It adds a negative literal, and this phase has at most $r_0+1$ moves.

Let $|P|>r_0$. Since $r_0>16$, every compatible row lies in a unique block: a compatible row of $H$ has
more than $r_0$ negatives, so it is a block row or a completion row, with negative set in one support
$A_b$, and $P$ cannot lie in two supports because
distinct supports intersect in at most $16$ positions. This block $b$ stays the same at all
subsequent positions, because $P$ only grows.

If that block has level greater than $3/4$ and
$|J\cap A_b|\le r_1$, predict the completion row negative exactly on
$A_b\setminus(J\cap A_b)$. It extends the position, since $P\subseteq A_b\setminus J$. A valid counterexample must add a
positive literal inside $A_b\setminus(P\cup J)$. Indeed, the literals $+j$, $j\in P$, and $-j$, $j\in J$, are
inconsistent; and a negative outside the block would
leave no compatible row, since the negative set of a compatible row would contain $P$ and lie in the unique support
$A_b$. After at most $r_1+1$ such moves we have $|J\cap A_b|>r_1$, and then every completion row of this block has been
excluded, since it is positive on at most $r_1$ coordinates of $A_b$.

In the last case the compatible rows of $H$ are original block rows, and at most $V$ of them are compatible,
by (iv): for level at most $3/4$ apply it to an $r_0$-subset of $P$, and for level greater than $3/4$
to an $r_1$-subset of $J\cap A_b$. Predict any one of them; each move eliminates that row and
leaves a compatible row, so this phase has fewer than $V$ moves.

The game has at most $r_0+r_1+V+2=O(\ln n)$ moves, since $\ln m_0=10\ln n$ and $\ln w=O(\ln n)$.
Lemma~\ref{lem:hub-game} proves $u(C)=O(\ln n)$.
\end{proof}

\section{Stabilization by independent coordinates}\label{app:stable}
Theorem~\ref{thm:main} bounds the list number of every class by a logarithmic product with the
constant $805$. This section lowers the explicit constant to $105.09$, at the cost of adjoining
independent binary coordinates, each of which adds one to the bound. The additional coordinates
provide room for general position. In the proof of Theorem~\ref{thm:main} the carriers must be highly
connected inside cube faces of nearly the same dimension, so general position does not make their
maps injective, and the weighted hulls of Section~\ref{app:hulls} control the multiple points over many
stages. When the number of coordinates is at least twice the number of prescribed literals, compatible
maps can be perturbed to embeddings and their images are themselves carriers, so a single connected
family and a single critical-point estimate suffice. The same construction with fewer prescribed
literals also covers small numbers of adjoined coordinates, including none.

For a class with $m$ rows on $n$ coordinates, put $\mathcal B_t=\{\pm1\}^t$, and define
\begin{align*}
 p&=2\lceil\ln(2m)\rceil,&q&=1+\frac1{2\ln(2n)},&\sigma&=\frac1{4n},\\
 D_*&=\frac{2(p-1)}{q-1}m^{2/p}n^{2-2/q}(4/3)^2,
 &D&=\lfloor D_*\rfloor,&a_0&=n-D-2.
\end{align*}
The integer $p\ge2$ is even, $1<q\le1+1/(2\ln2)<2$, and $D$ is an integer, as the range
conditions of the following result require. The exponents are chosen so that $m^{2/p}$ and
$n^{2-2/q}$ stay bounded, which gives $D_*$ the order of $\ln(2m)\ln(2n)$.

\begin{theorem}[Stabilized bound]\label{thm:stable}
Let $C$ be a nonempty class with $m$ rows on $n$ coordinates, and let $t$ be a nonnegative integer.
Then
\begin{align*}
 \operatorname{lr}(C\times\mathcal B_t)&\le t+D+2
 &&\text{if }t\ge\max\{0,n-2D-4\},\\
 u(C\times\mathcal B_t)&\le t+D+2
 &&\text{if }t\ge\max\{0,n-2D-3\},
\end{align*}
and $D_*<105.09\ln(2m)\ln(2n)$.
Also, if $n+t\ge5$ and $\lfloor(n+t)/2\rfloor\le a_0$, then
\[
 \operatorname{lr}(C\times\mathcal B_t)\le\lceil(n+t)/2\rceil.
\]
\end{theorem}
For comparison, Theorem~\ref{thm:main}, the additive join inequality
of~\cite[Theorem 1.12]{BHHLT26}, and the cube value of Theorem~\ref{res:5.1} give
$\operatorname{lr}(C\times\mathcal B_t)\le t+805\ln(2m)\ln(2n)$ for every $t$. Theorem~\ref{thm:stable}
replaces the term $805\ln(2m)\ln(2n)$ by $D+2$ when $t\ge\max\{0,n-2D-4\}$.

The last bound of the theorem treats the complementary range of $t$. Its hypothesis
$\lfloor(n+t)/2\rfloor\le a_0$ is equivalent to $t\le n-2D-3$, and for $t=n-2D-4$ and
$t=n-2D-3$ the numbers $t+D+2$ and $\lceil(n+t)/2\rceil$ coincide. At $t=0$ it gives
$\operatorname{lr}(C)\le\lceil n/2\rceil$ for $n\ge2D+3$, with no additional coordinate.
This is smaller than the bound of Theorem~\ref{thm:main} as long as
$\lceil n/2\rceil<805\ln(2m)\ln(2n)$, that is, for $n$ up to about $1610\ln(2m)\ln(2n)$.

The proof has three parts. We first construct highly connected sets outside the realizable
directions. We then transfer them to the cube and show how enough coordinates turn compatible maps
into carriers. Finally, we pass to the product and prove the three assertions.

\subsection{A connected family outside the realizable directions}
We work on a smoothed $\ell_q$ sphere $\Sigma$, on which the function $F$, a sum of even powers of the
row correlations, is large in every realizable direction. The goal is a sublevel set of a small
perturbation of $F$ that avoids these directions and stays highly connected after the signs of a few coordinates are prescribed. As in
Section~\ref{app:morse}, the connectivity comes from a trace bound on the number of nonnegative Hessian
directions at high critical points. In this section the prescribed signs are imposed by linear walls
instead of logarithmic barriers, and no stages are needed.

Write $r_j(y)=(y_j^2+\sigma^2)^{1/2}$ and set
\[
 G(y)=\sum_jr_j(y)^q,\quad\Sigma=\{G=1\},\qquad
 F(y)=\sum_{c\in C}\langle c,y\rangle^p.
\]
Choose $\delta>0$ with $n\delta\le3/(8e)$. For a consistent literal set $S$ on the coordinates of $C$, put
$Z_S=\{y\in\Sigma:sy_j\ge\delta\ ((j,s)\in S)\}$,
with $Z_\varnothing=\Sigma$.
For $x$ in the realizable space $X(C)$ of Section~\ref{app:cube-map}, the
\emph{realizable radial direction} of $x$ is the point of $\Sigma$ on the ray through $x$.

\begin{lemma}[Connected sublevel sets]\label{lem:stable-connectivity}
There are an arbitrarily small linear perturbation
$\widetilde F(y)=F(y)+\langle\ell,y\rangle$ and a common regular
level $b>(3/4)^p$ such that
$\mathcal K=\{y\in\Sigma:\widetilde F(y)\le b\}$ misses all
realizable radial directions, and
\[
 \mathcal K\cap Z_S\text{ is }(a_0-|S|-1)\text{-connected}
 \quad (|S|\le a_0),
\]
including the empty literal set. Connectivity $-1$ means nonemptiness.
\end{lemma}
\begin{proof}
\proofstep{Convexity of the body.}
The function $t\mapsto(t^2+\sigma^2)^{q/2}$ has second derivative
$q(t^2+\sigma^2)^{q/2-2}(\sigma^2+(q-1)t^2)>0$. So $G$ is smooth, strictly convex and
proper. If $G(y^0)<1$, then each ray from $y^0$ meets $\Sigma$ exactly once, and
convexity gives
\begin{equation}\label{eq:G-gradient}
 \langle\nabla G(y),y-y^0\rangle\ge G(y)-G(y^0)>0\qquad(y\in\Sigma).
\end{equation}
Since $r_j^q\le G=1$ on $\Sigma$, we have $r_j\le1$ and hence $|y_j|<1$ on $\Sigma$. Also $G(0)=n(4n)^{-q}<1$.

\proofstep{A level separating the realizable directions.}
For $x\in X(C)$, $\|x\|_1=1$ and some row satisfies
$\langle c,x\rangle=1$. Subadditivity of the $q/2$ power gives
$r_j(y)^q\le|y_j|^q+\sigma^q$, and $|x_j|^q\le|x_j|$. Therefore
\[
 G(3x/4)\le(3/4)^q\sum_j|x_j|^q+n\sigma^q\le(3/4)^q+n(4n)^{-q}<\frac34+\frac14=1.
\]
Since $G(0)<1$, the ray through $x$ meets $\Sigma$ at a single point $\lambda_xx$, and the last
inequality gives $\lambda_x>3/4$. All summands of $F$ are nonnegative because $p$ is even, so
$F(\lambda_xx)\ge\lambda_x^p\langle c,x\rangle^p=\lambda_x^p>(3/4)^p$.
The function $x\mapsto\lambda_x$ is continuous by \eqref{eq:G-gradient} and the implicit
function theorem. Compactness of $X(C)$ gives a strict gap above this threshold: the
minimum $b_0$ of $F(\lambda_xx)$ over $X(C)$ satisfies $b_0>(3/4)^p$.

\proofstep{The coordinate walls.}
The point $y^S$ with coordinates $s\delta$ on $S$ and zero elsewhere is
strictly inside $\{G\le1\}$, because $n^2(\delta^2+\sigma^2)\le(3/(8e))^2+1/16<1/9$ gives
$G(y^S)\le n(\delta^2+\sigma^2)^{q/2}<n(3n)^{-q}<1$.
Since $sy_j\ge\delta$ is the same as $s(y_j-y^S_j)\ge0$ on $S$,
radial projection from this point identifies
$Z_S$, for nonempty $S$, with a closed spherical orthant sector.
It is a ball of dimension $n-1$, whose boundary consists of coordinate walls.

These walls are transverse. For a consistent set $T$ let
$W_T=\{y\in\Sigma:sy_j=\delta\ ((j,s)\in T)\}$, with $W_\varnothing=\Sigma$. By
\eqref{eq:G-gradient} with $y^0=y^T$, the vector $y-y^T$, which is tangent to the affine subspace
$\{sy_j=\delta\ ((j,s)\in T)\}$, has positive derivative of $G$. This vector is orthogonal to the
$e_j$ for $j$ on $T$. Hence the vectors $\nabla G(y)$
and $e_j$, for $j$ on $T$, are linearly independent on $W_T$.
Every stratum $W_T$ is a smooth sphere of
dimension $n-1-|T|$, by radial projection from $y^T$ within the affine
subspace, or is empty, when $|T|=n$.

\proofstep{A generic linear perturbation.}
Choose a generic sufficiently small $\ell$. The restriction of
$\widetilde F$ to each of the finitely many wall strata $W_T$ is Morse for almost every $\ell$
\cite[Chapter 1, Section 7]{GP74}.
For almost every $\ell$, the normal derivatives of these restrictions at their critical points are
also nonzero. To prove this, consider a zero normal derivative at a critical point $y$ of
$\widetilde F|_{W_T}$, for a literal $v\in T$. It makes the
point critical on the stratum $W_{T-v}$ with that wall removed. That means
$\nabla F(y)+\ell=\nu$ with $\nu$ in the span of $\nabla G(y)$ and $e_j$, $j$ on $T-v$.
The pairs $(y,\nu)$ with $y\in W_T$ form the normal bundle of $W_{T-v}$ restricted to $W_T$, a manifold
of dimension $(n-1-|T|)+|T|=n-1$. So the image of the map
$(y,\nu)\mapsto\nu-\nabla F(y)$ has measure zero in the space of
linear perturbations, by Sard's theorem.

At a critical point $y$ of $\widetilde F|_{W_T}$ there are unique multipliers with
\begin{equation}\label{eq:stable-kkt}
 \nabla\widetilde F(y)=\lambda\nabla G(y)+
              \sum_{(j,s)\in T}\mu_{j,s}s e_j,
\end{equation}
and the nonzero normal derivatives say that $\mu_v\ne0$ for all $v\in T$, where $\mu_v=\mu_{j,s}$
for $v=(j,s)$.
Such a point lies on no further wall. Indeed, if $y\in W_{T+v}$ with $v\notin T$, then $y$ would be
critical on $W_{T+v}$ and on $W_T$, which is a zero normal derivative for $W_{T+v}$. Also, a literal
opposite to one in $T$ cannot be active at $y$.

Fix $\ell$ so small that $b_1:=\sup_\Sigma|\langle\ell,y\rangle|$ satisfies
$3b_1<b_0-(3/4)^p$ and $n\|\ell\|_\infty\le p(3/4)^p/10$. There are finitely many critical
values on the finitely many compact strata. Choose a level $b$ regular on every stratum,
below the minimum of $\widetilde F$ on the realizable radial directions, and with
$b>(3/4)^p+2b_1$. This is possible since that minimum is at least $b_0-b_1>(3/4)^p+2b_1$. If
$\widetilde F(y)\ge b$, then $F(y)\ge b-b_1>(3/4)^p$. By construction,
$\mathcal K$ misses all realizable radial directions.

\proofstep{A lower bound for the multiplier.}
Let $y$ be a critical point of a stratum $W_T$
with $\widetilde F(y)\ge b$, satisfying \eqref{eq:stable-kkt}.
Pair this equation with $y$ only on the free coordinates $j\notin T$.
Since $\partial_jG=qy_jr_j^{q-2}$, the walls contribute nothing and
\begin{equation}\label{eq:free-sum}
 \sum_{j\notin T}y_j(\partial_jF(y)+\ell_j)=\lambda q\sum_{j\notin T}y_j^2r_j^{q-2}.
\end{equation}
Euler's identity gives $\sum_jy_j\partial_jF=pF$.
H\"older's inequality gives
$|\partial_jF|\le p\sum_c|\langle c,y\rangle|^{p-1}\le pm^{1/p}F^{1-1/p}$. Since
$p\ge2\ln(2m)$ we have $m^{1/p}\le\sqrt e$, and $F^{-1/p}<4/3$, so
\[
 \|\nabla F\|_\infty\le p m^{1/p}F^{1-1/p},\qquad
 n\delta m^{1/p}F^{-1/p}\le\frac3{8e}\cdot\sqrt e\cdot\frac43=\frac1{2\sqrt e}.
\]
On coordinates of $T$ we have $|y_j|=\delta$, so
$|\sum_{j\in T}y_j\partial_jF|\le n\delta\|\nabla F\|_\infty\le\tfrac1{2\sqrt e}pF$.
Also $|\sum_{j\notin T}y_j\ell_j|\le n\|\ell\|_\infty\le\tfrac1{10}pF$, because $|y_j|<1$ and
$F>(3/4)^p$.
Since $1-\tfrac1{2\sqrt e}-\tfrac1{10}>\tfrac12$,
the left side of \eqref{eq:free-sum} is at least
$pF/2>0$. The coefficient of
$\lambda q$ on the right lies in $[0,1]$: indeed $y_j^2\le r_j^2$ gives
$y_j^2r_j^{q-2}\le r_j^q$, and $\sum_jr_j^q=1$. This proves $\lambda q\ge pF/2$.

\proofstep{The number of nonnegative Hessian directions.}
Let $A$ have rows $c\in C$, $W=\operatorname{diag}
(\langle c,y\rangle^{p-2})$, with diagonal entries one if $p=2$, and
$R=\operatorname{diag}(r_j^{q-2})$. The walls are linear
constraints, so
the Hessian of the constrained function, on the tangent space of $W_T$ at $y$, is the
restriction of $\nabla^2F-\lambda\nabla^2G$. We have $\nabla^2F=p(p-1)A^{\mathsf T}WA$. Also
$\nabla^2G\succeq q(q-1)R$, because the $j$th diagonal entry of $\nabla^2G$ is
$qr_j^{q-4}(\sigma^2+(q-1)y_j^2)\ge q(q-1)r_j^{q-2}$, as $q-1\le1$. With $\lambda q\ge pF/2$, the
Hessian is bounded above by
\[
 p\bigl((p-1)A^{\mathsf T}WA-(q-1)FR/2\bigr).
\]
A subspace of tangent directions on which the Hessian is nonnegative is a subspace on which this
form is nonnegative. By the Courant--Fischer theorem and Sylvester's law of inertia, after
conjugation by $R^{-1/2}$,
the number of nonnegative directions
is at most the number of eigenvalues at least $(q-1)F/2$ of the positive semidefinite matrix
$(p-1)R^{-1/2}A^{\mathsf T}WAR^{-1/2}$. Hence it is bounded by the trace of this
matrix divided by $(q-1)F/2$.
Because $c_j^2=1$, the trace is
$(p-1)\bigl(\sum_c\langle c,y\rangle^{p-2}\bigr)\bigl(\sum_jr_j^{2-q}\bigr)$.
H\"older's inequality supplies
\[
 \sum_c|\langle c,y\rangle|^{p-2}\le m^{2/p}F^{1-2/p},
 \qquad \sum_jr_j^{2-q}\le n^{2-2/q}\Bigl(\sum_jr_j^q\Bigr)^{(2-q)/q}=n^{2-2/q}.
\]
The number of nonnegative directions is therefore at most
$2(p-1)(q-1)^{-1}m^{2/p}n^{2-2/q}F^{-2/p}<D_*$. Being an integer, it is at most $D$.
This bound uses no information on the signs of the wall multipliers.

\proofstep{The local form at a critical point.}
Fix $S$ with $|S|\le a_0$. Then $|S|<n$, so $Z_S$ is a ball for nonempty $S$, and
$Z_\varnothing=\Sigma$ is a sphere.
Let $y\in Z_S$ be a critical point on walls $T\subseteq S$, that is, a critical point of
$\widetilde F|_{W_T}$. By the choice of $\ell$, no other wall of $S$ is active at $y$, the
multipliers $\mu_v$ are nonzero, and the restriction of $\widetilde F$
to the stratum is Morse. Let
$w_v=s_vy_{j_v}-\delta$ for $v=(j_v,s_v)\in T$ and let $x$ be coordinates on $W_T$, extended to a
neighborhood of $y$ in $\Sigma$. Then $(w,x)$ is a
chart of $\Sigma$ in which $Z_S$ is $\{w\ge0\}$. By \eqref{eq:stable-kkt}, the differential of
$\widetilde F$ on $T_y\Sigma$ is $\sum_v\mu_v\,dw_v$. Hadamard's lemma in the wall-normal variables
gives
$\widetilde F(w,x)=f(x)+\sum_vw_vg_v(w,x)$, with $f=\widetilde F|_{W_T}$ and $g_v(0,0)=\mu_v$. The new
variables $w_v|g_v(w,x)|$, again denoted $w_v$, preserve each wall and the sign of $w_v$. The
Morse lemma \cite[Lemma 2.2]{Milnor63} for $f$ in the wall-tangent variables gives local coordinates
$w_v\ge0,u_+,u_-$ with
\[
 \widetilde F=\widetilde F(y)+
       \sum_{v\in T}\operatorname{sign}(\mu_v)w_v
       +|u_+|^2-|u_-|^2.
\]
For a critical point with $\widetilde F(y)>b$, the dimension of $u_+$ is at most $D$ by the preceding
step.

\proofstep{A decreasing vector field.}
In these coordinates, the local field
$-2u_+\partial_{u_+}+2u_-\partial_{u_-}
-\sum_v\operatorname{sign}(\mu_v)w_v\partial_{w_v}$
is tangent to every wall. It decreases $\widetilde F$ on $\{w\ge0\}$ away from the origin, since the
derivative of $\widetilde F$ along it is $-4|u_+|^2-4|u_-|^2-\sum_vw_v$.
Away from critical points choose tangent
decreasing fields: at a point of a stratum which is not critical for that stratum, use a chart
$(w,x)$ of the same kind and a constant field in the $x$-variables along which $\widetilde F$
decreases. Combine them by a partition of unity, using the local fields above alone near the
critical points. The resulting smooth field $V$ is tangent to all walls of $S$, and
$V\widetilde F<0$ away from the finitely many critical points. Its flow $\phi_t$ therefore preserves every
wall and is defined on $Z_S$ for all $t\in\mathbb R$.

\proofstep{Stable sets and the exit retraction.}
Let $\mathcal P$ be the finite set of critical points in $Z_S$ with $\widetilde F>b$.
In the chart at $y\in\mathcal P$, the flow multiplies $u_+$ by $e^{-2t}$, $u_-$ by $e^{2t}$, and
$w_v$ by $e^{-\operatorname{sign}(\mu_v)t}$.
So the local stable set of $y$ has $u_-=0$ and has $w_v=0$ whenever
$\mu_v<0$, and at most $|S|$ wall directions are left. The global stable set is the increasing union of
the images of this local piece under $\phi_{-t}$, $t\ge0$. Thus the stable sets of the points of
$\mathcal P$ are finitely many immersed
submanifolds of dimension at most $D+|S|$. Write $\mathrm{Sp}$ for the union of these
stable sets. Then $\widetilde F>b$ on $\mathrm{Sp}$.

Every point of $Z_S\setminus\mathrm{Sp}$ above level $b$ reaches this level in finite time.
Otherwise its decreasing trajectory stays above level $b$. Its limit set is connected and
consists of critical points, so it is a single point of the finite set
of critical points, with value above $b$ by regularity, and the original point
belongs to a stable set. Regularity makes the
exit time continuous: on the level $\{\widetilde F=b\}\cap Z_S$ we have $V\widetilde F<0$, so the implicit
function theorem applies. The flow thus gives a retraction
\[
 \mathcal R:Z_S\setminus\mathrm{Sp}\to\mathcal K\cap Z_S,
\]
the identity on $\mathcal K\cap Z_S$ and the exit point on the rest.

\proofstep{Spheres in the sublevel set.}
Let $0\le i\le a_0-|S|-1$. Push a sphere in $\mathcal K\cap Z_S$ slightly below $b$, by a short
time of the flow, and into
the relative interior of $Z_S$, by a small push along a collar of the ball. These homotopies stay in
$\mathcal K\cap Z_S$. Extend the sphere to a disk in the relative interior of the ball $Z_S$.
For $S=\varnothing$, extend in the sphere $\Sigma$, which is possible
in all the dimensions at issue, since $i<n-1$. Relative transversality
\cite[Chapter 2, Section 3]{GP74} makes this
disk avoid all stable sets, which are countable unions of embedded pieces, whenever
\[
 (i+1)+(D+|S|)<n-1.
\]
The boundary sphere misses them already, being below level $b$, so the perturbation can be taken
relative to a neighborhood of the boundary.
The displayed inequality holds for $i\le a_0-|S|-1$, as required, since then
$(i+1)+D+|S|\le a_0+D=n-2$.
Composing with the retraction gives a null homotopy inside the
sublevel set.

The same dimension argument with a point gives nonemptiness: $D+|S|<n-1$ for
$|S|\le a_0$, so $\mathrm{Sp}$ does not fill $Z_S$, and $\mathcal R$ sends a point outside
$\mathrm{Sp}$ into $\mathcal K\cap Z_S$.
\end{proof}

\subsection{Compatible sections and embeddings}
This subsection transfers the connected family to the cube. A coordinate clamp and an induction over
an order complex give a map that is compatible with the faces of the cube and avoids the class. When
the number of coordinates is large compared with the number of prescribed literals, general position
then turns such a map into the nested acyclic carriers required by Theorem~\ref{thm:carrier}.

Fix $\mathcal K$ as in Lemma~\ref{lem:stable-connectivity}.
Let $\mathcal Q_S(C)$ be the union of closed cube faces in $Q_S$
whose vertices avoid $C$, where $S$ is a consistent literal set and $S=\varnothing$ is allowed, with
$Q_\varnothing=Q$. For a scale $\zeta>0$ define the coordinate clamp $\psi_\zeta$ by
$\psi_\zeta(y)_j=\max\{-1,\min\{1,y_j/\zeta\}\}$.

\begin{lemma}[Clamping into faces that avoid the class]\label{lem:stable-clamp}
\textup{(a)} Every $\mathcal Q_S(C)$ is disjoint from $N[C]$, and more generally from every point $y\in Q$ with
$c_jy_j\ge0$ for all $j$ and some $c\in C$.

\textup{(b)} There is a scale $\zeta$ with $0<\zeta<\delta$ such that $\psi_\zeta$ sends
$\mathcal K\cap Z_S$ into $\mathcal Q_S(C)$ for all consistent literal sets $S$ on the coordinates of $C$,
including $S=\varnothing$.
\end{lemma}
\begin{proof}
For (a), suppose that such a point lay in a closed face $B$ of $\mathcal Q_S(C)$.
Then $c$ would agree with the fixed coordinates of $B$, so $c$ would be a vertex of $B$, which is
excluded.

For (b), given $y\in\Sigma$, let $B_\zeta(y)$ be the cube face
fixing coordinate $j$ at $\operatorname{sign}y_j$ when $|y_j|\ge\zeta$ and leaving it free otherwise. Then
$\psi_\zeta(y)\in B_\zeta(y)$. We claim that for small $\zeta$, no $B_\zeta(y)$ with $y\in\mathcal K$ has
a vertex in $C$. A vertex $c$ of $B_\zeta(y)$ has $c_jy_j\ge-\zeta$ for every $j$. If the claim failed for
arbitrarily small clamp scales, compactness
would give $y\in\mathcal K$ and $c\in C$ with $c_jy_j\ge0$ for
every $j$. Since $y\ne0$, this gives $y/\|y\|_1\in X(C)$, so $y$ is a realizable radial direction,
contradicting Lemma~\ref{lem:stable-connectivity}.
Choose the scale smaller than $\delta$ so the prescribed faces are
preserved: if $y\in\mathcal K\cap Z_S$, then $B_\zeta(y)\subseteq Q_S$, so $\psi_\zeta(y)\in\mathcal Q_S(C)$.
\end{proof}

The prescribed literals are organized by an order complex. For an integer $a$ with $0\le a\le a_0$, take
the order complex of the consistent literal sets
of size at most $a$, including the empty set. Let $e_S$ be its
subcomplex of chains whose least member contains $S$.
It is a cone of dimension $a-|S|$ with apex the vertex $S$, with
$\partial e_S=\bigcup_v e_{S+v}$: a chain whose least member strictly contains $S$ lies in some
$e_{S+v}$, and adding $S$ to such a chain gives a chain in $e_S$. Moreover
$e_{S+v}\cap e_{S+w}=e_{S+v+w}$ when $S+v+w$ is consistent of size at most $a$, and this
intersection is empty otherwise.

\begin{lemma}[Compatible section]\label{lem:stable-section}
For every integer $a$ with $0\le a\le a_0$ there is a continuous map
\begin{equation}\label{eq:stable-section}
 \Phi_a:|e_\varnothing|\to Q,\qquad\Phi_a(e_S)\subseteq\mathcal Q_S(C),
\end{equation}
for every consistent literal set $S$ of size at most $a$, including $S=\varnothing$.
\end{lemma}
\begin{proof}
We map a chain with least member $S$ into $\mathcal K\cap Z_S$, by
induction on its dimension, using
Lemma~\ref{lem:stable-connectivity}. A vertex $S$ goes to a point of the nonempty set
$\mathcal K\cap Z_S$. Let a chain $S=S_0\subsetneq\cdots\subsetneq S_j$ be a $j$-simplex, so
$j\le a-|S|$. Its faces with least member $S$ are mapped into $\mathcal K\cap Z_S$, and the face
deleting $S_0$ is mapped into $\mathcal K\cap Z_{S_1}\subseteq\mathcal K\cap Z_S$. Thus
the boundary of the $j$-simplex
lies in this set and $j-1\le a_0-|S|-1$, so the map extends over the simplex.

Let $\Phi_a$ be this map composed with the clamp of Lemma~\ref{lem:stable-clamp}(b). A simplex of $e_S$ is a
chain with least member $S'\supseteq S$. It is mapped into
$\mathcal K\cap Z_{S'}\subseteq\mathcal K\cap Z_S$, and the clamp sends this set into $\mathcal Q_S(C)$.
\end{proof}

The next lemma is general position in the relative form that we need. It keeps a map fixed on a
subcomplex on which it is already an affine embedding, and it pushes the rest of the image into the
interior of the ambient convex set. This second property separates the images of different faces in the
induction that follows.

\begin{lemma}[Relative general position]\label{lem:relative-general-position}
Let $B\subseteq\mathbb R^M$ be a compact convex set with nonempty interior, $V\subseteq B$ relatively
open, $X$ a finite simplicial complex of dimension $k$ with $2k+1\le M$, and $X_0\subseteq X$ a
subcomplex. Let $g:|X|\to V$ be continuous, injective on $|X_0|$ and affine on each simplex of
$X_0$. Then there is an injective map $f:|X|\to V$, affine on each simplex of a subdivision of
$X$, equal to $g$ on $|X_0|$, sending $|X|\setminus|X_0|$ into $\operatorname{int}B$, and homotopic
to $g$ relative to $|X_0|$ inside $V$.
\end{lemma}
\begin{proof}
Choose $\varepsilon>0$ such that the $\varepsilon$-neighborhood of $g(|X|)$ in $B$ lies in $V$, and a
subdivision $X'$ of $X$ on whose simplices $g$ oscillates by less than $\varepsilon/2$ and in which the
subdivision $X_0'$ of $X_0$ is a full subcomplex. List the vertices of $X'$ outside $X_0'$ as
$u_1,\ldots,u_N$. Put $f=g$ on the vertices of $X_0'$. Choose successively
$f(u_i)\in\operatorname{int}B$ within $\varepsilon/2$ of $g(u_i)$, outside every proper affine subspace of
$\mathbb R^M$ spanned by points already chosen. Extend $f$ affinely over the simplices of $X'$. Then
$f=g$ on $|X_0|$ and $\|f-g\|<\varepsilon$, so $f$ maps into $V$ and the straight-line homotopy stays in
$V$. A point outside $|X_0|$ has, by fullness, a positive barycentric coordinate at some $u_i$, so
its image is in $\operatorname{int}B$ by convexity.

For injectivity, let $x,x'$ lie in the relative interiors of simplices $\sigma,\tau$ of $X'$ with
$f(x)=f(x')$. Split each vertex set into its part in $X_0'$ and its part among the $u_i$:
$\sigma_0\sqcup\sigma_1$ and $\tau_0\sqcup\tau_1$; by fullness, $\sigma_0$ and $\tau_0$ span simplices of
$X_0'$ when nonempty. The set $\sigma\cup\tau$ has at most $2k+2\le M+1$ vertices, so before each
$u\in\sigma_1\cup\tau_1$ is placed, the earlier points of $f(\sigma\cup\tau)$ span a proper
affine subspace, which $f(u)$ avoids. Let $\alpha$ and $\alpha'$ be the barycentric coordinates of $x$
and $x'$, extended by zero, so that $\sum_w(\alpha_w-\alpha'_w)f(w)=0$ with coefficients of sum zero. If
they differed at some $u\in\sigma_1\cup\tau_1$, then the last such $u$ in the order of placement
would exhibit $f(u)$ as an affine combination of earlier points of $f(\sigma\cup\tau)$, which is
impossible. So $\sigma_1=\tau_1$ with equal coordinates. Removing these equal parts and rescaling
leaves a point of $g(|\sigma_0|)$ equal to a point of $g(|\tau_0|)$, unless $\sigma_0$ and $\tau_0$ are
empty. Injectivity of $g$ on $|X_0|$ then gives $x=x'$.
\end{proof}

When the number of coordinates is at least twice the number of prescribed literals, general position
converts a compatible map into a list bound, as follows.

\begin{lemma}[From compatible maps to list bounds]\label{lem:stable-embedding}
Let $C'\subseteq\{\pm1\}^{n'}$ be nonempty, $a\ge1$, $n'\ge2a$, and
$n'-a-1\ge2$. Form $e_S$ in the order complex of the nonempty consistent literal sets of size at most
$a$ on $n'$ coordinates. Suppose there is a compatible map, that is, a continuous map $\Phi$ on this
complex with $\Phi(e_S)\subseteq Q_S\setminus N[C']$ for all nonempty $S$ of size at most $a$. Then
$\operatorname{lr}(C')\le n'-a$.
\end{lemma}
\begin{proof}
Write $N=N[C']$. The set $Q_S\setminus N$ is relatively open in $Q_S$.
We modify the maps by downward induction on $|S|$, together with
compatible homotopies, so that each $e_S$ is embedded. That is, we construct PL injections
$\iota_S:e_S\to Q_S\setminus N$ with $\iota_S|_{e_{S+v}}=\iota_{S+v}$ and
$\iota_S(e_S\setminus\partial e_S)\subseteq\operatorname{int}Q_S$, and homotopies from
$\Phi|_{e_S}$ to $\iota_S$ inside $Q_S\setminus N$ whose restriction to $e_{S+v}$ is the homotopy of
$S+v$. Homotopies of one level have equal duration, and an earlier homotopy is extended by a constant
in time when it is continued.

\proofstep{The top level.}
For $|S|=a$, the complex $e_S$ is a single point. Move
this point inside $Q_S\setminus N$ into the relative interior of the face.

\proofstep{The boundary is embedded.}
Let $|S|<a$ and suppose the construction is done for all larger literal sets. The boundary
$\partial e_S$ is already embedded, since its pieces
have disjoint relative interiors and agree on their intersections. Indeed, a point of
$\partial e_S$ has a carrier chain with least member $R\supsetneq S$, and $\iota_R$ sends it into
$\operatorname{int}Q_R$. The interiors of distinct faces are disjoint, and points with the same $R$ are
separated by the injection $\iota_R$.

\proofstep{The extension over $e_S$.}
The homotopy extension property, for the simplicial pair $(e_S,\partial e_S)$,
extends the already chosen boundary
homotopy over $e_S$, inside $Q_S\setminus N$. Let $g_S:e_S\to Q_S\setminus N$ be its end map, which equals
the PL injection $\bigcup_v\iota_{S+v}$ on $\partial e_S$. Subdivide $e_S$ as a cone over a subdivision of
$\partial e_S$ on which this injection is simplexwise affine.
Move its relative interior into
$\operatorname{int}Q_S$ and approximate there in relative PL general
position, by applying Lemma~\ref{lem:relative-general-position} with $B=Q_S$, $M=n'-|S|$,
$V=Q_S\setminus N$, $X=e_S$, $X_0=\partial e_S$, and $g=g_S$. The dimension inequality
\[
 2\dim e_S+1=2(a-|S|)+1\le n'-|S|
\]
holds because $|S|\ge1$ and $n'\ge2a$. It
ensures that the resulting map $\iota_S$ is an embedding. The lemma keeps the changes
in the open complement of $N$ and relative
to the already fixed boundary, and it gives a homotopy from $g_S$ to $\iota_S$ relative to
$\partial e_S$. Following the extended homotopy by this one completes the inductive step.

\proofstep{The carriers.}
The image $T_S=\iota_S(e_S)$ is a compact polyhedron homeomorphic to the cone $e_S$, so it is
contractible and semialgebraic, with $T_S\subseteq Q_S\setminus N$.
The images are nested, since $\iota_S$ extends $\iota_{S'}$ for $S\subseteq S'$.
Theorem~\ref{thm:carrier}, with the integers $a$ and $d=n'-a-1\ge2$,
therefore applies and gives $\mathrm{LR}(C')\le n'-a$.
Proposition~\ref{prop:cover-rule} gives $\operatorname{lr}(C')\le\mathrm{LR}(C')$, which proves the bound.
\end{proof}

\subsection{Proof of Theorem~\ref{thm:stable}}
The independent coordinates enter through one observation: a cube face avoids $C\times\mathcal B_t$
exactly when its first factor avoids $C$. So the section of Lemma~\ref{lem:stable-section} lifts to the
product, and Lemma~\ref{lem:stable-embedding} applies as soon as $n+t$ is at least twice the number of
prescribed literals. For the face width we use the anchoring identity of Theorem~\ref{thm:anchor-width}
and apply the same argument to the anchored class, with one more prescribed literal.

\begin{proof}[Proof of Theorem~\ref{thm:stable}]
\proofstep{Sections for the product.}
Let $n'=n+t$, and write a literal set on $n'$ coordinates as $S=S_1\sqcup S_2$, the parts on the coordinates
of $C$ and of $\mathcal B_t$. A cube face is a product $B_1\times B_2$.
For $C\times\mathcal B_t$, a cube face avoids the class exactly when its
first factor avoids $C$. Thus
$\mathcal Q_{(S_1,S_2)}(C\times\mathcal B_t)
=\mathcal Q_{S_1}(C)\times Q_{S_2}$, with $Q_{S_2}\subseteq[-1,1]^t$.
Fix an integer $a$ with $0\le a\le a_0$. The map $S\mapsto S_1$ preserves inclusions, so it induces a simplicial map from the
order complex on $n'$ coordinates to the one on $n$ coordinates, and a chain with least member containing $S$
goes to a chain with least member containing $S_1$.
Compose the section \eqref{eq:stable-section} with this map, and multiply the result by the map taking each literal set
$S_2$ to the center of its cube face and interpolating affinely on chains. This
gives a section at level $a$ in $n+t$ coordinates:
\begin{equation}\label{eq:product-section}
 \Phi_a'(e_S)\subseteq\mathcal Q_{S_1}(C)\times Q_{S_2}=\mathcal Q_S(C\times\mathcal B_t),
\end{equation}
for every consistent $S$ of size at most $a$, including $S=\varnothing$. The case $S_1=\varnothing$ may occur
for nonempty $S$, which is why Lemma~\ref{lem:stable-connectivity} includes the empty set.
Since $\mathcal Q_S$ misses $N[C\times\mathcal B_t]$, by Lemma~\ref{lem:stable-clamp}(a) applied to the first
factor, the restriction of $\Phi_a'$ to the nonempty literal sets is a compatible map as in
Lemma~\ref{lem:stable-embedding}.

We also use $D\ge4$: since $p-1\ge1$, $(q-1)^{-1}=2\ln(2n)\ge2\ln2$, and $m^{2/p},n^{2-2/q}\ge1$, we have
$D_*\ge2\cdot2\ln2\cdot16/9>4$.

\proofstep{The bound for $\operatorname{lr}$.}
Let $t\ge\max\{0,n-2D-4\}$, and suppose $a_0\ge1$.
Lemma~\ref{lem:stable-embedding} applies with $a=a_0$ when $n+t\ge2a_0$, that is, $t\ge n-2D-4$,
and $n+t-a_0-1=t+D+1\ge2$. It gives the bound
$n+t-a_0=t+D+2$. If $a_0\le0$, use the elementary bound
$\operatorname{lr}(C\times\mathcal B_t)\le n+t$ instead: Lemma~\ref{lem:cube-cover} applied to the inclusion
of $N[C\times\mathcal B_t]$ in $\partial Q$, with Proposition~\ref{prop:cover-rule}, gives it, and
$n\le D+2$.

\proofstep{The bound for $u$: the anchored class.}
Let $E=(C\times\mathcal B_t)^+$, with anchor coordinate $0$ and anchor literals $\alpha_\pm=(0,\pm1)$.
For the anchored class use level $a_0+1$ if $a_0\ge0$, and write $\mathcal N=N[E]$. Every row of $E$ has
$+1$ at the anchor. So a point with anchor coordinate $-1$ is outside $\mathcal N$, and a point
$(y_0,y',y'')$ with $y'\in\mathcal Q_{S_1}(C)$ is outside $\mathcal N$ by Lemma~\ref{lem:stable-clamp}(a).

\proofstep{The bound for $u$: the compatible map.}
First define the map on all chains whose least member contains the positive anchor,
using the section just constructed: a chain $\alpha_++S^0\subsetneq\alpha_++S^1\subsetneq\cdots$ goes to
$\{1\}\times\Phi'_{a_0}(S^0\subsetneq S^1\subsetneq\cdots)$, inside
$\{1\}\times\mathcal Q_{S^0_1}(C)\times Q_{S^0_2}$. For a chain whose least member $S^0$ contains the negative anchor, the
set $Q_{S^0}\setminus\mathcal N$ is a whole cube face, hence contractible: interpolate affinely between the centers of the
faces $Q_S$ of the members $S$ of the chain, so the chain goes into the convex face $Q_{S^0}$.
For a chain whose least member $S=S_1\sqcup S_2$ contains neither anchor literal, use the carrier
\[
 \mathcal C(S)=(\{-1\}\times Q_{S_1}\times Q_{S_2})
 \cup([-1,1]\times\mathcal Q_{S_1}(C)\times Q_{S_2}).
\]
It is contained in $Q_S\setminus\mathcal N$. It retracts to its negative-anchor face, by moving the anchor
coordinate linearly to $-1$, and this face is convex, so $\mathcal C(S)$ is contractible. It
contains both previously
used carriers for every chain whose least member contains $S$: the two faces
$\{1\}\times\mathcal Q_{S'_1}(C)\times Q_{S'_2}$ and $\{-1\}\times Q_{S'_1}\times Q_{S'_2}$, for $S'\supseteq S$. Also
$\mathcal C(S')\subseteq\mathcal C(S)$ for anchor-free $S'\supseteq S$.
Extend over the chains with anchor-free least member in increasing
dimension. Such a chain with least member $S$ has its boundary in $\mathcal C(S)$: faces with the same
least member by induction, and the face deleting $S$ by the two inclusions just stated.
This specifies the map in an order that respects every
intersection, and each $e_S$ is mapped into $Q_S\setminus\mathcal N$.

\proofstep{The bound for $u$: conclusion.}
The embedding condition is now
$n+t+1\ge2(a_0+1)$, that is, $t\ge n-2D-3$. Also $(n+t+1)-(a_0+1)-1=t+D+1\ge2$. Lemma~\ref{lem:stable-embedding}
gives $\operatorname{lr}(E)\le n+t-a_0$, and the result is again $t+D+2$.

If $a_0<0$, the identity model gives the bound: $u(C\times\mathcal B_t)\le n+t+1\le t+D+2$, since $n\le D+1$.
If $a_0=0$, the
positive-anchor part needs only $\mathcal K\ne\varnothing$, supplied
by Lemma~\ref{lem:stable-connectivity}, and the same construction works.
Theorem~\ref{thm:anchor-width} identifies $\operatorname{lr}(E)$ with $u(C\times\mathcal B_t)$.

\proofstep{The third assertion.}
Taking instead $a=\lfloor(n+t)/2\rfloor\le a_0$ in \eqref{eq:product-section} proves the third
assertion. The dimensional side conditions of Lemma~\ref{lem:stable-embedding} hold, since $a\ge2$,
$n+t\ge2a$, and $n+t-a-1=\lceil(n+t)/2\rceil-1\ge2$ because $n+t\ge5$. The lemma gives
$\operatorname{lr}(C\times\mathcal B_t)\le n+t-a=\lceil(n+t)/2\rceil$.

\proofstep{The constant.}
Finally, write $L=\ln(2m)$ and $k=\lceil L\rceil$, so $p=2k$ and
$(p-1)m^{2/p}=(2k-1)e^{(L-\ln2)/k}$. We first show that this quantity is at most $2e\ln(2m)$.
For $m\ge2$, we have $L>1$, and the function $k\mapsto(2k-1)e^{(L-\ln2)/k}$ is increasing
for $k\ge L$: its derivative has the sign of $2k^2-(2k-1)(L-\ln2)\ge2k(k-L+\ln2)>0$.
With $k\le L+1$ it is therefore at most $(2L+1)e^{(L-\ln2)/(L+1)}$, and this is at most $2eL$.
The latter inequality is equivalent to
$\ln(1+1/(2L))\le(1+\ln2)/(L+1)$, which for $L\ge1$ follows from
$\ln(1+1/(2L))\le1/(2L)$ and $L+1\le2L\le2L(1+\ln2)$.
For $m=1$ the claim holds directly: $(p-1)m^{2/p}=1\le2e\ln2$.

Also $2-2/q<2(q-1)=1/\ln(2n)$ gives
$n^{2-2/q}<e$, and
$(q-1)^{-1}=2\ln(2n)$. Therefore
\[
 D_*\le2\cdot2\ln(2n)\cdot2e\ln(2m)\cdot e\cdot\frac{16}9
   =8e^2(16/9)\ln(2m)\ln(2n)
       <105.09\ln(2m)\ln(2n).
\]
This proves the bound on $D_*$ and completes the proof of Theorem~\ref{thm:stable}.
\end{proof}

\section{Orientation classes and the index gap}\label{app:orientations}
The inequality $\operatorname{Ind}\le2\operatorname{lr}-1$ of~\cite[Theorem 3.1]{BHHLT26} bounds the
equivariant index by the list number. This section shows that it has no additive converse, which
answers negatively the question of~\cite[Section 9.2]{Zhang26} whether
$\operatorname{lr}\le\operatorname{Ind}+O(1)$: Theorem~\ref{thm:index-gap} gives explicit classes $D_k$,
with $32^k$ rows on $6k+1$ coordinates, such that $\operatorname{lr}(D_k)=4k+1$ and
$\operatorname{Ind}S(D_k)=3k$. Combined with the index bound of~\cite[Theorem 1]{FHMNV26v1}, an additive
converse would give the rate of Theorem~\ref{thm:main} directly. The classes $D_k$ are built
from source-free orientations of graphs, on which the face width $u$ of Section~\ref{app:width} can
be computed exactly: it equals the cycle rank plus the minimum size of a feedback vertex set, and
$u-1$ is additive on products of such classes. On the index side, the cycles of $K_4$ pairwise
intersect, which yields an odd map to the cycle space and hence an index bound involving only the
cycle rank. The feedback-vertex term of the face width has no counterpart in this bound, and this
difference produces the gap.

We first fix notation for the faces of $\operatorname{conv}C$, which is also used in
Sections~\ref{app:thresholds} and~\ref{app:low-index}.
Throughout, $P=\operatorname{conv}C$ for a class $C$, in the $[-1,1]$ or
the $[0,1]$ coordinates; an endpoint value means $\pm1$, respectively
$0,1$. For a consistent literal set $T$ let $Q_T$ be the cube face it
fixes. A convex combination of rows with positive weights lies on
$\{x_j=s\}$, with $s$ an endpoint value, exactly when all those rows have
the value $s$ at $j$. Hence
\begin{equation}\label{eq:literal-face}
 P\cap Q_T=\operatorname{conv}\{c\in C:\ c\text{ contains }T\},
\end{equation}
a face of $P$, nonempty exactly when $T$ is contained in a row.
A triangulation of $P$ is \emph{compatible} if every face of $P$ is a
subcomplex; a pulling triangulation is compatible and uses only the
vertices of $P$, and subdivisions of compatible triangulations are
compatible. For a simplex $\sigma$ of a compatible triangulation let
$U(\sigma)$ be the set of literals $(j,s)$ with $\sigma\subseteq\{x_j=s\}$.
Then $\sigma\subseteq Q_T$ if and only if $T\subseteq U(\sigma)$, and
$U(\sigma)$ is contained in a row by \eqref{eq:literal-face}. If $x$ lies
in the relative interior of $\sigma$ and $x_j$ is an endpoint value $s$,
then the affine function $x_j$ has an extreme value on $\sigma$ at a
relative interior point, so it is constant on $\sigma$. Thus
\begin{equation}\label{eq:fixed-at-interior}
 \{(j,s):x_j=s\}=U(\sigma)\qquad(x\in\operatorname{relint}\sigma),
\end{equation}
and if $\tau$ is a face of $\sigma$ then $U(\tau)\supseteq U(\sigma)$.

\subsection{A common-filling obstruction}
This subsection gives the tool for lower bounds on face width. Consider several unions of coordinate
facets of the cube and a cycle in their intersection that does not bound there. If
this cycle and its successive fillings can be filled inside the polytope, in the unions of few of these
sets, then a low-dimensional face-separating model would produce a filling in the intersection, which
is impossible.

All chains in this subsection have coefficients in $\mathbb F_2$, so a
chain is a finite set of cells. Reduced homology of a nonempty complex is
computed from the augmented chain complex, and an augmented $0$-cycle is a
$0$-chain with an even number of vertices. A \emph{face-separating model}
of a polytope $P\subseteq[0,1]^M$ is a triple $(f,g,K)$ as in
Section~\ref{app:width}: $K$ is a finite polyhedron and $f:P\to K$,
$g:K\to[0,1]^M$ are continuous with $(gf(x))_j=s$ whenever $x_j=s\in\{0,1\}$.
Its dimension is $\dim K$.

The combinatorial core is a statement about chains indexed by subsets. It says that a system of
compatible fillings in the unions $W_J$ whose top level closes up produces a filling in
the common intersection of the $W_i$.

\begin{lemma}[Support-preserving descent]\label{lem:support-descent}
Let $\mathcal X$ be a finite simplicial or cubical complex and let $W_i$,
$i\in A$, be subcomplexes indexed by a finite nonempty set. Write
$W_J=\bigcup_{i\in J}W_i$. Let $t,q\ge0$ be integers and $b\in C_t(\mathcal X)$.
Suppose that, with $z_\varnothing=b$, there are chains
$z_J\in C_{t+|J|}(W_J)$ for $\varnothing\ne J\subseteq A$, $|J|\le q$, with
\[
 \partial z_J=\sum_{i\in J}z_{J\setminus\{i\}}
 \quad(\varnothing\ne J\subseteq A,\ |J|\le q),\qquad
 \sum_{i\in I}z_{I\setminus\{i\}}=0
 \quad(I\subseteq A,\ |I|=q+1).
\]
Then there is $w\in C_{t+1}\bigl(\bigcap_{i\in A}W_i\bigr)$ with $\partial w=b$.
\end{lemma}
\begin{proof}
Put $p_0=\min\{q,|A|\}$. If $p_0=0$ then $q=0$, and the second hypothesis
for a singleton $I$ says $b=0$; take $w=0$. Let $p_0\ge1$. For
$1\le p\le p_0$ let $\mathcal S_p$ be the statement that there are chains
$z_J\in C_{t+|J|}(W_J)$, $\varnothing\ne J\subseteq A$, $|J|\le p$, such that,
with $z_\varnothing=b$,
\[
 \partial z_J=\sum_{i\in J}z_{J\setminus\{i\}}\quad(1\le|J|\le p),
 \qquad
 \sum_{i\in I}z_{I\setminus\{i\}}=0\quad(I\subseteq A,\ |I|=p+1).
\]
The hypotheses give $\mathcal S_{p_0}$: when $q\ge|A|$ there is no set $I$
of size $p_0+1$, and when $q<|A|$ we have $p_0=q$.

\proofstep{From $\mathcal S_p$ to $\mathcal S_{p-1}$ for $p\ge2$.}
For a cell $\sigma$ of dimension $t+p$ let $z_I(\sigma)\in\mathbb F_2$ be its
coefficient in $z_I$, $|I|=p$. Put $M_\sigma=\{i\in A:\sigma\in W_i\}$ and
$U_\sigma=A\setminus M_\sigma$. A cell of a union of subcomplexes is a cell
of one of them, so
\begin{equation}\label{eq:vanish-on-U}
 z_I(\sigma)=0\quad\text{whenever }I\subseteq U_\sigma.
\end{equation}
Choose $j_\sigma\in U_\sigma$ if $U_\sigma\ne\varnothing$, and any
$j_\sigma\in A$ otherwise. For $|J|=p-1$ define a chain $b_J$ of degree
$t+p$ by
\[
 b_J(\sigma)=
 \begin{cases}
 z_{J\cup\{j_\sigma\}}(\sigma),& j_\sigma\notin J,\\
 0,& j_\sigma\in J.
 \end{cases}
\]
We claim that $z_I=\sum_{i\in I}b_{I\setminus\{i\}}$ for $|I|=p$. Fix
$\sigma$ and write $j=j_\sigma$. If $j\in I$, the set $I\setminus\{i\}$
contains $j$ unless $i=j$, so the coefficient of the sum at $\sigma$ is
$b_{I\setminus\{j\}}(\sigma)=z_I(\sigma)$. If $j\notin I$, the coefficient
is $\sum_{i\in I}z_{(I\setminus\{i\})\cup\{j\}}(\sigma)$; the second relation
of $\mathcal S_p$ for the $(p+1)$-set $I\cup\{j\}$ says that this sum plus
$z_I(\sigma)$ is zero.

Next, $b_J\in C_{t+p}(W_J)$. Let $b_J(\sigma)\ne0$. Then $j_\sigma\notin J$
and $z_{J\cup\{j_\sigma\}}(\sigma)\ne0$, so $J\cup\{j_\sigma\}$ meets
$M_\sigma$ by \eqref{eq:vanish-on-U}. If $U_\sigma\ne\varnothing$ then
$j_\sigma\notin M_\sigma$, so $J$ meets $M_\sigma$. If $U_\sigma=\varnothing$
then $M_\sigma=A$, and $J$ meets it because $|J|=p-1\ge1$. In both cases
$\sigma$ is a cell of $W_J$.

Define $z'_J=z_J+\partial b_J$ for $|J|=p-1$ and $z'_J=z_J$ for $|J|<p-1$.
Since $W_J$ is a subcomplex, $z'_J\in C_{t+p-1}(W_J)$. For $|J|=p-1$ we have
$\partial z'_J=\partial z_J=\sum_{i\in J}z_{J\setminus\{i\}}
=\sum_{i\in J}z'_{J\setminus\{i\}}$, because the families of size below
$p-1$ are unchanged. For $|I|=p$,
\[
 \sum_{i\in I}z'_{I\setminus\{i\}}
 =\sum_{i\in I}z_{I\setminus\{i\}}
  +\partial\sum_{i\in I}b_{I\setminus\{i\}}
 =\partial z_I+\partial z_I=0.
\]
This is $\mathcal S_{p-1}$.

\proofstep{The end of the descent.}
By $\mathcal S_1$ we have chains $z_i\in C_{t+1}(W_i)$ with $\partial z_i=b$
and $z_i+z_{i'}=0$ for $i\ne i'$. Thus all $z_i$ are one chain $w$. Each cell
of $w$ is a cell of every $W_i$, so $w\in C_{t+1}(\bigcap_iW_i)$.
\end{proof}

We now state the obstruction. Its proof pushes the fillings forward along a simplicial model and applies
the descent lemma in the target.

\begin{lemma}[Common filling]\label{lem:common-filling}
Let $P\subseteq[0,1]^M$ be a nonempty convex polytope, and let
$R_i$, $i\in A$, be finitely many unions of closed coordinate facets, with $A$ nonempty.
Write $Y_i=P\cap R_i$, $Z=\bigcap_iR_i$, and $B=P\cap Z$.
Suppose a class $\gamma\in\widetilde H_d(B)$, $d\ge0$, has nonzero
image in $\widetilde H_d(Z)$. If, for some integer $q\ge0$,
\[
 \widetilde H_{d+|I|-1}\Big(\bigcup_{i\in I}Y_i\Big)=0
 \quad(\varnothing\ne I\subseteq A,\ |I|\le q),
\]
then every face-separating model of $P$ has dimension at least $d+q+1$.
\end{lemma}
\begin{proof}
The set $B$ is nonempty because $\gamma\ne0$. Also $B\subseteq Y_i$ for every $i$.
Suppose $(f_0,g_0,K)$ is a model with $\dim K\le d+q$.

\proofstep{Simplicial reduction.}
Write $F_{j,s}=P\cap\{x_j=s\}$ for $s\in\{0,1\}$. Take a compatible
triangulation $\mathcal T_0$ of $P$. Each $Y_i$ is a union of sets $F_{j,s}$,
and these are faces of $P$ or empty; so every $Y_i$ and $B=\bigcap_iY_i$ is a
subcomplex. The two sets $f_0(F_{j,0})$ and $f_0(F_{j,1})$ are compact and
disjoint, because $g_0$ takes the $j$th coordinate value $0$ on the first and
$1$ on the second. Fix a metric on $K$ and let $\eta>0$ be smaller than the
distance between these two sets, for all $j$ for which both are nonempty.
Triangulate $K$ with simplices of diameter less than $\eta/3$. The simplicial
approximation theorem~\cite[Theorem 2C.1]{Hatcher02} gives an iterated
barycentric subdivision $\mathcal T$ of $\mathcal T_0$ and a simplicial map
$f:\mathcal T\to K$ such that $f(x)$ and $f_0(x)$ lie in a common closed
simplex, for every $x$. Then the distance between $f(F_{j,0})$ and $f(F_{j,1})$
exceeds $\eta/3$, so these are disjoint subcomplexes. By the Tietze theorem
there are continuous $g_j:K\to[0,1]$ with $g_j=s$ on $f(F_{j,s})$. Put
$g=(g_j)_j$. Then $(f,g,K)$ is again a model. Write $W_i=f(Y_i)$, a subcomplex
of $K$. A point of $W_i$ has the form $f(x)$ with $x\in F_{j,s}$ for some facet
$\{x_j=s\}\subseteq R_i$, and then $g(f(x))_j=s$. Hence
\begin{equation}\label{eq:g-into-R}
 g(W_i)\subseteq R_i\quad(i\in A),\qquad
 g\Bigl(\bigcap_{i}W_i\Bigr)\subseteq Z.
\end{equation}

\proofstep{Compatible fillings.}
Represent $\gamma$ by an augmented simplicial cycle $a\in C_d(B)$ of
$\mathcal T$, and put $E_\varnothing=a$. By recursion on $|I|$, for
$1\le|I|\le q$, we choose
\[
 E_I\in C_{d+|I|}\Big(\bigcup_{i\in I}Y_i\Big),\qquad
 \partial E_I=c_I:=\sum_{i\in I}E_{I\setminus\{i\}}.
\]
The chain $c_I$ has degree $d+|I|-1$ and lies in $\bigcup_{i\in I}Y_i$: for
$|I|=1$ it equals $a$, which lies in $B\subseteq Y_i$, and for $|I|\ge2$ each
$E_{I\setminus\{i\}}$ lies in the smaller union. It is an augmented cycle: for
$|I|=1$ because $a$ is one, and for $|I|\ge2$ because
\[
 \partial c_I=\sum_{i\in I}\ \sum_{h\in I\setminus\{i\}}E_{I\setminus\{i,h\}}=0,
\]
each double deletion occurring twice. The homology hypothesis for $I$ provides
the filling $E_I$. For $|I|=q+1$ the same computation shows that $c_I$ is an
augmented cycle in $P$ of degree $d+q$. Since $P$ is contractible, there is
$E_I\in C_{d+q+1}(P)$ with $\partial E_I=c_I$.

\proofstep{Descent in the target.}
Put $z_J=f_\#E_J$ for $|J|\le q$, so that $z_J\in C_{d+|J|}(\bigcup_{i\in J}W_i)$
and $\partial z_J=\sum_{i\in J}z_{J\setminus\{i\}}$, with
$z_\varnothing=f_\#a$. For $|I|=q+1$,
\[
 \sum_{i\in I}z_{I\setminus\{i\}}=f_\#c_I=\partial f_\#E_I=0,
\]
because $K$ has no simplices of dimension $d+q+1$. Lemma~\ref{lem:support-descent},
with $\mathcal X=K$, $t=d$, and $b=f_\#a$, gives a chain
$w\in C_{d+1}(\bigcap_iW_i)$ with $\partial w=f_\#a$.

\proofstep{Contradiction.}
The map $f$ sends $B$ into $W=\bigcap_iW_i$, and $W\ne\varnothing$. The chain
$f_\#a$ represents $(f|_B)_*\gamma\in\widetilde H_d(W)$, and it is the boundary
of $w$, so this class is zero. By \eqref{eq:g-into-R}, $g|_W$ maps into $Z$.
Thus $gf|_B:B\to Z$ sends $\gamma$ to zero in $\widetilde H_d(Z)$.

On the other hand, let $x\in B$ and let $D_x$ be the smallest coordinate face
of $[0,1]^M$ containing $x$. For each $i$, the point $x$ lies on a facet
$\{x_j=s\}\subseteq R_i$; this facet contains $D_x$, so $D_x\subseteq R_i$.
Hence $D_x\subseteq Z$. Since $(f,g,K)$ is a model, $gf(x)\in D_x$. As $D_x$ is
convex, $(x,t)\mapsto(1-t)x+t\,gf(x)$ is a homotopy in $Z$ from the inclusion
$B\hookrightarrow Z$ to $gf|_B$. Therefore the image of $\gamma$ under the
inclusion is zero, contrary to the hypothesis.
\end{proof}

\subsection{Exact values from feedback vertex sets}
We now compute the face width of source-free orientation classes, from which the gap is built. The lower
bound applies Lemma~\ref{lem:common-filling} to one union of facets for each vertex of the graph. The
upper bound constructs a completion that protects only the vertices outside a minimum feedback
vertex set and retracts coordinate sections along a forest. The two bounds agree.

Let $G=(V,E)$ be a finite connected simple graph containing a cycle, and fix
reference directions for the edges. The source-free orientation
class $C_G\subseteq\{\pm1\}^E$ of $G$ consists of the orientations in which
every vertex has at least one incoming edge.
A vertex set whose removal leaves a forest is called a \emph{feedback vertex set}.
Write $r=|E|-|V|+1$ for the cycle rank, and let $\tau(G)$ be the minimum size of a
feedback vertex set.

A literal of this class is an oriented edge $y\to x$, that is, the edge $yx$
with the sign directing it from $y$ to $x$; its \emph{head} is $x$. A consistent
literal set is a \emph{partial orientation}. It \emph{covers} a vertex if that
vertex is the head of one of its literals. We use $[0,1]$ edge coordinates. A
literal $\ell$ determines a facet $F(\ell)$ of $[0,1]^E$, and a partial orientation
$T$ determines the face $Q_T=\bigcap_{\ell\in T}F(\ell)$, of dimension $|E|-|T|$.
Every face of the cube has this form, and $Q_W\subseteq Q_T$ exactly when
$T\subseteq W$. Put
\[
 P=\operatorname{conv}C_G,\qquad
 R_v=\bigcup_{\text{head of }\ell\text{ is }v}F(\ell),\qquad
 Y_v=P\cap R_v,\qquad
 X=\bigcap_{v\in V}R_v.
\]
For nonempty $U\subseteq V$ let $O_U$ be the set of literals $y\to x$ with
$y\in U$ and $x\notin U$. The complex $K(C_G)$ of Section~\ref{app:literal}
consists of the partial orientations contained in a row of $C_G$.

The first lemma describes this complex. A partial orientation fails to extend to a row only because it
forces a source in a set inducing a tree.

\begin{lemma}\label{lem:source-free-obstructions}
A partial orientation is contained in a row of $C_G$ if and only if it contains
no set $O_U$ with $\varnothing\ne U\subseteq V$ and $G[U]$ a tree.
\end{lemma}
\begin{proof}
Suppose $T\supseteq O_U$ with $G[U]$ a tree, and let $c$ be an orientation
extending $T$. Every edge between $U$ and its complement is directed out of $U$,
so the incoming edges at vertices of $U$ are edges of $G[U]$. There are
$|U|-1$ of these, each with one head, so some vertex of $U$ has no incoming edge,
and $c\notin C_G$.

Conversely, let $T$ contain no such set. Let $\Phi$ be the graph on $V$ formed
by the edges not fixed by $T$, and let $D$ be the vertex set of a connected
component of $\Phi$. If $\Phi[D]$ contains a cycle, orient that cycle cyclically.
Then repeatedly take an unoriented edge of $\Phi[D]$ from the vertices reached
so far to a new vertex and direct it toward the new vertex, and orient all other
edges of $\Phi[D]$ arbitrarily. Every vertex of $D$ then has an incoming edge. If $\Phi[D]$ is
a tree and $T$ covers a vertex $w\in D$, direct the tree away from $w$. Every
vertex of $D$ other than $w$ then has an incoming tree edge, and $w$ has an incoming
edge in $T$. In the last case $\Phi[D]$ is a tree and $T$ covers
no vertex of $D$. Then no edge of $G$ inside $D$ is fixed by $T$, since its head
would be covered, so $G[D]=\Phi[D]$ is a tree. Every edge between $D$ and its
complement is fixed, since $D$ is a component of $\Phi$, and it is directed out
of $D$, since $T$ covers no vertex of $D$. Thus $T\supseteq O_D$, which is
excluded. Treating all components in this way extends $T$ to a row of $C_G$.
\end{proof}

The next lemma supplies the vanishing required by Lemma~\ref{lem:common-filling}: the facets with head
in a set that does not meet all cycles have a contractible union inside $P$.

\begin{lemma}\label{lem:source-free-unions}
If $\varnothing\ne I\subseteq V$ is not a feedback vertex set, then
$P\cap\bigcup_{v\in I}R_v$ is contractible.
\end{lemma}
\begin{proof}
Let $\Lambda_I$ be the set of literals with head in $I$ that are contained in a
row of $C_G$. By \eqref{eq:literal-face}, the set in question is the union of
the nonempty faces $P\cap F(\ell)$, $\ell\in\Lambda_I$. For $S\subseteq\Lambda_I$
the intersection of the corresponding faces is $P\cap Q_S$ if $S$ is consistent
and empty otherwise; by \eqref{eq:literal-face} it is nonempty exactly when
$S\in K(C_G)$, and then it is a convex polytope. These faces are subcomplexes of a
compatible triangulation. Passing to open regular neighborhoods in its second
barycentric subdivision, which have the same intersection pattern and deformation
retract onto these subcomplexes and their intersections, the nerve
lemma~\cite[Corollary 4G.3]{Hatcher02} shows that the union is homotopy equivalent
to the induced subcomplex $K(C_G)[\Lambda_I]$.

We show that this complex is a cone. Since $I$ is not a feedback vertex set, some
connected component $D$ of $G-I$ contains a cycle. As $I\ne\varnothing$ and $G$ is
connected, there is an edge $yx$ with $y\in D$ and $x\notin D$, and $x\in I$
because $D$ is a component of $G-I$. Put $\ell=(y\to x)$. Let
$S\subseteq\Lambda_I$ be a face of $K(C_G)$, possibly empty. The set $S\cup\{\ell\}$
is consistent: the reverse of $\ell$ has head $y\notin I$, so it is not in $S$.
Suppose $S\cup\{\ell\}$ is not contained in a row. By
Lemma~\ref{lem:source-free-obstructions} it contains $O_U$ with $G[U]$ a tree. As
$S$ is contained in a row, it does not contain $O_U$, so $\ell\in O_U$, whence
$y\in U$. All literals of $O_U$ have heads in $I$. Let $y'\in U\cap D$ and let $z$ be a
neighbor of $y'$ in $D$. If $z\notin U$, then $y'\to z$ belongs to $O_U$, with head
$z\notin I$, a contradiction. So $z\in U$. Starting from $y$ and using connectivity
of $D$ gives $D\subseteq U$. Then $G[U]$ contains the cycle of $D$, contradicting
that it is a tree. Therefore $S\cup\{\ell\}\in K(C_G)$ for every such $S$. In
particular $\ell\in\Lambda_I$, and the induced complex is a cone with apex $\ell$.
\end{proof}

\begin{theorem}[Exact face width of source-free classes]\label{thm:graph-budget}
For every finite connected simple graph $G$ containing a cycle,
\[
 u(C_G)=\operatorname{lr}(C_G^+)=r+\tau(G).
\]
In particular $u(C_{K_s})=\binom{s}{2}-1$ for $s\ge3$,
and $u(C_{W_k})=k+2$ for the wheel with a rim of length $k\ge3$.
\end{theorem}
\begin{proof}
\proofstep{The complex $X$.}
A point $z\in[0,1]^E$ lies on $F(\ell)$ exactly when $\ell$ is one of its fixed
literals. So $z\in R_v$ exactly when the fixed literals of $z$ cover $v$, and
$X$ is the union of the faces $Q_T$ with $T$ covering $V$. All Boolean points of such
a face are rows of $C_G$, so $Q_T\subseteq P$, and therefore $X\subseteq P$.
A partial orientation covering $V$ has at least $|V|$ literals, because a literal
has one head. So $\dim X\le|E|-|V|=r-1$. Equality holds: choose a cycle, enlarge it
to a connected spanning subgraph by repeatedly adding an edge to a new vertex,
orient the cycle cyclically and direct each added edge toward its new vertex.
This partial orientation has $|V|$ literals and covers each vertex exactly once.

\proofstep{A nonzero class in $\widetilde H_{r-1}(X)$.}
Let $r\ge2$ and let $a_0$ be the sum of all $(r-1)$-dimensional cubes of $X$, a
nonzero cubical chain over $\mathbb F_2$. An $(r-2)$-cube of $X$ is $Q_{T'}$ with
$T'$ covering $V$ and $|T'|=|V|+1$. By counting heads, exactly one vertex $w$ is the
head of two literals of $T'$, and every other vertex is the head of one. The face
$Q_{T'}$ lies in a top cube $Q_{T'\setminus\{\ell\}}$ of $X$ exactly when
$T'\setminus\{\ell\}$ still covers $V$, that is, when $\ell$ is one of the two
literals with head $w$. So the coefficient of $Q_{T'}$ in $\partial a_0$ is zero,
and $a_0$ is a cycle. Since $X$ has no $r$-dimensional cubes, $a_0$ is not a boundary,
and it represents a nonzero class in $H_{r-1}(X;\mathbb F_2)$.

Let $r=1$, so $G$ is unicyclic and a top face of $X$ is a single orientation with
exactly one incoming edge at each vertex. The edge at a leaf must point into the
leaf; deleting leaves repeatedly reduces to the cycle, on which indegree one at
each vertex forces one of the two cyclic orientations. Conversely, each cyclic orientation with the
deleted edges directed away from the cycle has indegree one everywhere. So $X$ consists of two
points. Let $a_0$ be their sum, a nonzero augmented $0$-cycle representing a nonzero
class in $\widetilde H_0(X;\mathbb F_2)$.

\proofstep{Lower bound.}
Apply Lemma~\ref{lem:common-filling} with $M=E$, $A=V$, the sets $R_v$, the class of
$a_0$, and $d=r-1$, $q=\tau(G)-1$. We have $\tau(G)\ge1$ because $G$ has a cycle. Each
$R_v$ is a nonempty union of facets because $G$ is connected. We have $Z=X$ and
$B=P\cap X=X$, so $\gamma$ has nonzero image. Every $I\ne\varnothing$ with
$|I|\le\tau(G)-1$ is not a feedback vertex set, so
Lemma~\ref{lem:source-free-unions} gives the required vanishing in all degrees.
Hence every model has dimension at least $(r-1)+(\tau(G)-1)+1$, that is,
$u(C_G)\ge r+\tau(G)$.

\proofstep{An ample completion.}
Choose a minimum feedback vertex set $F$, put $S=V\setminus F$, and let $H$ be the set
of orientations in which every vertex of $S$ has an incoming edge. Then
$C_G\subseteq H$. In the notation of Proposition~\ref{prop:carrier-extension},
a face $Q_T$ lies in $\operatorname{Cub}(H)$ if and only if $T$ covers $S$. Indeed, if it does, all Boolean points of
$Q_T$ lie in $H$, and if $T$ does
not cover $s\in S$, directing all edges at $s$ not fixed by $T$ away from $s$ gives a
Boolean point of $Q_T$ outside $H$. A partial orientation covering $S$ has at least $|S|$ literals, so
\[
 \dim\operatorname{Cub}(H)\le|E|-|S|=|E|-|V|+\tau(G)=r+\tau(G)-1.
\]

\proofstep{Contractible sections.}
Let $W$ be a partial orientation with nonempty section
$\Sigma_W=\operatorname{Cub}(H)\cap Q_W$. We prove by induction on the number of
vertices of $S$ not covered by $W$ that $\Sigma_W$ is contractible. The set
$\Sigma_W$ is the union of the faces $Q_T$ with $T\supseteq W$ and $T$ covering $S$:
if $z\in\operatorname{Cub}(H)\cap Q_W$ and $z\in Q_{T'}$ with $T'$ covering $S$, then
$T'\cup W$ is consistent because $z$ lies on all its facets, it covers $S$, and $z$
lies in its face. If $W$ covers $S$, then $\Sigma_W=Q_W$ is a cube.

Otherwise let $S_W\ne\varnothing$ be the set of vertices of $S$ not covered by $W$.
The graph on $S_W$ formed by the edges not fixed by $W$ with both endpoints in $S_W$ is a subgraph of
the forest $G[S]$. Let $D$ be the vertex set of one of its components, a tree with
$|D|-1$ edges. Since $\Sigma_W\ne\varnothing$, some $T\supseteq W$ covers $S$, so the
vertices of $D$ are heads of $|D|$ distinct literals of $T\setminus W$. These lie on
$|D|$ distinct edges not fixed by $W$, so at least one of these edges, say $e$, is not an edge of the
tree. Write $e=\{s,w'\}$ with $s\in D$ the head. If $w'\in S_W$, then $e$ would join two
vertices of $S_W$, so it would be an edge of the component on $D$; hence $w'\notin S_W$.
Let $\ell=(w'\to s)$ and let $\bar\ell$ be its reverse.

Define $\rho_t$ by moving the coordinate of $e$ linearly to the endpoint value of
$\ell$ at time $t=1$, keeping the other coordinates. We claim that $\rho_t$ maps
$\Sigma_W$ into itself, that $\rho_1(\Sigma_W)\subseteq\Sigma_W\cap F(\ell)$, and
$\rho_t$ is the identity on $\Sigma_W\cap F(\ell)$. The last assertion is immediate.
Let $z\in Q_T$ with $T\supseteq W$ covering $S$. If $\bar\ell\in T$, then $T_0=
T\setminus\{\bar\ell\}$ still covers $S$: the head of $\bar\ell$ is $w'$; if $w'\in S$,
then $w'$ is covered by $W$ because $w'\notin S_W$, and $\bar\ell\notin W$ since $e$ is
not fixed by $W$. So we may replace $T$ by $T_0$ and assume that $\bar\ell\notin T$. If
$\ell\in T$ then $\rho_t(z)=z$. If $e$ is not fixed by $T$, then $\rho_t(z)\in Q_T$ and
$\rho_1(z)\in Q_{T\cup\{\ell\}}$, and $T\cup\{\ell\}$ contains $W$ and covers $S$. This
proves the claim, so $\rho$ is a strong deformation retraction of $\Sigma_W$ onto
$\Sigma_W\cap F(\ell)=\Sigma_{W\cup\{\ell\}}$. This smaller section is nonempty and its
partial orientation covers $s$ in addition, so its set of uncovered vertices is
$S_W\setminus\{s\}$. By induction it is contractible, and so is $\Sigma_W$.

\proofstep{Upper bound.}
Thus $H$ is an ample completion of $C_G$, and Proposition~\ref{prop:carrier-extension} gives
$u(C_G)\le\dim\operatorname{Cub}(H)+1\le r+\tau(G)$.
The anchoring identity in Theorem~\ref{thm:anchor-width} gives
$\operatorname{lr}(C_G^+)=u(C_G)$.

\proofstep{Complete graphs and wheels.}
For $K_s$ we have $r=\binom s2-s+1$. Removing fewer than $s-2$ vertices leaves a clique
on at least three vertices, and removing $s-2$ vertices leaves one edge, so $\tau=s-2$
and $r+\tau=\binom s2-1$. For $W_k$ we have $|V|=k+1$, $|E|=2k$, and $r=k$. Removing the
hub leaves the rim cycle, and removing a rim vertex leaves a triangle formed by the hub
and two consecutive rim vertices among those left, since $k\ge3$. Removing the hub and one rim
vertex leaves a path. So $\tau=2$ and $r+\tau=k+2$.
\end{proof}

\subsection{Products of orientation classes}
Theorem~\ref{thm:anchor-width} bounds the face width of a product from above, and to compute the face
width of a product of source-free classes we need a matching lower bound. A model of a product need
not be a product of models, so the lower bound of Theorem~\ref{thm:graph-budget} does not pass to
products directly. We therefore introduce an algebraic quantity $a_{\mathbb F}(C)$, the least degree in
which a natural comparison map between two diagrams of chain complexes can be truncated. It is at most
$u(C)-1$ for every class, it is additive on all products by a K\"unneth argument, and the obstruction of
Lemma~\ref{lem:common-filling} shows that it equals $r+\tau(G)-1$ on source-free classes. Additivity is
proved for this quantity and not for the face-separating dimension of arbitrary classes.

\paragraph{Diagrams and resolutions.}
Let $\widehat K(C)$ be the poset of realizable literal sets including
the empty set. Work over a field $\mathbb F$. A \emph{diagram} assigns to each
$T\in\widehat K(C)$ a vector space $D(T)$ and to each inclusion $T\subseteq T'$ a linear
map $D(T')\to D(T)$, functorially. In other words, it is a functor on the opposite-inclusion
incidence category. Chain complexes of diagrams, chain maps, and chain homotopies are
natural in $T$.

On this category, the representable functor $P_U$ has value $\mathbb F$ at $T\subseteq U$
and zero elsewhere, with identity maps between the nonzero values. A natural map $P_U\to D$
is the same as an element of $D(U)$, the image of the generator $1\in P_U(U)$. Evaluation at $U$
is exact, so a finite direct sum of representables is projective. A \emph{finite projective
resolution} of the constant functor $\underline{\mathbb F}$ is a bounded complex
$\mathcal F_k$, $k\ge0$, of finite direct sums of representables with an augmentation
$\varepsilon:\mathcal F_0\to\underline{\mathbb F}$, such that the augmented complex is exact
at every object.

The diagram $\mathcal S:T\mapsto C_*(P\cap Q_T;\mathbb F)$, from a compatible triangulation,
is such a resolution, with transition maps the inclusions of chain groups. A simplex
$\sigma$ belongs to $P\cap Q_T$ exactly when $T\subseteq U(\sigma)$, so
$\mathcal S_k=\bigoplus_{\dim\sigma=k}P_{U(\sigma)}$, with $U(\sigma)\in\widehat K(C)$ since
$U(\sigma)$ is contained in a row. For $T\in\widehat K(C)$ the section $P\cap Q_T$ is a
nonempty convex polytope by \eqref{eq:literal-face}, so the augmented complex is exact
there. We call $\mathcal S$ the simplicial resolution.

\paragraph{The cubical diagram and comparison maps.}
Let $\mathcal F$ be any finite projective resolution of that constant functor. The diagram
of cubical chains $\mathcal C(T)=C_*(Q_T;\mathbb F)$ has the coordinate faces $Q_W$,
$W\supseteq T$ consistent, as a basis, with $Q_W$ in degree $n-|W|$, and it has the boundary
of the tensor product of $n$ copies of the chain complex of an interval. It is augmented
and objectwise acyclic, because $Q_T$ is a cube. There is an augmentation-preserving chain
map $\varphi:\mathcal F\to\mathcal C$, unique up to chain homotopy, by the following standard comparison. Let
$\mathcal D$ be an augmented complex of diagrams that is exact at every object. To define an
augmentation-preserving chain map $\mathcal F\to\mathcal D$, take a generator $e$ of a summand
$P_U$ of $\mathcal F_k$ and assume the map is defined in lower degrees. For $k=0$ choose an element
of $\mathcal D_0(U)$ of augmentation $\varepsilon(e)$. For $k\ge1$ choose an element of
$\mathcal D_k(U)$ whose boundary is the image of $\partial e$; that image is an augmented cycle in the exact
complex $\mathcal D(U)$. If $\varphi,\varphi'$ are two such maps, define a
homotopy $h$ by induction on $k$: $h(e)\in\mathcal D_{k+1}(U)$ is chosen with
\[
 \partial h(e)=\varphi(e)-\varphi'(e)-h(\partial e),
\]
whose right side is a cycle by the inductive identity, and has augmentation zero for $k=0$.
By the correspondence for representables these choices define natural maps.

\paragraph{The truncation degree.}
Define the \emph{truncation degree} $a_{\mathbb F}(C)$ as the least $d$ for which $\varphi$ can be chosen
with image in the truncation $\mathcal C_{\le d}$, the subcomplex with
$\mathcal C_{\le d}(T)_k=\mathcal C(T)_k$ for $k\le d$ and zero for $k>d$. This does not
depend on $\mathcal F$: a comparison map from another resolution to $\mathcal F$ composed with
a truncated map is a truncated map from that resolution. Also $a_{\mathbb F}(C)\le n$, since
$\mathcal C_{\le n}=\mathcal C$. A model of $C$ produces a truncated map, as the next proof shows, so the
truncation degree is an algebraic analogue of the dimension of a model.

\begin{lemma}[Properties of the truncation degree]\label{lem:algebraic-product}
For classes $C$ and $C'$ on disjoint domains and every field $\mathbb F$,
\[
 a_{\mathbb F}(C)\le u(C)-1,
 \qquad a_{\mathbb F}(C\times C')=a_{\mathbb F}(C)+a_{\mathbb F}(C').
\]
For source-free orientation classes,
$a_{\mathbb F_2}(C_G)=r+\tau(G)-1$.
\end{lemma}
\begin{proof}
\proofstep{Comparison with a model.}
Let $d=u(C)-1$. By Lemma~\ref{lem:cover-model} there is a finite open cover $\{W_\alpha\}$ of $P$ of
multiplicity at most $d+1$, none of whose members meets two opposite coordinate faces. Its
nerve $N$ has dimension at most $d$. For each $\alpha$ choose a Boolean vector $h_\alpha$
with $(h_\alpha)_j=s$ whenever $W_\alpha$ meets $\{x_j=s\}$. This is possible because no member meets both
opposite faces. For $T\in\widehat K(C)$ let
$N_T$ be the full subcomplex of $N$ on the vertices $\alpha$ with $h_\alpha\in Q_T$.

Choose a compatible triangulation of $P$ whose closed vertex stars are smaller than a
Lebesgue number of the cover, and for each vertex $v$ a member $W_{\alpha(v)}$ containing
the closed star of $v$. A simplex $\sigma$ lies in the closed star of each of its vertices,
so $\sigma\subseteq\bigcap_vW_{\alpha(v)}$; hence $v\mapsto\alpha(v)$ is a simplicial map
$\nu$ to $N$. If $\sigma\subseteq Q_T$, each vertex $v$ of $\sigma$ lies in
$W_{\alpha(v)}$ and on every facet of $T$, so $h_{\alpha(v)}\in Q_T$, and $\nu(\sigma)$ is a
simplex of $N_T$. Thus $\nu_\#$ is a natural augmentation-preserving chain map from
$\mathcal S(T)$ to $C_*(N_T;\mathbb F)$.

For a simplex $\omega$ of $N$ let $Q(\omega)$ be the smallest coordinate face containing all
labels $h_\alpha$, $\alpha\in\omega$; faces of $\omega$ have smaller coordinate faces. Define
$\psi:C_*(N;\mathbb F)\to\mathcal C(\varnothing)$ by induction on $\dim\omega$, with
$\psi(\omega)\in C_{\dim\omega}(Q(\omega))$: a vertex $\alpha$ goes to the cube vertex $h_\alpha$, and
for $\dim\omega\ge1$ the chain $\psi(\partial\omega)$ is an augmented cycle in the acyclic complex
$C_*(Q(\omega))$, so it bounds a chain $\psi(\omega)$. Since $\dim N\le d$, the image lies in
degrees at most $d$. For $\omega\in N_T$ we have $Q(\omega)\subseteq Q_T$, so
$\psi$ maps $C_*(N_T;\mathbb F)$ into $\mathcal C_{\le d}(T)$. Then $\psi\nu_\#$ is an
augmentation-preserving natural chain map $\mathcal S\to\mathcal C_{\le d}$. Hence
$a_{\mathbb F}(C)\le d$.

\proofstep{A truncation criterion.}
For $e\ge0$ write $\mathcal B_e=\mathcal C/\mathcal C_{\le e-1}$, with $\mathcal C_{\le-1}=0$, and
let $\pi_e$ be the quotient map. Its terms are $\mathcal C_k$ for $k\ge e$ and zero for $k<e$.
Fix an augmentation-preserving $\varphi:\mathcal F\to\mathcal C$. We claim that $\pi_e\varphi$ is
null-homotopic if and only if $e\ge1$ and there is an augmentation-preserving map
$\mathcal F\to\mathcal C_{\le e-1}$.

If such a map $\varphi'$ exists, then $\varphi'\simeq\varphi$ by uniqueness, and $\pi_e\varphi'=0$,
so $\pi_e\varphi\simeq0$. Conversely, let $\pi_e\varphi=\partial\bar h+\bar h\partial$ with
$\bar h_k:\mathcal F_k\to(\mathcal B_e)_{k+1}$. If $e=0$, then $\pi_0=\mathrm{id}$ and
$\varepsilon\varphi_0=\varepsilon\partial\bar h_0=0$, which contradicts
$\varepsilon\varphi_0=\varepsilon_{\mathcal F}\ne0$; so $e\ge1$. As $(\mathcal B_e)_{k+1}=\mathcal C_{k+1}$
for $k\ge e-1$, the maps $\bar h_k$, $k\ge e-1$, are also natural maps
$h_k:\mathcal F_k\to\mathcal C_{k+1}$; set $h_k=0$ for $k<e-1$. Then
$\varphi'=\varphi-\partial h-h\partial$ is a chain map. It preserves the augmentation because
$\varepsilon\partial=0$ and $h_{-1}=0$. For $k\ge e$ the boundary of $\mathcal B_e$ from degree
$k+1$ to degree $k$ is that of $\mathcal C$, so the homotopy identity in degree $k$ says
$\varphi_k=\partial h_k+h_{k-1}\partial$, that is, $\varphi'_k=0$. Thus $\varphi'$ has image in
$\mathcal C_{\le e-1}$.

Consequently, with $a=a_{\mathbb F}(C)$, the map $\bar\varphi=\pi_a\varphi:\mathcal F\to\mathcal B_a$
is not null-homotopic: for $a\ge1$ by minimality of $a$, and for $a=0$ by the case $e=0$ above.

\proofstep{Products: the upper bound.}
Let $C$, $C'$ have disjoint domains. A literal set of the product is a disjoint union
$T\sqcup T'$, and it is realizable exactly when $T$ and $T'$ are; so the poset of the product is
$\widehat K(C)\times\widehat K(C')$. For diagrams $D,D'$ let $D\boxtimes D'$ be the diagram
$(T,T')\mapsto D(T)\otimes D'(T')$, and use total complexes with the Koszul sign.
We have $P_U\boxtimes P_{U'}=P_{U\sqcup U'}$ and
$\mathcal C^{C\times C'}=\mathcal C\boxtimes\mathcal C'$, since cubical chains of a product of cubes
are the tensor product of the cubical chains. If $\mathcal F,\mathcal F'$ are finite projective
resolutions, then $\mathcal F\boxtimes\mathcal F'$ is a bounded complex of finite direct sums of
representables, and by the K\"unneth theorem over a field, at each object its homology is
$\mathbb F$ in degree zero, identified by $\varepsilon\otimes\varepsilon'$. So it is a finite projective
resolution for the product.

Let $\varphi,\varphi'$ have images in $\mathcal C_{\le a}$ and $\mathcal C'_{\le a'}$, for
$a=a_{\mathbb F}(C)$ and $a'=a_{\mathbb F}(C')$. Then $\varphi\boxtimes\varphi'$ preserves the
augmentation and has image in the sum of the $\mathcal C_i\boxtimes\mathcal C'_{i'}$ with
$i\le a$, $i'\le a'$, which lies in total degree at most $a+a'$. Thus
$a_{\mathbb F}(C\times C')\le a+a'$.

\proofstep{Products: the lower bound.}
If $a+a'=0$ there is nothing to prove. Otherwise suppose there is an augmentation-preserving map
$\Phi':\mathcal F\boxtimes\mathcal F'\to(\mathcal C\boxtimes\mathcal C')_{\le a+a'-1}$. A summand
$\mathcal C_i\boxtimes\mathcal C'_{i'}$ of total degree at most $a+a'-1$ has $i<a$ or $i'<a'$, so
it maps to zero under $\pi_a\boxtimes\pi'_{a'}:\mathcal C\boxtimes\mathcal C'\to
\mathcal B_a\boxtimes\mathcal B'_{a'}$. Hence $(\pi_a\boxtimes\pi'_{a'})\Phi'=0$. By uniqueness up to
homotopy, $\Phi'\simeq\varphi\boxtimes\varphi'$, so $\bar\varphi\boxtimes\bar\varphi'$ is
null-homotopic as a map $\mathcal F\boxtimes\mathcal F'\to\mathcal B_a\boxtimes\mathcal B'_{a'}$.

Homotopy classes of chain maps are the degree-zero homology of the Hom complex with
$\operatorname{Hom}(\mathcal F,\mathcal B)_k=\prod_i\operatorname{Nat}(\mathcal F_i,\mathcal B_{i+k})$
and differential $\theta\mapsto\partial\theta-(-1)^k\theta\partial$. The Hom complex for the two
external products is the tensor product of the two Hom complexes, through
$\theta\otimes\theta'\mapsto\theta\boxtimes\theta'$ with the Koszul sign. For representable
projectives this is the evaluation identity
\[
 \operatorname{Nat}(P_U,D)\otimes\operatorname{Nat}(P_{U'},D')
 =D(U)\otimes D'(U')
 =\operatorname{Nat}(P_U\boxtimes P_{U'},D\boxtimes D'),
\]
and it extends to finite direct sums. All these spaces are finite-dimensional. The K\"unneth
theorem over a field says that the external product of the two nonzero degree-zero classes
$[\bar\varphi]$ and $[\bar\varphi']$ is nonzero, which contradicts the previous paragraph. Thus the
product map cannot be truncated in degree $a+a'-1$, and $a_{\mathbb F}(C\times C')\ge a+a'$.

\proofstep{Source-free classes: reduction.}
The first part of this lemma and Theorem~\ref{thm:graph-budget} give
$a_{\mathbb F_2}(C_G)\le r+\tau(G)-1$. Suppose a map to $\mathcal C_{\le D}$ exists with
$D=r+\tau(G)-2$. Composing with a comparison map from the simplicial resolution gives an
augmentation-preserving natural chain map $\varphi:\mathcal S\to\mathcal C_{\le D}$, for a compatible
triangulation $\mathcal T$ of $P$. A simplex $\sigma$ of $\mathcal T$ is an element of
$\mathcal S(U(\sigma))$, and the transition maps are inclusions, so naturality gives
\begin{equation}\label{eq:section-preservation}
 \varphi(\sigma)\in C_{*}\bigl(Q_{U(\sigma)};\mathbb F_2\bigr)
 \quad\text{for every simplex }\sigma\text{ of }\mathcal T.
\end{equation}

\proofstep{Source-free classes: the image of the cycle $a_0$.}
We use the notation of the proof of Theorem~\ref{thm:graph-budget}. Every cube $c=Q_W$ of $X$ equals the face
$P\cap Q_W$ of $P$, so it is a subcomplex of $\mathcal T$. Let $\operatorname{sd}(c)$ be the sum of the
$(\dim c)$-simplices of $\mathcal T$ in $c$. This defines a chain map from cubical chains of $X$ to
simplicial chains of $\mathcal T$, since the boundary of the sum of the top simplices of a triangulated
cube is the sum of the top simplices of its facets. We claim that $\varphi(\operatorname{sd}c)=c$. If
$\dim c=0$, then $c$ is a vertex, $Q_{U(c)}=c$, and \eqref{eq:section-preservation} with the augmentation
gives $\varphi(c)=c$. If $\dim c\ge1$, each simplex $\sigma$ of $\operatorname{sd}(c)$ has
$U(\sigma)\supseteq W$, so $\varphi(\operatorname{sd}c)$ is a chain of degree $\dim c$ on the cube $c$; this
chain is $0$ or $c$. By induction, $\partial\varphi(\operatorname{sd}c)=\varphi(\operatorname{sd}\partial c)
=\partial c\ne0$, which forces the value $c$. Therefore the augmented simplicial cycle
$a=\operatorname{sd}(a_0)$ satisfies $\varphi(a)=a_0$, where $a_0$ is the cubical cycle from the proof of
Theorem~\ref{thm:graph-budget}.

\proofstep{Source-free classes: descent in the cube.}
Put $d=r-1$ and $q=\tau(G)-1$. A nonempty $J\subseteq V$ with $|J|\le q$ is not a feedback vertex set. So, as in the
proof of Lemma~\ref{lem:common-filling}, using
Lemma~\ref{lem:source-free-unions}, there are chains $E_\varnothing=a$ and
$E_J\in C_{d+|J|}(\bigcup_{v\in J}Y_v)$, $|J|\le q$, with
$\partial E_J=\sum_{v\in J}E_{J\setminus\{v\}}$, and for $|I|=q+1$ chains
$E_I\in C_{d+q+1}(P)$ with the same boundary relation. Put $z_J=\varphi(E_J)$ for $|J|\le q$. Regard
each $R_v$ as a cubical subcomplex of $\mathcal X=[0,1]^E$; its cells are the faces $Q_{T}$
with $T$ containing a literal with head $v$. A simplex $\sigma$ of $\bigcup_{v\in J}Y_v$ lies in a
face $P\cap F(\ell)$ with the head of $\ell$ in $J$. Then $\ell\in U(\sigma)$, so by
\eqref{eq:section-preservation} the chain $\varphi(\sigma)$ lies on faces $Q_{T}$ with $\ell\in T$, which
are cells of $R_v$ for the head $v\in J$ of $\ell$. Thus $z_J$ has the support required in
Lemma~\ref{lem:support-descent}, with $W_v=R_v$. For $|I|=q+1$,
\[
 \sum_{v\in I}z_{I\setminus\{v\}}=\varphi(\partial E_I)=\partial\varphi(E_I)=0,
\]
because $\varphi(E_I)$ has degree $d+q+1=D+1$ and $\varphi$ takes values in $\mathcal C_{\le D}$. The
lemma, with $t=d$ and $b=\varphi(a)=a_0$, gives a cubical chain $w$ of degree $r$ in $\bigcap_vR_v=X$
with $\partial w=a_0$. But $X$ has no cubes of dimension $r$, so $w=0$ and $a_0=0$, a contradiction.
Thus the cycle $a_0$ and the contractibility of the unions in Lemma~\ref{lem:source-free-unions} obstruct
truncation below $r+\tau(G)-1$, which proves the equality.
\end{proof}

Together with the upper bound of Theorem~\ref{thm:anchor-width}, the lemma gives the exact face width of
products.

\begin{corollary}\label{cor:graph-products}
For finite connected simple graphs $G_1,\ldots,G_k$, each containing a cycle, with cycle ranks
$r_1,\ldots,r_k$,
\[
 u\Big(\prod_{i=1}^kC_{G_i}\Big)
 =1+\sum_{i=1}^k(r_i+\tau(G_i)-1).
\]
\end{corollary}
\begin{proof}
By Lemma~\ref{lem:algebraic-product} over $\mathbb F_2$, applied $k-1$ times to the additivity
statement, and by its first inequality,
$u(\prod_iC_{G_i})\ge1+a_{\mathbb F_2}(\prod_iC_{G_i})=1+\sum_i(r_i+\tau(G_i)-1)$.
The product bound in Theorem~\ref{thm:anchor-width} and Theorem~\ref{thm:graph-budget} give
$u(\prod_iC_{G_i})\le1+\sum_i(u(C_{G_i})-1)=1+\sum_i(r_i+\tau(G_i)-1)$.
\end{proof}

\subsection{The index gap}
We now compare the face width with the index. The sign complex of an anchored class is the union of two
cones over $K(C)$ and $-K(C)$ that meet in $K(C)\cap(-K(C))$, so its index can be bounded from above
through an odd map on this intersection. For an orientation class the natural candidate is the orthogonal
projection onto the cycle space. A zero of this projection is a gradient of a potential on the vertices,
and the extreme level sets of the potential then contain cycles of $G$.

The complex $K(C_G)$ and its negative $-K(C_G)$, the complex of negated faces, are realized in the
boundary of the cross-polytope: $|K|$ is the set of $x\in\mathbb R^E$ with $\|x\|_1=1$ whose
signed support $\{(e,\operatorname{sgn}x_e):x_e\ne0\}$ is a face of $K$. The involution is $x\mapsto-x$.
For $\varphi:V\to\mathbb R$ put
$(\nabla\varphi)_e=\varphi(e^+)-\varphi(e^-)$, where $e^-\to e^+$ is the reference direction of $e$, the direction of the coordinate value $+1$.
Let $\mathcal Z\subseteq\mathbb R^E$ be the cycle space, the kernel of the incidence map $\mathbf 1_e\mapsto\mathbf 1_{e^+}-\mathbf 1_{e^-}$, and let $\Pi$ be
the orthogonal projection onto $\mathcal Z$. The incidence map is the adjoint of $\nabla$, so the
image of $\nabla$ is the orthogonal complement of $\mathcal Z$, and $\ker\nabla$ is the constants since
$G$ is connected. So $\dim\mathcal Z=r$, and each $x$ decomposes as $x=\Pi x+\nabla\varphi(x)$, where
$\varphi(x)$ is the unique such potential with sum zero, linear in $x$.

\begin{lemma}\label{lem:level-cycles}
If $x$ is a point of $|K(C_G)\cap(-K(C_G))|$ with $\Pi x=0$, then the minimum and the maximum level sets
of the potential $\varphi(x)$ are disjoint, and each contains a cycle of $G$.
\end{lemma}
\begin{proof}
Write $\varphi=\varphi(x)$. As $x\ne0$ and $x=\nabla\varphi$, the potential is nonconstant, so the two level
sets $S_{\min},S_{\max}$ are disjoint. Since $x\in|K(C_G)|$, there is a row $c$ containing the signed support
of $x$: every edge with $\varphi(e^+)\ne\varphi(e^-)$ is directed by $c$ from its lower endpoint to its higher
endpoint. Let $v\in S_{\min}$ and let $y\to v$ be an incoming edge of $v$ in $c$. If $y\notin S_{\min}$, then
$\varphi(y)>\varphi(v)$, and $c$ directs this edge from $v$ to $y$, which is impossible. So every vertex of
$S_{\min}$ has an incoming edge from $S_{\min}$. Walking backward along incoming edges inside $S_{\min}$ gives
a repeated vertex and hence a directed cycle, of length at least three since $G$ is simple. Since
$-x\in|K(C_G)|$, there is a row $c'$ directing every such edge from its higher endpoint to its lower endpoint.
Let $v\in S_{\max}$ and let $y\to v$ be an incoming edge of $v$ in $c'$. If $y\notin S_{\max}$, then $\varphi(y)<\varphi(v)$, and $c'$
directs this edge from $v$ to $y$, which is impossible. So the backward walk along incoming edges of $c'$ stays in
$S_{\max}$ and gives a cycle there.
\end{proof}

Any two cycles of $K_4$ share a vertex, so for $G=K_4$ the projection $\Pi$ has no zero on
$|K(C_G)\cap(-K(C_G))|$. This gives the upper bound on the index in the main result of this section.

\begin{theorem}[Unbounded additive index gap]\label{thm:index-gap}
Let $C_0=C_{K_4}$ and, for an integer $k\ge1$, let $D_k=(C_0^k)^+$. Then $C_0$ has $32$ rows on
$6$ coordinates, and
\[
 |D_k|=32^k,\qquad |\operatorname{dom}D_k|=6k+1,\qquad
 \operatorname{lr}(D_k)=4k+1,\qquad\operatorname{Ind}S(D_k)=3k.
\]
Consequently $\operatorname{lr}(D_k)-(\operatorname{Ind}S(D_k)+1)=k$,
and $\operatorname{lr}(D_k)/(\operatorname{Ind}S(D_k)+1)\to4/3$.
\end{theorem}
\begin{proof}
\proofstep{Counts and the list number.}
In a tournament on four vertices, at most one vertex is a source: two sources would both be tails of the
edge joining them. A specified source fixes three edges, leaving $2^3=8$ tournaments, so there are
$2^6-4\cdot8=32$ source-free orientations. The constant coordinate adds nothing to the number of rows, so
$|D_k|=32^k$ and $|\operatorname{dom}D_k|=6k+1$. The graph $K_4$ has $r=3$ and $\tau=2$. By
Corollary~\ref{cor:graph-products}, $u(C_0^k)=1+k(3+2-1)=4k+1$, and
Theorem~\ref{thm:anchor-width} gives $\operatorname{lr}(D_k)=u(C_0^k)=4k+1$.

\proofstep{An odd map to the sphere $S^{3k-1}$.}
Let $K=K(C_0)$ be its literal complex and $L=K\cap(-K)$. The projection $\Pi$ onto the cycle space
of $K_4$, of dimension three, has no zero on $|L|$: a zero would give, by Lemma~\ref{lem:level-cycles}, two
vertex-disjoint cycles in $K_4$, whereas a cycle has at least three of the four vertices.

A consistent literal set on the product is a disjoint union of literal sets on the factors, and it lies in a
row of $C_0^k$ exactly when each part lies in a row of $C_0$. So
$K(C_0^k)$ and $-K(C_0^k)$ are the $k$-fold joins of $K$ and $-K$, and
$L_k:=K(C_0^k)\cap(-K(C_0^k))$ is the $k$-fold join of $L$. Write a point of $|L_k|$ as
$x=(x^1,\ldots,x^k)$. If $x^i\ne0$ then $x^i/\|x^i\|_1\in|L|$, so $\Pi x^i\ne0$. Since some $x^i\ne0$, the
linear map $x\mapsto(\Pi x^1,\ldots,\Pi x^k)\in\mathbb R^{3k}$ has no zero on $|L_k|$. Normalization gives an
odd map $h:|L_k|\to S^{3k-1}$.

\proofstep{The upper bound on the index.}
Let $a$ be the anchor coordinate and write $K_k=K(C_0^k)$. A face of the sign complex of $D_k$ is a subset of
the literal set of a row $(c,+)$ or of its negative. Hence
\[
 S(D_k)=(K_k*\{a^+\})\cup((-K_k)*\{a^-\}),
\]
and the two cones meet exactly in $L_k$, since a common face contains no anchor literal. Let $\mathrm{sd}K_k$
be the barycentric subdivision. Define a simplicial map on $\mathrm{sd}K_k*\{a^+\}$ by fixing $a^+$ and the
barycenter of each face of $L_k$ and sending the barycenter of each other face to $a^+$. A simplex is a chain
of faces of $K_k$, possibly with $a^+$. Because $L_k$ is a simplicial complex, the members of the chain lying
in $L_k$ form an initial segment, itself a chain in $L_k$. So the image is a simplex of $\mathrm{sd}L_k*\{a^+\}$,
and this is a retraction $\rho^+$ of the positive cone onto $|L_k|*\{a^+\}$. On the negative cone put
$\rho^-(x)=-\rho^+(-x)$. Both maps are the identity on $|L_k|$, so they agree on the common base and define a
continuous equivariant retraction of $|S(D_k)|$ onto the suspension
$\Sigma|L_k|=(|L_k|*\{a^+\})\cup(|L_k|*\{a^-\})$. The formula
\[
 (1-t)y+ta^\pm\longmapsto\bigl(\cos(\tfrac{\pi t}2)h(y),\ \pm\sin(\tfrac{\pi t}2)\bigr)
 \in S^{3k}\subset\mathbb R^{3k}\times\mathbb R
\]
is a well-defined continuous odd map $\Sigma|L_k|\to S^{3k}$. Hence
$\operatorname{Ind}S(D_k)\le3k$.

\proofstep{The lower bound on the index.}
Fix a spanning tree of $K_4$. Its complement, a cotree, has three edges, and these three coordinates are
shattered by $C_0$: after prescribing them, choose the head of one as the root and direct the spanning tree away
from it. The root has the prescribed incoming cotree edge and every other vertex has an incoming tree edge. The
product shatters the $3k$ cotree coordinates. Together with the anchor these are $3k+1$ signed-shattered columns
of $D_k$: a pattern with anchor value $+$ is the restriction of a row $(c,+)$, and a pattern with anchor value $-$
is the restriction of $-(c,+)$ for the row $c$ having the negated pattern on the cotrees. By the remark on
signed-shattered columns in Section~\ref{ssec:2.1}, $S(D_k)$ contains an equivariant $S^{3k}$, so
$\operatorname{coInd}S(D_k)\ge3k$, and $\operatorname{Ind}S(D_k)\ge\operatorname{coInd}S(D_k)$ by Borsuk--Ulam.

\proofstep{The gap.}
We get $\operatorname{lr}(D_k)-(\operatorname{Ind}S(D_k)+1)=4k+1-3k-1=k$, and the ratio is
$(4k+1)/(3k+1)\to4/3$.
\end{proof}

The odd map in this proof uses only the absence of two disjoint cycles in $K_4$. For a
general graph the projection $\Pi$ can vanish on $|K(C_G)\cap(-K(C_G))|$, at a point whose extreme level
sets contain disjoint cycles. The same potential argument then bounds the index of this complex in terms
of the number of pairwise-intersecting families needed to cover the cycles, with one additional
coordinate for each family beyond the first. For $K_4$ one family suffices, and the following bound is
the value $r-1=2$ used above for $k=1$.

\begin{proposition}\label{prop:cycle-coloring}
Let $G$ be a finite connected simple graph containing a cycle. Call a family of cycles of $G$
pairwise intersecting if any two of its cycles share a vertex, and
let $\chi_c(G)$ be the least number of pairwise-intersecting families
covering all cycles of $G$. Then
\[
 \operatorname{Ind}\bigl(K(C_G)\cap(-K(C_G))\bigr)
 \le r+\chi_c(G)-2.
\]
\end{proposition}
\begin{proof}
Use the linear decomposition $x=\Pi x+\nabla\varphi(x)$ above. Let $\mathcal F_1,\ldots,\mathcal F_{\chi}$,
$\chi=\chi_c(G)$, be nonempty families of cycles, each pairwise intersecting, whose union contains all cycles.
For $\varphi:V\to\mathbb R$ and $i\le\chi$ set
\[
 l_i(\varphi)=\min_{C\in\mathcal F_i}\max_{v\in C}\varphi(v),\quad
 h_i(\varphi)=\max_{C\in\mathcal F_i}\min_{v\in C}\varphi(v),\quad
 y_i=l_i+h_i.
\]
Pairwise intersection implies $h_i\le l_i$: for $C,C'\in\mathcal F_i$ pick $w\in C\cap C'$. Then
$\min_C\varphi\le\varphi(w)\le\max_{C'}\varphi$. These functions are continuous, and
$l_i(-\varphi)=-h_i(\varphi)$, $h_i(-\varphi)=-l_i(\varphi)$, so $y_i$ is odd. Also
$y_i(\lambda\varphi+t)=\lambda y_i(\varphi)+2t$ for $\lambda>0$ and $t\in\mathbb R$. Therefore the vector
$Y=(y_i-y_{\chi})_{i<\chi}\in\mathbb R^{\chi-1}$ is continuous, odd, invariant under adding constants, and
positively homogeneous.

Let $x\in|K(C_G)\cap(-K(C_G))|$ with $\Pi x=0$. By Lemma~\ref{lem:level-cycles} the minimum level of
$\varphi(x)$ contains a cycle $C$ and the maximum level contains a cycle $C'$, and they are disjoint. So
$\chi\ge2$ in this case, and $C\in\mathcal F_i$, $C'\in\mathcal F_j$ with $i\ne j$. By the homogeneity, we may
normalize the extrema of $\varphi$ to zero and one. Then $\max_C\varphi=0$ gives $l_i=0$, and $0\le h_i\le l_i$
gives $y_i=0$. Likewise $\min_{C'}\varphi=1$ gives $h_j=1$, and $1=h_j\le l_j\le1$ gives $y_j=2$. So
$y_1,\ldots,y_\chi$ are not all equal, and $Y(\varphi(x))\ne0$.

Thus the map $x\mapsto(\Pi x,Y(\varphi(x)))$ is continuous, odd, and has no zero on the complex, with values
in a vector space of dimension $r+\chi-1$. For $\chi=1$ the second component is absent and the preceding paragraph
shows $\Pi x\ne0$. Normalizing to the sphere gives an odd map to $S^{r+\chi-2}$.
\end{proof}

\subsection{The unanchored product family}
By Theorem~\ref{thm:anchor-width}, appending a constant coordinate raises the list number by at most one,
so the value $\operatorname{lr}(D_k)=4k+1$ leaves two possibilities for the list number of $C_{K_4}^{\,k}$.
We show that it is $4k$, so that on this family the anchor raises the list number by exactly one. The
lower bound follows from Corollary~\ref{cor:graph-products}. The upper bound rests on the integral link
criterion of Theorem~\ref{thm:literal-upper} for $C_{K_4}$, which requires the homology of all links of
its literal complex, and on the additive join inequality of~\cite[Theorem 1.12]{BHHLT26} for the products.

The complex $K(C_{K_4})$ is small enough for its links to be computed exactly.

\begin{lemma}\label{lem:k4-links}
Let $K=K(C_{K_4})$. The following table lists, for each size $|S|$ of a face $S$ of $K$, including the
empty face, the nonzero reduced integral homology of the link $\operatorname{lk}_KS$ and the number of faces
with this homology. The symbol $0$ means that all reduced homology groups vanish. All displayed groups are
torsion-free.
\begin{center}
\upshape
\begin{tabular}{clr}
\toprule
$|S|$ & Nonzero reduced homology of $\operatorname{lk}_KS$ & Number of faces\\
\midrule
0 & $H_3=\mathbb Z^3$ &1\\
1 & $H_2=\mathbb Z$ &12\\
2 & $H_1=\mathbb Z$ &12\\
2 & $0$ &48\\
3 & $H_0=\mathbb Z$ &4\\
3 & $0$ &152\\
4 & $H_1=\mathbb Z$ &30\\
4 & $0$ &180\\
5 & $H_0=\mathbb Z$ &60\\
5 & $0$ &72\\
6 & $H_{-1}=\mathbb Z$ &32\\
\bottomrule
\end{tabular}
\end{center}
\end{lemma}
\begin{proof}
The table is the result of a finite computation over the integers, verified exactly
(Appendix~\ref{sec:computation}), which we now describe. Among the $64$ orientations of $K_4$,
those having positive indegree at every vertex are the $32$ rows of $C_{K_4}$. All
subsets of their literal sets form $K$, which has $603$ faces, including the
empty face. The link of a face $S$ consists of the faces
$T$ with $T\cap S=\varnothing$ and $T\cup S\in K$.

Permutations of the four graph vertices act on $K$ by simplicial automorphisms, so faces in one orbit have
isomorphic links. It therefore suffices to treat one representative from each of the
$33$ orbits, which partition all $603$ faces, and to weight the corresponding table entry
by the orbit size.

To obtain the boundary matrices, order the six edges of $K_4$ lexicographically and label the two
directions of each edge by consecutive integers, which orders the twelve literals. With every face
written in increasing order, the augmented boundary is
\[
 \partial[v_0,\ldots,v_j]
   =\sum_{i=0}^j(-1)^i[v_0,\ldots,\widehat v_i,\ldots,v_j].
\]
These integer matrices are determined by the enumeration and this formula, so the computation is exact.
Integer row and column elimination of the matrices of each representative gives the table.
In this Smith reduction
every nonzero invariant factor is one. Thus there is no torsion, and the ranks of consecutive
boundary matrices determine the free ranks by
$b_j=f_j-\operatorname{rank}\partial_j-
\operatorname{rank}\partial_{j+1}$, where $f_j$ is the number of $j$-dimensional faces of the link.
\end{proof}

\begin{proposition}\label{prop:unanchored-products}
For every $k\ge1$,
\[
 \operatorname{lr}(C_{K_4}^{\,k})=4k,
 \qquad u(C_{K_4}^{\,k})=4k+1.
\]
\end{proposition}
\begin{proof}
Write $K=K(C_{K_4})$. Lemma~\ref{lem:k4-links} gives
$L^\circ_{\mathbb Z}(K)=3$ and $L_{\mathbb Z}(K)=4$. Indeed the table shows that all reduced
homology groups of links are free. By the universal coefficient theorem, the reduced cohomology of a link is
then free and in the same degrees as its reduced homology. The largest degree with a nonzero group among
nonempty faces is $2$, attained at $|S|=1$, and for the empty face it is $3$.

Since $K$ is a completion of itself, Theorem~\ref{thm:literal-upper} with $d=3$ therefore gives
$\operatorname{lr}(C_{K_4})\le4$. The reverse inequality follows
from $u(C_{K_4})=5$, the value $r+\tau=3+2$ of Theorem~\ref{thm:graph-budget}, and the inequality
$u\le\operatorname{lr}+1$ of Theorem~\ref{thm:anchor-width}.
For products, the additive join inequality of
\cite[Theorem 1.12]{BHHLT26}, namely
$\operatorname{lr}(C_1*C_2)\le\operatorname{lr}(C_1)+\operatorname{lr}(C_2)$ for classes on disjoint domains,
where $C_1*C_2$ is the Cartesian product of the two classes, gives by induction on $k$
\[
 \operatorname{lr}(C_{K_4}^{\,k})\le4k.
\]
Corollary~\ref{cor:graph-products} gives $u(C_{K_4}^{\,k})=1+4k$, and the same inequality
$u\le\operatorname{lr}+1$ gives $\operatorname{lr}(C_{K_4}^{\,k})\ge4k$.
\end{proof}

\section{Oriented-matroid representations}\label{app:om}
A real sign representation places the rows of a class and their negatives among the
regions of a central hyperplane arrangement, and such a representation yields the bounds
$\operatorname{lr}\le\operatorname{signrk}$~\cite[Corollary 2.9]{BHH26a} and
$\operatorname{Ind}+1\le\operatorname{signrk}$~\cite[Corollary 2]{FHV26}.
This section replaces hyperplanes by pseudospheres, that is, real representations
by oriented matroids, and shows that both bounds hold with the rank of the oriented
matroid in place of the sign rank. Given a labeled
pseudosphere arrangement containing the prescribed rows as regions,
its covector complex gives an intermediate finite model, from which we
obtain an accurate cover whose multiplicity is at most the rank of the
arrangement. On the full tope class of an oriented matroid this bound is an equality.
The oriented-matroid rank can be strictly smaller than the sign rank: the Hadamard
matrix of order eight has list number and oriented-matroid rank four and sign rank five.
For larger Hadamard matrices, Forster's spectral bound limits the size of every
realizable restriction of such an arrangement.

\subsection{The oriented-matroid rank}
We first define the parameter and compare it with the sign rank.
We use the covector and tope conventions of~\cite{BLSWZ99}.
For a loopless oriented matroid $M$ on $[n]$, write $\mathcal V^*(M)$
for its covectors and $\mathcal T(M)\subseteq\{\pm1\}^n$ for its
topes. Covectors are ordered by extension of their nonzero entries.
Define
\[
 \operatorname{omr}(C)=\min\{\operatorname{rank}M:
                       C\cup(-C)\subseteq\mathcal T(M)\},
\]
where the ground set of $M$ is exactly the set of columns.
Since the tope set is closed under negation, this is the oriented-matroid rank of a sign-vector set defined
in~\cite[Definition 4.9]{Lien23}.

A real sign representation gives an oriented matroid, so
$\operatorname{omr}(C)\le\operatorname{signrk}(C)$. Indeed, if
$c(x)\langle w_c,v_x\rangle>0$ for vectors in $\mathbb R^\rho$, then every $v_x$ is nonzero, so the
oriented matroid of the configuration $(v_x)_x$ is loopless, of rank at most $\rho$. Its covectors are the sign
vectors of $x\mapsto\langle w,v_x\rangle$, $w\in\mathbb R^\rho$, so $\pm c$ is a covector of
full support, that is, a tope.

In a real arrangement, the functionals with prescribed signs on a set of columns form
an open convex cone. The construction below uses the following substitute for this
convexity. A sign vector $Y$ \emph{extends} a set $S$ of literals if $Y(j)=s$ for
every literal $(j,s)\in S$.

\begin{lemma}\label{lem:om-regions}
Let $M$ be a loopless oriented matroid, and let $S$ be a nonempty literal set extendable to a tope of $M$.
The order complex of
\[
 P_S=\{Y\in\mathcal V^*(M)\setminus\{0\}:Y\text{ extends }S\}
\]
is nonempty and contractible.
\end{lemma}
\begin{proof}
The set $P_S$ contains a tope extending $S$, so its order complex is nonempty.
The topological representation theorem identifies the nonzero covector
poset with the face poset of a regular pseudosphere arrangement
$(S_j)_{j\in[n]}$ on $S^{r-1}$, where $r=\operatorname{rank}M$. Each $S_j$ has two open
sides $S_j^\pm$, and the open cell of a covector $Y$ is the set of points with
sign vector $Y$. If $r=1$, the sphere $S^0$ has two points, $P_S$ has one element, and there is
nothing to prove. Let $r\ge2$.

\proofstep{The closed region is a ball.}
The intersection $A$ of the closed sides prescribed by $S$ is a sphere or a ball, by the
closed-side intersection property of the representation~\cite[Chapters 4--5]{BLSWZ99}.
It contains the closed cell of a tope extending $S$, a ball of dimension $r-1$. So if $A$ were a
sphere, its dimension would be $r-1$, and invariance of domain would give $A=S^{r-1}$. But $S$ is nonempty,
and for a literal $(j,s)\in S$ the opposite open side $S_j^{-s}$ is nonempty and misses $A$. So $A$ is a closed
ball of dimension $r-1$. By invariance of domain its manifold interior is open in $S^{r-1}$.

\proofstep{The open region is contractible.}
The intersection of the open sides prescribed by $S$ is $U=A\setminus E$, where
$E$ is the set of points of $A$ lying on the pseudosphere of at least one literal of $S$, a closed subset. We show
$E\subseteq\partial A$. A point on one of the prescribed pseudospheres $S_j$ has every neighborhood
meeting its opposite side, because a pseudosphere is the image of an equator under a homeomorphism of the
sphere, and that side is disjoint from $A$. So the point is not interior to $A$ in $S^{r-1}$, and it lies in
the manifold boundary $\partial A$.
Under a homeomorphism from $A$ to a closed ball, $U$ is a ball with
some boundary points removed. Radial contraction to its center stays
inside this set. Thus $U$ is contractible.

\proofstep{The order complex.}
The cells of $U$ are exactly those indexed by $P_S$, and $P_S$ is an upper set of the nonzero covector poset.
The barycentric subdivision of the regular cell complex is the order complex of its face poset, and the open
simplex of a chain lies in the open cell of the largest member of the chain.
So a point of $U$ lies in the open simplex of a chain whose
largest cell belongs to $P_S$, and the members of the chain in $P_S$ form an upper segment.
Erase the barycentric coordinates of
cells outside $P_S$ and renormalize. The sum of the coordinates that are kept is positive, since the largest member has a positive
coordinate. Linear interpolation before normalization,
\[
 \sum_i\lambda_ib_{Y_i}\ \longmapsto\
 \frac{\sum_{Y_i\in P_S}\lambda_ib_{Y_i}+(1-t)\sum_{Y_i\notin P_S}\lambda_ib_{Y_i}}
 {\sum_{Y_i\in P_S}\lambda_i+(1-t)\sum_{Y_i\notin P_S}\lambda_i},
\]
is continuous on $U\times[0,1]$, because the formula on a simplex restricts to the formula on its faces. It
keeps the largest member of the chain, so it stays in $U$, and it fixes $\Delta(P_S)$ pointwise. This gives a
deformation retraction of $U$ onto $\Delta(P_S)$, which proves the lemma.
\end{proof}

\subsection{An accurate cover from the covector complex}
We now bound the list number by the oriented-matroid rank.
Chepoi, Knauer, and Philibert~\cite[Theorem 2]{CKP21} complete a
rank-$r$ tope class to an ample class of VC dimension $r$.
Combined with~\cite[Theorem B]{BHH26b}, this gives
$\operatorname{lr}(C)\le\operatorname{omr}(C)+1$.
The following face-compatible construction removes the extra one
and determines the value on full tope classes.

\begin{theorem}\label{thm:om-bound}
For every finite nonempty class,
\[
 \operatorname{lr}(C)\le\operatorname{omr}(C)
 \le\operatorname{signrk}(C),\qquad
 \operatorname{Ind}S(C)+1\le\operatorname{omr}(C).
\]
If $C=\mathcal T(M)$ and $M$ has rank $r$, then
\[
 \operatorname{lr}(C)=\operatorname{omr}(C)=r,
 \qquad\operatorname{Ind}S(C)=\operatorname{coInd}S(C)=r-1.
\]
\end{theorem}
\begin{proof}
The inequality $\operatorname{omr}(C)\le\operatorname{signrk}(C)$ was shown above. Let $M$ be a loopless oriented
matroid of rank $r=\operatorname{omr}(C)$ with $C\cup(-C)\subseteq\mathcal T(M)$.

\proofstep{The index bound.}
Let $L=\Delta(\mathcal V^*(M)\setminus\{0\})$. By the topological representation theorem it has
dimension $r-1$. The map $Y\mapsto-Y$ is a simplicial involution with no invariant simplex, because
comparable covectors agree on the support of the smaller one, so $Y$ and $-Y$ never lie in one chain. Thus
$L$ is a free $\mathbb Z_2$-complex of dimension $r-1$. Hence $\operatorname{Ind}L\le r-1$: define an odd map to $S^{r-1}$ on
orbit representatives of simplices, in increasing dimension, using that $S^{r-1}$ is $(r-2)$-connected.

We next construct an equivariant map from $S(C)$ to $L$.
On the barycentric subdivision of $S(C)$, a simplex is a chain
$S_0\subsetneq\cdots\subsetneq S_k$ of nonempty literal faces. Every such face is extendable to a tope of $M$,
because $C\cup(-C)\subseteq\mathcal T(M)$. We map this simplex into $\Delta(P_{S_0})$,
by induction over simplex dimension. A vertex $S_0$ goes to a point of the nonempty complex $\Delta(P_{S_0})$. A boundary
subchain has least face containing $S_0$, so its previously defined
image is in $\Delta(P_{S'})\subseteq\Delta(P_{S_0})$, where $S'\supseteq S_0$ is that least face.
Lemma~\ref{lem:om-regions} extends the boundary map over the simplex.
Make choices in antipodal pairs to obtain an equivariant map: the involution on chains has no invariant simplex since
$-S_0\ne S_0$, and $-\Delta(P_{S_0})=\Delta(P_{-S_0})$. We get an equivariant map
$F:S(C)\to L$. Together with the odd map from $L$ to $S^{r-1}$, this proves the index bound $\operatorname{Ind}S(C)\le r-1$.

\proofstep{Maximal coordinates.}
The map $F$ controls the coordinates of largest mass, as follows.
Write $x$ for the signed coordinate masses of a realizable distribution,
so $\|x\|_1=1$ and $x$ is a point of the row simplex of a row of $C$. Let $m_1>\cdots>m_t>0$ be the distinct values
of $|x_j|$ and $S_i=\{(j,\operatorname{sgn}x_j):|x_j|\ge m_i\}$. Then, with $m_{t+1}=0$,
$x=\sum_i(m_i-m_{i+1})|S_i|\,b_{S_i}$, where $b_{S_i}$ is the barycenter of the face $S_i$. So in its
barycentric carrier chain, the least face $S_1$ consists exactly of the coordinates
of maximum absolute mass. Hence, if $j$ is a coordinate of maximum absolute mass and $x_j$ has sign $s$, then
$F(x)$ lies in
$L_{j,s}:=\Delta(P_{\{(j,s)\}})$.
These two compact subcomplexes are disjoint for opposite signs, because they are full subcomplexes on the disjoint vertex
sets $P_{\{(j,+)\}}$ and $P_{\{(j,-)\}}$.

\proofstep{A coordinate clamp.}
To extend this control from the largest coordinates to all coordinates of non-negligible mass, fix $0<\delta<1/n$. Define a continuous coordinate clamp by
$z_j=\operatorname{sgn}(x_j)\min\{|x_j|/\delta,1\}$ and
$\rho(x)=z/\|z\|_1$, where $z\ne0$ because some $|x_j|\ge1/n>\delta$. It preserves the support and signs of $x$,
and hence preserves realizability. If $|x_j|\ge\delta$, then $|z_j|=1$ is the largest absolute value of a coordinate of
$z$, so $j$ is a maximal coordinate of $\rho(x)$. Consequently
$F(\rho(x))\in L_{j,\operatorname{sgn}x_j}$.

\proofstep{The cover.}
Let $\eta>0$ be smaller than the distance between $L_{j,+}$ and $L_{j,-}$ for all $j$.
Subdivide $L$ so finely that every closed vertex star has diameter less than $\eta$. Then a star meets at most
one of $L_{j,+},L_{j,-}$ for each $j$. The maximum-barycentric-coordinate
regions $D_v$ of Lemma~\ref{res:2.0} give a closed cover of multiplicity
at most $r$, each member lying in one such star. Label a member $D_v$ by the Boolean vector $h_v$ whose value at
coordinate $j$ is the sign of the subcomplex it meets, with an arbitrary
choice if it meets neither. Its pullback $R_v=(F\circ\rho)^{-1}(D_v)$ is a closed subset of the realizable space.
These sets cover it with multiplicity at most $r$.

If $x\in R_v$ and $|x_j|\ge\delta$, then
$F(\rho(x))\in D_v\cap L_{j,\operatorname{sgn}x_j}$, so $(h_v)_j=\operatorname{sgn}x_j$. Thus $h_v$ agrees
with every coordinate of mass at least $\delta$. Its loss is therefore
at most $\sum_{|x_j|<\delta}|x_j|\le n\delta$. Given $\varepsilon>0$, choose $\delta$ with $n\delta<\varepsilon/2$ and apply
Proposition~\ref{prop:cover-rule} with $\alpha=\tau=\varepsilon/2$.
This proves $\operatorname{lr}(C,\varepsilon)\le r$ for every $\varepsilon>0$, that is, $\operatorname{lr}(C)\le r$.

\proofstep{Tope classes.}
For the exact statement, let $C=\mathcal T(M)$ with $\operatorname{rank}M=r$. A basis $B$ of $M$ is
shattered by its topes: the restriction $M|_B$ is free, so every sign pattern on $B$ is the restriction of a
covector $Y$ of $M$, and the composition $Y\circ T$ with any tope $T$ is a tope extending this pattern.
Restricting to that basis gives the full
$r$-cube. Restriction of a rule shows that restricting the class to a set of columns does not increase the list number,
so $\operatorname{lr}(C)\ge r$ by Theorem~\ref{res:5.1}.
The set $B$ is signed-shattered, so $S(C)$ contains an equivariant $S^{r-1}$ subcomplex
(Section~\ref{ssec:2.1}), and $\operatorname{coInd}S(C)\ge r-1$. Moreover $\operatorname{omr}(C)\le r$ because
$M$ itself is admissible. Therefore $r\le\operatorname{lr}(C)\le\operatorname{omr}(C)\le r$ and
$r-1\le\operatorname{coInd}S(C)\le\operatorname{Ind}S(C)\le\operatorname{omr}(C)-1\le r-1$.
\end{proof}

\subsection{Uniform completion and contraction}
Two operations on oriented matroids are used in the sequel. The first makes an oriented
matroid uniform without changing its rank or losing topes. It reduces the minimum
defining $\operatorname{omr}$ to uniform oriented matroids, for which duality takes the
simple form of Lemma~\ref{lem:uniform-duality} in Section~\ref{app:low-index}.
The same-rank uniform completion is standard; see~\cite[Section 4.2]{CKP21}.
The following proof shows how the prescribed topes are retained.
\begin{lemma}[Uniform extension]\label{lem:uniform-om}
Every loopless rank-$r$ oriented matroid on $n$ elements has a uniform
rank-$r$ oriented matroid on the same elements containing all its topes.
Thus the minimum defining $\operatorname{omr}$ may be taken over
uniform oriented matroids.
\end{lemma}
\begin{proof}
Process the elements in a fixed order, maintaining a loopless
rank-$r$ oriented matroid containing every original tope, such that every set of at most $r$
previously processed elements is independent. For the next element $e$, which is a nonloop, choose
$f_1,\ldots,f_{r-1}$ so that $(e,f_1,\ldots,f_{r-1})$ is a basis of the
current matroid. Take the lexicographic extension
$e'=[e^+,f_1^+,\ldots,f_{r-1}^+]$ of~\cite[Section 7.2]{BLSWZ99}. It has the same rank, and
for every cocircuit $Y$ of the current matroid the vector $(Y,\sigma(Y))$ is a covector of the extension, where
$\sigma(Y)$, the new entry, is the first nonzero entry of $Y$ in that list. A cocircuit vanishes on a hyperplane and
cannot vanish on the spanning list, so $\sigma(Y)\ne0$.

\proofstep{Topes are retained.}
Write a tope $T$ as a composition of cocircuits $Y_i\le T$.
Since $T(e)\ne0$, choose the first cocircuit with $Y_1(e)=T(e)$.
Its lifted entry at $e'$ is $T(e)$. Composing the lifted cocircuits,
with this one first, gives the covector $(T,T(e))$. Thus $T$ has a
tope extension with its old sign at $e$. Now delete $e$ and rename $e'$ as $e$:
the original tope is retained.

\proofstep{Independence.}
Let $I$ be a set of size less than $r$, independent in the current matroid. Its closure is contained in a hyperplane;
let $Y$ be a cocircuit vanishing on that hyperplane. Then $(Y,\sigma(Y))$ vanishes on $I$ and is nonzero at $e'$, so $e'$
does not belong to the span of $I$, and $I\cup\{e'\}$ is independent.
With $I=\{f_1,\ldots,f_{r-1}\}$ this shows that
$\{e',f_1,\ldots,f_{r-1}\}$ is a basis avoiding $e$.
Deleting $e$ therefore preserves rank $r$. The other old
elements are nonloops and $e'$ lies in this basis, so looplessness
also persists.

Let $A$ be a set of at most $r$ elements processed so far, including the new element. If $e'\notin A$, then $A$ is
independent by the induction hypothesis, since deletion of $e$ does not change the restriction to the other old elements.
If $e'\in A$, the set $I=A\setminus\{e'\}$ has size less than $r$ and is independent by the induction
hypothesis, so $A$ is independent by the preceding argument. This holds also when fewer than $r-1$ elements have been
processed.

Subsequent steps
leave the restriction to the processed elements unchanged. After all elements
are processed, every $r$-set is a basis, so the matroid is uniform.
\end{proof}

The second operation is contraction, the combinatorial form of projecting a configuration
along one of its vectors. Two topes that differ at a coordinate $j$ and agree on a set $A$
give a tope of the contraction of $j$ on $A$. On the Sylvester Hadamard matrices of Section~\ref{ssec:5}, this shows that the
oriented-matroid rank and the sign rank both grow with each doubling of the order.

\begin{proposition}\label{prop:om-contraction}
Let $M$ be a loopless oriented matroid.
Suppose two topes of $M$ agree on $A$ and have opposite signs at $j\notin A$.
Their common restriction to $A$ is a tope of $(M/j)|_A$.
In particular, for the Sylvester Hadamard matrices,
\[
 \operatorname{omr}(H_{2^{k+1}})\ge\operatorname{omr}(H_{2^k})+1,
 \qquad
 \operatorname{signrk}(H_{2^{k+1}})\ge\operatorname{signrk}(H_{2^k})+1.
\]
\end{proposition}
\begin{proof}
\proofstep{Elimination.}
Covector elimination at $j$ gives a covector with zero entry at $j$
and the common signs at every agreeing coordinate. Deleting $j$ gives a covector of $M/j$, whose restriction to
$A$ has full support and is consequently a tope of the contraction
restricted to $A$. The contraction lowers rank by one, since $j$ is a nonloop, and restriction does not raise it.

\proofstep{The oriented-matroid rank of Hadamard matrices.}
The signed rows of $H_{2^{k+1}}$ are the affine sign functions on $\mathbb F_2^{k+1}$, that is, the functions
$x\mapsto(-1)^{a\cdot x+b}$ with $a\in\mathbb F_2^{k+1}$ and $b\in\mathbb F_2$.
Choose an affine hyperplane $A\subset\mathbb F_2^{k+1}$
and $j\notin A$, and let $\chi$ be the affine sign function equal to $+1$ on $A$ and $-1$ on its complement.
Each affine sign function $f$ has the partner $f\chi$, agreeing on
$A$ and reversed on its complement. The restrictions to $A\cong\mathbb F_2^k$ comprise all
affine sign functions on $\mathbb F_2^k$, since each of them extends to $\mathbb F_2^{k+1}$. Let $M$ be a loopless
oriented matroid of rank $\operatorname{omr}(H_{2^{k+1}})$ having all affine sign functions as topes. Apply the first
assertion to each pair $f,f\chi$: all affine sign functions on $A$ are topes of $(M/j)|_A$, which is
therefore loopless, of rank at most $\operatorname{rank}M-1$. This gives the first inequality.

\proofstep{The sign rank of Hadamard matrices.}
In a real representation $(w_f,v_x)$ of rank $\rho=\operatorname{signrk}(H_{2^{k+1}})$, with $w_{-f}=-w_f$, the same
elimination is a positive linear
combination of the two row functionals that vanishes at column $j$. Indeed $\langle w_f,v_j\rangle$ and
$\langle w_{f\chi},v_j\rangle$ have opposite signs, so
\[
 w=|\langle w_{f\chi},v_j\rangle|\,w_f+|\langle w_f,v_j\rangle|\,w_{f\chi}
\]
satisfies $\langle w,v_j\rangle=0$, and $f(x)\langle w,v_x\rangle>0$ for $x\in A$ because $f\chi=f$ on $A$.
Replacing $v_x$ by its orthogonal projection to the hyperplane $v_j^\perp$ does not change $\langle w,v_x\rangle$.
This gives a representation of all affine sign functions on $A$, of rank $\rho-1$, which proves the
sign-rank assertion.
\end{proof}

\subsection{The Hadamard matrix of order eight}
We can now compare the three parameters on a concrete matrix. For the Sylvester Hadamard
matrix $H_{2^m}$, Theorem~\ref{res:5.5} gives list number $m+1$, and Forster's
bound~\cite{Forster02} gives sign rank at least $2^{m/2}$. At order eight this lower
bound is three. The next theorem determines the oriented-matroid rank and the sign rank
at this order and shows that the two differ. Write $H_8^\pm$ for the class of the rows
of $H_8$ and their negatives.

\begin{theorem}[The Hadamard matrix of order eight]\label{thm:h8-om}
The matrix $H_8$ satisfies
\[
 \operatorname{lr}(H_8)=\operatorname{lr}(H_8^\pm)
 =\operatorname{omr}(H_8)=4,
 \qquad \operatorname{signrk}(H_8)=5.
\]
Consequently $6\le\operatorname{signrk}(H_{16})\le7$.
\end{theorem}
\begin{proof}
\proofstep{Learning identities and the lower bound on rank.}
The learning identities follow from Theorem~\ref{res:5.5}, which gives $\operatorname{lr}(H_8,\varepsilon)=4$ for
$\varepsilon\in(0,1/2)$, and the
affine-chain proof for the symmetric class in
Theorem~\ref{res:7.6} of Section~\ref{ssec:7}, which gives $\operatorname{lr}(H_8^\pm,\varepsilon)\le4$. A rule for
$H_8^\pm$ is a rule for its subclass $H_8$, and $\operatorname{lr}(C,\varepsilon)$ is nonincreasing in $\varepsilon$,
so these fixed-error values give the two identities. Theorem~\ref{thm:om-bound} gives
$\operatorname{omr}(H_8)\ge\operatorname{lr}(H_8)=4$.

This lower bound can also be seen directly. Four columns form an
affine basis, so every sign pattern on them extends to an affine sign function, and they are
shattered by $H_8^\pm$. A dependent set contains a circuit, and no tope restricts to that circuit on its support, because
topes are orthogonal to circuits. So the four columns are independent, giving the rank lower bound four.

\proofstep{The upper bound on the oriented-matroid rank.}
Take the following uniform rank-four chirotope on the ordered elements
$0,\ldots,7$, identified with their binary expansions in $\mathbb F_2^3$. The signs, indexed by the $70$ four-subsets in
lexicographic order, are
\begin{center}\small\ttfamily
+++-+-++-+-++-+--+--+-+-++-++-++-++-++-++-+-++-++-++-++----++-+++++-++
\end{center}
For each two-subset $\lambda$ and four further distinct elements
$a,b,c,d$, the three signs
\[
 \chi(\lambda ab)\chi(\lambda cd),\quad
 -\chi(\lambda ac)\chi(\lambda bd),\quad
 \chi(\lambda ad)\chi(\lambda bc)
\]
contain both signs, where $\chi$ is extended to ordered tuples as an alternating function. For a nowhere-zero
alternating map these three-term Grassmann--Pl\"ucker relations characterize uniform
chirotopes~\cite[Section 3.6]{BLSWZ99}. For each five-subset $S=(e_0,\ldots,e_4)$, its
circuit has signs $X_S(e_i)=(-1)^i\chi(S\setminus e_i)$, and a full-support sign vector is a tope exactly when it is
orthogonal to all these circuits, that is, when its restriction to no $S$ equals $\pm X_S$.
Direct enumeration of the $2^8$ sign vectors shows that the $128$ patterns orthogonal to every
such circuit contain all $16$ affine sign functions. This proves
$\operatorname{omr}(H_8)\le4$. The three-term relations and this enumeration are finite checks,
verified by exact computation (Appendix~\ref{sec:computation}).

\proofstep{Exclusion of real rank four.}
We exclude real rank four by a finite certificate, verified by exact
computation (Appendix~\ref{sec:computation}). Impose the displayed
three-term sign relations on all $70$ basis variables, together with
the circuit orthogonality conditions for the $16$ affine functions,
and normalize the first basis sign to positive. Exhaustive enumeration gives $13\,888$ assignments. The enumeration was carried
out by backtracking with unit propagation and, independently, by satisfiability solvers with
blocking clauses, with identical results.

Realizability is invariant under the following symmetries. Let $\pi$ be an affine permutation of $\mathbb F_2^3$ and $\beta$ an affine
sign function. The assignment $\chi'(\pi(b_1),\ldots,\pi(b_4))=\chi(b_1,\ldots,b_4)\prod_i\beta(b_i)$, up to the global sign
restoring the normalization, is realized by $v'_{\pi(x)}=\beta(x)v_x$ whenever $\chi$ is realized by $(v_x)$, since the global sign
is changed by a reflection of $\mathbb R^4$. These operations
map the enumerated assignments to themselves and
split them into seven
orbits of sizes
\[
 448,\ 1344,\ 1344,\ 2688,\ 2688,\ 2688,\ 2688.
\]
It therefore suffices to exclude one representative of each orbit.

For a real realization, write $x_B=\ln|\det B|$, where $B$ also denotes the matrix of the four
vectors with indices in $B$.
The three signed products above are the terms of a Grassmann--Pl\"ucker identity with sum zero, and
they are all nonzero, so exactly one of them has the sign opposite to the other two.
That term has magnitude equal to the sum of the magnitudes of the two others, so it is
strictly larger than either of the other terms.
Taking logarithms gives a strict linear inequality in the $x_B$ for each of the two other terms.
For every orbit representative, there are rational weights that are
positive, sum to one, and combine these strict inequalities to $0>0$.
The seven certificates have respectively $45,48,31,44,49,49,49$
nonzero weights. Their identities over the rationals and the closure of the
enumerated assignments under the group action are verified by exact computation
(Appendix~\ref{sec:computation}).
Hence no enumerated chirotope is realizable.

Now suppose that $H_8$ has a real sign representation of rank at most four. Pad it with zero coordinates if its rank is smaller,
and let the negatives of the row functionals represent the negated rows, so that all $16$ affine sign functions are sign vectors of functionals.
The representation can be perturbed,
without changing any strict matrix sign, to one whose $4\times4$ minors of the vectors $v_x$ are all nonzero,
since these minors are nonzero polynomials. After a reflection of $\mathbb R^4$ applied to vectors and functionals, if necessary, the first
minor is positive. Its chirotope satisfies the three-term relations, since a real Grassmann--Pl\"ucker
identity has nonzero terms summing to zero. For a five-subset $S$, Cramer's rule gives
$\sum_i(-1)^i\det(S\setminus e_i)v_{e_i}=0$. Pairing with a functional $w$ shows that the sign vector
of $\langle w,v_x\rangle$ is orthogonal to the circuit $X_S$. So the chirotope
would be among the enumerated assignments, a contradiction. Thus the sign rank of $H_8$ is at least five.

\proofstep{Sign-rank upper bounds.}
Index the rows and columns of $H_8$ by $\{0,\ldots,7\}$, so that its entries are $(-1)^{r\cdot j}$ in binary. The integer matrices
\[
 P=\begin{pmatrix}
0&2&5&3&0\\
1&2&-2&-4&-3\\
-4&3&0&2&-3\\
-2&-1&-3&-3&4\\
3&2&-4&3&1\\
-2&-1&-3&2&-4\\
-4&3&-1&1&4\\
1&6&0&-2&0
\end{pmatrix},\qquad
 Q=\begin{pmatrix}
-4&6&-2&1&1\\
-2&1&3&6&4\\
6&-2&2&2&-6\\
6&0&3&-2&6\\
-5&-3&6&-6&-2\\
0&3&6&5&-5\\
2&6&6&-3&3\\
-2&-6&1&4&3
\end{pmatrix}
\]
satisfy $\operatorname{sign}(PQ^{\mathsf T})=H_8$, and the product has no zero entry. This is a real sign representation of $H_8$ of rank at most five, so the sign rank is five.

Proposition~\ref{prop:om-contraction} gives the lower bound six for $H_{16}$. For the upper bound, index its rows and columns by $\{0,\ldots,15\}$ in the same way. The matrices
\begingroup\small
\[
 P'=\begin{pmatrix}
1&19&-15&-4&-1&11&-30\\
-3&33&1&-3&8&-21&0\\
8&7&10&-3&32&19&2\\
-3&-9&19&-29&-7&-12&-10\\
-13&3&17&7&-27&17&-7\\
-15&-31&-5&-3&-3&19&-3\\
6&3&-19&-33&4&4&-8\\
4&-11&0&5&-20&-28&-17\\
-9&3&-29&4&-19&0&18\\
-17&19&-25&-2&-14&-6&8\\
28&16&10&-3&-17&8&11\\
-14&-12&-8&17&11&-19&-21\\
16&-11&-18&-21&14&-10&13\\
14&-8&-15&31&2&13&1\\
-24&5&5&-6&13&27&6\\
18&-1&-15&1&-7&16&-27
\end{pmatrix},\quad
 Q'=\begin{pmatrix}
8&15&-27&-18&-5&19&-40\\
9&19&3&-18&13&40&26\\
-14&9&-40&14&-21&1&22\\
15&-23&-5&3&-34&-40&-29\\
-12&40&5&-4&12&-30&10\\
40&-11&-14&40&-6&19&1\\
-18&40&-26&-17&-10&-11&-7\\
-40&-35&-14&2&18&8&-15\\
-1&-9&40&-39&-2&-12&-28\\
14&14&-26&30&8&24&-40\\
-33&24&40&29&-23&26&-26\\
-16&11&32&28&-40&7&-11\\
11&2&24&0&40&15&-9\\
-40&-22&-14&8&33&-14&-17\\
23&11&-7&7&17&-40&-8\\
12&1&-2&-40&-11&5&-2
\end{pmatrix}
\]
\endgroup
satisfy $\operatorname{sign}(P'Q'^{\mathsf T})=H_{16}$, and the product has no zero entry. As $P'$ and $Q'$ have seven columns, this gives $\operatorname{signrk}(H_{16})\le7$. These products are checked by direct multiplication (Appendix~\ref{sec:computation}).
\end{proof}

By the theorem, every rank-four oriented matroid containing the signed rows of $H_8$ as
topes is nonrealizable. For larger orders the same comparison holds in a local form:
Forster's bound applies to every realizable part of an arrangement, so if the
oriented-matroid rank of a Hadamard matrix is much smaller than the square root of its
order, an arrangement attaining it has only small realizable restrictions.

\begin{proposition}\label{prop:om-hereditary}
If a rank-$r$ oriented matroid contains all signed rows of a Hadamard
matrix $H_n$ of order $n$ as topes, every realizable restriction of it has at most $r^2$ elements.
\end{proposition}
\begin{proof}
Let $M$ be such an oriented matroid. If $J$ is a realizable restriction, it is the oriented matroid of vectors $v_x$, $x\in J$, in a real space of dimension at most
$r$. The restriction of a tope to $J$ is a tope of $M|_J$, so it is the sign vector of a linear functional on these vectors.
These vectors and functionals give a real sign
representation of $H_n|_J$ of rank at most $r$.
The restricted columns remain orthogonal, so
$(H_n|_J)^{\mathsf T}(H_n|_J)=nI_J$ and $\|H_n|_J\|=\sqrt n$. Forster's spectral bound~\cite{Forster02},
$\operatorname{signrk}(A)\ge\sqrt{N'M'}/\|A\|$ for an $N'\times M'$ sign matrix $A$, gives
$\operatorname{signrk}(H_n|_J)\ge\sqrt{n|J|}/\sqrt n
=\sqrt{|J|}$. Thus $|J|\le r^2$.
\end{proof}

\section{Circle representations and complements of antipodal paths}\label{app:low-index}
The inequality $\operatorname{Ind}+1\le\operatorname{signrk}$ has no converse in general:
Theorem~\ref{res:5.6} gives unbounded sign rank already at index four.
This section gives two exact comparisons between the topology of a
sign complex and the representations used to construct learning rules,
at the two ends of the range of the index.
At index one, topology forces a rank-two real representation, because an odd map to
the circle imposes one threshold order on the columns. When the index is close to the
number $n$ of columns, the relevant object is the complement of the sign complex, which
retracts to a cubical complex on the missing cube vertices. There a path between
antipodal missing vertices characterizes index at most $n-3$, a geodesic path
characterizes the oriented-matroid rank bound $n-2$, and
the difference between the two conditions separates the index from both
the oriented-matroid rank and the list number.

\subsection{An odd circle map orders the columns}
We first describe all total classes of index at most one. A \emph{threshold} in an order of the columns is a row positive on a prefix and negative on the complementary
suffix, the prefix possibly empty or everything. A \emph{reverse threshold} is the negative of a threshold.
The rank-two threshold normal form appears in
\cite[Lemma 3.1]{BK15}, applied to the transpose.
The implication from an equivariant
circle map below connects that normal form to the sign complex.
\begin{theorem}\label{thm:index-one}
For a nonempty total class $C$, the following are equivalent:
\begin{enumerate}[label=(\roman*),leftmargin=2em]
\item $\operatorname{Ind}S(C)\le1$;
\item $\operatorname{signrk}(C)\le2$;
\item after column sign changes and permutation, every row is a
threshold or a reverse threshold in one order of the columns.
\end{enumerate}
In this case $\operatorname{lr}(C)\le2$. If the index is one,
the list number is exactly two.
\end{theorem}
Totality is essential: \cite[Section 6.2]{FHV26} gives partial matrices of
index at most one and sign rank $\Omega(\sqrt N)$.
\begin{proof}
\proofstep{Normalization for (i)$\Rightarrow$(iii).}
Multiply each column by its entry in a fixed row, so that the constant
positive row belongs to $C$. This relabels the literals of each column and is an equivariant isomorphism of sign complexes.
Next keep one column from each set of repeated columns.
Deletion of columns gives an equivariant subcomplex, so $\operatorname{Ind}$ does not increase and (i) still holds. The normal form (iii) for the reduced class
gives it for $C$, by placing repeated columns next to each other and undoing the column sign changes, which (iii) allows.
If one column remains, (iii) holds trivially. So let the columns be distinct, and write $\mathbf e(t)=e^{\sqrt{-1}\,t}$.

\proofstep{(i)$\Rightarrow$(iii): integers on cross edges.}
The order of the columns will be read off from the way an odd map winds along edges joining a positive vertex to a negative one.
Let $f:S(C)\to S^1$ be odd; such a map exists when the index is at most one, since $S^0\subset S^1$ equivariantly.
On the positive simplex, the simplex on all vertices $i^+$, lift $f$ through
$\mathbf e:\mathbb R\to S^1$ to a function $\theta$, and denote the lifted value
at $e_i=i^+$ by $\theta_i$. Since columns $i,j$ are distinct, some row
separates them. The cross edge from $e_i$ to $-e_j=j^-$ therefore belongs
to $S(C)$: it lies in the row simplex $\sigma_c^+$ if $c(i)=+1$, $c(j)=-1$ and in $\sigma_c^-$ if $c(i)=-1$, $c(j)=+1$.
Let $\alpha_{ij}:[0,1]\to\mathbb R$ be its lift starting at $\theta_i$. Then
$\mathbf e(\alpha_{ij}(1))=f(-e_j)=-\mathbf e(\theta_j)$, so $\alpha_{ij}(1)=
\theta_j+\pi n_{ij}$ for an odd integer $n_{ij}$.

Reversing the edge and applying the antipodal map turns this edge into the cross edge from $e_j$ to $-e_i$.
The function $t\mapsto\alpha_{ij}(1-t)-\pi n_{ij}$ starts at $\theta_j$, lifts $f$ along the new
edge because $f$ is odd and $n_{ij}$ is odd, and ends at $\theta_i-\pi n_{ij}$. This gives
$n_{ji}=-n_{ij}$.

\proofstep{(i)$\Rightarrow$(iii): one sign across a row cut.}
For a nonconstant row $c$, let $P$ and $Q$ be its positive and negative
coordinates. Lift $f$ on its row simplex $\sigma_c^+$ to $\psi_c$. On the positive face, $\psi_c$ and $\theta$ both lift $f$, and
this face is connected, so this
lift differs from the previously chosen lift by $2\pi a_c$ with $a_c\in\mathbb Z$. On the
negative face, $x\mapsto\theta(-x)+\pi$ also lifts $f$, because $f$ is odd; there $\psi_c$ differs from this antipodal lift by $2\pi b_c$,
$b_c\in\mathbb Z$. Comparing the lifts on each cross edge, from $\psi_c(e_i)=\theta_i+2\pi a_c$ to
$\psi_c(-e_j)=\theta_j+\pi+2\pi b_c$, gives
\begin{equation}\label{eq:row-cut-integer}
 n_{ij}=1+2(b_c-a_c)\qquad(i\in P,\ j\in Q).
\end{equation}

\proofstep{(i)$\Rightarrow$(iii): a transitive tournament.}
Define a tournament on the columns by $i\to j$ when $n_{ij}>0$; since $n_{ij}$ is odd and $n_{ji}=-n_{ij}$, exactly one of $i\to j$, $j\to i$
holds. By \eqref{eq:row-cut-integer}, all edges across a row cut have the same orientation.
There is no directed triangle. Indeed, distinctness of its three
columns supplies a row that cuts it into a singleton $x$ and a pair $\{y,z\}$. In a directed triangle
one of the edges at $x$ enters $x$ and the other leaves $x$; these two edges cross that cut and
have opposite orientations across it, contradicting \eqref{eq:row-cut-integer}.
A tournament without a directed triangle is transitive: if $i\to j\to h$ and not $i\to h$, then $h\to i$ gives a directed
triangle. So it is a total order. In this order, for every row, either all of $P$ precedes all of $Q$ or
all of $Q$ precedes all of $P$. So every row's positive set is a prefix or a suffix, and the constant rows have this form as well.
This proves (i)$\Rightarrow$(iii).

\proofstep{(iii)$\Rightarrow$(ii)$\Rightarrow$(i).}
Number the columns in the order of (iii). For a prefix ending at $k$, $0\le k\le n$, pair the column vector $(1,j)$ with
$(k+1/2,-1)$; the inner product $k+1/2-j$ is positive exactly for $j\le k$. Negating the row vector gives the reverse threshold.
Multiplying the vector of a column by the sign of its change in (iii) gives a representation of $C$. This proves (iii)$\Rightarrow$(ii).
A rank-two sign representation gives a nowhere-zero odd linear map
on the sign complex, as noted after Lemma~\ref{res:5.0}; normalization
proves (ii)$\Rightarrow$(i).
The list upper bound follows from the sign-rank bound
$\operatorname{lr}\le\operatorname{signrk}$.

\proofstep{Index one.}
If all rows are $\pm c$ for one row $c$, then $S(C)$ is a disjoint pair of simplices, exchanged by the involution, and has index zero.
So if the index is one, there are two rows $c,c'$ neither equal nor opposite. These two rows agree at some coordinate $a$ and
disagree at another coordinate $b$. Restricting to the coordinates $a,b$ and to the rows $c,c'$ gives an
anchored one-bit cube, up to negating the column $a$. Negating a column does not change the list number, and restriction does not increase it.
Theorem~\ref{res:5.1} gives list number at
least two, proving equality. Conversely, two such rows give index at least one: the simplices
$\sigma_{c'}^+$ and $\sigma_{c'}^-=-\sigma_{c'}^+$ both meet $\sigma_c^+$, at the literals of the coordinates $a$ and $b$ respectively. So a connected
component of $S(C)$ contains a point and its antipode, and no odd map to $S^0$ exists.
\end{proof}

The implication (i)$\Rightarrow$(iii) shows that an arbitrary continuous
odd map to the circle already forces the rank-two threshold normal form, and
the integers $n_{ij}$ explain why the topology imposes a common order on all rows.
Now~\cite[Question 7]{FHV26} asks whether the sign rank of a total matrix is bounded by a
function of its index. The answer is negative by~\cite[Theorem 1.2]{BHHLT26}, and
Theorem~\ref{res:5.6} gives unbounded sign rank already at index four, but by
Theorem~\ref{thm:index-one} the answer is positive at index one.

\subsection{Cyclic-interval classes}
Cyclic intervals give a family on which the criterion just proved can be compared with the
other parameters. Half-circle intervals are cut out of a regular polygon by lines through
its center, which is a rank-two representation in the sense of Theorem~\ref{thm:index-one}.
Intervals of other lengths are cut out by affine lines, which gives rank three. The next
proposition determines the list number, the index, and the face width of the whole family.
The three parameters change at different places: the list number two occurs with
index and face width equal to $(1,2)$, to $(2,2)$, and to $(1,3)$. In particular the inequality
$\operatorname{lr}\le u$ of Theorem~\ref{thm:anchor-width} is strict on half-circle intervals of length
$2\le k\le n-2$.

\begin{proposition}\label{prop:cyclic-intervals}
Let $C_{n,k}$ consist of the $n$ cyclic intervals of length $k$ on
$\mathbb Z/n\mathbb Z$, with $n\ge3$ and $1\le k<n$. Then
\[
 \operatorname{lr}(C_{n,k})=
 \begin{cases}
 2,& k\in\{1,n-1\}\text{ or }|2k-n|\le1,\\
 3,&2\le k\le n-2\text{ and }|2k-n|\ge2.
 \end{cases}
\]
Its index is one when $|2k-n|\le1$ and two otherwise.
Its face width is two for $k\in\{1,n-1\}$ and three otherwise.
\end{proposition}
\begin{proof}
Write $I_i=\{i,i+1,\ldots,i+k-1\}$ for the interval with starting point $i$; its row is positive on $I_i$ and negative
elsewhere.
Complementing all labels replaces $k$ by $n-k$, and it preserves the list number, the sign complex, and, through $x\mapsto-x$, the face width.
So assume $k\le n/2$.

\proofstep{Two lower bounds.}
The rows of $I_0$ and $I_1$ are distinct because $k<n$, and they are not opposite: for $k\ge2$ both contain the point $1$, and for $k=1$ the complement of a
singleton is not a singleton because $n\ge3$. By the last step of the proof of Theorem~\ref{thm:index-one},
$\operatorname{lr}(C_{n,k})\ge2$ and the index is at least one.

\proofstep{Intervals of length one.}
Let $k=1$. With the all-negative background the error sets are disjoint singletons, so
Theorem~\ref{res:5.2} with $s=0$ gives list number at most two.
For $n\ge4$, three columns $0,1,2$ are
signed-shattered: three rows supply the single-positive patterns,
and the row of the outside point $3$ supplies the all-negative pattern. Their
antipodes give the other four patterns, so the index is at least
two. Theorem~\ref{res:6.6} of Section~\ref{ssec:6}, with disjoint singleton error sets and $s=0$, gives the reverse inequality.
For $n=3$ the half-circle representation below gives index one.

\proofstep{The half-circle representation.}
Let $k=\lfloor n/2\rfloor$, so $|2k-n|\le1$. Take equally spaced points on a circle and unit vectors at the midpoints $m=i+(k-1)/2$ of
the intervals $I_i$:
\[
 v_j=\bigl(\cos(2\pi j/n),\sin(2\pi j/n)\bigr),\qquad
 w_i=\bigl(\cos(2\pi m/n),\sin(2\pi m/n)\bigr).
\]
Then $\langle w_i,v_j\rangle=\cos(2\pi(j-m)/n)$. For $j\in I_i$ the angle has absolute value at most $\pi(k-1)/n<\pi/2$, and for $j\notin I_i$ its distance
from every multiple of $2\pi$ is at least $\pi(k+1)/n>\pi/2$, using $k+1>n/2$. So
these vectors realize the interval rows by homogeneous halfspaces.
This gives sign rank at most two and hence list number at most two.
By Theorem~\ref{thm:index-one} the index is at most one, and hence exactly one by the first step.

\proofstep{Lengths far from the half-circle.}
Let $2\le k\le(n-2)/2$, and consider columns $0,1$
and the negative anchor $k+1$. The intervals with starting points
$0,1,1-k,k+2$ give the patterns $(+,+),(-,+),(+,-),(-,-)$ on $0,1$ and miss the anchor. Indeed
$I_0$ and $I_1$ lie in $\{0,\ldots,k\}$. The interval
$I_{1-k}=\{1-k,\ldots,0\}$ does not contain $1$ since $k<n$, and it contains $k+1$ only if $2k\ge n$. The interval $I_{k+2}=\{k+2,\ldots,2k+1\}$ contains $0$ or $1$ only if $2k+1\ge n$.
Both exceptions are excluded by $2k\le n-2$.
This is an anchored two-bit cube after negating the anchor column, so the list number is at least
three by Theorem~\ref{res:5.1}. The three columns are signed-shattered, since a pattern with anchor value $+$ is the negative of a pattern with anchor
value $-$, so the sign complex has coindex at least two.

Every cyclic interval is cut off from the regular polygon by an
affine line. Homogenizing gives a representation by the vectors $(1,p_j)\in\mathbb R^3$, where $p_j$ are the polygon vertices,
hence sign rank at most three and index
at most two, by the remark after Lemma~\ref{res:5.0}. The index is therefore two. The values of the index are now determined
for all $k\le n/2$, because $k=\lfloor n/2\rfloor$ is the case $|2k-n|\le1$.

\proofstep{A two-dimensional model.}
For an upper bound that also determines $u$, let $Z$ be the cone over a cycle with vertices $v_i$, $i\in\mathbb Z/n\mathbb Z$, adjacent in
cyclic order, with apex $o$. Label $v_i$ by the row of $I_i$, and label the apex
$o$ by the constant negative row. For a realizable literal set $T$ let $Z_T$ be the induced subcomplex on the vertices whose labels contain $T$.
We first show that these sections are nonempty and contractible.

A literal set prescribing only negative literals selects the cone over the permitted
cycle vertices, with the apex included even if there are no such vertices. So its section is a cone, and the section of the empty literal set is all of $Z$.

Suppose now that at least one positive literal is prescribed, so that the apex is excluded. Let the positive columns be $A\ne\varnothing$ and the negative columns $B$.
The starting points whose interval contains a fixed positive column $a$
form an arc of $k$ consecutive vertices. Two arcs of $\mathbb Z/n\mathbb Z$ can have a disconnected intersection only if they cover the whole circle
and overlap at both ends, which requires that their lengths sum to at least $n+2$.
Since $k\le n/2$, a nonempty
intersection of such arcs is a single arc, of length at most $k$. If negative columns are
also prescribed, the interval must lie in the component of their
complement containing all prescribed positive columns. This component has $g\ge k$ points, because the literal set is realizable, and the
permitted starting points in that component again form one arc, of length $g-k+1\le n-k$.
Its length and the length of the previous arc sum to at most $n$, so the set of permitted starting points is one arc, of length at most $k<n$.
Thus a realizable section prescribing a positive literal is a
nonempty path. Every section
needed for the extension is therefore nonempty and contractible.

Now define a map $f$ to $|Z|$ over a compatible triangulation of $\operatorname{conv}C_{n,k}$,
in increasing simplex dimension, mapping each simplex $\sigma$ into the
section $Z_{U(\sigma)}$ determined by its fixed coordinates, in the notation of the opening of Section~\ref{app:orientations}. The set $U(\sigma)$ is realizable. If $\tau$ is a face of $\sigma$ then
$U(\tau)\supseteq U(\sigma)$, so $Z_{U(\tau)}\subseteq Z_{U(\sigma)}$. Thus the boundary of $\sigma$ is already mapped into the contractible
set $Z_{U(\sigma)}$, and the map extends. This is the inductive extension in the proof of
Proposition~\ref{prop:carrier-extension}, with this cone complex as target.

Let $g:|Z|\to[-1,1]^n$ be the affine decoding
of the labels, the map affine on simplices and sending each vertex to its label. It preserves every extreme coordinate: if $x_j=s$, then
$(j,s)\in U(\sigma)$ for the supporting simplex $\sigma$ of $x$ by~\eqref{eq:fixed-at-interior}, so $f(x)$ lies in a simplex all of whose labels have value $s$ at $j$. This is a
model of dimension two in the sense of Section~\ref{app:width}, and it gives $u\le3$.

\proofstep{The lower bound on $u$ and the case $k=1$.}
If $2\le k\le n-2$, the adjacent columns $0,1$ are shattered, by the rows of $I_0$, $I_{1-k}$, $I_1$, $I_2$. Choose four rows $c^{st}$
realizing the four patterns $(s,t)$ and interpolate them with the bilinear
weights $(1+sy_1)(1+ty_2)/4$ of the square. The resulting point has coordinates $y_1$ and $y_2$ at the columns $0$ and $1$, so this gives a continuous section of the
two-coordinate projection that preserves its four sides. Suppose there were a face-separating model of dimension at most one. By
Lemma~\ref{lem:cover-model}, $\operatorname{conv}C_{n,k}$ would have a finite open cover of multiplicity at most two, no
member meeting two opposite coordinate faces. Pulling it back along the section would give such a cover of the square, no member meeting two
opposite sides. This contradicts the square covering theorem of Lebesgue. Hence $u\ge3$.

For singleton intervals, $k=1$, use the star with center $o$ labeled by the
all-negative row and with the rows as leaves. If a realizable literal set prescribes a positive column $a$, its section is the leaf $v_a$. Otherwise its section
contains $o$ and is a star. These literal sections are contractible, so the extension and affine decoding of the preceding step, with this star as target, give face width at most two. A model
of dimension zero has a finite target, so $f$ is constant on the connected set $\operatorname{conv}C_{n,1}$, which is impossible
since some coordinate takes both values on rows. So a nonconstant
coordinate requires face width at least two.

\proofstep{List values.}
Theorem~\ref{thm:anchor-width} gives $\operatorname{lr}\le u$. For $k=1$ this gives $\operatorname{lr}\le2$, and the first step
gives equality. For $k=\lfloor n/2\rfloor$ the half-circle representation and the first step give the value two. For $2\le k\le(n-2)/2$ the anchored
two-bit cube and $\operatorname{lr}\le u\le3$ give the value three.
\end{proof}

\subsection{Missing vertices and the corank-two distinction}
We now turn to the other end of the range. A nonempty proper antipodal class on $n$
columns always has oriented-matroid rank at most $n-1$, as we show in the proof of
Theorem~\ref{thm:complement-family}, and hence index at most $n-2$ by
Theorem~\ref{thm:om-bound}. So the first nontrivial
conditions are $\operatorname{omr}(C)\le n-2$ and $\operatorname{Ind}S(C)\le n-3$, and we
characterize both through the vertices missing from the class.
For an antipodal class $C=-C\subsetneq\{\pm1\}^n$, write
$C'=\{\pm1\}^n\setminus C$, and let $G'$ be the subgraph of the
cube induced by $C'$. An \emph{antipodal path} is a path in $G'$ from a vertex to its antipode. It is
\emph{geodesic} when its length is $n$.

Oriented-matroid duality exchanges a representation of $C$ with a tope set in $C'$.
To state this we let $\mathcal T$ denote the set of full-support covectors also for
oriented matroids with loops. It is empty for the rank-zero uniform matroid, consisting of loops.

\begin{lemma}\label{lem:uniform-duality}
For a uniform rank-$r$ oriented matroid $M$ on $n$ elements,
\[
 \{\pm1\}^n=\mathcal T(M)\mathbin{\dot\cup}\mathcal T(M^*).
\]
Consequently, for a nonempty antipodal class $C\subsetneq\{\pm1\}^n$, $\operatorname{omr}(C)\le r$ if and only if $C'$ contains
the tope set of a uniform oriented matroid of rank at least $n-r$.
\end{lemma}
\begin{proof}
\proofstep{The tope sets are disjoint.}
A covector and a vector are orthogonal. In particular a full-support
sign vector $X$ cannot be both a tope of $M$ and a tope of its dual: orthogonality of $X$ to itself would require
two coordinates with opposite products $X_eX_e$.

\proofstep{Counting topes.}
A uniform rank-$r$ arrangement has
$t(n,r)=2\sum_{i=0}^{r-1}\binom{n-1}{i}$ topes, by deletion--contraction
(or the region formula in~\cite{BLSWZ99}). Explicitly, let $e$ be an element and $1\le r<n$. A tope of $M\setminus e$ has one or two extensions to a
tope of $M$, and it has two exactly when it is a tope of $M/e$, by covector elimination at $e$ and by composing with a covector nonzero at $e$. As $M\setminus e$ and
$M/e$ are uniform of ranks $r$ and $r-1$, we get $t(n,r)=t(n-1,r)+t(n-1,r-1)$. With $t(n,0)=0$ and $t(n,n)=2^n$, the formula follows from Pascal's rule. The dual is uniform of
rank $n-r$, and substituting $i\mapsto n-1-i$ gives
\[
 t(n,n-r)=2\sum_{i=r}^{n-1}\binom{n-1}{i}.
\]
The two counts sum to $2^n$, and the tope sets are disjoint, proving the partition.

\proofstep{The criterion for $\operatorname{omr}$.}
Let $\operatorname{omr}(C)\le r$, with a representation of rank $s\le r$.
Lemma~\ref{lem:uniform-om} permits uniformizing any representation
without losing its topes. Its dual has rank $n-s\ge n-r$ and,
by the partition, its topes lie in $C'$. Conversely, the dual of
a rank-$s$ uniform oriented matroid with $s\ge n-r$ whose topes lie in $C'$ has rank
$n-s\le r$ and, by the partition, contains $C$ among its topes. It is loopless since $C\ne\varnothing$ forces $n-s\ge1$. As $C=-C$, it is admissible in the definition of $\operatorname{omr}(C)$.
\end{proof}

The next theorem shows that the two conditions differ only in the length of an antipodal path among the missing vertices.
Its second equivalence is a path form of the sign-vector duality criterion of \cite[Theorems 2 and 18]{Arav13}.

\begin{theorem}\label{thm:corank-paths}
For $n\ge5$ and a nonempty proper antipodal class $C$ on $n$ columns,
\begin{align*}
 \operatorname{Ind}S(C)\le n-3
 &\quad\Longleftrightarrow\quad G'\text{ contains an antipodal path},\\
 \operatorname{omr}(C)\le n-2
 &\quad\Longleftrightarrow\quad G'\text{ contains a geodesic antipodal path}.
\end{align*}
\end{theorem}
\begin{proof}
\proofstep{The oriented-matroid condition.}
A geodesic and its antipode together give the $2n$ topes of a
uniform rank-two oriented matroid: reorient its starting vertex to all-positive,
and order the coordinates by their unique flips along the geodesic. Explicitly, in this order take
$v_i=(\cos\vartheta_i,\sin\vartheta_i)\in\mathbb R^2$ with $\vartheta_i=\pi-i\pi/(n+1)$. These are $n$ pairwise independent vectors, and a functional rotating from the
angle $\pi/2$ to $-\pi/2$ changes the sign of $\langle w,v_i\rangle$ in the order $i=1,\ldots,n$. The $2n$ regions of the
$n$ lines are the geodesic and its antipode. Since $C'=-C'$, all these vertices lie in $C'$.
Lemma~\ref{lem:uniform-duality} then gives $\operatorname{omr}(C)\le n-2$.

Conversely, that lemma supplies in $C'$ a uniform tope set of rank
at least two. A uniform matroid of rank at least two is simple, so two adjacent topes differ in exactly one element,
and the tope graph is isometric in the cube
\cite[Section 4.2]{BLSWZ99}. So it contains a path of length $n$
between any tope and its antipode. This path is a geodesic antipodal path in $G'$.

In this corank-two setting,
$\operatorname{omr}(C)\le n-2$ is also equivalent to
$\operatorname{signrk}(C)\le n-2$: the preceding argument replaces any
representation of rank at most $n-2$ by the dual of a uniform rank-two
oriented matroid, and a rank-two oriented matroid is realizable, as is
its dual.

\proofstep{The complementary cubical complex.}
We turn to the index, which is controlled by the part of the ambient sphere missed by the sign complex.
Let $W(C')$ be the union of cube faces all of whose vertices lie in
$C'$. Polar duality identifies the complement of $S(C)$ in the
ambient sphere with a space retracting equivariantly onto $W(C')$.
We use the following description by barycentric subdivision. Let $\mathcal P$ be the poset of nonempty consistent literal sets.
A cross-polytope face
is such a set, and its dual cube face $Q_T$ fixes those
literals. Inclusion of literal sets is reverse inclusion of cube faces. Thus the order complex $\Delta(\mathcal P)$ is the barycentric subdivision of both boundary spheres, and it gives an
equivariant identification of them with one sphere $\Sigma$. A set $T$ belongs to $S(C)$ exactly when $Q_T$ has a vertex in $C$, because $C=-C$. So it is
absent from $S(C)$ exactly when that cube face
has no vertex in $C$, that is, when all vertices of $Q_T$ lie in $C'$.

Write $\mathcal P=\mathcal P_S\sqcup\mathcal P_W$ accordingly. Then $\mathcal P_S$ is a lower set and
$\mathcal P_W$ an upper set, $|\Delta(\mathcal P_S)|=|S(C)|$, and $|\Delta(\mathcal P_W)|=|W(C')|$. On $\Sigma$ let $\mathsf t(x)\in[0,1]$ be the sum of the
barycentric coordinates of $x$ on vertices of $\mathcal P_W$. It is continuous and invariant. Remove and renormalize the coordinates
of faces in $\mathcal P_S$ on the set $\{\mathsf t>0\}$. Interpolating before normalization gives an equivariant deformation
retraction of $\{\mathsf t>0\}$, the complement of $S(C)$, onto $W(C')$. Likewise $\{\mathsf t<1\}$ retracts equivariantly onto $S(C)$, by removing
the coordinates on $\mathcal P_W$. Write $\rho_S,\rho_W$ for the end maps of these retractions.

\proofstep{Invariant components.}
The one-skeleton of $W(C')$ is $G'$, so the components of $W(C')$ correspond to those of $G'$.
An invariant component of $W(C')$ is equivalent to an antipodal
path in its one-skeleton $G'$. Indeed an invariant component contains $p$ and $-p$ for each of its vertices $p$, and conversely a component
containing $p$ and $-p$ meets its image under the involution, which is a component. If there is no such component, the
components occur in disjoint antipodal pairs. Assigning opposite
signs to each pair gives an odd map $W(C')\to S^0$.

\proofstep{An index inequality for complementary subcomplexes.}
The two complementary invariant subcomplexes satisfy
\begin{equation}\label{eq:complementary-index}
 \operatorname{Ind}S(C)+\operatorname{Ind}W(C')\ge n-2.
\end{equation}
To obtain it, let $h_1:S(C)\to S^{\alpha}$ and $h_2:W(C')\to S^{\beta}$ be odd maps. The sets $\{\mathsf t<2/3\}$ and $\{\mathsf t>1/3\}$ are invariant neighborhoods
retracting to the two spaces. Let $\mu:\Sigma\to[0,1]$ be the invariant function of $\mathsf t$ that is $0$ for $\mathsf t\le1/3$, is $1$ for
$\mathsf t\ge2/3$, and is linear in between. The formula
\[
 x\mapsto\bigl(\sqrt{1-\mu(x)}\,h_1\rho_S(x),\ \sqrt{\mu(x)}\,h_2\rho_W(x)\bigr)\in S^{\alpha+\beta+1},
\]
with a term omitted where its coefficient vanishes, is a continuous odd map from $\Sigma\cong S^{n-1}$ to the join of
the two target spheres. Borsuk--Ulam requires that join to have
dimension at least $n-1$. Thus an index at most $n-3$ forces an
antipodal path, because with no invariant component $\beta=0$ is available, and then $\alpha\ge n-2$.

\proofstep{The converse: the exterior of an invariant circle.}
Conversely, suppose that $G'$ contains an antipodal path.
Choose a shortest antipodal path $\gamma=(p_0,\ldots,p_L)$ in $G'$, over all starting vertices, so $p_L=-p_0$. It meets its antipode
only at its endpoints: if $p_i=-p_{i'}$ with $i<i'$ and $(i,i')\ne(0,L)$, then $p_i,\ldots,p_{i'}$ would be a shorter
antipodal path. Hence $\Gamma=\gamma\cup(-\gamma)$ is an
embedded invariant polygonal circle in the cubical sphere $\Sigma$.
It lies in $W(C')$, so it is disjoint from the dual copy of $S(C)$.
We build an odd map to $S^{n-3}$ on the exterior of this circle.

The complex $\Delta(\mathcal P)$ is an invariant triangulation of $\Sigma$ with $\Gamma$ and $S(C)$ as subcomplexes. Let $N$ be the simplicial neighborhood of $\Gamma$ in
its second barycentric subdivision. This is an invariant regular neighborhood of $\Gamma$ disjoint
from that compact copy, and $N$ collapses to $\Gamma$. Let $E$ be its exterior, the closure of $\Sigma\setminus N$, so that $\partial E=\partial N=N\cap E$ by the PL regular neighborhood theorem.
The involution is free on $E$, and the quotient $\bar E=E/\mathbb Z_2$ is a compact $(n-1)$-manifold
with boundary $\partial\bar E=\partial E/\mathbb Z_2$.

Since $n-1\ge4$, general position makes $E$ connected and simply
connected: a path in $E$ joins two points of $\Sigma$ and a loop in $E$ bounds a disk in the sphere, and rel boundary the path or the disk may
be perturbed off the one-dimensional core $\Gamma$, because $2+1<n-1$. Then it may be pushed from $\Sigma\setminus\Gamma$ into $E$.
Thus $E\to\bar E$ is the universal cover, and $\pi_1(\bar E)=\mathbb Z_2$.
Its boundary is connected: in the reduced Mayer--Vietoris sequence for $\Sigma=N\cup E$,
\[
 H_1(\Sigma)\to\widetilde H_0(\partial E)\to\widetilde H_0(N)\oplus\widetilde H_0(E),
\]
the outer groups vanish. So the inverse image $\partial E$ of $\partial\bar E$ in the universal cover is connected, and a path in $\partial E$ from a point to its antipode projects
to a boundary loop whose lift exchanges sheets. Hence the map
$\pi_1(\partial\bar E)\to\pi_1(\bar E)=\mathbb Z_2$ is onto.

\proofstep{Vanishing with local coefficients.}
Let $\Lambda=\mathbb Z[\mathbb Z_2]$, and let $C_*=C_*(E,\partial E)$ be the simplicial chain complex of an invariant triangulation, a complex of free $\Lambda$-modules.
Since $E$ is connected, $\partial E$ is nonempty and connected, and $H_1(E)=0$, the exact sequence of the pair gives $H_0(C_*)=H_1(C_*)=0$.
Then $\partial_1:C_1\to C_0$ is onto a free module, so it splits and $\ker\partial_1$ is projective. Moreover $\partial_2:C_2\to\ker\partial_1$ is onto, so
it splits as well. These splittings persist after tensoring with a $\Lambda$-module $\mathcal A$, so
the pair $(\bar E,\partial\bar E)$ has
vanishing relative homology in degrees zero and one with every
local coefficient system. Poincar\'e--Lefschetz duality with local coefficients, $H^j(\bar E;\mathcal A)\cong H_{n-1-j}(\bar E,\partial\bar E;\mathcal A\otimes\mathcal O)$, with $\mathcal O$ the
orientation system, gives
\[
 H^j(\bar E;\mathcal A)=0\qquad(j\ge n-2)
\]
for every local coefficient system $\mathcal A$; compare~\cite[Section 3.H]{Hatcher02}.

\proofstep{An odd map from the exterior.}
Odd maps $E\to S^{n-3}$ are the sections of the associated
$S^{n-3}$ bundle $E\times_{\mathbb Z_2}S^{n-3}\to\bar E$. Its fiber is simply connected because $n\ge5$, so obstruction theory for sections
applies with the local coefficient systems $\pi_i(S^{n-3})$~\cite[Chapter 8]{Spanier66}. The bundle has no obstruction below degree
$n-2$, because its fiber is $(n-4)$-connected. All higher obstruction
groups vanish by the preceding step, with their actual monodromy
coefficients. A section therefore exists and gives an odd map
$E\to S^{n-3}$. Since $S(C)\subset E$, restricting it proves the required index upper bound.
\end{proof}

A missing set that is an induced cycle, with antipodal arcs slightly longer than
$n$, satisfies the first condition of Theorem~\ref{thm:corank-paths} and not the second.
For such a class all the parameters can be computed, which gives the following family.

\begin{theorem}[Explicit complement families]\label{thm:complement-family}
For every $n\ge7$, there is an antipodal class with
$2^n-2(n+2)$ rows satisfying
\[
 \operatorname{vc}(C)=n-2,\qquad
 \operatorname{Ind}S(C)=n-3,\qquad
 \operatorname{lr}(C)=\operatorname{omr}(C)=n-1.
\]
\end{theorem}
\begin{proof}
\proofstep{The cycle.}
Partition $n-1$ coordinates into ordered blocks $A,B,E$, each of
size at least two, and let $j$ be the coordinate outside these blocks. This is possible because $n-1\ge6$.
Starting at the all-negative vertex, flip coordinates in the sequence
\[
 j,\ A,\ j,\ B,\ j,\ E.
\]
The path has length $n+2$ and ends at the antipode, since $j$ flips three times and every other coordinate once. Include its
antipodal path to form $\Gamma$, and let $C$ be the cube vertices
outside $\Gamma$. The antipodal path flips the same sequence, so $\Gamma$ is the closed walk whose direction sequence repeats this
sequence twice.

This is an induced cycle. To check that there is no chord, consider
a cyclic segment of at most $n+2$ steps; any two positions on the closed walk of length $2(n+2)$ are joined by such a segment. Coordinates other than $j$
are all distinct on such a segment. If it contains $c$ occurrences
of $j$ and has length $\ell\ge1$, its endpoint Hamming distance is
$\ell-c+(c\bmod2)$. Two occurrences of $j$ are never adjacent, so $\ell>c$ unless $\ell=c=1$, and hence this distance is positive.
It equals one only for a single step,
or for three steps containing two occurrences of $j$. The latter
is excluded because consecutive occurrences are separated by a
block of at least two other coordinates. The two antipodal paths
therefore give precisely $2(n+2)$ distinct vertices and no additional
edges. So $C$ is an antipodal class with $2^n-2(n+2)$ rows, and $G'$ is this cycle. The antipode is the rotation by $n+2$ steps. So every antipodal path in $G'$
is one of two arcs, of length $n+2$, and there is no geodesic antipodal path.

\proofstep{Index and rank.}
The cubical complex $W(C')$ is exactly this circle: a square in $W(C')$ would be a $4$-cycle in the cycle $G'$ of length $2(n+2)$. The
involution on it is rotation by half the circle, so it maps equivariantly to $S^1$. Thus it has index at most
one. The complementary-index inequality~\eqref{eq:complementary-index} in the proof of Theorem~\ref{thm:corank-paths} gives $\operatorname{Ind}S(C)\ge n-3$, and
that theorem gives $\operatorname{Ind}S(C)\le n-3$, since $G'$ contains an antipodal path.
The same theorem gives $\operatorname{omr}(C)\ge n-1$.
Any proper antipodal class has an oriented-matroid representation
of rank at most $n-1$: choose one missing antipodal pair as the
topes of a rank-one uniform matroid and apply
Lemma~\ref{lem:uniform-duality}. Hence $\operatorname{omr}(C)=n-1$.

\proofstep{The list number.}
For the list lower bound apply Corollary~\ref{cor:facet-test} to
the literal $v=(j,+)$ and its facet $x_j=+1$, in the notation of Section~\ref{app:codimension}. Following the closed walk, the vertices with $x_j=+1$ form three arcs:
\begin{itemize}[leftmargin=1.6em]
\item the coordinates of $A$ vary, with $B$ and $E$ all negative;
\item the coordinates of $E$ vary, with $A$ and $B$ all positive;
\item the coordinates of $B$ vary, with $A$ all negative and $E$ all positive.
\end{itemize}
Since $\Gamma$ has no chords, these are the components of $G_v$.
Each varies only the coordinates of one of $A,B,E$, and therefore
fixes the coordinates in the other two blocks. Every component
is contained in a proper coordinate facet of $Q_v$. The corollary gives
$\operatorname{lr}(C)\ge n-1$, while Theorem~\ref{thm:om-bound}
gives $\operatorname{lr}(C)\le\operatorname{omr}(C)=n-1$.

\proofstep{The VC dimension.}
Finally, $\Gamma$ contains an edge of each direction. For a direction $i$, the common pattern of the two endpoints of such an edge on $[n]\setminus\{i\}$
has no extension in $C$, so deleting
any one coordinate misses a projected pattern. Thus no $(n-1)$-set is
shattered, and neither is $[n]$. A pattern on $[n]\setminus\{a,b\}$ determines a two-dimensional cube face, and it has an extension in
$C$ exactly when that face contains a vertex of $C$. The cycle $\Gamma$ contains no square, so each two-dimensional cube face
contains a vertex of $C$. Every $(n-2)$-set is consequently shattered,
and the VC dimension is $n-2$.
\end{proof}

\part{Analytic and combinatorial upper bounds}\label{part:analytic}
This part gives constructions of accurate covers that do not pass through the
compatible carriers of Part~\ref{part:upper}. They retain information that
Theorem~\ref{thm:main} does not provide: an explicit encoding of the realizable
distributions with controlled entropy, or a bound that reflects the structure of
the matrix. Section~\ref{ssec:3} encodes a distribution by a power map and transports
the resulting cover by exponential tilting. Section~\ref{app:two-sided} gives a
continuous two-sided Gibbs model with a smaller fixed-scale entropy.
Sections~\ref{ssec:6} and~\ref{app:static} label the closed sets of realizable
constraints by Boolean hypotheses and bound the list number by the number of labels
along chains, which is stronger than Theorem~\ref{thm:main} on structured matrices.
Section~\ref{app:families} computes these parameters exactly on further families.
Section~\ref{ssec:8} applies the closure-chain bounds, together with spectral and
shattering estimates, to the Paley matrices, the basic explicit pseudorandom family. Their index lies between the orders \(\log p\) and \(\log^2p\), its
growth is open, and that section describes what controls it.

\section{Power maps and entropy tilting}\label{ssec:3}

This section develops an analytic approach to list replicability, independent of the carrier construction of Part~\ref{part:upper}. A realizable distribution is encoded by its signed column profile, and a coordinatewise power map with exponent of order \(\ln N\) sends the profile to a Hilbert sphere. The image has an explicit finite net at a fixed scale, consisting of power images of empirical averages of signed columns, whose logarithmic size is at most \(s_N\ln(2M)\). A Gaussian projection of the net yields a closed cover of multiplicity at most \(1+\kappa(N,M)\) whose labels are accurate up to a fixed margin. The main problem is to amplify this fixed margin to an arbitrarily small error without increasing the multiplicity. We solve it by an exponential tilt of the distribution: the tilt is a support-preserving homeomorphism of \(\Delta_C\), so it transports the cover without the factor of order \(\ln M\) that boosting costs.

The resulting bound \(2^{31}\ln^2(2N)\ln(2M)\) of Theorem~\ref{res:3.8}, from~\cite{Zhang26}, was the first polylogarithmic upper bound on list replicability. A short self-contained proof of a bound of this order is given in~\cite{BHHLTpoly26}. Theorem~\ref{thm:main} removes the additional factor \(\ln(2N)\) by a different argument. The construction of this section provides information that Theorem~\ref{thm:main} does not: an explicit encoding with a fixed-scale entropy bound, which Section~\ref{app:two-sided} refines, and a transport of covers by an explicit homeomorphism, which applies to every cover satisfying the low-vote condition of Section~\ref{ssec:7} (Lemma~\ref{res:7.11}). A second transport of the same kind gives the identity \(\operatorname{lr}(C,\eta)=\operatorname{lr}(C)\) for every \(\eta\in(0,\min(1/2,1/M))\) (Theorem~\ref{res:3.9}).

Throughout this section, $p=p_N$ and $u=u_{p_N}$ denote the
power-map parameters of Section~\ref{ssec:2.5}. As there, $c_r(i)=B_{ri}$,
and an atom is a labeled point $(i,y)\in[M]\times\{\pm1\}$. For a distribution
$\mu$ on atoms and a function $f:[M]\to\mathbb R$ we write
\[
\mathbb E_\mu[yf]=\mathbb E_\mu[yf(i)]=\sum_{(i,y)}\mu_{(i,y)}\,yf(i).
\]
Since $(a_{(i,y)})_r=yB_{ri}$, the definition of $\Phi$ gives
$\mathbb E_\mu[yB_{ri}]=\Phi(\mu)_r$ for every row $r$. Consequently, for a unit
vector $w\in\mathbb R^N$,
\begin{equation}\label{eq:vote-expectation}
\mathbb E_\mu[yf_w]=\sum_r\Phi(\mu)_r\,w_r|w_r|.
\end{equation}
We write $\operatorname{sign}$ for the coordinatewise sign with
$\operatorname{sign}(0)=+1$. If $h=\operatorname{sign}f$ and $h(i)\ne y$, then
$yf(i)\le0$. A \emph{hypothesis} is a Boolean function on $[M]$.

\subsection{An index bound from power maps}\label{ssec:3.1}

A high coordinatewise power of a point \(q\in Q\) with \(\|q\|_\infty=1\) concentrates on the coordinates of largest absolute value. At such a coordinate, the \(p\)-th power is insensitive to an additive error of order \(1/p\), and an empirical average of order \(p^2\) signed columns has coordinate errors of this order in mean square. Thus a sample of order \(p^2\) signed columns determines the normalized power image of \(q\) up to a small Euclidean error (Lemma~\ref{res:3.0}). A Gaussian projection reduces the resulting finite net to a dimension proportional to the logarithm of its size (Lemma~\ref{res:3.0prime}), and a Lipschitz extension gives an odd map with small fibers on the whole image. This proves an index bound (Theorem~\ref{res:3.1}) and supplies the metric estimates for the covers of Sections~\ref{ssec:3.3} and~\ref{ssec:3.4}.

\begin{resultenv}{Lemma}{res:3.0}{Empirical power-map net}
Let \(N\ge2\), let \(v_1,\ldots,v_L\in[-1,1]^N\), and let \(Q\) be the convex hull of these points. Put
\[
p=\lceil16\ln(2N)\rceil,
\qquad
s=2^{16}p^2,
\]
and let \(Y_s\) be the set of averages of \(s\) of the points \(v_j\), repetitions allowed. For every \(q\in Q\) with \(\|q\|_\infty=1\) there is \(y\in Y_s\setminus\{0\}\) with
\[
\|u_p(q)-u_p(y)\|_2\le\frac1{16}.
\]
Moreover \(|Y_s|\le L^s\).
\end{resultenv}

\begin{proof}
We find \(y\) by the probabilistic method: we bound the expected squared distance between the power images of \(q\) and of a random average of \(s\) of the points \(v_j\). The point \(y\) depends on \(q\), whereas the set \(Y_s\) does not.

\proofstep{The random average.} Write \(q=\sum_j\theta_jv_j\) with \(\theta\) a probability vector, let \(V_1,\ldots,V_s\) be independent with \(\mathbb P(V_\nu=v_j)=\theta_j\), and put \(\overline V=s^{-1}\sum_\nu V_\nu\in Y_s\), so that \(\mathbb E\overline V=q\). The estimates below use the independence of the \(s\) draws and require no independence between coordinates.

\proofstep{Moment bounds for one coordinate.} Fix a coordinate \(r\), write \(x=q_r\), \(X_\nu=V_{\nu,r}-x\) and \(\delta=\overline V_r-x=s^{-1}\sum_\nu X_\nu\). The variables \(X_\nu\) are independent and centered, with values in the interval \([-1-x,1-x]\) of length \(2\). Hoeffding's lemma states that a centered random variable \(X\) with values in an interval of length \(\ell\) satisfies \(\mathbb Ee^{\tau X}\le e^{\tau^2\ell^2/8}\). Applied with \(\tau=t/s\) and \(\ell=2\), it gives the bound \(e^{t^2/(2s^2)}\) for each factor, so
\[
\mathbb Ee^{t\delta}=\prod_{\nu=1}^s\mathbb Ee^{(t/s)X_\nu}\le e^{t^2/(2s)}
\qquad(t\in\mathbb R).
\]
By the Chernoff bound with \(t=s/4\), \(\mathbb P(\delta>1/4)\le e^{-s/16}e^{s/32}=e^{-s/32}\). The same bound holds for \(-\delta\), so \(\mathbb P(|\delta|>1/4)\le2e^{-s/32}\).

We next bound the fourth moment of \(\delta\). Since $X_\nu=V_{\nu,r}-x$ is centered and $V_{\nu,r}\in[-1,1]$, we have $\mathbb EX_\nu^2=\operatorname{Var}V_{\nu,r}\le1$. Since also $|X_\nu|\le2$, we have $\mathbb EX_\nu^4\le4\mathbb EX_\nu^2\le4$. Expand the fourth power of the centered sum. A product in which some index occurs exactly once has expectation zero. There remain the \(s\) products \(X_\nu^4\) and the \(3s(s-1)\) ordered products of the form \(X_\nu^2X_{\nu'}^2\) with \(\nu\ne\nu'\). Hence
\[
\mathbb E\delta^4\le\frac{4s+3s(s-1)}{s^4}=\frac{3s^2+s}{s^4}\le\frac7{s^2}.
\]
For \(t\ge0\) we have \(e^{2t|\delta|}\le e^{2t\delta}+e^{-2t\delta}\), so by Cauchy--Schwarz and the bound on \(\mathbb Ee^{t\delta}\),
\begin{equation}\label{eq:delta-square-exp}
\mathbb E\bigl[\delta^2e^{t|\delta|}\bigr]
\le
(\mathbb E\delta^4)^{1/2}\bigl(\mathbb Ee^{2t\delta}+\mathbb Ee^{-2t\delta}\bigr)^{1/2}
\le\frac{\sqrt7}s\bigl(2e^{2t^2/s}\bigr)^{1/2}
\le\frac4se^{t^2/s}.
\end{equation}

\proofstep{Coordinates with \(|x|\ge1/2\).} These coordinates can carry a large part of the norm of \(\phi_p(q)\), so we bound the error relative to \(|x|^{2p}\). The function \(\phi_p\) is continuously differentiable with \(\phi_p'(\tau)=p|\tau|^{p-1}\), because \(p\ge2\). A point between \(x\) and \(x+\delta\) has absolute value at most \(|x|+|\delta|\). By the mean value theorem, and by \(1+\tau\le e^\tau\),
\[
|\phi_p(x+\delta)-\phi_p(x)|
\le p|\delta|(|x|+|\delta|)^{p-1}
\le p|\delta|\,|x|^{p-1}e^{(p-1)|\delta|/|x|}.
\]
Square this, take expectations, and use \eqref{eq:delta-square-exp} with \(t=2(p-1)/|x|\), for which \(0\le t\le4p\). With \(|x|^{2p-2}\le4|x|^{2p}\) this gives
\[
\mathbb E|\phi_p(x+\delta)-\phi_p(x)|^2
\le\frac{4p^2}s|x|^{2p-2}e^{t^2/s}
\le16\frac{p^2}se^{16p^2/s}|x|^{2p}
\le\frac{|x|^{2p}}{2048},
\]
because \(16p^2/s=2^{-12}\) and \(e^{2^{-12}}\le2\).

\proofstep{Coordinates with \(|x|<1/2\).} Both power values are exponentially small in \(p\) unless \(\delta\) is large, so an absolute bound suffices. On \(\{|\delta|\le1/4\}\) both \(|x|\) and \(|x+\delta|\) are below \(3/4\), so the squared difference is at most \(4(3/4)^{2p}\). On the complement it is at most \(4\), since both values lie in \([-1,1]\). Hence
\[
\mathbb E|\phi_p(x+\delta)-\phi_p(x)|^2
\le4(3/4)^{2p}+4\,\mathbb P(|\delta|>1/4)
\le4(3/4)^{2p}+8e^{-s/32}.
\]

\proofstep{Summation.} Put \(R=\sum_r|q_r|^{2p}=\|\phi_p(q)\|_2^2\). Then \(R\ge1\) because some \(|q_r|=1\). Since \(\ln(4/3)\ge1/4\) and \(p\ge16\ln(2N)\), and since \(N\ge2\),
\[
4N(3/4)^{2p}\le4Ne^{-p/2}\le4N(2N)^{-8}=2(2N)^{-7}\le2\cdot4^{-7}=\frac1{8192}.
\]
Also \(s/32=2^{11}p^2\ge2^{11}p\ge2^{15}\ln(2N)\ge9\ln(2N)\), so
\[
8Ne^{-s/32}\le8N(2N)^{-9}=4(2N)^{-8}\le4\cdot4^{-8}\le\frac1{8192}.
\]
Summing the two cases over the coordinates, with at most \(N\) coordinates in the second case and \(\sum_{|q_r|\ge1/2}|q_r|^{2p}\le R\) in the first,
\[
\mathbb E\|\phi_p(\overline V)-\phi_p(q)\|_2^2
\le\frac R{2048}+\frac1{4096}
\le\frac{3R}{4096}
\le\frac R{1024},
\]
where the second inequality uses \(R\ge1\).

\proofstep{Normalization.} Some outcome \(y=\overline V\in Y_s\) therefore satisfies \(\|\phi_p(y)-\phi_p(q)\|_2\le\sqrt R/32=\|\phi_p(q)\|_2/32\). In particular \(\phi_p(y)\ne0\), so \(y\ne0\). The elementary inequality
\[
\left\|\frac a{\|a\|}-\frac b{\|b\|}\right\|
\le2\frac{\|a-b\|}{\|a\|}
\qquad(a,b\ne0)
\]
follows from
\[
\frac a{\|a\|}-\frac b{\|b\|}
=\frac{a-b}{\|a\|}+\frac b{\|b\|}\cdot\frac{\|b\|-\|a\|}{\|a\|}
\]
and \(\bigl|\|b\|-\|a\|\bigr|\le\|a-b\|\). With \(a=\phi_p(q)\), \(b=\phi_p(y)\) it gives \(\|u_p(q)-u_p(y)\|_2\le1/16\). Finally \(|Y_s|\le L^s\) because an average is determined by the sequence of \(s\) chosen points.
\end{proof}

The logarithm of the size of a finite set determines the dimension of a Gaussian projection that preserves its distances. We record this form of the Johnson--Lindenstrauss lemma with explicit constants, together with the two one-sided tail bounds, because Section~\ref{ssec:7.2} also uses them at other values of the distortion.

\begin{resultenv}{Lemma}{res:3.0prime}{Gaussian distortion on a finite set}
Let \(T\) be a \(k\times D\) matrix with independent \(N(0,1/k)\) entries. For fixed \(y\ne0\) and every \(\eta>0\),
\[
\mathbb P\left(\frac{\|Ty\|_2^2}{\|y\|_2^2}\ge1+\eta\right)\le\exp\bigl(-\tfrac k2(\eta-\ln(1+\eta))\bigr),
\]
and for \(0<\eta<1\),
\[
\mathbb P\left(\frac{\|Ty\|_2^2}{\|y\|_2^2}\le1-\eta\right)\le\exp\bigl(-\tfrac k2(-\eta-\ln(1-\eta))\bigr);
\]
for \(0<\eta<1\) both bounds are at most \(e^{-k\eta^2/8}\), so that
\[
\mathbb P\left(\left|\frac{\|Ty\|_2^2}{\|y\|_2^2}-1\right|\ge\eta\right)\le2e^{-k\eta^2/8}.
\]
Consequently, if \(Z\subset\mathbb R^D\) is finite with \(|Z|=L\ge1\) and
\[
k=\lceil32(2\ln L+\ln16)\rceil,
\]
then with probability at least \(1/2\) every nonzero difference \(z-z'\) with \(z,z'\in Z\cup\{0\}\) satisfies
\[
\tfrac12\|z-z'\|_2^2\le\|Tz-Tz'\|_2^2\le\tfrac32\|z-z'\|_2^2.
\]
\end{resultenv}

\begin{proof}
The \(k\) coordinates of \(Ty\) are independent, and each is a centered Gaussian of variance \(\|y\|_2^2/k\). Thus the ratio \(\|Ty\|_2^2/\|y\|_2^2\) has the law of \(\chi_k^2/k\), where \(\chi_k^2\) is a sum of \(k\) independent squares of standard Gaussians, and \(\mathbb Ee^{tg^2}=(1-2t)^{-1/2}\) for a standard Gaussian \(g\) and \(t<1/2\). By Markov's inequality, for \(0<t<1/2\),
\[
\mathbb P(\chi_k^2\ge k(1+\eta))\le e^{-tk(1+\eta)}(1-2t)^{-k/2}.
\]
The optimal parameter \(t=\eta/(2(1+\eta))<1/2\) has \(1-2t=1/(1+\eta)\), and it gives, for every \(\eta>0\),
\[
\mathbb P(\chi_k^2/k\ge1+\eta)\le\exp\bigl(-\tfrac k2(\eta-\ln(1+\eta))\bigr).
\]
Applied to \(-\chi_k^2\), Markov's inequality gives, for \(t>0\) and \(0<\eta<1\),
\[
\mathbb P(\chi_k^2\le k(1-\eta))\le e^{tk(1-\eta)}(1+2t)^{-k/2}.
\]
The optimal parameter \(t=\eta/(2(1-\eta))\) has \(1+2t=1/(1-\eta)\), and it gives
\[
\mathbb P(\chi_k^2/k\le1-\eta)\le\exp\bigl(-\tfrac k2(-\eta-\ln(1-\eta))\bigr).
\]
For \(0<\eta<1\) both exponents are at most \(-k\eta^2/8\). For the first, the function \(\eta-\ln(1+\eta)-\eta^2/4\) vanishes at \(0\) and has derivative \(\eta(1-\eta)/(2(1+\eta))\ge0\) on \([0,1]\), so \(\eta-\ln(1+\eta)\ge\eta^2/4\). For the second, \(-\eta-\ln(1-\eta)=\sum_{j\ge2}\eta^j/j\ge\eta^2/2\).

Now take \(\eta=1/2\). Each nonzero difference \(y=z-z'\) violates the displayed two-sided inequality with probability at most \(2e^{-k/32}\), because \(Tz-Tz'=Ty\). There are at most \((L+1)^2\le4L^2\) ordered pairs in \(Z\cup\{0\}\), so the union bound gives failure probability at most
\[
8L^2e^{-k/32}\le8L^2e^{-2\ln L-\ln16}=\frac12
\]
by the choice of \(k\).
\end{proof}

We now combine the two lemmas; a similar combination of a net and the Johnson--Lindenstrauss lemma appears in~\cite{HHM23}. The theorem has three branches: the rank of \(B\), and the two dimensions that the construction produces for \(B\) and for its transpose.

\begin{resultenv}{Theorem}{res:3.1}{Polylogarithmic index bound}
For every \(B\in\{\pm1\}^{N\times M}\), \(N,M\ge2\), let
\[
p_N=\lceil16\ln(2N)\rceil,
\qquad
s_N=2^{16}p_N^2,
\]
\[
\kappa(N,M)
=
\left\lceil32\bigl(2s_N\ln(2M)+\ln16\bigr)\right\rceil.
\]
Then
\[
\operatorname{Ind}S(B)+1
\le
\min\{\operatorname{rank}B,\kappa(N,M),\kappa(M,N)\}
\]
\[
\le
2^{31}\ln(2N)\ln(2M)
\min\{\ln(2N),\ln(2M)\}.
\]
\end{resultenv}

\begin{proof}
The required odd map is a composite of three maps: the profile map from \(|S(B)|\) to \(X_B\), the power map from \(X_B\) to a sphere, and a Lipschitz map built from a Gaussian projection of a net. Steps 1--4 construct these maps, and Step 5 treats the transpose and the constants.

\proofstep{Step 1: the profile map.} Send vertex \(i^b\) to \(bq_i\) and extend affinely on each simplex. Under the identification \(|S(B)|=\Delta_{C_B^\pm}\) of Section~\ref{ssec:2.1}, this is the profile map \(\Phi\) of Section~\ref{ssec:2.5}. If a simplex belongs to signed row \(r\), say it is contained in \(\sigma_r^s\), then \((bq_i)_r=bB_{ri}=s\) for each of its vertices \(i^b\) by \eqref{eq:signed-row-simplex}, so the corresponding profile coordinate is constantly \(+1\) or constantly \(-1\). The image of the simplex also lies in \(Q\subseteq[-1,1]^N\). Thus the image lies in \(X_B\). The map is continuous, being affine on each closed simplex with values at the vertices fixed, and it is odd, because the involution sends \(i^b\) to \(i^{-b}\) and acts affinely on simplices.

Since \(X_B\) lies in the column span of \(B\) and avoids zero (every point of \(X_B\) has a coordinate \(\pm1\)), normalization in that span gives the rank bound: with \(\rho=\operatorname{rank}B\) and a linear isomorphism \(\iota\) of the column span onto \(\mathbb R^\rho\), the map \(x\mapsto\iota(x)/\|\iota(x)\|_2\) is continuous and odd on \(X_B\), and its composite with the profile map gives \(\operatorname{Ind}S(B)\le\rho-1\).

\proofstep{Step 2: the empirical net.} On \(X_B\) the map \(u_{p_N}\) is defined, because \(0\notin X_B\). It is continuous and odd, because \(\phi_{p_N}\) is continuous and odd coordinatewise and \(\|\phi_{p_N}(v)\|_2\ge1\) on \(X_B\). Write
\[
U=u_{p_N}(X_B).
\]
This set is symmetric, because \(Q\) and \(\partial[-1,1]^N\) are symmetric and \(u_{p_N}\) is odd.

Put \(s=s_N=2^{16}p_N^2\). Let \(Y_s\) be the set of averages of \(s\) signed columns \(\pm q_i\), repetitions allowed. It is symmetric, since the negative of an average of \(s\) signed columns is the average of their negatives. Apply Lemma~\ref{res:3.0} to the list of \(L=2M\) signed columns, whose convex hull is \(Q\), with the parameters \(p=p_N\) and \(s=s_N\) of that lemma. Its assumption \(N\ge2\) holds, and repeated points in the list do no harm. Every \(q\in X_B\) lies in \(Q\) and has \(\|q\|_\infty=1\), so every point of \(U\) lies within Euclidean distance \(1/16\) of the finite set
\[
Z=u_{p_N}(Y_s\setminus\{0\})\subset S^{N-1},
\]
which is symmetric because \(Y_s\setminus\{0\}\) is symmetric and \(u_{p_N}\) is odd, and satisfies
\[
|Z|\le|Y_s|\le(2M)^s.
\]
The set \(Z\) need not be contained in \(U\). It is nonempty, because \(X_B\ne\varnothing\) and each point of \(U\) has a point of \(Z\) within distance \(1/16\).

\proofstep{Step 3: a Gaussian map on the finite net.} Let \(L=|Z|\ge1\) and
\[
k=\left\lceil32(2\ln L+\ln16)\right\rceil.
\]
By Lemma~\ref{res:3.0prime}, with \(D=N\) and the set \(Z\), the good event has positive probability, so there is a linear map \(T:\mathbb R^N\to\mathbb R^k\) with
\[
\tfrac12\|z-z'\|_2^2\le\|Tz-Tz'\|_2^2\le\tfrac32\|z-z'\|_2^2
\qquad(z,z'\in Z\cup\{0\}).
\]
Hence the data \(z\mapsto Tz\) on \(Z\cup\{0\}\) are \(\sqrt{3/2}\)-Lipschitz and, taking \(z'=0\) and using \(\|z\|_2=1\), satisfy \(\|Tz\|_2\ge1/\sqrt2\) for \(z\in Z\).

\proofstep{Step 4: Lipschitz extension and odd symmetrization.} The map \(T\) controls distances only on the net, whereas we need a map on all of \(U\). By the Kirszbraun theorem the data \(z\mapsto Tz\) extend to \(\mathbb R^N\) with the same Lipschitz constant. For completeness we include the short direct argument for finite data, which Section~\ref{ssec:7.2} also uses with another Lipschitz constant. For finite data \(z_i\mapsto y_i\), \(z_i\in\mathbb R^N\), \(y_i\in\mathbb R^k\), with Lipschitz constant \(L_0\), extension to a new point \(x\in\mathbb R^N\) asks for a common point of the balls
\[
\overline B(y_i,L_0\|x-z_i\|).
\]
Consider
\[
F(y)
=
\max_iF_i(y),
\qquad
F_i(y)=\|y-y_i\|^2-L_0^2\|x-z_i\|^2.
\]
This function is continuous and coercive, because \(F\ge F_1\) and \(F_1(y)\to\infty\) as \(\|y\|\to\infty\), so it attains a minimum. Let \(y_*\) be a minimizer and \(I\) the set of active indices, those with \(F_i(y_*)=F(y_*)\).

We claim that \(y_*\) lies in the convex hull of the active centers: otherwise nearest-point separation supplies a direction decreasing all active terms. To prove this, suppose \(y_*\notin\operatorname{conv}\{y_i:i\in I\}\), let \(c\ne y_*\) be the nearest point of this hull to \(y_*\), and put \(d=c-y_*\). The nearest-point property gives \(\langle y_*-c,y_i-c\rangle\le0\), and therefore
\[
\langle y_*-y_i,d\rangle
=-\|d\|^2+\langle y_i-c,y_*-c\rangle
\le-\|d\|^2
\qquad(i\in I).
\]
Thus \(F_i(y_*+\tau d)=F_i(y_*)+2\tau\langle y_*-y_i,d\rangle+\tau^2\|d\|^2<F_i(y_*)\) for \(i\in I\) and \(0<\tau<2\). The inactive terms stay below \(F(y_*)\) for small \(\tau\) by continuity. So \(F(y_*+\tau d)<F(y_*)\) for small \(\tau>0\), a contradiction, which proves the claim.

By the claim there are convex weights $\alpha_i\ge0$, with $\sum_i\alpha_i=1$,
$y_*=\sum_i\alpha_i y_i$, and positive weights only at active terms.
Every such term equals $F(y_*)$. For points $w_i$ with barycenter
$\overline w=\sum_i\alpha_iw_i$ and for any point $w$, expanding squares
around $\overline w$ and using $\sum_i\alpha_i(w_i-\overline w)=0$ gives
\begin{equation}\label{eq:weighted-variance}
\sum_i\alpha_i\|w-w_i\|^2
=\|w-\overline w\|^2+\frac12\sum_{i,j}\alpha_i\alpha_j\|w_i-w_j\|^2.
\end{equation}
Using \eqref{eq:weighted-variance} first for the points $y_i$ with $w=y_*=\overline w$,
then the Lipschitz inequalities $\|y_i-y_j\|\le L_0\|z_i-z_j\|$, and finally
\eqref{eq:weighted-variance} for the points $z_i$ with $w=x$, we obtain
\begin{align*}
F(y_*)&=\sum_i\alpha_i\|y_i-y_*\|^2
          -L_0^2\sum_i\alpha_i\|x-z_i\|^2\\
&=\frac12\sum_{i,j}\alpha_i\alpha_j\|y_i-y_j\|^2
          -L_0^2\sum_i\alpha_i\|x-z_i\|^2\\
&\le L_0^2\left(\frac12\sum_{i,j}\alpha_i\alpha_j\|z_i-z_j\|^2
                  -\sum_i\alpha_i\|x-z_i\|^2\right)\\
&=-L_0^2\left\|x-\sum_i\alpha_i z_i\right\|^2\le0.
\end{align*}
Since $F_i(y_*)\le F(y_*)\le0$ for every $i$, every required ball contains $y_*$, and assigning the value $y_*$ to
$x$ keeps the Lipschitz constant $L_0$. Add the points of a countable
dense subset of $\mathbb R^N$ one at a time. Each finite extension preserves the same
Lipschitz constant. The resulting map on the dense set has a unique
continuous extension to the whole domain $\mathbb R^N$, because $\mathbb R^k$ is complete, and passing to limits in the Lipschitz inequalities shows that the extension has
that constant.

Apply this construction to the Gaussian data \(z\mapsto Tz\) on \(Z\cup\{0\}\), with \(L_0=\sqrt{3/2}\). If \(\widetilde G:\mathbb R^N\to\mathbb R^k\) is an extension, define
\[
G(w)=\frac{\widetilde G(w)-\widetilde G(-w)}2.
\]
It is odd, so \(G(0)=0\). It has the same Lipschitz bound, by the triangle inequality and \(\|(-w)-(-w')\|=\|w-w'\|\). It agrees with the Gaussian data on the symmetric net: for \(z\in Z\) also \(-z\in Z\), so \(\widetilde G(-z)=-Tz\) and \(G(z)=Tz\).

For \(w\in U\), choose a net point \(z\in Z\) within \(1/16\). Then
\[
\|G(w)\|
\ge
\|G(z)\|-\|G(w)-G(z)\|
\ge
\frac1{\sqrt2}
-
\frac{\sqrt{3/2}}{16}
>
0.
\]
Normalization gives a continuous odd map \(U\to S^{k-1}\).

With \(L\le(2M)^{s_N}\), the selected dimension is at most \(\kappa(N,M)\). Composing with the profile and power maps gives the corresponding index bound: the odd maps \(|S(B)|\to X_B\to U\to S^{k-1}\) show \(\operatorname{Ind}S(B)+1\le k\le\kappa(N,M)\).
From now on we append zero coordinates to regard $G$ as
a map into $\mathbb R^{\kappa(N,M)}$. This is an isometric embedding,
so the Lipschitz bound and the small-fiber estimate \eqref{eq:res-3.1} below are not affected.

\proofstep{Step 5: transposition and constants.} Theorem~\ref{res:2.1} gives \(\operatorname{Ind}S(B)=\operatorname{Ind}S(B^T)\). Applying Steps 1--4 to \(B^T\in\{\pm1\}^{M\times N}\), now using \(p_M,s_M\), we obtain \(\operatorname{Ind}S(B)+1\le\kappa(M,N)\). For $N,M\ge2$, we have $\ln(2N),\ln(2M)\ge\ln4>1$, so $p_N\le16\ln(2N)+1\le17\ln(2N)$. Also $32\cdot2s_N=2^{22}p_N^2$, and the ceiling adds at most $1$. Therefore
\[
\begin{aligned}
1+\kappa(N,M)
&\le 2+32\ln16+2^{22}17^2\ln^2(2N)\ln(2M)\\
&\le(2^{22}17^2+98)\ln^2(2N)\ln(2M)\\
&<2^{31}\ln^2(2N)\ln(2M).
\end{aligned}
\]
The second inequality uses $2+32\ln16<98$ and $\ln^2(2N)\ln(2M)>1$, and the third uses $2^{22}17^2+98=1212153954<2^{31}$.
Interchanging $N$ and $M$ gives the other bound. Taking their minimum
proves the claimed product, because
$\min\{a^2b,ab^2\}=ab\min\{a,b\}$ for $a=\ln(2N)$ and $b=\ln(2M)$.
\end{proof}

The branches \(\kappa(N,M)\) and \(\kappa(M,N)\) of Theorem~\ref{res:3.1} improve on the rank only asymptotically. Since \(p_N\), \(s_N\) and \(\ln(2M)\) are nondecreasing in \(N\) and \(M\), and since \(p_2=23\) and \(s_2=34{,}668{,}544\), one has \(\kappa(N,M)\ge\kappa(2,2)=3{,}075{,}891{,}741>2^{31}\). Hence for every matrix of rank below \(3\times10^9\) the minimum in Theorem~\ref{res:3.1} is attained by \(\operatorname{rank}B\). We have not tried to optimize the sample count \(s_N=2^{16}p_N^2\) or the constants \(2^{31}\) of Theorem~\ref{res:3.1} and \(2^{37}\) of Theorem~\ref{res:3.4}.

The proof also yields the metric estimates on which the covers of Sections~\ref{ssec:3.3} and~\ref{ssec:3.4} are built. The map \(G:\mathbb R^N\to\mathbb R^{\kappa(N,M)}\) is odd and \(\sqrt{3/2}\)-Lipschitz, with \(G(0)=0\), so
\begin{equation}\label{eq:G-cube}
\|G(w)\|_2\le\sqrt{3/2}<2,
\qquad\text{hence}\qquad
G(w)\in[-2,2]^{\kappa(N,M)}
\qquad(\|w\|_2=1).
\end{equation}
Before normalization, \(G\) has the small-fiber estimate, for \(w,w'\in U\),
\[
\|w-w'\|
\le
\frac{2(1+\sqrt3)}{16}
+
\sqrt2\,\|G(w)-G(w')\|.
\refstepcounter{equation}\tag{\theequation}\label{eq:res-3.1}
\]
To prove it, approximate \(w,w'\) by net points \(z,z'\in Z\) at distance at most \(1/16\). The lower bound of Step 3 gives
\[
\|z-z'\|\le\sqrt2\,\|Tz-Tz'\|=\sqrt2\,\|G(z)-G(z')\|,
\]
and the Lipschitz bound for \(G\), applied to the two approximation errors, gives
\[
\|G(z)-G(z')\|\le\|G(w)-G(w')\|+\frac{2\sqrt{3/2}}{16}.
\]
Since \(\sqrt2\cdot\sqrt{3/2}=\sqrt3\), we conclude
\[
\|w-w'\|\le\frac2{16}+\|z-z'\|
\le\frac2{16}+\frac{2\sqrt3}{16}+\sqrt2\,\|G(w)-G(w')\|.
\]

A finite free simplicial complex is an equivariant subcomplex of the sign complex of a total matrix, with one row for each pair of maximal faces and one column for each pair of vertices. The index bound therefore passes to all such complexes.

\begin{resultenv}{Corollary}{res:3.2}{Finite free simplicial complexes}
Every nonempty finite free \(\mathbb Z_2\)-simplicial complex \(K\), with \(n\) vertex pairs and \(f\) maximal-face pairs, \(n,f\ge2\), satisfies
\[
\operatorname{Ind}(K)+1
\le
2^{31}\ln(2f)\ln(2n)
\min\{\ln(2f),\ln(2n)\}.
\]
If \(n=1\) or \(f=1\), then \(\operatorname{Ind}=0\).
\end{resultenv}

\begin{proof}
By~\cite[Lemma 15]{FHV26}, the complex \(K\) is the sign complex of an \(f\times n\) partial matrix: label the vertex pairs \(i^\pm\), choose one maximal face \(F_j\) from each pair, and put \(A_{ji}=\pm1\) if \(i^\pm\in F_j\) and \(A_{ji}=*\) otherwise. This is well defined because freeness excludes antipodal pairs from simplices. The row simplices are \(\sigma_j^+=F_j\) and \(\sigma_j^-=-F_j\), and every simplex of \(K\) lies in a maximal face, so \(S(A)=K\). Complete that matrix without changing its dimensions. The original complex is an equivariant subcomplex of the completion (Section~\ref{ssec:2.3}), so an equivariant map from the completion to a sphere restricts to \(K\), and Theorem~\ref{res:3.1} applies with \((N,M)=(f,n)\), using \(f,n\ge2\).

If \(n=1\), freeness excludes \(\{1^+,1^-\}\) as a simplex, so \(K=S^0\) and \(\operatorname{Ind}=0\). If \(f=1\), the two maximal faces \(\sigma\) and \(-\sigma\) are disjoint, because a nonempty intersection would be an invariant simplex, whose barycenter would be a fixed point. Hence \(|K|\) maps equivariantly to \(S^0\) by sending \(\sigma\) to \(+1\) and \(-\sigma\) to \(-1\), and again \(\operatorname{Ind}=0\).
\end{proof}

This analytic construction, from~\cite{Zhang26}, gives the simplicial bound with one extra logarithmic factor; the sharper index estimate for strongly regular CW complexes in~\cite{FHMNV26v1} (see also~\cite{FHMNV26}) appeared first. Theorem~\ref{thm:main} removes that factor through a different construction, and Proposition~\ref{res:4.4} records the resulting product bound.

Theorem~\ref{res:7.1}(iv) gives, for the \(n\times n\) Hadamard matrices of that theorem, a lower bound of order \(\log^3n\) on the target dimension of the power-map construction of this subsection. The bound applies to the combination of a coordinate power map, a \(1/16\)-net of its whole image, and a linear map preserving the distances of that net. It does not apply to arbitrary continuous maps with small fibers.

\subsection{A finite closed cover with margin gives a learner}\label{ssec:3.2}

All covers of this section are converted into learning rules by the following lemma. It is the form of Proposition~\ref{prop:cover-rule} used here: it works with the \(\ell_1\) distance on atoms, it asks for loss at most \(\varepsilon/2\) on the cover members, which leaves a margin for the passage from the empirical to the true distribution, and it makes the list and the sample size explicit in terms of the domain size.

\begin{resultenv}{Lemma}{res:3.3}{From a closed cover with margin to a learner}
Let \(C\) be a finite class on a domain of size \(M\). Let \(\{G_h\}\) be a finite closed cover of \(\Delta_C\), indexed by hypotheses, with multiplicity at most \(L\), and suppose
\[
G_h\subseteq\{\mu:\operatorname{loss}_\mu(h)\le\varepsilon/2\}.
\]
Then there is an \((\varepsilon,L)\)-list-replicable learner.

More precisely, there is a constant \(d_*>0\) depending only on the cover such that, with
\[
\rho=\min\{\varepsilon/16,d_*/4\},
\]
the list of a distribution \(\mu\) may be taken to consist of the hypotheses whose cover members lie within \(2\rho\) of \(\mu\). Its members have loss at most
\[
5\varepsilon/8<\varepsilon.
\]
With \(\delta_*=\min(\delta,1/2)\), the sample size may be any
\[
n(\delta)\ge
\frac{2M^2}{\rho^2}
\ln\frac{4M}{\delta_*}.
\]
\end{resultenv}

\begin{proof}
We use the notation and the facts of Lemma~\ref{lem:distribution-facts}.

\proofstep{The separation.} Merge cover members carrying the same hypothesis, taking finite unions. The merged sets are closed and satisfy the same loss bound, and their multiplicity is still at most \(L\), because a point in \(m\) merged sets with distinct labels lies in at least \(m\) original members. Because the cover is finite and compact and no \(L+1\) distinct members intersect, sufficiently small common neighborhoods of any \(L+1\) members also have empty intersection. In quantitative form, Lemma~\ref{lem:distribution-facts}(e) with \(b=L\) gives a positive separation \(d_*\) such that for every \(\mu\in\Delta_C\) at most \(L\) hypotheses \(h\) satisfy \(\operatorname{dist}_1(\mu,G_h)<d_*\). Put
\[
\rho=\min\{\varepsilon/16,d_*/4\}.
\]

\proofstep{The rule.} Fix an ordering of the hypotheses. On a sample \(S\) with empirical distribution \(\widehat\mu=\widehat\mu_S\in\Delta_C\), output the first hypothesis whose closed cover member is within \(\rho\) of \(\widehat\mu\); such a hypothesis exists, because \(\widehat\mu\) lies in some member. For an empirical distribution outside the realizable space, and on the empty sample, output a fixed hypothesis. This rule is deterministic and does not depend on \(\delta\).

\proofstep{The list.} For a realizable population distribution \(\mu\), define its list to contain precisely the hypotheses whose cover members are within \(2\rho\) of \(\mu\):
\[
\mathcal L_\mu=\{h:\operatorname{dist}_1(\mu,G_h)\le2\rho\}.
\]
Since \(2\rho\le d_*/2<d_*\), the choice of \(d_*\) shows that the list contains at most \(L\) distinct hypotheses. It depends only on \(\mu\), \(\varepsilon\) and the cover.

Let \(S\) be a sample of positive probability under \(\mu\), and suppose \(\|\widehat\mu-\mu\|_1\le\rho\). By Lemma~\ref{lem:distribution-facts}(a), \(\widehat\mu\in\Delta_C\). Moreover some cover member containing \(\mu\) is within \(\rho\) of \(\widehat\mu\), so the learner has an eligible output. Every eligible output has its cover member within \(\rho\) of \(\widehat\mu\), hence within \(2\rho\) of \(\mu\), and therefore belongs to the designated list.

For a hypothesis \(h\) on that list, compare \(\mu\) with a distribution \(\nu\) in its cover member at distance at most \(2\rho\). Loss is Lipschitz under this distribution distance (Lemma~\ref{lem:distribution-facts}(c)), so
\[
\operatorname{loss}_\mu(h)
\le
\operatorname{loss}_\nu(h)+2\rho
\le
\varepsilon/2+2\rho
\le5\varepsilon/8.
\]

\proofstep{The sample size.} There are \(2M\) labeled atoms. Hoeffding's inequality for each atom at precision \(\rho/(2M)\), followed by a union bound over the \(2M\) atoms, gives by Lemma~\ref{lem:distribution-facts}(d)
\[
\mathbb P(\|\widehat\mu-\mu\|_1>\rho)\le4M\exp\!\left(-\frac{n\rho^2}{2M^2}\right)\le\delta_*
\]
at the displayed sample size, which is uniform in \(\mu\) and at least \(1\), because \(\ln(4M/\delta_*)>0\). Thus the learner outputs an element of the list with probability at least \(1-\delta_*\ge1-\delta\).
\end{proof}

\subsection{The boosting bound}\label{ssec:3.3}

The maps of Section~\ref{ssec:3.1} provide a closed cover of bounded multiplicity at a fixed margin: a vertex of a triangulation of the target cube carries the vote of a nearby point of \(U\), and this vote has correlation above \(1/2\) with the labels of every distribution mapped near that vertex. We give two ways to amplify this fixed margin to an arbitrarily small error. The first, in this subsection, is boosting. It runs multiplicative weights on the distribution, and it reads the labels from the resulting trajectory of \(T=\lceil8\ln(8M)\rceil\) points in the cube, so the multiplicity is that of a triangulation of a space of dimension \(kT\), where \(k=\kappa(N,M)\). The second, in Section~\ref{ssec:3.4}, is a single change of variables, which leaves the dimension at \(k\) and removes the boosting factor. The comparison shows that this factor comes from the dimension of the trajectory and not from the fixed-margin cover, whose estimates are common to both arguments.

The covers below are built on a standard triangulation of a cube, which we fix now. Let \(D\ge1\) and \(\ell>0\) with \(2/\ell\in\mathbb Z\), and put \(K=[-2,2]^D\). A \emph{Freudenthal simplex} of step \(\ell\) is a set \(\{a_0,\ldots,a_m\}\subset\ell\mathbb Z^D\cap K\), \(m\ge0\), for which
\begin{equation}\label{eq:freudenthal-chain}
a_j-a_{j-1}=\ell\,\mathbf1_{T_j}\qquad(1\le j\le m),
\end{equation}
with pairwise disjoint nonempty sets \(T_1,\ldots,T_m\subseteq[D]\), where \(\mathbf1_T\) is the indicator vector of \(T\). The \emph{Freudenthal subdivision} of \(K\) with step \(\ell\) is the family of the convex hulls of these sets.

\begin{lemma}[Freudenthal subdivision]\label{lem:freudenthal}
The Freudenthal subdivision of \(K=[-2,2]^D\) with step \(\ell\) is a finite simplicial complex of dimension at most \(D\) with union \(K\). Moreover:
\begin{enumerate}[label=(\alph*),leftmargin=2em]
\item any two points of a simplex have \(\ell_\infty\)-distance at most \(\ell\); consequently a point \(z\) of a simplex having \(a\) as a vertex satisfies \(\|z-a\|_\infty\le\ell\), and every closed vertex star has Euclidean diameter at most \(2\ell\sqrt D\);
\item if \(W:K\to\mathbb R^{M'}\) is affine on each simplex and all its values at vertices lie in \([-1,1]^{M'}\), then
\[
\|W(q)-W(q')\|_\infty\le\frac{2D}\ell\,\|q-q'\|_\infty
\qquad(q,q'\in K).
\]
\end{enumerate}
\end{lemma}

\begin{proof}
We first determine the coordinates of a point in a Freudenthal simplex, and deduce that the relative interiors of the simplices partition \(K\).

\proofstep{Coordinates in a simplex.} Consider a set as in \eqref{eq:freudenthal-chain}. The vectors \(\mathbf1_{T_j}\) are linearly independent, so its convex hull is a simplex of dimension \(m\le D\). A subset of the chain is again of the form \eqref{eq:freudenthal-chain}, so faces of Freudenthal simplices are Freudenthal simplices. A point \(q=\sum_j\lambda_ja_j\) with all \(\lambda_j>0\) has the form
\begin{equation}\label{eq:freudenthal-coordinates}
q=a_0+\ell\sum_{j=1}^mt_j\mathbf1_{T_j},
\qquad
t_j=\sum_{l\ge j}\lambda_l,
\qquad
1>t_1>\cdots>t_m>0.
\end{equation}
Thus \(a_0=\ell\lfloor q/\ell\rfloor\) coordinatewise, the numbers \(t_j\) are the distinct nonzero values of the coordinates of \((q-a_0)/\ell\), and the sets \(T_j\) are their level sets. So the simplex having \(q\) in its relative interior is unique.

\proofstep{The partition of the cube.} Conversely, let \(q\in K\) be arbitrary and define \(a_0\), \(t_j\) and \(T_j\) by these formulas, with \(a_j=a_0+\ell\mathbf1_{T_1\cup\cdots\cup T_j}\). In every coordinate \(-2\le(a_0)_i\le q_i\le2\), because \(q_i\ge-2\in\ell\mathbb Z\). In a coordinate \(i\in\bigcup_jT_j\), the number \(q_i\) is not in \(\ell\mathbb Z\) and \(2\in\ell\mathbb Z\), so \((a_0)_i+\ell\le2\). Hence all \(a_j\) lie in \(K\), and \(q\) is in the relative interior of \(\{a_0,\ldots,a_m\}\), with coefficients \(\lambda_0=1-t_1\), \(\lambda_j=t_j-t_{j+1}\) and \(t_{m+1}=0\). Thus the relative interiors of the Freudenthal simplices partition \(K\). A point common to two simplices lies in the relative interior of one common face, so the intersection of two simplices is the face spanned by their common vertices, and the family is a simplicial complex.

\proofstep{Part (a).} For \(i<j\), \eqref{eq:freudenthal-chain} and the disjointness of the sets \(T_l\) give \(a_j-a_i\in\ell\{0,1\}^D\). So vertices of a simplex are at \(\ell_\infty\)-distance at most \(\ell\), and convexity of the norm extends this to all points of the simplex. Every point of the closed star of \(a\) is then at Euclidean distance at most \(\ell\sqrt D\) from \(a\), which gives the diameter bound.

\proofstep{Part (b).} Let \(q,q'\) lie in a common simplex, with coordinates \(t_j,t_j'\) as in \eqref{eq:freudenthal-coordinates} (some of them may vanish or equal \(1\) if the points lie on the boundary of the simplex). Then
\[
W(q)=\sum_{j=0}^m\lambda_jW(a_j)=W(a_0)+\sum_{j=1}^mt_j\bigl(W(a_j)-W(a_{j-1})\bigr).
\]
Reading a coordinate \(i\in T_j\) in \eqref{eq:freudenthal-coordinates} gives \(|t_j-t_j'|=|q_i-q_i'|/\ell\le\|q-q'\|_\infty/\ell\). Since \(\|W(a_j)-W(a_{j-1})\|_\infty\le2\) and \(m\le D\), the claimed bound holds on each simplex. For arbitrary \(q,q'\), the segment \([q,q']\) lies in the convex set \(K\). It is a finite union of closed subsegments, each contained in one simplex, with consecutive endpoints \(q=q_0,q_1,\ldots,q_n=q'\) in order along the segment. Since \(\sum_i\|q_{i-1}-q_i\|_\infty=\|q-q'\|_\infty\), adding the bounds along the subsegments gives the global bound.
\end{proof}

We can now state the bound obtained by boosting.

\begin{resultenv}{Theorem}{res:3.4}{Boosting bound}
For every \(B\in\{\pm1\}^{N\times M}\), \(N,M\ge2\), and every \(\varepsilon\in(0,1/2)\),
\[
\operatorname{lr}(C_B,\varepsilon)
\le
1+\kappa(N,M)\lceil8\ln(8M)\rceil.
\]
Hence
\[
\operatorname{lr}(B)
=
\sup_\varepsilon\operatorname{lr}(B,\varepsilon)
\le
2^{37}\ln^2(2N)\ln^2(2M).
\]
\end{resultenv}

\begin{proof}
We use the set \(U=u(X_B)\) and the map \(G\) constructed in the proof of Theorem~\ref{res:3.1}, with the properties \eqref{eq:G-cube} and \eqref{eq:res-3.1}. The small-fiber estimate first gives a weak learner that depends continuously on the distribution. Multiplicative weights then boost it along a continuous trajectory, and a triangulation of the space of trajectories gives the cover.

\proofstep{The square-root loss.} We measure the quality of a vote by a quantity that bounds its loss and whose square root is Lipschitz in the vote parameter. For \(v\in[-1,1]^N\) and unit \(w\in\mathbb R^N\), put
\[
e_v(w)
=
\frac{1-\sum_rv_rw_r|w_r|}{2}
=
\|D_vJ(w)\|_2^2,
\]
where
\[
J(w)=(w_+,w_-),
\qquad
D_v=\operatorname{diag}\!\left(\sqrt{(1\mp v_r)/2}\right).
\]
The vectors \(w_\pm\) have coordinates \((w_\pm)_r=\max\{\pm w_r,0\}\), so \(J(w)\in\mathbb R^{2N}\), and \(D_v\) acts by the factor \(\sqrt{(1-v_r)/2}\) on \((w_+)_r\) and by \(\sqrt{(1+v_r)/2}\) on \((w_-)_r\). The second equality holds because \(w_r^2=(w_+)_r^2+(w_-)_r^2\), \(w_r|w_r|=(w_+)_r^2-(w_-)_r^2\) and \(\sum_rw_r^2=1\). The map \(J\) is nonexpansive: for reals \(\alpha,\beta\) of equal sign, \((\alpha_+-\beta_+)^2+(\alpha_--\beta_-)^2=(\alpha-\beta)^2\), and for \(\alpha\ge0>\beta\) this sum is \(\alpha^2+\beta^2\le(\alpha-\beta)^2\). The diagonal entries of \(D_v\) lie in \([0,1]\), so \(\|D_v\|\le1\). The triangle inequality therefore gives
\[
\sqrt{e_v(w)}
\le
\sqrt{e_v(w')}+\|w-w'\|_2.
\]
If \(\Phi(\mu)=v\), then \eqref{eq:vote-expectation} gives
\begin{equation}\label{eq:soft-vote-loss}
\mathbb E_\mu[yf_w]=1-2e_v(w).
\end{equation}
For \(v\in X_B\), choose a distribution \(\nu\) on atoms with \(\Phi(\nu)=v\). The entropy estimate \eqref{eq:res-3.3} of Lemma~\ref{res:3.7} in Section~\ref{ssec:3.4}, whose proof does not depend on the present theorem, together with \eqref{eq:soft-vote-loss} and \(p\ge16\ln(2N)\), gives
\[
e_v(u(v))\le\frac{\ln N}{4p}\le\frac1{64}.
\]
Finally, for \(h=\operatorname{sign}f_w\) we have \(1-yf_w(i)\ge0\) at every atom and \(1-yf_w(i)\ge1\) at an atom with \(h(i)\ne y\). Taking expectations and using \eqref{eq:soft-vote-loss},
\[
\operatorname{loss}_\mu(\operatorname{sign}f_w)\le2e_v(w).
\]

\proofstep{A weak advantage from the small-fiber estimate.} Let \(v\in X_B\), let \(w'\in U\), and let \(\mu\) be a distribution on atoms with \(\Phi(\mu)=v\). The small-fiber estimate \eqref{eq:res-3.1} implies
\[
\|G(w')-G(u(v))\|\le1/256
\quad\Longrightarrow\quad
\mathbb E_\mu[yf_{w'}]\ge\frac{4471}{8192}>\frac12.
\]
To see this, note that under this assumption \eqref{eq:res-3.1}, \(\sqrt3\le7/4\) and \(\sqrt2\le2\) give
\begin{equation}\label{eq:45-128}
\|w'-u(v)\|_2\le\frac{2(1+7/4)}{16}+\frac2{256}=\frac{11}{32}+\frac1{128}=\frac{45}{128}.
\end{equation}
The square-root inequality and the bound on \(e_v(u(v))\) then give \(\sqrt{e_v(w')}\le1/8+45/128=61/128\), and \eqref{eq:soft-vote-loss} gives \(\mathbb E_\mu[yf_{w'}]\ge1-2(61/128)^2=4471/8192\).

\proofstep{A continuous weak learner.} Set \(k=\kappa(N,M)\), and regard \(G\) as a map into \(\mathbb R^k\). By \eqref{eq:G-cube}, \(G(U)\subset K\), where
\[
K=[-2,2]^k.
\]
Triangulate \(K\) by the Freudenthal subdivision with step \(1/(512k)\) (Lemma~\ref{lem:freudenthal} with \(D=k\)). The step is admissible, because \(2\cdot512k\in\mathbb Z\). By Lemma~\ref{lem:freudenthal}(a) every closed vertex star has Euclidean diameter at most \(2\sqrt k/(512k)\le1/256\). At a vertex \(a\) whose star meets \(G(U)\), choose \(w_a\in U\) with \(G(w_a)\in\operatorname{St}(a)\), and use the vote $f_{w_a}$ there. Assign the zero vote to every other vertex, then extend affinely on each simplex of $K$. All vertex votes lie in $[-1,1]^M$, since $|f_w(i)|\le\sum_rw_r^2=1$. So we obtain a continuous map \(W:K\to[-1,1]^M\). Put
\[
F(\mu)=G(u(\Phi(\mu)))\in G(U)\subset K
\qquad(\mu\in\Delta_{C_B}),
\]
which is defined because \(\Phi(\Delta_{C_B})=Y\subset X_B\). Then
\[
\mathbb E_\mu[yW(F(\mu))]\ge1/2
\]
for every realizable \(\mu\).

To see this, let \(\sigma\) be the simplex having \(q=F(\mu)\) in its relative interior. Every vertex \(a\) of \(\sigma\) has \(q\in\sigma\subseteq\operatorname{St}(a)\). Thus the simplex carrying a point of $G(U)$ has only vertices whose stars meet $G(U)$, so the additional zero votes do not affect the weak-advantage estimate. Both \(G(w_a)\) and \(q=G(u(\Phi(\mu)))\) lie in \(\operatorname{St}(a)\), so they are at distance at most \(1/256\), and the weak-advantage estimate gives \(\mathbb E_\mu[yf_{w_a}]>1/2\). As \(W(q)\) is a convex combination of the votes \(f_{w_a}\) at the vertices of \(\sigma\), the claim follows.

The following interpolation estimate holds on all of $K$, by Lemma~\ref{lem:freudenthal}(b) with \(D=k\) and \(\ell=1/(512k)\):
\[
\|W(q)-W(q')\|_\infty
\le1024k^2\|q-q'\|_\infty.
\]

\proofstep{Flattening and multiplicative weights.} Fix \(\varepsilon\in(0,1/2)\). Multiplicative weights will produce a vote whose errors have small mass under its starting distribution. To transfer this to the target error at \(\mu\), we start from a flattened distribution, in which every atom of \(\mu\)-mass above \(\varepsilon/(2M)\) has mass above \(1/(2M)\). For \(\mu\in\Delta_{C_B}\), put
\[
\nu_0\propto\mu^\alpha,
\qquad
\alpha=\frac{\ln2}{\ln(2M/\varepsilon)},
\]
with the convention \(0^\alpha=0\). Then \(\alpha>0\) and \((\varepsilon/(2M))^\alpha=1/2\). The map \(\mu\mapsto\nu_0\) is continuous and preserves the support. Use
\[
T=\lceil8\ln(8M)\rceil
\]
multiplicative-weight rounds
\[
\nu_{t+1}
\propto
\nu_t e^{-yW(F(\nu_t))/2},
\qquad\text{that is,}\qquad
\nu_{t+1}(i,y)=\frac{\nu_t(i,y)e^{-yW(F(\nu_t))(i)/2}}{Z_t},
\]
where \(Z_t=\mathbb E_{\nu_t}e^{-yW(F(\nu_t))(i)/2}\). Support is preserved, so every intermediate distribution is realizable by Lemma~\ref{lem:distribution-facts}(a), and \(F(\nu_t)\) is defined. The weak advantage applies at every round: the variable \(X_t=yW(F(\nu_t))(i)\) has values in \([-1,1]\) and mean at least \(1/2\) under \(\nu_t\), so Hoeffding's lemma, applied to \(X_t-\mathbb E_{\nu_t}X_t\) with \(\tau=-1/2\) and interval length \(2\), gives
\[
Z_t\le e^{-\mathbb E_{\nu_t}X_t/2+1/8}\le e^{-1/8}.
\]
With \(q_t=F(\nu_t)\) and \(S=\sum_{t<T}W(q_t)\in\mathbb R^M\), iterating the update gives \(\nu_T(i,y)=\nu_0(i,y)e^{-yS(i)/2}/\prod_{t<T}Z_t\). Summing over the atoms, we obtain
\[
\mathbb E_{\nu_0}e^{-yS/2}=\prod_{t<T}Z_t\le e^{-T/8}\le\frac1{8M}.
\]

\proofstep{A cover from the trajectory.} The trajectory \(z(\mu)=(q_0,\ldots,q_{T-1})\in K^T\), \(q_t=F(\nu_t)\), depends continuously on \(\mu\), because \(F\) is continuous and each round is a continuous function of the previous distribution. For \(z\in K^T\) put \(S_z=\sum_{t<T}W(q_t)\in\mathbb R^M\). Subdivide \(K^T=[-2,2]^{kT}\) by the Freudenthal subdivision with coordinate step \(\ell'=1/(1024k^2T)\) (Lemma~\ref{lem:freudenthal} with \(D=kT\), which applies because \(2/\ell'\in\mathbb Z\)), so that \(\|S_z-S_a\|_\infty\le1\) whenever \(z\) lies in a simplex having \(a\) as a vertex: by Lemma~\ref{lem:freudenthal}(a), \(\|z-a\|_\infty\le\ell'\), and the interpolation estimate applied to each of the \(T\) terms gives \(\|S_z-S_a\|_\infty\le T\cdot1024k^2\ell'=1\).

Label the vertex \(a\) by \(h_a=\operatorname{sign}S_a\) with \(\operatorname{sign}0=+1\). Let \(D_a\) be the closed region of \(K^T\) where the barycentric coordinate of \(a\) is maximal (Lemma~\ref{res:2.0}), and put \(F_h=\bigcup_{a:h_a=h}z^{-1}(D_a)\).

Suppose \(z(\mu)\in D_a\) and \(h_a(i)\ne y\). Then \(yS_a(i)\le0\). Since \(D_a\subseteq\operatorname{St}(a)\), we get \(yS_{z(\mu)}(i)\le1\), and with \(S=S_{z(\mu)}\) also \(e^{-yS(i)/2}\ge e^{-1/2}\) at that atom. So the atoms erroneous for \(h_a\) contribute at least \(e^{-1/2}\operatorname{loss}_{\nu_0}(h_a)\) to \(\mathbb E_{\nu_0}e^{-yS/2}\). Comparing with the multiplicative-weight bound, and using \(e^{1/2}<4\),
\[
\operatorname{loss}_{\nu_0}(h_a)
\le
\frac{e^{1/2}}{8M}
<
\frac1{2M}.
\]
If \(\operatorname{loss}_\mu(h_a)>\varepsilon/2\), an erroneous atom \(\zeta\) has original mass \(\mu_\zeta>\varepsilon/(2M)\) by Lemma~\ref{lem:distribution-facts}(b), and flattening gives it \(\nu_0\)-mass \(>1/(2M)\), a contradiction: indeed \(\mu_\zeta^\alpha>(\varepsilon/(2M))^\alpha=1/2\), while the normalizing sum \(\sum_{\zeta'}\mu_{\zeta'}^\alpha\) is at most \(M\), because \(\mu\) has at most \(M\) atoms in its support (Lemma~\ref{lem:distribution-facts}(a)) and each term is at most \(1\). The atom \(\zeta\) is erroneous for \(h_a\), so this contradicts the bound on \(\operatorname{loss}_{\nu_0}(h_a)\).

The sets \(F_h\) are closed, as finite unions of preimages of closed sets, and they cover \(\Delta_{C_B}\), because the regions \(D_a\) cover \(K^T\). If \(\mu\) lies in \(F_h\) for \(m\) distinct hypotheses \(h\), then \(z(\mu)\) lies in \(D_a\) for \(m\) distinct vertices \(a\), so \(m\le kT+1\) by Lemma~\ref{res:2.0}. The resulting finite closed cover has multiplicity at most \(kT+1\), and each label has loss at most \(\varepsilon/2\) on its member. Lemma~\ref{res:3.3} completes the proof of the first inequality.

\proofstep{The constant.} Since \(\ln(2M)\ge\ln4>1\), we have \(T\le8\ln(2M)+8\ln4+1\le17\ln(2M)<2^5\ln(2M)\). Step 5 of the proof of Theorem~\ref{res:3.1} gives \(1+\kappa(N,M)<2^{31}\ln^2(2N)\ln(2M)\), so
\[
1+\kappa(N,M)T\le(1+\kappa(N,M))T<2^{36}\ln^2(2N)\ln^2(2M)<2^{37}\ln^2(2N)\ln^2(2M).
\]
This holds for each \(\varepsilon\in(0,1/2)\). A rule that is \((\varepsilon,L)\)-list replicable is \((\varepsilon',L)\)-list replicable for \(\varepsilon'\ge\varepsilon\), so \(\operatorname{lr}(B,\varepsilon)\) is nonincreasing in \(\varepsilon\), and the bound holds for the supremum over all positive errors.
\end{proof}

\subsection{Relative-entropy tilting}\label{ssec:3.4}

This subsection removes the entire factor \(\lceil8\ln(8M)\rceil\) of Theorem~\ref{res:3.4}. Boosting reweights a distribution \(T\) times, because each round gains only a constant factor. We instead use a single exponential reweighting with a large exponent, applied in reverse: given \(\mu\), we find the distribution \(\nu\) whose tilt is \(\mu\), and we apply the fixed-margin cover to \(\nu\). The label chosen at \(\nu\) errs only at atoms whose potential vote is below its \(\nu\)-average by a fixed gap, so the tilt reduces the mass of these atoms exponentially in the exponent (Lemmas~\ref{res:3.6} and~\ref{res:3.7}). The tilt is a homeomorphism of \(\Delta_C\) (Lemma~\ref{res:3.5}), so the transported cover has the same multiplicity as the fixed-margin cover.

\subsubsection*{The tilt homeomorphism}

The tilt uses the potential \(\Psi\) and the profile map \(\Phi\) of Section~\ref{ssec:2.5}. Its inverse is a proximal step of the convex function \(\Psi\circ\Phi\) for relative entropy, and this variational description is what gives invertibility.

\begin{resultenv}{Lemma}{res:3.5}{Tilt homeomorphism}
For \(\lambda>0\),
\[
\Theta_\lambda(\nu)_z
\propto
\nu_z
\exp\!\left(
\frac{\langle\nabla\Psi(\Phi(\nu)),a_z\rangle}{\lambda}
\right)
\]
is a homeomorphism of each \emph{support face}, the closed face of distributions supported in a given nonempty set of atoms. Its inverse is
\[
\Theta_\lambda^{-1}(\mu)
=
\mathop{\rm argmin}_{\nu}
\left\{
\Psi(\Phi(\nu))
+
\lambda\operatorname{KL}(\nu\|\mu)
\right\}.
\]
The map \(\Theta_\lambda\) preserves support and restricts to a homeomorphism of \(\Delta_C\) onto itself.
\end{resultenv}

\begin{proof}
Let \(\Omega\) be the set of atoms. In this proof \(\nu\) and \(\mu\) range over the distributions on \(\Omega\), and \(\Delta(S)\) denotes the support face of a nonempty set \(S\subseteq\Omega\). We use \(\operatorname{KL}(\nu\|\mu)=\sum_{z:\nu_z>0}\nu_z\ln(\nu_z/\mu_z)\), which is \(+\infty\) unless the support of \(\nu\) lies in that of \(\mu\). The normalizing sum in the definition of \(\Theta_\lambda(\nu)\) is positive and continuous in \(\nu\), because \(\nabla\Psi\) and \(\Phi\) are continuous. So \(\Theta_\lambda\) is a continuous self-map of the probability simplex on \(\Omega\). All exponential factors are positive, so \(\Theta_\lambda(\nu)\) has the same support as \(\nu\).

\proofstep{The minimizer.} Fix the support \(S\) of \(\mu\), and write \(\mathcal J_\mu(\nu)=\Psi(\Phi(\nu))+\lambda\operatorname{KL}(\nu\|\mu)\). Only points of \(\Delta(S)\) have a finite objective. On this closed probability simplex, the objective is continuous with the usual boundary convention \(0\ln0=0\) for relative entropy, and therefore has a minimizer. The function \(\Psi\circ\Phi\) is convex: \(\Phi\) is linear and \(\Psi\) is a sum of convex functions of the coordinates. Relative entropy is strictly convex on this face, since \(t\mapsto t\ln t\) is strictly convex and \(\operatorname{KL}(\nu\|\mu)\) differs from \(\sum_z\nu_z\ln\nu_z\) by a linear function of \(\nu\). Hence the minimizer \(\nu^*\) is unique.

\proofstep{Positivity.} The minimizer is positive at every atom in the support of \(\mu\). Suppose a supported coordinate \(z_0\) were zero, and choose \(z_1\) with \(\nu^*_{z_1}>0\). Moving mass \(t>0\) from \(z_1\) into \(z_0\) would create a contribution \(\lambda t\ln(t/\mu_{z_0})\) to the objective. All remaining terms are differentiable at \(t=0\), because \(\Psi\circ\Phi\) is continuously differentiable and \(\nu^*_{z_1}>0\), so they change by \(O(t)\). The one-sided derivative \(-\infty\) of the first contribution dominates this bounded first-order change. This would decrease the objective for small \(t\).

\proofstep{Stationarity.} At the interior minimizer, with \(\nu=\nu^*\), the derivatives of \(\mathcal J_\mu\) along the directions \(e_z-e_{z'}\), \(z,z'\in S\), vanish. Thus the Lagrange-multiplier equations are
\[
\langle\nabla\Psi(\Phi(\nu)),a_z\rangle
+
\lambda\bigl(\ln(\nu_z/\mu_z)+1\bigr)
=
\text{constant}
\qquad(z\in S).
\]
Rearranging gives that \(\mu_z\) is a constant multiple of \(\nu_z\exp(\langle\nabla\Psi(\Phi(\nu)),a_z\rangle/\lambda)\) on \(S\). Both vanish off \(S\). Normalizing gives exactly
\[
\mu=\Theta_\lambda(\nu).
\]

\proofstep{The inverse.} Conversely, suppose \(\nu\) has support \(S\) and \(\mu=\Theta_\lambda(\nu)\). Then \(\mu\) has support \(S\), and taking logarithms shows that these equations hold at \(\nu\), with some constant \(c\). For \(\nu'\in\Delta(S)\), the convex function \(g(\tau)=\mathcal J_\mu(\nu+\tau(\nu'-\nu))\), \(\tau\in[0,1]\), is differentiable at \(\tau=0\), with \(g'(0)=c\sum_z(\nu'_z-\nu_z)=0\), so \(g(1)\ge g(0)\). So these equations imply optimality by convexity, and uniqueness by strict convexity gives \(\nu=\nu^*\). Thus the formula defines the inverse: \(\Theta_\lambda\) is a bijection of the probability simplex on \(\Omega\) onto itself, and \(\Theta_\lambda^{-1}(\mu)\) is the minimizer.

\proofstep{Topology.} The forward map is continuous and preserves support, as noted above. Hence it maps each support face \(\Delta(S)\) bijectively onto itself: a preimage of a point of \(\Delta(S)\) has the same support. It is a continuous bijection on a compact simplex, hence a homeomorphism. The support characterization of realizability (Lemma~\ref{lem:distribution-facts}(a)) shows that both it and its inverse preserve \(\Delta_C\). So \(\Theta_\lambda\) restricts to a continuous bijection of the compact space \(\Delta_C\) onto itself, which is a homeomorphism.
\end{proof}

The hypothesis \(\lambda>0\) is needed for support preservation: at \(\lambda=0\) the limiting map can lose support, since in the limit all mass concentrates on the atoms maximizing the gradient pairing.

\subsubsection*{Loss transfer}

The next lemma gives the effect of the tilt on the loss of a hypothesis \(h\). Its hypotheses say that \(h\) errs only at atoms whose potential vote at \(\nu\) is smaller, by a fixed gap, than the \(\nu\)-average of the vote. Section~\ref{ssec:7} studies this low-vote condition in its own right.

\begin{resultenv}{Lemma}{res:3.6}{Loss transfer}
Let \(\nu\in\Delta_C\) and \(D=Z(\Phi(\nu))=\|\nabla\Psi(\Phi(\nu))\|_1\), as in Section~\ref{ssec:2.5}. Suppose
\[
yf_{u(\Phi(\nu))}(i)\le b
\]
at every erroneous atom of \(h\), and
\[
\mathbb E_\nu[yf_{u(\Phi(\nu))}]\ge a>b.
\]
If \(D\ge D_0\), then
\[
\operatorname{loss}_{\Theta_\lambda(\nu)}(h)
\le
e^{-D_0(a-b)/\lambda}.
\]
\end{resultenv}

\begin{proof}
With \(v=\Phi(\nu)\), the definitions of Section~\ref{ssec:2.5} give \(\nabla\Psi(v)_r=A_r(v)\) and \(u(v)_r|u(v)_r|=A_r(v)/Z(v)\) by \eqref{eq:squared-power}, so the atom-gradient identity is
\[
\langle\nabla\Psi(\Phi(\nu)),a_{(i,y)}\rangle
=\sum_rA_r(v)\,yB_{ri}
=
D\,y f_{u(\Phi(\nu))}(i).
\]
Write \(g(i,y)=yf_{u(\Phi(\nu))}(i)\). Then \(\Theta_\lambda(\nu)\) assigns to the erroneous atoms of \(h\) the total mass
\[
\operatorname{loss}_{\Theta_\lambda(\nu)}(h)
=
\frac{\sum_{z\ \mathrm{erroneous}}\nu_ze^{Dg(z)/\lambda}}{\sum_z\nu_ze^{Dg(z)/\lambda}}.
\]
Since \(\nu\) is realizable, \(\Phi(\nu)\in X_B\), so \(D\ge1>0\) by \eqref{eq:res-3.2}. The numerator of the tilted error probability is at most \(e^{Db/\lambda}\), because the erroneous atoms have total \(\nu\)-mass at most \(1\) and \(g\le b\) on them. Only erroneous atoms of positive \(\nu\)-mass enter, so the assumption on \(b\) is used only at those atoms. Jensen's inequality, with \(D>0\), gives the denominator lower bound
\[
\mathbb E_\nu
e^{D y f_{u(\Phi(\nu))}/\lambda}
\ge
e^{D\mathbb E_\nu[yf_{u(\Phi(\nu))}]/\lambda}
\ge
e^{Da/\lambda}.
\]
Their ratio is at most \(e^{-D(a-b)/\lambda}\), which is at most the displayed expression, because \(D\ge D_0\) and \(a-b>0\).
\end{proof}

\subsubsection*{The potential estimates}

Lemma~\ref{res:3.6} needs lower bounds on the normalization \(D\) and on the average vote. For the power potential, both follow from the fact that a point of \(X_B\) has a coordinate of absolute value one. The bound on the average vote is an entropy estimate: the weights \(\pi_r\) below form a Gibbs distribution with energies \(-\ln|v_r|\) at inverse temperature \(2p\), so the bound \(\ln N\) on its entropy forces the average energy to be at most \(\ln N/(2p)\). We also record the Lipschitz dependence of the vote on its parameter.

\begin{resultenv}{Lemma}{res:3.7}{Potential and vote estimates}
With \(p=\lceil16\ln(2N)\rceil\),
\[
\nabla\Psi(v)_r
=
\operatorname{sgn}(v_r)|v_r|^{2p},
\]
\[
D(v)=Z(v)=\|\nabla\Psi(v)\|_1
=
\sum_r|v_r|^{2p}\ge1
\qquad(v\in X_B),
\refstepcounter{equation}\tag{\theequation}\label{eq:res-3.2}
\]
for \(v=\Phi(\nu)\in X_B\) with \(a_r=|v_r|\) and \(\pi_r=a_r^{2p}/Z(v)\),
\[
\mathbb E_\nu\bigl[yf_{u(v)}(i)\bigr]
=\sum_r\pi_ra_r
\ge
1-\frac{\ln N}{2p}
\ge\frac{31}{32},
\refstepcounter{equation}\tag{\theequation}\label{eq:res-3.3}
\]
and
\[
\|f_w-f_{w'}\|_\infty
\le2\|w-w'\|_2.
\refstepcounter{equation}\tag{\theequation}\label{eq:res-3.4}
\]
\end{resultenv}

\begin{proof}
The derivative formula follows directly from the definition of \(\Psi\) (Section~\ref{ssec:2.5}). A coordinate of \(v\in X_B\) has absolute value \(1\), so \(D(v)\ge1\).

For the identity in \eqref{eq:res-3.3}, \(\mathbb E_\nu[yB_{ri}]=\Phi(\nu)_r=v_r\) for every row \(r\), so \eqref{eq:vote-expectation} and \eqref{eq:squared-power} give
\[
\mathbb E_\nu\bigl[yf_{u(v)}(i)\bigr]
=\sum_rv_r\,u(v)_r|u(v)_r|
=\sum_r\frac{|v_r|^{2p+1}}{Z(v)}
=\sum_r\pi_ra_r.
\]
For the entropy identity, consider the rows with \(\pi_r>0\). The others have \(\pi_r=a_r=0\) and contribute nothing, with the convention \(0\ln0=0\). From \(\ln\pi_r=2p\ln a_r-\ln Z(v)\), multiplication by \(\pi_r\) and summation give
\[
\sum_r\pi_r(-\ln a_r)=\frac{H(\pi)-\ln Z(v)}{2p},
\qquad
H(\pi)=-\sum_r\pi_r\ln\pi_r.
\]
With \(H(\pi)\le\ln N\), by concavity of the logarithm, together with \(\ln Z(v)\ge0\) from \eqref{eq:res-3.2}, this gives the estimate
\[
\sum_r\pi_r(-\ln a_r)\le\frac{\ln N}{2p}.
\]
Using \(1-a_r\le-\ln a_r\) gives \(1-\sum_r\pi_ra_r\le\ln N/(2p)\), the first inequality in \eqref{eq:res-3.3}. The choice of \(p\) gives \(\ln N/(2p)\le\ln N/(32\ln(2N))\le1/32\), the second.

For \eqref{eq:res-3.4}, let \(w,w'\) be unit vectors. The scalar inequality \(\bigl|\alpha|\alpha|-\beta|\beta|\bigr|\le|\alpha-\beta|(|\alpha|+|\beta|)\) holds with equality when \(\alpha,\beta\) have equal sign; for opposite signs its left side is \(\alpha^2+\beta^2\le(|\alpha|+|\beta|)^2\) and \(|\alpha-\beta|=|\alpha|+|\beta|\). Apply Cauchy--Schwarz to the coordinate differences of \(w_r|w_r|\) and \(w'_r|w'_r|\), using that both vectors are unit vectors:
\[
\sum_r\bigl|w_r|w_r|-w'_r|w'_r|\bigr|
\le\sum_r|w_r-w'_r|\,(|w_r|+|w'_r|)
\le\|w-w'\|_2\bigl(\|w\|_2+\|w'\|_2\bigr)
=2\|w-w'\|_2.
\]
Since \(|c_r(i)|=1\), this sum of absolute differences bounds the difference \(|f_w(i)-f_{w'}(i)|\) at every column.
\end{proof}

\subsubsection*{The cover and the final constant}

We now apply the tilt to the fixed-margin cover given by the small-fiber estimate.

\begin{resultenv}{Theorem}{res:3.8}{General list-replicability bound}
For every \(B\in\{\pm1\}^{N\times M}\), \(N,M\ge2\), with concept class \(C_B\), and every \(\varepsilon\in(0,1/2)\),
\[
\operatorname{lr}(C_B,\varepsilon)\le1+\kappa(N,M).
\]
Consequently,
\[
\operatorname{lr}(B)
=
\sup_{\eta>0}\operatorname{lr}(B,\eta)
\le
2^{31}\ln^2(2N)\ln(2M).
\]
\end{resultenv}

\begin{proof}
Put \(k=\kappa(N,M)\). Use the unnormalized map \(G\) constructed in the proof of Theorem~\ref{res:3.1} and its small-fiber estimate \eqref{eq:res-3.1}. Since \(G(0)=0\) and \(G\) is \(\sqrt{3/2}\)-Lipschitz, its image on the unit sphere is contained in
\[
K=[-2,2]^k,
\]
by \eqref{eq:G-cube}; in particular \(G(U)\subset K\).

\proofstep{The fixed-margin cover.} Triangulate \(K\) by the Freudenthal subdivision (Lemma~\ref{lem:freudenthal} with \(D=k\)) with step
\[
\ell=\frac1{512k},
\qquad\text{so that }2/\ell\in\mathbb Z.
\]
Let \(D_a\) be the region on which the barycentric coordinate of vertex \(a\) is maximal (Lemma~\ref{res:2.0}). These regions are closed, cover \(K\), and have multiplicity at most \(k+1\). Call a vertex \(a\) \emph{relevant} if \(\operatorname{St}(a)\) meets \(G(U)\). The closed stars of the relevant vertices have diameter at most \(2\ell\sqrt k\le1/256\) by Lemma~\ref{lem:freudenthal}(a), and \(D_a\subseteq\operatorname{St}(a)\).

For each relevant vertex, choose \(w_a\in U\) with
\[
G(w_a)\in\operatorname{St}(a),
\]
and let
\[
h_a=\operatorname{sign}f_{w_a},
\qquad
\operatorname{sign}(0)=+1.
\]
Define, for \(\nu\in\Delta_{C_B}\),
\[
F(\nu)=G(u(\Phi(\nu))),
\qquad
R_a=F^{-1}(D_a).
\]
Since \(F(\nu)\in G(U)\), a vertex with \(R_a\ne\varnothing\) is relevant, and we discard the other vertices. If \(\nu\in R_a\), both \(F(\nu)\) and \(G(w_a)\) lie in \(\operatorname{St}(a)\), so \eqref{eq:res-3.1} and the star-diameter bound give, by the computation \eqref{eq:45-128},
\[
\|w_a-u(\Phi(\nu))\|_2\le\frac{45}{128}.
\]
At an atom where \(h_a(i)\ne y\), the signed vote \(yf_{w_a}(i)\) is nonpositive. Hence \eqref{eq:res-3.4} gives
\[
y f_{u(\Phi(\nu))}(i)\le yf_{w_a}(i)+2\cdot\frac{45}{128}\le\frac{45}{64}.
\]

\proofstep{The tilt.} Now set
\[
\lambda_\varepsilon
=
\frac{17}{64\ln(4/\varepsilon)},
\qquad
A_a=\Theta_{\lambda_\varepsilon}(R_a),
\]
where \(\lambda_\varepsilon>0\) because \(\varepsilon<4\). Apply Lemma~\ref{res:3.6} to \(\nu\in R_a\) and \(h=h_a\), taking the two parameters called \(a\) and \(b\) in that lemma to be \(31/32\) and \(45/64\), whose difference is \(17/64\), and taking \(D_0=1\). Since \(\Phi(\nu)\in Y\subset X_B\), the hypotheses of the lemma hold: the bound \(yf_{u(\Phi(\nu))}(i)\le45/64\) was obtained above at the atoms with \(h_a(i)\ne y\), \eqref{eq:res-3.3} gives \(\mathbb E_\nu[yf_{u(\Phi(\nu))}]\ge31/32\), and \eqref{eq:res-3.2} gives \(D=Z(\Phi(\nu))\ge1\). For \(\mu\in A_a\), take
\[
\nu=\Theta_{\lambda_\varepsilon}^{-1}(\mu)\in R_a.
\]
Then
\[
\operatorname{loss}_\mu(h_a)
\le
e^{-(17/64)/\lambda_\varepsilon}
=
\varepsilon/4.
\]

\proofstep{The final cover.} The sets \(R_a\) are closed in the compact realizable-distribution space, because \(F\) is continuous. Thus \(A_a\) is closed as a compact image. The regions \(D_a\) cover \(K\) and \(F\) maps into \(K\), so the sets \(R_a\) cover \(\Delta_{C_B}\), and surjectivity of the tilt on \(\Delta_{C_B}\) (Lemma~\ref{res:3.5}) gives coverage. Injectivity preserves the multiplicity bound \(k+1\): if \(\mu\in A_a\), then \(\Theta_{\lambda_\varepsilon}^{-1}(\mu)\in R_a\), so a point \(\mu\) lies in \(A_a\) only for those \(a\) with \(F(\Theta_{\lambda_\varepsilon}^{-1}(\mu))\in D_a\). Merging equal labels does not increase the multiplicity. Since \(\varepsilon/4\le\varepsilon/2\), Lemma~\ref{res:3.3} therefore gives
\[
\operatorname{lr}(C_B,\varepsilon)\le k+1.
\]

\proofstep{The constant.} For the numerical estimate, Step 5 of the proof of Theorem~\ref{res:3.1} gives
\[
1+\kappa(N,M)
\le
(2^{22}\cdot17^2+98)\ln^2(2N)\ln(2M),
\]
with
\[
2^{22}\cdot17^2+98
=
1212153954
<
2^{31}.
\]
Taking the supremum over positive error parameters yields the result: \(\operatorname{lr}(B,\eta)\) is nonincreasing in \(\eta\), so the supremum over \(\eta>0\) equals the supremum over \(\eta\in(0,1/2)\).
\end{proof}

The proof uses the map \(G\), the estimates \eqref{eq:G-cube}, \eqref{eq:res-3.1} and \eqref{eq:45-128} obtained in the proofs of Theorems~\ref{res:3.1} and~\ref{res:3.4}, the cover construction, and the tilt. The index bound of Theorem~\ref{res:3.1} itself is not used.

\subsection{Error stabilization}\label{ssec:3.5}

The function \(\varepsilon\mapsto\operatorname{lr}(C,\varepsilon)\) is nonincreasing, and~\cite[Question 2.4]{BHH26c} asks whether it can actually vary on \((0,1/2)\). Upper bounds on \(\operatorname{lr}(C)\), such as Theorem~\ref{thm:main}, concern its supremum and give no information on this question. The flattening step in the proof of Theorem~\ref{res:3.4} is a second support-preserving homeomorphism of \(\Delta_C\). Used as a transport of covers, it shows that the function is constant on \((0,\min(1/2,1/M))\).

\begin{resultenv}{Theorem}{res:3.9}{Stabilization below the \(1/M\) scale}
For every finite class \(C\) on a domain of size \(M\), and
\[
\eta\in(0,\min(1/2,1/M)),
\]
\[
\operatorname{lr}(C,\eta)
=
\operatorname{lr}(C)
=
\sup_{\eta'>0}\operatorname{lr}(C,\eta').
\]
In particular,
\[
\operatorname{lr}(B)=\operatorname{lr}(B,1/(4M)).
\]
\end{resultenv}

\begin{proof}
Fix such an \(\eta\) and a target error \(\varepsilon\in(0,1)\). We pull back a cover at error \(\eta\) by a power map that raises every atom of mass above \(\varepsilon/(2M)\) to mass above \(\eta\). Set
\[
c=\frac{1+M\eta}{2},
\qquad
\alpha=\frac{\ln(1/c)}{\ln(2M/\varepsilon)},
\]
and define the support-preserving power homeomorphism
\[
P_\alpha(\mu)\propto\mu^\alpha,
\]
with \(0^\alpha=0\). We have \(M\eta<c<1\), because \(M\eta<1\), and \(2M/\varepsilon>1\), so \(\alpha>0\) and \((\varepsilon/(2M))^\alpha=c\). The map \(P_\alpha\) is continuous on the probability simplex of atoms, and its inverse is \(P_{1/\alpha}\). It preserves supports, so \(P_\alpha\) restricts to a homeomorphism of \(\Delta_C\) onto itself (Lemma~\ref{lem:distribution-facts}(a)).

Let \(\mu\in\Delta_C\). If a hypothesis has \(\mu\)-loss \(>\varepsilon/2\), some erroneous atom has, by Lemma~\ref{lem:distribution-facts}(b),
\[
\mu_z>\frac{\varepsilon}{2M}.
\]
Its unnormalized powered mass exceeds \((\varepsilon/(2M))^\alpha=c\). Since a realizable distribution has at most \(M\) atoms in its support and each powered mass is at most \(1\), the normalizing sum is at most \(M\), and
\[
P_\alpha(\mu)_z>\frac cM>\eta.
\]
Therefore
\[
\operatorname{loss}_{P_\alpha(\mu)}(h)>\eta.
\]

Let \(L=\operatorname{lr}(C,\eta)\); we may assume \(L<\infty\). Take the closed cover \(\{G_h\}\) obtained from an \((\eta,L)\)-learner: by the first inequality of Theorem~\ref{res:b} it has multiplicity at most \(L\), and \(G_h\subseteq\{\mu:\operatorname{loss}_\mu(h)\le\eta\}\). Pull it back by \(P_\alpha\). The preceding implication shows that every pulled-back member \(P_\alpha^{-1}(G_h)\) has loss at most \(\varepsilon/2\), and the homeomorphism preserves closedness, coverage and multiplicity. Lemma~\ref{res:3.3} gives an \((\varepsilon,L)\)-learner.

Taking the supremum over target errors proves the equality: since \(\operatorname{lr}(C,\varepsilon)\) is nonincreasing in \(\varepsilon\), we get \(\operatorname{lr}(C)=\sup_{\varepsilon\in(0,1)}\operatorname{lr}(C,\varepsilon)\le L=\operatorname{lr}(C,\eta)\le\operatorname{lr}(C)\). Finally \(1/(4M)<\min(1/2,1/M)\), which gives the last assertion.
\end{proof}

This gives a partial answer to the question of~\cite{BHH26c}: below the \(1/M\) scale, the value no longer changes. The theorem concerns positive errors, and the distinction between small positive error and zero error remains. It is already present for the anchored class with one free bit. By Theorem~\ref{res:5.1} its value is \(2\) at every error in \((0,1/2)\). At error zero no finite list suffices: consider the distributions with mass \(1-t\) on the anchor atom and mass \(t\) on an atom of the free coordinate. For a fixed confidence a rule uses one sample size for all of these distributions, and as \(t\to0\) its samples become indistinguishable for the two labels of the free coordinate, while the two labels require disjoint sets of outputs.

\section{Two-sided Gibbs models}\label{app:two-sided}
This section develops an analytic alternative to the carrier construction of Part~\ref{part:upper}. Its main object is a continuous face-separating model, in the sense of Section~\ref{app:width}, with a controlled
Hilbert-space covering number.
The question is how small the metric entropy of such an encoding can be at a fixed scale.
The power-map encoding of Section~\ref{ssec:3}, from~\cite{Zhang26}, and the encoding of~\cite[Proposition 2.5]{BHHLTpoly26} have nets of logarithmic size of order $\ln^2m\ln n$, and Theorem~\ref{res:7.1} shows that a coordinate power map cannot do better on square Hadamard matrices.
We define a different encoding on the row polytope by an entropy-regularized concave maximization problem, whose optimizer couples a Gibbs distribution on rows with a Gibbs distribution on signed coordinates.
A single relative-entropy identity controls the two sides together, and with empirical row sampling and a joint coding of a coordinate sample it gives entropy of order $\ln^{3/2}m\ln n$.
Theorems~\ref{thm:main} and~\ref{thm:anchor-width} give a bound of smaller order on $u(C)$, but the carrier construction behind them gives no metric information of this kind: the entropy estimates of this section
describe the approximation of the model itself.

\subsection{The model and its optimality equations}
We define the model as the solution of a concave maximization problem and write its optimality conditions as a pair of Gibbs distributions.

Let $m=|C|\ge2$, let $A$ be the $m\times n$ matrix with rows $C$,
and set $P=\operatorname{conv}C$ and $H(q)=-\sum_cq_c\ln q_c$.
For $\Lambda,\gamma>0$, define
\[
 f_\gamma(\rho)=\gamma^{-1}\ln\sum_{j,s=\pm1}e^{\gamma s\rho_j},\qquad
 q(v)=\arg\max_{q\in\Delta_m}
 \{H(q)-\Lambda f_\gamma(A^{\mathsf T}q-v)\}.
\]
Here $f_\gamma$ is a smooth approximation of the norm $\|\rho\|_\infty$, so the problem asks for a distribution on rows of large entropy whose mean is close to $v$ in that norm.

The function $f_\gamma$ is smooth and convex, as the logarithm of a sum of exponentials of
linear functions, and $H$ is strictly concave.
The objective is continuous on the compact simplex, and its strict concavity gives a unique maximizer. The entropy derivative
at a zero coordinate makes the maximizer interior: at a point with $q_c=0$, the one-sided derivative of the
objective along the segment toward the uniform distribution is $+\infty$, because of the term
$-(\tau/m)\ln(\tau/m)$ at coordinate $c$. Compactness and uniqueness
give continuity in $v$: if $v_k\to v$, any limit point of $q(v_k)$ maximizes the limit objective,
by joint continuity of the objective in $(q,v)$.

Put $z(v)=A^{\mathsf T}q(v)$,
$\rho(v)=z(v)-v$, and let $a(v)$ be the Gibbs distribution on signed
coordinates with weights $e^{\gamma s\rho_j(v)}$.
Its signed mean is $\widetilde a(v)=\sum_{j,s}a_{j,s}(v)s e_j$, which equals
$\nabla f_\gamma(\rho(v))$ and satisfies $\|\widetilde a(v)\|_1\le1$.
At the interior maximizer the $q_c$-derivative of the objective,
$-\ln q_c-1-\Lambda\langle c,\widetilde a(v)\rangle$, does not depend on $c$.
The optimality equations are therefore
\[
 q(v)_c\ \propto\ \exp(-\Lambda\langle c,\widetilde a(v)\rangle).
\]
Thus $q(v)$ is a Gibbs distribution on rows whose exponent is determined by $a(v)$, and $a(v)$ is a Gibbs distribution on signed coordinates whose exponent is determined by $q(v)$ and $v$, which explains the name of the model.

\subsection{The coupled relative-entropy identity}
Each of the two distributions is an exponential family whose parameter is determined by the other distribution and $v$, and this gives an exact expression for a weighted sum of their symmetrized relative entropies. The identity yields the stability of the model in $v$ and its behavior under sampling of rows.
Write $J(\mu,\nu)=\operatorname{KL}(\mu\|\nu)+\operatorname{KL}(\nu\|\mu)$ for symmetrized relative entropy.

\begin{lemma}[Coupled relative-entropy identity]\label{lem:two-sided-kl}
For $v,v'\in\mathbb R^n$,
\begin{equation}\label{eq:two-sided-kl}
 J(q(v),q(v'))+\frac\Lambda\gamma J(a(v),a(v'))
 =\Lambda\langle\widetilde a(v)-\widetilde a(v'),v'-v\rangle.
\end{equation}
Consequently, with $t=\|\widetilde a(v)-\widetilde a(v')\|_1$,
\[
 t\le\gamma\|v-v'\|_\infty,\qquad
 J(q(v),q(v'))\le\frac{\Lambda\gamma}{4}\|v-v'\|_\infty^2.
\]
For $v\in P$,
$\|z(v)-v\|_\infty\le\ln m/\Lambda+\ln(2n)/\gamma$.
\end{lemma}
\begin{proof}
Write $q,a,\widetilde a,z,\rho$ at $v$ and $q',a',\widetilde a',z',\rho'$ at $v'$.

\proofstep{The identity.}
For probability vectors, $J(\mu,\nu)=\sum_i(\mu_i-\nu_i)(\ln\mu_i-\ln\nu_i)$.
Subtract the two row optimality equations and pair with the difference
of the row distributions. The normalizing constants cancel because $q-q'$ has sum zero. This gives
\[
 J(q,q')=-\Lambda\sum_c(q_c-q'_c)\langle c,\widetilde a-\widetilde a'\rangle
 =\Lambda\langle\widetilde a-\widetilde a',z'-z\rangle.
\]
The exponential-family identity for the column distributions, obtained in the same way from
$\ln a_{j,s}=\gamma s\rho_j+\text{const}$, gives
\[
 J(a,a')=\gamma\sum_{j,s}(a_{j,s}-a'_{j,s})s(\rho_j-\rho'_j)
 =\gamma\langle\widetilde a-\widetilde a',\rho-\rho'\rangle.
\]
Multiplying the second identity by $\Lambda/\gamma$, adding the first, and using $z'-z+\rho-\rho'=v'-v$ proves~\eqref{eq:two-sided-kl}.

\proofstep{Stability in $v$.}
Pinsker's inequality $\operatorname{KL}(\mu\|\nu)\ge\tfrac12\|\mu-\nu\|_1^2$ implies
$J(a,a')\ge\|a-a'\|_1^2\ge t^2$, since
$\|\widetilde a-\widetilde a'\|_1\le\|a-a'\|_1$ by the triangle inequality.
The identity and H\"older's inequality therefore bound the first term of
\eqref{eq:two-sided-kl} by
$\Lambda t\|v-v'\|_\infty-(\Lambda/\gamma)t^2$.
Nonnegativity of $J(q,q')$ gives the first estimate. Maximizing this quadratic in $t$, at
$t=\gamma\|v-v'\|_\infty/2$, gives the second.

\proofstep{Accuracy.}
For the accuracy assertion, compare the optimum with any row
distribution $q^*$ lifting $v$, so $A^{\mathsf T}q^*-v=0$. Its objective is
$H(q^*)-\Lambda f_\gamma(0)\ge-\Lambda\ln(2n)/\gamma$. Since the sum defining $f_\gamma(\rho)$ contains
the term $e^{\gamma\|\rho\|_\infty}$, we have $f_\gamma(\rho)\ge\|\rho\|_\infty$, so the optimum is at most
$\ln m-\Lambda\|z(v)-v\|_\infty$. Comparing the two bounds and dividing by $\Lambda$ gives the assertion.
\end{proof}

We next replace $v$ by the mean $V_T$ of $T$ sampled rows. A direct use of the first estimate of Lemma~\ref{lem:two-sided-kl} loses a factor of order $\ln(2n)$, because $\mathbb E\|V_T-v\|_\infty^2$ can be of order $\ln(2n)/T$. The next lemma avoids this factor by taking expectations in the identity itself and using the exchangeability of the rows.

\begin{lemma}[Row sampling]\label{lem:row-sampling}
Let $v\in P$ and let $V_T$ be the mean of $T$ independent identically distributed rows with
mean $v$. Then
\[
 \mathbb E\left[J(q(v),q(V_T))+
          \frac\Lambda\gamma J(a(v),a(V_T))\right]
 \le\frac{4\Lambda\gamma}{T}.
\]
\end{lemma}
\begin{proof}
Let $c_1,\ldots,c_T$ be the rows.
Take expectations in~\eqref{eq:two-sided-kl} with $v'=V_T$.
The term involving the fixed $\widetilde a(v)$ has expectation zero, since
$\mathbb EV_T=v$.
Since $V_T$ is symmetric in the rows and $V_T-v=T^{-1}\sum_i(c_i-v)$,
exchangeability shows that the other term is
$-\Lambda\mathbb E\langle\widetilde a(V_T),c_1-v\rangle$.
Replace $c_1$ by an independent copy $c_1'$ to form $V_T^{(1)}=V_T+(c_1'-c_1)/T$.
The vector $V_T^{(1)}$ is independent of $c_1$, so
$\mathbb E\langle\widetilde a(V_T^{(1)}),c_1-v\rangle=0$, and this term equals
$-\Lambda\mathbb E\langle\widetilde a(V_T)-\widetilde a(V_T^{(1)}),c_1-v\rangle$.
The first bound in Lemma~\ref{lem:two-sided-kl} and
$\|V_T-V_T^{(1)}\|_\infty\le2/T$ give
$\|\widetilde a(V_T)-\widetilde a(V_T^{(1)})\|_1\le2\gamma/T$. Since $\|c_1-v\|_\infty\le2$, the
difference is bounded by
$\Lambda(2\gamma/T)\cdot2$, which is the claimed bound.
\end{proof}

\subsection{Sampling the exponent and coding the sample}
By the optimality equations, $q(v)$ is determined by the vector $-\Lambda\widetilde a(v)$, whose $\ell_1$ norm is at most $\Lambda$. We approximate this vector by an average of signed coordinates, at a cost that depends only on its $\ell_1$ norm, and we count the possible averages by coding them relative to a reference distribution. The first lemma gives the approximation in Hellinger distance.

\begin{lemma}[Sampling the exponent of a Gibbs distribution]\label{lem:gibbs-sample}
Let $q_y(c)\propto e^{\langle c,y\rangle}$.
Let $b>0$, $\|\Delta\|_1\le b$, and let $X_1,\ldots,X_s$ be
independent identically distributed variables in $\{0,\pm e_1,\ldots,
\pm e_n\}$ with mean $\Delta/b$. Then
$\widehat\Delta=(b/s)\sum_tX_t$ has mean $\Delta$ and
\[
 \mathbb E\|\sqrt{q_{y+\Delta}}-\sqrt{q_{y+\widehat\Delta}}\|_2^2
 \le b^2/(2s).
\]
\end{lemma}
\begin{proof}
Such a law exists: give $\operatorname{sgn}(\Delta_j)e_j$ mass
$|\Delta_j|/b$ and the remaining mass to zero. The following argument
applies to every law satisfying the stated support and mean conditions.
For fixed $c$, $d_c=\langle c,\widehat\Delta-\Delta\rangle$ is centered
and is an average of the $s$ independent centered variables
$b\langle c,X_t\rangle-\langle c,\Delta\rangle$, each of range at most $2b$, because
$c\in\{\pm1\}^n$.
Thus Hoeffding's lemma gives
$\mathbb E e^{d_c}\le\bigl(\exp((2b/s)^2/8)\bigr)^s=e^{b^2/(2s)}$.
For $Q=q_{y+\Delta}$, the sampled Gibbs law has weights proportional to $Q_ce^{d_c}$, so the relative
entropy to the sampled Gibbs law is
$\ln\sum_cQ_ce^{d_c}-\sum_cQ_cd_c$. Since $Q$ is deterministic and $\mathbb Ed_c=0$, the second sum has expectation zero.
Jensen's inequality for the logarithm and the last exponential bound give expected relative entropy
at most $b^2/(2s)$. Squared Hellinger distance is bounded by relative
entropy: for the Bhattacharyya sum $\beta=\sum_i\sqrt{\mu_i\nu_i}$, one has
$\|\sqrt\mu-\sqrt\nu\|_2^2=2(1-\beta)\le-2\ln\beta\le\operatorname{KL}(\mu\|\nu)$, the last step by Jensen.
This proves the assertion.
\end{proof}

To code a sample from one distribution using samples from a nearby reference distribution, we use a shared random sequence of samples from the reference. The cost of the code is the logarithm of the index of the selected sample, and the next lemma bounds its expectation by the relative entropy of the two distributions, up to an additive constant.

\begin{lemma}[Poisson sampling representation]\label{lem:poisson}
Let $\pi\ll\rho$ be probability distributions on a finite set. There is a
shared random sequence of $\rho$-marks and a positive integer $K$ whose
selected mark has distribution $\pi$, with
\[
 \mathbb E\ln K\le\operatorname{KL}(\pi\|\rho)+2.
\]
\end{lemma}
This is the Poisson representation underlying the strong functional
representation method of Li and El Gamal \cite{LiEG18}.
\begin{proof}
We first show that the selected mark has law $\pi$, and then bound the index.

\proofstep{The selected mark.}
Take a Poisson process $(T_i,Y_i)$ with intensity $dt\,\rho$, ordered by
$T_i$. Put $r=\pi/\rho$ and select the smallest $T_i/r(Y_i)$, discarding
marks with $r=0$. Let $K$ be the index of the selected point. For each mark $y$ with $r(y)>0$, the
points with this mark form a Poisson process of rate $\rho(y)$ in $t$, hence of rate
$\rho(y)r(y)=\pi(y)$ in the rescaled time $t/r(y)$. Rescaling the marked times thus produces intensity
$dt\,\pi$, so its first mark $Y$ has law $\pi$, independent of its first time
$X\sim\operatorname{Exp}(1)$.

\proofstep{The index.}
Conditional on the first rescaled point $(X,Y)=(x,y)$, the selected point has original time
$xr(y)$, and $K-1$ counts the original points preceding this time. The expected number
of these points is at most $xr(y)$. Points with mark $y'$ and $r(y')=0$ are
independent of the rescaled process and have rate $\rho(y')$. Other points are restricted, by the
conditioning, to original times above $xr(y')$. Thus removing the
forbidden region preceding the first rescaled point only decreases
this intensity, and the total rate before time $xr(y)$ is at most $\sum_{y'}\rho(y')=1$. Therefore
$\mathbb E[K\mid X,Y]\le1+Xr(Y)\le(1+X)(1+r(Y))$.
Jensen, $\mathbb E\ln(1+X)\le\ln(1+\mathbb EX)=\ln2$, and $\ln(1+r)\le\ln2+\ln^+r$ give
\[
 \mathbb E\ln K\le\ln2+\mathbb E_\pi\ln(1+r)
 \le2\ln2+\mathbb E_\pi\ln^+r
 \le\operatorname{KL}(\pi\|\rho)+2\ln2+1/e<\operatorname{KL}(\pi\|\rho)+2.
\]
For the third inequality, write $\mathbb E_\pi\ln^+r=\operatorname{KL}(\pi\|\rho)+\mathbb E_\pi[\ln(1/r);\,r<1]$
and use $r\ln(1/r)\le1/e$ on $0<r<1$, integrated against $\rho$. The last inequality holds because $2\ln2+1/e<1.76$.
\end{proof}

\subsection{The entropy bound and the model}
We now combine the lemmas. Row sampling places the coordinate distribution $a(v)$ close, in relative entropy, to one of at most $m^T$ reference distributions. Coding a coordinate sample relative to that reference then leaves few possible values of the sampled exponent, so a separated subset of the image is small. A Gaussian projection of a small net gives a continuous map with small fibers, and this map yields the model.

\begin{theorem}[Fixed-scale entropy of the two-sided model]\label{thm:two-sided-entropy}
Take $\Lambda=4\ln m$, $\gamma=4\ln(2n)$, and
$U=\{\sqrt{q(v)}:v\in P\}$.
Let $\mathcal N(U,\eta)$ denote the least number of closed Euclidean
balls of radius $\eta$, with centers in $U$, covering $U$.
For any fixed sufficiently small $\eta>0$ there is a constant $K_\eta$, depending only on $\eta$, with
\[
 \ln\mathcal N(U,\eta)\le K_\eta\,\ln^{3/2}m\,\ln(2n).
\]
In particular $u(C)\le K\ln^{3/2}m\ln(2n)$ for an absolute constant $K$.
\end{theorem}
\begin{proof}
Write $G_0=\ln^{3/2}m\ln(2n)$. Since $m\ge2$ and $n\ge1$, we have $G_0\ge(\ln2)^{5/2}>2/5$ and
$\ln m\le G_0/(\ln2)^{3/2}<7G_0/4$.

\proofstep{A separated set and the row tuples.}
Fix $0<\eta\le1$ and an integer $T\ge1$, to be chosen below.
Let $Z_0\subseteq U$ be maximal with pairwise distances greater than
$2\eta$, and write $N_0=|Z_0|$. Compactness of $U$, the continuous image of $P$, makes $Z_0$ finite. We bound $N_0$ by a counting argument.
For every ordered $T$-tuple of rows $\tau$, let $v_\tau$ be its mean
and $a'_\tau=a(v_\tau)$. This distribution has full support on the $2n$ signed coordinates.
For each point of $Z_0$, fix a target $v\in P$ with $\sqrt{q(v)}$ equal to it, and draw rows from a
distribution with mean $v$.
Lemma~\ref{lem:row-sampling} supplies a tuple
$\tau$ with $(\Lambda/\gamma)J(a(v),a'_\tau)\le4\Lambda\gamma/T$, since $J(q(v),q(v_\tau))\ge0$. Hence
\[
 \operatorname{KL}(a(v)\|a'_\tau)\le J(a(v),a'_\tau)\le4\gamma^2/T=:\kappa.
\]

\proofstep{Coding the column sample.}
Set $s=\lceil2\Lambda^2/\eta^2\rceil$. For each $\tau$ use one Poisson process $\Omega_\tau$
whose marks are $s$-tuples $((j_t,s_t))_{t\le s}$ of signed coordinates with law
$(a'_\tau)^{\otimes s}$. Relative entropy adds over products, so
$\operatorname{KL}(a(v)^{\otimes s}\|(a'_\tau)^{\otimes s})\le s\kappa$.
Lemma~\ref{lem:poisson}, applied jointly to all $s$ coordinates,
selects a mark with law $a(v)^{\otimes s}$, at an index $K_v$ with
$\mathbb E\ln K_v\le s\kappa+2$. The additive constant occurs once, whereas coding the coordinates one at a time would cost it $s$ times.

Use the selected tuple to form
$\widehat y=-(\Lambda/s)\sum_{t=1}^s s_te_{j_t}$.
Apply Lemma~\ref{lem:gibbs-sample} with $y=0$,
$\Delta=-\Lambda\widetilde a(v)$, $b=\Lambda$, and
$X_t=-s_te_{j_t}$. We have $\|\Delta\|_1\le\Lambda$. These variables are independent with mean
$-\widetilde a(v)=\Delta/b$, and
$q_{\Delta}=q(v)$ by the optimality equations. The expected squared approximation error is at
most $\Lambda^2/(2s)\le\eta^2/4$.
By two applications of Markov's inequality, each event failing with probability at most $1/4$,
with probability at least
$1/2$ both the error is at most $\eta$ and
$\ln K_v\le L_0:=4(s\kappa+2)$.

\proofstep{Counting.}
For each $\tau$, keep all code indices at most $e^{L_0}$. This keeps at most
$m^Te^{L_0}$ approximations $\sqrt{q_{\widehat y}}$ in total, for one realization of the processes
$(\Omega_\tau)_\tau$, and the set kept does not depend on the target $v$.
Take $R=\lceil\log_2N_0\rceil+1$ independent copies of these processes.
For one point of $Z_0$, the probability that all $R$ copies fail is at most
$2^{-R}\le1/(2N_0)$.
A union bound shows that some fixed choice covers every point of
$Z_0$ within $\eta$. Each kept approximation serves at most one
point of $Z_0$, by its separation: an approximation within $\eta$ of two points would put them
within $2\eta$. Thus, with $L_0=16s\gamma^2/T+8$ and $R\le\log_2N_0+2$,
\[
 N_0\le Rm^Te^{L_0},\qquad
 \ln N_0\le T\ln m+16s\gamma^2/T+8+
             \ln(\log_2N_0+2).
\]
Choose $T=\lceil4\gamma\sqrt{s/\ln m}\rceil$. Then
$T\ln m\le4\gamma\sqrt{s\ln m}+\ln m$ and $16s\gamma^2/T\le4\gamma\sqrt{s\ln m}$, where the ceiling
contributes the term $\ln m$.
Since $\Lambda\ge4\ln2$ and $\eta\le1$, we have $s\le2\Lambda^2/\eta^2+1\le(9/4)\Lambda^2/\eta^2$, so
\[
 8\gamma\sqrt{s\ln m}\le12\gamma\Lambda\sqrt{\ln m}/\eta=192\,G_0/\eta.
\]
To absorb the final logarithm, we use $\ln(x/\ln2+2)\le x/2+1$ for $x\ge0$, with $x=\ln N_0$. Together with $\ln m<7G_0/4$ and $1<5G_0/2$, this gives
\[
 \ln N_0\le2\bigl(192\,G_0/\eta+\ln m+8\bigr)+2\le(384/\eta+49)\,G_0,
\]
which has the form
$\ln N_0\le K_\eta\Lambda\gamma\sqrt{\ln m}$.
Maximality makes $Z_0$ a $2\eta$-net: closed balls of radius $2\eta$ about $Z_0$ cover $U$.
Rescaling the fixed accuracy
parameter proves the displayed covering-number bound:
$\ln\mathcal N(U,2\eta)\le(384/\eta+49)G_0$ for $0<\eta\le1$.

\proofstep{The map to a Euclidean space.}
To obtain a model, we use a continuous map with small fibers. A map of this kind,
built from a net, a Johnson--Lindenstrauss projection, and a Lipschitz
extension, is constructed in the proof of Theorem~\ref{res:3.1},
from~\cite{Zhang26}; see~\eqref{eq:res-3.1}. We use the averaging form
of~\cite[Lemma 2.4, inequality (4)]{BHHLTpoly26}: a compact subset of a
Hilbert unit sphere with a $\delta$-net $Z$ contained in that subset
admits a continuous map to $\mathbb R^{O(1+\ln|Z|)}$ whose fibers
have diameter at most $(2+6\sqrt3)\delta$.
The set $U$ lies on the unit sphere, and $Z_0\subseteq U$ is a
$2\eta$-net. Choose $0<\eta<1/50$, so
$(2+6\sqrt3)2\eta<1/2$. We take $\eta=1/60$.

For completeness we write the map out. With $\delta=2\eta$, Lemma~\ref{res:3.0prime} gives
$k=\lceil32(2\ln N_0+\ln16)\rceil$ and a linear map $\Pi:\mathbb R^m\to\mathbb R^k$ with
$\tfrac12\|z-z'\|_2^2\le\|\Pi z-\Pi z'\|_2^2\le\tfrac32\|z-z'\|_2^2$ for $z,z'\in Z_0$.
For $w\in U$ put
$\lambda_z(w)=(2\delta-\|w-z\|_2)_+/\sum_{z'\in Z_0}(2\delta-\|w-z'\|_2)_+$ and
$E(w)=\sum_{z\in Z_0}\lambda_z(w)\Pi z$. The denominator is positive because $Z_0$ is a
$\delta$-net, so $E$ is continuous.
Let $z_w\in Z_0$ be within $\delta$ of $w$. If $\lambda_z(w)>0$, then
$\|z-z_w\|_2<3\delta$, so $\|E(w)-\Pi z_w\|_2\le3\sqrt{3/2}\,\delta$. If
$E(w)=E(w')$, then $\|\Pi z_w-\Pi z_{w'}\|_2\le6\sqrt{3/2}\,\delta$ and
\[
 \|w-w'\|_2\le2\delta+\|z_w-z_{w'}\|_2\le2\delta+\sqrt2\cdot6\sqrt{3/2}\,\delta=(2+6\sqrt3)\delta.
\]

\proofstep{The model.}
Let $f(v)=E(\sqrt{q(v)})$, a continuous map from $P$ to $\mathbb R^k$. By the choice of $\Lambda,\gamma$,
Lemma~\ref{lem:two-sided-kl} gives $\|z(v)-v\|_\infty\le1/4+1/4$ on $P$.
Also $\|A^{\mathsf T}(q-q')\|_\infty\le\|q-q'\|_1$, since the entries of $A$ are $\pm1$, and
$\|q-q'\|_1\le\|\sqrt q-\sqrt{q'}\|_2\|\sqrt q+\sqrt{q'}\|_2\le2\|\sqrt q-\sqrt{q'}\|_2$ by the
Cauchy--Schwarz inequality.
For $v,v'$ in one fiber of $f$, the points $\sqrt{q(v)}$ and $\sqrt{q(v')}$ are at distance less than $1/2$, by the fiber bound and the choice of $\eta$, and this gives
\[
 \|v-v'\|_\infty
 \le1+\|q(v)-q(v')\|_1
 \le1+2\|\sqrt{q(v)}-\sqrt{q(v')}\|_2<2.
\]
Thus no fiber meets opposite coordinate faces.
The images of each opposite pair are therefore disjoint compact sets, and continuous
coordinate separators on the finite-dimensional target produce the
map $g$ in the definition of $u(C)$ in Section~\ref{app:width}. To construct them, let $\mathsf K$ be a closed cube in $\mathbb R^k$ containing
$f(P)$, a finite polyhedron of dimension $k$, and for each $j$ choose a continuous
$g_j:\mathsf K\to[-1,1]$ equal to $s$ on $f(P\cap\{x_j=s\})$, by Urysohn's lemma, or a suitable constant if
one of the two sets is empty. This completes the model construction: $u(C)\le k+1$.
With $\eta=1/60$, the counting step gives $\ln N_0\le23089\,G_0$, and
$k+1\le64\ln N_0+32\ln16+2$. Since $G_0>2/5$, this is at most an absolute constant times
$G_0$.
\end{proof}

\section{Closure chains and chain labels}\label{ssec:6}

Theorem~\ref{thm:main} bounds the list number of a sign matrix in terms of its size alone. This section gives purely combinatorial upper bounds that depend on the structure of the matrix: the height of the poset of closed faces of its sign complex, the number of labels needed along chains of closures, and the switching binary rank. The last of these is the rank over \(\mathbb F_2\) of the matrix after one row and one column have been switched to positive signs. On structured matrices these bounds are stronger than Theorem~\ref{thm:main}: for the Sylvester Hadamard matrix of order \(2^m\) they give list number at most \(m+1\) in place of a bound of order \(m^2\), and the logarithmic-product bound holds with constant one on all matrices generated from small matrices by signed joins and Kronecker products. Sections~\ref{ssec:7.6} and~\ref{ssec:7.7} characterize the list number through covers correct on heavy atoms and bound the switching rank on classes generated from finitely many matrices.

These parameters are tied directly to both invariants studied in this paper: the height of the poset of closed faces gives an odd map from the sign complex to a sphere, and a labeling of closures gives a closed cover of the realizable distributions and hence a learning rule. The resulting inequalities sharpen the height bound of \cite{FHV26} and, for total classes, refine \cite[Theorem 1.3]{BHHLT26}. In Sections~\ref{ssec:6.4} and~\ref{ssec:6.5} we partition the faces of the sign complex and join odd maps on the pieces, which gives a sharp index bound for bounded error intersections and isolates the faces on which further structure of the matrix is needed, as for the Paley matrices of Section~\ref{ssec:8}.

\subsection{Signed closure height and switching binary rank}\label{ssec:6.1}

The face poset of a sign complex can be replaced by the smaller poset of closed faces, in which a face is determined by its compatible signed rows. We show that the index is bounded by the height of this poset, and that the height is bounded by a binary rank that is invariant under switching, transposition, and repetition.

Throughout this section, \(B\in\{\pm1\}^{N\times M}\) is a total matrix with \(N,M\ge1\), and \(C=C_B\) is its row class. A \emph{signed row} is a pair \((r,s)\) with \(r\in[N]\) and \(s\in\{\pm1\}\); we put \(-(r,s)=(r,-s)\). We write \(i^b\), \(b\in\{\pm1\}\), for the vertices \(i^+,i^-\) of \(S(B)\). The definition of \(\sigma_r^{\pm}\) in Section~\ref{ssec:2.1} says that
\begin{equation}\label{eq:comb-membership}
i^b\in\sigma_r^s
\quad\Longleftrightarrow\quad
sB_{ri}=b .
\end{equation}
For a set \(F\) of vertices put \(F^\pm=\{i:i^\pm\in F\}\) and \(-F=\{i^{-b}:i^b\in F\}\). By~\eqref{eq:comb-membership}, \(\sigma_r^{-s}=-\sigma_r^s\), and a simplex of \(S(B)\) never contains both \(i^+\) and \(i^-\); hence \(F\ne-F\) for every nonempty face \(F\).

Since \(B\) is total, every \(\sigma_r^s\) has exactly \(M\) vertices, one over each column. Every simplex of \(S(B)\) lies in some \(\sigma_r^s\), and no \(\sigma_r^s\) properly contains another, so the facets of \(S(B)\) are exactly the sets \(\sigma_r^s\). For a nonempty face \(F\) let
\[
R(F)=\{(r,s):F\subseteq\sigma_r^s\}
\]
be its set of compatible signed rows, which is the set \(R_B(F)\) of the proof of Theorem~\ref{res:2.1}. It is nonempty, and the facets containing \(F\) are the \(\sigma_r^s\) with \((r,s)\in R(F)\).

For a nonempty face \(F\), define its signed closure by
\[
c(F)=\bigcap\{\text{facets containing }F\}=\bigcap_{(r,s)\in R(F)}\sigma_r^s.
\]
It is equivariant, order-preserving, and idempotent; the next lemma records these properties.

\begin{lemma}[Signed closure]\label{lem:comb-signed-closure}
Let \(F,G\) be nonempty faces of \(S(B)\).
\begin{enumerate}[label=(\alph*),leftmargin=2em]
\item \(F\subseteq c(F)\), the set \(c(F)\) is a face, and \(R(c(F))=R(F)\).
\item \(c(c(F))=c(F)\).
\item If \(F\subseteq G\), then \(R(G)\subseteq R(F)\) and \(c(F)\subseteq c(G)\).
\item \(R(-F)=-R(F)\) and \(c(-F)=-c(F)\).
\end{enumerate}
\end{lemma}

\begin{proof}
(a) Every facet containing \(F\) contains \(F\), so \(F\subseteq c(F)\). As \(R(F)\ne\varnothing\), the set \(c(F)\) lies in a facet and is a face. From \(F\subseteq c(F)\) we get \(R(c(F))\subseteq R(F)\). If \((r,s)\in R(F)\), then \(\sigma_r^s\) is one of the sets intersected in \(c(F)\), so \(c(F)\subseteq\sigma_r^s\) and \((r,s)\in R(c(F))\).

(b) The set \(c(F)\) is determined by \(R(F)\), and \(R(c(F))=R(F)\) by (a).

(c) If \(G\subseteq\sigma_r^s\), then \(F\subseteq\sigma_r^s\); thus \(R(G)\subseteq R(F)\). The intersection defining \(c(F)\) therefore runs over a larger family than the one defining \(c(G)\), so \(c(F)\subseteq c(G)\).

(d) Since \(\sigma_r^{-s}=-\sigma_r^s\), we have \(F\subseteq\sigma_r^s\) if and only if \(-F\subseteq\sigma_r^{-s}\). This gives \(R(-F)=-R(F)\), and then \(c(-F)=\bigcap_{(r,s)\in R(F)}\sigma_r^{-s}=-c(F)\).
\end{proof}

A face \(F\) is \emph{signed closed} if \(c(F)=F\). By Lemma~\ref{lem:comb-signed-closure}(b), the signed closed faces are exactly the values of \(c\). They form the \emph{signed closure poset} \(\mathcal P(B)\), ordered by inclusion, and \(h(B)\) is the largest number of members of a chain of \(\mathcal P(B)\) (Section~\ref{ssec:2.4}).

To relate \(\mathcal P(B)\) to \(S(B)\) we use order complexes. For a finite poset \(Q\) let \(\Delta(Q)\) be its order complex: its vertices are the elements of \(Q\), and its simplices are the nonempty chains. An order-preserving map \(Q\to Q'\) sends a chain to a chain after repetitions are deleted, so it induces a simplicial map \(\Delta(Q)\to\Delta(Q')\). Let \(\mathcal F(B)\) be the poset of nonempty faces of \(S(B)\). Then \(\Delta(\mathcal F(B))=\operatorname{sd}S(B)\) is the barycentric subdivision (\cite[Section 2.1]{Hatcher02}), and sending each face to its barycenter and extending affinely gives a homeomorphism \(|\operatorname{sd}S(B)|\to|S(B)|\) that commutes with the involutions. The signed closure order complex is \(\Delta(\mathcal P(B))\). The map \(F\mapsto-F\) is an order automorphism of \(\mathcal F(B)\), and by Lemma~\ref{lem:comb-signed-closure}(d) of \(\mathcal P(B)\).

The closure map retracts every family of faces that it preserves onto the closed members of that family. We apply this below to the whole face poset, and in Section~\ref{ssec:6.5} to a smaller family.

\begin{lemma}[Closure retraction]\label{lem:comb-closure-retraction}
Let \(P\subseteq\mathcal F(B)\) satisfy \(c(P)\subseteq P\), and let \(j:c(P)\to P\) be the inclusion. Then \(c\circ j=\operatorname{id}\) on \(\Delta(c(P))\), and \(j\circ c\) is homotopic to the identity on \(|\Delta(P)|\). If \(-P=P\), then all these maps and the homotopy are equivariant.
\end{lemma}

\begin{proof}
Elements of \(c(P)\) are signed closed, so \(c\circ j=\operatorname{id}\). Give \(P\times\{0<1\}\) the product order and define
\[
H:P\times\{0<1\}\to P,
\qquad
H(F,0)=F,
\quad
H(F,1)=c(F).
\]
The values lie in \(P\) because \(c(P)\subseteq P\); no other face is involved, so \(P\) need not be a downset. The map \(H\) is order-preserving: for \(F\subseteq G\) we have \(H(F,0)=F\subseteq G=H(G,0)\), \(H(F,1)\subseteq H(G,1)\) by Lemma~\ref{lem:comb-signed-closure}(c), and \(H(F,0)=F\subseteq G\subseteq c(G)=H(G,1)\) by Lemma~\ref{lem:comb-signed-closure}(a). The order complex of \(P\times\{0<1\}\) is the prism triangulation of \(|\Delta(P)|\times[0,1]\) (see \cite[proof of Theorem 2.10]{Hatcher02}): over a chain \(F_0\subsetneq\cdots\subsetneq F_k\) of \(P\), the top simplices have the vertices
\[
(F_0,0),\ldots,(F_i,0),(F_i,1),\ldots,(F_k,1),
\qquad
0\le i\le k.
\]
Hence the simplicial map induced by \(H\) is a homotopy from the identity to \(j\circ c\). If \(-P=P\), then \(H(-F,e)=-H(F,e)\) by Lemma~\ref{lem:comb-signed-closure}(d), so the homotopy is equivariant.
\end{proof}

A family of faces invariant under the involution admits an odd map to a sphere whose dimension is controlled by the longest chain of the family.

\begin{lemma}[Level map]\label{lem:comb-level-map}
Let \(Q\subseteq\mathcal F(B)\) be nonempty with \(-Q=Q\), and let \(h\) be the largest number of members of a chain of \(Q\). Then there is a continuous odd map \(|\Delta(Q)|\to S^{h-1}\).
\end{lemma}

\begin{proof}
For \(F\in Q\) let \(\rho(F)\in\{1,\ldots,h\}\) be the largest number of members of a chain of \(Q\) with top element \(F\). Then \(\rho\) strictly increases along chains, and \(\rho(-F)=\rho(F)\) because \(F\mapsto-F\) is an order automorphism of \(Q\). Since \(F\ne-F\), we can choose signs \(\eta(F)\in\{\pm1\}\) with \(\eta(-F)=-\eta(F)\) on each orbit. Send the vertex \(F\) of \(\Delta(Q)\) to \(\eta(F)e_{\rho(F)}\in\mathbb R^h\) and extend affinely on simplices. A point \(\sum_j\lambda_jF_j\) of a chain simplex goes to \(\sum_j\lambda_j\eta(F_j)e_{\rho(F_j)}\). The indices \(\rho(F_j)\) are distinct, so this vector is nonzero. The map is odd by construction, and normalizing gives the required map.
\end{proof}

These lemmas bound the index by the height of \(\mathcal P(B)\). To bound the height we bring the matrix to a normal form. Writing \(B=(-1)^A\) with \(A\) binary, the matrix \(D\) of the next theorem is the binary matrix obtained from \(B\) by negating rows and columns until its first row and first column are positive. Its rank \(r_2(B)\), which is defined for total matrices only, is the \emph{switching binary rank} of \(B\).

\begin{resultenv}{Theorem}{res:6.1}{Closure height and switching rank}
The signed closure order complex is equivariantly homotopy equivalent to \(S(B)\). If \(h(B)\) is the largest number of members of a chain of \(\mathcal P(B)\), then
\[
\operatorname{Ind}S(B)\le h(B)-1,
\qquad
h(B)=h(B^T).
\]
Write
\[
B=(-1)^A,
\qquad
D=A+A_{r1}+A_{1i}+A_{11},
\]
with the sum taken entrywise in \(\mathbb F_2\), and
\[
r_2(B)=\operatorname{rank}_{\mathbb F_2}D.
\]
Then
\[
h(B)\le r_2(B)+1,
\qquad
\operatorname{Ind}S(B)\le r_2(B).
\]
\end{resultenv}

\begin{proof}
The assertions on \(h(B)\) and the index follow from the three lemmas above and a duality between closed faces and their compatible signed rows. For the rank bound we first show that \(r_2\) is invariant under switching. We then encode the signed rows in a binary space of dimension \(r_2(B)+1\), in which a strict chain of closed faces gives a strictly decreasing chain of affine subspaces.

\proofstep{The homotopy equivalence.} Apply Lemma~\ref{lem:comb-closure-retraction} with \(P=\mathcal F(B)\), so that \(c(P)=\mathcal P(B)\). With the equivariant homeomorphism \(|\operatorname{sd}S(B)|\cong|S(B)|\), this gives \(S(B)\simeq_{\mathbb Z_2}\Delta(\mathcal P(B))\).

\proofstep{The height bound for the index.} Lemma~\ref{lem:comb-level-map} with \(Q=\mathcal P(B)\) gives an odd map \(|\Delta(\mathcal P(B))|\to S^{h(B)-1}\). Composing it with the odd map \(|S(B)|\cong|\operatorname{sd}S(B)|\to|\Delta(\mathcal P(B))|\) induced by \(c\) gives \(\operatorname{Ind}S(B)\le h(B)-1\).

\proofstep{Duality of heights.} The vertices of \(S(B^T)\) are the signed rows of \(B\), and~\eqref{eq:comb-membership} applied to \(B^T\) says that \((r,s)\in\sigma_i^b(B^T)\) if and only if \(bB_{ri}=s\), that is, \(sB_{ri}=b\). Thus
\[
i^b\in\sigma_r^s(B)
\quad\Longleftrightarrow\quad
(r,s)\in\sigma_i^b(B^T).
\]
For a nonempty face \(F\) of \(S(B)\) this gives \(R(F)=\bigcap_{i^b\in F}\sigma_i^b(B^T)\), a nonempty face of \(S(B^T)\). For a nonempty face \(G\) of \(S(B^T)\) let \(R^T(G)=\{i^b:G\subseteq\sigma_i^b(B^T)\}=\bigcap_{(r,s)\in G}\sigma_r^s(B)\) be its set of compatible signed columns, and let \(c^T\) be the signed closure of \(S(B^T)\). The same display gives
\[
c(F)=R^T(R(F)),
\qquad
c^T(G)=R(R^T(G)).
\]
If \(F\in\mathcal P(B)\) and \(G=R(F)\), then \(R^T(G)=c(F)=F\) and \(c^T(G)=R(F)=G\), so \(G\in\mathcal P(B^T)\). Exchanging the roles of \(B\) and \(B^T\) shows that \(R:\mathcal P(B)\to\mathcal P(B^T)\) and \(R^T:\mathcal P(B^T)\to\mathcal P(B)\) are mutually inverse. Both reverse inclusion by Lemma~\ref{lem:comb-signed-closure}(c). These are the compatibility maps of the proof of Theorem~\ref{res:2.1}, restricted to closed faces. An order-reversing bijection preserves the largest number of members of a chain, so \(h(B)=h(B^T)\).

\proofstep{Invariance of the switching rank.} The matrix \(D\) has entries \(D_{ri}=A_{ri}+A_{r1}+A_{1i}+A_{11}\) in \(\mathbb F_2\), and its first row and first column vanish. We use the pivot \((1,1)\) in this definition. The invariance of its rank is seen most easily on an augmented matrix, on which the operations of switching act by elementary row and column operations. Index the rows of the augmented binary matrix
\[
E(A)=\begin{pmatrix}0&\mathbf1^T\\\mathbf1&A\end{pmatrix}
\]
by \(0,1,\ldots,N\) and its columns by \(0,1,\ldots,M\). To compare its rank with that of \(D\), perform the following elementary operations:
\begin{enumerate}[leftmargin=2em]
\item For \(r\ge2\), add row \(1\) to row \(r\). Row \(r\) becomes \((0,A_{r,\cdot}+A_{1,\cdot})\), and column \(0\) now has a single nonzero entry, in row \(1\).
\item For every \(i\ge1\) with \(A_{1i}=1\), add column \(0\) to column \(i\). This changes row \(1\) alone, which becomes \((1,0,\ldots,0)\).
\item For \(i\ge2\), add column \(1\) to column \(i\). Row \(0\) becomes \((0,1,0,\ldots,0)\), and for \(r,i\ge2\) the entry in position \((r,i)\) becomes \(A_{ri}+A_{1i}+A_{r1}+A_{11}=D_{ri}\).
\item For every \(r\ge2\) with \(A_{r1}+A_{11}=1\), add row \(0\) to row \(r\). This clears column \(1\) in the rows \(r\ge2\) and changes nothing else.
\end{enumerate}
The result is
\[
\begin{pmatrix}0&1&0\\1&0&0\\0&0&D'\end{pmatrix},
\qquad
D'=(D_{ri})_{r,i\ge2},
\]
where \(D'\) is empty if \(N=1\) or \(M=1\). Hence \(\operatorname{rank}_{\mathbb F_2}E(A)=2+\operatorname{rank}_{\mathbb F_2}D'=2+\operatorname{rank}_{\mathbb F_2}D\).

Negating row \(r\) of \(B\) adds \(\mathbf1\) to row \(r\) of \(A\), which adds row \(0\) of \(E(A)\) to its row \(r\). Negating a column of \(B\) adds column \(0\) of \(E(A)\) to the corresponding column. Row and column permutations of \(B\) permute the rows and columns of \(E(A)\), and repeating a row or column of \(B\) repeats one of \(E(A)\). Finally \(E(A^T)=E(A)^T\). All these operations preserve \(\operatorname{rank}_{\mathbb F_2}E(A)\), so \(r_2(B)\) is invariant under row and column negations, permutations, repetitions, and transposition. The matrix built with another pivot \((r_0,i_0)\) is, up to a permutation, the matrix \(D\) of the matrix obtained from \(A\) by exchanging rows \(1,r_0\) and columns \(1,i_0\); it therefore also has rank \(\operatorname{rank}_{\mathbb F_2}E(A)-2\). Thus \(r_2(B)\) does not depend on the pivot.

\proofstep{The normalization.} Negating rows and columns does not change \(r_2\), by the invariance just proved. It does not change \(h\) either: negating row \(r\) exchanges \(\sigma_r^+\) and \(\sigma_r^-\) and leaves \(S(B)\) unchanged, while negating column \(i\) renames \(i^+\) and \(i^-\), which is an isomorphism of complexes carrying facets to facets and closures to closures. Negate the rows \(r\) with \(B_{r1}=-1\), and then the columns whose entry in the first row is now \(-1\). After this, \(A_{r1}=A_{1i}=0\) for all \(r,i\), so \(D=A\).

Let \(d_r\) be row \(r\) of \(D\), and let \(V\subseteq\mathbb F_2^M\) be its row space. Every vector of \(V\) has first coordinate \(0\), because the first column of \(D\) vanishes, whereas \(\mathbf1\) has first coordinate \(1\). So the sum
\[
W=V\oplus\langle\mathbf1\rangle
\]
is direct and \(\dim W=r_2(B)+1\). Encode signed rows by \(x(r,+)=d_r\) and \(x(r,-)=d_r+\mathbf1\). Then \(sB_{ri}=(-1)^{x(r,s)_i}\). Let \(\beta(+1)=0\) and \(\beta(-1)=1\). By~\eqref{eq:comb-membership},
\[
i^b\in\sigma_r^s
\quad\Longleftrightarrow\quad
x(r,s)_i=\beta(b).
\]
Let $\mathcal R=\{d_r,d_r+\mathbf1:r\in[N]\}\subseteq W$. For a nonempty face \(F\) let
\[
L_F=\{x\in W:x_i=\beta(b)\text{ for every }i^b\in F\},
\qquad
X(F)=\mathcal R\cap L_F .
\]
The previous display shows that \((r,s)\in R(F)\) if and only if \(x(r,s)\in L_F\). So \(X(F)\) is the set of compatible signed-row encodings of \(F\), it is nonempty, and \(R(F)\) is the full preimage of \(X(F)\) under the encoding. The set \(L_F\subseteq W\) is the affine subspace cut out by the coordinate equations of the face. Because \(\mathbf1\in W\), the coordinate functional \(x\mapsto x_i\) of a vertex \(i^b\in F\) is nonzero on \(W\), so \(L_F\) lies in an affine hyperplane of \(W\) and \(\dim L_F\le r_2(B)\).

\proofstep{The chain bound.} The compatible set \(X(F)\) need not be an affine subspace, so the argument uses its affine hull and its description as \(\mathcal R\cap L_F\). Let \(F\subsetneq F'\) be signed closed faces. Then \(R(F')\subseteq R(F)\), hence \(X(F')\subseteq X(F)\). Equality is impossible: \(X(F')=X(F)\) would give \(R(F')=R(F)\), and then \(F'=c(F')=c(F)=F\). Their affine hulls in \(W\) also change strictly: \(L_{F'}\) is an affine subspace containing \(X(F')\), so it contains \(\operatorname{aff}X(F')\); if \(\operatorname{aff}X(F')=\operatorname{aff}X(F)\), then \(X(F)\subseteq\mathcal R\cap L_{F'}=X(F')\), a contradiction. Thus \(\operatorname{aff}X(F')\) is a proper affine subspace of \(\operatorname{aff}X(F)\), and \(\dim\operatorname{aff}X(F)\) strictly decreases along a strict chain of closed faces. It lies between \(0\) and \(\dim L_F\le r_2(B)\). Therefore a chain has at most \(r_2(B)+1\) members, that is, \(h(B)\le r_2(B)+1\). The height bound for the index then gives \(\operatorname{Ind}S(B)\le r_2(B)\).
\end{proof}

Theorem~\ref{res:6.1} sharpens the height bound of \cite{FHV26} and extends its affine
binary construction. To compare the two heights, let \(h_{\mathrm{FHV}}\) be the height of \cite[Theorem 27]{FHV26}, the largest number of members of a chain among the nonempty intersections of the sets \(\{i:B_{ri}=+1\}\) and \(\{i:B_{ri}=-1\}\). For a signed closed face \(F\), the equality \(F=\bigcap_{(r,s)\in R(F)}\sigma_r^s\) gives \(F^+=\bigcap_{(r,s)\in R(F)}\{i:B_{ri}=s\}\) and \(F^-=\bigcap_{(r,s)\in R(F)}\{i:B_{ri}=-s\}\). So each of \(F^+\) and \(F^-\) is empty or such an intersection. Along a strict chain the pairs \((F^+,F^-)\) strictly increase in the product order and never equal \((\varnothing,\varnothing)\). Each coordinate changes at most \(h_{\mathrm{FHV}}\) times, and at most \(h_{\mathrm{FHV}}-1\) times if it starts nonempty, and at least one coordinate starts nonempty. So a chain has at most \(2h_{\mathrm{FHV}}-1\) steps, and \(h(B)\le2h_{\mathrm{FHV}}\). Thus \(\operatorname{Ind}S(B)\le h(B)-1\) implies the bound \(2h_{\mathrm{FHV}}-1\) of that theorem.

The switching-rank and sign-rank bounds are
incomparable: the Sylvester Hadamard matrix of order $2^m$ has $r_2=m$ and
sign rank at least $2^{m/2}$~\cite{Forster02}, whereas the threshold matrix $T_n$, $n\ge2$ (defined in Theorem~\ref{res:6.4}), has $r_2=n-1$ and sign
rank two.

For $H_{2^m}(r,j)=(-1)^{r\cdot j}$, the pivot $(0,0)$ gives $D_{rj}=r\cdot j$, that is, $D=XX^T$, where the rows of the $2^m\times m$ matrix $X$ are all the vectors of $\mathbb F_2^m$. Since $X$ has rank $m$, the map $X$ is injective and $X^T$ is surjective, so $r_2=\operatorname{rank}_{\mathbb F_2}D=m$. The sign-rank estimate is Forster's bound $\sqrt{NM}/\|H\|_{2\to2}$ with $N=M=2^m$ and $\|H\|_{2\to2}=2^{m/2}$.

For $T_n$ the first row is positive, so $D_{ri}=A_{ri}+A_{r1}$. For $r\ge2$ we have $A_{r1}=1$, and $D_{ri}=1$ exactly when $i\ge r$. The rows $2,\ldots,n$ of $D$ are the indicator vectors of the intervals $[r,n]$, which are linearly independent, so $r_2(T_n)=n-1$. A sign representation of dimension two is given after the proof of Theorem~\ref{res:6.4}. Dimension one is impossible for $n\ge2$, because it would make any two rows equal up to sign, whereas $c_2$ is neither $c_1$ nor $-c_1$.

The real rank, in contrast, always dominates the switching rank: $r_2(B)+1\le\operatorname{rank}_{\mathbb R}B$
holds for every \(B\). Indeed, negating rows and columns preserves both sides, by the invariance of \(r_2\) proved above and because negation does not change a real rank. So we may switch the first row and column to
positive signs. Write $B=\mathbf1-2D$ as an integer matrix, with the first row and column
of $D$ zero. The first row of $B$ is $\mathbf1$, and row $r$ of $B$ minus the first row is $-2d_r$. Hence the real row space of $B$ is $\mathbb R\mathbf1$ plus the real row space of $D$. This sum is direct, because the vectors of the second space have first coordinate zero and \(\mathbf1\) does not, so $\operatorname{rank}_{\mathbb R}B=1+
\operatorname{rank}_{\mathbb R}D$. A nonzero minor modulo two
is also a nonzero integer minor, so $\operatorname{rank}_{\mathbb R}D\ge r_2(B)$.

The switching binary rank behaves additively under Kronecker products and signed joins, and this carries the bound \(r_2+1\le\log_2(2N)\log_2(2M)\) from small matrices to every matrix they generate. Let $J(A,B)$ be the signed join defined in Section~\ref{ssec:4.3}, and let $A\otimes B$ be the Kronecker product, with rows $(r,u)$, columns $(i,j)$, and entries $A_{ri}B_{uj}$. A repetition adds a copy of an existing row or column.

The class \(\mathcal G\) of the next theorem contains the recursive family \(A_{t+1}=J(A_t,A_t)^{\mathsf T}\) of Theorem~\ref{res:4.3}(i), on which the index attains the logarithmic-product order, so the bound is of the right order on this class. The restriction \(\min(N,M)\le16\) serves only to make the generators satisfy the bound. Theorem~\ref{res:7.10} treats an arbitrary finite set of generators, with a constant depending on that set.

\begin{resultenv}{Theorem}{res:6.2}{Operations and the generated class}
For total matrices \(A,B\),
\[
r_2(A\otimes B)=r_2(A)+r_2(B),
\]
\[
r_2(J(A,B))=r_2(A)+r_2(B)+1.
\]
Let \(\mathcal G\) be generated from arbitrary total matrices with \(\min(N,M)\le16\) by finitely many signed joins, transpositions, Kronecker products, row/column permutations, row/column negations, and repetitions. Every \(B\in\mathcal G\) satisfies
\[
\operatorname{Ind}S(B)+1
\le
r_2(B)+1
\le
\log_2(2N)\log_2(2M).
\]
\end{resultenv}

\begin{proof}
Let \(A,B\) have sizes \(N_1\times M_1\) and \(N_2\times M_2\), let \(\mathsf a,\mathsf b\) be their binary matrices (\(A=(-1)^{\mathsf a}\), \(B=(-1)^{\mathsf b}\)), let \(D^A,D^B\) be their matrices \(D\) for the pivot \((1,1)\), with rows \(d^A_r,d^B_u\), and let \(V_A,V_B\) be their row spaces. All vectors of \(V_A\) and \(V_B\) vanish at the first coordinate, and \(d^A_1=0\), \(d^B_1=0\).

\proofstep{Kronecker products.} The binary matrix of \(A\otimes B\) has entries \(\mathsf a_{ri}+\mathsf b_{uj}\). With the pivot \(((1,1),(1,1))\), its normalized matrix has the entries
\[
(\mathsf a_{ri}+\mathsf b_{uj})+(\mathsf a_{r1}+\mathsf b_{u1})+(\mathsf a_{1i}+\mathsf b_{1j})+(\mathsf a_{11}+\mathsf b_{11})
=D^A_{ri}+D^B_{uj}.
\]
For \(f\in\mathbb F_2^{M_1}\) and \(g\in\mathbb F_2^{M_2}\) define the functions \(\pi_1^*f(i,j)=f(i)\) and \(\pi_2^*g(i,j)=g(j)\). Row \((r,u)\) of the normalized matrix is \(\pi_1^*d^A_r+\pi_2^*d^B_u\). Taking \(u=1\) and \(r=1\) shows that \(\pi_1^*d^A_r\) and \(\pi_2^*d^B_u\) are themselves rows, so the row space is \(\pi_1^*V_A+\pi_2^*V_B\). The maps \(\pi_1^*,\pi_2^*\) are injective. If \(\pi_1^*f=\pi_2^*g\) with \(f\in V_A\), \(g\in V_B\), then \(f(i)=g(j)\) for all \(i,j\); putting \(i=1\) gives \(g=f(1)=0\), and putting \(j=1\) gives \(f=0\). So the sum is direct, proving the first identity.

\proofstep{Signed joins.} Both sides of the second identity are unchanged if a row or column of \(A\) or \(B\) is negated. Negating a column of \(A\) or \(B\) negates a column of \(J(A,B)\). Negating row \(u\) of \(B\) exchanges the rows \((r,u,+)\) and \((r,u,-)\) of \(J(A,B)\). Negating row \(r\) of \(A\) replaces row \((r,u,s)\) by \((-A_{r,\cdot},sB_{u,\cdot})=-(A_{r,\cdot},-sB_{u,\cdot})\), which is the negative of the old row \((r,u,-s)\). In each case \(J(A,B)\) changes by negations and permutations of rows and columns. By the invariance proved in Theorem~\ref{res:6.1}, this changes none of \(r_2(J(A,B))\), \(r_2(A)\), \(r_2(B)\). We may therefore normalize \(A\) and \(B\) as in that proof, and assume \(\mathsf a=D^A\) and \(\mathsf b=D^B\). The binary matrix of \(J(A,B)\) then has the rows
\[
(d^A_r,\ d^B_u+\tau\mathbf1),
\qquad
\tau\in\mathbb F_2,
\]
where \(\tau=1\) for \(s=-1\). The row with \(r=u=1\), \(\tau=0\) vanishes, and so does the first column of the block of \(A\), so this binary matrix is its own normalized matrix for that pivot. Its row space contains \((d^A_r,0)\), \((0,d^B_u)\), and \((0,\mathbf1)\) (take \(u=1,\tau=0\); then \(r=1,\tau=0\); then \(r=u=1,\tau=1\)), and every row is a sum of these. Thus the row space is \((V_A\oplus0)\oplus(0\oplus(V_B\oplus\langle\mathbf1\rangle))\), where \(\mathbf1\notin V_B\) by the first coordinate. Its dimension is \(r_2(A)+r_2(B)+1\).

\proofstep{The generators.} Let \(B\) be a total matrix with \(k=\min(N,M)\le16\). The matrix \(D\) has a zero row and a zero column, so \(r_2(B)\le k-1\). Also \(\log_2(2N)\log_2(2M)\ge(1+\log_2k)^2\), and \(k\le(1+\log_2k)^2\) holds on the intervals of \(k\): for \(k=1\) it reads \(1\le1\); for \(2\le k\le4\) the right side is at least \(4\); for \(5\le k\le8\) it is at least \((1+2)^2=9\); for \(9\le k\le16\) it is at least \((1+3)^2=16\). Hence \(r_2(B)+1\le k\le\log_2(2N)\log_2(2M)\).

\proofstep{Propagation through the operations.} We prove \(r_2+1\le\log_2(2N)\log_2(2M)\) by induction on the number of operations. For the factors put $a_i=\log_2(2N_i)\ge1$ and $b_i=\log_2(2M_i)\ge1$, and write \(a=\log_2(2N)\), \(b=\log_2(2M)\) for the result.

Under Kronecker products, the new parameters are
$a=a_1+a_2-1$ and $b=b_1+b_2-1$. By the identity for Kronecker products and the induction hypothesis, \(r_2+1=(r_2(A)+1)+(r_2(B)+1)-1\le a_1b_1+a_2b_2-1\). The required numerical inequality is
\[
ab-(a_1b_1+a_2b_2-1)
=
(a_1-1)(b_2-1)+(a_2-1)(b_1-1)
\ge0,
\]
where the identity follows by expanding both sides in \(a_i-1\) and \(b_i-1\).

For a signed join the number of rows is $2N_1N_2$ and the number of
columns is $M_1+M_2$. Thus $a=a_1+a_2$ and
$b=\log_2(2(M_1+M_2))\ge\max\{b_1,b_2\}$, giving
$ab\ge a_1b_1+a_2b_2$. The right side is the sum of the inductive
bounds for the two terms of \(r_2(J)+1=(r_2(A)+1)+(r_2(B)+1)\), which is the identity for signed joins.

Transposition interchanges $a,b$; permutations and sign changes
preserve them, and repetitions can only increase them. By the invariance proved in Theorem~\ref{res:6.1}, none of these operations changes \(r_2\). These facts
propagate the bound through every operation generating \(\mathcal G\). Finally \(\operatorname{Ind}S(B)\le r_2(B)\) by Theorem~\ref{res:6.1}.
\end{proof}

\subsection{Chain labels}\label{ssec:6.2}

A learning rule outputs hypotheses, so to bound the list number we attach to every closure of the class, that is, to every intersection of atom sets of concepts, a Boolean hypothesis that extends it. The relevant quantity is then the number of distinct hypotheses along a chain of closures, which can be much smaller than the length of the chain. The main result of this subsection shows that this number bounds the list number, and its proof turns a labeling into an explicit closed cover of the realizable distributions.

We write \((i,y)\) and \(i^y\) for the same labeled point, and call it an atom. A \emph{positive face} is a set \(F\) of atoms, possibly empty, with \(F\subseteq\sigma_c\) for at least one \(c\in C\).

For positive faces, put
\[
c_+(F)=\bigcap\{\sigma_c:\sigma_c\supseteq F\}.
\]
The positive closures are exactly the members of \(P_+^0(B)\) of Section~\ref{ssec:2.4}, the possibly empty intersection \(\bigcap_{c\in C}\sigma_c\) included; \(P_+^*(B)\) omits the empty set. A labeling assigns to every \(R\) in the poset a label \(\ell(R)\in\{\pm1\}^M\) that agrees with every atom of \(R\); labels may be improper, that is, they need not belong to \(C\). Let \(L(\ell)\) be the largest number of distinct labels appearing along a chain of the poset. Then
\[
\chi^*(B)=\min_\ell L(\ell)\ \text{over labelings of }P_+^*(B),
\qquad
\chi^0(B)=\min_\ell L(\ell)\ \text{over labelings of }P_+^0(B).
\]
For a positive closure \(R\) we write \(V(R)=\{c\in C:R\subseteq\sigma_c\}\) for its version space; distinct closures have distinct version spaces, and a strictly increasing chain of closures corresponds to a strictly decreasing chain of version spaces. These posets and numbers depend on the class \(C\) alone, so we also write \(P_+^0(C)\), \(\chi^0(C)\), and so on. Both conventions are needed: the list number is bounded by \(\chi^*\) in Theorem~\ref{res:6.3}, whereas the product bound of Theorem~\ref{res:6.4}(e) requires the empty closure.

The next lemma records the elementary properties of positive closures.

\begin{lemma}[Positive closures]\label{lem:comb-positive-closure}
Let \(F\subseteq G\) be positive faces.
\begin{enumerate}[label=(\alph*),leftmargin=2em]
\item \(F\subseteq c_+(F)\in P_+^0(B)\), \(V(c_+(F))=\{c\in C:F\subseteq\sigma_c\}\), and \(c_+(F)\subseteq c_+(G)\).
\item Every \(R\in P_+^0(B)\) satisfies \(V(R)\ne\varnothing\) and \(R=c_+(R)=\bigcap_{c\in V(R)}\sigma_c\).
\item The map \(V\) is injective on \(P_+^0(B)\), and \(R\subsetneq R'\) implies \(V(R')\subsetneq V(R)\).
\end{enumerate}
\end{lemma}

\begin{proof}
(a) The inclusion \(F\subseteq c_+(F)\) holds by definition, and \(c_+(F)\) is the intersection over the nonempty family \(\{c:F\subseteq\sigma_c\}\), so it belongs to \(P_+^0(B)\). If \(F\subseteq\sigma_c\), then \(\sigma_c\) is one of the intersected sets, so \(c_+(F)\subseteq\sigma_c\); conversely \(c_+(F)\subseteq\sigma_c\) gives \(F\subseteq\sigma_c\). This identifies \(V(c_+(F))\). Since \(\{c:G\subseteq\sigma_c\}\subseteq\{c:F\subseteq\sigma_c\}\), we get \(c_+(F)\subseteq c_+(G)\).

(b) Let \(R=\bigcap_{c\in S}\sigma_c\) with \(\varnothing\ne S\subseteq C\). Then \(S\subseteq V(R)\), so \(c_+(R)=\bigcap_{c\in V(R)}\sigma_c\subseteq\bigcap_{c\in S}\sigma_c=R\). With (a) this gives \(c_+(R)=R\).

(c) By (b), \(R\) is recovered from \(V(R)\). If \(R\subsetneq R'\), then \(V(R')\subseteq V(R)\), and the inclusion is strict by injectivity.
\end{proof}

The next theorem shows that chain labels bound the list number, and it places the chain-label number below the heights of Section~\ref{ssec:6.1}.

\begin{resultenv}{Theorem}{res:6.3}{Chain labels bound list replicability}
For every total \(B\) and every \(\varepsilon\in(0,1/2)\),
\[
\operatorname{lr}(C_B,\varepsilon)
\le\chi^*(B)\le\chi^0(B)\le\chi^*(B)+1,
\]
\[
\operatorname{lr}(C_B,\varepsilon)
\le\chi^*(B)\le h_+^*(B)\le h(B)\le r_2(B)+1.
\]
Consequently, for \(B\in\mathcal G\),
\[
\operatorname{lr}(B,\varepsilon)
\le
\log_2(2N)\log_2(2M).
\]
\end{resultenv}

\begin{proof}
Fix a labeling \(\ell\) of \(P_+^*(B)\) with chain-label bound \(L=L(\ell)=\chi^*(B)\). The poset is nonempty because \(\sigma_c\ne\varnothing\), so \(L\ge1\). The atoms of a realizable distribution, sorted by mass, determine a chain of closures, and the labels along this chain give a point of a complex of dimension at most \(L-1\). A power map then makes the masses of all non-negligible atoms nearly equal, and this forces the label with the maximal barycentric coordinate to be correct on these atoms.

\proofstep{The chain-label complex.} Let \(\mathcal H\subseteq\{\pm1\}^M\) be the set of values of \(\ell\), and let \(K_\ell\) be the simplicial complex on the vertex set \(\mathcal H\) whose simplices are the nonempty subsets of the label sets \(\{\ell(R_1),\ldots,\ell(R_k)\}\) of chains \(R_1\subsetneq\cdots\subsetneq R_k\) of \(P_+^*(B)\). Every simplex has at most \(L\) vertices, so \(\dim K_\ell\le L-1\). We realize \(|K_\ell|\subseteq\mathbb R^{\mathcal H}\) with the basis \((e_h)\), and write \(\beta_h\) for the barycentric coordinates.

\proofstep{The mass-prefix map.} Let \(\nu\in\Delta_C\), let \(v_1>\cdots>v_q>0\) be the distinct positive values of \(\nu\), put \(v_{q+1}=0\), and let \(G_k=\{z:\nu_z\ge v_k\}\). These sets are nonempty, nested, and contained in \(\sigma_c\) for any concept \(c\) realizing \(\nu\). So they are positive faces, and by Lemma~\ref{lem:comb-positive-closure}(a) the sets \(c_+(G_k)\in P_+^*(B)\) increase with \(k\). Define
\[
F_\ell(\nu)=\sum_{k=1}^q|G_k|(v_k-v_{k+1})\,e_{\ell(c_+(G_k))}.
\]
The coefficients are positive. An atom \(z\) with \(\nu_z=v_m\) lies in \(G_k\) exactly for \(k\ge m\), and \(\sum_{k\ge m}(v_k-v_{k+1})=v_m\); summing over \(z\) shows that the coefficients have sum \(\sum_z\nu_z=1\). The labels that occur lie on a chain, so \(F_\ell(\nu)\in|K_\ell|\).

We use the following sorted form. Sort the nonzero atom masses of \(\nu\) as \(\nu_1\ge\cdots\ge\nu_s>0\), with atoms \(z_1,\ldots,z_s\) and ties broken arbitrarily, put \(\nu_{s+1}=0\), and let \(F_j=\{z_1,\ldots,z_j\}\) be the nonempty prefixes. Put
\[
t_j=j(\nu_j-\nu_{j+1}).
\]
Then
\[
\nu=\sum_jt_j\operatorname{Unif}(F_j),
\]
since the mass at \(z_m\) on the right is \(\sum_{j\ge m}t_j/j=\nu_m\). If \(\nu_j>\nu_{j+1}\), then \(F_j\) is one of the sets \(G_k\), with \(|G_k|=j\), \(v_k=\nu_j\), and \(v_{k+1}=\nu_{j+1}\); if \(\nu_j=\nu_{j+1}\), then \(t_j=0\). Therefore
\[
F_\ell(\nu)
=
\sum_jt_j e_{\ell(c_+(F_j))},
\]
for every choice of tie-breaking. The map \(F_\ell\) thus takes values in the finite complex \(K_\ell\), whose simplices are chain-label sets and whose dimension is at most \(L-1\).

\proofstep{Continuity of the mass-prefix map.} When atom masses tie, the coefficients of the ambiguous intermediate prefixes vanish, and this makes the map continuous. To see this, fix \(c\in C\) and an enumeration \(z_1,\ldots,z_M\) of \(\sigma_c\), and consider the closed set of distributions on \(\sigma_c\) with \(\nu_{z_1}\ge\cdots\ge\nu_{z_M}\). On this set, with \(\nu_{z_{M+1}}=0\),
\[
F_\ell(\nu)=\sum_{j=1}^Mj(\nu_{z_j}-\nu_{z_{j+1}})\,e_{\ell(c_+(\{z_1,\ldots,z_j\}))},
\]
because the terms with \(j\) beyond the support have coefficient \(0\). The right side is affine in \(\nu\). Finitely many such closed sets, over all \(c\) and all enumerations, cover \(\Delta_C\), so \(F_\ell\) is continuous by the pasting lemma.

\proofstep{Flattening and the cover.} We flatten the masses by a small power. Put
\[
P_\theta(\mu)\propto\mu^\theta,
\qquad
\gamma=1-\frac1{2L},
\qquad
\theta=\frac{\ln(1/\gamma)}{\ln(2M/\varepsilon)},
\]
where \(\theta>0\) because \(\gamma<1\) and \(2M/\varepsilon>1\), and \(P_\theta(\mu)_z=\mu_z^\theta/\sum_t\mu_t^\theta\) with \(0^\theta=0\). The map \(P_\theta\) preserves supports, so it maps \(\Delta_C\) to itself, and it preserves the order of the masses. It is continuous, because \(x\mapsto x^\theta\) is continuous on \([0,1]\) and the denominator is at least \(M^{-\theta}\): a realizable distribution has at most \(M\) atoms in its support, hence an atom of mass at least \(1/M\).

For each hypothesis \(h\in\mathcal H\), pull back its maximum-barycentric-coordinate region \(D_h\) of Lemma~\ref{res:2.0}:
\[
A_h=(F_\ell\circ P_\theta)^{-1}(D_h).
\]
These sets are closed, cover \(\Delta_C\), and have multiplicity at most \(L\): Lemma~\ref{res:2.0} says that the sets \(D_h\) are closed, cover \(|K_\ell|\), and have multiplicity at most \(\dim K_\ell+1\le L\), and \(F_\ell\circ P_\theta\) is continuous.

\proofstep{The loss estimate.} Let \(\mu\in A_h\) and suppose \(\mu_z\ge\varepsilon/(2M)\). Write the unnormalized flattened masses as \(u_i=\mu_i^\theta\in(0,1]\), for the atoms of the support sorted by decreasing mass as in the sorted form above, and put \(Z=\sum_iu_i\). Since \(x\mapsto x^\theta\) is increasing,
\[
u_z\ge\Bigl(\frac\varepsilon{2M}\Bigr)^\theta
=\exp\bigl(-\theta\ln(2M/\varepsilon)\bigr)=\gamma.
\]
Suppose \(h\) is wrong at \(z\), and let \(z\) be at position \(j\), so that \(u_j\ge\gamma\). If \(k\ge j\), then \(z\in F_k\subseteq c_+(F_k)\), and the label \(\ell(c_+(F_k))\) agrees with \(z\), so it differs from \(h\). Thus no prefix contributing to label \(h\) contains \(z\). The map \(P_\theta\) preserves the order of the masses, so the sorted form applies to \(\nu=P_\theta(\mu)\), with \(\nu_k=u_k/Z\). Together with the summation by parts \(\sum_{k<j}k(u_k-u_{k+1})=\sum_{i<j}\sum_{k=i}^{j-1}(u_k-u_{k+1})=\sum_{i<j}(u_i-u_j)\), this shows that the coordinate \(\beta_h(F_\ell(\nu))\) is at most
\[
\sum_{k<j}\frac{k(u_k-u_{k+1})}{Z}
=
\frac{\sum_{i<j}(u_i-u_j)}{Z}
\le1-\gamma
=
\frac1{2L}.
\]
The inequality holds because \(u_j\ge\gamma\) and \(u_i\le1\) give \(u_i-u_j\le u_i-\gamma\le u_i-\gamma u_i=(1-\gamma)u_i\), and \(\sum_{i<j}u_i\le Z\).

On \(D_h\), however, the maximal coordinate is at least \(1/L\): the point \(F_\ell(\nu)\) has at most \(L\) positive barycentric coordinates, with sum one. Since \(1/(2L)<1/L\), this contradiction shows that every atom of mass at least \(\varepsilon/(2M)\) is labeled correctly by \(h\). The support of \(\mu\) lies in some \(\sigma_{c'}\), so at most \(M\) atoms have positive mass, and each atom where \(h\) errs has mass below \(\varepsilon/(2M)\). Thus loss is at most \(\varepsilon/2\) on \(A_h\), and Lemma~\ref{res:3.3}, applied to the finite closed cover \(\{A_h\}_{h\in\mathcal H}\) of \(\Delta_C\), gives
\[
\operatorname{lr}(C_B,\varepsilon)\le\chi^*(B).
\]

\proofstep{The comparison with heights.} Label every \(R\in P_+^*(B)\) by a concept of \(V(R)\), which is nonempty by Lemma~\ref{lem:comb-positive-closure}(b). This is a labeling with at most one label per member of a chain, so it gives \(\chi^*\le h_+^*\). A nonempty positive closure \(R\) is signed closed: \(R\) is a nonempty face, and the facets containing \(R\) include all \(\sigma_c\) with \(c\in V(R)\), so
\[
R\subseteq c(R)\subseteq c_+(R)=R.
\]
Thus \(P_+^*(B)\) is a subposet of \(\mathcal P(B)\). Hence \(h_+^*\le h\). Theorem~\ref{res:6.1} supplies the inequality \(h\le r_2+1\). The comparison \(\chi^*\le\chi^0\le\chi^*+1\) is Theorem~\ref{res:6.4}(a), whose proof does not use the present theorem. Finally, for $B\in\mathcal G$, Theorem~\ref{res:6.2}
gives $r_2(B)+1\le\log_2(2N)\log_2(2M)$, proving the last claim.
\end{proof}

We compare Theorem~\ref{res:6.3} with the junction height of \cite{BHHLT26}. For finite partial classes, \cite[Theorem 1.3]{BHHLT26} gives a learner whose output hypothesis is likely to lie on a single chain of the junction closure. For a total class \(C\) this yields
\[
\operatorname{lr}(C)\le H(C),
\qquad
H(C)=h_+^0\ \text{(Lemma~\ref{res:2.2})}.
\]
A chain of \(P_+^0(B)\) has at most one member outside \(P_+^*(B)\), so \(h_+^0\le h_+^*+1\le h+1\le r_2+2\).
Thus \cite[Theorem 1.3]{BHHLT26} and Theorems~\ref{res:6.1} and~\ref{res:6.2} give the bound \(\log_2(2N)\log_2(2M)\) on \(\mathcal G\) with an additional \(+1\). Theorem~\ref{res:6.3} removes that \(+1\), replaces the length of a chain by its number of distinct labels, and adds the distinction between proper and improper labels.

\subsection{Exact chain-label values and limitations}\label{ssec:6.3}

We now compute the chain-label numbers of basic classes and record how they behave under operations on classes. The values show how this parameter differs from the list number and the index. On threshold matrices it grows with the size, while the list number and the index stay bounded, so it is not bounded by a function of either. On anchored singletons it equals two, while labelings with values in the class need as many labels as there are rows.

For classes \(C_1,C_2\) on disjoint domains \(X_1,X_2\), the disjoint-domain product \(C_1*C_2\) is the class of all functions on \(X_1\sqcup X_2\) whose restrictions lie in \(C_1\) and \(C_2\). For a class \(C\) we put \(-C=\{-c:c\in C\}\) and \(C^\pm=C\cup(-C)\). For a matrix \(B\), the matrix \(B^\pm\) consists of the rows of \(B\) and their negatives; its row class is \(C_B^\pm\). The full cube \(\{\pm1\}^k\), \(k\ge1\), is the class of all functions on \([k]\). The proper chain-label number \(\chi^{0,\mathrm{proper}}\) is the minimum of \(L(\ell)\) over labelings of \(P_+^0\) with values in the class.

Lower bounds for chain-label numbers come from an adversary that repeatedly extends a closure by an atom contradicting its current label. The next lemma states this argument in the form we use.

\begin{lemma}[Coordinate adversary]\label{lem:comb-coordinate-adversary}
Let \(\ell\) be a labeling of \(\mathcal Q\in\{P_+^0(B),P_+^*(B)\}\), let \(R_0\in\mathcal Q\), and let \(J\) be a set of \(k\) columns such that the restriction of \(V(R_0)\) to \(J\) is all of \(\{\pm1\}^J\). Then \(\mathcal Q\) has a chain \(R_0\subsetneq R_1\subsetneq\cdots\subsetneq R_k\) carrying \(k+1\) distinct labels. In particular \(L(\ell)\ge k+1\).
\end{lemma}

\begin{proof}
Enumerate \(J=\{i_1,\ldots,i_k\}\). We construct \(R_t\) and signs \(y_1,\ldots,y_t\) with
\[
V(R_t)=\{c\in V(R_0):c(i_j)=y_j\text{ for }j\le t\}.
\]
Given \(R_t\) with \(t<k\), put \(h_t=\ell(R_t)\), \(y_{t+1}=-h_t(i_{t+1})\), and \(z=(i_{t+1},y_{t+1})\). Since \(V(R_0)\) restricts onto \(\{\pm1\}^J\), the set \(V(R_t)\) contains concepts with both values at \(i_{t+1}\). So \(R_t\cup\{z\}\subseteq\sigma_c\) for some \(c\in V(R_t)\), and \(z\notin R_t\). Put \(R_{t+1}=c_+(R_t\cup\{z\})\). Lemma~\ref{lem:comb-positive-closure}(a) gives \(R_t\subsetneq R_{t+1}\) and the stated form of \(V(R_{t+1})\). As \(R_{t+1}\supseteq R_0\cup\{z\}\) is nonempty, \(R_{t+1}\in\mathcal Q\). For \(t'>t\), the label \(\ell(R_{t'})\) agrees with \(z\in R_{t'}\), whereas \(h_t(i_{t+1})\ne y_{t+1}\). Thus \(h_0,\ldots,h_k\) are pairwise distinct.
\end{proof}

The next theorem compares the two conventions, computes the threshold values, and gives the general upper bounds, the behavior under products and unions, and the exact values on the basic classes.

\begin{resultenv}{Theorem}{res:6.4}{Chain-label identities and bounds}
In this theorem and in the discussion following it, \(\chi\) without superscript denotes \(\chi^0\), the chain-label number with the empty-intersection convention of Section~\ref{ssec:6.2}; the assertions marked \((*)\) hold verbatim for \(\chi^*\) as well. Let \(T_n\) be the \(n\times n\) threshold matrix with rows and columns indexed by \([n]\), \(c_r(i)=-1\) for \(i<r\) and \(c_r(i)=+1\) for \(i\ge r\); its last column is a common positive anchor. Let \(B_n\) be the anchored singleton class of Section~\ref{ssec:5.4}.

(a) For every total \(B\),
\[
\chi^*(B)\le\chi^0(B)\le\chi^*(B)+1.
\]

(b) For thresholds,
\[
\chi^0(T_n)=\chi^*(T_n)=\lfloor\log_2n\rfloor+1.
\]

(c) \((*)\) Under the error-intersection hypothesis of Theorem~\ref{res:5.2} with parameter \(s\in\mathbb Z_{\ge0}\),
\[
\chi\le s+2,
\]
and for a real parameter the bound is \(\lfloor s\rfloor+2\).

(d) \((*)\) In general,
\[
\chi\le1+\lfloor N/2\rfloor,
\]
and
\[
\chi^0
\le
\min\{M+1,1+\lfloor N/2\rfloor,r_2+2\}.
\]

(e) For disjoint-domain products,
\[
\chi(C_1*C_2)
\ge
\chi(C_1)+\chi(C_2)-1\qquad(*),
\]
and
\[
\chi^0(C_1*C_2)\le\chi^0(C_1)\chi^0(C_2).
\]
The product upper bound fails for \(\chi^*\): \(\chi^*(\{\pm1\}^1)=1\) while \(\chi^*(\{\pm1\}^2)=2\); the correct \(\chi^*\) bound is \(\chi^*(C_1*C_2)\le\chi^0(C_1)\chi^0(C_2)\).

(f) \((*)\) For classes on a common domain,
\[
\chi(C_1\cup C_2)\le\chi(C_1)+\chi(C_2),
\]
and consequently
\[
\chi(C)\le\chi(C^\pm)\le2\chi(C).
\]

(g) For the signed join,
\[
\chi^0(J(A,B))
\in
[\chi^0(A)+\chi^0(B)-1,\ 2\chi^0(A)\chi^0(B)].
\]

(h) Exact values: for \(n\ge3\), \(\chi^0(B_n)=2\) while the proper (concept-labeled) chain-label number of \(B_n\) is \(n\); \(\chi^0(H_{2^m})=m+1\); and \(\chi^0(\{\pm1\}^k)=k+1\), \(\chi^*(\{\pm1\}^k)=k\).
\end{resultenv}

\begin{proof}
\proofstep{Part (a).} If \(\varnothing\notin P_+^0(B)\), the two posets coincide. Otherwise \(\varnothing\) is the least element of \(P_+^0(B)\). Adding the empty-intersection member to a labeling of \(P_+^*(B)\) and giving it an arbitrary label adds at most one distinct label to every chain, because \(\varnothing\) has no atoms to agree with and a chain contains it at most once; conversely, restricting a labeling of \(P_+^0(B)\) to \(P_+^*(B)\) removes that member, and chains of \(P_+^*(B)\) are chains of \(P_+^0(B)\). This gives the two inequalities.

\proofstep{Part (b).} A concept \(c_r\) contains the atom \((i,+)\) if and only if \(i\ge r\), and \((i,-)\) if and only if \(i<r\). Let \(\varnothing\ne S\subseteq[n]\), \(a=\min S\), \(b=\max S\). For \(a\le i<b\) we have \(c_a(i)=+\) and \(c_b(i)=-\), so no atom over \(i\) is common to the rows of \(S\); for \(i<a\) all are negative, and for \(i\ge b\) all are positive. Hence the positive closures of \(T_n\) are the sets
\[
R_{a,b}=\{(i,-):i<a\}\cup\{(i,+):i\ge b\},
\qquad1\le a\le b\le n,
\]
and \(R_{a,b}\subseteq\sigma_{c_r}\) if and only if \(a\le r\le b\), so the version space is \([a,b]\). The common positive anchor makes every intersection nonempty, so \(P_+^0=P_+^*\) and \(\chi^0=\chi^*\). By Lemma~\ref{lem:comb-positive-closure}(b),(c), chains are strictly nested chains of integer intervals, with the closures increasing as the intervals shrink.

For the upper bound, label \(R_{a,b}\) by the threshold \(c_{r(a,b)}\), where \(r(a,b)\) is the point of \([a,b]\) with maximal \(2\)-adic valuation; the maximum is unique, since two distinct integers \(2^vx<2^vy\) of the same valuation \(v\), with \(x,y\) odd, have the multiple \(2^v(x+1)\) of \(2^{v+1}\) between them. Consider a nested chain \(I_1\supsetneq I_2\supsetneq\cdots\) with selected points \(r_k\). The label stays unchanged while the selected point remains in the interval: if \(r_k\in I_{k+1}\), then \(r_k\) still has maximal valuation in \(I_{k+1}\), so \(r_{k+1}=r_k\). Once that point is deleted, the valuation of the new selected point strictly decreases: \(r_{k+1}\in I_k\) differs from \(r_k\), and \(r_k\) is the unique point of maximal valuation in \(I_k\). So the selected point is determined along the chain by its valuation, and at most \(1+\max_{r\le n}v_2(r)=1+\lfloor\log_2n\rfloor\) labels occur, since \(2^{v_2(r)}\le r\le n\).

For the lower bound, fix any labeling \(\ell\). For a label \(h\) consistent with \(R_{a,b}\) put \(r(h)=\min\{i:h(i)=+1\}\); the anchor guarantees that \(r(h)\) exists, and consistency gives \(a\le r(h)\le b\), since \(h(i)=-1\) for \(i<a\) and \(h(b)=+1\). The value \(r(h)\) is a function of \(h\) alone, so distinct values of \(r(h)\) force distinct labels. An adversary starts from \(I_1=[1,n]\). Given \(I_k\) with \(m\ge2\) points and label \(h_k\), it deletes \(r(h_k)\in I_k\) and keeps as \(I_{k+1}\) the larger of the two sides of the deleted point, which has at least \(\lceil(m-1)/2\rceil=\lfloor m/2\rfloor\ge1\) points. It stops at an interval with one point. By induction \(|I_k|\ge\lfloor n/2^{k-1}\rfloor\), so if it stops at \(I_K\), then \(n<2^K\) and \(K\ge1+\lfloor\log_2n\rfloor\). The old label cannot reappear, because for \(k'>k\) we have \(r(h_{k'})\in I_{k'}\subseteq I_{k+1}\), and \(r(h_k)\notin I_{k+1}\). This forces \(1+\lfloor\log_2n\rfloor\) distinct labels on the chain of closures of \(I_1\supsetneq I_2\supsetneq\cdots\supsetneq I_K\).

\proofstep{Part (c).} Negating a column \(i\) renames the atoms \((i,\pm)\) in all \(\sigma_c\) and in all labels, and replaces \(g(i)\) by \(-g(i)\) without changing the sets \(D_c\). It therefore changes neither \(\chi^0\) nor \(\chi^*\) nor the hypothesis, and we can switch the background to all negative. Then \(D_c=\{i:c(i)=+1\}\). If a positive closure \(R\) has at least two compatible concepts, its forced positive set \(I(R)=\{i:(i,+)\in R\}\) is contained in the intersection of two distinct error sets, so \(|I(R)|\le s\). Label such an \(R\) by the hypothesis that is positive on \(I(R)\) and negative elsewhere, and a closure with a single compatible concept by that concept. Both labels are consistent with \(R\): in the first case, an atom \((i,-)\in R\) has \(i\notin I(R)\), because \(R\) lies in some \(\sigma_c\) and so has one atom per column at most.

Along a chain, the version spaces strictly decrease by Lemma~\ref{lem:comb-positive-closure}(c), so at most the top member has a single compatible concept. The sets \(I(R)\) of the other members are nested subsets of size at most \(s\), which take at most \(s+1\) distinct values and yield at most \(s+1\) distinct labels before the version space becomes a singleton, after which at most one further label occurs. The empty intersection, if it is a closure, has \(|V(\varnothing)|=|C|\ge2\) (for \(|C|=1\) the intersection is \(\sigma_c\ne\varnothing\)), so it receives the background label, which coincides with the label of every closure with at least two compatible concepts and \(I(R)=\varnothing\); the bound holds for \(\chi^0\) as well, and restricting the labeling gives it for \(\chi^*\). For a real parameter, replace \(s\) by \(\lfloor s\rfloor\), since the intersection sizes are integers.

The bound is attained: a negative anchor with \(s+1\) free coordinates has pairwise error intersections of size at most \(s\), and Lemma~\ref{lem:comb-coordinate-adversary}, with \(R_0=\bigcap_{c\in C}\sigma_c=\{(0,-)\}\) and \(J\) the free coordinates, forces \(s+2\) labels.

\proofstep{Part (d).} Fix the coordinatewise majority \(g\) of the class, ties broken arbitrarily, and a total order of the concepts. Label \(R\) by \(g\) if \(g\) is consistent with \(R\), and by the first concept of \(V(R)\) otherwise. Along an increasing chain, once \(g\) is inconsistent it stays inconsistent. At the first inconsistency some atom \((i,y)\) of the member \(R\) requires the sign opposite to the majority. At most half of the concepts have \(c(i)=y\), so \(|V(R)|\le\lfloor|C|/2\rfloor\le\lfloor N/2\rfloor\), and from then on all labels are concepts of \(V(R)\), since version spaces decrease; at most that many distinct concept labels occur. This gives \(1+\lfloor N/2\rfloor\) for both conventions.

The bound \(M+1\) holds because a chain of closures is a strictly increasing chain of subsets of a set of \(M\) atoms with at most one sign per column, namely of \(\sigma_c\) for any \(c\) in the version space of its top member; concept labels therefore give at most \(M+1\) labels. The bound \(r_2+2\) follows from \(\chi^0\le\chi^*+1\le h_+^*+1\le h+1\le r_2+2\) (part (a) and Theorems~\ref{res:6.3} and~\ref{res:6.1}).

\proofstep{Part (e).} For disjoint domains, \(\sigma_{(c_1,c_2)}=\sigma_{c_1}\sqcup\sigma_{c_2}\). For nonempty \(S\subseteq C_1\times C_2\) with projections \(S_1,S_2\) this gives
\[
\bigcap_{(c_1,c_2)\in S}\sigma_{(c_1,c_2)}
=
\bigcap_{c_1\in S_1}\sigma_{c_1}\ \sqcup\ \bigcap_{c_2\in S_2}\sigma_{c_2},
\]
and every pair of closures of the factors arises in this way, with \(S=S_1\times S_2\). So the positive-closure poset with empty intersections is the Cartesian product \(P_+^0(C_1*C_2)=P_+^0(C_1)\times P_+^0(C_2)\), and the version space of \((R_1,R_2)\) is \(V(R_1)\times V(R_2)\).

For the lower bound, let \(\ell\) be an arbitrary labeling of the product, and write a label as a pair of blocks. Fix the minimal closure \(R_{2,0}=\bigcap_{c\in C_2}\sigma_c\) of the second factor; projecting the labels of \((R_1,R_{2,0})\) to the first block gives a labeling of \(P_+^0(C_1)\), so some chain in the first factor forces at least \(\chi(C_1)\) distinct first blocks. Let \(c_1\) be a concept in the version space of its top member, and extend the chain to the maximal closure \(\sigma_{c_1}\) (a single concept of \(C_1\)). At least \(\chi(C_1)-1\) of these labels have first block different from \(c_1\).

Next freeze the first block at \(c_1\): a label of \((\sigma_{c_1},R_2)\) agrees with all atoms of \(\sigma_{c_1}\), so its first block equals \(c_1\). The second blocks of these labels form a labeling of \(P_+^0(C_2)\), so we can follow a chain in the second factor forcing at least \(\chi(C_2)\) distinct second blocks; all labels of this stage have first block \(c_1\), so none of them coincides with the \(\chi(C_1)-1\) labels selected before. Since \(R_{2,0}\) is contained in every closure of the second factor, the two stages concatenate to one chain, which carries at least \(\chi(C_1)+\chi(C_2)-1\) labels.

For \(\chi^*\), let \(\ell\) be a labeling of \(P_+^*(C_1*C_2)\); the argument changes in the two stages alone. The first stage runs over the nonempty closures \(R_1\) of the first factor, since \((R_1,R_{2,0})\) is then nonempty, and forces at least \(\chi^*(C_1)\) distinct first blocks. The second stage uses the labeling of the second factor that includes its empty closure, since \((\sigma_{c_1},R_2)\) is nonempty for every \(R_2\in P_+^0(C_2)\). It forces at least \(\chi^0(C_2)\ge\chi^*(C_2)\) labels (part (a)).

For the upper bound, labeling \((R_1,R_2)\) by the pair of optimal labels of the factors is consistent. A chain of the product projects to a chain of each factor, with possible repetitions, so at most \(\chi^0(C_1)\chi^0(C_2)\) distinct labels occur on any chain; the counterexample for \(\chi^*\) in the statement, whose values are part (h), shows that the empty closures cannot be omitted here. Part (a) gives \(\chi^*(C_1*C_2)\le\chi^0(C_1*C_2)\le\chi^0(C_1)\chi^0(C_2)\).

\proofstep{Part (f).} Given labelings \(\ell_1,\ell_2\) of the two classes, optimal for the convention under consideration, label a closure \(R\) of \(C_1\cup C_2\) as follows: if some concept of \(C_1\) is compatible with \(R\), take the closure of \(R\) inside \(C_1\) and use its \(\ell_1\)-label; otherwise some concept of \(C_2\) is compatible with \(R\), and we use the \(\ell_2\)-label of the closure of \(R\) inside \(C_2\). The closure inside \(C_j\) contains \(R\), so these labels are consistent with \(R\), and it is nonempty whenever \(R\) is, so the labeling is defined for both conventions. Along a chain, once \(C_1\) becomes incompatible it stays incompatible, so the chain splits into two consecutive stages, each producing at most \(\chi(C_j)\) labels (closures inside a factor are monotone along the chain, by Lemma~\ref{lem:comb-positive-closure}(a) applied to \(C_j\)). This gives the union bound for both conventions.

The map \(R\mapsto-R\) is an isomorphism of the posets of \(C\) and \(-C\), and negating all labels preserves consistency, so \(\chi(-C)=\chi(C)\). The union bound therefore gives \(\chi(C^\pm)\le2\chi(C)\). For the lower bound, every positive closure of \(C\) is an intersection of atom sets of a nonempty subfamily of \(C\), hence also of \(C^\pm\), so \(P_+^0(C)\subseteq P_+^0(C^\pm)\) as sets of atoms, with the induced order and with consistency of labels preserved; restricting a labeling of \(P_+^0(C^\pm)\), or of \(P_+^*(C^\pm)\), therefore gives a labeling of \(P_+^0(C)\), or of \(P_+^*(C)\), with no more labels on any chain, and \(\chi(C)\le\chi(C^\pm)\).

\proofstep{Part (g).} By the definition of the signed join, the rows of \(J(A,B)\) are all pairs \((c,sd)\) with \(c\in C_A\), \(d\in C_B\), \(s=\pm1\). So the class of \(J(A,B)\) is the disjoint-domain product \(C_A*C_B^\pm\), and \(P_+^0(J(A,B))=P_+^0(A)\times P_+^0(B^\pm)\). The lower bound follows from (e) and \(\chi^0(B^\pm)\ge\chi^0(B)\), and the upper bound from (e) and \(\chi^0(B^\pm)\le2\chi^0(B)\). These two inequalities are part (f).

\proofstep{Part (h): the class \(B_n\), \(n\ge3\).} The columns are \(0,1,\ldots,n\). For \(S\subseteq[n]\) with \(|S|\ge2\) and \(j\in S\), the concepts of \(S\) disagree at \(j\), while for \(j\notin S\) they are all negative at \(j\), and all are positive at \(0\). So the intersection of the atom sets of any two or more concepts is \(\{(0,+)\}\cup\{(j,-):j\notin S\}\) for the corresponding index set \(S\), with version space \(S\). All closures contain \((0,+)\), so \(P_+^0=P_+^*\). The hypothesis \((+,-,\ldots,-)\) is consistent with every closure having at least two compatible concepts, and a chain has at most its top member with a single compatible concept. So labeling every closure with at least two compatible concepts by the hypothesis \((+,-,\ldots,-)\) and every single-concept closure by its concept gives at most two labels on any chain, while \(\operatorname{lr}(B_n)=2\) (Theorem~\ref{res:5.4}, for any \(\varepsilon\in(0,1/2)\)) and Theorem~\ref{res:6.3} give the lower bound.

For proper labels, a labeling must choose some \(i\in S\), since a concept consistent with the closure of \(S\) lies in its version space, and the adversary that deletes that \(i\) forces \(n\) labels: starting from \(S_1=[n]\) and deleting the chosen point each time produces \(S_1\supsetneq\cdots\supsetneq S_n\), and a deleted point is never chosen again. Conversely \(|S|\) strictly decreases along a chain, from at most \(n\) to at least \(1\), so any proper labeling has at most \(n\) labels on a chain.

\proofstep{Part (h): the class \(H_{2^m}\).} The version space of a closure \(R\) is the set of \(r\in\mathbb F_2^m\) with \((-1)^{r\cdot j}=y\) for all \((j,y)\in R\). So the version spaces are nonempty affine subspaces of \(\mathbb F_2^m\) (proof of Theorem~\ref{res:5.5}), and they strictly decrease along a chain by Lemma~\ref{lem:comb-positive-closure}(c). Their dimensions therefore strictly decrease, so a chain has at most \(m+1\) members, and labeling each closure by any compatible row uses at most \(m+1\) labels along a chain. For the lower bound, the column \(0\) is constant, so \(R_0=\bigcap_r\sigma_{c_r}\) is nonempty, and \(V(R_0)\) is the whole class, whose restriction to the columns \(e_1,\ldots,e_m\) is the full cube. Lemma~\ref{lem:comb-coordinate-adversary} with this \(R_0\) and \(J=\{e_1,\ldots,e_m\}\) (choose, at each step, an undetermined coordinate and add the atom opposite to the current label) forces \(m+1\) labels.

\proofstep{Part (h): the full cube.} The closures of \(\{\pm1\}^k\) are all sets of atoms with at most one atom per column, the empty set included, so chains have at most \(k+1\) members, and at most \(k\) in \(P_+^*\). The same adversary starting from the empty closure, with \(R_0=\varnothing\) and \(J=[k]\), forces \(k+1\) labels, and any concept label on each closure achieves \(k+1\). Omitting the empty closure gives \(\chi^*=k\), since Lemma~\ref{lem:comb-coordinate-adversary} with \(R_0=\{(1,+)\}\) and \(J=[k]\setminus\{1\}\) forces \(k\) labels and concept labels use at most \(k\).
\end{proof}

We next compare \(\chi\) with the list number and the index on thresholds. The value of the list number appeared first in~\cite{CMY23}, and we include a direct proof. For the threshold classes $T_n$, $n\ge2$, the vectors
$v_i=(i,1)$ and $w_r=(1,-r+1/2)$ give a strict sign representation
of dimension two: $\langle w_r,v_i\rangle=i-r+1/2$ is positive for $i\ge r$ and negative for $i<r$. After doubling $w_r$, all products $c_r(i)\langle 2w_r,v_i\rangle$ are at least one. Hence $\operatorname{Ind}\le1$ and
$\operatorname{lr}\le2$, by Lemma~\ref{res:5.0} and the odd map recorded after it. Columns $\{1,n\}$ are signed-shattered
and restrict to the anchored one-bit cube: row $1$ restricts to $(+,+)$, the rows $r\ge2$ restrict to $(-,+)$, and the negated rows give the two other patterns. They give respectively
$\operatorname{Ind}\ge1$ and $\operatorname{lr}\ge2$: the first by the signed-shattering estimate of Section~\ref{ssec:2.1}, the second
by Theorem~\ref{res:5.1} with $a=k=1$, because a learner for $T_n$, used on distributions supported on these two columns, with its outputs restricted to them, is a learner for the restricted class. Together with part (b), this proves
\[
\operatorname{lr}=2,
\qquad
\operatorname{Ind}=1,
\qquad
\chi=\lfloor\log_2n\rfloor+1.
\]
There is no bound \(\chi\le F(\operatorname{lr})\) or \(\chi\le F(\operatorname{Ind})\) depending only on the indicated parameter, since these values of \(\chi\) are unbounded.

The most direct label of a closure is the coordinatewise majority of its version space: it is always consistent with the closure, and part (d) uses it for the whole class. On thresholds it needs far more labels than the optimal value of part (b). Indeed, for \(a\le i<b\), the rows of \([a,b]\) have \(i-a+1\) positive and \(b-i\) negative entries in column \(i\), so the majority is positive if and only if \(i\ge\lceil(a+b-1)/2\rceil=\lfloor(a+b)/2\rfloor\); for \(i<a\) all these entries are negative, and for \(i\ge b\) all are positive. Thus the majority label of \(R_{a,b}\) (ties positive) is the threshold \(c_{\lfloor(a+b)/2\rfloor}\). Along \([1,n]\supset[2,n]\supset\cdots\supset[n,n]\) the index \(\lfloor(a+n)/2\rfloor\) runs from \(\lfloor(n+1)/2\rfloor\) to \(n\) in steps of at most one, so at least \(\lfloor n/2\rfloor+1\) distinct thresholds occur.

Part (h) also separates proper from improper labels. On anchored singletons the two numbers are far apart, whereas on Hadamard classes and cubes the optimal labels in the proof are proper. The values are
\[
\chi^0(B_n)=2,
\qquad
\chi^{0,\mathrm{proper}}(B_n)=n,
\]
and
\[
\chi^0(H_{2^m})=m+1,
\qquad
\chi^0(\{\pm1\}^k)=k+1.
\]

\subsection{The sharp two-piece join bound}\label{ssec:6.4}

We return to the index. If the rows of a class differ from a common background on sets with pairwise intersections of size at most \(s\), then Theorem~\ref{res:5.2} gives list number at most \(s+2\), and the inequality \(\operatorname{Ind}\le2\operatorname{lr}-1\) of \cite[Theorem 3.1]{BHHLT26} (Theorem~\ref{res:a}) turns this into \(\operatorname{Ind}\le2s+3\). We show that the index is at most \(2s+2\), and that this is sharp. The argument partitions the face poset into two pieces, constructs an odd map on each piece, and joins the maps. We state the joining step in general, because Section~\ref{ssec:6.5} uses it for three pieces.

For a finite complex \(K\) with free involution, the order-complex piece of a set \(A\) of nonempty faces is \(\Delta(A)\), the full subcomplex of \(\operatorname{sd}K\) on the vertex set \(A\). An empty piece has \(d_i=-1\) and needs no map (Section~\ref{ssec:2.1}).

\begin{resultenv}{Theorem}{res:6.5}{Piecewise join principle}
If the nonempty faces of a finite free complex \(K\) are partitioned into \(\mathbb Z_2\)-invariant pieces \(A_1,\ldots,A_t\), and each corresponding order-complex piece has an odd map of sphere dimension \(d_i\), then
\[
\operatorname{Ind}K
\le
\sum_i(d_i+1)-1.
\]
\end{resultenv}

\begin{proof}
Let \(f_i:|\Delta(A_i)|\to S^{d_i}\) be the given odd maps. Every face chain splits into its subchains in the pieces. Therefore
\[
\operatorname{sd}K
\subseteq
\mathop{*}_i\operatorname{sd}K[A_i],
\qquad
\operatorname{sd}K[A_i]=\Delta(A_i).
\]
Join the given odd maps, weighted by the barycentric masses of the pieces, and normalize. Explicitly, for \(x=\sum_Fx_Fe_F\in|\operatorname{sd}K|\) let \(\alpha_i(x)=\sum_{F\in A_i}x_F\). If \(\alpha_i(x)>0\), the point \(x_i=\alpha_i(x)^{-1}\sum_{F\in A_i}x_Fe_F\) lies in \(|\Delta(A_i)|\), because the support of \(x\) is a chain. Define
\[
G(x)=(G_1(x),\ldots,G_t(x))\in\mathbb R^{\sum_i(d_i+1)},
\quad
G_i(x)=\alpha_i(x)f_i(x_i),
\]
with \(G_i(x)=0\) if \(\alpha_i(x)=0\). Each \(G_i\) is continuous: \(\alpha_i\) and \(x_i\) are continuous where \(\alpha_i>0\), and \(\|G_i(x)\|=\alpha_i(x)\) tends to zero as \(\alpha_i(x)\to0\). At least one mass is positive, since \(\sum_i\alpha_i(x)=1\), so the joined vector never vanishes. The involution permutes the faces and preserves each piece, so \(\alpha_i\) is invariant and \(x_i\) is mapped to its antipode; hence \(G\) is odd. Normalizing and composing with the equivariant homeomorphism \(|K|\cong|\operatorname{sd}K|\) gives an odd map \(|K|\to S^{\sum_i(d_i+1)-1}\).
\end{proof}

Two kinds of faces admit odd maps of low dimension. A face with few vertices of one sign is separated from the origin by a low-degree polynomial on the moment curve indexed by the columns. A face with few compatible signed rows maps to signed combinations of linearly independent vectors on the moment curve indexed by the rows. The next two lemmas construct these maps, which are used on the pieces of Theorems~\ref{res:6.6} and~\ref{res:6.7}.

\begin{lemma}[Column moment curve]\label{lem:comb-column-moment}
For an integer \(s\ge0\), let \(A_s=\{F\in\mathcal F(B):\min(|F^+|,|F^-|)\le s\}\). There is a continuous odd map \(|\Delta(A_s)|\to S^{2s}\).
\end{lemma}

\begin{proof}
Send column \(i\) to \(v_i=(1,i,\ldots,i^{2s})\in\mathbb R^{2s+1}\), and send a face \(F\in A_s\) to
\[
z_F=\frac1{|F|}\Bigl(\sum_{i\in F^+}v_i-\sum_{i\in F^-}v_i\Bigr).
\]
Then \(z_{-F}=-z_F\). Extend affinely on the simplices of \(\Delta(A_s)\), and let \(x=\sum_j\lambda_jF_j\) be a point of a chain \(F_0\subsetneq\cdots\subsetneq F_k\) in \(A_s\). Let \(F=F_k\). If \(|F^+|\le s\), use
\[
P_F(t)=1-2\prod_{i\in F^+}(t-i)^2.
\]
Its degree is at most \(2s\), so \(P_F(t)=\langle a,(1,t,\ldots,t^{2s})\rangle\) for a coefficient vector \(a\) and \(\langle a,v_i\rangle=P_F(i)\). It is \(1\) on \(F^+\), and it is at most \(-1\) on \(F^-\): the integer \(i\in F^-\) is not in \(F^+\), so every factor \((i-i')^2\) with \(i'\in F^+\) is at least \(1\), and for \(F^+=\varnothing\) the polynomial is the constant \(-1\). Thus \(\langle a,bv_i\rangle\ge1\) for every vertex \(i^b\in F\). The same separating polynomial controls affine extensions along the chain: since each \(F_j\subseteq F\), we get \(\langle a,z_{F_j}\rangle\ge1\) for all \(j\), so \(\langle a,\sum_j\lambda_jz_{F_j}\rangle\ge1\) and the affine extension does not vanish. If \(|F^-|\le s\), apply this to \(-F\), which has \(|(-F)^+|\le s\), and use \(z_{-F_j}=-z_{F_j}\). Normalization gives an odd map to \(S^{2s}\).
\end{proof}

For faces with few compatible signed rows, we use the moment curve indexed by the rows.

\begin{lemma}[Row moment curve]\label{lem:comb-row-moment}
Let \(t\ge1\) be an integer, and let \(Q\subseteq\mathcal F(B)\) satisfy \(-Q=Q\) and \(|R(F)|\le t\) for all \(F\in Q\). There is a continuous odd map \(|\Delta(Q)|\to S^{t-1}\).
\end{lemma}

\begin{proof}
Map row \(r\) to \(u_r=(1,r,\ldots,r^{t-1})\), and let
\[
z_F=\frac1{|R(F)|}\sum_{(r,s)\in R(F)}su_r
\]
be the signed average over \(R(F)\). Lemma~\ref{lem:comb-signed-closure}(d) gives \(z_{-F}=-z_F\). Extend affinely, and let \(x=\sum_j\lambda_jF_j\) be a point of a chain \(F_0\subsetneq\cdots\subsetneq F_k\) in \(Q\). By Lemma~\ref{lem:comb-signed-closure}(c), the sets \(R(F_j)\) are nested in \(R(F_0)\). The set \(R(F_0)\) has at most \(t\) members, and it has no antipodal pair, since \(\sigma_r^+\cap\sigma_r^-=\varnothing\) and \(F_0\ne\varnothing\). So its members \((r_1,s_1),\ldots,(r_m,s_m)\), \(m\le t\), have distinct row indices. The image of \(x\) has the form \(\sum_{l=1}^m\gamma_ls_lu_{r_l}\) with \(\gamma_l\ge0\) and \(\sum_l\gamma_l=1\). The vectors \(u_{r_1},\ldots,u_{r_m}\) are linearly independent, because their first \(m\) coordinates form a nonsingular Vandermonde matrix. Hence the image is nonzero, and normalization gives an odd map to \(S^{t-1}\).
\end{proof}

Under the error-intersection hypothesis, these two maps cover all faces, and we obtain the announced bound.

\begin{resultenv}{Theorem}{res:6.6}{Sharp bounded-intersection index bound}
Let \(C\) be a finite total class with background \(g\), and
\[
D_c=\{i:c(i)\ne g(i)\},
\qquad
|D_c\cap D_{c'}|\le s\quad(c\ne c',\ c,c'\in C),
\]
where $s\in\mathbb Z_{\ge0}$. Then
\[
\operatorname{Ind}S(C)\le2s+2.
\]
For the polynomial-graph class over an odd prime field \(\mathbb F_q\), in which the rows are the polynomials of degree at most \(s\), the columns are the points of \(\mathbb F_q^2\), and a row is positive precisely on the graph of its polynomial, with
\[
N=q^{s+1},
\qquad
M=q^2,
\qquad
q\ge2s+4,
\]
the index is exactly \(2s+2\).
\end{resultenv}

\begin{proof}
We first prove the upper bound by joining maps on two pieces, and then find a large signed-shattered set of columns in the polynomial-graph class.

\proofstep{The two pieces.} Switch columns so that \(g\equiv-1\): renaming \(i^+\) and \(i^-\) for the columns with \(g(i)=+1\) is an equivariant isomorphism of sign complexes, and it does not change the sets \(D_c\). Now \(D_c\) is the positive support of \(c\). Index the rows by the concepts, so that distinct rows are distinct concepts. Partition the face poset into
\[
A=\{F:\min(|F^+|,|F^-|)\le s\},
\]
and
\[
D=\{F:|F^+|,|F^-|\ge s+1\}.
\]
Both pieces are invariant under \(F\mapsto-F\), which exchanges \(F^+\) and \(F^-\).

\proofstep{The low piece.} Lemma~\ref{lem:comb-column-moment} gives an odd map to \(S^{2s}\).

\proofstep{The high piece.} For \(F\in D\) there is at most one compatible positive row and at most one compatible negative row. Indeed, a compatible row \((r,+)\) has \(F^+\subseteq D_{c_r}\), and a compatible row \((r,-)\) has \(F^-\subseteq D_{c_r}\) by~\eqref{eq:comb-membership}. Since \(|F^+|,|F^-|\ge s+1\), two distinct compatible rows of the same sign would have error supports sharing \(s+1\) positive errors, contrary to the hypothesis. Hence
\[
1\le|R(F)|\le2.
\]
Lemma~\ref{lem:comb-row-moment} with \(t=2\) and \(Q=D\) gives an odd map to \(S^1\).

\proofstep{The join of the two maps.} Theorem~\ref{res:6.5} with \(d_1=2s\) and \(d_2=1\) now gives \(\operatorname{Ind}S(C)\le(2s+1)+2-1\). With \(f,f'\) the maps of the two pieces and \(\alpha,\beta\) their barycentric masses, the joined map is
\[
\frac{(\alpha f(x_A),\beta f'(x_D))}
{\|(\alpha f(x_A),\beta f'(x_D))\|},
\]
with target \(S^{2s+2}\).

\proofstep{Sharpness.} Take rows to be degree-at-most-\(s\) polynomials and columns to be \((x,y)\in\mathbb F_q^2\), with sign positive precisely on the graph. There are \(q^{s+1}\) rows and \(q^2\) columns. Distinct graphs intersect in at most \(s\) points, since a nonzero polynomial of degree at most \(s\) has at most \(s\) roots; so the upper bound applies with \(g\equiv-1\). Since \(q\ge2s+4\), choose \(2s+3\) distinct field elements \(a_1,\ldots,a_{2s+3}\) and one further element \(b\), and consider the columns \((a_j,a_j^{s+1})\). For a set \(S\) of at most \(s+1\) of the elements \(a_j\) put
\[
p_S(x)=x^{s+1}-\prod_{a\in S}(x-a)\,(x-b)^{s+1-|S|};
\]
the leading terms cancel, so \(\deg p_S\le s\), and on the chosen columns the row of \(p_S\) is positive exactly at \(a_j\in S\) (because \(a_j\ne b\)). Patterns with more than \(s+1\) positive entries have at most \(s+1\) negative entries among the \(2s+3\) columns, so they are the negated rows of the complementary patterns. Hence the \(2s+3\) columns are signed-shattered, giving
\[
\partial\Diamond_{2s+3}\subseteq S(C).
\]
This subcomplex supplies coindex \(2s+2\) by Section~\ref{ssec:2.1}, and \(\operatorname{coInd}\le\operatorname{Ind}\), which meets the upper bound.
\end{proof}

\subsection{A three-piece index bound and its limit on thresholds}\label{ssec:6.5}

The proof of Theorem~\ref{res:6.6} uses the error-intersection hypothesis only on faces with many vertices of both signs, where it bounds the number of compatible signed rows. In a general matrix such a face may have many compatible signed rows, and neither moment curve applies to it. We therefore partition the faces into three pieces. The low piece, with few vertices of one sign, and the high piece, with few compatible signed rows, are treated by the maps of Section~\ref{ssec:6.4}. All other faces form the residual piece: they have many vertices of both signs and many compatible signed rows. Closure preserves this piece, and its contribution to the bound is the height of its closed faces.

The residual piece is where further structure of the matrix enters. If it is empty, the height term vanishes, and for Paley matrices Corollary~\ref{res:8.7} relates this emptiness to the size of switchable monochromatic squares. Threshold matrices show the limit of the bound when the residual piece is treated by its height alone: for them, relative to the all-negative background, the terms of the bound add up to at least the number of rows for all values of the parameters, whereas the index equals one.

\begin{resultenv}{Theorem}{res:6.7}{The three-piece index bound}
For a total \(B\) and nonnegative integers \(s,t\), let
\[
A=\{F:\min(|F^+|,|F^-|)\le s\},
\]
\[
B'=\{F\notin A:|R(F)|\le t\},
\]
\[
E=E_{s,t}=\{F:|F^+|,|F^-|\ge s+1,\ |R(F)|\ge t+1\},
\]
where \(R(F)\) is the set of signed rows compatible with \(F\) and \(F^\pm\) are the vertices of \(F\) with the respective signs; we write \(E_{s,t}\) when the parameters matter. If \(h_E\) is the largest number of members of a chain of signed closed faces (Section~\ref{ssec:6.1}) in \(E\), with \(h_E=0\) when \(E\) is empty, then
\[
\operatorname{Ind}S(B)+1
\le
2s+1+t+h_E.
\]
\end{resultenv}

\begin{proof}
A face outside \(A\) has \(|F^+|,|F^-|\ge s+1\), and it lies in \(B'\) or in \(E\) according to the size of \(R(F)\). So the three sets partition \(\mathcal F(B)\). They are invariant under \(F\mapsto-F\), because \(|(-F)^\pm|=|F^\mp|\) and \(|R(-F)|=|R(F)|\) (Lemma~\ref{lem:comb-signed-closure}(d)). We construct an odd map on each piece and join the three maps.

\proofstep{The low and high pieces.} The first piece maps to \(S^{2s}\) by the column moment curve (Lemma~\ref{lem:comb-column-moment}). For the second, if \(t\ge1\), use the row moment curve
\[
u_r=(1,r,\ldots,r^{t-1}).
\]
Along a chain in \(B'\), the compatible signed rows number at most \(t\) and contain no antipodal pair, so their signed combinations along the chain do not vanish. This gives a map to \(S^{t-1}\) (Lemma~\ref{lem:comb-row-moment} with \(Q=B'\)). If \(t=0\), then \(B'\) is empty because \(R(F)\ne\varnothing\) for every face, and its contribution is \(0=t\).

\proofstep{The residual piece.} Closure preserves compatibility sets and can only add forced vertices: \(R(c(F))=R(F)\) and \(F\subseteq c(F)\) by Lemma~\ref{lem:comb-signed-closure}(a). Thus \(c(E)\subseteq E\). The closure prism homotopy therefore remains inside the residual subposet, even though the latter need not be a downset (Lemma~\ref{lem:comb-closure-retraction} with \(P=E\)). The set \(c(E)\) is exactly the set of signed closed faces in \(E\), and it is invariant under \(F\mapsto-F\). If \(E\) is nonempty, the level map, that is, Lemma~\ref{lem:comb-level-map} with \(Q=c(E)\), gives a map from the order complex of \(c(E)\) to \(S^{h_E-1}\). Composing it with the odd simplicial map \(\Delta(E)\to\Delta(c(E))\) induced by \(c\) gives an odd map \(|\Delta(E)|\to S^{h_E-1}\).

\proofstep{The join of the three maps.} Apply Theorem~\ref{res:6.5} and add the three numbers \(d_i+1\), which are \(2s+1\), \(t\), \(h_E\). The empty-piece conventions give the same formula when \(t=0\) or \(E\) is empty.
\end{proof}

The residual piece \(E\) need not be a downset of \(\mathcal F(B)\), and the proof does not require it: the closure retraction of Lemma~\ref{lem:comb-closure-retraction} on a subposet \(P\) of \(\mathcal F(B)\) requires only \(c(P)\subseteq P\), because the prism homotopy uses the faces of \(P\) and their closures alone. It is equivariant when \(P\) is \(\mathbb Z_2\)-invariant, since \(c\) is equivariant.

Under the hypothesis of Theorem~\ref{res:6.6}, after the column switching used in its proof, the piece \(E_{s,2}\) is empty. So Theorem~\ref{res:6.7} with \(t=2\) gives back the upper bound \(2s+2\).

The sets \(F^\pm\), and therefore the pieces of Theorem~\ref{res:6.7}, depend on the signs of the columns. Negating columns is an equivariant isomorphism of sign complexes, so the theorem can be applied relative to any background \(g\), that is, after negating the columns with \(g(i)=+1\), as in the proof of Theorem~\ref{res:6.6}. For the threshold matrix, the all-negative background means that the theorem is applied to \(T_n\) itself. The error sets are then the intervals \([r,n]\), and the closed faces in the residual piece can be listed explicitly.

\begin{resultenv}{Theorem}{res:6.8}{Threshold matrices and the three-piece bound}
For the threshold matrix \(T_n\) with the all-negative background \(g\equiv-1\) and all integers \(s,t\ge0\),
\[
h_E=\max\{0,n-2s-t-1\}.
\]
Hence, for this background, the three-piece bound of Theorem~\ref{res:6.7} satisfies
\[
2s+1+t+h_E\ \ge\ n,
\]
although
\[
\operatorname{Ind}S(T_n)=1\qquad(n\ge2).
\]
\end{resultenv}

\begin{proof}
We describe the closed faces of the residual piece explicitly and then read off the longest chain.

\proofstep{Compatible rows.} With \(g\equiv-1\), the error set of row \(r\) is its positive support \([r,n]\). Let \(F\) be a face with \(F^+\ne\varnothing\ne F^-\); every face in \(E\) has this form, because \(s+1\ge1\). The unsigned row \(r\) is compatible with \(F\) if and only if \(\max F^-<r\le\min F^+\), and the negated row \(-r\) is compatible if and only if \(\max F^+<r\le\min F^-\). At most one of these two conditions can be satisfiable: together they would give \(\min F^+\le\max F^+<\min F^-\le\max F^-<\min F^+\). So all compatible signed rows of \(F\) carry the same sign. The second case is the image of the first under \(F\mapsto-F\), by Lemma~\ref{lem:comb-signed-closure}(d).

\proofstep{Closed faces in the residual piece.} Consider the first case and put \(a=\max F^-\), \(b=\min F^+\). Then
\[
R(F)=\{r:a<r\le b\},
\qquad
c(F)=\{i^-:i\le a\}\cup\{i^+:i\ge b\},
\]
since \(c_r(i)=-\) for all \(r\in(a,b]\) exactly when \(i\le a\), and \(c_r(i)=+\) for all these \(r\) exactly when \(i\ge b\). Write \(F_{a,b}\) for this closed face. It has \(|F_{a,b}^-|=a\), \(|F_{a,b}^+|=n-b+1\), and \(|R(F_{a,b})|=b-a\). Consequently the closed faces in \(E\) are the sets \(F_{a,b}=c(F)\) with
\[
a\ge s+1,\qquad b\le n-s,\qquad b-a\ge t+1,
\]
together with their antipodes, and \(F_{a,b}\subseteq F_{a',b'}\) if and only if \(a\le a'\) and \(b\ge b'\).

\proofstep{The longest chain.} A face \(F_{a,b}\) is not comparable with an antipode \(-F_{a',b'}\), since the compatible rows of the smaller face would carry both signs. So a chain consists of faces \(F_{a,b}\), or of their antipodes. Along a strict chain the difference \(b-a\) decreases by at least one at every step, from at most \(n-2s-1\) to at least \(t+1\); hence \(h_E\le n-2s-t-1\). The chain \(a=s+1\), \(b=n-s,n-s-1,\ldots,s+t+2\) attains this length. If \(n-2s-1<t+1\), there are no closed faces in \(E\), so by \(c(E)\subseteq E\) the piece \(E\) is empty and \(h_E=0\).

\proofstep{Substitution in the bound.} Substitution in Theorem~\ref{res:6.7} gives \(2s+1+t+h_E=n\) when \(h_E>0\), and \(2s+1+t+h_E=2s+1+t\ge n\) when \(n-2s-t-1\le0\). The value \(\operatorname{Ind}S(T_n)=1\) was proved in the discussion following Theorem~\ref{res:6.4}.
\end{proof}

When \(h_E>0\), the proof gives the equality \(2s+1+t+h_E=n\): increasing \(s\) or \(t\) removes from the height term exactly what it adds to the other two terms. Majority labels behave in the same way: by Section~\ref{ssec:6.3}, the largest number of majority labels on a chain of a threshold class, which bounds \(\operatorname{lr}\) by Theorem~\ref{res:6.3}, is at least \(\lfloor n/2\rfloor+1\). Further partitioning into closure-invariant pieces does not decrease the height term: if the closed faces of \(E\) are divided among several pieces, the members of a chain with \(h_E\) members are distributed among them, so the longest-chain lengths of the pieces have sum at least \(h_E\).

This height term comes from the level map, and a different map on the residual piece can do better. For the all-negative background the compatible rows of every face in the residual piece have the same sign, as shown in the proof above. Their signed average therefore never vanishes and gives a scalar map to \(S^0\) on that piece, because compatible row sets are nested along a chain and so the sign is constant on the chain. For \(s=t=0\), this map, the map of Lemma~\ref{lem:comb-column-moment}, and Theorem~\ref{res:6.5} give back \(\operatorname{Ind}S(T_n)\le1\). The argument depends on the all-negative background: for \(n=3\), \(g=(+,-,-)\) and \(s=t=0\), the face \(\{1^+,2^+\}\) (which becomes \(\{1^-,2^+\}\) after the column switching that makes \(g\equiv-1\)) lies in \(E\), and its compatible signed rows are \(c_1\) and \(-c_3\), of opposite signs.

\subsection{Covers correct on heavy atoms and label reuse}\label{ssec:7.6}

The accuracy of a cover is usually expressed through its loss. The covers built from closure chains have a more combinatorial property: the label of a member containing a distribution is correct at every atom whose mass is at least a fixed fraction of the largest mass (see the loss estimate in the proof of Theorem~\ref{res:6.3}). In this subsection we show that this property characterizes list replicability, and then examine a direct way of producing few labels along chains.

For \(\tau\in(0,1)\), call an atom \(z\) \emph{heavy} for a distribution \(\nu\) if \(\nu_z\ge\tau\max\nu\). The next theorem shows that asking for correct labels on heavy atoms is exactly as strong as accuracy at every error level, for each fixed \(\tau\).

\begin{resultenv}{Theorem}{res:7.8}{Covers correct on heavy atoms}
Let \(\chi_{\mathrm{heavy}}^\tau(C)\) be the minimum multiplicity of a finite closed cover \(\{G_h\}\) of \(\Delta_C\) such that
\[
\nu\in G_h,\quad
\nu_z\ge\tau\max\nu
\quad\Longrightarrow\quad
h\text{ is correct at }z.
\]
For every \(\tau\in(0,1)\),
\[
\chi_{\mathrm{heavy}}^\tau(C)=\operatorname{lr}(C).
\]
\end{resultenv}

\begin{proof}
Let \(M\) be the domain size. A realizable distribution has at most \(M\) atoms of positive mass, one per column, so \(\max\nu\ge1/M\).

\proofstep{The inequality \(\operatorname{lr}(C)\le\chi_{\mathrm{heavy}}^\tau(C)\).} Let \(\{G_h\}\) be a finite closed cover as in the definition, with multiplicity \(L=\chi_{\mathrm{heavy}}^\tau(C)\), and fix \(\varepsilon\in(0,1/2)\). Put
\[
b=(1+\tau)/2\in(\tau,1),
\qquad
\theta=\frac{\ln(1/b)}{\ln(2M/\varepsilon)}>0,
\qquad
P_\theta(\mu)\propto\mu^\theta.
\]
The map \(P_\theta\) preserves supports and is a homeomorphism of \(\Delta_C\), with inverse \(P_{1/\theta}\), so the sets \(A_h=P_\theta^{-1}(G_h)\) form a finite closed cover of \(\Delta_C\) with multiplicity \(L\). If \(\mu_z\ge\varepsilon/(2M)\), the unnormalized powered mass satisfies \(\mu_z^\theta\ge(\varepsilon/(2M))^\theta=b\), while all powered masses are at most \(1\). So \(P_\theta(\mu)_z\ge b\max P_\theta(\mu)\ge\tau\max P_\theta(\mu)\). Thus, for \(\mu\in A_h\), the hypothesis \(h\) is correct at every atom of original mass at least \(\varepsilon/(2M)\), and its errors lie on at most \(M\) atoms of smaller mass. Hence \(\operatorname{loss}_\mu(h)\le\varepsilon/2\) on \(A_h\), and Lemma~\ref{res:3.3} gives \(\operatorname{lr}(C,\varepsilon)\le L\). As \(\operatorname{lr}(C,\varepsilon)\) is nonincreasing in \(\varepsilon\), taking the supremum gives \(\operatorname{lr}(C)\le L\).

\proofstep{The inequality \(\chi_{\mathrm{heavy}}^\tau(C)\le\operatorname{lr}(C)\).} The first inequality of Theorem~\ref{res:b} at error \(\alpha=\tau/(2M)\) gives a finite closed cover \(\{G_h\}\) of \(\Delta_C\) with multiplicity at most \(\operatorname{lr}(C,\alpha)\le\operatorname{lr}(C)\) and \(\operatorname{loss}_\nu(h)\le\tau/(2M)\) on \(G_h\). If \(\nu\in G_h\) and \(\nu_z\ge\tau\max\nu\), then \(\nu_z\ge\tau/M\). An error of \(h\) at \(z\) would give loss at least \(\tau/M>\tau/(2M)\), a contradiction. Thus this cover is correct on heavy atoms.
\end{proof}

The threshold classes \(T_n\), \(n\ge2\), therefore have \(\chi_{\mathrm{heavy}}^\tau=2\) for every \(\tau\in(0,1)\) (they have \(\operatorname{lr}=2\) by the discussion following Theorem~\ref{res:6.4}), while their chain-label number grows logarithmically by Theorem~\ref{res:6.4}(b). So covers correct on heavy atoms need not come from chain labels with the same number of labels.

We now turn to the labels themselves. Whether the chain-label number admits a general bound of order \(\log N\log M\) is open (Section~\ref{sec:open}). If a label may depend on the chain, and not only on the closure, then \(O(\log N)\) labels suffice on every chain. Indeed, along a chain of positive closures the version spaces strictly decrease (Lemma~\ref{lem:comb-positive-closure}(c)), so we can group the closures into layers in which \(|V(R)|\) varies by a factor of at most \(4/3\), and reuse one label within each layer. A chain meets \(O(\log N)\) layers. One label serves a whole layer: if \(R\subseteq R'\) lie in one layer, then \(|V(R')|>\frac34|V(R)|\), so for every atom \(z\) of \(R'\) more than three quarters of the concepts in \(V(R)\) agree with \(z\). Hence a label that takes, wherever possible, the value shared by at least three quarters of \(V(R)\) agrees with every atom of \(R'\). Reusing this label of the first closure of each layer gives \(O(\log N)\) consistent labels on every chain.

The label of a closure obtained in this way depends on which closure of its layer comes first along the chain, and not on the closure alone. So it is not a labeling in the sense of Section~\ref{ssec:6.2}, and the continuity of the mass-prefix map in the proof of Theorem~\ref{res:6.3}, which relies on labels being functions of closures, does not apply. The other steps of that proof, the flattening and the loss estimate, use only that the map is continuous, that each prefix carries a label agreeing with all atoms of the prefix, and that at most \(L\) labels occur on a chain. The label-reuse rule has the last two properties with \(L=O(\log N)\), so continuity is the decisive question for this rule. If the rule were continuous on a matrix, those steps would bound the list number of that matrix by \(O(\log N)\), with an absolute implied constant. Theorem~\ref{res:4.3}(iii) excludes such a bound on all matrices, for large \(N\), so continuity fails on some matrices. The next proposition shows how it fails on a small explicit matrix: a tie between two atoms changes the first closure of a layer, and with it the label of every later closure of that layer.

To describe the rule, define the \emph{size layer} of a positive closure \(R\) with version space \(V(R)\) as the integer \(k\) with \((4/3)^k<|V(R)|\le(4/3)^{k+1}\), and its \(3/4\)-\emph{label} \(h^{3/4}_R\) by \(h^{3/4}_R(x)=-\) if at least three quarters of the concepts in \(V(R)\) are negative at \(x\) and \(h^{3/4}_R(x)=+\) otherwise. The \emph{same-size-layer first-prefix rule} maps a realizable \(\nu\) with mass-sorted prefix closures \(R_1\subseteq R_2\subseteq\cdots\) and barycentric coefficients \(t_j\) (as in the proof of Theorem~\ref{res:6.3}) to \(\sum_jt_je_{h(R_j)}\), where \(h(R_j)\) is the \(3/4\)-label of the first prefix closure lying in the size layer of \(R_j\).

Consider the \(14\times17\) matrix with columns \(a,b,x,d_1,\ldots,d_{14}\), where
\[
c_r(d_j)=+\quad\Longleftrightarrow\quad r=j,
\]
and whose entries on the first three columns are:

\begin{center}\small\begin{tabular}{llll}
\toprule
Rows & \(a\) & \(b\) & \(x\) \\
\midrule
\(1,\ldots,8\) & \(+\) & \(+\) & \(-\) \\
\(9,10,11\) & \(+\) & \(+\) & \(+\) \\
\(12\) & \(+\) & \(-\) & \(-\) \\
\(13\) & \(-\) & \(+\) & \(+\) \\
\(14\) & \(-\) & \(-\) & \(+\) \\
\bottomrule\end{tabular}\end{center}

Let \(R_a,R_b,R_{ab}\) be the positive closures of \(\{(a,+)\}\), \(\{(b,+)\}\) and \(\{(a,+),(b,+)\}\).

\begin{resultenv}{Proposition}{res:7.9}{A discontinuous label-reuse rule}
On this matrix the same-size-layer first-prefix rule is discontinuous. The closures \(R_a,R_b,R_{ab}\) lie in one size layer, since their version spaces have \(11\) or \(12\) elements and
\[
(4/3)^8<11\le12<(4/3)^9,
\]
but along
\[
\nu_t(a^+)=1/2+t,
\qquad
\nu_t(b^+)=1/2-t,
\]
the output jumps at \(t=0\).
\end{resultenv}

\begin{proof}
\proofstep{The three closures.} The compatible concepts are read from the table, since a closure has the same compatible concepts as the atom set it is generated by. So \(V(R_a)\) consists of the rows \(1,\ldots,12\), \(V(R_b)\) of the rows \(1,\ldots,11,13\), and \(V(R_{ab})\) of the rows \(1,\ldots,11\):
\[
|V(R_a)|=|V(R_b)|=12,
\qquad
|V(R_{ab})|=11.
\]
Since \((4/3)^8=65536/6561<10\) and \((4/3)^9>13\), they belong to the same size layer, the eighth.

\proofstep{The two labels.} On \(V(R_a)\), the rows negative at \(x\) are \(1,\ldots,8\) and \(12\), a fraction \(9/12=3/4\), so \(h_a(x)=-\) for \(h_a:=h^{3/4}_{R_a}\). On \(V(R_b)\), they are \(1,\ldots,8\), a fraction \(8/12=2/3\), so \(h_b(x)=+\) for \(h_b:=h^{3/4}_{R_b}\). The atoms of \(R_{ab}\) are \((a,+),(b,+),(d_{12},-),(d_{13},-),(d_{14},-)\). On each of \(V(R_a)\), \(V(R_b)\) at most one row disagrees with such an atom, a fraction at most \(1/12\), so \(h_a\) and \(h_b\) both agree with all atoms of \(R_{ab}\).

\proofstep{The jump.} For \(t>0\) the atom \(a^+\) has the larger mass, the prefix closures are \(R_a\subseteq R_{ab}\), and the first same-size-layer representative of both is \(R_a\). So the image is \(e_{h_a}\). For \(t<0\) the prefix closures are \(R_b\subseteq R_{ab}\), and the image is \(e_{h_b}\). The two labels differ at \(x\). The prefix coefficients
\[
t_1=2|t|,
\qquad
t_2=1-2|t|,
\]
are continuous in \(t\). The rule selects the first representative for the size layer, and this selection causes the jump. Flattening \(\nu\mapsto P_\theta(\nu)\) preserves the order of the two masses on each side of \(t=0\), so the composition of the rule with flattening has the same jump.
\end{proof}

\subsection{Matrices generated from finitely many generators}\label{ssec:7.7}

The lower bound of Theorem~\ref{res:4.3} comes from matrices built from the single matrix \([+]\) by signed joins, transpositions, and repetitions, and Theorem~\ref{res:6.2} gives \(r_2(B)+1\le\log_2(2N)\log_2(2M)\) for all matrices generated from matrices with \(\min(N,M)\le16\). Can other generators, under the same operations, produce an index larger than the order \(\log N\log M\)? The next theorem answers this directly, without Theorem~\ref{thm:main}: for every finite set of generators, the switching binary rank \(r_2\), which bounds both the index and the list number from above, is at most a constant multiple of \(\log_2(2N)\log_2(2M)\), with a constant determined by the generators. This gives information that Theorem~\ref{thm:main} does not, since \(r_2\) can be far larger than the index: \(r_2(T_n)=n-1\), while \(\operatorname{Ind}S(T_n)=1\) for \(n\ge2\). The theorem also shows that Kronecker powers of a generator stay below this order. Along the recursive Hanner family of Theorem~\ref{res:4.3}(i) the corresponding ratio stays bounded away from zero, so it is the signed joins that attain the order.

\begin{resultenv}{Theorem}{res:7.10}{Classes generated from finitely many matrices}
Let \(\mathcal S\) be a finite set of total sign matrices and put \(G(B)=\log_2(2N)\log_2(2M)\) for an \(N\times M\) matrix \(B\). All matrices generated from \(\mathcal S\) by signed joins, Kronecker products, transposition, permutations, sign changes, and repetitions satisfy
\[
r_2(B)\le K_{\mathcal S}(G(B)-1),
\qquad
K_{\mathcal S}=\max\Bigl\{1,\ \max_{B\in\mathcal S,\ G(B)>1}\frac{r_2(B)}{G(B)-1}\Bigr\}.
\]
(If no generator has \(G>1\), the inner maximum is omitted.) Consequently \((\operatorname{Ind}S(B)+1)/G(B)\le K_{\mathcal S}\) on this family, this ratio tends to zero along the Kronecker powers of a fixed generator with \(G>1\), and
\[
\operatorname{lr}(B,\varepsilon)\le r_2(B)+1\le K_{\mathcal S}(G(B)-1)+1
\qquad(\varepsilon\in(0,1/2)).
\]
\end{resultenv}

\begin{proof}
We argue by induction on the number of generation steps. Note that \(G(B)\ge1\) always, and \(K_{\mathcal S}\ge1\).

\proofstep{Generators.} A generator with \(G>1\) satisfies the bound by the definition of \(K_{\mathcal S}\). A generator with \(G=1\) has \(\log_2(2N)=\log_2(2M)=1\), so it is \(1\times1\) and \(r_2=0\).

\proofstep{Kronecker products.} The rank \(r_2\) is additive under these products (Theorem~\ref{res:6.2}). Write \(a_j=\log_2(2N_j)\), \(b_j=\log_2(2M_j)\), so \(a_j,b_j\ge1\). The product has parameters \(a=a_1+a_2-1\) and \(b=b_1+b_2-1\), and the identity from the proof of Theorem~\ref{res:6.2},
\[
(a_1+a_2-1)(b_1+b_2-1)-(a_1b_1+a_2b_2-1)=(a_1-1)(b_2-1)+(a_2-1)(b_1-1)\ge0,
\]
gives \(G-1\ge(G_1-1)+(G_2-1)\). Therefore \(r_2=r_2(B_1)+r_2(B_2)\le K_{\mathcal S}(G_1-1)+K_{\mathcal S}(G_2-1)\le K_{\mathcal S}(G-1)\).

\proofstep{Signed joins.} By Theorem~\ref{res:6.2}, \(r_2\) increases by one. The join has \(2N_1N_2\) rows and \(M_1+M_2\) columns, so \(\log_2(2N)=a_1+a_2\) and \(\log_2(2M)\ge\max(b_1,b_2)\). Hence \(G\ge(a_1+a_2)\max(b_1,b_2)\ge G_1+G_2\), and, as \(K_{\mathcal S}\ge1\),
\[
K_{\mathcal S}(G-1)\ge K_{\mathcal S}(G_1-1)+K_{\mathcal S}(G_2-1)+K_{\mathcal S}\ge r_2(B_1)+r_2(B_2)+1.
\]

\proofstep{Transposition, permutations, sign changes, and repetitions.} Transposition, permutations and sign changes leave \(r_2\) and \(G\) unchanged (Theorem~\ref{res:6.1} and its proof). Repetitions duplicate rows or columns of the switched matrix \(D\), so they leave \(r_2\) unchanged, and they increase \(G\). Hence the bound propagates through every allowed operation.

\proofstep{Consequences.} Theorem~\ref{res:6.1} gives \(\operatorname{Ind}S(B)+1\le r_2(B)+1\le K_{\mathcal S}(G(B)-1)+1\le K_{\mathcal S}G(B)\), since \(K_{\mathcal S}\ge1\). Let \(B\) be a fixed generator of size \(N\times M\) with \(G(B)>1\). Under repeated Kronecker products of \(B\), the \(k\)-th power has \(r_2=kr_2(B)\) and \(G=(1+k\log_2N)(1+k\log_2M)\). If \(N,M\ge2\), then \(G\) grows quadratically in \(k\), and \((\operatorname{Ind}+1)/G\le(kr_2(B)+1)/G\to0\). Otherwise \(N=1\) or \(M=1\), so \(D=0\) and \(r_2=0\), so \(\operatorname{Ind}+1\le1\) for all powers, while \(G\) still tends to infinity. This gives the vanishing ratio. The list bound is Theorem~\ref{res:6.3}, which gives \(\operatorname{lr}(B,\varepsilon)\le r_2(B)+1\) for \(\varepsilon\in(0,1/2)\).
\end{proof}

\section{Conservative labels on the closure poset}\label{app:static}
A chain labeling (Section~\ref{ssec:6.2}) assigns to each closed set of realizable
coordinate constraints
a hypothesis extending it, and Theorem~\ref{res:6.3} converts a uniform bound on
the labels along nested constraints into a learning guarantee. The most direct
way to obtain such a labeling is to fix a priority order on Boolean functions and
to assign to each closed set the first function extending it, so that one finite
list determines the learning rule. A labeling of this kind keeps each label for as
long as it remains consistent, and this section studies the parameter $q(C)$: the
least possible worst-case number of label changes along a chain under this
requirement, which we call conservativity.

The parameter is bounded below by the
Littlestone dimension and, by the main result of this section, above by a function of the Littlestone
dimension alone, with optimal values for Littlestone dimension at most two. Since
$\operatorname{lr}(C)\le q(C)+1$, this gives list bounds that, in contrast with
Theorem~\ref{thm:main}, do not depend on the size of the class; conversely, no
function of the list number bounds $q(C)$, so the parameter records structure that the
list number alone does not determine. We also
obtain exact bounds when the columns have small Littlestone dimension and
characterize when two changes suffice, which refines the general priority-order
construction (see Section~\ref{sec:static} for an overview).

\subsection{Definitions and elementary constructions}
This subsection fixes the language of closures and chains, defines conservative
labels and the parameter $q(C)$, and gives two constructions of conservative
labels. It then collects the properties of the Littlestone dimension used
throughout and proves the elementary bounds on $q$.

A \emph{literal} is a pair $(i,y)\in[n]\times\{\pm1\}$; in the notation of
Section~\ref{ssec:2.4} it is the atom $i^{y}$. A Boolean function $f$
\emph{extends} a set $R$ of literals if $f(i)=y$ for every $(i,y)\in R$, and a
literal $(i,y)$ \emph{opposes} $f$ if $f(i)=-y$.
For a set $A$ of Boolean functions, $A_R$ is the set of members of $A$
extending $R$, and $A_{i,y}=A_{\{(i,y)\}}$ is the \emph{branch} of $A$ at
$(i,y)$.

A \emph{version} is a nonempty subset $V\subseteq C$, and $R(V)$ is the set of
literals common to its rows. A version is \emph{closed} if
$V=C_{R(V)}$.
If $C_S\ne\varnothing$, then $S\subseteq R(C_S)$ and every row of $C_S$
satisfies $R(C_S)$, so $C_{R(C_S)}=C_S$. Thus $C_S$ is closed, and we call
$R(C_S)$ the \emph{closure} of $S$. Taking $S=R(V)$ for a nonempty $V$ gives
$V\subseteq C_S$ and hence $R(C_S)\subseteq R(V)=S\subseteq R(C_S)$. So the
sets $R(V)$ are exactly the closures, and $R=R(C_R)$
for each of them. We call $C_R$ the \emph{version} of $R$; it is the version space
$V(R)$ of Section~\ref{ssec:6.2}.

Ordered by inclusion, the closures form the \emph{closure poset},
which is the poset $P_+^0$ of Section~\ref{ssec:2.4}. For closures,
$R\subseteq R'$ holds exactly when $C_{R'}\subseteq C_R$: the forward implication is
immediate, and the converse follows from the identity $R=R(C_R)$, because a smaller
version has a larger set of common literals.
The closure poset has the least element $R(C)$, which may be
empty. A closure whose version is a singleton $\{c\}$ has a literal at
every coordinate; it is the set $\sigma_c$ of literals of $c$, and $c$ is the
only Boolean function extending it.

A \emph{label} assigns to each closure $R$ a Boolean function
$\ell(R)\in\{\pm1\}^n$, and it is \emph{consistent} if $\ell(R)$ extends $R$ for
every closure $R$, which makes it a labeling of $P_+^0$ in the sense of
Section~\ref{ssec:6.2}. A consistent label is \emph{conservative} if
\begin{equation}\label{eq:conservative}
 R\subseteq R',\quad \ell(R)\text{ extends }R'
 \quad\Longrightarrow\quad\ell(R')=\ell(R).
\end{equation}
Thus a conservative label is replaced only when the growing closure contradicts
it. Arbitrary Boolean labels are permitted; a label, priority-list entry, or rule is
called \emph{proper} when its value is a row of $C$.

A \emph{chain} is a strictly increasing sequence of closures. Its versions
strictly decrease, so only its last member can have a singleton version.
A label \emph{changes} at a consecutive pair of a chain if it differs on the
pair, and a \emph{run} is a maximal block of consecutive members with one
label; the number of runs is the number of changes plus one. Inserting further
closures into a chain does not decrease either number. In particular every
chain may be extended to start at $R(C)$ and to end at a closure with a
singleton version.

Let $q(C)$ be the minimum, over conservative labels, of the maximum
number of label changes along a chain; we call it the \emph{conservative-label
parameter} of $C$. Let $\mu(C)$, the \emph{run number}, be the minimum, over all
consistent labels, of the maximum number of constant runs along a chain. The
empty-prefix parameter $\chi^0(C)$ of Section~\ref{ssec:6.2} counts distinct
labels instead of runs. These definitions give
\begin{equation}\label{eq:static-chain}
 \chi^0(C)\le\mu(C)\le q(C)+1,\qquad
 \operatorname{lr}(C)\le q(C)+1.
\end{equation}
For a fixed consistent label the number of distinct labels on a chain is at
most the number of runs, which gives the first inequality after minimizing.
A conservative label is consistent, and its runs on a chain number its changes
plus one, which gives the second. The third, the learning inequality, is the
chain-label construction of Theorem~\ref{res:6.3}. An
$(\varepsilon,b)$-list-replicable rule is also $(\varepsilon',b)$-list replicable
for every $\varepsilon'\ge\varepsilon$, so $\operatorname{lr}(C,\varepsilon)$ is
nonincreasing in $\varepsilon$ and $\operatorname{lr}(C)$ is its supremum over
$\varepsilon\in(0,1/2)$. For such $\varepsilon$,
Theorem~\ref{res:6.3} gives $\operatorname{lr}(C,\varepsilon)\le\chi^*(C)
\le\chi^0(C)\le q(C)+1$.

A conservative label cannot return to a value it has left, because
that value would extend every intermediate closure, contradicting
\eqref{eq:conservative}. Indeed, if $R_1\subset R_2\subset R_3$ and
$\ell(R_1)=\ell(R_3)=h$, then $h$ extends $R_3\supseteq R_2$, and
\eqref{eq:conservative} for $R_1\subseteq R_2$ gives $\ell(R_2)=h$.

We use two ways of producing conservative labels. The first takes the first
consistent entry of a fixed list of hypotheses.

\begin{lemma}[Priority lists]\label{lem:priority-list}
Let $L$ be a finite list of Boolean functions containing every row of $C$, with
the first occurrence of each function retained and later occurrences deleted. For
a closure $R$ let $\ell_L(R)$ be the first entry of $L$ extending $R$. Then $\ell_L$
is a consistent conservative label. Along a chain the position of $\ell_L$ in $L$
is weakly increasing, and the number of changes equals the number of distinct
labels minus one.
\end{lemma}
\begin{proof}
The version $C_R$ is nonempty, and its rows extend $R$ and occur in $L$, so
$\ell_L(R)$ exists. Let $R\subseteq R'$. Every entry extending $R'$ also extends
$R$, so no entry before $\ell_L(R)$ extends $R'$. Hence the position of
$\ell_L(R')$ is at least that of $\ell_L(R)$, and $\ell_L(R')=\ell_L(R)$ if
$\ell_L(R)$ extends $R'$. This is \eqref{eq:conservative}. The entries of $L$ are
distinct and positions are monotone, so each label occupies one run of a chain.
\end{proof}

The second reverses a fixed background on the coordinates where the closure forces
the opposite sign.

\begin{lemma}[Background completion]\label{lem:background-label}
Fix $g\in\{\pm1\}^n$. For a closure $R$ put $I_g(R)=\{i:(i,-g(i))\in R\}$ and let
$\ell_g(R)$ be $g$ with its signs reversed on $I_g(R)$. Then $\ell_g$ is a consistent
conservative label, $R\subseteq R'$ implies $I_g(R)\subseteq I_g(R')$, and a label
change on a chain is a strict increase of $I_g$.
\end{lemma}
\begin{proof}
A closure is realizable, so it has at most one literal at each coordinate.
A literal $(i,g(i))$ of $R$ therefore has $i\notin I_g(R)$ and agrees with
$\ell_g(R)$, and a literal $(i,-g(i))$ has $i\in I_g(R)$ and also agrees with it.
The inclusion $I_g(R)\subseteq I_g(R')$ holds by definition, and $\ell_g(R)$ is
determined by $I_g(R)$, which gives the statement on changes.
Let $R\subseteq R'$ and suppose that $\ell_g(R)$ extends $R'$. If $i\in I_g(R')$,
then $\ell_g(R)(i)=-g(i)$, so $i\in I_g(R)$. Thus $I_g(R')=I_g(R)$ and
$\ell_g(R')=\ell_g(R)$.
\end{proof}

Our bounds on $q$ are expressed through the Littlestone dimension of the rows and of
the columns.
Throughout this section $d=\operatorname{Ldim}(C)$ and
$e=\operatorname{Ldim}(C^*)$, where $C^*$ is the class of distinct
column functions on the rows; we call $e$ the \emph{dual dimension} of $C$. The
Littlestone dimension is the maximum depth of a complete binary tree of coordinate
tests whose every root-to-leaf answer sequence is realized by a row; such a tree is
called a \emph{mistake tree}. We also call the Littlestone dimension the \emph{rank},
and we set it to $-1$ for the empty class. A
tree of depth zero is a single leaf, so a nonempty class has rank at least zero.
The following elementary properties of the rank are used throughout.

\begin{lemma}[Elementary properties of the rank]\label{lem:rank-facts}
Let $A$ be a nonempty finite set of distinct Boolean functions.
\begin{enumerate}[label=\textup{(\alph*)},leftmargin=2em]
\item If $A'\subseteq A$, then $\operatorname{Ldim}(A')\le\operatorname{Ldim}(A)$.
Also $|A|\ge2^{\operatorname{Ldim}(A)}$, and $\operatorname{Ldim}(A)=0$ exactly when
$|A|=1$.
\item If both branches $A_{i,+},A_{i,-}$ have rank at least $s\ge0$, then
$\operatorname{Ldim}(A)\ge s+1$. If $\operatorname{Ldim}(A)=r\ge1$, then some
coordinate has both branches of rank at least $r-1$.
\item $\operatorname{Ldim}(A)\le1$ exactly when every coordinate has a branch with
at most one member.
\item If $A,A'$ are nonempty classes on disjoint domains, then
$\operatorname{Ldim}(A\times A')=\operatorname{Ldim}(A)+\operatorname{Ldim}(A')$.
\end{enumerate}
\end{lemma}
\begin{proof}
(a) A tree for $A'$ is a tree for $A$. Two leaves of a complete tree of depth $r$
have different answers at the test where their paths separate, so they are realized
by different functions, and there are $2^r$ leaves. If $|A|\ge2$, two members differ
at a coordinate, and that test is a tree of depth one. If $|A|=1$, the bound
$|A|\ge2^{\operatorname{Ldim}(A)}$ gives rank zero.

(b) Truncating a complete tree at depth $s$ leaves a complete tree whose answer
sequences are realized, so each branch has a tree of depth exactly $s$. Attach the
two trees below a test at $i$; a sequence with first answer $y$ is realized in
$A_{i,y}$. Conversely, the two subtrees below the root of a tree of depth $r$ have
depth $r-1$, and their sequences are realized in the two branches at the root test.

(c) If a coordinate has two members on each side, both branches have rank at least
one by (a), and (b) gives rank at least two. Conversely, if the rank is at least two,
(b) gives a coordinate whose two branches have rank at least one, and (a) gives each
of them at least two members.

(d) Attaching a tree for $A'$ at every leaf of a tree for $A$ gives a complete tree
of depth $\operatorname{Ldim}(A)+\operatorname{Ldim}(A')$ whose sequences are realized
by pairs. For the reverse inequality argue by induction on $|A|+|A'|$. If the
product has rank zero, the inequality holds because both ranks on the right are
nonnegative. Otherwise let a complete
tree of depth $D\ge1$, equal to the rank of $A\times A'$, have its root test at a
coordinate $x$ of the
domain of $A$; the other case exchanges the roles of $A$ and $A'$. Both branches
$A_{x,\pm}$ are nonempty and proper, and $(A\times A')_{x,y}=A_{x,y}\times A'$ has
rank at least $D-1$. By induction
$D-1\le\operatorname{Ldim}(A_{x,y})+\operatorname{Ldim}(A')$ for both $y$. By (b) some
branch has rank at most $\operatorname{Ldim}(A)-1$, so
$D\le\operatorname{Ldim}(A)+\operatorname{Ldim}(A')$.
\end{proof}

Additivity of the rank suggests the corresponding question for $q$ on products
over disjoint domains, which requires a description of their closures.

\begin{lemma}[Closures of a product]\label{lem:product-poset}
Let $C_1,C_2$ be classes on disjoint domains, with projections $\pi_1,\pi_2$ of
$C_1\times C_2$. For a nonempty $V\subseteq C_1\times C_2$,
$R(V)=R(\pi_1V)\sqcup R(\pi_2V)$. The closures of $C_1\times C_2$ are exactly the
unions $R_1\sqcup R_2$ of closures of the factors, the version of $R_1\sqcup R_2$ is
$(C_1)_{R_1}\times(C_2)_{R_2}$, and $R_1\sqcup R_2\subseteq R_1'\sqcup R_2'$ holds
exactly when $R_1\subseteq R_1'$ and $R_2\subseteq R_2'$.
\end{lemma}
\begin{proof}
A literal on the domain of $C_j$ is common to $V$ exactly when it is common to
$\pi_jV$, which gives the first formula. Since each $R(\pi_jV)$ is a closure of
$C_j$, every closure of $C_1\times C_2$ is such a union. Conversely, for closures
$R_1,R_2$ the
members of $C_1\times C_2$ extending $R_1\sqcup R_2$ form the product of the two
versions. By the first formula and the identity $R=R(C_R)$ in each factor, the
literal set of this product is
$R_1\sqcup R_2$. The statement on inclusions is read off blockwise.
\end{proof}

The next proposition collects the elementary bounds on $q$. An adversary who
answers against the current label along a mistake tree forces a change at every
step, so $q$ is at least the Littlestone dimension. The upper bounds come from
explicit priority lists and background completions. Relative to a fixed background
$g\in\{\pm1\}^n$, the \emph{error support} of a row $c$ is the set of coordinates
on which $c$ differs from $g$.

\begin{proposition}\label{prop:static-elementary}
For $m$ distinct rows on $n$ coordinates,
\[
 d\le q(C)\le
 \min\{\lfloor m/2\rfloor,n,\dim_{\mathbb F_2}\operatorname{aff}(C)\}.
\]
The affine dimension uses binary encodings of the signs. Also
$\chi^0(C)\ge d+1$. If error
supports relative to a fixed background have pairwise intersection
size at most $s$, then $q(C)\le s+1$. On disjoint domains,
\[
 q(C_1\times C_2)=q(C_1)+q(C_2).
\]
\end{proposition}
\begin{proof}
We prove the lower bounds first, then the upper bounds, and finally additivity on
products.

\proofstep{The lower bounds.}
Let $\ell$ be any consistent label and fix a complete tree of depth $d$ whose
answer sequences are all realized.
We follow this depth-$d$ mistake tree, choosing at each node the answer
opposite to the label of the current closure. Put
$S_0=\varnothing$ and start at the root. At the node of depth $j<d$ reached with
answer set $S_j$, let $R_j=R(C_{S_j})$, let $i_j$ be the test at this node, and put
$y_j=-\ell(R_j)(i_j)$ and $S_{j+1}=S_j\cup\{(i_j,y_j)\}$. Both continuations
remain realizable, so $C_{S_{j+1}}$ is a nonempty proper subset of $C_{S_j}$, and
$R(C)=R_0\subset R_1\subset\cdots\subset R_d$ is a chain.

The new closure excludes the previous label, since
$(i_j,y_j)\in R_{j+1}$ while $\ell(R_j)(i_j)=-y_j$, so
each of the $d$ steps changes it. This opposing literal
also excludes the old label from all subsequent closures, because it lies in
$R_k$ for every $k>j$.
The $d+1$ labels on this chain are therefore distinct, even without
conservativity. Thus $\chi^0(C)\ge d+1$ and $q(C)\ge d$.

\proofstep{The bound $\lfloor m/2\rfloor$.}
A fixed finite priority list containing all rows defines a
conservative label by taking its first consistent entry
(Lemma~\ref{lem:priority-list}). Put the
coordinatewise majority $g$ of $C$ first, with ties broken arbitrarily, followed by
the rows in any fixed order. Let $R$ be the first closure of a chain not extended
by $g$; if there is none, the chain has no change. Then $R$ contains a literal
$(i,y)$ with $g(i)=-y$. This literal contradicts the majority, so it leaves
at most $\lfloor m/2\rfloor$ rows, and $|C_R|\le\lfloor m/2\rfloor$.

The entry $g$
extends no closure $R'\supseteq R$, so the labels from $R$ on are rows of $C$. A row of
$C$ extends such an $R'$ exactly when it lies in $C_{R'}\subseteq C_R$. Each such
label is permanently removed when it changes, so
these labels are distinct members of $C_R$ (Lemma~\ref{lem:priority-list}).
There are consequently at most
$\lfloor m/2\rfloor$ changes, including departure from the majority.

\proofstep{The coordinate and intersection bounds.}
Fix a background $g$ and extend each closure by it
on all unforced coordinates; this is the label $\ell_g$ of
Lemma~\ref{lem:background-label}. A change requires a new forced coordinate
opposite to that background. Each coordinate can be charged once, so $q(C)\le n$.

Suppose now that the error supports $D_c=\{i:c(i)\ne g(i)\}$ satisfy
$|D_c\cap D_{c'}|\le s$ for distinct rows.
If $c\ne c'$ belong to the version of a closure $R$, then
$I_g(R)\subseteq D_c\cap D_{c'}$. So
every nonsingleton
version has at most $s$ common error coordinates. Along a chain the sets $I_g$ are
nested, so there are at most
$s$ changes before reaching a singleton, and at most one further
change on entering it. This proves $q\le s+1$.

\proofstep{The affine bound.}
Encode $+1,-1$ as $0,1\in\mathbb F_2$, write $\beta(c)\in\mathbb F_2^n$ for the
encoding of a row, and let $\operatorname{aff}(V)$ be the affine hull of
$\beta(V)$. The bound holds for any proper priority order. For a set $R$ of
literals let $L_R$ be the affine subspace of those $x\in\mathbb F_2^n$ with
$x_i=\beta(y)$ for all $(i,y)\in R$.
After identifying rows with their encodings, a closed version $V$ has the form
$C\cap L$, with $L=L_{R(V)}$ defined by coordinate
affine equations
over $\mathbb F_2$. If $W\subsetneq V$ are closed versions with the
same affine hull, every equation defining $W$ also holds on $V$, because
$\beta(V)\subseteq\operatorname{aff}(V)=\operatorname{aff}(W)\subseteq L_{R(W)}$.
Then $V\subseteq C_{R(W)}=W$,
contradicting strict inclusion. Thus each strict version change
lowers affine dimension. In particular every label change does so, and the number
of changes on a chain is at most $\dim_{\mathbb F_2}\operatorname{aff}(C)$.

\proofstep{Products.}
The closure poset of a product is the product of the factor
posets (Lemma~\ref{lem:product-poset}). Concatenating optimal labels $\ell_1,\ell_2$
proves the upper bound. The concatenation $\ell_1(R_1)\sqcup\ell_2(R_2)$ extends
$R_1\sqcup R_2$. If it extends
$R_1'\sqcup R_2'\supseteq R_1\sqcup R_2$, then $\ell_j(R_j)$ extends $R_j'$, so
$\ell_j(R_j')=\ell_j(R_j)$ for both $j$, and the concatenation is conservative. Along
a product chain each factor closure is weakly increasing, each change of the
concatenated label is a change of some component, and the components change at most
$q(C_1)$ and $q(C_2)$ times.

For
the reverse inequality, let $\ell$ be a conservative label of the product and
fix the second closure at its least element $R_2^0=R(C_2)$.
The first component $\ell^1(R_1)$ of $\ell(R_1\sqcup R_2^0)$ is a
conservative label on the first factor. Indeed, it extends $R_1$, and if it
extends $R_1'\supseteq R_1$, then
$\ell(R_1\sqcup R_2^0)$ extends $R_1'\sqcup R_2^0$, because its second component
extends $R_2^0$. So \eqref{eq:conservative} applies with the second closure fixed
and gives
$\ell^1(R_1')=\ell^1(R_1)$.
By the definition of $q(C_1)$, the label $\ell^1$ has at least $q(C_1)$ changes on
some chain of the first factor. Extend this
chain to a singleton first version $\{c_1\}$, with closure $\sigma_{c_1}$, and keep
that version fixed.

The second component $\ell^2(R_2)$ of $\ell(\sigma_{c_1}\sqcup R_2)$ now gives a
conservative label on the second
factor. It extends $R_2$, and the first component of $\ell(\sigma_{c_1}\sqcup R_2)$
extends $\sigma_{c_1}$, so if $\ell^2(R_2)$ extends $R_2'\supseteq R_2$, then
$\ell(\sigma_{c_1}\sqcup R_2)$ extends $\sigma_{c_1}\sqcup R_2'$ and
\eqref{eq:conservative} applies. So $\ell^2$ has at least $q(C_2)$ changes on some
chain of the second factor, which we extend to start at $R_2^0$.
The concatenation of the two chains at $\sigma_{c_1}\sqcup R_2^0$ is a chain of the
product on which $\ell$ changes at least $q(C_1)+q(C_2)$ times. This
proves additivity.
\end{proof}

Together with~\eqref{eq:static-chain}, the bound $\lfloor m/2\rfloor$ gives the
list bound
of~\cite[Theorem E]{CCMY24} in the stronger form of a conservative label,
which does not follow from that theorem. The proof of Theorem E also starts
with a majority vote; compare the chain coloring of~\cite[Theorem B]{CCMY24}.

The affine bound is attained on the basic examples: a full $k$-cube and the Sylvester
Hadamard class $H_{2^k}$ both have
$q=k$. Their binary affine dimension is $k$, and $k$ independent
coordinate tests give a complete tree. For the cube the encoded rows are all of
$\mathbb F_2^k$. For the Hadamard class the encoded rows are the image of the linear
injection $r\mapsto(r\cdot j)_{j\in\mathbb F_2^k}$, and the tests $e_1,\ldots,e_k$ read
the free bits of $r$. Hence $k\le d\le q\le k$.

The affine dimension is closely related to the switching rank of
Theorem~\ref{res:6.1}: if a matrix has a constant
column, its binary affine dimension equals its switching rank.
To see this, write $B=(-1)^A$ and use the constant column as
the pivot column in the definition of $r_2$ in Theorem~\ref{res:6.1}, which does not
depend on the pivot. Then $A_{r1}+A_{11}=0$ for all $r$, so the rows of
$D$ are the differences $A_{r,\cdot}+A_{1,\cdot}$ of the encoded rows, and
$\operatorname{rank}_{\mathbb F_2}D$ is the affine dimension.

This identifies the affine bound on the Hanner matrices $A_t$, the lower-bound
family of Theorem~\ref{res:4.3}. The recursion in that theorem gives
them both a constant row and a constant column.
This holds for $A_0=[+]$. If $A$ has a constant row $r$ and a constant column, then
$J(A,A)$ has the constant column coming from the first block and the constant row
$(A_{r,\cdot},A_{r,\cdot})$. Transposition exchanges the two properties, so the
induction passes from $A_t$ to $A_{t+1}=J(A_t,A_t)^{\mathsf T}$. Further $r_2(A_0)=0$.
Since $r_2$ is invariant under transposition,
Theorems~\ref{res:6.1} and~\ref{res:6.2} give $r_2(A_{t+1})=2r_2(A_t)+1$, so
$r_2(A_t)=2^t-1$. By the preceding paragraph this is the affine dimension of $A_t$ and
of $A_t^{\mathsf T}$.
Thus they satisfy $q(A_t),q(A_t^{\mathsf T})\le2^t-1$.

\paragraph{Conservative labels and mistake-driven rules.}
Condition~\eqref{eq:conservative} resembles the requirement, familiar from online
learning, that a learner change its hypothesis only after a mistake. The two
requirements differ, because consistency is tested against all consequences of the
observations and not only against the observations themselves. The anchored
singleton classes show how large the difference can be.

For the anchored singleton class $B_N$, $N\ge2$, of Section~\ref{ssec:5.4},
predict the negative background (with a positive anchor)
until the closure is a singleton, then return its row. This is the label $\ell_g$
of Lemma~\ref{lem:background-label}, and the error supports are the disjoint
singletons $\{i\}$, so Proposition~\ref{prop:static-elementary} with $s=0$ applies.
This class has
$q=1$, the lower bound being $q\ge d\ge1$. A sequence of correct negative observations
can nevertheless
force the last unobserved coordinate to be positive and hence change
the label: the observations $(j,-)$, $j\ne i$, leave the version $\{c_i\}$, whose
closure contains $(i,+)$.

In contrast, consider a rule whose only state is the version: it assigns a Boolean
function $h(V)$ to each version reached, predicts $h(V)(i)$ when coordinate $i$ is
presented, receives the value $y$ of an unknown row at $i$, and passes to $V_{i,y}$.
Suppose it changes its prediction only after an observed mistake, that is,
$h(V_{i,y})=h(V)$ whenever $h(V)(i)=y$. If such a rule has a finite worst-case
mistake bound with repeated observations allowed, then it must
always return a member of the current version. To see this, suppose
its output $h$ is outside $V$. Choose $c\in V$ with inclusion-maximal
agreement set $F=\{i:c(i)=h(i)\}$. Any row of $V$ agreeing with $h$ on $F$
has the same agreement set by maximality, and hence is $c$, since
labels are binary: it equals $h$ on $F$ and $-h$ off $F$. Present the coordinates in
$F$ with the values of $c$. All predictions
are correct, so the output remains $h$, while the version becomes
$\{c\}$. Since $h\notin V$, the functions $h$ and $c$ differ somewhere.
Repeating a coordinate on which $h$ and $c$ differ then
causes infinitely many errors with the same state.

Such a rule is therefore proper. On $m$ anchored
singletons it can be forced to eliminate its selected row
one at a time, for exactly $m-1$ mistakes: while the version has at least two rows
and the selected row is $c_i$, present $i$ with value $-$, and the version loses
exactly $c_i$. Conversely, each mistake of such a rule deletes its selected row, so
there are at most $m-1$ mistakes. The closure parameter
therefore measures a different constraint: on these classes it equals one, whereas
the mistake bound of a rule of this kind grows with the number of rows.

\subsection{A uniform bound from robust low-rank subclasses}
We now bound $q(C)$ by a function of the Littlestone dimension alone. The priority
list is built from the predictions of the standard optimal algorithm on subclasses
of $C$, in order of decreasing rank. A subclass contributes its prediction only if
it cannot be covered by few classes of smaller rank. This robustness forces the
prediction to coincide with that of a member of any small cover of the current
version, which limits the number of labels a chain can use at each rank.

For nonempty $A\subseteq C$, let $h_A$ select, at each coordinate,
the branch of larger Littlestone dimension, with a fixed tie convention.
This is the prediction of the standard optimal algorithm of
Littlestone~\cite{Littlestone88} on the version $A$.
If a coordinate has the constant value $y$ on $A$, the branch $A_{i,-y}$ is empty, so
$h_A(i)=y$; thus $h_A$ extends $R(A)$. We call $h_A$ the \emph{label} of $A$ and
$A_{i,-h_A(i)}$ the \emph{error branch} of $A$ at $i$.
If $\operatorname{Ldim}(A)=r$, then
\begin{equation}\label{eq:soa-drop}
 \operatorname{Ldim}(A_{i,-h_A(i)})\le r-1.
\end{equation}
Otherwise both branches have dimension at least $r$, and attaching
their trees below coordinate $i$ gives dimension at least $r+1$
(Lemma~\ref{lem:rank-facts}(b)).

A rank-$r$ class is \emph{$K$-robust} if it cannot be covered by at
most $K$ subclasses of dimension at most $r-1$. A set covering a part of $A$ may be
intersected with $A$ without raising its rank (Lemma~\ref{lem:rank-facts}(a)), so
the covering sets may be taken inside $A$. A singleton is $K$-robust for every $K$,
since the sets of rank at most $-1$ are empty. For $K\ge2$ a $K$-robust class has a
unique larger-rank branch at every coordinate: two lower-rank
branches would cover it, and two rank-$r$ branches would increase
its dimension. In particular $h_A$ then does not depend on the tie convention.
A $K$-robust class is $(K-1)$-irreducible in the sense
of~\cite[Definition 4.1]{GGKM21}, in the formulation
of~\cite[Definition 4.1]{Lyu26} that is also used in~\cite{BHLT26}:
if agreeing with $h_A$ at $K-1$ coordinates lowered the rank, that
agreement class and the $K-1$ error branches would cover $A$ by $K$
lower-rank classes.

The next lemma is the main property of robust classes. Suppose that few classes of
rank at most $r$ cover the rows satisfying a small set $S$ of literals. Then the
label of any robust rank-$r$ class that agrees with $S$ is already the label of one
of the covering classes.

\begin{lemma}[Robust labels and covers]\label{lem:robust-capture}
Suppose $C_S\subseteq B_1\cup\cdots\cup B_t$, each $B_i$ has
dimension at most $r$, and $A$ is a rank-$r$, $K$-robust subclass of $C$.
If $h_A$ agrees with the literal set $S$ and $|S|+2t\le K$, then
$h_A=h_{B_i}$ for some rank-$r$ component $B_i$.
\end{lemma}
\begin{proof}
The part of $A$ outside $C_S$ is covered by at most $|S|$ error
branches of $h_A$, all of dimension below $r$: a member of $A$ outside $C_S$ violates
some $(i,y)\in S$, so it lies in $A_{i,-y}=A_{i,-h_A(i)}$, and \eqref{eq:soa-drop}
applies. The part of $A$ inside $C_S$ is covered by the sets $A\cap B_i$.

Let $B_i$ have rank $r$. If $h_A\ne h_{B_i}$,
choose a coordinate $x$ on which they differ. Then
\[
 A\cap B_i\subseteq
 A_{x,-h_A(x)}\cup (B_i)_{x,h_A(x)},
\]
because a member of $A\cap B_i$ either errs against $h_A$ at $x$ or agrees with it.
Both sets on the right have dimension below $r$, by
\eqref{eq:soa-drop}; for the second, note that $h_A(x)=-h_{B_i}(x)$.
A component of dimension below $r$ can instead
be used whole. If no rank-$r$ component has the label $h_A$,
these sets cover $A$ with at most $|S|+2t\le K$ lower-rank classes,
contrary to robustness.
\end{proof}

The uniform bound is obtained by arranging the labels of robust subclasses in layers
of decreasing rank. The robustness threshold of each layer is chosen so that the
hypothesis $|S|+2t\le K$ of Lemma~\ref{lem:robust-capture} holds there, and this
choice determines the recurrence in the next theorem.

\begin{theorem}\label{thm:static-recurrence}
Let $a_0=1$ and $a_{j+1}=a_j(2a_j+1)$. A class of dimension $d=0$
has $q=0$. For $d\ge1$ there is a fixed priority order with
\[
 q(C)\le a_{d-1}\le3^{\,2^{d-1}-1}.
\]
In particular the optimal uniform bounds at $d=1,2$ are respectively
one and three.
\end{theorem}
\begin{proof}
The rank-zero case is a singleton (Lemma~\ref{lem:rank-facts}(a)), whose closure
poset has one element, so $q=0$.

\proofstep{The priority list.}
For $d\ge1$, put $h_C$ first; this is the root layer.
For $j=0,\ldots,d-2$, in that order, list the labels of all
$(2a_j+1)$-robust subclasses of rank $r=d-1-j$; this is layer $j$. Finish with all
proper rows; this is layer $d-1$. Use a fixed order within each layer and retain the
first occurrence of every duplicate, so that an entry belongs to the layer of its
first occurrence. This finite list defines one
conservative label for every closure (Lemma~\ref{lem:priority-list}), and the layer of
the label is weakly increasing along a chain.

\proofstep{Reduction to counting labels.}
We bound the number of label changes along an arbitrary chain, extended to start at
$R(C)$. The label of $R(C)$ is $h_C$, since $h_C$ extends $R(C)$. By
Lemma~\ref{lem:priority-list} the number of changes equals the number of distinct
nonroot labels on the chain, that is, labels in layers $0,\ldots,d-1$.
If $h_C$ never fails there is nothing to prove.
For $0\le j\le d-1$, let $R^{[j]}$ be the first closure of the chain whose label
lies in layer $j$ or later, when such a closure exists, and let $c_j$ be the number
of distinct labels of the chain in layers $0,\ldots,j-1$.

\proofstep{The invariant.}
For each $j$ such that $R^{[j]}$ exists, we construct a set $S\subseteq R^{[j]}$ of
literals with $|S|\ge1$
and a
cover of $C_S$ by $t$ components, which are subclasses of $C$ of dimension at most
$r=d-1-j$, and we let
$c=c_j$ count nonroot labels already used. A component of rank exactly $r$ is called
\emph{full-rank}. The invariant is
\[
 P=t+\frac{|S|-1}{2}+c\le a_j.
\]
For $j=0$, the label $h_C$ fails at $R^{[0]}$, so we may choose a literal
$a\in R^{[0]}$ opposing it and start with this single literal in $S$. The class
$C_S=C_a$ has dimension at most $d-1$ by \eqref{eq:soa-drop}, so the invariant holds
initially with $t=|S|=1$ and $c=0$.

\proofstep{Selected labels in layer $j\le d-2$.}
Let $R^{[j]}$ exist, assume the invariant, and put $K=2a_j+1$. Call the labels of
the chain in layer $j$ its \emph{selected labels}.
Every selected label agrees with $S$, because it extends a closure
containing $R^{[j]}\supseteq S$, and
\[
 |S|+2t=2(P-c)+1\le2a_j+1=K.
\]
A selected label has the form $h_A$ with $A$ a $K$-robust rank-$r$ subclass of $C$.
Lemma~\ref{lem:robust-capture}
matches it to the label of a full-rank component. Distinct selected
labels match distinct components, since a component has one label. Let $s$ be the
number of selected labels. If the chain ends in this layer,
its $s$ selected labels give at most $c+s\le c+t\le P$ nonroot
labels and hence that many changes.

\proofstep{Passing to the next layer.}
If the chain leaves the layer, let $R'=R^{[j+1]}$. No entry in the root layer or in
layers $0,\ldots,j$ extends $R'$. Take the $s$ matched components and
all other $K$-robust full-rank components. Let their number be $b$,
so $s\le b\le t$. None of their labels extends $R'$. For
matched components the label has already been used; for the other robust
components it appears in this or an earlier layer. For each of these components, add
to $S$ a literal of $R'$ opposing its label and restrict the component to its error
branch at that coordinate, which has rank at most $r-1$ by \eqref{eq:soa-drop}. Split
each other
full-rank component into at most $K$
lower-rank classes inside the component, which is possible because it is not
$K$-robust. Keep the components of lower rank as they are.

A row of the new $C_{S'}$ lies in some old component, and
then in the piece just assigned to that component, so the new family covers $C_{S'}$;
its members have dimension at most $r-1=d-1-(j+1)$, and $S'\subseteq R'$.
There are $b$ error branches, at most $K$ pieces for each other component, and
$K\ge1$. The new quantities therefore satisfy
\[
 t'\le b+K(t-b)=Kt-(K-1)b,\qquad |S'|\le|S|+b,\qquad c'=c+s.
\]
Consequently
\begin{align*}
 P'&\le Kt-(K-1)b+\frac{|S|+b-1}{2}+c+s\\
   &= Kt+\frac{|S|-1}{2}+c-(K-\tfrac32)b+s\\
   &\le Kt+\frac{|S|-1}{2}+c
    \le KP\le Ka_j=a_{j+1}.
\end{align*}
The successive inequalities use $s\le b$, $K\ge3$, and $(|S|-1)/2+c\ge0$: the first
two give $-(K-\tfrac32)b+s\le-(K-\tfrac52)b\le0$,
and the third gives $(|S|-1)/2+c\le K\bigl((|S|-1)/2+c\bigr)$.
A skipped layer has $s=0$ and the same
argument applies at the same closure $R'=R^{[j]}$. Earlier occurrences of a duplicate
are already
inconsistent at $R'$, so removing duplicates does not affect the argument.

\proofstep{The last layer and the total.}
At the proper layer the covering components have rank zero, so each has one row at
most and $|C_S|\le t$. The labels of this layer are rows of versions contained in
$C_S$, because their closures contain $R^{[d-1]}\supseteq S$, so there are at most
$t$ of them. Thus the total number of
nonroot labels is at most $c+t\le P\le a_{d-1}$. A chain that ends in an earlier
layer $j$ has at most $P\le a_j\le a_{d-1}$ nonroot labels, since the sequence
$(a_j)$ is nondecreasing. The sets $S$ and
the covers were used only to analyze a chain; the priority list itself
was fixed before any chain was chosen.

\proofstep{The estimate and optimality.}
The inequality $a_{j+1}=2a_j^2+a_j\le3a_j^2$ proves the displayed elementary
estimate by induction: $a_0=3^0$, and $a_j\le3^{2^j-1}$ gives
$a_{j+1}\le3\cdot3^{2^{j+1}-2}=3^{2^{j+1}-1}$. At $d=1$ the lower bound in
Proposition~\ref{prop:static-elementary} is attained.

For the dictator class $c_i(x)=x_i$ on $\{\pm1\}^{m}$, every sign pattern on the
$m$ rows occurs as a column, so a set of $s\ge2$ rows can be split into parts of
sizes $\lfloor s/2\rfloor$ and $\lceil s/2\rceil$. Recursive splitting gives a
complete tree of depth $\lfloor\log_2m\rfloor$, and
Lemma~\ref{lem:rank-facts}(a) gives equality for the rank.
At $d=2$,
the six-row dictator class $c_i(x)=x_i$ on $\{\pm1\}^6$ has list
number four by~\cite[Theorem E]{CCMY24},
and Littlestone dimension two:
six leaves forbid depth three, while its columns permit balanced
splits. Equation~\eqref{eq:static-chain} gives $q\ge3$, and $q\le a_1=3$.
\end{proof}

\subsection{Comparison with bounds in terms of Littlestone dimension}\label{sec:static-comparison}
This subsection compares Theorem~\ref{thm:static-recurrence} with the known bounds on
list replicability in terms of the Littlestone dimension, and it derives a second
bound on $q$, smaller for large $d$, from the arguments behind those bounds.

Theorem~\ref{thm:static-recurrence} bounds the conservative-label parameter $q(C)$ of a
finite class in terms of its Littlestone dimension alone, and
\eqref{eq:static-chain} then gives $\operatorname{lr}(C)\le a_{d-1}+1$. Bounds on $\operatorname{lr}$ in terms of
$d$ appeared earlier. The globally stable learner of Bun, Livni, and
Moran~\cite{BLM20} gives $\operatorname{lr}(C)\le2^{2^{O(d)}}$.
Blondal, P.~Hatami, Lalov, and Tretiak~\cite[Theorems 3.1 and 3.2]{BHLT26}
prove, for arbitrary classes on countable domains,
\[
 \operatorname{lr}(C)\le(d+2)^d,\qquad
 \operatorname{lr}(C)\le2^{O(dv^*)},
\]
where $v^*$ is the dual VC dimension, and they note, from the finite-class
extremal result of~\cite{CCMY24}, that $2^{\Omega(d)}$ is necessary.

Iterating the recursion
$\operatorname{lr}(C)\le(d+1)^d+\operatorname{lr}(\widetilde C)$
of~\cite[Section 5.3]{BHLT26}, where $\widetilde C$ is their auxiliary class
of dimension at most $d-1$, without the final simplification made there, gives
$\operatorname{lr}(C)\le\sum_{r=0}^d(r+1)^r$. Indeed, a class of dimension zero has
$\operatorname{lr}=1=(0+1)^0$, and each application of the recursion at a dimension
$r\ge1$ adds $(r+1)^r$ and strictly lowers the dimension. The values of the three
bounds are as follows.
\begin{center}\small
\begin{tabular}{lrrrrr}
\toprule
$d$ & $1$ & $2$ & $3$ & $4$ & $5$\\
\midrule
$a_{d-1}+1$ (Theorem~\ref{thm:static-recurrence}) & $2$ & $4$ & $22$ & $904$ & $1631722$\\
$\sum_{r=0}^d(r+1)^r$ (iterating~\cite[Section 5.3]{BHLT26}) & $3$ & $12$ & $76$ & $701$ & $8477$\\
$(d+2)^d$ (\cite[Theorem 3.1]{BHLT26}) & $3$ & $16$ & $125$ & $1296$ & $16807$\\
\bottomrule
\end{tabular}
\end{center}
The first row uses $a_1=1\cdot3$, $a_2=3\cdot7$, $a_3=21\cdot43$, and
$a_4=903\cdot1807$. The second uses the summands $(r+1)^r=1,2,9,64,625,7776$ for
$r=0,\ldots,5$.

For $d\le3$ the bound of Theorem~\ref{thm:static-recurrence} is the smallest of
these. On finite classes it gives the exact suprema $2$ and $4$ of
$\operatorname{lr}$ at $d=1$ and $d=2$, attained by the anchored one-bit cube
(Theorem~\ref{res:5.1}) and the six-row dictator class, and
$\operatorname{lr}(C)\le22$ at $d=3$, where the fourteen-row dictator class gives
the lower bound $8$. The anchored one-bit cube has two rows, so $d=1$, and
Theorem~\ref{res:5.1} gives $\operatorname{lr}=2$. The fourteen-row dictator class has
$d=\lfloor\log_214\rfloor=3$ by the proof of Theorem~\ref{thm:static-recurrence}
and $\operatorname{lr}=1+\lfloor14/2\rfloor=8$ by~\cite[Theorem E]{CCMY24}.

From $d=4$ on, the argument of~\cite{BHLT26} gives the
smaller list (the stated bound $(d+2)^d$ from $d=5$ on), with $2^{O(d\log d)}$
against a doubly exponential bound for large $d$, and it also applies to
infinite classes.

By~\eqref{eq:static-chain} a bound on $q$ also bounds
$\operatorname{lr}$, but the converse fails: the threshold classes $T_n$,
$n\ge2$, have list number two, while $q(T_n)\ge\chi^0(T_n)-1
=\lfloor\log_2n\rfloor$ by Theorem~\ref{res:6.4}. So a bound on $\operatorname{lr}$ in terms of $d$ does not by itself bound $q$, and
Theorem~\ref{thm:static-recurrence} bounds $q$ directly. Its proof and the proof
of~\cite[Theorem 3.1]{BHLT26} are structurally close: both use standard-optimal-algorithm
hypotheses as centers and descend through lower-rank covers. They differ in
that~\cite{BHLT26} collects these hypotheses into a new class of smaller
dimension, with an irreducibility threshold equal to the current dimension,
whereas the proof above stays among subclasses of $C$, with thresholds $2a_j+1$
that grow; this accounts for the doubly exponential recurrence.

The key combinatorial lemma of~\cite{BHLT26} and their covering descent also
yield conservative labels. We derive this consequence here from their results; closure-poset labels lie outside
the scope of their paper. See also the related tower of Moran and
Yan~\cite[Definition 1 and Lemma 4]{MY26} on private learning.

\begin{proposition}[based on {\cite[Lemmas 5.5 and 5.7]{BHLT26}}]\label{prop:static-soa-tower}
For every nonempty finite class of Littlestone dimension $d$, a fixed
priority order achieves
\[
 q(C)\le\sum_{r=0}^d(r+1)^r-1.
\]
\end{proposition}
\begin{proof}
The definition of $h_A$ and \eqref{eq:soa-drop} apply verbatim to any finite
set of Boolean functions.
A class $G$ of positive rank is \emph{$k$-irreducible} if, for all coordinates
$x_1,\ldots,x_k$, the members of $G$ agreeing with $h_G$ at every $x_i$ have
the same rank as $G$. If $G$ is not $k$-irreducible, those members and the $k$
error branches of $h_G$ at the $x_i$ cover $G$ by $k+1$ classes of smaller
rank, by~\eqref{eq:soa-drop}: a member either agrees with $h_G$ at every $x_i$ or
lies in an error branch.

\proofstep{The tower.}
Put $\mathcal T_d=C$ and, for $r=d,d-1,\ldots,1$,
\[
 \mathcal T_{r-1}=\{h_G:G\subseteq\mathcal T_r,\ \operatorname{Ldim}(G)\ge1,\
 G\text{ is $r$-irreducible}\},
\]
where the members of $\mathcal T_r$ may be arbitrary Boolean functions.
By~\cite[Lemma 5.5]{BHLT26} and descending induction,
$\operatorname{Ldim}(\mathcal T_r)\le r$. We use that lemma in this form: if
$\operatorname{Ldim}(\mathcal T_r)\le r$ and $r\ge1$, then
$\operatorname{Ldim}(\mathcal T_{r-1})\le r-1$. The argument is as follows.
A complete tree of depth $r$ for $\mathcal T_{r-1}$ assigns to each leaf a function
$h_G$ whose values at the tests $x_1,\ldots,x_r$ on the path are the answers. By
irreducibility the members of $G$ with these values have positive rank, so two of them
differ at a coordinate. A test at that coordinate below each leaf gives a complete
tree of depth $r+1$ for $\mathcal T_r$, a contradiction.

An $r$-irreducible class with $r\ge1$
has a unique larger-rank branch at every coordinate $x$: taking
$x_1=\cdots=x_r=x$ shows that the branch selected by $h_G$ has full rank, and the other
branch has smaller rank by \eqref{eq:soa-drop}. So the sets $\mathcal T_r$ do not
depend
on the tie convention. Further $|\mathcal T_0|\le1$, by
Lemma~\ref{lem:rank-facts}(a). Use the priority list
$\mathcal T_0,\mathcal T_1,\ldots,\mathcal T_d$, with fixed orders
and first occurrences retained, and call the entries that first occur in
$\mathcal T_r$ the layer of $\mathcal T_r$. The list contains all rows of $C$, so its
first
consistent entry defines a conservative label on the closure poset
(Lemma~\ref{lem:priority-list}).

\proofstep{Counting the labels of one layer.}
Fix a chain and $r\ge1$, suppose some closure of the chain has its label in
the layer of $\mathcal T_r$, and let $R$ be the first such closure. No entry of
$\mathcal T_{r-1}$ extends $R$. Let $G\subseteq(\mathcal T_r)_R$ have positive
rank. All members of $G$ extend $R$, so the branch of $G$ opposite to a literal
of $R$ is empty and $h_G$ extends $R$. Hence $G$ is not $r$-irreducible, since
otherwise $h_G\in\mathcal T_{r-1}$ would extend $R$. So $G$
has a cover by $r+1$ subclasses of smaller rank, again contained in
$(\mathcal T_r)_R$.

Iterating from rank at most $r$
down to singletons, as in~\cite[Section 4.1]{Lyu26} and the proof
of~\cite[Lemma 5.7]{BHLT26}, gives $|(\mathcal T_r)_R|\le(r+1)^r$. Indeed, a
subclass of rank zero has one member, and a subclass of rank $u\ge1$ has at most
$r+1$ times the maximal size of a subclass of rank at most $u-1$, so a subclass of
rank $u$ has at most $(r+1)^u$ members, and
$\operatorname{Ldim}\bigl((\mathcal T_r)_R\bigr)\le r$.

Every subsequent
label of the chain from this layer extends $R$, so it lies in
$(\mathcal T_r)_R$, and a conservative label does not return to an earlier
value. The chain therefore uses at most $(r+1)^r$ labels from this layer, and at
most one label from $\mathcal T_0$. Summing over the layers bounds the number of
labels by $\sum_{r=0}^d(r+1)^r$, and the number of changes is one less.
\end{proof}

At $d=1,\ldots,5$ this bound has the values $2$, $11$, $75$, $700$, $8476$,
and $a_{d-1}$ has the values $1$, $3$, $21$, $903$, $1631721$. So
Theorem~\ref{thm:static-recurrence} gives the smaller bound on $q$ for $d\le3$
and is optimal for $d\le2$, and Proposition~\ref{prop:static-soa-tower} gives the
smaller bound for $d\ge4$, with dependence $2^{O(d\log d)}$.

The last two claims follow from these values at $d=4$ and in general from two
estimates. First, $\sum_{r=0}^d(r+1)^r\le\sum_{r=0}^d(d+1)^r\le(d+1)^{d+1}$
for $d\ge1$, by the geometric sum. Second, for $d\ge5$ induction
gives $a_{d-1}\ge2^{2^{d-1}}\ge(d+1)^{d+1}$. The first of these inequalities holds at
$d=5$ and is
preserved by $a_{j+1}\ge a_j^2$. The second holds at $d=5$, where $6^6<2^{16}$. When $d$
increases by one its left side is multiplied by $2^{2^{d-1}}\ge(d+1)^{d+1}$, and its
right side, with $\mathrm{e}$ denoting Euler's number, by
$(d+2)^{d+2}/(d+1)^{d+1}=(d+2)\bigl(1+\tfrac1{d+1}\bigr)^{d+1}\le\mathrm{e}(d+2)\le(d+1)^{d+1}$,
so it is preserved. Also $(d+2)^d\le(d+1)^{d+1}\le a_{d-1}$ for $d\ge5$, because
$\bigl(1+\tfrac1{d+1}\bigr)^d<\mathrm{e}<d+1$; this is the comparison with $(d+2)^d$
made above.

Through~\eqref{eq:static-chain}, Proposition~\ref{prop:static-soa-tower} recovers
on finite classes the numerical bound obtained by
iterating~\cite[Section 5.3]{BHLT26}, without using their transfer lemma.

\subsection{Exact low-dual-dimension bounds}
Theorem~\ref{thm:static-recurrence} uses only the Littlestone dimension of the rows.
When the columns also have small Littlestone dimension, the rows are confined to
simple configurations around explicit centers, and the first consistent entry of a
short list already gives sharp bounds. The next two theorems supply these short
priority orders for dual dimension at most one and at most two, and the resulting
uniform bounds depend on neither the Littlestone dimension nor the size of the
class.

Identical columns may be merged
for the construction and their predictions copied back.
To this end, keep one representative of each set of identical columns, which gives a
class $\bar C$ on $t\le n$ columns with $\bar C^*=C^*$. Restriction is a bijection
$C\to\bar C$, and two literals $(i,y),(i',y)$ at identical columns have the same
branch. Hence restriction of literal sets identifies the closure posets of $C$ and
$\bar C$, trees transfer between the two classes, and
$\operatorname{Ldim}(\bar C)=\operatorname{Ldim}(C)$. Let $\hat f$ be the copy of a
function $f$ on the representatives, with equal values at identical columns. Then
$f$ extends a closure of $\bar C$ exactly when $\hat f$ extends the corresponding
closure of $C$. So a conservative label or a priority list for $\bar C$ is copied to
one for $C$ with the same changes on corresponding chains, and
$q(C)\le q(\bar C)$. We also use that negating all signs maps closures to closures,
labels to labels, and trees to trees, and preserves $q$, $d$, and $e$.

For dual dimension at most one, the Littlestone dimension determines $q$ completely,
and it is itself at most three.

\begin{theorem}\label{thm:dual-one}
If $e\le1$, then $q(C)=d\le3$. All three positive values occur.
\end{theorem}
\begin{proof}
After merging columns, let their number be $t$. By the preceding reduction we may
assume in the construction that the columns of $C$ are distinct.

\proofstep{Two stars.}
A row with at least
two positive and two negative columns would split the dual class
into two subclasses each containing two distinct functions. Each
can be split again, giving a depth-two dual tree
(Lemma~\ref{lem:rank-facts}(b)), contrary to $e\le1$. Thus every row
has at most one minority sign.

Suppose $t\ge3$, let $p_0,n_0$ be
the constant positive and negative vectors, respectively, and let
$p_i,n_i$ be their single-flip neighbors: $p_i$ is negative exactly at $i$, and $n_i$
is positive exactly at $i$. For $t\ge3$ the vector $p_i$ has at least two positive
entries and $n_j$ at most one. Put
$P=C\cap\{p_0,p_1,\ldots,p_t\}$ and
$N=C\cap\{n_0,n_1,\ldots,n_t\}$, so that $C=P\sqcup N$ lies in the
union of two disjoint stars, with centers $p_0$ and $n_0$.
The constant vectors may be used as labels even when absent from $C$.
Let $I=\{i\in[t]:p_i\in C\}$ and $J=\{j\in[t]:n_j\in C\}$.
Two facts are used repeatedly. A negative literal at $i$ leaves at most $p_i$ in $P$,
and all of $N\setminus\{n_i\}$ in $N$. A positive literal at $j$ leaves at most $n_j$
in $N$, and all of $P\setminus\{p_j\}$ in $P$.

\proofstep{Three changes.}
The priority order $p_0,n_0$, followed by all proper rows, has at
most three changes. Once both constant labels fail, the closure contains
a negative
literal at some $i$ and a positive literal at some $j$, with $j\ne i$ by consistency,
and these leave at most
$p_i,n_j$. So a chain has at most four labels. Hence $q\le3$, and
Proposition~\ref{prop:static-elementary} gives $d\le q\le3$.
If $d\le1$, Theorem~\ref{thm:static-recurrence}
already gives equality, and if $d=3$, the inequality $d\le q\le3$ does.
Suppose $d=2$. It remains to find a list with at most two changes.

\proofstep{Structure when $d=2$.}
If $|P|,|N|\ge4$ and $i\in I\cap J$, the negative branch at $i$
contains $p_i$ and at least three rows of $N\setminus\{n_i\}$.
Among the latter at most one is $n_0$; choose a unit row $n_j$, so that $j\ne i$. At
$j$, the positive side
contains $p_i,n_j$ and the negative side at least two rows of
$N\setminus\{n_i,n_j\}$. That
branch has dimension at least two by Lemma~\ref{lem:rank-facts}(a),(b).
The positive branch at $i$ has
the same property after negating all signs, which exchanges $p_k$ with $n_k$. This
would give $d\ge3$.
Therefore either one star has at most three rows or $I\cap J$ is
empty.

\proofstep{A star with at most three rows.}
In the first case negate all signs if necessary to assume $|N|\le3$.
Let $h_J$ be positive exactly on $J$, and let $\gamma_j$ be positive
exactly on $J\setminus\{j\}$. Use the order
\[
 h_J,\quad p_0,\quad (\gamma_j:j\in J),\quad C,
\]
with fixed orders and duplicates removed. Let $R$ be the first closure of a chain
not extended by $h_J$. Then $R$ contains a positive literal outside $J$ or a negative
literal inside $J$, and the same holds for later closures $R'$.

If $h_J$ fails on a
positive literal outside $J$, only $P$ survives, since $n_k$ is positive only at $k$;
every $\gamma_j$
also fails there. The label $p_0$ extends every closed version that is a nonsingleton
subset of $P$, because a common negative literal at $i$ would leave at most
$p_i$, and its failure leaves a singleton. So the labels from $R$ on are $p_0$ and at
most the one row of a singleton version.

If $h_J$ fails on a negative literal at $j\in J$, the version is in
\[
 B_j=(N\setminus\{n_j\})\cup(\{p_j\}\cap C).
\]
At each coordinate at most one row of $B_j$ disagrees with
$\gamma_j$. At $j$ there are none, since every row of $B_j$ is negative at $j$.
Outside $J$ only $p_j$ can
disagree, since $\gamma_j$ is negative there and the rows of $N$ are negative outside
$J$. At $i\in J\setminus\{j\}$ the value of $\gamma_j$ is positive, $p_j$ and $n_i$
agree with it, and the possible disagreements
are in $N\setminus\{n_i,n_j\}$, which has size at most one.
Thus $\gamma_j$ extends every nonsingleton closed version inside
$B_j$, because a literal of its closure opposing $\gamma_j$ would make every row of
the version disagree at that coordinate. Its failure leaves one row. The earlier
entries
$p_0$ and all
$\gamma_k$ for $k\ne j$ are already excluded at $j$, being positive there.

Every chain
therefore uses at most a root, a star center, and a singleton label, which is at
most two changes.

\proofstep{Disjoint index sets.}
In the other case $I\cap J=\varnothing$, use
$h_J,p_0,n_0,C$. Again let $R$ be the first closure of a chain not extended by $h_J$.
A positive literal of $R$ outside $J$ leaves
only $P$; a negative literal of $R$ inside $J$ leaves only $N$, since $p_j\notin C$
for $j\in J$, and it excludes $p_0$.
A nonsingleton version inside one star is extended by the constant center of that
star, by the two facts above.
After its constant center fails, either star has a singleton
version. This again gives two changes.

\proofstep{Small $t$ and attainability.}
When $t\le2$, the coordinate bound of Proposition~\ref{prop:static-elementary} gives
$q(C)\le q(\bar C)\le t\le2$, and this together with $q\ge d$ and the $d\le1$ case
suffices.
Finally, the full cubes of dimensions one, two, and three have
$q=d$ by the example after Proposition~\ref{prop:static-elementary} and dual
dimension at most one, by Lemma~\ref{lem:rank-facts}(a), since they have at most
three column functions. This proves
attainability.
\end{proof}

For dual dimension at most two we obtain a uniform bound on both $q$ and $d$: a list
of polynomial length gives at most seven changes, and the seven-dimensional cube
shows that this is sharp.

\begin{theorem}\label{thm:dual-two}
For classes with $e\le2$, the sharp uniform bounds are
\[
 \sup q(C)=\sup d=7.
\]
A fixed priority list of length $O(n^2+m)$ achieves $q\le7$.
Throughout $e\le2$, one also has $q(C)\le d(e+1)$.
\end{theorem}
\begin{proof}
We describe the structure of the columns, then the list, and then count its label
changes. Merge identical
columns and let their number be $t\le n$; a list for the merged class is copied back.

\proofstep{Orienting the rows.}
A class of distinct binary functions has dimension at most one
exactly when every test has a side of size at most one
(Lemma~\ref{lem:rank-facts}(c)). Since the
dual has dimension at most two, each row has a sign side of dual
dimension at most one: otherwise both sides have dual rank at least two, and
Lemma~\ref{lem:rank-facts}(b) gives dual rank three. Call that side $S_c$, and call
the other
sign its background; if both sides qualify, choose one. For any two chosen sides
$S=S_c$, $T=S_{c'}$,
\begin{equation}\label{eq:dual-two-sets}
 |S\cap T|\le1\quad\text{or}\quad
 |S\setminus T|\le1\text{ and }|T\setminus S|\le1.
\end{equation}
Indeed the row $c'$ splits the column set $S$ into $S\cap T$ and $S\setminus T$, so
the rank-one criterion on $S$ says
$\min(|S\cap T|,|S\setminus T|)\le1$, and the criterion on $T$
gives its symmetric counterpart $\min(|S\cap T|,|T\setminus S|)\le1$.
If $|S\cap T|\ge2$, these give both inequalities on the right.
Conversely these conditions ensure
that the chosen side of every row has dual dimension at most one,
so no row can root a depth-three dual tree.

Let $\mathcal F$ be the distinct sets $S_c$, viewed as positive
supports: a set $X$ is identified with the sign vector positive exactly on $X$. The
original class is contained in
$\mathcal F\cup(-\mathcal F)$, since a row with negative background is the vector of
$S_c$ and a row with positive background is the negative of that vector.

\proofstep{Radius-one balls.}
Call $S\setminus T$ and $T\setminus S$ the directed differences of two sets.
A family with both directed differences of size at most one between every
pair lies in a Hamming ball of radius one, and we call the center of such a ball a
center of the family. To prove this, note first that
two members satisfy $|S|-|T|=|S\setminus T|-|T\setminus S|\in\{-1,0,1\}$.
If sizes $h,h+1$ both occur, every smaller set is contained in every
larger one, since then $|T\setminus S|-|S\setminus T|=1$ forces $S\setminus T=\varnothing$
for $|S|=h$, $|T|=h+1$. Two distinct larger sets have intersection of size $h$,
forcing a unique smaller set, which is a center. If there is only
one larger set, it is a center.

If all sizes agree and there are at least two members,
choose distinct
$A,B$ and put $I=A\cap B$, $U=A\cup B$, so that $|U\setminus I|=2$. Every other member
either
contains $I$ or is contained in $U$: omitting a point $p$ of $I$ forces
both points of $U\setminus I$, because $A\setminus S=B\setminus S=\{p\}$, and then
$S=U\setminus\{p\}$ by size. A member containing $I$ and a point
outside $U$ cannot coexist with one omitting a point of $I$, since
their directed difference would have size at least two. Thus either
$I$ or $U$ is a center: if no member omits a point of $I$, all members are $I$ plus
one point; otherwise all members are $U$ minus one point. A family with one member is its own
center.

With three or more
distinct members, their coordinatewise majority determines the
center uniquely, since the members at distance one from a center $H$ differ from $H$
at distinct coordinates, so at each coordinate at most one member disagrees with $H$.

\proofstep{Centers of the large-set components.}
Call a member of $\mathcal F$ large if its size is at least four.
Join two large sets if they intersect in at least two points.
This relation is transitive: if $S$ and $U$ are both adjacent to
$T$, then \eqref{eq:dual-two-sets} gives $|T\setminus S|,|T\setminus U|\le1$ and
$|S\cap U|\ge|T|-2\ge2$. The components are therefore cliques.

Add to each clique every three-element set of $\mathcal F$ meeting one of its
members $T$ in at least two points, and call the result an enlarged clique. Such a
set $S$ is contained in that
member, whose size must be four: by \eqref{eq:dual-two-sets}, $|T\setminus S|\le1$
gives
$|T|\le|S\cap T|+1\le4$, and $|T|\ge4$ then forces equality throughout. The set
$S$ then meets every member $T'$ of
the clique in at least two points, because $S\subset T$ and $|T\setminus T'|\le1$, so
the same conclusion holds
for all of them. It cannot belong to two cliques, since two large sets containing it
would meet in three points.

Inside an enlarged clique all directed differences have size at
most one: for two large members by \eqref{eq:dual-two-sets}, for a large member and
an added set by the inclusion, and for two added sets because they lie in one
four-element set. Each enlarged clique has a radius-one center $H$: take its unique
majority center
if it has at least three members, their intersection if it has two,
and its sole member otherwise. Each enlarged clique contains a large member, so
$|H|\ge3$.

Every pair of points in $H$ lies in an original large member of
that clique. For three or more members, at most two can omit one
of the points, so a third contains the pair; if that third member
has size three, a large member contains it. The one- and two-member
cases follow directly from the definition of $H$, because $H$ is contained in every
member when there are at most two.

Further, any
$S\in\mathcal F$ containing that pair belongs to the same enlarged
clique. To verify this, use a large set $T$ containing the pair.
Equation~\eqref{eq:dual-two-sets} either puts another large $S$
in its clique or forces $|S|=3$, $|T|=4$, and $S\subset T$.
A smaller $S$ would leave at least two points of $T$ outside $S$.
Consequently distinct centers intersect in at most one point: two centers
sharing a pair would give large members of the two cliques that meet in that pair,
and these members would then lie in one clique.

\proofstep{A five-stage list.}
Consider the fixed list
\[
 L_{\mathcal F}=(\varnothing,\text{ all singletons},
 \text{ the centers},\text{ all pairs},\mathcal F).
\]
Remove duplicates, retaining their first occurrence. On any growing
consistent set of positive and negative constraints, denote the
forced positive and negative coordinates by $P,Q$; these are disjoint, and a set $X$
in the list extends the constraints exactly when $P\subseteq X$ and
$X\cap Q=\varnothing$. The empty
support is used at most once, and exactly when $P=\varnothing$.
In the singleton stage the only
possible support is $P$ with $|P|=1$, and $P$ itself extends the constraints, so it is
used at most once.
Entering the center stage requires $|P|\ge2$; distinct centers
intersect in at most one point and $P$ only grows, so at most one is possible. In the
pair stage the only possible support is $P$ with $|P|=2$, and it remains the same
while $|P|=2$.

At the final stage $|P|\ge3$, since otherwise an entry of an earlier stage extends the
constraints, and every center is excluded. If a
large set $T$ survives, then $|P\setminus H|\le|T\setminus H|\le1$ for the center $H$ of
its clique, so $P\cap H$ contains a pair, and
all surviving members contain this common pair
of $P$ and hence lie in the enlarged clique, inside its radius-one ball. A literal
opposing the center, at a coordinate $x$, leaves at most one member of that clique,
namely the
one differing from $H$ at $x$. If no large
set survives, any surviving member has size at most three and
contains $P$, so again there is at most one. The list therefore
uses at most five labels before exhaustion.

Suppose now that the growing sets of constraints are closures of the class
$\mathcal F$. Then the center and pair stages cannot both occur.
Indeed, if a center $H\supseteq P$ was used with $|P|=2$ and the pair stage follows,
then $P$ is unchanged, the surviving members contain the pair $P\subseteq H$, and after
the center fails they lie in its ball and differ from $H$ at a fixed coordinate.
After a center fails, the nonempty surviving version in its ball
is therefore a singleton and its closure specifies its full row, so $P$ is that row.
That row
has size at least three, since it belongs to the enlarged clique, which contradicts
$|P|=2$.
Thus there are at most four labels on closures of $\mathcal F$, or at most
three if a positive coordinate is forced from the start, since the empty
support is then not used.

\proofstep{Combining the two backgrounds.}
Use $L_{\mathcal F}$ followed by its sign reversal. It contains
all rows of $C$ and is conservative by Lemma~\ref{lem:priority-list}. On a chain of
closures of $C$, the first block uses at most
five labels. If it uses five, the center-to-pair transition still
has $|P|=2$ with $P$ inside the center, so a negative constraint excludes the center
and
$Q\ne\varnothing$.

When the first block is exhausted all surviving
rows belong to $-\mathcal F$. The current closure $R$ of $C$ is
also a closure of the class $-\mathcal F$. Indeed, putting
$W=C_R$ and $V=(-\mathcal F)_R$, one has $\varnothing\ne W\subseteq V$
and
\[
 R\subseteq R(V)\subseteq R(W)=R.
\]
Negating all signs makes these closures of $\mathcal F$, on which the second block
acts as $L_{\mathcal F}$.
The second block consequently uses at most four labels. If the
first used five, its nonempty $Q$ is a positive constraint after
sign reversal, and the second uses at most three. In either case
there are at most eight labels, hence at most seven changes.

The list has $O(n^2+m)$ entries, as asserted: there are at most $\binom t2$ pairs, $t$
singletons, and $|\mathcal F|\le m$ further supports, and distinct centers share no
pair of points, so there are at most $\binom t2/3$ centers.

\proofstep{Sharpness and the last assertion.}
The full seven-dimensional cube has $q=d=7$, and its seven column
functions can be divided as evenly as desired by a row. Its dual
dimension is $\lfloor\log_2 7\rfloor=2$, proving sharpness: seven functions forbid dual
depth three by Lemma~\ref{lem:rank-facts}(a), and the splits $7\to(3,4)\to(1,2),(2,2)$
give a dual tree of depth two. Since $d\le q\le7$ for all these classes, both suprema
are seven.

It remains to prove $q(C)\le d(e+1)$.
For $e=2,d\ge3$, $7\le3d$; for $d=1,2$, use
Theorem~\ref{thm:static-recurrence}, which gives $q\le1$ and $q\le3$. Also $d=0$ gives
$q=0$. Theorem~\ref{thm:dual-one}
handles $e=1$. If $e=0$, all columns are identical, so $d=q\le1$.
\end{proof}

\subsection{Rank plateaus and an exact product family}
Along a chain the rank of the version is weakly decreasing, so a chain splits into
blocks on which the rank is constant. The number of label changes on a chain is at
most the number of rank drops plus the changes inside these blocks, so adding bounds
for the separate ranks gives a bound on $q$. This subsection determines the changes at
each fixed rank exactly for one product family. For this family the sum over the ranks
is quadratic in the Littlestone dimension, whereas $q$ equals the Littlestone
dimension, so this way of adding local bounds can lose a factor that grows with the
dimension.

A \emph{rank-$r$ plateau} is a chain all of whose versions have dimension $r$. For a
consistent label $\ell$, let $K_r(\ell)$ be the largest number
of changes on a rank-$r$ plateau.
A label with at most $Q$ changes on every chain has $K_r\le Q-r$: append to
a rank-$r$ plateau a depth-$r$ mistake tree, which forces $r$ further
changes. The tree is taken inside the last version $V$ of the chain, of rank $r$, and
is followed as in the proof of Proposition~\ref{prop:static-elementary}, since a branch
of $V$ at a literal is again a closed version of $C$.
The following family determines the minimum of $K_r$ exactly at every rank. For that
family, the bound $K_r\le Q-r$ is attained for $r\ge k$, with $Q=2k$.

\begin{theorem}\label{thm:plateau-family}
Let $E=\{0,1\}^3\setminus\{000,111\}$ and $C_k=E^k$ on $3k$
coordinates. Then
\[
 \operatorname{Ldim}(C_k)=q(C_k)=\operatorname{lr}(C_k)=2k,
 \qquad \operatorname{Ldim}(C_k^*)=\lfloor\log_2(3k)\rfloor.
\]
Every consistent label, even without conservativity, satisfies
\[
 K_r(\ell)\ge\min\{r,2k-r\}\quad(1\le r\le2k).
\]
One conservative label attains all these lower bounds simultaneously.
Consequently $\min_\ell\max_rK_r=k$ and
$\min_\ell\sum_{r=1}^{2k}K_r=k^2$.
\end{theorem}
\begin{proof}
The lower bound on $K_r$ comes from a topological argument on a cube of closures,
built from a path of rank-one versions of $E$. We then give a label attaining these
bounds and compute the other parameters.

\proofstep{A rank-one path in $E$.}
Encode the rows of $E$ by integers $1,\ldots,6$, with the lowest
bit the first coordinate and with the bits $1,0$ meaning $+,-$. Its closed rank-one
versions contain the
following path in their comparability graph:
\[
 \{1,5\}\subset\{1,4,5\}\supset\{4,5\}
 \subset\{4,5,6\}\supset\{4,6\}.
\]
The common literals verify closure directly. They are, in order, with $x_i^{\pm}$
denoting the literal $(i,\pm1)$,
\[
 Z_0=\{x_1^+,x_2^-\},\quad Z_1=\{x_2^-\},\quad Z_2=\{x_2^-,x_3^+\},\quad
 Z_3=\{x_3^+\},\quad Z_4=\{x_1^-,x_3^+\},
\]
and the rows of $E$ extending each set are exactly the displayed ones, because $000$
and $111$ are absent from $E$. Every two-row version has dimension one.
A three-row version cannot support four leaves,
so it also has dimension one. The endpoints force opposite values
of the first coordinate; nonconsecutive vertices of this displayed
path are incomparable, as a comparison of the five row sets shows. As closures,
$Z_1\subset Z_0,Z_2$ and $Z_3\subset Z_2,Z_4$.

\proofstep{A cube of closures.}
Fix $a\ge1$, choose $a$ active factors of $C_k$, and fix a closure in each other
factor. For $v\in\{0,\ldots,4\}^a$ let $R_v$ be the closure of $C_k$
(Lemma~\ref{lem:product-poset}) with $Z_{v_i}$ in the $i$-th active factor and the fixed
closures elsewhere. Write $v\preceq v'$ if $R_v\subseteq R_{v'}$; this is the
product of the orders on the path.

Take the product of $a$ copies of the path. Its order complex
triangulates a product of intervals: triangulate each small product
cell by its monotone chains in the product order. To see this, note that a set of
pairwise
comparable grid points takes at most two consecutive integer values in each
coordinate, so it lies in a unit cube of the grid. After a reflection of each factor
of such a cube that sends the even endpoint to $1$ and the odd endpoint to $0$, the
order becomes the coordinatewise order on $\{0,1\}^a$. A point $x\in[0,1]^a$ is a
convex combination, with unique positive coefficients, of a chain of vertices: the
indicator vectors of its superlevel sets $\{i:x_i\ge t\}$, $t>0$, and the zero vector
if $\max_ix_i<1$. So the chains triangulate the cube.
The triangulations
agree on common faces, because the chains inside a face are the chains of its vertex
set in the same global order.

\proofstep{A chain with $a$ changes.}
Let $\ell$ be a consistent label.
Project its value at each vertex
to the first coordinate of each of the $a$ active factors and extend
affinely over each simplex. This yields a continuous map $f$ from
$[0,4]^a$ to $[-1,1]^a$ with opposite fixed coordinate signs on
each pair of opposite faces: $f_i=+1$ on $\{x_i=0\}$ and $f_i=-1$ on $\{x_i=4\}$,
because $Z_0$ and $Z_4$ force these values and the faces are triangulated by their own
vertices. The Poincar\'e--Miranda theorem shows
that the image contains $(-1,1)^a$. To derive this from the Brouwer fixed point
theorem~\cite[Corollary 2.15]{Hatcher02}, fix $z\in(-1,1)^a$ and put $g=f-z$. The map
sending $x$ to the coordinatewise truncation of $x+g(x)$ to $[0,4]$ has a fixed
point $x^*$. If $g_i(x^*)>0$, the fixed-point equation forces $x^*_i=4$, where $g_i<0$;
if $g_i(x^*)<0$, it forces $x^*_i=0$, where $g_i>0$. So $g(x^*)=0$ and $f(x^*)=z$.

If each simplex chain changed
labels fewer than $a$ times, it would carry at most $a$ distinct labels, and
its image would lie in the convex hull
of at most $a$ points and have dimension at most $a-1$. A finite
union of those images cannot contain an open $a$-cube, since a finite union of closed
sets with empty interior has empty interior. Some chain
therefore has at least $a$ changes.

\proofstep{The lower bound on every plateau.}
Dimensions add on products over disjoint domains
(Lemma~\ref{lem:rank-facts}(d)). The lower bound
comes from concatenating mistake trees, and the upper bound is the induction in that
lemma. Also $\operatorname{Ldim}(E)=2$: two coordinates are shattered and six rows
forbid depth three.

To realize total rank $r\le k$, use $r$
active rank-one path factors and singleton versions elsewhere.
For $k\le r\le2k$, use $2k-r$ active path factors and $r-k$
full rank-two factors; these are $k$ factors of total rank
$(2k-r)+2(r-k)=r$. All closures $R_v$ then have rank $r$, so the preceding
step proves the
claimed lower bound on every plateau. For $r=2k$ the bound is zero.

\proofstep{A label attaining the bounds.}
For the matching upper bound, label a closure in $E$ by its forced
positive coordinates and put zero elsewhere. This is conservative
(Lemma~\ref{lem:background-label} with the negative background).
Every row has at most two ones, so a chain has at most two changes.
The sole rank-two version is $E$ itself, since every other closed version lies in a
branch of $E$ at a literal, which has three rows. Rank-zero versions are
singletons, and any rank-one chain has at most one strict inclusion,
because strict nonsingleton subversions have two or three rows.
Thus $K_0=K_2=0$ and $K_1\le1$.

Use the product label on $C_k$, which is conservative
by the proof of Proposition~\ref{prop:static-elementary}.
On a chain of constant total rank every factor rank is constant, because factor ranks
are weakly decreasing along the chain and their sum is constant.
If $a$ factors have rank one and $b$ have rank two, then
$a+2b=r$ and $a+b\le k$. Hence $a\le r$ and $2a\le 2k-2b=2k-r+a$, that is,
$a\le\min(r,2k-r)$, and only
these $a$ factors can change, at most once each. This attains all the lower bounds.
Their maximum is $k$, attained at $r=k$, and their sum is
$\sum_{r=1}^{k}r+\sum_{r=k+1}^{2k}(2k-r)=k^2$.

\proofstep{The number of changes and the list number.}
Two coordinates in each factor are shattered: the first two coordinates of $E$ take
all four values. Thus $2k$ coordinates
are shattered, giving $d\ge2k$ and list number at least $2k$ by
the cube value in Theorem~\ref{res:5.1}, applied to the restriction to these
coordinates, because a rule for $C_k$ restricts to a rule for the cube. The product
label gives
$q\le2k$ and hence $d=q=2k$ with Proposition~\ref{prop:static-elementary}.

For the
list upper bound, $E$ has
a rank-two sign representation with columns
$(1,0),(0,1),(-1,-1)$ and rows, in integer order,
\[
 (2,-1),\ (-1,2),\ (1,1),\ (-1,-1),\ (1,-2),\ (-2,1).
\]
All dot products are nonzero and have the required signs: the three values on the
six rows are $(2,-1,-1)$, $(-1,2,-1)$, $(1,1,-2)$, $(-1,-1,2)$, $(1,-2,1)$, and
$(-2,1,1)$. Put each
factor into its own two-dimensional block to obtain sign rank at
most $2k$, with integer dot products of absolute value at least one. Lemma~\ref{res:5.0}
gives $\operatorname{lr}(C_k,\varepsilon)\le2k$ for $\varepsilon\in(0,1/2)$, and this
range of $\varepsilon$ determines $\operatorname{lr}(C_k)$ by the monotonicity recorded
after~\eqref{eq:static-chain}. This proves the
list equality.

\proofstep{The dual dimension.}
There are $3k$ distinct columns, so the dual dimension is at most
$\lfloor\log_2(3k)\rfloor$ by Lemma~\ref{lem:rank-facts}(a). To attain it, take any
subset of at
least two columns. Let it contain $b$ complete three-column blocks
and $p$ columns from incomplete blocks. A row of $C_k$ may put one
or two positive signs in each complete block and arbitrary signs
on the selected columns of an incomplete block, since a row of $E$ can take any values
at two coordinates. Its total number
of positive signs can therefore be any integer from $b$ to $2b+p$.
This interval contains $\lfloor(3b+p)/2\rfloor$, because
$b\le\lfloor3b/2\rfloor$ and $(3b+p)/2\le2b+p$. Every nonsingleton
dual version can thus be split as evenly as possible. Recursive
balanced splitting gives a complete tree of the required depth, since $s$ columns
split into parts of sizes at least $\lfloor s/2\rfloor$, and
$\lfloor\log_2\lfloor s/2\rfloor\rfloor=\lfloor\log_2s\rfloor-1$ for $s\ge2$.
\end{proof}

This family distinguishes the total number of label changes from estimates
that add the worst case of every rank plateau. Its total number of label changes is
$2k$, although that sum is $k^2$, and a single plateau requires $k$ changes
while the dual dimension is only $\lfloor\log_2(3k)\rfloor$.

\subsection{A complete criterion for two changes}
Proposition~\ref{prop:static-elementary} and Theorem~\ref{thm:static-recurrence} show
that $q(C)\le1$ exactly when $d\le1$. At the next level the Littlestone dimension no
longer decides: at $d=2$ the full two-dimensional cube has $q=2$, whereas the six-row
dictator class has $q=3$. The next proposition gives an exact criterion for $q\le2$.
It also shows that at this level conservativity is no restriction: $q(C)\le2$ is
equivalent to the corresponding bound $\mu(C)\le3$ on runs of arbitrary consistent
labels.

For a consistent root label $h$, that is, a function extending $R(C)$, call a closure
\emph{bad} if $h$ does not extend it, and consider the graph whose vertices are the
bad closures with nonsingleton version, two of them being adjacent when comparable.
This is the comparability graph in the next statement. The forced literals of a
component are jointly consistent if their union contains no two literals at one
coordinate with opposite values.

\begin{proposition}\label{prop:two-changes}
The following are equivalent: $q(C)\le2$; $\mu(C)\le3$; and
there is a consistent root label $h$ such that each component of
the comparability graph of nonsingleton closed versions whose
closures contradict $h$ has jointly consistent forced literals.
\end{proposition}
\begin{proof}
The first condition implies the second by \eqref{eq:static-chain}.

Suppose a consistent label $\ell$ has at most three runs on every chain, and put
$h=\ell(R(C))$. Suppose that two comparable bad nonsingleton versions have
different labels, and let their closures be $R\subset R'$. Passing
from the root to $R$ and then
to $R'$ gives two changes, since each bad closure
excludes $h$. The later version has at least two rows, so some
realizable literal contradicts its current label: choose a coordinate on which the
version is not constant, and the value opposite to $\ell(R')$ there. Adding it forces
a third change, or four runs, a contradiction. Thus each component
has a common label, which extends the union of its literals.

Conversely, given $h$ as in the third condition, label closures agreeing with $h$ by
$h$. Label each
bad nonsingleton component by one common extension, which exists because its forced
literals are jointly consistent, and label
the bad closures with singleton version by their unique row.

This label is conservative. Let $R\subset R'$ with $\ell(R)$
extending $R'$. If $h$ extends $R$, then $\ell(R)=h$ extends $R'$, so $\ell(R')=h$.
Otherwise $R'$ is bad as well.
Comparable bad nonsingleton versions belong to the same component, so they have the
same label. If the
version becomes a singleton while the old label remains consistent,
that label already equals its unique row.

Closures extended by $h$ form an initial segment of a chain, and the nonsingleton bad
closures of a chain lie in one component. So a chain uses at most
the root label, one component label, and one singleton label, for
at most two changes.
\end{proof}

\section{Further exact families}\label{app:families}
This section computes exactly, on a few explicit families, the parameters studied so far:
the list number $\operatorname{lr}$, the conservative-label parameter $q$, the run number
$\mu$, which counts constant runs of a label along a chain, the chain-label number $\chi^0$ of
Sections~\ref{ssec:6} and~\ref{app:static}, and the index and coindex of
the sign complex. These calculations extend the exact families of
Section~\ref{ssec:5}, and the values show that the list number, the number of labels, and the
index measure different things. A cube with its two constant rows removed has
$\operatorname{lr}=q$ and $\chi^0=\mu=q+1$. A Hamming ball of radius $k$ in dimension
$M\ge2k+1$ has $\operatorname{lr}=\chi^0=\mu=q+1=k+1$, while its sign complex, which is also the sign complex of its
symmetrization (the ball together with the negatives of its rows), has index $2k$. On a product of threshold classes with $n_i$ rows the number of labels is
$1+\sum_i\lfloor\log_2 n_i\rfloor$, although the threshold matrix $T_n$ itself has list number two for $n\ge2$ (Section~\ref{ssec:6}).
The list numbers of extremal classes use the extremal-class theorem
of~\cite{BHH26b}, and the other bounds below follow from explicit
constructions, shattering, or complete mistake trees.

A class is \emph{extremal} if each shattered coordinate
set is strongly shattered, meaning that all its patterns occur
with one common fixed pattern on the other coordinates.
We use the following form of the extremal-class theorem
\cite[Corollary 5.5]{BHH26b} (Theorem B there): a nonempty extremal class on a finite
domain that is not the full cube has list number $\operatorname{vc}+1$. The learning model
of~\cite{BHH26b} differs slightly from ours, and the comparison of models after
Theorem~\ref{res:5.1} shows that the two give the same value.
The full cube on $M\ge1$ coordinates has list number $M$ (Theorem~\ref{res:5.1}), and
its sign complex is $\partial\Diamond_M$, which is equivariantly $S^{M-1}$.

\subsection{A cube with its two constant rows removed}
We begin with the smallest change to the cube that keeps it closed under negation. Up to column sign changes, deleting one
antipodal pair of vertices gives the following class. Each parameter below is exactly one less
than on the full cube, and the list number equals $q$, so the inequality
$\operatorname{lr}\le q+1$ of~\eqref{eq:static-chain} is strict. The class is not extremal: a set of
$d-1$ coordinates is shattered, but for either fixed sign on the other coordinate the constant pattern of that sign
does not occur. Its list number is the VC dimension, and not the value
$\operatorname{vc}+1$ that the extremal-class theorem gives for extremal classes other than the full cube.

\begin{proposition}\label{prop:punctured-cube}
For $d\ge2$, let
$E_d=\{\pm1\}^d\setminus\{(1,\ldots,1),(-1,\ldots,-1)\}$.
Then
\[
 \operatorname{vc}(E_d)=q(E_d)=\operatorname{lr}(E_d)=d-1,
 \qquad \operatorname{coInd}S(E_d)=\operatorname{Ind}S(E_d)=d-2,
 \qquad \chi^0(E_d)=\mu(E_d)=d.
\]
\end{proposition}
\begin{proof}
\proofstep{The VC dimension.}
Every pattern on $d-1$ coordinates extends to a nonconstant row:
if the pattern is constant, give the other coordinate the
opposite sign. The full set of $d$ coordinates is not shattered, since the two
constant patterns are missing.
Thus the VC dimension is $d-1$, and the VC lower bound
$\operatorname{vc}\le\operatorname{lr}$ of Section~\ref{ssec:2.1} gives
$\operatorname{lr}(E_d)\ge d-1$.

\proofstep{The index and coindex.}
Since $-E_d=E_d$, the faces of $S(E_d)$ are the sets of signed coordinates, with at
most one sign per coordinate, that extend to a row of $E_d$. A proper subset of the
coordinates with any signs extends, by the first step, and a full sign vector
extends exactly when it is nonconstant.
Represent $S(E_d)$ as the $\ell_1$ unit sphere with the relative
interiors of the two constant-sign facets removed. Orthogonal
projection onto $\mathbf1^\perp$ never vanishes on this space.
Indeed its only zeros on the full $\ell_1$ sphere are
$\pm\mathbf1/d$, which lie in the removed interiors. Normalizing
the projection gives an odd map to $S^{d-2}$. Conversely, the
$\ell_1$ sphere in $\mathbf1^\perp$ lies in $S(E_d)$, because its points have
positive and negative entries, and it is an
odd copy of $S^{d-2}$, by radial projection in $\mathbf1^\perp$. Thus
$d-2\le\operatorname{coInd}S(E_d)\le\operatorname{Ind}S(E_d)\le d-2$.
This proves both index equalities.

\proofstep{The list upper bound.}
For a row $h$ with $k$ positive entries, define
\[
 w_i=\begin{cases}d-k,&h_i=1,\\-k,&h_i=-1.\end{cases}
\]
All entries have the required strict signs, because $1\le k\le d-1$, and
$\sum_iw_i=k(d-k)-(d-k)k=0$.
Using the first $d-1$ entries as row vectors, and columns
$e_1,\ldots,e_{d-1},-\mathbf1$, gives a sign representation in
$\mathbb R^{d-1}$: the inner product with $e_i$ is $w_i$, and with $-\mathbf1$ it is
$-\sum_{i<d}w_i=w_d$. All these integers have absolute value at least one.
This is a case of~\cite[Theorem 18(a)]{Arav13}. Hence
$\operatorname{lr}(E_d)\le d-1$, by Lemma~\ref{res:5.0} for
$\varepsilon\in(0,1/2)$ and monotonicity in $\varepsilon$.

\proofstep{The conservative-label and chain parameters.}
The Littlestone dimension is at least $d-1$ by shattering and at
most $\lfloor\log_2(2^d-2)\rfloor=d-1$, because a depth-$r$
complete tree requires at least $2^r$ rows. Complete every
closure by the negative background on unforced coordinates.
This labeling is conservative, by Lemma~\ref{lem:background-label}: if its label
remains consistent after
the closure grows, no newly forced coordinate can be positive, so the
completed label remains the same.
The label changes only when another positive coordinate is
forced. A realizable closure has at most $d-1$ such coordinates, since a row of
$E_d$ has at most $d-1$ positive entries,
so $q(E_d)\le d-1$. Proposition~\ref{prop:static-elementary}
gives equality, because $q\ge\operatorname{Ldim}$. Its distinct-label lower bound
$\chi^0\ge\operatorname{Ldim}+1$, together with
\eqref{eq:static-chain}, proves the last two equalities:
$d\le\chi^0\le\mu\le q+1=d$.
\end{proof}

\subsection{Hamming balls and their symmetrizations}
Hamming balls show the opposite behavior. For them $\operatorname{lr}=q+1$, so the inequality
$\operatorname{lr}\le q+1$ is attained. The sign complex of a class is also the sign complex
of its symmetrization, although the two classes can have different list numbers.
For $M\ge2k+1$, the list number $2k+1$ of the symmetrization and the index $2k$ of the common sign
complex are about twice the list number $k+1$ of the ball.

\begin{theorem}\label{thm:hamming-balls}
Let $0\le k<M$ and
$E_{M,k}=\{h\in\{\pm1\}^M:|\{i:h_i=1\}|\le k\}$.
Then
\[
 \operatorname{lr}(E_{M,k})=k+1,\qquad q(E_{M,k})=k,
 \qquad\mu(E_{M,k})=\chi^0(E_{M,k})=k+1,
\]
and
\[
 \operatorname{coInd}S(E_{M,k})=\operatorname{Ind}S(E_{M,k})
       =\min\{2k,M-1\},
 \qquad
 \operatorname{lr}(E_{M,k}\cup-E_{M,k})=\min\{2k+1,M\}.
\]
\end{theorem}
\begin{proof}
\proofstep{The class itself.}
The shattered subsets of $E_{M,k}$ are exactly those of size at
most $k$: a set of size at least $k+1$ does not carry the all-positive pattern. Each
set of size at most $k$ is strongly shattered by fixing every outside
coordinate negative. Thus the class is extremal and, since $k<M$, is not the
full cube. The extremal-class theorem of~\cite{BHH26b} gives
its list number $k+1$ (see the comparison of models after
Theorem~\ref{res:5.1}). Completing closures by the negative
background is conservative by Lemma~\ref{lem:background-label}, and changes a label at most $k$ times,
since a realizable closure forces at most $k$ positive coordinates. Since the
Littlestone dimension is at least the VC dimension $k$, the
bounds of Section~\ref{app:static} give $q=k$ and $\chi^0=\mu=k+1$: indeed
$k\le\operatorname{Ldim}\le q\le k$ and
$k+1\le\chi^0\le\mu\le q+1=k+1$ by Proposition~\ref{prop:static-elementary}
and~\eqref{eq:static-chain}. The list number also follows from these bounds:
\eqref{eq:static-chain} gives $\operatorname{lr}\le k+1$, and the subclass with $M-k$
fixed negative coordinates is, after negating those columns, the anchored cube of
Theorem~\ref{res:5.1} with $k$ free coordinates.

\proofstep{The upper bounds from sign rank.}
The bound $\operatorname{signrk}(E_{M,k})\le2k+1$ follows
from the sign-change estimate of~\cite[Lemma 18]{AMY16}, and we give an explicit
polynomial realization below. Together with the sign-rank bounds for list
replicability and index, this bound gives
the upper bounds in the second display of the theorem.

\proofstep{The coindex.}
Put $d=\min\{2k+1,M\}$. Every sign pattern on $d$ chosen
coordinates has at most $k$ occurrences of one of its signs, because $d\le2k+1$.
It extends either to $E_{M,k}$ or to $-E_{M,k}$ by giving the
other coordinates the other sign. Therefore the
symmetrization shatters $d$ coordinates. Its sign complex
contains an odd $S^{d-1}$, the boundary $\partial\Diamond_d$
(Section~\ref{ssec:2.1}), giving coindex at least $d-1$.

\proofstep{The index.}
For the matching upper bound, realize the sign complex on the
$\ell_1$ sphere in $\mathbb R^M$ and take distinct real numbers
$t_1,\ldots,t_M$. Consider the moment map
\[
 L(x)=\left(\sum_i x_i,\sum_i t_ix_i,\ldots,
                        \sum_i t_i^{2k}x_i\right).
\]
Suppose $L(x)=0$ for a nonzero point of the complex. Its first
coordinate implies that $x$ has both positive and negative
entries. The point lies in the simplex of a signed row, so after a global sign change
its positive support $S$
has size at most $k$. The polynomial
$P(t)=\prod_{i\in S}(t-t_i)^2$ has degree at most $2k$, vanishes
on the positive support, and is strictly positive at every
negative-support coordinate. Thus $\sum_i x_iP(t_i)<0$. This sum is the inner product
of the coefficient vector of $P$ with $L(x)$,
contradicting $L(x)=0$. Normalization gives an odd sphere map
of dimension $2k$. The ambient sphere also bounds the index
by $M-1$, the inclusion of $S(E_{M,k})$ in $\partial\Diamond_M$ being odd. This proves
the two index equalities, with $\operatorname{coInd}\le\operatorname{Ind}$.

\proofstep{The symmetrization.}
We finally prove the list upper bound for the symmetrization. If
$M\le2k+1$, it is the full cube and has list number $M$: a sign vector has at
most $k$ positive entries or at most $k$ negative entries.
Otherwise set $t_i=i$. For a positive support $S$ of size at
most $k$, the polynomial
\[
 Q_S(t)=\frac12-\prod_{i\in S}(t-i)^2
\]
is positive exactly on the coordinates of $S$ and negative on
the others, since the product vanishes on $S$ and is a positive integer at any other
integer coordinate. For $S=\varnothing$ the empty product is one, as
required. Its degree is at most $2k$. Negating the polynomial
represents the opposite row. Evaluation on
$(1,t_i,\ldots,t_i^{2k})$ therefore gives sign rank at most
$2k+1$. The list upper bound follows, by Lemma~\ref{res:5.0} after rescaling the row
vectors. The lower bound
$d$ is the VC lower bound, because the symmetrization shatters $d$ coordinates and
$\operatorname{vc}\le\operatorname{lr}$.
\end{proof}

\subsection{Products of extremal classes and of threshold classes}
We now place the preceding families in a broader setting and assemble larger examples from small
ones. Hamming balls are extremal, as the proof of Theorem~\ref{thm:hamming-balls} shows.
For every extremal class the list number is controlled by the coindex of its sign complex, and the
list number of a product of two extremal classes has a closed form.
\begin{proposition}\label{prop:extremal-products}
If $E\subseteq\{\pm1\}^M$ is a nonempty extremal class with
VC dimension $k$ and $M\ge1$, then
$\operatorname{lr}(E)\le\operatorname{coInd}S(E)+1$.
For two such classes on disjoint nonempty domains, put $b_i=0$
if $E_i$ is the full cube and $b_i=1$ otherwise. Then
\[
 \operatorname{lr}(E_1\times E_2)
 =k_1+k_2+\max\{b_1,b_2\}
 =\operatorname{lr}(E_1)+\operatorname{lr}(E_2)-b_1b_2.
\]
\end{proposition}
\begin{proof}
\proofstep{The inequality.}
For the full cube, the coindex is $M-1$ and the list number is
$M$. Otherwise $k<M$: shattering all $M$ coordinates would
force the full cube. Choose a strongly shattered $k$-set $A$, with fixed outside
pattern $f$, and
one coordinate $j$ outside it. Its fixed outside sign is reversed
in $-E$, so $E\cup-E$ shatters these $k+1$ coordinates: a pattern $(\pi,f(j))$ occurs
in $E$, and a pattern $(\pi,-f(j))$ is the negative of the pattern $(-\pi,f(j))$.
Consequently $\operatorname{coInd}S(E)\ge k$ (Section~\ref{ssec:2.1}).
The extremal-class theorem~\cite{BHH26b} gives
$\operatorname{lr}(E)=k+1$ and proves the inequality.

\proofstep{The product formula.}
In a product, a coordinate set is shattered, or strongly
shattered, exactly when its intersections with both factor
domains have that property: the patterns of a product row on the two blocks are
chosen independently, and a fixed outside pattern is a pair of fixed outside patterns.
Hence the product is extremal
with VC dimension $k_1+k_2$. It is a full cube exactly when
both factors are. Thus $\operatorname{lr}(E_i)=k_i+b_i$ for each factor, and the
product has list number $k_1+k_2+\max\{b_1,b_2\}$. Applying the same list formula and
$\max\{b_1,b_2\}=b_1+b_2-b_1b_2$ proves the display.
\end{proof}

For extremal classes other than the full cube, this improves the inequality
$\operatorname{vc}(E)-1\le\operatorname{coInd}S(E)$
in~\cite[Equation (5)]{BHHLT26} by one, and Theorem~\ref{thm:hamming-balls}
shows that the upper bound $\operatorname{Ind}S(E)\le2\operatorname{vc}(E)+1$
there is attained up to one.

The label parameters obey their own product rules. The parameter $q$ is additive on products by
Proposition~\ref{prop:static-elementary}, and the next proposition shows that the run number is
additive up to the constant one. Together with a binary-search argument, this gives exact values on
products of threshold classes.

A \emph{threshold class} with $n$ distinct rows is a class whose rows can be listed as
$c_1,\ldots,c_n$ so that every column changes sign at most once along the list. The
matrix $T_n$ of Theorem~\ref{res:6.4} is an example. Every branch at a literal is then
an initial or a final segment of the list. Consecutive rows differ at some column,
and that column separates the initial segment ending at the first of them from its
complement. Hence the closed versions are exactly the intervals $[a,b]$ of the list.

\begin{proposition}\label{prop:run-products}
For nonempty finite classes on disjoint domains,
\[
 \mu(C_1\times C_2)=\mu(C_1)+\mu(C_2)-1.
\]
For nonempty finite classes on the same domain,
$\mu(C_1\cup C_2)\le\mu(C_1)+\mu(C_2)$.
If $T_i$ is a threshold class with $n_i$ distinct rows, then
\[
 q\Big(\prod_iT_i\Big)=\sum_i\lfloor\log_2 n_i\rfloor,
 \qquad
 \mu\Big(\prod_iT_i\Big)=\chi^0\Big(\prod_iT_i\Big)
       =1+\sum_i\lfloor\log_2 n_i\rfloor.
\]
\end{proposition}
\begin{proof}
\proofstep{Products.}
The closures of the product are described by Lemma~\ref{lem:product-poset}.
Concatenate labels that attain the two factor run numbers.
Along a product chain each factor closure is weakly increasing, so each factor can
change at most
$\mu(C_i)-1$ times, giving at most
$\mu(C_1)+\mu(C_2)-1$ runs in total.

Conversely, fix the
second factor at its least closure $R_2^0$. The first components of
any product labeling form a consistent labeling of the first
factor, so some chain forces at least $\mu(C_1)$ runs of the first components.
Extend it to a singleton first version $\{c_1\}$, then hold that version
fixed. The second components of the labels of $\sigma_{c_1}\sqcup R_2$ give a consistent
labeling of
the second factor; a chain with at least $\mu(C_2)$ runs of these components, extended
to start at $R_2^0$, forces a further $\mu(C_2)-1$ changes.
On the concatenated chain the product label has at least
$\mu(C_1)+\mu(C_2)-2$ changes.
This proves equality, and the argument uses only the consistency of the labels.

\proofstep{Unions.}
For a union, use an optimal labeling $\ell_1$ of $C_1$ as long as its
version is nonempty, applying it to the closure within $C_1$: if $R$ is a closure
of $C_1\cup C_2$ with $(C_1)_R\ne\varnothing$, put $\ell(R)=\ell_1(R((C_1)_R))$.
Once that version becomes empty, use the analogous labeling
of $C_2$. These are at most two successive phases, with at
most the stated total number of runs, because along a chain the closures within each
class are weakly increasing and an empty version stays empty. Each chosen label
extends the original constraints, since $R\subseteq R((C_1)_R)$,
so the construction is valid.

\proofstep{Threshold classes.}
A threshold version is an interval in the ordered list of
rows. Recursively choose a middle row and then middle rows
of the two intervals on its sides, and use this recursive order as
a fixed priority list, with every row preceding the rows of the two intervals on its
sides. By Lemma~\ref{lem:priority-list} this list defines a conservative label. In the resulting binary search tree, the first row of a version
interval is the least common ancestor of its endpoints: this ancestor lies between
the endpoints, and the subtree below it is an interval containing the version. Excluding
that row places the new interval inside one of its child subtrees,
so the depth of the chosen row strictly increases at every change.
Each child subtree has at most half as many rows as its parent
subtree, and the tree has height $\lfloor\log_2 n_i\rfloor$.
This bounds the number of changes by that height.

A balanced binary search
tree gives a complete mistake tree of that depth: an interval of $s\ge2$ rows is split
by a column between two consecutive rows into parts of sizes $\lfloor s/2\rfloor$ and
$\lceil s/2\rceil$. So equality
holds for $q(T_i)$, by $q\ge\operatorname{Ldim}$ in
Proposition~\ref{prop:static-elementary}. Proposition~\ref{prop:static-elementary}
gives product additivity of $q$, by induction on the number of factors. A complete
mistake tree in
each factor can be followed successively, giving Littlestone
dimension at least the sum of the depths (Lemma~\ref{lem:rank-facts}(d)). Finally
$\operatorname{Ldim}+1\le\chi^0\le\mu\le q+1$ makes all the
inequalities equalities.
\end{proof}

\section{Paley matrices}\label{ssec:8}
The Paley matrices are the standard explicit pseudorandom sign matrices: \(P_p\) has operator norm \(\sqrt p+1\), and any two of its rows have inner product \(1\) or \(-3\). Random matrices, with high probability, and the Sylvester Hadamard matrices have index \(O(\log N)\) by~\cite{FHV26}. For \(P_p\), the general index bound of~\cite{FHMNV26v1}, whose simplicial case also follows from Theorem~\ref{thm:main} and Theorem~\ref{res:a}, gives \(\operatorname{Ind}S(P_p)=O(\log^2p)\), while signed shattering gives \(\operatorname{Ind}S(P_p)=\Omega(\log p)\) (Proposition~\ref{res:8.5}). The exact position of the Paley matrices between these two orders is open, and this section describes what controls it. We compute their parameters, bound their closure height by spectral inequalities, and relate the residual piece of the three-piece construction of Section~\ref{ssec:6.5} to the largest switchable monochromatic square, and hence to the clique number of the Paley graph. We also show that spectral pseudorandomness alone does not force logarithmic index, and we determine the index of \(P_{13}\) exactly.

\subsection{Parameters}\label{ssec:8.1}
We fix the notation of this section and evaluate on \(P_p\) the parameters that enter the general bounds of Sections~\ref{ssec:5} and~\ref{ssec:6}.

Let \(p\equiv1\pmod4\) be prime, and put
\[
P_p=Q+I,
\qquad
Q_{ij}=\chi(i-j).
\]
Rows and columns are indexed by \(\mathbb F_p\). Throughout this section \(\chi\) denotes the quadratic character modulo \(p\), with \(\chi(0)=0\); it is unrelated to the chain-label numbers of Section~\ref{ssec:6.2}. In matrix formulas, \(I\) is the identity matrix and \(J\) the all-ones matrix of order \(p\). Set
\[
L_p=(\sqrt p+1)^2,
\qquad
C_p=\lfloor L_p\rfloor.
\]

Closed faces of a sign complex are controlled by rectangles that become constant after switching the signs of rows and columns. An \(r\times c\) \emph{switchable monochromatic rectangle} (a switchable rectangle, for short) of a sign matrix \(B\) is a pair \((I,C)\) of row and column sets with \(|I|=r\), \(|C|=c\) such that \(B_{IC}=\sigma\tau^T\) for some \(\sigma\in\{\pm1\}^I\), \(\tau\in\{\pm1\}^C\). Equivalently, the rows in \(I\) agree on \(C\) up to a global sign, or the face with vertex set \(\{j^{\tau_j}:j\in C\}\) has at least \(r\) compatible signed rows. A switchable square is a switchable rectangle with \(r=c\).

For a background \(g\in\{\pm1\}^p\) and a row \(c\), viewed as a concept, put \(D_c=\{i:c(i)\ne g(i)\}\), and let
\[
s_{\cap,\min}(B)=\min_g\max_{c\ne c'}|D_c\cap D_{c'}|.
\]
This is the least intersection parameter admissible in Theorems~\ref{res:5.2} and~\ref{res:6.6}. It is a different quantity from the minimum star number \(s_{\min}\) of~\cite{BHHLT26}.

For a sign matrix \(B\), the number \(h(B)\) is the signed closure height of Section~\ref{ssec:6.1}, and \(R(F)\) is the set of signed rows compatible with a face \(F\). We write \(r^s\) for the signed row \((r,s)\). A nonempty face is not compatible with both a row and its negative, so the members of \(R(F)\) have distinct underlying rows. In the rest of this section, \(B\) denotes \(P_p\).

As \(p\equiv1\pmod4\), we have \(\chi(-1)=1\), so \(Q\) and \(B\) are symmetric. Further \(QJ=JQ=0\), because \(\sum_a\chi(a)=0\). We use the character identity
\begin{equation}\label{eq:paley-character}
\sum_{a\in\mathbb F_p}\chi(i-a)\chi(j-a)=
\begin{cases}p-1,&i=j,\\-1,&i\ne j,\end{cases}
\qquad\text{that is,}\qquad
Q^2=pI-J.
\end{equation}
For \(i=j\) every term with \(a\ne i\) equals \(1\) and the term with \(a=i\) vanishes. For \(i\ne j\), substitute \(u=a-i\) and \(d=i-j\ne0\): the sum becomes \(\sum_{u\ne0}\chi(u)\chi(u+d)=\sum_{u\ne0}\chi(1+d/u)\), and \(1+d/u\) runs over \(\mathbb F_p\setminus\{1\}\), so the sum is \(-\chi(1)=-1\). Consequently
\begin{equation}\label{eq:paley-gram}
BB^T=(Q+I)^2=(p+1)I-J+2Q .
\end{equation}

The next theorem collects the parameters of \(P_p\). It shows that the switching binary rank, the intersection parameter, and the sign-rank upper bound are all of order \(p\), and that the rectangle bound forces \(h(P_p)=O(\sqrt p)\). So the resulting index bounds, of order \(p\) and \(\sqrt p\), stay far above the general bound \(O(\log^2p)\), and the following subsections use finer information about the matrix.

\begin{resultenv}{Theorem}{res:8.1}{Paley parameters}
(i) With \(A=(J-I-Q)/2\) and \(a=(p-1)/4\), we have \(A^2+A=a(I+J)\).

(ii) The switching binary rank is \(r_2(P_p)=p-1\) for \(p\equiv5\pmod8\), and \(r_2(P_p)=(p-1)/2\) for \(p\equiv1\pmod8\).

(iii) The least intersection parameter is \(s_{\cap,\min}(P_p)=a\).

(iv) We have \(\|P_p\|=\sqrt p+1\), every switchable rectangle satisfies \(|I||C|\le p+1+\lfloor2\sqrt p\rfloor\), and \(\lfloor(h(P_p)+1)^2/4\rfloor\le C_p\).

(v) The matrix
\[
E_+
=
\frac12\left(I-J/p+Q/\sqrt p\right)
\]
is a rank-\((p-1)/2\) projection with \(\operatorname{sign}(E_+)=P_p\), so \(\operatorname{signrk}(P_p)\le(p-1)/2\) and \(\operatorname{Ind}S(P_p)\le(p-3)/2\).
\end{resultenv}

\begin{proof}
We begin with the quadratic relation. Using \(QJ=0\), \(J^2=pJ\) and \eqref{eq:paley-character},
\[
4A^2=(J-I-Q)^2=pJ+I+(pI-J)-2J+2Q=(p-3)J+(p+1)I+2Q.
\]
Since \(4A=2J-2I-2Q\), this gives \(4(A^2+A)=(p-1)(I+J)\). This is the strongly-regular-graph relation of the Paley graph.

\proofstep{The binary rank of \(A\).} The entries of \(A\) are \(0\) on the diagonal and \((1-\chi(i-j))/2\) off it, so \(B=(-1)^A\) in the notation of Theorem~\ref{res:6.1}. Reduce the relation modulo \(2\). Over \(\mathbb F_2\), \(A\mathbf 1=0\) because every row of \(A\) has \((p-1)/2=2a\) ones. Put \(W=\mathbf 1^\perp\), so that \(\mathbb F_2^p=\langle\mathbf1\rangle\oplus W\) as \(p\) is odd. Note that \(J\) vanishes on \(W\), and \(A\) maps \(\mathbb F_2^p\) into \(W\), since \(A\) is symmetric with \(A\mathbf1=0\). Hence \(\operatorname{im}A=A(W)\).

If \(a\) is odd (\(p\equiv5\pmod8\)), the relation becomes \(A(A+I)=I\) on \(W\), so \(A|_W\) is invertible and \(\operatorname{rank}A=p-1\).

If \(a\) is even (\(p\equiv1\pmod8\)), then \(A|_W\) is idempotent. Let \(\Pi\) be the permutation matrix of \(i\mapsto n_0i\) for a nonresidue \(n_0\). Then \(\Pi Q\Pi^T=-Q\), because \(\chi(n_0)=-1\), so \(\Pi A\Pi^T=(J-I+Q)/2=J-I-A\), which is \(J+I+A\) modulo \(2\) and equals \(I+A\) on \(W\). As \(\Pi\) preserves \(W\), the maps \(A|_W\) and \((I+A)|_W\) are similar and have equal rank. Being complementary idempotents, their ranks sum to \(p-1\). Thus \(\operatorname{rank}A=(p-1)/2\).

\proofstep{From \(A\) to the switched matrix.} The number \(r_2\) does not depend on the pivot, as shown in the proof of Theorem~\ref{res:6.1}, so we use the pivot \((0,0)\). The matrix \(D\) of Theorem~\ref{res:6.1} then has zero first row and column, and for \(r,i\ne0\) its entry is \(A_{ri}+A_{r0}+A_{0i}+A_{00}=(e_r+e_0)^TA(e_i+e_0)\). So \(D\), with these zero lines deleted, is the Gram matrix \(G^TAG\) of the bilinear form \((x,y)\mapsto x^TAy\) on the basis \(\{e_i+e_0:i\ne0\}\) of \(W\), where \(G\) has these basis vectors as columns. The column space of \(AG\) is \(A(W)=\operatorname{im}A\). The map \(G^T:x\mapsto(x_i+x_0)_{i\ne0}\) has kernel \(\langle\mathbf1\rangle\), which meets \(\operatorname{im}A\subseteq W\) only in zero. Hence \(r_2(P_p)=\operatorname{rank}D=\operatorname{rank}A\), which gives the two cases.

\proofstep{The parameter \(s_{\cap,\min}\).} Take the all-positive background, so that \(D_r\) is the set of negative positions of the row \(r\), with \(|D_r|=(p-1)/2\). By \eqref{eq:paley-gram}, two distinct rows have inner product \(-1+2\chi(r-s)\in\{1,-3\}\). The inner product also equals \(p-2|D_r\triangle D_s|=p-2(p-1)+4|D_r\cap D_s|\). So
\[
|D_r\cap D_s|=\frac{p-3+2\chi(r-s)}4\in\{a,a-1\},
\]
hence \(s_{\cap,\min}\le a\).

Conversely, consider any background. Every column contains \((p+1)/2\) entries \(+1\) and \((p-1)/2\) entries \(-1\), so at least \((p-1)/2\) rows err at every column. If \(m_i\) rows err at column \(i\), then counting ordered pairs gives
\[
\sum_{r\ne s}|D_r\cap D_s|=\sum_im_i(m_i-1)\ge p\cdot\frac{p-1}2\cdot\frac{p-3}2,
\]
since \(m(m-1)\) is nondecreasing for \(m\ge1\). So the average intersection over the \(p(p-1)\) ordered pairs is at least \((p-3)/4=a-\frac12\). The maximum is an integer, hence at least \(a\).

\proofstep{Norm and rectangles.} By \eqref{eq:paley-character}, \(Q\) has eigenvalue \(0\) on \(\mathbf 1\) and eigenvalues \(\pm\sqrt p\) on \(\mathbf1^\perp\), where the multiplicities are both \((p-1)/2\) because \(\operatorname{tr}Q=0\). So the symmetric matrix \(P_p=Q+I\) has eigenvalues \(1,1\pm\sqrt p\), and \(\|P_p\|=\sqrt p+1\).

Let \((I,C)\) be a switchable rectangle with signs \(\sigma,\tau\), and let \(x,y\) be \(\sigma,\tau\) extended by zero. Then \(x^TBy=|I||C|\) and \(\|x\|_2\|y\|_2=\sqrt{|I||C|}\), so the operator norm gives \(\sqrt{|I||C|}\le\sqrt p+1\), that is, \(|I||C|\le L_p\). As \(|I||C|\) is an integer, \(|I||C|\le\lfloor L_p\rfloor=p+1+\lfloor2\sqrt p\rfloor=C_p\).

\proofstep{Height.} Let \(F_1\subsetneq\cdots\subsetneq F_h\) be a strict closed-face chain, with \(h=h(P_p)\). The \(j\)-th face has at least \(j\) vertices, with distinct columns. Its compatible signed-row set has at least \(h-j+1\) elements: these sets strictly decrease along the chain, since a closed face is determined by its compatible signed rows, and the final set is nonempty. Hence the columns of \(F_j\) and the underlying rows of \(R(F_j)\) form a switchable rectangle of area at least \(j(h-j+1)\). Taking \(j=\lfloor(h+1)/2\rfloor\) gives area at least \(\lfloor(h+1)^2/4\rfloor\), which proves \(\lfloor(h+1)^2/4\rfloor\le C_p\).

\proofstep{Sign rank.} Let \(V_+\) be the \(\sqrt p\)-eigenspace of \(Q\). The matrix \((Q^2+\sqrt pQ)/(2p)\) acts as \(1\) on \(V_+\), as \(0\) on the \(-\sqrt p\)-eigenspace and on \(\mathbf1\). So it is the orthogonal projection onto \(V_+\), of rank \((p-1)/2\), and by \eqref{eq:paley-character} it equals \(E_+\). Its diagonal is \(\frac12(1-1/p)>0\), and its entry at \(i\ne j\) is \(\frac12(-1/p+\chi(i-j)/\sqrt p)\), of the same sign as \(\chi(i-j)\), because \(1/\sqrt p>1/p\).

Writing \(E_+=VV^T\) with \((p-1)/2\) columns gives a strict sign representation of \(P_p\) of the displayed dimension, so \(\operatorname{signrk}(P_p)\le(p-1)/2\). The odd map described after Lemma~\ref{res:5.0} gives \(\operatorname{Ind}S(P_p)\le\operatorname{signrk}(P_p)-1=(p-3)/2\).
\end{proof}

The two cases of \(r_2(P_p)\) are the classical \(2\)-ranks of the Paley adjacency matrix computed by Brouwer and van Eijl~\cite{BvE92}, transported through the switching normalization: \(\operatorname{rank}_{\mathbb F_2}A=p-1\) for \(p\equiv5\pmod 8\) and \((p-1)/2\) for \(p\equiv1\pmod 8\), and \(\operatorname{rank}_{\mathbb F_2}(A+I)\) is one larger in each case.

\subsection{Parity-sensitive spectral inequalities}\label{ssec:8.2}
The rectangle bound of Theorem~\ref{res:8.1} uses only the entries inside a rectangle, and it ignores the columns outside. When the number of rows is odd, each of these columns still contributes at least one to \(\|B^Tx\|^2\), where \(x\) is the vector of row signs extended by zero, because a sum of an odd number of signs is nonzero. This parity gain excludes rectangles whose area is within the bound of Theorem~\ref{res:8.1}.

\begin{resultenv}{Theorem}{res:8.2}{Odd rectangle inequality}
If there is an \(r\times c\) switchable monochromatic rectangle and \(r\) is odd, then
\[
p+c(r^2-1)\le rL_p.
\]
In integer form, if \(r\) is odd, \(\Lambda:=p+c(r^2-1)-r(p+1)>0\) and \(\Lambda^2>4r^2p\), then no \(r\times c\) switchable monochromatic rectangle exists. If \(c\) is odd, the same criterion with \(r\) and \(c\) interchanged applies.
\end{resultenv}

\begin{proof}
Let \(B_{ij}=\sigma_i\tau_j\) on the rectangle, with row set \(I\) of size \(r\) and column set \(C\) of size \(c\), and let \(x\) be \(\sigma\) extended by zero. On the \(c\) rectangle columns,
\[
(B^Tx)_j=\sum_{i\in I}\sigma_i\sigma_i\tau_j=r\tau_j.
\]
On every other column, \((B^Tx)_j\) is the sum of \(r\) signs. As \(r\) is odd, it is a nonzero odd integer, whose square is at least one. Thus
\[
\|B^Tx\|^2
\ge
cr^2+(p-c)
=
p+c(r^2-1).
\]
Since \(\|x\|^2=r\) and \(\|B\|^2=L_p\) by Theorem~\ref{res:8.1}, the spectral upper bound is \(\|B^Tx\|^2\le rL_p=r(p+1)+2r\sqrt p\). So a rectangle gives \(\Lambda\le2r\sqrt p\). If \(\Lambda>0\), squaring shows that \(\Lambda^2>4r^2p\) excludes the rectangle. As \(B\) is symmetric, an \(r\times c\) rectangle is also a \(c\times r\) rectangle with rows and columns exchanged, so the criterion with \(r,c\) interchanged applies when \(c\) is odd.
\end{proof}

An excluded rectangle bounds the height, because a long chain of closed faces contains a face with many vertices and many compatible signed rows. Through Theorem~\ref{res:6.1} this also bounds the index.

\begin{resultenv}{Theorem}{res:8.3}{No rectangle implies a height bound}
Let \(r,c\ge1\). If \(B\) has no \(r\times c\) switchable monochromatic rectangle, then \(h(B)\le r+c-2\) and \(\operatorname{Ind}S(B)\le r+c-3\).
\end{resultenv}

\begin{proof}
Let \(F_1\subsetneq\cdots\subsetneq F_h\) be a strict chain of nonempty closed faces with \(h\ge r+c-1\). The face \(F_c\) contains at least \(c\) columns. The sets \(R(F_j)\) strictly decrease along the chain and \(R(F_h)\) is nonempty, so \(|R(F_c)|\ge h-c+1\ge r\). Choosing \(c\) columns of \(F_c\) and \(r\) members of \(R(F_c)\), which have distinct underlying rows, creates the prohibited rectangle. So \(h(B)\le r+c-2\). The index bound follows from \(\operatorname{Ind}S(B)\le h(B)-1\) in Theorem~\ref{res:6.1}.
\end{proof}

For example, Theorem~\ref{res:8.2} excludes a \(15\times14\) rectangle at \(p=193\), although its area \(210\) is below \(C_{193}=221\): here \(\Lambda=193+14\cdot224-15\cdot194=419\) and \(419^2=175561>173700=4\cdot15^2\cdot193\). Theorem~\ref{res:8.3} then yields
\[
\operatorname{Ind}S(P_{193})\le26.
\]
The area bound alone excludes a rectangle only when \(rc\ge222\), which forces \(r+c\ge30\) and gives the weaker value \(27\).

\subsection{Shattering lower bounds for coindex}\label{ssec:8.3}
We turn to lower bounds. By Section~\ref{ssec:2.1}, \(k\) signed-shattered columns give coindex at least \(k-1\). The number of rows realizing a signed pattern on given columns expands into correlations of these columns, and the small operator norm of \(P_p\) makes the high-order correlations small on average. This gives four and five signed-shattered columns above explicit thresholds on \(p\). A probabilistic argument then gives the lower bound \(\Omega(\log p)\), and a count shows that signed shattering cannot exceed this order.

\begin{resultenv}{Theorem}{res:8.4}{Signed shattering in Paley matrices}
For \(p\ge29\), some four columns are signed-shattered, so \(\operatorname{coInd}(P_p)\ge3\). For \(p\ge89\), some five columns are signed-shattered, so \(\operatorname{coInd}(P_p)\ge4\).
\end{resultenv}

\begin{proof}
Put
\[
T_S=\sum_r\prod_{i\in S}B_{ri}
\]
for a column set \(S\). By \eqref{eq:paley-gram} and the symmetry of \(B\), the pair correlations satisfy
\[
T_{ij}=-1+2\chi(i-j)\in\{1,-3\}.
\]

\proofstep{A selection estimate.} The choice of columns rests on the following estimate, for a three-column set \(I\):
\[
\sum_{d\notin I}T_{I\cup\{d\}}^2
\le
\|B^Tv\|^2
\le
pL_p,
\]
where \(v_r=\prod_{i\in I}B_{ri}\), because \((B^Tv)_d=\sum_rB_{rd}\prod_{i\in I}B_{ri}=T_{I\cup\{d\}}\) for \(d\notin I\), \(\|v\|^2=p\) and \(\|B\|^2=L_p\). The sum is only over \(d\notin I\), since for \(d\in I\) the coordinate \((B^Tv)_d\) is a pair correlation.

\proofstep{Patterns and correlations.} For a \(k\)-column set \(J\) and a pattern \(\varepsilon\in\{\pm1\}^J\), let \(N(\varepsilon)\) be the number of rows whose restriction to \(J\) equals \(\varepsilon\). The indicator of this event for the row \(r\) is \(\prod_{i\in J}(1+\varepsilon_iB_{ri})/2\). Expanding gives \(N(\varepsilon)=2^{-k}\sum_{T\subseteq J}\varepsilon_TT_T\), where \(T_\varnothing=p\). Summing a pattern and its antipode cancels all odd correlations:
\[
N(\varepsilon)+N(-\varepsilon)
=
2^{1-k}\Bigl(p+\sum_{\varnothing\ne T\subseteq J,\ |T|\ \text{even}}\varepsilon_TT_T\Bigr),
\qquad
\varepsilon_T=\prod_{i\in T}\varepsilon_i .
\refstepcounter{equation}\tag{\theequation}\label{eq:res-8.1}
\]
The pattern \(\varepsilon\) is realized by a signed row exactly when the left side is positive. For four or five columns, only pair and four-column correlations occur on the right, and \(|T_{ij}|\le3\).

\proofstep{Four columns.} Let \(p\ge29\) and fix three columns \(I\). Averaging the selection estimate over the \(p-3\) candidates, some \(d\notin I\) satisfies
\[
|T_{I\cup\{d\}}|\le\sqrt{\frac{pL_p}{p-3}}<2\sqrt p.
\]
The last inequality is equivalent to \(L_p<4(p-3)\), that is, to \(3p-2\sqrt p-13>0\), which holds for \(p\ge13\). For \(J=I\cup\{d\}\) the six pair terms contribute at most \(18\) in absolute value, so \eqref{eq:res-8.1} gives \(8\,[N(\varepsilon)+N(-\varepsilon)]> p-18-2\sqrt p\) for every \(\varepsilon\). The right side is increasing for \(p\ge1\) and positive at \(p=29\), where it exceeds \(0.2\).

\proofstep{Five columns.} Let \(p\ge89\), and let \(J\) be the four columns just chosen. Apply the selection estimate to each of the four three-element subsets \(S\subset J\), and sum over \(e\notin J\). Averaging over the \(p-4\) candidates, some \(e\) satisfies \(\sum_{S}T_{S\cup\{e\}}^2\le4pL_p/(p-4)\), and by Cauchy--Schwarz \(\sum_S|T_{S\cup\{e\}}|\le4\sqrt{pL_p/(p-4)}\). Together with \(|T_J|\le\sqrt{pL_p/(p-3)}\), the five four-column terms of \(J\cup\{e\}\) contribute at most \(5\sqrt{pL_p/(p-4)}\), and the ten pair terms at most \(30\). So \eqref{eq:res-8.1} gives
\[
16\,[N(\varepsilon)+N(-\varepsilon)]\ge p-30-5\sqrt{\frac{pL_p}{p-4}} .
\]
This is positive for \(p\ge89\). Indeed \(\sqrt{p/(p-4)}\le21/20\), valid for \(p\ge44\), and \(\sqrt{L_p}=\sqrt p+1\), so the subtracted term is at most \(\frac{21}4(\sqrt p+1)\). Further \(f(p)=p-30-\frac{21}4(\sqrt p+1)\) has \(f'(p)=1-\frac{21}{8\sqrt p}>0\) for \(p\ge7\), and \(f(89)>4\).

\proofstep{Conclusion.} All signed patterns therefore occur on the chosen columns, and choosing at each step the smallest admissible column gives a deterministic construction. By Section~\ref{ssec:2.1}, \(k\) signed-shattered columns give the cross-polytope boundary subcomplex \(\partial\Diamond_k\) and \(\operatorname{coInd}\ge k-1\), which yields the stated coindex bounds.
\end{proof}

A count limits what signed shattering can give. If \(d_\pm\) denotes the largest number of signed-shattered columns, then
\[
d_\pm\le\lfloor\log_2(2p)\rfloor,
\]
because \(2^{d_\pm}\) patterns must occur among at most \(2p\) signed rows. Thus a cross-polytope boundary in \(S(P_p)\) gives an index lower bound of at most \(\lfloor\log_2(2p)\rfloor-1\). The next proposition shows that this order is attained, already by the VC dimension. Its proof uses only the fact that distinct rows are at Hamming distance at least \((p-1)/2\).

\begin{resultenv}{Proposition}{res:8.5}{Logarithmic lower bound}
For every prime \(p\equiv1\pmod4\), \(p\ge5\),
\[
\operatorname{VC}(P_p)>\frac{\ln p}{10},
\qquad
\operatorname{Ind}(P_p)\ge\operatorname{coInd}(P_p)>\frac{\ln p}{10}-1 .
\]
In particular \(\operatorname{Ind}(P_p)=\Omega(\log p)\).
\end{resultenv}

\begin{proof}
\proofstep{A small separating set of columns.} By \eqref{eq:paley-gram}, two distinct rows have inner product \(1\) or \(-3\), so their Hamming distance is \((p-1)/2\) or \((p+3)/2\), in both cases at least \((p-1)/2\). Choose \(k=\lceil6\ln p\rceil\) columns independently and uniformly at random. Two fixed rows agree on one random column with probability at most \((p+1)/(2p)\le3/5\) for \(p\ge5\), so they agree on all chosen columns with probability at most \((3/5)^k\). By the union bound over pairs of rows, and since \(6\ln(3/5)<-3\),
\[
\binom p2\left(\frac35\right)^k\le\frac{p^2}2\,p^{6\ln(3/5)}<\frac1{2p}<1,
\]
so some choice of \(k\) columns separates all \(p\) rows.

\proofstep{The VC dimension of the restriction.} Let \(D\) be the VC dimension of the rows restricted to these columns. Then \(D\ge1\), because \(p\ge2\) distinct restricted rows differ at some column. The Sauer--Shelah lemma, for a domain of at most \(k\) points, gives \(p\le\sum_{j\le D}\binom kj\le(ek/D)^D\).

Suppose \(D\le(\ln p)/10\). This already contradicts \(D\ge1\) when \((\ln p)/10<1\). Otherwise, use \(k\le6\ln p+1\le7\ln p\), which holds because \(\ln p>1\), and that \(y\mapsto(ek/y)^y\) is increasing on \([1,k]\), since the derivative of \(y\ln(ek/y)\) is \(\ln(k/y)\ge0\). Evaluating at \(y=(\ln p)/10\),
\[
p\le(70e)^{(\ln p)/10}=p^{\ln(70e)/10}<p^{0.53}<p,
\]
a contradiction. Hence \(\operatorname{VC}(P_p)\ge D>(\ln p)/10\).

\proofstep{Coindex and index.} Finally \(\operatorname{coInd}(S(A))\ge\operatorname{VC}(A)-1\) by~\cite{FHV26}, Proposition 6, and \(\operatorname{Ind}\ge\operatorname{coInd}\).
\end{proof}

\subsection{Inversion mixing and the three-piece residual}\label{ssec:8.4}
The three-piece bound of Theorem~\ref{res:6.7} charges the faces with few vertices of one sign and the faces with few compatible signed rows at a fixed cost. The residual piece \(E_{s,t}\), which consists of the faces with at least \(s+1\) vertices of each sign and at least \(t+1\) compatible signed rows, is charged by its height. A switchable rectangle whose column signs are nearly constant does not by itself give a face of this piece. For Paley matrices, however, the inversions \(x\mapsto1/(x-a)\) act by switching, and a suitable choice of \(a\) balances the signs of a rectangle. As a result, the residual piece is empty at a given scale essentially when the largest switchable square is below that scale.

We call \(a\) the \emph{pole} of the inversion \(\iota_a\) defined next. It is a bijection of \(\mathbb F_p\setminus\{a\}\) onto \(\mathbb F_p\setminus\{0\}\).

\begin{resultenv}{Theorem}{res:8.6}{Inversion mixing}
Let \(\iota_a(x)=1/(x-a)\) and \(\delta_a(x)=\chi(x-a)\). For \(x,y\ne a\),
\[
B_{\iota_a(x),\iota_a(y)}
=
\delta_a(x)\delta_a(y)B_{xy}.
\]
Let a switchable rectangle have row set $I$, column set $C$, and
$B_{xi}=\sigma_x\tau_i$ for $x\in I$, $i\in C$, where
$|I|=r$, $|C|=c$, and $u=|I\cup C|<p$.
There is a pole $a\notin I\cup C$ such that the face
\[
 F_a=\{\iota_a(i)^{\tau_i\delta_a(i)}:i\in C\}
\]
satisfies \(|R(F_a)|\ge r\) and
\[
\min(|F_a^+|,|F_a^-|)
\ge
\frac{c-\sqrt{pc/(p-u)}}2.
\]
If \(u\le p/2\), there is a pole \(a\notin I\cup C\) at which both the signs \(\sigma_x\delta_a(x)\) of the \(r\) signed rows \(\iota_a(x)^{\sigma_x\delta_a(x)}\in R(F_a)\), \(x\in I\), and the signs of the vertices of \(F_a\) are balanced: each of the two sign counts is at least \((r-2\sqrt r)/2\) for these rows, and at least \((c-2\sqrt c)/2\) for these vertices.
\end{resultenv}

\begin{proof}
\proofstep{The switching identity.} For distinct nonpole points,
\[
\iota_a(x)-\iota_a(y)
=
\frac{y-x}{(x-a)(y-a)}.
\]
As \(\chi\) is multiplicative, \(\chi(1/t)=\chi(t)\) for \(t\ne0\), and \(\chi(-1)=1\), this gives \(\chi(\iota_a(x)-\iota_a(y))=\chi(x-y)\delta_a(x)\delta_a(y)\), which is the asserted identity for \(x\ne y\). It also holds on the diagonal, since \(P_p\) has positive diagonal and \(\delta_a(x)^2=1\). This is a switched permutation equivalence between two \((p-1)\times(p-1)\) submatrices, not a global automorphism of all \(p\) points.

\proofstep{Averaging over poles.} For column signs \(\tau_i\), put
\[
S_C(a)=\sum_{i\in C}\tau_i\chi(i-a).
\]
Expanding the square and using \eqref{eq:paley-character} gives
\[
\sum_aS_C(a)^2
=(p-1)|C|-\sum_{i\ne j}\tau_i\tau_j
=
p|C|-\left(\sum_i\tau_i\right)^2
\le p|C|.
\]
Averaging over the \(p-u\) permitted poles \(a\notin I\cup C\), some pole has
\[
|S_C(a)|\le\sqrt{pc/(p-u)}.
\]

\proofstep{The inverted face.} The map \(\iota_a\) is injective, so \(F_a\) has \(c\) distinct vertices, and its vertex \(\iota_a(i)\) has sign \(\tau_i\delta_a(i)=\tau_i\chi(i-a)\). So the difference between the two new column sign counts is \(S_C(a)\), and \(\min(|F_a^+|,|F_a^-|)=(c-|S_C(a)|)/2\), which proves the first balancing bound. Moreover, the \(r\) distinct signed rows $\iota_a(x)^{\sigma_x\delta_a(x)}$, $x\in I$, are compatible with $F_a$: by the displayed switching identity, for \(i\in C\),
\[
\sigma_x\delta_a(x)B_{\iota_a(x),\iota_a(i)}
=\sigma_x\delta_a(x)^2\delta_a(i)B_{xi}
=\sigma_x\delta_a(i)\sigma_x\tau_i
=\tau_i\delta_a(i).
\]
Hence \(|R(F_a)|\ge r\).

\proofstep{Balancing rows and columns together.} When \(u\le p/2\), apply the same square-sum argument to the row sums \(S_I(a)=\sum_{x\in I}\sigma_x\chi(x-a)\) as well: \(\sum_aS_I(a)^2\le pr\) and \(\sum_aS_C(a)^2\le pc\), so \(\sum_a\bigl(S_I(a)^2/(pr)+S_C(a)^2/(pc)\bigr)\le2\). Averaging over the \(p-u\ge p/2\) permitted poles gives one pole at which this summand is at most \(4/p\), hence with \(S_I(a)^2\le4r\) and \(S_C(a)^2\le4c\) simultaneously. As \(S_I(a)\) is the difference of the two sign counts of the new signed rows, and \(S_C(a)\) is that of the vertices of \(F_a\), this yields the two stated lower bounds.
\end{proof}

We now apply this to the residual piece. If \(E_{q,q}\) is empty, Theorem~\ref{res:6.7} with \(s=t=q\) reads \(\operatorname{Ind}S(P_p)\le3q\). A face of \(E_{q,q}\) contains a switchable square of side \(q+1\), so a bound on the square side empties the residual. Theorem~\ref{res:8.6} gives the converse up to a constant factor, because it turns a large square into a face of \(E_{q,q}\).

\begin{resultenv}{Corollary}{res:8.7}{The residual of the three-piece construction on Paley matrices}
Let \(\beta_p\) be the maximum side length of a switchable monochromatic square, and let \(E_{q,q}\) be the residual piece of Theorem~\ref{res:6.7} with \(s=t=q\). For \(p\ge29\) and \(q\ge1\),
\[
\beta_p\le q
\ \Longrightarrow\
E_{q,q}=\varnothing
\ \Longrightarrow\
\beta_p<4(q+1).
\]
\end{resultenv}

\begin{proof}
A face \(F\) in \(E_{q,q}\) has at least \(q+1\) compatible signed rows, with distinct underlying rows, and at least \(q+1\) vertices of both signs. So \(q+1\) of its columns and \(q+1\) of these rows form a switchable square of side \(q+1>q\). This proves the first implication.

For the second, suppose \(\beta_p\ge k:=4(q+1)\ge8\) and take a \(k\times k\) switchable monochromatic square \(I\times C\). The rectangle bound of Theorem~\ref{res:8.1} gives \(k^2\le p+1+2\sqrt p\), so \(k\le\sqrt p+1\). As \(\sqrt p+1\le p/4\) for \(p\ge29\), we get \(u=|I\cup C|\le2k\le p/2<p\). By the first estimate of Theorem~\ref{res:8.6} there is a pole \(a\notin I\cup C\) with \(|S_C(a)|\le\sqrt{pk/(p-u)}\le\sqrt{2k}\le k/2\), the last inequality because \(k\ge8\). Hence the face \(F_a\) has \(\min(|F_a^+|,|F_a^-|)\ge(k-k/2)/2=q+1\) and \(|R(F_a)|\ge k\ge q+1\), so \(F_a\in E_{q,q}\), a contradiction.
\end{proof}

Thus \(E_{q,q}\) is empty for some \(q=O(\log^2p)\) if and only if \(\beta_p=O(\log^2p)\). This equivalence concerns the residual piece and not the index: when \(E_{q,q}\) is nonempty, Theorem~\ref{res:6.7} still gives an index bound through the height \(h_E\).

The number \(\beta_p\) is related to a classical quantity. A clique of the Paley graph is a set \(K\) with \(\chi(x-y)=1\) for distinct \(x,y\in K\). As \(P_p\) has positive diagonal, a clique of size \(k\) is a switchable square of side \(k\), with \(I=C=K\) and all signs positive. So \(\beta_p\) is at least the clique number of the Paley graph, and an estimate \(\beta_p=O(\log^2p)\) would in particular bound this clique number polylogarithmically. The size of the largest clique in a Paley graph is a classical open problem.

For a clique of size \(k<p\) the estimate of Theorem~\ref{res:8.6} can be sharpened. The all-positive weighted character sum of its proof has \(\sum_aS_C(a)^2=pk-k^2=k(p-k)\), and averaging over the \(p-k\) poles \(a\notin K\) gives a pole with \(|S_C(a)|\le\sqrt k\). After inversion this gives
\[
|R|\ge k,
\qquad
\min(|F^+|,|F^-|)\ge(k-\sqrt k)/2.
\]
So inversion turns a clique, whose original signs are all positive, into a face with nearly balanced signs. In this way cliques contribute to the residual piece: \(E_{q,q}\) is nonempty as soon as a clique of size \(k<p\) satisfies \((k-\sqrt k)/2\ge q+1\).

\subsection{Spectral pseudorandomness does not force logarithmic index}\label{ssec:8.5}
Random matrices, with high probability, and the Sylvester Hadamard matrices have operator norm \(O(\sqrt N)\), the smallest possible order for an \(N\times N\) sign matrix, and their index is \(O(\log N)\) by~\cite{FHV26}. Since \(\|P_p\|=\sqrt p+1\), it is natural to ask whether a small operator norm alone forces logarithmic index. The next family shows that it does not. It plants a Hanner matrix, with its rows repeated to reach side \(\sqrt N\), in a Sylvester Hadamard matrix of order \(N\). The planted block changes the norm by at most \(2\sqrt N\), but it brings a sphere of dimension of order \(\log^2N\) into the sign complex.

\begin{resultenv}{Theorem}{res:8.8}{A pseudorandom family with large index}
Let \(k\ge0\) be an integer, and let \(t=2^k\), \(d=t^2\), \(n_0=2^{3(t-1)}\), \(N=n_0^2=2^m\), and \(m=6(t-1)\). Let \(A_{2k}\) be the Hanner matrix of Theorem~\ref{res:4.3}(i), of size \(2^{2(t-1)}\times2^{3(t-1)}\), with \(\operatorname{coInd}=d-1\) and \(\operatorname{rank}=d\). Repeat its rows to obtain an \(n_0\times n_0\) matrix \(\overline A\). Define
\[
(D_t)_{ij}
=
\begin{cases}
\overline A_{ij},&i,j<n_0,\\
(-1)^{i\cdot j},&\text{otherwise},
\end{cases}
\]
for \(0\le i,j<N\), with the \(m\)-bit inner product. Then \(\|D_t\|\le3\sqrt N\) and \(d-1\le\operatorname{Ind}(D_t)\le2d+4m+3\), and \(\operatorname{Ind}(D_t)=\Theta(\log^2N)\) as \(k\to\infty\).
\end{resultenv}

\begin{proof}
\proofstep{Sizes.} Theorem~\ref{res:4.3}(i) with index \(2k\) gives \(N_{2k}=2^{2(t-1)}\), \(M_{2k}=2^{3(t-1)}\), and \(\operatorname{Ind}=\operatorname{coInd}=d-1\), \(\operatorname{rank}=d\). So \(\overline A\) repeats each row \(2^{t-1}\) times.

\proofstep{The norm.} We have \(D_t=H_N+E\), where \(E\) is supported on the block of the first \(n_0\) rows and the first \(n_0\) columns, with entries in \(\{0,\pm2\}\). So \(\|E\|\le\|E\|_F\le2n_0\), and as \(\|H_N\|=\sqrt N=n_0\),
\[
\|D_t\|
\le
\sqrt N+2n_0
=
3\sqrt N.
\]

\proofstep{The lower bound.} Every simplex of \(S(\overline A)=S(A_{2k})\) is a subset of a signed row of the replaced block, hence a simplex of \(S(D_t)\). So \(S(A_{2k})\) is an equivariant subcomplex of \(S(D_t)\). This gives the coindex lower bound \(d-1\), and \(\operatorname{Ind}\ge\operatorname{coInd}\).

\proofstep{The upper bound.} Split the domain into the first \(n_0\) columns \(X_1\) and the other columns \(X_2\). On \(X_2\) the matrix \(D_t\) coincides with \(H_N\). On \(X_1\) its rows are those of \(\overline A\), for \(i<n_0\), together with rows of \(H_N\) restricted to \(X_1\), for \(i\ge n_0\). The concatenation bounds of~\cite{BHHLT26} (Corollary 1.13 and Lemma 7.1) apply, first to the disjoint domains and then to the union of classes on \(X_1\); for the latter, the proof of~\eqref{eq:union-list} keeps track of the error margin. They give
\[
\operatorname{lr}(D_t)\le\operatorname{lr}(D_t|_{X_1})+\operatorname{lr}(D_t|_{X_2}),
\qquad
\operatorname{lr}(D_t|_{X_1})\le\operatorname{lr}(\overline A)+\operatorname{lr}(H_N|_{X_1}).
\]
The first inequality is the horizontal case of~\cite[Corollary 1.13]{BHHLT26}, for a matrix \([A_3\ A_4]\), and the second is its vertical case, which is~\cite[Lemma 7.1]{BHHLT26}.

Further \(\operatorname{lr}(\overline A)\le\operatorname{signrk}(\overline A)\le\operatorname{rank}\overline A=d\) (Section~\ref{ssec:2.1}). Restriction does not increase list replicability, since a learner for the whole domain serves distributions supported on a subset of it, and \(\operatorname{lr}(H_N)=m+1\) by Theorem~\ref{res:5.5}. Hence
\[
\operatorname{lr}(D_t)\le d+2(m+1).
\]
Applying Theorem~\ref{res:a} yields
\[
\operatorname{Ind}(D_t)\le2\operatorname{lr}(D_t)-1\le2d+4m+3.
\]
Since \(d=t^2\) and \(m=6(t-1)=\log_2N\), both bounds are of order \(t^2=\Theta(\log^2N)\), which gives the asserted order.
\end{proof}

Thus, for $N\times N$ sign matrices $D$, the implication
\[
\|D\|=O(\sqrt N)\Longrightarrow\operatorname{Ind}S(D)=O(\log N)
\]
is false. The matching general $O(\log^2N)$ index bound for square matrices is due to~\cite{FHMNV26v1}, and it also follows from Theorem~\ref{thm:main} and Theorem~\ref{res:a}. Consequently the spectral information \(\|P_p\|=\sqrt p+1\) cannot by itself decide the order of \(\operatorname{Ind}S(P_p)\).

\subsection{\texorpdfstring{The index of \(P_{13}\) and small primes}{The index of P13 and small primes}}\label{ssec:8.6}
At small primes the index can be examined directly. We determine it exactly at \(p=5\) and \(p=13\), by combining low-dimensional sign representations with an integral obstruction computation. We then record what the results of this section give at all primes \(p\equiv1\pmod4\) up to \(197\), and what a complete enumeration of closed faces gives at the five smallest of them.

For \(P_5\), choose the two-dimensional vectors \(v_0,\ldots,v_4\) to be
\[
(2,0),\quad(1,3),\quad(-3,2),\quad(-3,-2),\quad(1,-3).
\]
The matrix of the inner products \(\langle v_r,v_i\rangle\) is
\[
\begin{pmatrix}
4&2&-6&-6&2\\
2&10&3&-9&-8\\
-6&3&13&5&-9\\
-6&-9&5&13&3\\
2&-8&-9&3&10
\end{pmatrix},
\]
and since the nonzero squares modulo \(5\) are \(\pm1\), they satisfy
\[
\operatorname{sign}\langle v_r,v_i\rangle=(P_5)_{ri}.
\]
They give a sign representation of dimension \(2\), hence \(\operatorname{Ind}(P_5)\le1\) by the odd map described after Lemma~\ref{res:5.0}. The columns \(0,1\) are signed-shattered: the signed rows \(0^+\) and \(4^+\) realize the patterns \((+,+)\) and \((+,-)\), and their negatives realize the other two. So \(\operatorname{coInd}(P_5)\ge1\) and \(\operatorname{Ind}(P_5)=1\).

For \(P_{13}\), a sign representation comes from the constant function and the characters of \(\mathbb Z_{13}\) of frequencies \(1\) and \(3\). We use the following scaling, in which \(c_r\) and \(s_r\) are the values \(1000\cos(2\pi r/13)\) and \(1000\sin(2\pi r/13)\), \(r=0,\ldots,12\), rounded to integers:
\[
c=(1000,885,568,121,-355,-749,-971,-971,-749,-355,121,568,885),
\]
\[
s=(0,465,823,993,935,663,239,-239,-663,-935,-993,-823,-465).
\]
Indices are modulo \(13\), and
\[
U_r=(937500,5c_r,5s_r,4c_{3r},4s_{3r}),
\]
\[
V_i=(1,c_i,s_i,c_{3i},s_{3i}).
\]
Up to the rounding, \(\langle U_r,V_i\rangle\) equals \(10^6\) times the value at \(r-i\) of the trigonometric polynomial \(\frac{15}{16}+5\cos(2\pi x/13)+4\cos(6\pi x/13)\), whose sign on \(\mathbb Z_{13}\) is the first column of \(P_{13}\). The next proposition records the exact margin of this representation and determines the index. Dimension five alone gives only \(\operatorname{Ind}S(P_{13})\le4\), and the proof lowers this bound by one through a vanishing Euler class.

\begin{resultenv}{Proposition}{res:8.9}{The index of $P_{13}$}
Let $B=P_{13}$. For all $169$ entries,
\[
 B_{ri}\langle U_r,V_i\rangle\ge102714,
\]
with equality exactly at $(5,7),(6,8),(7,5),(8,6)$.
Moreover, with $\operatorname{swh}$ the Stiefel--Whitney height defined in Section~\ref{ssec:2.1},
\[
 \operatorname{coInd}S(P_{13})
 =\operatorname{swh}S(P_{13})
 =\operatorname{Ind}S(P_{13})=3.
\]
\end{resultenv}

\begin{proof}
The integer inequalities and the list of equality cases are verified by evaluating the $169$ exact dot products (Appendix~\ref{sec:computation}).
They give a sign representation in dimension five, and so an odd map to $S^4$. To obtain the
sharper topological upper bound, we compute the integral obstruction
to a map to $S^3$.

\proofstep{The lower bound.}
The columns $\{0,1,2,4\}$ are signed-shattered: the signed rows
\[
1^+,\ 7^-,\ 0^+,\ 10^+,\ 3^+,\ 12^+,\ 2^-,\ 9^+
\]
realize, in this order, the patterns $(+,+,+,+)$, $(+,+,+,-)$, $(+,+,-,+)$, $(+,+,-,-)$, $(+,-,+,+)$, $(+,-,+,-)$, $(+,-,-,+)$, $(+,-,-,-)$, and their negatives realize the other eight of the $16$ patterns. Hence their cross-polytope boundary
is an equivariant copy of $S^3$ in $X=S(P_{13})$. This gives
$\operatorname{coInd}X\ge3$ and, by naturality, $w_1^3\ne0$ on the quotient
$Y=X/\mathbb Z_2$.

\proofstep{The integral obstruction.}
Let $\lambda$ be the real line bundle associated with $X\to Y$.
Sections of $4\lambda$ are the odd maps $X\to\mathbb R^4$, so an equivariant map $X\to S^3$ is equivalent to a nowhere-zero section
of $4\lambda$. This bundle is oriented, since its transition maps
have determinant $(-1)^4=1$. Its Euler class is therefore an ordinary
integer cohomology class in degree four.

\proofstep{A cocycle for the Euler class.}
Order the vertices of every simplex by column number. A simplex has at most one vertex in each column, and the antipodal map preserves this order. So $Y$ is an ordered $\Delta$-complex, and cochains on $Y$ are the antipodally invariant cochains on $X$. Represent a
quotient simplex by the lift whose first vertex has positive sign. Let $s$ be the sign of a vertex, and let $a$ be the cochain with value $(t-s)/2$ on a lifted edge with endpoint
signs $s,t$. It changes sign under the antipodal map, and it represents the twisted Euler class of $\lambda$. One obtains it by assigning sign $s$ to the
lifted vertex section: the signed number of zeros on the edge is zero
when the signs agree, and is $1$ or $-1$ when they differ.
Multiplicativity of Euler classes, with tensor products of these local
coefficient systems, gives the following ordinary integer cocycle for
$e(4\lambda)=e(\lambda)^4$:
\[
 E([v_0,\ldots,v_4])
       =\prod_{i=0}^3\frac{s_{i+1}-s_i}{2}.
\]
Changing the lift negates each factor and leaves the product unchanged.

\proofstep{An identification without orientations.}
The identification of $E$ with $e(4\lambda)$ can also be verified without tracking orientations. The complex $X$ lies in the boundary $\partial\Diamond_{13}$ of the cross-polytope, on the same vertices, and the column order and the antipodal map extend to it. So the same formula defines a cochain $\widehat E$ on the ordered $\Delta$-complex $\partial\Diamond_{13}/\mathbb Z_2=\mathbb{RP}^{12}$, with restriction $E$ to $Y$, and $\lambda$ on $Y$ is the restriction of the line bundle of $\partial\Diamond_{13}\to\mathbb{RP}^{12}$. The cochain $a=\delta(s/2)$ is closed, so $\widehat E=a\smile a\smile a\smile a$ is a cocycle.

Modulo two, $a$ represents $w_1$: along a lifted edge path its values telescope to $(s_{\mathrm{end}}-s_{\mathrm{start}})/2$, which is odd exactly when the lift of a loop ends at the antipode of its starting point. So $\widehat E$ reduces modulo two to $w_1^4\ne0$ in $H^4(\mathbb{RP}^{12};\mathbb F_2)$. The Euler class of $4\lambda$, for either orientation, reduces modulo two to the top Stiefel--Whitney class (\cite[Section 9]{MS74}), and $w_4(4\lambda)=w_1^4$ by the Whitney product formula $w(4\lambda)=(1+w_1)^4$. Finally $H^4(\mathbb{RP}^{12};\mathbb Z)=\mathbb Z_2$, and reduction modulo two is injective on it, since its kernel is $2H^4(\mathbb{RP}^{12};\mathbb Z)=0$. Hence $[\widehat E]=e(4\lambda)$ on $\mathbb{RP}^{12}$, and restricting to $Y$ gives $[E]=e(4\lambda)$.

\proofstep{Vanishing of the class.}
Exact computation (Appendix~\ref{sec:computation}) gives an integer
three-cochain $b$ on $Y$, found by elimination with pivots $1$ or $-1$, followed by back substitution. All signed-row simplices and their antipodal quotients are constructed, which gives
$13\,572$ quotient four-simplices, and for every one of them the computation verifies the equality
\[
 \sum_{i=0}^4(-1)^i
 b([v_0,\ldots,\widehat v_i,\ldots,v_4])
        =E([v_0,\ldots,v_4]).
\]
Thus $E=\delta b$, and $e(4\lambda)=0$ over the integers.

\proofstep{Dimension reduction.}
The same computation gives a sequence of elementary collapses in antipodal pairs.
For each pair $(\sigma,\tau)$, it verifies that $\tau$ is
maximal and that $\sigma$ is its codimension-one face with no other
coface, and it then deletes $(\sigma,\tau)$ and $(-\sigma,-\tau)$.
These conditions are tested against the current complex at every step,
with cofaces recomputed from the vertex set.

A simplex $\tau$ and its antipode $-\tau$ are disjoint, because a simplex contains no antipodal pair of vertices. So the two elementary collapses of a pair have disjoint supports and together form an equivariant strong deformation retraction. The final complex is verified to be closed under faces and under the antipodal map, and it has dimension four
and $15\,970$ nonempty faces. Hence $X$ equivariantly retracts to a
four-dimensional complex $X'$, by a map $\varrho$, and $Y$ retracts to $Y'=X'/\mathbb Z_2$.

\proofstep{Extension.}
The fiber of the unit sphere bundle of $4\lambda|_{Y'}$ is $S^3$.
Its homotopy groups in dimensions $0,1,2$ vanish, so a section extends
over the three-skeleton. In terms of odd maps, extend skeleton by skeleton, over one simplex of each antipodal pair by~\cite[Lemma 4.7]{Hatcher02}, and over the other simplex by oddness.

The only further obstruction over the
four-dimensional base is the primary obstruction, the class in $H^4(Y';\pi_3(S^3))$. The orientation action on $\pi_3(S^3)=\mathbb Z$ is trivial, because the antipodal map of $S^3$ has degree $(-1)^4=1$. By~\cite[Theorem 12.5]{MS74} the primary obstruction is the Euler class of $4\lambda|_{Y'}$, the restriction of $e(4\lambda)$.
The class just computed vanishes. So, after a change of the section on the three-skeleton, it extends over all of
$Y'$.
This yields an equivariant map $X'\to S^3$ and, by composition with $\varrho$, one on $X$. Therefore $\operatorname{Ind}X\le3$.
Together with
$\operatorname{coInd}X\le\operatorname{swh}X\le\operatorname{Ind}X$
and the lower bound, this proves the equalities.
\end{proof}

\subsubsection*{Analytic bounds and exact finite values}

We now tabulate the results of this section at small primes. For \(a\ge1\) let \(m_a(p)=\max\{|R(F)|:F\in S(P_p),\ |F^+|\ge a,\ |F^-|\ge a\}\), with value zero if there is no such face. This is the largest number of compatible signed rows of a face with at least \(a\) vertices of each sign. These numbers describe the residual piece of Theorem~\ref{res:6.7}, because \(E_{a-1,t}\) is nonempty if and only if \(m_a(p)\ge t+1\).

Table~\ref{tab:paley-exact} gives exact values at the five smallest primes \(p\equiv1\pmod4\). They are obtained by exact computation (Appendix~\ref{sec:computation}): we enumerate every nonempty signed closure, that is, every nonempty intersection of facets, and we compute from these closures the height, the maximum switchable-square size, and the numbers \(m_a\).

\begin{table}[ht]
\centering\small\begin{tabular}{lllll}
\toprule
\(p\) & Nonempty closed faces & \(h\) & \(\beta_p\) & \((m_2,m_3,m_4)\) \\
\midrule
5 & 50 & 5 & 3 & \((2,0,0)\) \\
13 & 2522 & 7 & 4 & \((4,2,2)\) \\
17 & 11152 & 7 & 4 & \((4,3,2)\) \\
29 & 240874 & 9 & 5 & \((6,4,3)\) \\
37 & 1304842 & 9 & 5 & \((8,4,4)\) \\
\bottomrule\end{tabular}
\caption{Exact values from a complete enumeration of the closed faces of \(S(P_p)\).}\label{tab:paley-exact}
\end{table}

The maximum switchable-square size is computed as the maximum of \(\min(|F|,|R(F)|)\) over closed faces \(F\). This suffices because passing from a face to its closure does not change its compatible signed rows and can only add vertices. So a \(k\times k\) switchable square yields a closed face with \(\min(|F|,|R(F)|)\ge k\), and conversely such a closed face contains a \(k\times k\) switchable square. The same remark shows that the maxima defining \(m_a\) are attained on closed faces.

Table~\ref{tab:paley-bounds} covers all primes \(p\equiv1\pmod4\) up to \(197\). It lists \(C_p\), the rank \(r_2\), an interval for the index, a pair \((r,c)\) for which no \(r\times c\) switchable rectangle exists, and upper bounds for \(m_2,m_3,m_4\). The integer inequalities behind all its rows are verified by exact computation (Appendix~\ref{sec:computation}).

\begin{table}[ht]
\centering\small\begin{tabular}{llllllll}
\toprule
\(p\) & \(C_p\) & \(r_2\) & \(\operatorname{Ind}\) interval & Excluded \((r,c)\) & \(m_2\le\) & \(m_3\le\) & \(m_4\le\) \\
\midrule
5 & 10 & 4 & \([1,1]\) & \((3,4)\) & \(=2\) & \(=0\) & \(=0\) \\
13 & 21 & 12 & \([3,3]\) & \((5,4)\) & \(=4\) & \(=2\) & \(=2\) \\
17 & 26 & 8 & \([3,6]\) & \((5,5)\) & 6 & 4 & \(=2\) \\
29 & 40 & 28 & \([4,8]\) & \((6,7)\) & 10 & 6 & 4 \\
37 & 50 & 36 & \([5,8]\) & \((7,7)\) & 12 & 8 & 6 \\
41 & 54 & 20 & \([3,12]\) & \((7,8)\) & 12 & 8 & 6 \\
53 & 68 & 52 & \([3,13]\) & \((7,9)\) & 16 & 10 & 8 \\
61 & 77 & 60 & \([3,14]\) & \((9,8)\) & 18 & 12 & 8 \\
73 & 91 & 36 & \([3,16]\) & \((9,10)\) & 22 & 14 & 10 \\
89 & 108 & 44 & \([4,18]\) & \((10,11)\) & 26 & 18 & 12 \\
97 & 117 & 48 & \([4,18]\) & \((11,10)\) & 28 & 18 & 14 \\
101 & 122 & 100 & \([4,19]\) & \((11,11)\) & 30 & 20 & 14 \\
109 & 130 & 108 & \([4,20]\) & \((11,12)\) & 32 & 20 & 16 \\
113 & 135 & 56 & \([4,20]\) & \((11,12)\) & 32 & 22 & 16 \\
137 & 161 & 68 & \([4,22]\) & \((13,12)\) & 40 & 26 & 20 \\
149 & 174 & 148 & \([4,23]\) & \((13,13)\) & 42 & 28 & 20 \\
157 & 183 & 156 & \([4,24]\) & \((13,14)\) & 44 & 30 & 22 \\
173 & 200 & 172 & \([4,25]\) & \((13,15)\) & 50 & 32 & 24 \\
181 & 208 & 180 & \([4,26]\) & \((14,15)\) & 52 & 34 & 26 \\
193 & 221 & 96 & \([4,26]\) & \((15,14)\) & 54 & 36 & 26 \\
197 & 226 & 196 & \([4,27]\) & \((15,15)\) & 56 & 36 & 28 \\
\bottomrule\end{tabular}
\caption{Bounds for the index and the numbers \(m_a\) at all primes \(p\equiv1\pmod4\) up to \(197\). An entry marked \(=\) is an exact value.}\label{tab:paley-bounds}
\end{table}

Table~\ref{tab:paley-bounds} is obtained as follows. The columns \(C_p\) and \(r_2\) are given by Theorem~\ref{res:8.1}. Call a pair \((r,c)\) \emph{excluded} if \(rc>C_p\) (Theorem~\ref{res:8.1}) or if one of the integer criteria of Theorem~\ref{res:8.2} holds. Then no \(r\times c\) switchable rectangle exists. Each pair in the column ``Excluded'' is an excluded pair with the least value of \(r+c\) among the excluded pairs, and Theorem~\ref{res:8.3} gives
\[
h\le r+c-2,
\qquad
\operatorname{Ind}\le r+c-3.
\]
This is the upper end of the interval for \(p\ge41\). At \(p=17,29,37\) it gives \(7\), \(10\), and \(11\), and the table records the smaller upper end \(h-1\) (Theorem~\ref{res:6.1}), with \(h\) from Table~\ref{tab:paley-exact}. The intervals at \(p=5\) and \(p=13\) are the exact values determined above, by the integer sign representation of \(P_5\) and by Proposition~\ref{res:8.9}.

Theorem~\ref{res:8.4} gives the lower end \(3\) for \(p\ge29\) and \(4\) for \(p\ge89\), and these are the lower ends recorded for \(p\ge41\). The lower ends \(3\), \(4\), and \(5\) at \(p=17,29,37\) come from signed-shattered column sets. These sets, together with the one for \(p=13\), are
\[
\{0,1,2,4\}\ (p=13),\quad\{0,1,2,3\}\ (p=17),\quad\{0,1,2,3,7\}\ (p=29),\quad\{0,1,2,4,12,29\}\ (p=37),
\]
each verified by a direct check of all signed patterns (Appendix~\ref{sec:computation}).

The bounds on \(m_a\) follow from excluded pairs. A face with \(|F^+|,|F^-|\ge a\) and \(|R(F)|\ge m+1\) contains an \((m+1)\times2a\) switchable rectangle. So if \((m+1,2a)\) is excluded, then \(m_a\le m\). Each entry \(m_a\le\) is the least such \(m\), and an entry marked \(=\) is the exact value from Table~\ref{tab:paley-exact}. So at \(p=17,29,37\) an unmarked entry records the bound from excluded pairs, and Table~\ref{tab:paley-exact} gives the exact value.

Finally, the residual piece at the scale of the general bound is empty at all these primes. Let $q=\lceil\log_2^2p\rceil$. A face in $E_{q,q}$ has at least $2(q+1)$ vertices and at least $q+1$ compatible signed rows, so it contains a $(q+1)\times2(q+1)$ switchable rectangle. By Theorem~\ref{res:8.1}, nonemptiness of $E_{q,q}$ therefore requires
\[
2(q+1)^2\le C_p.
\]
This inequality fails at every prime of Table~\ref{tab:paley-bounds}, so \(E_{q,q}\) is empty there. This reflects the size of these primes, since \(q\) exceeds \(\sqrt p\) throughout the table.

\part{Limits of specific representations}\label{part:limits}
The analytic construction of Section~\ref{ssec:3} passes through a specific encoding of
the realizable distributions, a power map together with a vote decoded from it. This
part studies what such encodings can and cannot achieve. Section~\ref{ssec:7} gives
lower bounds for the entropy of power maps and for the dimension of affine statistics,
and it relates the tilt dimension to the covering dimension of the vote image.
Section~\ref{app:vote-geometry} computes the sign-compatible geometry of the vote image
on Sylvester Hadamard matrices.

\section{Limits of power maps, affine statistics, and votes}\label{ssec:7}
The constructions of Section~\ref{ssec:3} encode a realizable distribution by a power map, decode a vote from that encoding, and transport the resulting labels by exponential tilting. Each step has a natural numerical measure: the covering entropy of the power-map image, the dimension of the statistic from which a vote is decoded, and the covering dimension of the vote image. The proof of Theorem~\ref{thm:main} follows a different path and uses none of these quantities. This section determines how far the quantities themselves can go. The answer shows where the net argument of Section~\ref{ssec:3} stops short of Theorem~\ref{thm:main}, and it measures what explicit encodings of a distribution cost.

On Hadamard matrices of order \(n=2^{64t}\) the power-map image has covering entropy at least \(10t^3\) for every exponent, so an argument through a net of this image and a distance-preserving linear map cannot reach the order \(\log N\log M\) of Theorem~\ref{thm:main}. On a symmetric class, affine statistics require dimension at least the sign rank, which on symmetrized Hadamard matrices exceeds the list number exponentially. These two bounds concern the power-map net and affine statistics; other encodings lie outside their scope. On the vote image, the estimates of Section~\ref{ssec:7.8} leave the fixed-scale covering dimension for \(H_{2^m}\) between \(m\) and \(O(m^3)\).

The same analysis gives two-sided estimates for the least value of each quantity. The least covering entropy of an odd image of the profile space is comparable to the index, so the large entropy above is a property of the power maps and not of the profile space. The least dimension of an affine statistic on a symmetric class is exactly the sign rank. The transport step is governed by an atomwise condition on the vote, the low-vote condition, and the least dimension of a parameterization satisfying it lies between \(\operatorname{lr}-1\) and the covering dimension of the vote image.

Sections~\ref{ssec:7.1}--\ref{ssec:7.4} treat the power map and other odd encodings of the profile, Section~\ref{ssec:7.5} affine statistics and the condition used for transport, and Section~\ref{ssec:7.8} the vote image. Section~\ref{app:vote-geometry} continues the comparison of Section~\ref{ssec:7.8} with sign-compatible covers.

For a metric subspace $A$, $\mathcal N(A,\rho)$ is the least number
of closed radius-$\rho$ balls with centers in $A$ that cover $A$, and we call
$\log_2\mathcal N(A,\rho)$ the \emph{covering entropy} of $A$ at scale $\rho$.
The metric is Euclidean on subsets of unit spheres (power-map images,
coordinate-function images, and images of odd maps to a sphere) and
$\ell_\infty$ on vote images. Packing lower bounds below also hold for
arbitrary centers, because two points of a closed ball of radius $\rho$
are at distance at most $2\rho$ whatever its center is.

Two elementary facts are used repeatedly. If points of $\mathbb R^K$ lie
in a ball of radius $R$ and have pairwise distances at least $\rho>0$,
then
\begin{equation}\label{eq:volume-packing}
 \#\{\text{points}\}\le\left(1+\frac{2R}\rho\right)^K,
\end{equation}
since the open balls of radius $\rho/2$ around them are disjoint and lie
in the ball of radius $R+\rho/2$. Consequently, a maximal subset of a set
$A$ in the unit sphere of a $d$-dimensional Euclidean space, among those
with pairwise distances greater than $\rho$, has at most
$(1+2/\rho)^d$ points, and the closed $\rho$-balls around its points
cover $A$. Further, for nonzero vectors $a,b$ of a normed space,
\begin{equation}\label{eq:normalization}
 \left\|\frac a{\|a\|}-\frac b{\|b\|}\right\|
 \le\frac{\|a-b\|}{\|a\|}
    +\|b\|\,\left|\frac1{\|a\|}-\frac1{\|b\|}\right|
 \le2\,\frac{\|a-b\|}{\|a\|}.
\end{equation}

\subsection{Packing lower bounds for power maps}\label{ssec:7.1}

The index bound of Theorem~\ref{res:3.1} has two stages. Lemma~\ref{res:3.0} approximates the image \(u_{p_N}(X_B)\) by the normalized powers of averages of \(s_N=2^{16}p_N^2\) signed columns, which form a net of at most \((2M)^{s_N}\) points. Lemma~\ref{res:3.0prime} then maps a net of \(L\) points to dimension of order \(\ln L\) by a linear map that preserves distances up to a constant factor. This gives a target dimension of order \(\log^2N\log M\). To reach the order \(\log N\log M\) along these lines, one would need averages of fewer columns, another exponent, or a smaller target dimension for such a linear map.

The next theorem shows that on Hadamard matrices none of these is available. For every exponent, the image \(u_p(X)\) contains \(2^{10t^3}\) points at pairwise distance greater than \(1/8\), where \(n=2^{64t}\). Consequently the number of averaged columns must be at least of order \(p_n^2\), as in Lemma~\ref{res:3.0}, and every linear map that preserves distances on a net has target dimension at least of cubic order in \(\log n\).

\begin{resultenv}{Theorem}{res:7.1}{A packing lower bound for power maps on Hadamard matrices}
Let \(t\ge1\) be an integer, let \(n=2^{64t}\), let \(H_n(r,j)=(-1)^{r\cdot j}\) for \(r,j\in\mathbb F_2^{64t}\), and put \(X=X_{H_n}\).

(i) For every real \(p\ge1\),
\[
\log_2\mathcal N(u_p(X),1/16)\ge10t^3.
\]

(ii) Let \(Y_s\) be the set of averages of \(s\) signed columns of \(H_n\), repetitions allowed. If \(u_p(Y_s\setminus\{0\})\) is a \(1/16\)-net of \(u_p(X)\), then
\[
s\ge\frac{10t^3}{64t+1}\ge\frac2{13}t^2.
\]

(iii) Under the net assumption of (ii), for \(p=p_n=\lceil16\ln(2n)\rceil\),
\[
s\ge2^{-22}p_n^2.
\]

(iv) Let \(Z\subset S^{n-1}\) be a finite set of unit vectors such that every point of \(u_p(X)\) lies within Euclidean distance \(1/16\) of \(Z\). Call a linear map \(T:\mathbb R^n\to\mathbb R^K\) \emph{distance-preserving on \(Z\)} if \(\tfrac12\|z-z'\|_2^2\le\|Tz-Tz'\|_2^2\le\tfrac32\|z-z'\|_2^2\) for all \(z,z'\in Z\cup\{0\}\), the property provided by Lemma~\ref{res:3.0prime}. Every such map satisfies
\[
K\ge\frac54t^3.
\]
\end{resultenv}

The packing points are convex combinations of the columns that put mass \(4p/(4p+t)\) on the constant column and spread the rest uniformly over a set \(S\) of \(t^2\) further columns. On such a point the power map acts, up to a scalar, as the function \(\phi_p(1+Z_S/p)\), where \(Z_S\) is a normalized sum of the characters in \(S\). For at least one eighth of the sets \(S\) and for every \(p\), this function is close to \(1+Z_S\) in a normalized \(L^2\) norm, and the orthogonality of characters then separates the images of sets with small intersection. We first fix the notation and establish these facts in four lemmas.

With \(t\) and \(n\) as in the theorem and a real \(p\ge1\), put \(G=\mathbb F_2^{64t}\), \(k=t^2\) and \(\chi_j(r)=(-1)^{r\cdot j}\), so that the column \(q_j\) of \(H_n\) is the function \(\chi_j\) of the row index. Use the normalized Hilbert norm \(\|f\|_H^2=\frac1n\sum_{r\in G}f(r)^2\), with inner product \(\langle f,f'\rangle_H=\frac1n\sum_rf(r)f'(r)\). The isometry \(f\mapsto f/\sqrt n\) transfers it to the Euclidean norm, that is, \(\|f\|_2=\sqrt n\,\|f\|_H\). The characters \(\chi_j\), \(j\in G\), are orthonormal for this inner product, and \(\chi_0\) is the constant function \(1\). The \emph{positive row-\(0\) simplex} is the set of convex combinations of the columns \(q_j\). For \(S\subseteq G\setminus\{0\}\) with \(|S|=k\) write
\[
Z_S(r)=\frac1{4t}\sum_{j\in S}\chi_j(r),
\qquad
F_S(r)=\phi_p(1+Z_S(r)/p).
\]
We call \(S\) a \emph{good support} if
\[
\|F_S-1-Z_S\|_H\le\frac1{16}.
\]
The set of good supports may depend on \(p\).

\begin{lemma}[Uniform scalar remainder]\label{lem:scalar-remainder}
Let \(Z=\frac1{4t}\sum_{i=1}^{t^2}\varepsilon_i\) with independent uniform signs. Then \(\mathbb EZ=0\) and \(\mathbb EZ^2=t^2/(16t^2)=\lambda:=1/16\), and for real \(a\),
\[
\mathbb Ee^{aZ}=\cosh\bigl(a/(4t)\bigr)^{t^2}\le e^{a^2/32}.
\]
For \(p\ge1\) put \(R_p(Z)=\phi_p(1+Z/p)-1-Z\). Then \(\mathbb ER_p(Z)^2<13/4000\).
\end{lemma}

\begin{proof}
The moments and the identity for the generating function follow from the independence of the signs, and the inequality from \(\cosh y\le e^{y^2/2}\).

\proofstep{A bound on \(R_p(Z)^2\) for each outcome.} If \(Z\ge-p\), then \(1+Z/p\ge0\) and \(\phi_p(1+Z/p)=(1+Z/p)^p\). The function \(x\mapsto(1+x/p)^p\) is convex on \([-p,\infty)\) and has tangent \(1+x\) at \(x=0\), so \((1+Z/p)^p\ge1+Z\). Also \(\ln(1+x)\le x\) gives \((1+Z/p)^p\le e^Z\). Hence \(0\le R_p(Z)\le e^Z-1-Z\).

If \(Z<-p\), put \(v=-Z>p\ge1\). Then \(\phi_p(1+Z/p)=-(v/p-1)^p\), so \(R_p(Z)=(v-1)-(v/p-1)^p\) is a difference of two nonnegative numbers, and \(R_p(Z)^2\le(v-1)^2+(v/p-1)^{2p}\). Moreover \(e^Z-1-Z=e^{-v}-1+v\ge v-1\ge0\), and \(y\le e^{y-1}\) with \(y=v/p-1\) gives \((v/p-1)^p\le e^{v-2p}\). Thus in this case
\[
R_p(Z)^2\le(e^Z-1-Z)^2+e^{2v-4p}=(e^Z-1-Z)^2+e^{-4p}e^{-2Z}.
\]
Together, on every outcome \(R_p(Z)^2\le(e^Z-1-Z)^2+e^{-4p}e^{-2Z}\mathbf 1_{\{Z<-p\}}\).

\proofstep{The tail term.} Put \(c=16p-2\ge14\). On \(\{Z<-p\}\) we have \(e^{-cZ}\ge e^{cp}\), so the moment generating function with \(a=-(2+c)=-16p\) gives
\[
\mathbb E\bigl[e^{-4p}e^{-2Z}\mathbf 1_{\{Z<-p\}}\bigr]
\le e^{-4p-cp}\,\mathbb Ee^{-(2+c)Z}
\le e^{-4p-(16p-2)p+8p^2}
=e^{-8p^2-2p}
\le e^{-10}<\frac1{20000},
\]
where the last inequality uses \(e^{10}>(27/10)^{10}>20000\).

\proofstep{The main term.} Let \(g\) be a centered Gaussian of variance \(\lambda\). For real \(y\), the power series \(\cosh(ay)=\sum_j(ay)^{2j}/(2j)!\) is dominated coefficientwise by \(e^{a^2y^2/2}=\sum_j(ay)^{2j}/(2^jj!)\), because \((2j)!\ge2^jj!\). Multiplying \(t^2\) such series with \(y=1/(4t)\), all of whose coefficients are nonnegative, shows that the power series in \(a\) of \(\mathbb Ee^{aZ}\) is dominated coefficientwise by that of \(\mathbb Ee^{ag}=e^{a^2/32}\). The coefficient of \(a^l\) in these series is \(\mathbb EZ^l/l!\), respectively \(\mathbb Eg^l/l!\), so \(\mathbb EZ^l\le\mathbb Eg^l\) for all \(l\ge0\).

The function \(f(z)=(e^z-1-z)^2\) is the square of \(\sum_{l\ge2}z^l/l!\), hence \(f(z)=\sum_{l\ge4}f_lz^l\) with \(f_l\ge0\). Summing termwise, which is legitimate since \(Z\) is bounded and \(\mathbb Ef(|g|)<\infty\), we get \(\mathbb Ef(Z)\le\mathbb Ef(g)\). Now \(f(z)=e^{2z}-2(1+z)e^z+(1+z)^2\), and \(\mathbb Ee^{2g}=e^{2\lambda}\), \(\mathbb Ee^g=e^{\lambda/2}\), \(\mathbb E[ge^g]=\lambda e^{\lambda/2}\) (the derivative of \(a\mapsto e^{\lambda a^2/2}\) at \(a=1\)), \(\mathbb E(1+g)^2=1+\lambda\). Therefore
\[
\mathbb E(e^Z-1-Z)^2
\le e^{2\lambda}-2(1+\lambda)e^{\lambda/2}+1+\lambda
=e^{1/8}-\frac{17}8e^{1/32}+\frac{17}{16}.
\]
Use the lower bound \(e^{1/32}\ge\sum_{j\le3}(1/32)^j/j!\). From \(\sum_{j\ge4}x^j/j!\le\frac{x^4}{4!}\sum_{i\ge0}(x/5)^i\) at \(x=1/8\), we also have the upper bound \(e^{1/8}\le\sum_{j\le3}(1/8)^j/j!+\frac{(1/8)^4}{24}\cdot\frac{40}{39}\). Exact rational arithmetic gives
\[
e^{1/8}-\frac{17}8e^{1/32}+\frac{17}{16}\le\frac{195913}{61341696}<\frac2{625},
\]
the last inequality being \(195913\cdot625=122445625<122683392=2\cdot61341696\). Adding the two contributions gives \(\mathbb ER_p(Z)^2<2/625+1/20000=13/4000\).
\end{proof}

\begin{lemma}[At least one eighth of the supports are good]\label{lem:good-supports}
For every \(p\ge1\), at least \(1/8\) of the \(k\)-subsets of \(G\setminus\{0\}\) are good supports.
\end{lemma}

\begin{proof}
Let \(j_1,\ldots,j_k\) be independent uniform elements of \(G\), and let \(\mathcal A\) be the event that they are distinct and nonzero. Conditionally on \(\mathcal A\), the set \(S=\{j_1,\ldots,j_k\}\) is a uniformly random \(k\)-subset of \(G\setminus\{0\}\), by symmetry. A union bound over the \(k\) events \(j_i=0\) and the \(\binom k2\) events \(j_i=j_{i'}\) shows that \(\delta:=\mathbb P(\mathcal A^c)\le k/n+\binom k2/n=k(k+1)/(2n)\). Write \(Z(r)=\frac1{4t}\sum_i\chi_{j_i}(r)\) and \(X_r=(\phi_p(1+Z(r)/p)-1-Z(r))^2\), which on \(\mathcal A\) equals \((F_S(r)-1-Z_S(r))^2\).

On every outcome \(|Z(r)|\le t^2/(4t)=t/4\), so \(|\phi_p(1+Z(r)/p)|\le(1+t/(4p))^p\le e^{t/4}\) and \(|1+Z(r)|\le e^{t/4}\). Hence \(0\le X_r\le4e^{t/2}\le M:=9e^{t/2}\).

Fix \(r\ne0\). As \(j\) is uniform on \(G\), the value \(\chi_j(r)\) is a uniform sign, and these signs are independent for \(j_1,\ldots,j_k\). So Lemma~\ref{lem:scalar-remainder} gives \(\mathbb EX_r<13/4000\). From \(\mathbb EX_r=(1-\delta)\mathbb E[X_r\mid\mathcal A]+\delta\mathbb E[X_r\mid\mathcal A^c]\) we get
\[
\mathbb E[X_r\mid\mathcal A]
=\mathbb EX_r+\delta\bigl(\mathbb E[X_r\mid\mathcal A]-\mathbb E[X_r\mid\mathcal A^c]\bigr)
\le\frac{13}{4000}+\delta M.
\]
For \(r=0\) we use \(X_0\le M\). Averaging over the \(n\) rows gives, for a uniformly random \(k\)-subset \(S\),
\[
\mathbb E_S\|F_S-1-Z_S\|_H^2
\le\frac{13}{4000}+\delta M+\frac Mn
\le\frac{13}{4000}+\frac{9e^{t/2}(t^4+t^2+2)}{2\cdot2^{64t}}.
\]
For integers \(t\ge1\) we have \(e^{t/2}\le2^t\) and \(t^4\le16^{t-1}\) (equality at \(t=1,2\), and the ratio of consecutive values of \(t^4\) is at most \(16\)), and \(t^4+t^2+2\le4t^4\). So the last term is at most \(9\cdot2^t\cdot2^{4t-2}/2^{64t+1}=9\cdot2^{-59t-3}<2^{-58}\). Since \(13/4000+2^{-58}<7/2048\), we obtain \(\mathbb E_S\|F_S-1-Z_S\|_H^2<7/2048\), and Markov's inequality gives \(\mathbb P(\|F_S-1-Z_S\|_H^2>1/256)<7/8\).
\end{proof}

\begin{lemma}[Good supports are separated]\label{lem:supports-separated}
If \(S,T\) are good supports with \(|S\cap T|\le k/4\), then
\[
\left\|\frac{F_S}{\|F_S\|_H}-\frac{F_T}{\|F_T\|_H}\right\|_H>\frac{61}{420}>\frac18.
\]
\end{lemma}

\begin{proof}
By orthonormality of the characters and \(0\notin S\), \(\|Z_S\|_H^2=k/(16t^2)=1/16\), \(\langle Z_S,Z_T\rangle_H=|S\cap T|/(16t^2)\) and \(\|1+Z_S\|_H^2=17/16\). Put \(A_S=\|F_S\|_H\) and \(E_S=F_S-1-Z_S\). The good-support condition and the triangle inequality give \(|A_S-\sqrt{17}/4|\le1/16\). Since \(4<\sqrt{17}<33/8\), this yields \(15/16<A_S<35/32\).

Let \(\Pi\) be the orthogonal projection onto the complement of the constants. It does not increase \(\|\cdot\|_H\), and \(\Pi F_S=Z_S+\Pi E_S\). Hence, with \(a=A_S\), \(b=A_T\),
\[
\left\|\frac{F_S}{a}-\frac{F_T}{b}\right\|_H
\ge\left\|\frac{Z_S}{a}-\frac{Z_T}{b}\right\|_H-\frac{\|E_S\|_H}{a}-\frac{\|E_T\|_H}{b}
\ge\left\|\frac{Z_S}{a}-\frac{Z_T}{b}\right\|_H-2\cdot\frac{1/16}{15/16}.
\]
With \(\rho=|S\cap T|/k\le1/4\), the inequality \(2/(ab)\le a^{-2}+b^{-2}\) gives
\[
\left\|\frac{Z_S}{a}-\frac{Z_T}{b}\right\|_H^2
=\frac1{16}\Bigl(a^{-2}+b^{-2}-\frac{2\rho}{ab}\Bigr)
\ge\frac{1-\rho}{16}(a^{-2}+b^{-2})
\ge\frac3{32}\Bigl(\frac{32}{35}\Bigr)^2.
\]
Since \(\sqrt{3/32}>39/128\) (indeed \(39^2=1521<1536=3\cdot2^{14}/32\)), the main term exceeds \(\frac{39}{128}\cdot\frac{32}{35}=\frac{39}{140}\), and the distance exceeds \(39/140-2/15=61/420\).
\end{proof}

\begin{lemma}[Many pairwise low-intersection good supports]\label{lem:many-supports}
For every \(p\ge1\) there are at least \(2^{10t^3}\) good supports that pairwise meet in at most \(k/4\) elements.
\end{lemma}

\begin{proof}
Let \(m=n-1\) and let \(\mathcal G\) be the family of good supports, so \(|\mathcal G|\ge\frac18\binom mk\) by Lemma~\ref{lem:good-supports}. Greedily select supports from \(\mathcal G\), deleting after each selection every support meeting the selected one in more than \(k/4\) elements. With \(r=\lfloor k/4\rfloor+1\), a deleted support \(T\) satisfies \(|T\cap S|\ge r\) for the selected \(S\). Choosing \(r\) elements of \(T\cap S\) and then the other \(k-r\) elements of \(T\) outside them shows that each step deletes at most \(\binom kr\binom{m-r}{k-r}\le2^k\binom{m-r}{k-r}\) supports. So at least
\[
L\ge\frac18\,2^{-k}\binom mk\Big/\binom{m-r}{k-r}
=\frac18\,2^{-k}\prod_{i=0}^{r-1}\frac{m-i}{k-i}
\ge\frac18\,2^{-k}(m/k)^r
\]
supports are selected, where \(r\le k\) and \(m\ge k\) were used, in the form \((m-i)/(k-i)\ge m/k\). Since \(t^2\le2^{2t}\), we have \(m/k=(2^{64t}-1)/t^2\ge2^{60t}\), and \(r>k/4=t^2/4\). Hence
\[
\log_2L\ge15t^3-t^2-3\ge10t^3 ,
\]
the last step being \(5t^3\ge t^2+3\) for \(t\ge1\). The selected supports are good and pairwise meet in at most \(k/4\) elements. Probability entered only in the count of good supports; the selection itself is deterministic.
\end{proof}

\begin{proof}[Proof of Theorem~\ref{res:7.1}]
We construct the packing points from the supports of Lemma~\ref{lem:many-supports} and then deduce the four items.

\proofstep{The packing points.} Put
\[
a=\frac{4p}{4p+t},
\qquad
b=\frac{t}{4p+t},
\qquad
x_S=aq_0+\frac bk\sum_{j\in S}q_j.
\refstepcounter{equation}\tag{\theequation}\label{eq:res-7.1}
\]
This is a convex combination of columns, and \(q_j(0)=1\) for all \(j\), so \(x_S(0)=1\). Thus \(x_S\in Q\cap\partial[-1,1]^n=X\), and all these points lie in the positive row-\(0\) simplex. Since \(b/k=1/(t(4p+t))\) and \(a/(4tp)=1/(t(4p+t))\), we have \(x_S(r)=a(1+Z_S(r)/p)\) for every row \(r\). As \(\phi_p\) is homogeneous of degree \(p\) on positive scalars, \(\phi_p(x_S)=a^pF_S\), so
\[
u_p(x_S)=\frac{F_S}{\|F_S\|_2}=\frac1{\sqrt n}\cdot\frac{F_S}{\|F_S\|_H},
\qquad
\|u_p(x_S)-u_p(x_T)\|_2=\left\|\frac{F_S}{\|F_S\|_H}-\frac{F_T}{\|F_T\|_H}\right\|_H.
\]

\proofstep{Item (i).} Lemmas~\ref{lem:supports-separated} and~\ref{lem:many-supports} give \(2^{10t^3}\) points of \(u_p(X)\) at pairwise distance greater than \(61/420>1/8\). A ball of radius \(1/16\), with any center, contains at most one of them, proving (i).

\proofstep{Items (ii) and (iii).} An average of \(s\) signed columns is determined by a sequence of \(s\) of the \(2n\) signed columns, so the net \(u_p(Y_s\setminus\{0\})\) has at most \((2n)^s\) points. If it is a \(1/16\)-net, each of the \(2^{10t^3}\) packing points lies in a ball of radius \(1/16\) around a net point, and such a ball contains at most one packing point. Hence \((2n)^s\ge2^{10t^3}\), that is, \(s(64t+1)=s\log_2(2n)\ge10t^3\). Further, \(10t^3/(64t+1)\ge2t^2/13\) is equivalent to \(130t^3\ge128t^3+2t^2\), which holds for \(t\ge1\). This proves (ii).

For (iii), \(p_n=\lceil16(64t+1)\ln2\rceil\le16\cdot65t\ln2+1\le736t\), so \(2^{-22}p_n^2\le\frac{736^2}{2^{22}}t^2=\frac{529}{4096}t^2\le\frac2{13}t^2\le s\), because \(529\cdot13=6877<8192\).

\proofstep{Item (iv).} Choose for each of the \(2^{10t^3}\) packing points a point of \(Z\) within distance \(1/16\). By the triangle inequality these net points are pairwise at distance greater than \(\eta:=61/420-1/8=17/840\). In particular they are distinct. Let \(T:\mathbb R^n\to\mathbb R^K\) be distance-preserving on \(Z\). Taking \(z'=0\) in that condition gives \(\|Tz\|_2\le\sqrt{3/2}\) for \(z\in Z\), so the images of these net points lie in the ball of radius \(\sqrt{3/2}\) in \(\mathbb R^K\). They are pairwise at distance at least \(\eta/\sqrt2\). By \eqref{eq:volume-packing},
\[
2^{10t^3}\le\left(1+\frac{2\sqrt{3/2}\,\sqrt2}{\eta}\right)^K=\left(1+\frac{2\sqrt3}{\eta}\right)^K<256^K,
\]
because \(2\sqrt3\cdot840/17<172\). Therefore \(K\ge10t^3/8=5t^3/4\).
\end{proof}

The lower bound of Theorem~\ref{res:7.1} is uniform in \(p\). At a given exponent the covering entropy can be larger, and at \(p=1\) it is at least of order \(n\). Indeed, for \(x=\sum_j\theta_jq_j\) in the positive row-\(0\) simplex, \(u_1(x)=x/\|x\|_2\) has coordinates \(\theta/\|\theta\|_2\) in the orthonormal basis \(q_j/\sqrt n\), so taking \(\theta\) uniform on a set \(S\) of size \(n/2\) gives the unit vector \(\mathbf 1_S/\sqrt{n/2}\) in that basis. Two such vectors are at squared distance \(|S\triangle S'|/(n/2)\). There are at least \(2^n/(n+1)\) subsets of size \(n/2\), and at most \(2^{H(1/4)n}\) subsets lie at symmetric difference less than \(n/4\) from a given one, where \(H(1/4)<0.82\) is the binary entropy. So a greedy choice (the Gilbert--Varshamov bound) gives a family \(\mathcal S\) of subsets of size \(n/2\) with
\begin{equation}\label{eq:code-family}
 |\mathcal S|\ge\frac{2^{n/6}}{n+1},
 \qquad
 |S\triangle S'|\ge\frac n4\quad(S\ne S'\in\mathcal S).
\end{equation}
Their images are pairwise at distance at least \(1/\sqrt2>1/8\).

The lower bounds of this subsection concern the maps \(u_p\) and linear maps that are distance-preserving on a net of their image. Continuous small-fiber encodings of other forms are outside their scope, and Section~\ref{ssec:7.3} determines the least covering entropy over all continuous odd maps of the profile space.

\subsection{Other coordinate functions and spherical nets}\label{ssec:7.2}

Theorem~\ref{res:7.1} concerns the functions \(\phi_p\). We first determine what other odd functions of a single coordinate achieve on Hadamard matrices, and then return to the general step from a net to an index bound.

Let \(n=2^m\) and \(X=X_{H_{2^m}}\). For a continuous odd function \(\psi:[-1,1]\to\mathbb R\) with \(\psi(1)\ne0\), the coordinate-function map is \(u_\psi(x)=\psi(x)/\|\psi(x)\|_2\), with \(\psi\) applied coordinatewise. It is defined, continuous and odd on \(X\), because some coordinate of \(x\in X\) equals \(\pm1\). We use the function
\[
\psi_\alpha(t)
=
\left(\frac{1+t}{2}\right)^\alpha
-
\left(\frac{1-t}{2}\right)^\alpha,
\qquad
\alpha=\frac1{4096n(m+1)},
\]
which is continuous and odd with \(\psi_\alpha(\pm1)=\pm1\).

The reason for this choice is the following. If \(x\) is a convex combination of signed columns, then \((1+x_r)/2\) and \((1-x_r)/2\) are the total masses of the columns that are positive and negative at the row \(r\). For \(\alpha\) this small, the power \(\alpha\) replaces each of these totals essentially by its largest term. Thus \(u_{\psi_\alpha}(x)\) is close to a vector determined by the order of the masses and by the largest mass on each side of each row. On a Hadamard matrix such vectors lie in few subspaces of low dimension. The next proposition shows that this reduces the covering entropy to quadratic order in \(m\), and that no coordinate function has smaller order.

To state it, write a point \(x\in X\) as a convex combination of signed columns with positive masses, order these signed columns by decreasing mass as \(v_1,\ldots,v_L\), and let \(c_j\in\{0,\pm1\}^n\) be the common-sign vector of the first \(j\) of them: \(+1\) where all of them are positive, \(-1\) where all are negative, \(0\) elsewhere. The sequence of the distinct vectors among \(c_1,\ldots,c_L\) is the \emph{common-sign chain} of the representation.

\begin{resultenv}{Proposition}{res:7.2}{Hadamard coordinate-function estimates}
(i) \(\log_2\mathcal N(u_{\psi_\alpha}(X),1/16)=O(m^2)\); more precisely, the common-sign chains have length at most \(m+1\) and number at most \((2n)^{m+2}\), and \(u_{\psi_\alpha}(X)\) has a symmetric \(1/16\)-net of size at most \(2(2n)^{m+2}129^{m+1}\).

(ii) For every continuous odd \(\psi\) with \(\psi(1)\ne0\),
\[
\log_2\mathcal N(u_\psi(X),1/16)\ge\lfloor m/2\rfloor\lceil m/2\rceil.
\]
\end{resultenv}

\begin{proof}
For part (i) we round \(\psi_\alpha(x)\) to a vector in the span of a short chain of sign vectors, and then cover the spans of all chains.

\proofstep{Part (i): the rounded vector.} Fix \(x\in X\). Since \(x\in Q\), we can write \(x=\sum_v\mu_vv\) with positive masses \(\mu_v\) of total mass \(1\) on signed columns \(v\in\{\pm q_j\}\). Some coordinate \(r_0\) of \(x\) equals a sign \(\varepsilon\), because \(x\in\partial[-1,1]^n\). An average of numbers \(\pm1\) equals \(\varepsilon\) only if all of them do, so \(v(r_0)=\varepsilon\) for every \(v\) in the representation. For a row \(r\) let \(A_r\) and \(B_r\) be the largest masses among the vertices whose \(r\)-th coordinate is \(+1\), respectively \(-1\) (zero if there are none), and put \(w_r=A_r^\alpha-B_r^\alpha\). Let \(P_r^\pm\) be the total mass of the vertices with \(r\)-th coordinate \(\pm1\). Then \((1\pm x_r)/2=P_r^\pm\), so \(\psi_\alpha(x_r)=(P_r^+)^\alpha-(P_r^-)^\alpha\). Each \(P_r^\pm\) has at most \(2n\) terms, so \(A_r\le P_r^+\le2nA_r\) and \(B_r\le P_r^-\le2nB_r\). As \(A_r,B_r\le1\), both \((P_r^+)^\alpha-A_r^\alpha\) and \((P_r^-)^\alpha-B_r^\alpha\) lie in \([0,(2n)^\alpha-1]\). Their difference is \(\psi_\alpha(x_r)-w_r\), and \(\alpha\ln(2n)=\ln2/(4096n)\le1\) together with \(e^y-1\le2y\) on \([0,1]\) gives
\[
|\psi_\alpha(x_r)-w_r|\le(2n)^\alpha-1\le\frac{2\ln2}{4096n}\le\frac1{2048n}.
\]
Hence \(\|\psi_\alpha(x)-w\|_2\le\sqrt n/(2048n)\le1/2048\). Further \(\|\psi_\alpha(x)\|_2\ge|\psi_\alpha(x_{r_0})|=1\), and \(w\ne0\) because at \(r_0\) exactly one of \(A_{r_0},B_{r_0}\) is positive. By \eqref{eq:normalization}, the unit vectors \(u_{\psi_\alpha}(x)\) and \(w/\|w\|_2\) differ by at most \(1/1024\).

\proofstep{Part (i): the chain.} Order the vertices of the representation by decreasing mass as \(v_1,\ldots,v_L\), put \(a_j=\mu_{v_j}^\alpha\) and \(a_{L+1}=0\), and let \(c_j\) be the common-sign vectors defined before the proposition. Then
\[
w=\sum_{j=1}^L(a_j-a_{j+1})c_j.
\]
To check the coordinate \(r\), suppose \(v_1(r)=+1\) and let \(j_-\) be the least \(j\) with \(v_j(r)=-1\) (\(j_-=L+1\) if there is none). Then \(c_j(r)=1\) for \(j<j_-\) and \(c_j(r)=0\) for \(j\ge j_-\), so the sum at \(r\) telescopes to \(a_1-a_{j_-}\). Moreover \(a_1=A_r^\alpha\) and \(a_{j_-}=B_r^\alpha\), since \(v_1\) has the largest mass overall and \(v_{j_-}\) the largest mass among the vertices negative at \(r\). The case \(v_1(r)=-1\) follows by negating all signs at \(r\).

Let \(E_j=\{r:v_1(r)=\cdots=v_j(r)\}\), so that \(c_j=v_1\mathbf 1_{E_j}\). On a Hadamard matrix \(v_i(r)=\varepsilon_i(-1)^{r\cdot h_i}\) for a sign \(\varepsilon_i\) and a column index \(h_i\), so \(E_j\) is the solution set of the affine equations \(r\cdot(h_i+h_1)=\text{const}_i\), \(i\le j\), over \(\mathbb F_2\). It contains \(r_0\), so it is a nonempty affine subspace, and \(E_1=\mathbb F_2^m\supseteq E_2\supseteq\cdots\). Every strict change of \(c_j\) is a strict shrinking of \(E_j\) and lowers its affine dimension by at least one. Hence at most \(m+1\) distinct vectors \(c_j\) occur, all nonzero because \(r_0\in E_j\), so the common-sign chain has length at most \(m+1\). The chain is determined by the signed column \(v_1\) and the signed columns \(v_j\) at which a strict change occurs, at most \(m+1\) signed columns in total, so the number of chains is at most \(\sum_{l=1}^{m+1}(2n)^l\le(2n)^{m+2}\).

\proofstep{Part (i): the net.} Each chain spans a space of dimension at most \(m+1\), and \(w/\|w\|_2\) is a unit vector of the span of the chain of \(x\). By \eqref{eq:volume-packing} with \(\rho=1/64\), the unit sphere of that span has a \(1/64\)-net of size at most \(129^{m+1}\). Taking the union of these nets over all chains and adding negatives gives a symmetric set \(Z\) of unit vectors with \(|Z|\le2(2n)^{m+2}129^{m+1}\), and every point of \(U=u_{\psi_\alpha}(X)\) lies within \(1/64+1/1024=17/1024\) of \(Z\).

To obtain centers in \(U\), note that \(U=-U\), since \(X=-X\) and \(u_{\psi_\alpha}\) is odd. A point \(z\in Z\) satisfies \(z\ne-z\). For each pair \(\{z,-z\}\subseteq Z\) such that the closed ball \(\overline B(z,17/1024)\) meets \(U\), choose \(y_z\) in the intersection and put \(y_{-z}=-y_z\), which lies in \(\overline B(-z,17/1024)\cap U\). The points \(y_z\) form a symmetric subset of \(U\) with at most \(|Z|\) points. Every \(u\in U\) lies in some ball around \(z\in Z\), and then \(\|u-y_z\|_2\le34/1024<1/16\). This proves (i): \(\log_2\mathcal N(U,1/16)\le1+(m+2)(m+1)+(m+1)\log_2129=O(m^2)\). The representations and their orderings enter only the proof of the covering, and no continuity in \(x\) is required of them; the map itself is the coordinatewise \(\psi_\alpha\), whose continuity and oddness were noted above.

\proofstep{Part (ii).} For a linear subspace \(W\le\mathbb F_2^m\) of dimension \(r=\lfloor m/2\rfloor\), the uniform average of the columns indexed by \(W^\perp\) has coordinate \(|W^\perp|^{-1}\sum_{j\in W^\perp}(-1)^{s\cdot j}=\mathbf 1_{(W^\perp)^\perp}(s)\) at the row \(s\). So it is \(x_W=\mathbf 1_W\), the indicator of \(W\) among the rows, and \(x_W\in X\) because \(x_W(0)=1\). Any \(\psi\) as in the statement has \(\psi(0)=0\) by oddness and maps \(x_W\) to \(\psi(1)\mathbf 1_W\), which normalizes to \(\operatorname{sgn}(\psi(1))\mathbf 1_W/\sqrt{|W|}\). Two distinct \(r\)-dimensional subspaces intersect in dimension at most \(r-1\), so these unit vectors have squared distance \(2-2|W\cap W'|/|W|\ge1\). A closed ball of radius \(1/16\) contains at most one of them. The number of such subspaces is \(\prod_{i<r}(2^m-2^i)/(2^r-2^i)\ge2^{r(m-r)}\), each factor being at least \(2^{m-r}\) since \(2^{m-r}(2^r-2^i)=2^m-2^{m-r+i}\le2^m-2^i\).
\end{proof}

The bound on the number of chains in part (i) uses the affine structure of the sets \(E_j\), and it has no analogue for general matrices. For the single-flip matrix \(B_{ij}=1-2\mathbf 1_{\{i=j\}}\) of even order \(n=2^m\ge4\) the same function \(\psi_\alpha\) gives exponentially many separated points, on a class of bounded index. The uniform average \(x_S\) of the columns in a set \(S\) of size \(n/2\) has coordinates \(1\) off \(S\) and \(\tau=1-4/n\) on \(S\), and \(x_S\in X_B\) since \(S^c\ne\varnothing\). As \(\tau\ge0\), we get \(0\le\psi_\alpha(\tau)\le1-(2/n)^\alpha\le\alpha\ln(n/2)\le1/(4096n)\), so \(\|\psi_\alpha(x_S)-\mathbf 1_{S^c}\|_2\le\sqrt{n/2}/(4096n)\), and \eqref{eq:normalization} shows that \(u_{\psi_\alpha}(x_S)\) lies within \(1/(2048n)\) of \(\mathbf 1_{S^c}/\sqrt{n/2}\). For a family as in \eqref{eq:code-family}, these images are pairwise at distance at least \(1/\sqrt2-1/(1024n)>1/8\), which gives \(e^{\Omega(n)}\) pairwise separated images.

This class has index bounded by an absolute constant: for the background \(g\equiv+1\) the error sets are disjoint singletons, so Theorem~\ref{res:5.2} with \(s=0\) gives \(\operatorname{lr}\le2\), and Theorem~\ref{res:a} gives \(\operatorname{Ind}\le3\). Thus the function \(\psi_\alpha\), which has smaller covering entropy than every power map on large Hadamard matrices, cannot replace the power maps on all matrices, since Lemma~\ref{res:3.0} provides a net of at most \((2M)^{s_N}\) points for \(u_{p_N}\) on every matrix.

We turn to the step from a net to an index bound. The proof of Theorem~\ref{res:3.1} uses a Gaussian map that preserves distances up to a small constant factor, because the learning bound of Theorem~\ref{res:3.8} needs the small-fiber estimate \eqref{eq:res-3.1}. For an index bound alone it suffices that the map be Lipschitz and keep the net away from the origin. Allowing a larger distortion then gives the following constant, which is used for the comparison between entropy and index in Theorem~\ref{res:7.4}.

\begin{resultenv}{Theorem}{res:7.3}{An index bound from a finite spherical net}
If a symmetric set \(U\subseteq S^{D-1}\) is \(1/16\)-close to a symmetric finite set \(Z\subset S^{D-1}\) of size \(L\) (every point of \(U\) lies within Euclidean distance \(1/16\) of \(Z\)), then
\[
\operatorname{Ind}(U)+1
\le
\left\lceil
\frac54\ln\bigl(4(L+1)^2\bigr)
\right\rceil.
\]
\end{resultenv}

\begin{proof}
If \(U\) is empty there is nothing to prove, so let \(U\ne\varnothing\) and put
\[
k=\left\lceil\frac54\ln\bigl(4(L+1)^2\bigr)\right\rceil.
\]

\proofstep{A Gaussian map with large distortion.} Let \(T\) be a \(k\times D\) Gaussian matrix with independent \(N(0,1/k)\) entries, and apply the tail bounds of Lemma~\ref{res:3.0prime} to a nonzero difference \(y\) of two points of \(Z\cup\{0\}\): the upper-tail bound with \(\eta=3\) and the lower-tail bound with \(\eta=15/16\). Since \(\|Ty\|_2^2/\|y\|_2^2\) has the law of \(\chi_k^2/k\), they give
\[
\mathbb P(\chi_k^2/k\ge4)\le e^{-k(3-\ln4)/2}\le e^{-4k/5},
\qquad
\mathbb P(\chi_k^2/k\le1/16)\le e^{-k(\ln16-15/16)/2}\le e^{-4k/5},
\]
because \((3-\ln4)/2>0.8\) and \((\ln16-15/16)/2>0.9\). There are at most \((L+1)^2\) ordered pairs in \(Z\cup\{0\}\), and \(e^{-4k/5}\le1/(4(L+1)^2)\) by the choice of \(k\), so by the union bound some difference violates one of these two bounds with probability at most \(2(L+1)^2e^{-4k/5}\le1/2\). Fix a matrix \(T\) for which all differences satisfy
\[
\tfrac1{16}\|z-z'\|_2^2\le\|Tz-Tz'\|_2^2\le4\|z-z'\|_2^2
\qquad(z,z'\in Z\cup\{0\}).
\]
Thus \(z\mapsto Tz\) is \(2\)-Lipschitz on \(Z\), and taking \(z'=0\) gives \(\|Tz\|_2\ge1/4\) for the unit vectors \(z\in Z\).

\proofstep{Extension and odd symmetrization.} Apply the Lipschitz extension of the proof of Theorem~\ref{res:3.1} to the finite data \(z\mapsto Tz\) on \(Z\), with Lipschitz constant \(L_0=2\). It extends the data to a \(2\)-Lipschitz map \(\widetilde G:\mathbb R^D\to\mathbb R^k\). Put \(G(w)=(\widetilde G(w)-\widetilde G(-w))/2\). Then \(G\) is odd and \(2\)-Lipschitz, and \(G(z)=Tz\) on \(Z\), since \(Z=-Z\) and \(T\) is linear. For \(u\in U\) choose \(z\in Z\) with \(\|u-z\|_2\le1/16\). Then
\[
\|G(u)\|\ge\|Tz\|-2\|u-z\|\ge1/4-2/16=1/8.
\]
Hence \(G/\|G\|\) is a continuous odd map \(U\to S^{k-1}\), and \(\operatorname{Ind}(U)\le k-1\).
\end{proof}

The logarithmic dependence on \(L\) is optimal among bounds that depend only on the cardinality of the net. Let \(g(L)\) denote the largest index of a symmetric spherical set admitting a symmetric \(1/16\)-net of cardinality at most \(L\). Then, for \(L\ge66\),
\[
g(L)\ge
\left\lfloor\frac{\ln(L/2)}{\ln33}\right\rfloor-1.
\]
Indeed, \(S^{d-1}\), of index \(d-1\), has a symmetric \(1/16\)-net of size at most \(2\cdot33^d\): by \eqref{eq:volume-packing}, a maximal subset \(A\) with pairwise distances greater than \(1/16\) has \(|A|\le33^d\), and \(A\cup-A\) is symmetric. Moreover \(2\cdot33^d\le L\) for \(d=\lfloor\ln(L/2)/\ln33\rfloor\), and \(d\ge1\) since \(L\ge66\). So no index bound depending only on the cardinality of the net can lie below this order.

\subsection{Reduction to the profile space}\label{ssec:7.3}

The encodings of Sections~\ref{ssec:7.1} and~\ref{ssec:7.2} factor through the profile map, which replaces a distribution on labeled points by the losses of the rows. One may ask whether this passage can lower the index. The next theorem shows that it cannot: the profile map has a continuous odd section and is an equivariant homotopy equivalence, so \(S(B)\) and \(X_B\) have the same index. It also shows that the large covering entropy in Theorem~\ref{res:7.1} is a property of the power maps and not of \(X_B\). Over all continuous odd maps from \(X_B\) to a sphere, the least covering entropy at scale \(1/16\) is comparable to the index.

Throughout this subsection \(\Phi\) is the profile map of Section~\ref{ssec:2.5}, on \(|S(B)|=\Delta_{C_B^\pm}\), and \(\operatorname{Ind}\) abbreviates \(\operatorname{Ind}S(B)\). Write \(\iota\) for the involution of \(|S(B)|\) that flips all labels; then \(\Phi(\iota\nu)=-\Phi(\nu)\).

\begin{resultenv}{Theorem}{res:7.4}{A section of the profile map and the entropy of odd images}
The map \(\Phi:|S(B)|\longrightarrow X_B\) has a continuous odd section \(j\), with \(\Phi\circ j=\operatorname{id}\), and \(|S(B)|\) has an equivariant strong deformation retraction onto \(j(X_B)\).

If
\[
\mathcal E(B)
:=
\inf_f\log_2\mathcal N(f(X_B),1/16),
\]
where \(f\) ranges over continuous odd maps from $X_B$ to a unit sphere and \(\mathcal N(A,1/16)\) denotes the least number of points of \(A\) whose closed \(1/16\)-balls cover \(A\) (centers in the covered set, hence unit vectors), then
\[
\frac{\operatorname{Ind}+1}{8}
\le
\mathcal E(B)
\le
7(\operatorname{Ind}+1).
\]
\end{resultenv}

\begin{proof}
The section and the retraction rest on the fact that the fibers of \(\Phi\) are convex.

\proofstep{The fibers.} For a row \(r\) and a sign \(\varepsilon\), let \(\sigma_{r,\varepsilon}=\{i^{\varepsilon B_{ri}}:i\in[M]\}\) be the signed row simplex and \(X_{r,\varepsilon}=Q\cap\{v_r=\varepsilon\}\). Let \(v\in X_B\), say \(v_r=\varepsilon\), and let \(\nu\) be a distribution on atoms with \(\Phi(\nu)=v\). Then \(\varepsilon=\Phi(\nu)_r=\sum_{(i,\tau)}\nu_{i,\tau}\tau B_{ri}\) is an average of numbers \(\pm1\), so every positive-weight atom satisfies
\[
\tau=\varepsilon B_{ri}.
\]
Thus \(\nu\in|\sigma_{r,\varepsilon}|\subseteq|S(B)|\). Conversely, a point of \(|\sigma_{r,\varepsilon}|\) has profile in \(X_{r,\varepsilon}\). Since \(Q\) is the image of the distributions on atoms under \(\Phi\), this shows that \(\Phi\) maps \(|S(B)|\) onto \(X_B\), and that the fiber over \(v\) is the intersection of the simplex \(|\sigma_{r,\varepsilon}|\) with an affine subspace. So the fiber is convex and lies in a single signed row simplex.

\proofstep{The section.} As \(Q\subseteq[-1,1]^N\), each nonempty \(X_{r,\varepsilon}\) is a face of the polytope \(Q\), and \(X_B\) is their union. The set of faces of \(Q\) contained in some \(X_{r,\varepsilon}\) is a polyhedral complex with support \(X_B\), and the map \(v\mapsto-v\) permutes it, with \(-X_{r,\varepsilon}=X_{r,-\varepsilon}\). Its barycentric subdivision \(\mathcal T\) is a simplicial complex on which \(v\mapsto-v\) acts simplicially, and every simplex of \(\mathcal T\) lies in some \(X_{r,\varepsilon}\). No vertex of \(\mathcal T\) is fixed, since \(0\notin X_B\). For each antipodal pair \(\{a,-a\}\) of vertices of \(\mathcal T\) choose \(\nu_a\in\Phi^{-1}(a)\) and put \(\nu_{-a}=\iota\nu_a\). For \(x\) in a simplex of \(\mathcal T\) with vertices \(a_0,\ldots,a_l\) and barycentric coordinates \(\lambda_i\), put \(j(x)=\sum_i\lambda_i\nu_{a_i}\). If this simplex lies in \(X_{r,\varepsilon}\), then each \(\nu_{a_i}\) lies in \(|\sigma_{r,\varepsilon}|\) by the previous step, so \(j(x)\in|S(B)|\). The definition is compatible on common faces, so \(j\) is continuous. It satisfies \(j(-x)=\iota j(x)\), and \(\Phi(j(x))=\sum_i\lambda_ia_i=x\) because \(\Phi\) is affine on distributions.

\proofstep{The retraction.} For \(\nu\in|S(B)|\) both \(\nu\) and \(j(\Phi(\nu))\) lie in the convex fiber of \(\Phi(\nu)\), so the straight-line homotopy
\[
(1-t)\nu+t\,j(\Phi(\nu))
\]
stays in that fiber, hence in \(|S(B)|\). It is continuous, it commutes with \(\iota\) because \(\iota\) is affine and \(j,\Phi\) are odd, and it fixes \(j(X_B)\) pointwise because \(j\Phi j=j\). It is the required strong deformation retraction.

\proofstep{The entropy lower bound.} Let \(f\) be a continuous odd map from \(X_B\) to a unit sphere, put \(U=f(X_B)\), and let \(L=\mathcal N(U,1/16)\). The set \(U\) is symmetric and nonempty. A single ball of radius \(1/16\) cannot cover it, since \(u,-u\) are at distance \(2\), so \(L\ge2\). Let \(Z_0\subseteq U\) be a covering set with \(|Z_0|=L\), and put \(Z=Z_0\cup-Z_0\), a symmetric subset of \(U\) with \(|Z|\le2L\), to which \(U\) is \(1/16\)-close. Theorem~\ref{res:7.3} gives \(\operatorname{Ind}(U)+1\le\lceil\frac54\ln(4(2L+1)^2)\rceil\), and the composition of \(f\circ\Phi\) with an odd map \(U\to S^{\operatorname{Ind}(U)}\) is odd on \(|S(B)|\), so \(\operatorname{Ind}\le\operatorname{Ind}(U)\). As \(L\ge2\), we have \(2L+1\le\frac52L\), so
\[
\operatorname{Ind}+1
\le
\left\lceil
\frac54\ln\bigl(4(2L+1)^2\bigr)
\right\rceil
\le\frac52\ln L+\frac54\ln25+1
\le\frac52\ln L+5.03
\le8\log_2L,
\]
where \(\frac54\ln25+1<5.03\), and the last step uses \(5.03<8-\frac52\ln2=(\frac8{\ln2}-\frac52)\ln2\le(\frac8{\ln2}-\frac52)\ln L\). Taking the infimum over \(f\) gives \(\mathcal E(B)\ge(\operatorname{Ind}+1)/8\).

\proofstep{The entropy upper bound.} Put \(d=\operatorname{Ind}+1\), so \(d\ge1\) since \(S(B)\) is nonempty, and let \(f\) be the composition of the odd section \(j\) with an odd map \(|S(B)|\to S^{d-1}\), which exists by the definition of the index. By \eqref{eq:volume-packing} with \(R=1\), \(\rho=1/16\), a maximal subset of \(f(X_B)\subseteq S^{d-1}\) with pairwise distances greater than \(1/16\) has at most \(33^d\) points, and the closed \(1/16\)-balls around them cover \(f(X_B)\). Its centers lie in \(f(X_B)\), so
\[
\mathcal N(f(X_B),1/16)\le2\cdot33^d.
\]
Since \(\log_2(2\cdot33^d)=1+d\log_2 33\le7d=7(\operatorname{Ind}+1)\), because \(\log_233<5.05\) and \(d\ge1\), this gives the stated upper bound.
\end{proof}

Combining the upper bound with Theorem~\ref{thm:main} and the inequality \(\operatorname{Ind}\le2\operatorname{lr}-1\) gives, for every sign matrix \(B\),
\[
 \mathcal E(B)\le7(\operatorname{Ind}S(B)+1)
 \le14\operatorname{lr}(B)
 \le11270\ln(2N)\ln(2M).
\]
Indeed, Theorem~\ref{res:a} gives \(\operatorname{Ind}S(B)+1\le2\operatorname{lr}(C_B,\varepsilon)\le2\operatorname{lr}(B)\), Theorem~\ref{thm:main} applies to the class of \(m\le N\) distinct rows on \(M\) points, and \(14\cdot805=11270\). So every sign matrix has an odd encoding of the profile space with covering entropy at most of the order of Theorem~\ref{thm:main}, whereas by Theorem~\ref{res:7.1} the encodings \(u_p\) do not have it on large Hadamard matrices.

The section preserves profiles and equivariant homotopy type, but not the losses of arbitrary improper hypotheses: the loss of a concept is a function of the profile, by \(\Phi(\nu)_r=1-2\operatorname{loss}_\nu(c_r)\), whereas the loss of an improper hypothesis need not be. So the reduction concerns the index and the covering entropy of odd images, and not accurate improper labels.

\subsection{Power maps on closure strata}\label{ssec:7.4}

A signed row \((a,\varepsilon)\) is \emph{compatible} with a point \(x\in X_B\) if \(x_a=\varepsilon\). If \(x\) is the profile of a distribution with support \(F\), then \((a,\varepsilon)\) is compatible with \(x\) exactly when \(F\subseteq\sigma_a^\varepsilon\), so the compatible signed rows of \(x\) form the set \(R(F)\) of Section~\ref{ssec:6.1}. The sets of points with a given compatible signed-row set stratify \(X_B\), and these \emph{closure strata} correspond to the signed closed faces of Section~\ref{ssec:6.1}.

The closure chains of Section~\ref{ssec:6} treat each closed face separately, which suggests treating the power map in the same way. If the packing of Theorem~\ref{res:7.1} came from mixing different strata, one could use a separate exponent or a separate orthogonal transformation on each stratum. The next theorem shows that this is not the case. A small perturbation moves the whole packing into a single stratum, the points whose only compatible signed row is \((0,+)\).

\begin{resultenv}{Theorem}{res:7.5}{Power maps on a single closure stratum}
For the Hadamard matrix of order \(n=2^{64t}\) and every \(p\ge1\), the relative interior of the positive row-\(0\) simplex contains at least \(2^{10t^3}\) points whose \(u_p\)-images are pairwise at distance greater than \(3799/26880\), which exceeds \(1/8\). Their exact compatible signed-row set is \(\{(0,+)\}\), of affine dimension \(0\). Let \(Z\subset S^{n-1}\) be a finite set of unit vectors such that each of these image points lies within distance \(1/16\) of \(Z\). A linear map \(T:\mathbb R^n\to\mathbb R^k\) that is distance-preserving on \(Z\) in the sense of Theorem~\ref{res:7.1}(iv) requires \(k>t^3\).
\end{resultenv}

\begin{proof}
Take the packing set of Theorem~\ref{res:7.1}: at least \(2^{10t^3}\) points \(x=x_S\) in the positive row-\(0\) simplex with pairwise \(u_p\)-distance greater than \(61/420\).

\proofstep{The perturbation.} Since \(\frac1n\sum_j(-1)^{r\cdot j}=\mathbf 1_{\{r=0\}}\), the average of all columns is the row-\(0\) indicator \(e_0=\frac1n\sum_jq_j\). Put
\[
\delta=\frac1{1024\lceil p\rceil n},
\qquad
\widetilde x=(1-\delta)x+\delta e_0.
\]
This adds positive uniform noise to the column coefficients, so \(\widetilde x\) lies in the relative interior of the positive row-\(0\) simplex. Further \(\widetilde x_0=1\), and for \(a\ne0\) we have \(e_0(a)=0\), so \(|\widetilde x_a|=(1-\delta)|x_a|\le1-\delta<1\). A signed row \((a,\varepsilon)\) is compatible with \(\widetilde x\) exactly when \(\widetilde x_a=\varepsilon\), so the exact compatible signed-row set is \(\{(0,+)\}\). On a Hadamard matrix the compatible signed rows of a point form an affine subspace over \(\mathbb F_2\), since each atom of a representing distribution gives one affine equation, and here this subspace is a single point, of affine dimension \(0\).

On \([-1,1]\) the function \(\phi_p\) is \(p\)-Lipschitz, both power vectors have norm at least \(1\) because their row-\(0\) coordinate is \(1\), and \(x-e_0\) has row-\(0\) coordinate \(0\) and all other coordinates in \([-1,1]\), so \(\|\widetilde x-x\|_2=\delta\|x-e_0\|_2\le\delta\sqrt n\). By \eqref{eq:normalization},
\[
\|u_p(\widetilde x)-u_p(x)\|_2\le2p\|\widetilde x-x\|_2\le2p\delta\sqrt n\le\frac1{512}.
\]
Distinct perturbed points are therefore at distance greater than \(61/420-1/256=3799/26880>1/8\), and the cardinality is preserved.

\proofstep{Distance-preserving maps.} Let \(Z\) be as in the statement. Let \(T:\mathbb R^n\to\mathbb R^k\) be distance-preserving on \(Z\). Choose for each perturbed point a point of \(Z\) within distance \(1/16\). These points of \(Z\) are distinct and pairwise at distance greater than \(\eta':=3799/26880-1/8=439/26880\). The map \(T\) sends them into the ball of radius \(\sqrt{3/2}\) in \(\mathbb R^k\) (take \(z'=0\)) while keeping their pairwise distances at least \(\eta'/\sqrt2\). By \eqref{eq:volume-packing}, \(2^{10t^3}\le(1+2\sqrt3/\eta')^k<214^k\), because \(2\sqrt3\cdot26880/439<213\). So \(k\ge10t^3/\log_2214>1.29\,t^3>t^3\), as \(\log_2214<7.75\).
\end{proof}

Every set \(Z\) as in Theorem~\ref{res:7.1}(iv) satisfies this hypothesis, because the perturbed points lie in \(X_{H_n}\), and the two lower bounds on the dimension have the same order \(t^3\). The hypothesis here is weaker: \(Z\) has to approximate only the perturbed points, which all lie in one stratum.

The theorem rules out two refinements of the power map. An orthogonal transformation applied after the power map does not remove the packing, since it preserves the distances between the \(u_p\)-images, and neither does a separate choice of exponent on each stratum, since the packing holds for every \(p\). A rotation applied before taking coordinate powers, and exponents and rotations depending continuously on the point within a stratum, are outside its scope.

\subsection{Affine statistics and average advantage}\label{ssec:7.5}

This subsection tests two simplifications of the constructions of Section~\ref{ssec:3}. The first replaces the nonlinear power map by a statistic that is affine in the distribution. The second replaces the atomwise condition used for transport by an average condition on the vote. Both fall short, and the first has an exact answer in terms of the sign rank.

An affine statistic is the simplest compression of a distribution, the kind computed by a linear sketch of the empirical measure. We measure it against the weakest reasonable task, namely a bounded vote with some positive advantage on every realizable distribution, and we allow an arbitrary decoder. The least dimension for this task on a symmetric class is exactly the sign rank, which can exceed the list number exponentially.

Formally, a \emph{weak learner in the population soft-vote model} consists of an affine distribution statistic
\[
S(\nu)=b+\sum_z\nu_z t_z\in\mathbb R^k
\]
followed by an arbitrary, not necessarily continuous, decoder \(W:S(\Delta_C)\to[-1,1]^M\) with positive advantage \(\mathbb E_{(i,y)\sim\nu}[yW(S(\nu))(i)]\ge\gamma>0\) for every realizable \(\nu\). This is a bounded real-valued vote on population distributions, not a binary-hypothesis learning rule in the sense of Section~\ref{sec:model}. We call \(k\) its affine dimension.

\begin{resultenv}{Theorem}{res:7.6}{Affine statistics require sign-rank dimension}
On a symmetric class \(C=-C\) with matrix \(B\), every weak learner in the population soft-vote model requires \(k\ge\operatorname{signrk}(B)\). The dimension \(k=\operatorname{signrk}(B)\) is attainable, with advantage depending on \(B\).

For the square family \(Q_m\) obtained by repeating twice the columns of the symmetric Hadamard class, with \(N=M=2^{m+1}\), the affine dimension is at least \(2^{m/2}\), while \(\operatorname{lr}(Q_m,\varepsilon)=m+1\) and \(\operatorname{Ind}=m\) for \(0<\varepsilon<1/2\).
\end{resultenv}

\begin{proof}
The idea of the lower bound is that if a statistic does not separate a distribution realized by \(c\) from its mirror image realized by \(-c\), then no decoder can have positive advantage on both.

\proofstep{The lower bound.} Write \(S(\nu)=b+\sum_{i,y}\nu_{i,y}t_{i,y}\) and \(d_i=(t_{i,+}-t_{i,-})/2\), so that \(t_{i,c(i)}-t_{i,-c(i)}=2c(i)d_i\) for a concept \(c\). Fix a concept \(c\). Suppose \(0\in\operatorname{conv}\{c(i)d_i:i\in[M]\}\), so there is a probability vector \(p\) with \(\sum_ip_ic(i)d_i=0\). Define \(\nu\) by \(\nu_{i,c(i)}=p_i\) and \(\bar\nu\) by \(\bar\nu_{i,-c(i)}=p_i\), with all other masses zero. They are realized by \(c\) and by \(-c\), and \(-c\in C\) by the symmetry \(C=-C\). Moreover
\[
S(\nu)-S(\bar\nu)=\sum_ip_i\bigl(t_{i,c(i)}-t_{i,-c(i)}\bigr)=2\sum_ip_ic(i)d_i=0.
\]
So the same vote \(w=W(S(\nu))\) is decoded on both distributions. Its advantages are \(\sum_ip_ic(i)w(i)\) on \(\nu\) and the negative of this number on \(\bar\nu\), so it cannot be positive on both. Hence each of these convex hulls avoids zero. Let \(a_c\ne0\) be the point of the hull nearest to zero. The nearest-point inequality for a convex set gives \(\langle a_c,c(i)d_i\rangle\ge\|a_c\|_2^2>0\) for all \(i\). Then \(R_{c,i}=\langle a_c,d_i\rangle\) is a strict sign representation of \(B\) by vectors of \(\mathbb R^k\), and \(k\ge\operatorname{signrk}(B)\). Continuity of the decoder was not used.

\proofstep{Attainability.} Conversely, given a strict sign representation \(\operatorname{sign}\langle a_c,d_i\rangle=c(i)\) of dimension \(r\), put \(S(\nu)=\sum_{i,y}\nu_{i,y}yd_i\). If \(\nu\) is realized by \(c\), then \(\langle a_c,S(\nu)\rangle=\sum_i\nu_{i,c(i)}c(i)\langle a_c,d_i\rangle>0\), so no concept simplex is mapped to zero. The realizable space is compact, so \(\rho:=\min_{\nu\in\Delta_C}\|S(\nu)\|_2>0\). Let \(R=\max_i\|d_i\|_2\), so \(\|S(\nu)\|_2\le R\), and let \(W(q)(i)=\langle q,d_i\rangle/R^2\), whose coordinates lie in \([-1,1]\) on the image. Then
\[
\mathbb E_\nu[yW(S(\nu))(i)]
=\frac1{R^2}\Bigl\langle S(\nu),\sum_{i,y}\nu_{i,y}yd_i\Bigr\rangle
=\frac{\|S(\nu)\|_2^2}{R^2}\ge\frac{\rho^2}{R^2}>0.
\]
Coordinatewise truncation extends \(W\) continuously to a cube containing the image.

\proofstep{The family \(Q_m\).} Let \(H=H_{2^m}\) and let \(H^\pm\) be its symmetric class, with \(2^{m+1}\) rows and \(2^m\) columns. Thus \(Q_m\) has \(N=M=2^{m+1}\) and is symmetric, so the first part applies. By Forster's bound \(\operatorname{signrk}(A)\ge\sqrt{N_AM_A}/\|A\|\) for an \(N_A\times M_A\) sign matrix \(A\) (\cite{Forster02}), and \(\|H\|=2^{m/2}\), the sign rank of the Hadamard block is at least \(2^m/2^{m/2}=2^{m/2}\). A submatrix has no larger sign rank, so the affine dimension is at least \(2^{m/2}\).

Column repetition preserves list replicability at each \(\varepsilon\). Indeed, a rule for \(H^\pm\) gives one for \(Q_m\): map each column of a sample to the column of \(H^\pm\) it repeats and compose the output with this map, which preserves losses and lists. Conversely, a rule for \(Q_m\) gives one for \(H^\pm\), by placing the samples on the first copy of the columns and restricting the output to that copy. Column repetition also preserves the index: collapsing the two copies and including the first copy are equivariant simplicial maps between \(S(Q_m)\) and \(S(H^\pm)\).

For the symmetric class, \(S(H^\pm)=S(H)\), because a sign complex contains both \(\sigma_r^+\) and \(\sigma_r^-\) for every row. So \(\operatorname{Ind}=m\) by Theorem~\ref{res:5.5}. Fix \(\varepsilon\in(0,1/2)\). We have \(\operatorname{lr}(H^\pm,\varepsilon)\ge\operatorname{lr}(H,\varepsilon)=m+1\), because a learner for the larger class serves the subclass. For the upper bound, a signed row is a pair \((s,r)\in\mathbb F_2\times\mathbb F_2^m\) with value \((-1)^{s+r\cdot j}\) at the column \(j\). An atom \((j,(-1)^e)\) gives the equation \(s+r\cdot j=e\), whose linear part \((1,j)\) is nonzero. So the version space of a nonempty positive closure is a nonempty proper affine subspace of \(\mathbb F_2^{m+1}\), of dimension at most \(m\). Along a strict chain of closures the version spaces strictly shrink, hence their dimensions strictly decrease. Thus \(h_+^*(H^\pm)\le m+1\), and Theorem~\ref{res:6.3} gives \(\operatorname{lr}(H^\pm,\varepsilon)\le\chi^*(H^\pm)\le h_+^*(H^\pm)\le m+1\).
\end{proof}

The Hadamard sign-rank lower bound used here is due to~\cite{Forster02}. The sign rank also bounds the index: the inequality
\[
\operatorname{Ind}S(B)\le\operatorname{signrk}(B)-1
\]
used elsewhere in this paper is the argument of~\cite[Corollary 2]{FHV26}: a strict signed realization gives a nonzero affine map on every face.

Theorem~\ref{res:7.6} concerns statistics that are affine in the distribution. The constructions of Section~\ref{ssec:3} compute the nonlinear weights \(u\circ\Phi\) before compressing, and such statistics lie outside its scope.

We now isolate what the transport step of Section~\ref{ssec:3} uses about a hypothesis. Let \(\nu\) be a realizable distribution, put \(D=Z(\Phi(\nu))\), write \(g_z=yg_\nu(i)\) at an atom \(z=(i,y)\), and let \(a\le\mathbb E_\nu[yg_\nu]\). By the atom-gradient identity in the proof of Lemma~\ref{res:3.6}, the tilt \(\Theta_\lambda\) multiplies the mass of \(z\) by \(e^{Dg_z/\lambda}\), up to normalization. In the transfer estimate of that lemma, the numerator of the tilted loss of a hypothesis is \(\sum_{\mathrm{err}}\nu_ze^{Dg_z/\lambda}\), and Jensen's inequality bounds the denominator below by \(e^{Da/\lambda}\). So the estimate bounds the tilted loss by
\[
\sum_{\mathrm{err}}
\nu_z\exp\!\left(\frac{D(g_z-a)}{\lambda}\right).
\]
This quantity tends to zero as \(\lambda\downarrow0\) exactly when \(g_z<a\) at every erroneous atom of positive mass, however small the mass of that atom. Thus the estimate needs a bound on the vote at each individual error, and not merely on average. With the values \(a=31/32\) of \eqref{eq:res-3.3} and \(b=45/64\) of the proof of Theorem~\ref{res:3.8}, this leads to the following definition.

\begin{definition}[Low-vote condition]\label{def:low-vote}
A hypothesis \(h\) satisfies the \emph{low-vote condition} at a realizable distribution \(\nu\) if every atom \((i,y)\) with \(\nu_{i,y}>0\) and \(h(i)\ne y\) has \(yg_\nu(i)\le45/64\).
\end{definition}

In words, \(h\) errs only on atoms where the vote is not high. Under this condition Lemma~\ref{res:3.6} bounds the tilted loss of \(h\), as the proof of Lemma~\ref{res:7.11} below shows in detail.

The average properties of a hypothesis, such as a large advantage and closeness to the vote in mean square, are what a stability argument for the vote would provide, and it is natural to ask whether they imply this atomwise condition. The next theorem shows that they do not. On the single-flip class, a single atom of small mass can carry the largest vote. A hypothesis that errs only at this atom has advantage close to one and is close to the vote in mean square, but the tilt concentrates on its error. This hypothesis is moreover the output at the given distribution of a continuous decoder whose advantage is bounded below on all realizable distributions.

\begin{resultenv}{Theorem}{res:7.7}{Average advantage does not imply the low-vote condition}
Let \(n\ge16\), let the domain be \(\{0,\ldots,n-1\}\), and let the concepts be \(c_0\equiv+1\) and, for \(0\le j<n\), \(c_{j+1}\) equal to \(c_0\) flipped at \(j\); this is the single-flip class of size \((n+1)\times n\). Let \(\nu\) be the distribution with all labels positive (its true concept is \(c_0\)),
\[
\delta:=\nu(0)=2/n,
\qquad
\nu(i)=\frac{n-2}{n(n-1)}\quad(i>0),
\]
and put \(p=p_{n+1}\), \(A=(1-2\delta)^{2p}\), \(B=(1-2\nu(1))^{2p}\) and \(\Delta=2(B-A)>0\). The hypothesis \(h=c_1\), which flips \(c_0\) at the point \(0\), satisfies
\[
\mathbb E_\nu[yh]=1-4/n\longrightarrow1,
\]
\[
\mathbb E_\nu(h-g_\nu)^2
\le
8/n+4/(n-1)^2
\longrightarrow0,
\]
but
\[
\operatorname{loss}_{\Theta_\lambda(\nu)}(h)
=
\frac{\delta e^{\Delta/\lambda}}
{1-\delta+\delta e^{\Delta/\lambda}}
\longrightarrow1
\qquad(\lambda\downarrow0).
\]
Moreover, \(h\) is the value at \(\nu\) of a continuous decoder with uniformly positive advantage: writing \(C\) for this class, there are continuous maps \(F:\Delta_C\to K\), where \(K=\partial B_1^{n+1}\) is the unit sphere of \(\ell_1^{n+1}\), and \(W:K\to[-1,1]^n\) with \(W(F(\nu))=h\) and \(\mathbb E_\tau[yW(F(\tau))]\ge23/32\) for every realizable \(\tau\).
\end{resultenv}

\begin{proof}
We compute the vote of \(\nu\) explicitly, and then show that \(h\) is the output at \(\nu\) of a continuous decoder with uniformly positive advantage.

\proofstep{The profile and the vote.} The masses sum to one, since \(2/n+(n-1)\cdot\frac{n-2}{n(n-1)}=1\). All labels of \(\nu\) are positive, so a row has profile coordinate \(1-2\nu(\text{points where it is negative})\): this gives \(\Phi_0=1\), \(\Phi_1=1-2\delta\) and \(\Phi_{i+1}=1-2\nu(i)\) for \(i>0\). As \(\delta\le1/8\), all these coordinates are positive, so the potential weights are \(1,A,B,\ldots,B\), with \(n-1\) copies of \(B\), and \(Z=D:=1+A+(n-1)B\). The column \(0\) is negative only in the row \(c_1\), and a column \(i>0\) is negative only in the row \(c_{i+1}\). Hence
\[
g_\nu(0)=1-\frac{2A}D,
\qquad
g_\nu(i)=1-\frac{2B}D\quad(i>0).
\]
Since \(\delta>\nu(1)\) (equivalently \(2(n-1)>n-2\)) we have \(A<B\), so \(g_\nu(0)>g_\nu(i)\) for \(i>0\): the atom at \(0\) carries the largest vote. Moreover \(D>nA\), because \(A<1\) and \(A<B\), and this gives \(g_\nu(0)>1-2/n\ge7/8>45/64\). Thus \(h\) violates the low-vote condition at \(\nu\).

\proofstep{The three displayed quantities.} The sole error of \(h=c_1\) is at the atom \((0,+)\), whence \(\mathbb E_\nu[yh]=1-2\delta=1-4/n\). At that atom \((h-g_\nu)^2\le4\), and at an atom \((i,+)\) with \(i>0\) it equals \((2B/D)^2\le(2/(n-1))^2\), because \(D\ge(n-1)B\). So \(\mathbb E_\nu(h-g_\nu)^2\le4\delta+4/(n-1)^2=8/n+4/(n-1)^2\). By the atom-gradient identity in the proof of Lemma~\ref{res:3.6}, the tilt multiplies the mass of an atom \(z=(i,+)\) by \(e^{\langle\nabla\Psi,a_z\rangle/\lambda}=e^{Dg_\nu(i)/\lambda}\), up to normalization. The gap between the atom at \(0\) and the others is \(D(g_\nu(0)-g_\nu(i))=2(B-A)=\Delta\), so the tilted mass of the error atom is
\[
\frac{\delta e^{Dg_\nu(0)/\lambda}}{\delta e^{Dg_\nu(0)/\lambda}+(1-\delta)e^{Dg_\nu(1)/\lambda}}
=\frac{\delta e^{\Delta/\lambda}}{1-\delta+\delta e^{\Delta/\lambda}},
\]
which converges to one as \(\lambda\downarrow0\), because \(\Delta>0\).

\proofstep{A continuous decoder with output \(h\) at \(\nu\).} Let \(a(\tau)=A(\Phi(\tau))/Z(\Phi(\tau))\) be the normalized signed potential weights of a realizable distribution \(\tau\), where \(A(\cdot)\) is the potential vector of Section~\ref{ssec:2.5}, not the number \(A\) above. It is continuous because \(Z\ge1\) by \eqref{eq:res-3.2}, and \(\|a(\tau)\|_1=1\). Let \(\zeta(\tau)=(1-8\|\tau-\nu\|_1/\delta)_+\). Suppose \(\zeta(\tau)>0\), so that \(\|\tau-\nu\|_1<\delta/8=1/(4n)\). The smallest mass of \(\nu\) is \(\frac{n-2}{n(n-1)}>\frac1{4n}\), because \(3n>7\). Hence every positive-label atom keeps positive mass under \(\tau\), and the unique realizing concept of \(\tau\) is \(c_0\). Further \(\max_i\tau(i)\le\delta+\delta/8=\frac98\delta\le\frac9{64}\), so all profile coordinates of \(\tau\) are at least \(1-\frac9{32}>0\), and \(a(\tau)\) is a probability vector.

Put \(F(\tau)=(1-\zeta(\tau))a(\tau)+\zeta(\tau)e_1\), where \(e_1\) is the basis vector of the row \(c_1\). Inside the bump both terms are multiples of probability vectors, and outside it \(F=a\). So \(F\) is a continuous map into \(K=\partial B_1^{n+1}\), the unit sphere of \(\ell_1^{n+1}\). With the continuous decoder \(W(q)(i)=\sum_rq_rc_r(i)\in[-1,1]\) we have \(W(a(\tau))=g_\tau\), and Lemma~\ref{res:3.7} gives \(\mathbb E_\tau[yg_\tau]\ge31/32\) for every realizable \(\tau\). Inside the bump also \(\mathbb E_\tau[yc_1]=1-2\tau(0)\ge1-\frac94\delta\ge\frac{23}{32}\). As \(W(F(\tau))=(1-\zeta(\tau))g_\tau+\zeta(\tau)c_1\) inside the bump and \(W(F(\tau))=g_\tau\) outside it, \(\mathbb E_\tau[yW(F(\tau))]\ge23/32\) for every realizable \(\tau\).

Every \(q\in K\) is the squared soft vote of a unit vector: \(W(q)=f_w\) with \(w_r=\operatorname{sgn}(q_r)\sqrt{|q_r|}\), since \(\|w\|_2=1\) and \(w_r|w_r|=q_r\). At \(\nu\) itself \(\zeta(\nu)=1\) and \(F(\nu)=e_1\), so the decoder outputs \(h\). The advantage \(1-4/n\) tends to one only along the distributions \(\nu=\nu_n\); the lower bound valid for all realizable distributions is \(23/32\).
\end{proof}

\subsection{Tilt dimension and vote-image covering dimension}\label{ssec:7.8}

The proof of Theorem~\ref{res:3.8} composes a continuous map from \(\Delta_C\) to a finite complex with fixed labels on the vertices of that complex, and its transport step uses only the low-vote condition of Definition~\ref{def:low-vote} for these labels. This suggests a numerical parameter, the least dimension of a complex that admits such a map, which we call the tilt dimension. Since this condition is a bound on the vote, a second parameter is natural, namely the covering dimension of the vote image at a fixed scale. In this subsection we place the two parameters in the chain \(\operatorname{lr}-1\le k_{\mathrm{tilt}}\le\dim_\eta(V_B)\), bound the covering dimension from above, and compute both parameters on explicit families.

\subsubsection*{The two parameters}

Throughout, \(C=C_B\), and we set
\[
a=31/32,
\qquad
b=45/64,
\qquad
\eta_0=(a-b)/2=17/128.
\]
Thus \(a\) and \(b\) are the constants of Section~\ref{ssec:7.5}, and \(\eta_0\) is half of the margin \(a-b\) available in Lemma~\ref{res:3.6}.

\begin{definition}[Tilt dimension]\label{def:tilt-dimension}
The parameter \(k_{\mathrm{tilt}}(B)\) is the minimum dimension of a nonempty finite simplicial complex \(K\) for which there are a continuous map
\[
F:\Delta_C\to|K|
\]
and fixed vertex hypotheses \(h_v\) satisfying
\[
F(\nu)\in D_v,\quad
\nu_{i,y}>0,\quad
h_v(i)\ne y
\quad\Longrightarrow\quad
yg_\nu(i)\le b,
\]
where \(D_v\) is the closed region of \(|K|\) on which the barycentric coordinate of \(v\) is maximal. We also call this implication the \emph{low-vote condition}: it says that \(h_v\) satisfies the condition of Definition~\ref{def:low-vote} at every \(\nu\in F^{-1}(D_v)\). A complex with such a map and such hypotheses is a \emph{low-vote parameterization}.
\end{definition}

In words, at every atom of positive mass where the label of the active vertex is wrong, the potential vote does not confidently support the true label. The map \(F\) need not be affine, \(K\) need not embed in \(\mathbb R^{\dim K}\), and the labels may be improper.

Let
\[
V_B=\{g_\nu:\nu\in\Delta_C\}\subseteq P_B:=\operatorname{conv}\{\pm c_r:r\in[N]\}\subseteq[-1,1]^M.
\]
The first inclusion holds because \(g_\nu=\sum_r(A_r/Z)c_r\) with \(\sum_r|A_r|/Z=1\). The map \(\nu\mapsto g_\nu\) is continuous, since \(Z\ge1\) on \(\Phi(\Delta_C)\) by \eqref{eq:res-3.2}, so \(V_B\) is compact. For a compact set \(V\subseteq[-1,1]^M\), the parameter \(\dim_\eta(V)\) is the minimum multiplicity minus one of a finite closed cover of \(V\) by sets of \(\ell_\infty\)-diameter \(<\eta\). Such covers exist, for instance by small cubes. Restricting a cover shows \(\dim_\eta(V_B)\le\dim_\eta(P_B)\), and \(\dim_\eta\) is nonincreasing in \(\eta\), because a cover admissible at \(\eta\) is admissible at every larger scale.

The first comparison is the transport argument of Section~\ref{ssec:3} applied to an arbitrary low-vote parameterization.

\begin{resultenv}{Lemma}{res:7.11}{The low-vote condition gives tilt covers}
Let \(K\) be a nonempty finite simplicial complex of dimension \(k\), and let \(F:\Delta_C\to|K|\) and vertex hypotheses \(h_v\) satisfy the low-vote condition. Then \(\operatorname{lr}(C,\varepsilon)\le k+1\) for every \(\varepsilon\in(0,1/2)\). Consequently
\[
\operatorname{lr}(B)-1\le k_{\mathrm{tilt}}(B).
\]
\end{resultenv}

\begin{proof}
Fix \(\varepsilon\in(0,1/2)\) and put \(R_v=F^{-1}(D_v)\). These sets are closed, cover \(\Delta_C\), and have multiplicity at most \(k+1\) by Lemma~\ref{res:2.0}. Let \(\nu\in R_v\). By the low-vote condition, every atom of positive \(\nu\)-mass at which \(h_v\) errs satisfies \(yg_\nu(i)\le b=45/64\), where \(g_\nu=f_{u(\Phi(\nu))}\) (Section~\ref{ssec:2.5}). Further \(\mathbb E_\nu[yg_\nu]\ge31/32=a\) by \eqref{eq:res-3.3} and \(D=Z(\Phi(\nu))\ge1\) by \eqref{eq:res-3.2}. Lemma~\ref{res:3.6} applies with \(D_0=1\) and \(\lambda_\varepsilon=17/(64\ln(4/\varepsilon))\). In its proof only atoms of positive mass enter, in the numerator as well as in the denominator, so atoms of zero mass contribute to neither side, and we may restrict its hypothesis to atoms of positive mass. Since \(a-b=17/64\), it gives
\[
\operatorname{loss}_{\Theta_{\lambda_\varepsilon}(\nu)}(h_v)\le e^{-(17/64)/\lambda_\varepsilon}=\varepsilon/4,
\]
exactly as in the proof of Theorem~\ref{res:3.8}. The sets \(A_v=\Theta_{\lambda_\varepsilon}(R_v)\) are closed as compact images, cover \(\Delta_C\) by surjectivity of the tilt (Lemma~\ref{res:3.5}), and have multiplicity at most \(k+1\) by injectivity. Merging equal labels does not increase the multiplicity. On \(A_v\) the loss of \(h_v\) is at most \(\varepsilon/4\le\varepsilon/2\), so Lemma~\ref{res:3.3} gives \(\operatorname{lr}(C,\varepsilon)\le k+1\). As \(\operatorname{lr}(C,\varepsilon)\) is nonincreasing in \(\varepsilon\), the supremum over \(\varepsilon>0\) is also at most \(k+1\). Taking the minimum over low-vote parameterizations gives the consequence.
\end{proof}

The covering dimension of the vote image depends on the vote map, whose topology is simple: on the profile space it is injective. This gives a bound on the covering dimension by the rank at every scale.

\begin{resultenv}{Lemma}{res:7.12}{The vote map is a homeomorphism}
Let \(Y=\Phi(\Delta_{C_B})=\{v\in Q:\max_rv_r=1\}\) and
\[
G_B(v)=\frac{B^TA(v)}{Z(v)},
\]
so that \(g_\nu=G_B(\Phi(\nu))\) and \(V_B=G_B(Y)\). Then \(G_B\) is a homeomorphism of \(Y\) onto \(V_B\), and likewise of \(X_B\) onto \(G_B(X_B)\). Consequently, for every \(\eta>0\),
\[
\dim_\eta(V_B)\le\operatorname{rank}_{\mathbb R}B-1.
\]
\end{resultenv}

\begin{proof}
On \(X_B\) we have \(Z\ge1\), so \(G_B\) is continuous there. It suffices to prove injectivity on \(X_B\), since \(Y\) and \(X_B\) are compact and a continuous injection of a compact space into a Hausdorff space is a homeomorphism onto its image.

\proofstep{Injectivity.} Suppose \(G_B(v)=G_B(v')\) with \(v,v'\in X_B\), and put
\[
c=\left(\frac{Z(v)}{Z(v')}\right)^{1/(2p)}>0.
\]
Since \(A(cv')=c^{2p}A(v')\) and \(c^{2p}Z(v')=Z(v)\), we have \(G_B(v')=B^TA(cv')/Z(v)\). So the assumption gives \(B^T\bigl(A(v)-A(cv')\bigr)=0\), that is, \(A(v)-A(cv')\in\ker B^T=(\operatorname{im}B)^\perp\). Both \(v\) and \(v'\) lie in \(Q\subseteq\operatorname{im}B\), so \(v-cv'\in\operatorname{im}B\) and
\[
\sum_r(v_r-cv'_r)\bigl(A_r(v)-A_r(cv')\bigr)=0 .
\]
The scalar function \(t\mapsto\operatorname{sgn}(t)|t|^{2p}\) is strictly increasing, so every summand is nonnegative and vanishes only if \(v_r=cv'_r\). Therefore \(v=cv'\). As \(\|v\|_\infty=\|v'\|_\infty=1\), we get \(c=1\).

\proofstep{The dimension bound.} Let \(r=\operatorname{rank}B\). As \(Q\subseteq[-1,1]^N\), each set \(Q\cap\{v_j=1\}\) is a face of \(Q\) or empty, and \(Y\) is the union of these faces. Each is contained in the affine space \(\operatorname{im}B\cap\{v_j=1\}\), which has dimension \(r-1\), because the coordinate \(v_j\) does not vanish identically on \(\operatorname{im}B\) (the columns of \(B\) have entries \(\pm1\)). So \(Y\) is the support of a finite polyhedral complex of dimension at most \(r-1\). By uniform continuity of \(G_B\) on the compact set \(Y\), we can subdivide this complex to a finite simplicial complex so fine that \(G_B\) oscillates by less than \(\eta\) in \(\ell_\infty\) on every closed vertex star. Let \(D_v\) be the regions of Lemma~\ref{res:2.0} for this subdivision. They are closed, cover \(Y\), have multiplicity at most \(r\), and each lies in the closed star of its vertex, so the compact set \(G_B(D_v)\) has \(\ell_\infty\)-diameter less than \(\eta\). Since \(G_B\) is a homeomorphism of \(Y\) onto \(V_B\), the sets \(G_B(D_v)\) form a closed cover of \(V_B\) with diameters below \(\eta\) and multiplicity at most \(r\).
\end{proof}

The vote map thus preserves the topology of the profile space. Its fixed-scale metric behavior can be very different, as the comparison of Theorem~\ref{res:7.1} with Theorem~\ref{res:7.15} below shows.

The lower bounds on covers below rest on the following form of Lebesgue's covering lemma.

\begin{lemma}\label{lem:cube-cover-multiplicity}
Let \(d\ge1\), and let \(\{F_j\}\) be a finite closed cover of the cube \([-1,1]^d\) in which no member meets two opposite facets. Then its multiplicity is at least \(d+1\).
\end{lemma}

\begin{proof}
Suppose the multiplicity is at most \(d\), and discard empty members. For every subfamily \(J\) of size \(d+1\), the function \(z\mapsto\max_{j\in J}\operatorname{dist}(z,F_j)\) is continuous and positive on the cube, so it has a positive minimum. Each \(F_j\) has positive distance from every facet it does not meet. Choose \(\rho>0\) smaller than all these finitely many numbers, and let \(U_j\) be the open \(\rho\)-neighborhood of \(F_j\) in the cube. The sets \(U_j\) form an open cover of multiplicity at most \(d\), and \(U_j\) meets the same facets as \(F_j\).

Label each \(j\) by a vertex \(\ell_j\in\{\pm1\}^d\) such that \((\ell_j)_i=\pm1\) whenever \(F_j\) meets the facet \(z_i=\pm1\). This is possible because \(F_j\) does not meet both, and the other coordinates of \(\ell_j\) are arbitrary. Let \(\{\varphi_j\}\) be a partition of unity subordinate to \(\{U_j\}\), and put \(\Psi(z)=\sum_j\varphi_j(z)\ell_j\). If \(z_i=1\), every \(j\) with \(\varphi_j(z)>0\) has \(z\in U_j\), hence \((\ell_j)_i=1\), and so \(\Psi_i(z)=1\). Likewise \(\Psi_i(z)=-1\) if \(z_i=-1\). For \(w\in(-1,1)^d\), the map \(\Psi-w\) therefore has its \(i\)-th coordinate positive on \(z_i=1\) and negative on \(z_i=-1\), for every \(i\). The Poincar\'e--Miranda theorem (a continuous map \(f:[-1,1]^d\to\mathbb R^d\) with \(f_i\ge0\) on \(z_i=1\) and \(f_i\le0\) on \(z_i=-1\) for all \(i\) has a zero) gives \(w\in\Psi([-1,1]^d)\). So the image of \(\Psi\) contains the open cube. But at each \(z\) at most \(d\) of the functions \(\varphi_j\) are positive, so the image lies in the finite union of the convex hulls of at most \(d\) labels. Each such hull has dimension at most \(d-1\), so the union has empty interior, a contradiction.
\end{proof}

\subsubsection*{Comparison of the parameters}

The next theorem gives the comparison. A cover of the vote image of mesh below \(b\) yields a low-vote parameterization, because a vote within \(b\) of a fixed vote cannot confidently contradict its signs. At the fixed scale \(\eta_0\), the net argument of Section~\ref{ssec:3} bounds the covering dimension, and this bound does not extend to small scales, where the covering dimension of a Hadamard vote image is exponential in \(m\).

\begin{resultenv}{Theorem}{res:7.13}{Tilt and vote dimensions}
(i) For every \(\eta\) with \(0<\eta\le b\),
\[
\operatorname{lr}(B)-1
\le
k_{\mathrm{tilt}}(B)
\le
\dim_\eta(V_B).
\]
(ii) At the fixed scale \(\eta_0=17/128\),
\[
\dim_{\eta_0}(V_B)\le2^{24}\kappa(N,M);
\]
by monotonicity the same bound holds for every \(\eta\in[\eta_0,b]\). For large \(m\) it does not extend to arbitrarily small \(\eta\): for \(H_{2^m}\), the vote image of the positive row-\(0\) simplex contains a homeomorphic copy of a simplex of dimension \(2^m-1\) (Lemma~\ref{res:7.12}), so \(\dim_\eta(V_{H_{2^m}})\ge2^m-1\) once \(\eta\) is small enough.

(iii) For every \(\eta>0\), \(\dim_\eta(V_B)\le\operatorname{rank}_{\mathbb R}B-1\), and the vote map is a homeomorphism of \(Y\) onto \(V_B\) (Lemma~\ref{res:7.12}).

(iv) For the anchored singleton class \(B_n\), \(n\ge3\), we have \(k_{\mathrm{tilt}}(B_n)=1\). For \(B_4\), we have \(\operatorname{lr}=2\), \(\operatorname{Ind}=2\), \(k_{\mathrm{tilt}}=1\), and \(\dim_{\eta_0}(V_{B_4})=3\).
\end{resultenv}

\begin{proof}
We prove the parts in order; part (iii) is Lemma~\ref{res:7.12}.

\proofstep{Part (i).} The first inequality is Lemma~\ref{res:7.11}. For the second, let \(d=\dim_\eta(V_B)\) and take a finite closed cover \(\{W_j\}\) of \(V_B\) by nonempty sets, of multiplicity at most \(d+1\) and with diameters less than \(\eta\). For every subfamily \(J\) of size \(d+2\), the function \(x\mapsto\max_{j\in J}\operatorname{dist}(x,W_j)\) has a positive minimum on the compact set \(V_B\). Choose \(\rho>0\) smaller than all these minima and smaller than \((\eta-\max_j\operatorname{diam}W_j)/2\), and let \(U_j\) be the open \(\rho\)-neighborhood of \(W_j\) in \(V_B\). The open cover \(\{U_j\}\) has multiplicity at most \(d+1\), and \(\operatorname{diam}U_j<\eta\).

Choose \(g^{(j)}\in U_j\), label \(h_j=\operatorname{sign}g^{(j)}\) with \(\operatorname{sign}0=+1\), take a partition of unity \(\{\varphi_j\}\) subordinate to \(\{U_j\}\), and put
\[
F(\nu)=\sum_j\varphi_j(g_\nu)e_j\in|K|,
\]
where \(K\) is the nerve of \(\{U_j\}\). The map \(F\) is continuous, and \(\dim K\le d\) because a nonempty intersection involves at most \(d+1\) members. If \(F(\nu)\in D_j\), then \(\varphi_j(g_\nu)>0\), since the maximal barycentric coordinate is positive, so \(g_\nu\in U_j\). If moreover \(h_j(i)\ne y\), then \(yg^{(j)}_i\le0\), in both cases \(y=\pm1\), and \(|g_\nu(i)-g^{(j)}_i|<\eta\), so \(yg_\nu(i)<\eta\le b\). This is the low-vote condition, atom by atom, and it gives \(k_{\mathrm{tilt}}(B)\le d\).

\proofstep{Part (ii): the setup.} If $N=1$, then $\Phi\equiv1$, $V_B$ is a single point, and its covering dimension is zero. Assume $N\ge2$. Put \(p=p_N\ge\lceil16\ln4\rceil=23\), \(s'=2^{40}p^2\) and \(\delta=2^{-10}\).

\proofstep{Part (ii): an empirical net.} Let \(v\in Y\), say \(v=\Phi(\nu)\) with \(\nu\in\Delta_C\). Let \(V_1,\ldots,V_{s'}\) be independent signed columns with \(\mathbb P(V_\ell=a_z)=\nu_z\), and put \(\widehat v=\frac1{s'}\sum_\ell V_\ell\), so that \(\mathbb E\widehat v=v\). A coordinate with \(v_r=1\) has \(a_z(r)=1\) on the support of \(\nu\), so it stays equal to \(1\) in \(\widehat v\). Hence \(\widehat v\ne0\). Fix a row \(r\), write \(x=v_r\) and \(h=\widehat v_r-x\). The moment estimates in the proof of Lemma~\ref{res:3.0}, which use only the independence of the draws and hold with \(s'\) in place of \(s\), give for real \(t\)
\[
\mathbb Ee^{th}\le e^{t^2/(2s')},
\qquad
\mathbb P(|h|>1/4)\le2e^{-s'/32},
\qquad
\mathbb E\bigl[h^2e^{|t||h|}\bigr]\le\frac4{s'}e^{t^2/s'}.
\]

If \(|x|\ge1/4\), the mean value theorem and \(1+y\le e^y\) give
\[
|\phi_p(x+h)-\phi_p(x)|\le p|h|(|x|+|h|)^{p-1}\le p|h|\,|x|^{p-1}e^{(p-1)|h|/|x|}.
\]
Square this, take \(t=2(p-1)/|x|\le8p\), and use \(|x|^{2p-2}\le16|x|^{2p}\):
\[
\mathbb E|\phi_p(x+h)-\phi_p(x)|^2
\le16p^2|x|^{2p}\cdot\frac4{s'}e^{64p^2/s'}
\le\frac{256p^2}{s'}|x|^{2p},
\]
because \(64p^2/s'=2^{-34}\) and \(e^{2^{-34}}\le4\). If \(|x|<1/4\), then on \(\{|h|\le1/4\}\) both \(|x|\) and \(|x+h|\) are below \(1/2\), so the squared difference is at most \(4\cdot4^{-p}\), and on the complement it is at most \(4\). So its expectation is at most \(4\cdot4^{-p}+8e^{-s'/32}\). Summing over the rows and dividing by \(\|\phi_p(v)\|_2^2=\sum_r|v_r|^{2p}\ge1\),
\[
\frac{\mathbb E\|\phi_p(\widehat v)-\phi_p(v)\|_2^2}{\|\phi_p(v)\|_2^2}
\le\frac{256p^2}{s'}+4N4^{-p}+8Ne^{-s'/32}.
\]
On the right, \(256p^2/s'=2^{-32}\). Since \(p\ge16\ln(2N)\), we have \(N\le e^{p/16}/2\). With \(p\ge23\) this gives \(4N4^{-p}\le2e^{p/16}4^{-p}\le2^{-42}\) and \(8Ne^{-s'/32}\le4e^{p/16}e^{-2^{35}p^2}\le2^{-42}\). So the right side is less than \(2^{-27}\). By \eqref{eq:normalization}, normalization to the unit sphere increases the squared error by a factor of at most \(4\), so some outcome satisfies \(\|u_p(\widehat v)-u_p(v)\|_2^2<2^{-25}\), hence \(\|u_p(\widehat v)-u_p(v)\|_2<\delta\).

\proofstep{Part (ii): the small-fiber map.} Let \(Y'\) be the set of averages of \(s'\) signed columns and \(Z'=u_p(Y'\setminus\{0\})\), a symmetric set of at most \((2M)^{s'}\) unit vectors within \(\delta\) of every point of \(u_p(Y)\). Apply to \(Z'\) the Gaussian map and the Lipschitz extension with odd symmetrization from the proof of Theorem~\ref{res:3.1}. The first gives a linear map \(T\) into a space of dimension at most \(k':=\lceil32(2s'\ln(2M)+\ln16)\rceil\) with \(\frac12\|z-z'\|_2^2\le\|Tz-Tz'\|_2^2\le\frac32\|z-z'\|_2^2\) on \(Z'\cup\{0\}\). The second gives an odd \(\sqrt{3/2}\)-Lipschitz map \(G\) agreeing with \(T\) on \(Z'\), which we regard, padded with zero coordinates, as a map into \(\mathbb R^{k'}\). For \(w,w'\in u_p(Y)\) choose \(z,z'\in Z'\) within \(\delta\) of them. Then
\[
\|w-w'\|_2\le2\delta+\|z-z'\|_2\le2\delta+\sqrt2\|G(z)-G(z')\|_2\le2(1+\sqrt3)\delta+\sqrt2\,\|G(w)-G(w')\|_2,
\]
using the Lipschitz bound for the two approximation errors, with \(\sqrt2\cdot2\sqrt{3/2}=2\sqrt3\). This is the analogue of \eqref{eq:res-3.1} with \(1/16\) replaced by \(\delta\). The vote Lipschitz bound \eqref{eq:res-3.4} gives
\[
\|f_w-f_{w'}\|_\infty\le4(1+\sqrt3)\delta+2\sqrt2\,\|G(w)-G(w')\|_2 ,
\]
and \(4(1+\sqrt3)\delta<0.011<\eta_0\).

\proofstep{Part (ii): the cover.} Since \(G(0)=0\), we have \(\|G(w)\|_2\le\sqrt{3/2}<2\) for unit \(w\), so \(G\circ u_p\) maps \(Y\) continuously into the cube \([-2,2]^{k'}\). Triangulate this cube so finely that \(2\sqrt2\) times the Euclidean diameter of every closed vertex star is less than \(\eta_0-4(1+\sqrt3)\delta\), and let \(D_a\) be the regions of Lemma~\ref{res:2.0} for this triangulation. The sets \(\{y\in Y:G(u_p(y))\in D_a\}\) are closed, cover \(Y\), and have multiplicity at most \(k'+1\). Since \(D_a\subseteq\operatorname{St}(a)\) and \(G_B(y)=f_{u_p(y)}\), two points of the same set have votes at \(\ell_\infty\)-distance less than \(\eta_0\). By Lemma~\ref{res:7.12}, the images of these sets under the homeomorphism \(G_B\) form a closed cover of \(V_B\) by compact sets of \(\ell_\infty\)-diameter less than \(\eta_0\), with the same multiplicity. Hence \(\dim_{\eta_0}(V_B)\le k'\). Finally \(s'=2^{24}s_N\), so \(32(2s'\ln(2M)+\ln16)\le2^{24}\cdot32(2s_N\ln(2M)+\ln16)\le2^{24}\kappa(N,M)\). As the right side is an integer, \(k'\le2^{24}\kappa(N,M)\). This bound is of order \(\log^2N\log M\).

\proofstep{Part (ii): small \(\eta\).} Let \(H=H_{2^m}\) and \(d=2^m-1\). The distributions realized by row \(0\) are mapped by \(\Phi\) onto the positive row-\(0\) simplex \(\Delta_0=\{H\theta:\theta\ge0,\ \sum_j\theta_j=1\}\). This is contained in \(Y\), since row \(0\) has coordinate \(1\), and it is a \(d\)-dimensional simplex because \(H\) is invertible. By Lemma~\ref{res:7.12}, composing a homeomorphism \([-1,1]^d\to\Delta_0\) with \(G_H\) gives a continuous injection \(\psi:[-1,1]^d\to V_H\). The images of two opposite facets under \(\psi\) are disjoint compact sets. Let \(\eta_*>0\) be the least distance among the \(d\) pairs. If \(\eta\le\eta_*\), a closed cover of \(V_H\) of mesh less than \(\eta\) pulls back along \(\psi\) to a closed cover of the cube in which no member meets two opposite facets, so Lemma~\ref{lem:cube-cover-multiplicity} gives multiplicity at least \(d+1\). Hence \(\dim_\eta(V_H)\ge2^m-1\) once \(\eta\) is small enough. For large \(m\) this exceeds \(2^{24}\kappa(2^m,2^m)=O(m^3)\).

\proofstep{Part (iv): anchored singletons.} The columns of \(B_n\) are \(0,1,\ldots,n\). Write \(h_0=(+,-,\ldots,-)\), positive on the anchor \(0\) only, and let \(O_t\) be the set of realizable \(\nu\) whose unique compatible concept is \(c_t\). If \(\nu\in O_t\), then each \(c_s\), \(s\ne t\), errs at an atom of positive \(\nu\)-mass, and such an atom keeps positive mass nearby. So the sets \(O_t\) are relatively open. They are pairwise disjoint.

Let \(\nu\) have at least two compatible concepts. Then there is no positive non-anchor atom, because \((j,+)\) with \(j\ge1\) is compatible with \(c_j\) only, and \((0,-)\) is compatible with no concept. So the atoms of \(\nu\) are \((0,+)\) and \((r,-)\), \(r\in[n]\). Writing \(\mu_r=\nu_{r,-}\), the row \(c_r\) errs exactly at \((r,-)\), so \(\Phi_r=1-2\mu_r\). Let \(P\) and \(Q\) be the total potential weights \(|\Phi_r|^{2p}\) of the rows with \(\Phi_r>0\) and with \(\Phi_r<0\), respectively. The compatible rows have profile \(1\), so \(P\ge2\). At most one \(\mu_r\) exceeds \(1/2\), so \(Q\le1\). For a compatible \(t\), we have \(A_t=1\), the row \(c_t\) is positive at the column \(t\), and all other rows are negative there. So \(g_\nu(t)=(2A_t-\sum_rA_r)/Z=(2-P+Q)/(P+Q)\), which is decreasing in \(P\) and, as \(P\ge1\), increasing in \(Q\). Therefore
\[
g_\nu(t)=\frac{2-P+Q}{P+Q}\le\frac{2-2+1}{2+1}=\frac13 .
\]

Define \(\ell_t(\nu)=\min\{1,4(g_\nu(t)-1/2)_+\}\) for \(\nu\in O_t\) and \(\ell_t(\nu)=0\) otherwise. On \(O_t\) this is continuous. A boundary point \(\nu^*\) of \(O_t\) is compatible with \(c_t\), because compatibility is a closed condition, and it does not lie in the open set \(O_t\). So it has at least two compatible concepts, and \(g_{\nu^*}(t)\le1/3\). By continuity of \(g\), we have \(g_\nu(t)<1/2\) near \(\nu^*\), so \(\ell_t\) vanishes near \(\nu^*\). Thus all \(\ell_t\) are continuous, and at most one of them is nonzero at each point.

Let \(K\) be the star graph with center \(0\) and \(n\) edges of length \(1\), and let \(F(\nu)\) be the point at parameter \(\ell_t(\nu)\) on edge \(t\) (the center if all \(\ell_t\) vanish); \(F\) is continuous. Subdivide each edge at \(1/4,1/2,3/4\), and label the center and the points \(1/4,1/2\) by \(h_0\), the points \(3/4\) and \(1\) of edge \(t\) by \(c_t\). We verify the low-vote condition on closed stars, which contain the regions \(D_v\) by Lemma~\ref{res:2.0}.

The closed star of an \(h_0\)-vertex consists of points whose parameter on their edge is at most \(3/4\). So if \(F(\nu)\) lies in it and \(\nu\in O_t\), then \(\ell_t(\nu)\le3/4\), hence \(g_\nu(t)\le\frac12+\frac3{16}=\frac{11}{16}<b\). In the unique-concept region \(O_t\), the only possible error atom of \(h_0\) is \((t,+)\), since \(h_0\) errs only at atoms \((j,+)\), \(j\ge1\), and \(c_t\) has no such atom with \(j\ne t\). In the multi-concept region \(h_0\) makes no error, as its atoms were listed above. The closed star of a \(c_t\)-vertex lies in the part of edge \(t\) with parameter at least \(1/2\). So \(\ell_t(\nu)\ge1/2>0\), hence \(\nu\in O_t\) and \(c_t\) makes no error. Thus the low-vote condition holds and \(k_{\mathrm{tilt}}(B_n)\le1\). Theorem~\ref{res:5.4} and Lemma~\ref{res:7.11} give \(k_{\mathrm{tilt}}(B_n)\ge\operatorname{lr}(B_n)-1=1\).

\proofstep{Part (iv): the class \(B_4\).} For \(y\in[-1/12,1/12]^3\) put \(w_i=\frac14+\frac{y_i}2\) (\(i=1,2,3\)) and \(w_4=\frac14-\frac{y_1+y_2+y_3}2\), so that \(\sum_iw_i=1\) and \(\frac18\le w_i\le\frac38\). With \(w_*=\max_iw_i\) and \(p=p_4=\lceil16\ln8\rceil=34\), put \(v_i=(w_i/w_*)^{1/68}\), \(\nu_{i,-}=(1-v_i)/2\) and \(\nu_{0,+}=1-\sum_{i=1}^4\nu_{i,-}\). Since \(w_i/w_*\ge1/3\), we have \(v_i\ge3^{-1/68}>1/2\), so \(\nu_{i,-}<1/4\) and the anchor mass is positive. The row of maximal weight has no negative atom, so \(\nu\) is realized by that row, and the construction is continuous in \(y\). Its profile is \(\Phi_i=1-2\nu_{i,-}=v_i\), all profile coordinates are positive, and \(v_i^{68}=w_i/w_*\) shows that the normalized potential weights are exactly \(w\). As the anchor column is positive in all rows and column \(i\) is positive in row \(i\) only, \(g_\nu(0)=1\) and \(g_\nu(i)=w_i-\sum_{j\ne i}w_j=-1+2w_i\), that is,
\[
(g_1,g_2,g_3)=(-1/2,-1/2,-1/2)+y .
\]
This is a continuous embedding of the cube into the vote image. Take any finite closed cover of \(V_{B_4}\) of mesh less than \(\eta_0\), and pull it back to the cube. Two points of opposite faces \(y_i=\pm1/12\) have votes at \(\ell_\infty\)-distance at least \(1/6>\eta_0\), so no member meets two opposite faces. After rescaling the cube to \([-1,1]^3\), Lemma~\ref{lem:cube-cover-multiplicity} gives multiplicity at least \(4\), so \(\dim_{\eta_0}(V_{B_4})\ge3\). Conversely, the rows of \(B_4\) are \(c_i=c_1+2(e_i-e_1)\) with \(c_1\) nonzero at the anchor, so \(\operatorname{rank}B_4=4\), and Lemma~\ref{res:7.12} gives \(\dim_{\eta_0}(V_{B_4})\le3\). The values \(\operatorname{lr}=2\) and \(\operatorname{Ind}=2\) are Theorem~\ref{res:5.4}, and \(k_{\mathrm{tilt}}=1\) is the anchored-singleton case.
\end{proof}

The class \(B_4\) thus separates \(k_{\mathrm{tilt}}\) from \(\dim_{\eta_0}\) by a constant, so the second inequality of Theorem~\ref{res:7.13}(i) can be strict.

The low-vote condition is one-sided. It does not require the label to follow the vote on every confident atom; it constrains only atoms where the vote confidently supports the true label. For example, in the anchored \(16\times17\) class \(B_{16}\), where \(p=p_{16}=\lceil16\ln32\rceil=56\), let
\[
\nu_{0,+}=1023/1024,
\qquad
\nu_{1,+}=1/1024.
\]
Then \(\Phi_1=1\) and \(\Phi_r=511/512\) for \(r\ne1\). The column \(1\) is positive in row \(1\) only, so with \(x=(511/512)^{112}\) we get \(g_\nu(1)=(1-15x)/(1+15x)\). Bernoulli's inequality gives \(x\ge1-112/512=25/32\), so \(15x\ge375/32\) and, as \(R\mapsto(1-R)/(1+R)\) is decreasing, one has
\[
g_\nu(1)\le-343/407<-45/64,
\]
while \(h=c_1\) makes no error. So this hypothesis satisfies the low-vote condition at \(\nu\), although the vote confidently opposes it at the atom \((1,+)\).

\subsubsection*{Balanced version spaces}

The labels of closure chains are correct on the heavy atoms of Section~\ref{ssec:7.6}, whereas the low-vote condition concerns all atoms of positive mass. The two are linked when the version space of the heavy atoms is balanced at every column that those atoms leave undetermined. In that case the vote nearly vanishes on such a column, so a concept that agrees with the heavy atoms errs only where the vote is small. The next theorem makes this precise through an involution of the rows. Its hypothesis holds for affine binary-code classes \(c_u(i)=c_0(i)(-1)^{\langle u,v_i\rangle}\), with rows indexed once by each \(u\in\mathbb F_2^d\), and on Hadamard matrices it gives the exact value \(k_{\mathrm{tilt}}(H_{2^m})=m\), far below the bound of Theorem~\ref{res:7.13}(ii).

\begin{resultenv}{Theorem}{res:7.14}{Balanced version spaces}
Let \(B\) be a total matrix. A column \(i\) is \emph{forced} by a realizable atom set \(T\) if the concepts consistent with \(T\) all agree at \(i\), and \emph{unforced} by \(T\) otherwise. We say that \(B\) satisfies the \emph{involution hypothesis} if, for every realizable atom set \(T\) and every column \(i\) unforced by \(T\), there is an involution of the whole set of row indices, not only of the indices of rows consistent with \(T\), that preserves the labels of the columns in \(T\) and reverses column \(i\).

(i) For all realizable \(\nu,\nu'\),
\[
\|g_\nu-g_{\nu'}\|_\infty
\le
4pN\|\nu-\nu'\|_1.
\]

(ii) With
\[
\tau=\frac{b}{16pNM},
\qquad
T(\nu)=\{z:\nu_z\ge\tau\max\nu\},
\]
and \(\nu_T\) the renormalized restriction of \(\nu\) to \(T(\nu)\), we have \(\|g_\nu-g_{\nu_T}\|_\infty\le b/2\).

(iii) Under the involution hypothesis, if \(\nu\) is supported on a realizable atom set \(T\) and the column \(i\) is not forced by \(T\), then \(g_\nu(i)=0\); and, with the notation of (ii), a column \(i\) not forced by \(T(\nu)\) satisfies \(|g_\nu(i)|\le b/2\).

(iv) Under the involution hypothesis, \(k_{\mathrm{tilt}}(B)\le h_+^*(B)-1\). In particular, \(k_{\mathrm{tilt}}(H_{2^m})=m=\operatorname{lr}-1\).
\end{resultenv}

\begin{proof}
We first show that the vote vanishes on unforced columns, then pass from \(\nu\) to its heavy atoms, and finally use the mass-prefix map of the closure chains.

\proofstep{Cancellation.} Let \(\nu\) be supported on \(T\), let \(i\) be unforced by \(T\), and let \(\pi\) be the involution of the row indices given by the hypothesis, so that \(c_{\pi(r)}(j)=c_r(j)\) on the columns \(j\) of \(T\) and \(c_{\pi(r)}(i)=-c_r(i)\). In particular \(\pi\) has no fixed point. The profile of \(\nu\) only involves the columns of \(T\), so \(\Phi(\nu)_{\pi(r)}=\Phi(\nu)_r\) and \(A_{\pi(r)}=A_r\). Hence the contributions \(A_rc_r(i)\) and \(A_{\pi(r)}c_{\pi(r)}(i)\) of a pair \(\{r,\pi(r)\}\) to the numerator of \(g_\nu(i)\) cancel, and \(g_\nu(i)=0\).

\proofstep{Affine binary-code classes.} These classes satisfy the involution hypothesis. Let \(W=\operatorname{span}\{v_j:j\text{ a column of }T\}\). The parameters \(u\) of the concepts consistent with \(T\) form a coset of \(W^\perp\), so if \(v_i\in W\), then \(\langle u,v_i\rangle\) is constant on this coset and \(i\) is forced. Thus if \(i\) is not forced, then \(v_i\notin W=(W^\perp)^\perp\), so there is \(u_0\in W^\perp\) with \(\langle u_0,v_i\rangle=1\), and the translation \(u\mapsto u+u_0\) is the required involution of all rows.

\proofstep{The Lipschitz bound.} Let \(\nu,\nu'\) be realizable and write \(v=\Phi(\nu)\), \(v'=\Phi(\nu')\). Every profile coordinate is an average of \(\pm1\) values, so \(\|v-v'\|_\infty\le\|\nu-\nu'\|_1\). The function \(t\mapsto\operatorname{sgn}(t)|t|^{2p}\) is \(2p\)-Lipschitz on \([-1,1]\), so \(\|A(v)-A(v')\|_1\le2pN\|\nu-\nu'\|_1\). The vote is \(g_\nu=B^T(A(v)/Z(v))\), where \(A(v)/Z(v)\) has \(\ell_1\)-norm one and \(Z(v)=\|A(v)\|_1\ge1\) by \eqref{eq:res-3.2}, since \(v\in X_B\). Since the entries of \(B\) are \(\pm1\), \eqref{eq:normalization} in the \(\ell_1\) norm gives
\[
|g_\nu(i)-g_{\nu'}(i)|
\le\left\|\frac{A(v)}{Z(v)}-\frac{A(v')}{Z(v')}\right\|_1
\le\frac{2\|A(v)-A(v')\|_1}{Z(v)}
\le4pN\|\nu-\nu'\|_1.
\]

\proofstep{Heavy atoms.} A realizable distribution has at most \(M\) atoms, and each atom outside \(T(\nu)\) has mass less than \(\tau\max\nu\le\tau\), so the deleted mass \(m_0\) satisfies \(m_0\le M\tau\). Moreover \(\|\nu-\nu_T\|_1=m_0+(1-m_0)\bigl(\frac1{1-m_0}-1\bigr)=2m_0\le2M\tau\), and \(\nu_T\) is realizable because its support lies in that of \(\nu\). The Lipschitz bound gives \(\|g_\nu-g_{\nu_T}\|_\infty\le8pNM\tau=b/2\). If \(i\) is not forced by \(T(\nu)\), the cancellation applied to \(\nu_T\) gives \(g_{\nu_T}(i)=0\), hence \(|g_\nu(i)|\le b/2\).

\proofstep{The tilt dimension.} Let \(L=h_+^*(B)\) and put
\[
c=1-\frac1{2L},
\qquad
\theta=\frac{\ln c}{\ln\tau}>0,
\qquad
P_\theta(\nu)\propto\nu^\theta ,
\]
so that \(\tau^\theta=c\); here \(\tau<1\) and \(1/2\le c<1\). The map \(P_\theta\) is continuous on \(\Delta_C\) and preserves supports. We use the mass-prefix barycentric map of the proof of Theorem~\ref{res:6.3}, with target the order complex \(\Lambda\) of the poset \(P_+^*(B)\) of nonempty positive closures, which has dimension \(L-1\). For \(\mu\in\Delta_C\) and \(s>0\) let \(F_s(\mu)=\{z:\mu_z\ge s\}\), and for \(R\in P_+^*(B)\) put
\[
\beta_R(\mu)=\int_0^\infty|F_s(\mu)|\,\mathbf 1\bigl\{F_s(\mu)\ne\varnothing,\ c_+(F_s(\mu))=R\bigr\}\,ds.
\]
If \(\mu_1\ge\mu_2\ge\cdots\ge\mu_l>\mu_{l+1}=0\) are the sorted masses and \(F_j\) the \(j\)-th prefix, then \(F_s(\mu)=F_j\) for \(s\in(\mu_{j+1},\mu_j]\), so \(\beta_R(\mu)=\sum_{j:c_+(F_j)=R}t_j\) with \(t_j=j(\mu_j-\mu_{j+1})\), the coefficients of that proof. The closures of the prefixes form a chain, of at most \(L\) members, and \(\sum_R\beta_R(\mu)=\int_0^\infty|F_s(\mu)|\,ds=\sum_z\mu_z=1\), so \(\beta(\mu)\in|\Lambda|\). The map \(\beta\) is continuous. Indeed, if \(\|\mu-\mu'\|_\infty\le\eta\), then \(F_s(\mu)=F_s(\mu')\) unless \(s\) lies within \(\eta\) of some \(\mu_z\), a set of measure at most \(4M\eta\) since there are \(2M\) atoms, and \(|F_s|\le M\) on realizable distributions. Thus \(|\beta_R(\mu)-\beta_R(\mu')|\le4M^2\eta\). In particular, ambiguous prefixes at mass ties have zero coefficient and vanishing atoms disappear continuously.

Put \(F_0=\beta\circ P_\theta:\Delta_C\to|\Lambda|\). Label every closed set \(R\) by a fixed compatible concept \(h_R\in V(R)\), so that the label depends on \(R\) alone. Let \(F_0(\nu)\in D_R\), and suppose that \(h_R\) is wrong at some \(z\in T(\nu)\). Put \(\mu=P_\theta(\nu)\), with sorted masses \(\mu_1\ge\mu_2\ge\cdots\), and let \(z\) be the \(j\)-th atom. Then \(\mu_j/\mu_1=(\nu_z/\max\nu)^\theta\ge\tau^\theta=c\). A prefix \(F_i\) with \(c_+(F_i)=R\) cannot contain \(z\), because \(h_R\) agrees with all atoms of \(R\). So the coordinate of \(R\) is at most
\[
\sum_{i<j}t_i=\sum_{i<j}(\mu_i-\mu_j)
=\sum_{i<j}\mu_i\Bigl(1-\frac{\mu_j}{\mu_i}\Bigr)
\le\Bigl(1-\frac{\mu_j}{\mu_1}\Bigr)\sum_{i<j}\mu_i
\le1-c=\frac1{2L}.
\]
On \(D_R\), however, the coordinate of \(R\) is at least \(1/L\), since it is the largest of at most \(L\) nonzero coordinates with sum one. This contradiction shows that on \(F_0^{-1}(D_R)\) the concept \(h_R\) agrees with all heavy atoms, that is, with all atoms of \(T(\nu)\).

Let \(c^*\) be a concept realizing \(\nu\); it is also consistent with \(T(\nu)\). If \(h_R\) errs at an atom \((i,y)\) of positive \(\nu\)-mass, then \(c^*(i)=y\ne h_R(i)\), so the two concepts consistent with \(T(\nu)\) differ at \(i\), and the column \(i\) is not forced by \(T(\nu)\). Thus \(|g_\nu(i)|\le b/2\): the low-vote condition holds with margin \(b/2\). Since \(\dim\Lambda\le L-1\), this proves \(k_{\mathrm{tilt}}(B)\le L-1\).

\proofstep{The Hadamard class.} The matrix \(H_{2^m}\) is an affine binary-code class, with \(c_0\equiv+1\) and \(v_i=i\), and \(h_+^*=m+1\) by the proof of Theorem~\ref{res:5.5}. So \(k_{\mathrm{tilt}}\le m\). Lemma~\ref{res:7.11} with \(\operatorname{lr}=m+1\) (Theorem~\ref{res:5.5}) gives equality.
\end{proof}

\paragraph{Closed stars.} Assume the involution hypothesis, and keep the notation \(L=h_+^*(B)\), \(F_0\), and \(h_R\) of the proof of Theorem~\ref{res:7.14}. The definition of \(k_{\mathrm{tilt}}\) asks for the low-vote condition only on the regions \(D_v\), and the construction gives more: the same labels also work on closed vertex stars of a finite complex, which contain these regions by Lemma~\ref{res:2.0}. Let \(d_*>0\) be such that the closed \(d_*\)-neighborhoods (in \(\ell_1\)) of any \(L+1\) members of the finite closed cover \(\{F_0^{-1}(D_R)\}_R\) with empty common intersection still have empty common intersection (such a \(d_*\) exists by finiteness and compactness, as in the proof of Lemma~\ref{res:3.3}), and choose
\[
\rho<
\min\{\tau/(2M),b/(16pN),d_*\}.
\]
Let \(\nu\in F_0^{-1}(D_R)\), and let \(\nu'\in\Delta_C\) satisfy \(\|\nu'-\nu\|_1<\rho\). Every heavy atom of \(\nu\) has mass at least \(\tau\max\nu\ge\tau/M\), so it has positive \(\nu'\)-mass. A concept realizing \(\nu'\) is therefore consistent with \(T(\nu)\), and so is \(h_R\). If \(h_R\) errs at an atom \((i,y)\) of positive \(\nu'\)-mass, then \(i\) is unforced by \(T(\nu)\), as in that proof, and parts (i) and (iii) of the theorem give
\[
|g_{\nu'}(i)|
\le
b/2+4pN\rho
<
3b/4 .
\]
This covers atoms of zero \(\nu\)-mass that appear in \(\nu'\). Thickening the finite closed cover by \(\rho<d_*\) preserves its multiplicity \(L\). As in the proof of Theorem~\ref{res:7.13}(i), take the partition-of-unity map to the nerve of the open \(\rho\)-neighborhoods. By a Lebesgue-number argument for the cover of the nerve by open vertex stars, subdivide the nerve so that every closed vertex star lies in the open star of a nerve vertex, and label each subdivision vertex by the hypothesis \(h_R\) of such a nerve vertex \(R\). If the image of \(\nu'\) lies in the closed star of a subdivision vertex with label \(h_R\), then the partition function of \(R\) is positive at \(\nu'\), so \(\nu'\) lies within \(\rho\) of \(F_0^{-1}(D_R)\), and the last display applies. This gives a finite-complex parameterization of dimension at most \(L-1\) for which the low-vote condition holds on closed stars.

\subsubsection*{Bounded-multiplicity encodings}

To bound \(\dim_\eta\) from above on specific families, we encode the points of \(P_B\) by continuous parameters of low dimension from which the point can be recovered up to a small error. The constructions apply to all of \(P_B\), hence to \(V_B\), by restriction of covers. They rest on the following encoding principle.

\begin{lemma}[Encoding principle]\label{lem:encoding-principle}
Let \(\eta>0\), let \(K\) be a finite polyhedral complex, and let \(F:P_B\to K\) and \(W:K\to\mathbb R^M\) be continuous, with \(\dim K\le d\) and \(\|x-W(F(x))\|_\infty\le e<\eta/2\) on \(P_B\). Then \(\dim_\eta(P_B)\le d\).
\end{lemma}

\begin{proof}
Triangulate \(K\) so finely that \(W\) oscillates by less than \(\eta-2e\) on every closed vertex star, and pull back the regions \(D_v\) of Lemma~\ref{res:2.0} for this triangulation. The sets \(F^{-1}(D_v)\) are closed, cover \(P_B\), and have multiplicity at most \(d+1\). Since \(D_v\subseteq\operatorname{St}(v)\), two points \(x,x'\) of the same set satisfy
\[
\|x-x'\|_\infty\le2e+\|W(F(x))-W(F(x'))\|_\infty<\eta,
\]
and by compactness the diameter of the set is less than \(\eta\). So \(\dim_\eta(P_B)\le d\).
\end{proof}

The parameterizations use row weights. The linear map \(w\mapsto B^Tw\) from the unit \(\ell_1\)-ball onto \(P_B\) has a continuous piecewise-affine section: choose preimages in the ball of the vertices of a triangulation of \(P_B\) and extend affinely on each simplex. Convexity keeps the section in the unit ball, and linearity of \(B^T\) makes it a section. So a continuous choice \(w=w(x)\) with \(B^Tw=x\) and \(\|w\|_1\le1\) is available.

A set \(\{x\in\mathbb R^D:\|x\|_1\le\rho,\ |\operatorname{supp}x|\le\ell\}\), a \emph{sparse \(\ell_1\)-ball}, is a finite union of \(\ell_1\)-balls of coordinate subspaces of dimension at most \(\ell\). It is a finite polyhedral complex of dimension at most \(\ell\), namely the cone over a skeleton of a cross-polytope. Below we write \(S_a(t)=\operatorname{sgn}(t)(|t|-a)_+\) for soft thresholding, applied coordinatewise. It is continuous and \(|S_a(t)-t|\le a\).

The next proposition gives three structural conditions that bound the covering dimension at a fixed scale. In parts (a) and (b) the bound does not depend on the size of the matrix, and in part (c) it depends only on the side of the rectangles that are required to be nearly unbiased.

\begin{proposition}[Bounded-multiplicity encodings]\label{prop:encodings}
Let \(B\) be an \(N\times M\) sign matrix and let \(\eta>0\).

(a) Interval columns. If the positive row set of every column is an interval of the ordered rows, then
\[
\dim_\eta(V_B)\le4\lceil16/\eta\rceil .
\]

(b) Bounded error intersections. If, after negating some columns, the positive supports \(D_r\) of the rows satisfy \(|D_r\cap D_s|\le s_0\) for \(r\ne s\), then
\[
\dim_\eta(V_B)
\le
1+\lceil128/\eta^2\rceil
+\lceil128s_0/\eta^2\rceil .
\]

(c) Unbiased rectangles. Let \(t=\eta/4\) and let \(q\ge1\) be an integer. If every \(q\times q\) submatrix, with row set \(I\) and column set \(J\), satisfies, for all row signs \(\tau\) and column signs \(\sigma\),
\[
\sum_{r\in I,\ i\in J}\tau_r\sigma_iB_{ri}\le tq^2 ,
\]
then \(\dim_\eta(P_B)\le2q\).
\end{proposition}

\begin{proof}
In each part we construct a parameter map and a decoder for Lemma~\ref{lem:encoding-principle}, or a direct cover.

\proofstep{Part (a).} For \(w=w(x)\) put \(A_j=\sum_{r\le j}(w_r)_+\), \(B_j=\sum_{r\le j}(w_r)_-\) and \(C_j=A_j-B_j\), for \(0\le j\le n\), where \(n\) is the number of rows. The coordinate of \(x=B^Tw\) at the column with interval \([a,b]\), for integers \(a\le b\), is
\[
\sum_{r\in[a,b]}w_r-\sum_{r\notin[a,b]}w_r=2(C_b-C_{a-1})-C_n.
\]
Let \(q=\lceil16/\eta\rceil\). Interpolate \(A\) and \(B\) piecewise linearly on \([0,n]\), and add the strictly increasing term \(\xi/n\) in the variable \(\xi\in[0,n]\): let \(H_A(\xi)=A(\xi)+\xi/n\), and define \(H_B\) in the same way from \(B\). Then \(H_A(0)=0\), \(H_A(n)=A_n+1\le2\), and on each segment the slope of \(H_A\) is \((w_j)_++1/n\in[1/n,2]\). Define the truncated inverse by \(I_A(h)=H_A^{-1}(h)\) for \(h\le H_A(n)\) and \(I_A(h)=n\) for \(h>H_A(n)\), and define \(I_B\) in the same way. Recording \(I_A(j/q)\) and \(I_B(j/q)\), \(j=1,\ldots,2q\), defines a map \(F:P_B\to[0,n]^{4q}\). It is continuous: if \(\|H-\widetilde H\|_\infty\le\varepsilon'\) for two such functions, the slope bound \(\ge1/n\) gives \(|I(h)-\widetilde I(h)|\le n\varepsilon'\) at every height, including heights where a truncation occurs, and \(H_A,H_B\) depend continuously on \(x\).

Let \(x,\widetilde x\) be two points whose parameters differ by at most \(\delta'\) in every coordinate. We claim that \(\|H_A-\widetilde H_A\|_\infty\le1/q+2\delta'\). If not, after possibly exchanging the roles of \(x\) and \(\widetilde x\), there is \(\xi\) with \(H_A(\xi)-\widetilde H_A(\xi)>1/q+2\delta'\). So there is a grid height \(h=j/q\) with \(\widetilde H_A(\xi)<h<H_A(\xi)-2\delta'\), and \(1\le j\le2q\) because the outer terms lie in \([0,2]\). The slope bound \(\le2\) gives \(H_A(\xi-\delta')>h\) (and \(\xi>\delta'\), since \(H_A(\xi)\le2\xi\)), so \(I_A(h)<\xi-\delta'\). On the other hand \(\widetilde H_A(\xi)<h\) gives \(\widetilde I_A(h)\ge\xi\). So the two recorded coordinates differ by more than \(\delta'\), a contradiction. The same holds for \(H_B\). Evaluating at integers, \(|A_j-\widetilde A_j|,|B_j-\widetilde B_j|\le1/q+2\delta'\), hence \(|C_j-\widetilde C_j|\le2/q+4\delta'\), and the coordinate formula gives
\[
\|x-\widetilde x\|_\infty\le5(2/q+4\delta')=\frac{10}q+20\delta'.
\]
On the right, \(10/q\le5\eta/8\). Triangulate \([0,n]^{4q}\) so finely that every closed vertex star has \(\ell_\infty\)-diameter at most \(\delta'\), where \(20\delta'<3\eta/8\). Pulling back along \(F\) the regions \(D_v\) of Lemma~\ref{res:2.0} gives a closed cover of \(P_B\) of mesh less than \(\eta\) and multiplicity at most \(4q+1\), which proves (a).

\proofstep{Part (b).} Negating columns acts on \(P_B\) and \(V_B\) by coordinate sign changes, which are \(\ell_\infty\)-isometries. So we may switch the background to all negative. Let \(D_r\) be the positive support of the row \(r\), with \(|D_r\cap D_s|\le s_0\) for \(r\ne s\). For \(g=B^Tw\) we have \(g_i=-W+2\sum_{r:i\in D_r}w_r\) with \(W=\sum_rw_r\). Put \(t=\eta/8\) and \(\delta=t^2/2=\eta^2/128\). Soft-threshold the row weights, \(w'_r=S_\delta(w_r)\) and \(r_r=w_r-w'_r\). Since \(\|w\|_1\le1\), the vector \(w'\) has fewer than \(1/\delta\) nonzero coordinates, \(|r_r|\le\delta\) and \(\sum_r|r_r|\le1\). With \(v_i=\sum_{r:i\in D_r}r_r\), \(\alpha_i=\sum_{r:i\in D_r}|r_r|\le1\) and \(H=\{i:\alpha_i>t\}\), expanding the square gives
\[
\sum_{i\in H}\alpha_i^2
=\sum_r|r_r|^2|D_r\cap H|+\sum_{r\ne s}|r_r||r_s||D_r\cap D_s\cap H|
\le\delta\sum_{i\in H}\alpha_i+s_0\Bigl(\sum_r|r_r|\Bigr)^2
\le\delta|H|+s_0 .
\]
As \(\alpha_i^2>t^2\) on \(H\), this yields \(|H|\le s_0/(t^2-\delta)=128s_0/\eta^2\).

Let \(v'=S_t(v)\). Since \(|v_i|\le\alpha_i\), it is supported in \(H\), and \(|v'_i|\le1\), so \(\|v'\|_1\le|H|\le128s_0/\eta^2\). The parameter \(F(g)=(W,w',v')\) and the decoder \(\widetilde g_i=-W+2\sum_{r:i\in D_r}w'_r+2v'_i\) are continuous, and \(g_i-\widetilde g_i=2(v_i-v'_i)\) has absolute value at most \(2t=\eta/4\). The parameter lies in the product of \([-1,1]\), the sparse \(\ell_1\)-ball of radius \(1\) with sparsity at most \(\lceil128/\eta^2\rceil\), and the sparse \(\ell_1\)-ball of radius \(128s_0/\eta^2+1\) with sparsity at most \(\lceil128s_0/\eta^2\rceil\). This is a finite polyhedral complex of dimension at most \(1+\lceil128/\eta^2\rceil+\lceil128s_0/\eta^2\rceil\), and Lemma~\ref{lem:encoding-principle} with \(e=\eta/4\) proves (b).

\proofstep{Part (c).} If \(q>\min(N,M)\), the hypothesis is void, and the bound holds because \(P_B\) is a polytope of dimension at most \(\operatorname{rank}B\le\min(N,M)<q\): the regions \(D_v\) of a sufficiently fine triangulation of \(P_B\) form a closed cover of mesh less than \(\eta\) and multiplicity at most \(\min(N,M)+1\). So let \(q\le\min(N,M)\). Soft-threshold the row weights at \(1/q\): put \(w'=S_{1/q}(w)\) and \(r=w-w'\). Then \(w'\) has fewer than \(q\) nonzero coordinates, and the residual \(r\) has \(\|r\|_1\le1\) and \(\|r\|_\infty\le1/q\).

Suppose \(B^Tr\) had at least \(q\) coordinates of absolute value greater than \(t\). Take a set \(J\) of \(q\) such coordinates, let \(\sigma\) be their signs, and put \(\Xi_u=\sum_{i\in J}\sigma_iB_{ui}\) for a row \(u\). Then
\[
tq<\sum_{i\in J}\sigma_i(B^Tr)_i=\sum_ur_u\Xi_u\le\sum_u|r_u||\Xi_u|.
\]
Under the constraints \(\sum_u|r_u|\le1\) and \(|r_u|\le1/q\), the last sum is at most the average of the \(q\) largest values of \(|\Xi_u|\). Let \(I\) be the corresponding set of \(q\) rows and \(\tau_u=\operatorname{sgn}\Xi_u\). Then \(\sum_{u\in I,i\in J}\tau_u\sigma_iB_{ui}=\sum_{u\in I}|\Xi_u|>tq^2\), which violates the hypothesis. Hence the residual vote \(B^Tr\) has fewer than \(q\) coordinates above \(t\) in absolute value, and \(y'=S_t(B^Tr)\) has fewer than \(q\) nonzero coordinates, all at most \(1\) in absolute value. The parameter \((w',y')\) lies in a product of two sparse \(\ell_1\)-balls, of total dimension at most \(2q\), and the decoder \((w',y')\mapsto B^Tw'+y'\) has error \(\|B^Tr-y'\|_\infty\le t=\eta/4\). Lemma~\ref{lem:encoding-principle} applies.
\end{proof}

Part (b) gives fixed-scale constant dimension for single-flip classes (\(s_0=0\)) and for \(\operatorname{PG}(2,q)\) (\(s_0=1\)), and \(O_\eta(s_0+1)\) under the hypothesis of Theorem~\ref{res:5.2}.

\subsubsection*{Values on explicit families}

We can now collect the values of the two parameters on the families treated in this paper. The dual-interval matrix \(D_n^{\mathrm{int}}\) has rows indexed by the points \(r\in[n]\), columns indexed by the nonempty intervals \([a,b]\subseteq[n]\), and entry \(+\) exactly when \(a\le r\le b\); it is the transpose of the interval class. The recursive Hanner family is that of Theorem~\ref{res:4.3}(i).

\begin{proposition}[Explicit values]\label{prop:tilt-values}
The parameters \(k_{\mathrm{tilt}}\) and \(\dim_{\eta_0}\) take the following values and orders. The Hanner entry is a common interval, not an equality of the two parameters.

\begin{center}\small\upshape\begin{tabular}{ll}
\toprule
Family & \(k_{\mathrm{tilt}}\) and \(\dim_{\eta_0}\) \\
\midrule
Full cube, \(d\ge1\) & both \(d-1\) \\
Anchored cube, \(d\ge0,\ a\ge1\) & both \(d\) \\
Random \(n\times n\) & both \(\Theta(\log n)\) with high probability \\
Thresholds and dual intervals (\(n\ge2\)), \(\operatorname{PG}(2,q)\) & \(\Theta(1)\) \\
Recursive Hanner family, \(d=4^s\) & each in \([\lceil d/2\rceil-1,d-1]\) \\
\bottomrule\end{tabular}\end{center}
\end{proposition}

\begin{proof}
We treat the families of the table in turn.

\proofstep{Cubes.} The full cube is the affine binary-code class with \(c_0\equiv+1\) and \(v_i=e_i\) on the \(d\) coordinates, and the anchored cube is the one with the additional \(a\) anchor columns, for which \(v_i=0\). In the full cube the nonempty positive closures are the subcubes with at least one fixed coordinate, so a chain has at most \(d\) members and \(h_+^*=d\). In the anchored cube they are all subcubes together with the anchor atoms, so \(h_+^*=d+1\). Theorem~\ref{res:7.14} therefore gives \(k_{\mathrm{tilt}}\le d-1\) and \(\le d\), respectively, and Theorem~\ref{res:5.1} with Lemma~\ref{res:7.11} gives the matching lower bounds. The real ranks of these matrices are \(d\) and \(d+1\), so Lemma~\ref{res:7.12} gives \(\dim_{\eta_0}\le d-1\) and \(\le d\), while Theorem~\ref{res:7.13}(i) with \(\eta_0\le b\) gives \(\dim_{\eta_0}\ge k_{\mathrm{tilt}}\).

\proofstep{Hanner matrices.} Let \(d=4^s\) and let \(A\) be the matrix of Theorem~\ref{res:4.3}(i) with \(\operatorname{rank}A=d\) and \(\operatorname{Ind}=d-1\). Theorem~\ref{res:a} gives \(2\operatorname{lr}(A)\ge\operatorname{Ind}+1=d\), so \(\operatorname{lr}(A)\ge\lceil d/2\rceil\). Lemma~\ref{res:7.11}, Theorem~\ref{res:7.13}(i) and Lemma~\ref{res:7.12} give
\[
\lceil d/2\rceil-1\le\operatorname{lr}(A)-1\le k_{\mathrm{tilt}}(A)\le\dim_{\eta_0}(V_A)\le\operatorname{rank}A-1=d-1.
\]

\proofstep{Thresholds, dual intervals, projective planes.} The upper bounds follow from Proposition~\ref{prop:encodings}. Threshold matrices have interval columns, since the positive rows of the column \(i\) of \(T_n\) are \([1,i]\). Dual-interval matrices have interval columns as well, by their definition. So part (a) gives \(\dim_{\eta_0}\le4\lceil16/\eta_0\rceil=484\). The matrix \(\operatorname{PG}(2,q)\) has error intersections at most \(s_0=1\) for the all-negative background, since two lines meet in one point. So part (b) gives \(\dim_{\eta_0}\le1+2\lceil128/\eta_0^2\rceil=14515\). These upper bounds also bound $k_{\mathrm{tilt}}$ by Theorem~\ref{res:7.13}(i).

For the lower bounds, thresholds with $n\ge2$ have list number two by Theorem~\ref{res:6.4} and the argument following it. Also for $n\ge2$, the dual-interval matrix restricted to columns $[1,1]$ and $[1,n]$ has the row patterns \((+,+)\) (row \(1\)) and \((-,+)\) (the other rows), which form the anchored one-bit cube. Restriction does not increase list replicability, so Theorem~\ref{res:5.1} gives list number at least two. Projective planes have list number three by Theorem~\ref{res:5.6}. Lemma~\ref{res:7.11} therefore gives positive lower constants for \(k_{\mathrm{tilt}}\) in these three cases, and Theorem~\ref{res:7.13}(i) gives them for \(\dim_{\eta_0}\).

\proofstep{Random matrices.} For the lower bound, fix
$d=\lfloor\log_2(n/(2\ln n))\rfloor$ columns, for sufficiently large $n$.
Each prescribed pattern on these columns is absent from all $n$ rows
with probability $(1-2^{-d})^n\le e^{-n/2^d}\le n^{-2}$, because \(n/2^d\ge2\ln n\).
A union bound over at most $2^d\le n$ patterns shows that these columns
are shattered with probability at least $1-1/n$. Thus
$\operatorname{lr}\ge\operatorname{vc}\ge d$ by Section~\ref{ssec:2},
and Lemma~\ref{res:7.11} and Theorem~\ref{res:7.13}(i) give both
required logarithmic lower bounds, since \(d=\log_2n-O(\log\log n)\).

For the upper bound, consider an \(n\times n\) matrix with independent uniform signs and, with \(t=\eta/4\),
\[
q=\lceil16t^{-2}\ln(2n)\rceil.
\]
Suppose first that \(q\le n\). For fixed \((I,J,\tau,\sigma)\), the sum in the hypothesis of Proposition~\ref{prop:encodings}(c) is a sum of \(q^2\) independent uniform signs, so Hoeffding's inequality bounds the probability that it exceeds \(tq^2\) by \(e^{-t^2q^2/2}\). There are at most \((2n)^{2q}\) choices of \((I,J,\tau,\sigma)\), and \(t^2q/2\ge8\ln(2n)\). So the failure probability of the hypothesis is at most \((2n)^{2q}e^{-t^2q^2/2}\le(2n)^{-6q}\). For \(q>n\) the rank bound of Lemma~\ref{res:7.12} gives \(\dim_{\eta_0}(V_B)\le n-1<q\). Taking \(\eta=\eta_0\), so that \(q=O(\log n)\), we get \(\dim_{\eta_0}(V_B)=O(\log n)\) with high probability, and with the lower bound \(\dim_{\eta_0}(V_B)=\Theta(\log n)\) and \(k_{\mathrm{tilt}}(B)=\Theta(\log n)\) with high probability.
\end{proof}

\subsubsection*{Covering entropy of the vote image}

The vote \(f_w\) is a Lipschitz function of \(w\) by \eqref{eq:res-3.4}, so a net of the power-map image gives a net of the vote image. On Hadamard matrices the vote image is smaller than this suggests: its covering entropy is exactly of quadratic order, while Theorem~\ref{res:7.1} shows that the power-map image has at least cubic order.

\begin{resultenv}{Theorem}{res:7.15}{Hadamard vote entropy}
For \(0<\rho<1/2\), using the $\ell_\infty$ metric,
\[
\log_2\mathcal N(V_{H_{2^m}},\rho)
=
\Theta_\rho(m^2).
\]
\end{resultenv}

\begin{proof}
Write \(H_{r,i}=(-1)^{\langle r,i\rangle}\) with \(r,i\in\mathbb F_2^m\).

\proofstep{The lower bound.} For a linear subspace \(W\le\mathbb F_2^m\), the uniform distribution with positive labels on the columns in \(W\) is realized by the row \(0\). Its profile is \(\Phi_r=|W|^{-1}\sum_{i\in W}(-1)^{\langle r,i\rangle}=\mathbf 1_{W^\perp}(r)\), so for every \(p\) the potential weights are uniform on \(W^\perp\), and the vote is
\[
g_W(i)=|W^\perp|^{-1}\sum_{r\in W^\perp}(-1)^{\langle r,i\rangle}=\mathbf 1_W(i).
\]
Distinct subspaces give votes at \(\ell_\infty\)-distance exactly one. With \(d=\lfloor m/2\rfloor\), the Gaussian binomial coefficient counts at least
\[
2^{d(m-d)}
\]
subspaces of dimension \(d\), as in the proof of Proposition~\ref{res:7.2}(ii), and a ball of radius \(\rho<1/2\) contains at most one of these votes, because two of its points are at distance at most \(2\rho<1\). This gives the lower bound of quadratic order in \(m\).

\proofstep{The upper bound.} Every vote \(g_\nu=\sum_r(A_r/Z)B_{r,\cdot}\) is a convex combination of signed rows, with probabilities \(|A_r|/Z\) on the signed rows \(\operatorname{sgn}(A_r)B_{r,\cdot}\). Average \(s\) signed rows drawn independently with these probabilities. Each coordinate is an average of \(s\) independent variables with values in \([-1,1]\) and mean \(g_\nu(i)\), so Hoeffding's inequality and a union bound over the \(M\) coordinates give \(\mathbb P(\|\widehat g-g_\nu\|_\infty>\rho/2)\le2Me^{-s\rho^2/8}\), which is at most \(1/2\) for \(s=\lceil8\rho^{-2}\ln(4M)\rceil\). Hence these averages give an external $\rho/2$-net of the vote image of size at most $(2N)^s$. For each ball of this net that meets the image, choose a vote in the intersection. By the triangle inequality the chosen votes form an internal $\rho$-net, with no increase in cardinality. On \(H_{2^m}\) we have \(N=M=2^m\), and this gives \(\log_2\mathcal N\le s(m+1)=O_\rho(m^2)\).
\end{proof}

The passage from the power-map image to the vote image thus reduces the covering entropy at a fixed scale. This reduction concerns the number of balls and not the multiplicity of a cover. The metric estimates of this subsection leave
\[
\dim_{\eta_0}(V_H)\in[m,O(m^3)]
\]
for \(H=H_{2^m}\): the lower bound is Theorem~\ref{res:7.14} with Theorem~\ref{res:7.13}(i), and the upper bound is Theorem~\ref{res:7.13}(ii), since \(\kappa(2^m,2^m)=O(m^3)\). The sign-compatible dimension, which only preserves confident signs, is computed exactly in Theorem~\ref{thm:walsh-sign-geometry} of Section~\ref{app:vote-geometry}.

\section{Sign-compatible geometry of vote images}\label{app:vote-geometry}
The entropy-tilting construction of Section~\ref{ssec:3} asks little of a cover: the cover lives on the vote image rather than on the whole signed hull \(P_B\), the convex hull of the signed concepts, and its labels only have to respect the signs on which the vote is confident. This section measures the quantitative effect of these two restrictions. We introduce the sign-compatible covering dimension, which lies between the tilt dimension of Section~\ref{ssec:7.8} and metric covering dimension. For a Sylvester Hadamard matrix of order $2^\ell$, the vote image has sign-compatible dimension exactly $\ell$, the value forced by list replicability, while the full signed hull has dimension at least $\ell+\lfloor\ell/6\rfloor+1$ at the same confidence threshold. We also give lower bounds for two direct compressions of the hull, by linear measurements and by two successive thresholds, and we bound the dimension that a coordinate cube argument can detect. These statements concern the vote representation itself; the proof of Theorem~\ref{thm:main} does not use it.

\subsection{Sign-compatible covers}
We first isolate the property of a closed cover that the low-vote condition of Section~\ref{ssec:7.8} uses, and we compare the resulting dimension with the tilt dimension and with metric covering dimension.

\begin{definition}[Sign-compatible covers]
Fix $0<b<1$ and a nonempty compact set $X\subseteq[-1,1]^n$.
A set $F\subseteq X$ is \emph{$b$-sign-compatible} if it contains
no pair $u,v$ with $u_i>b$ and $v_i<-b$ for the same coordinate.
Let $d_b^{\rm sign}(X)$ be the minimum multiplicity minus one
of a finite closed cover of $X$ by such sets.
\end{definition}

Such covers exist, since a set of $\ell_\infty$-diameter less than $2b$ is $b$-sign-compatible.
A set $F$ is $b$-sign-compatible if and only if there is $h\in\{\pm1\}^n$ with
$h_i u_i\ge-b$ for every $u\in F$ and every $i$. Indeed, if some $u\in F$ has
$u_i>b$, take $h_i=+1$, and otherwise take $h_i=-1$; conversely,
$h_i=+1$ excludes $v_i<-b$ and $h_i=-1$ excludes $u_i>b$.
So a sign-compatible set is exactly a set that admits one Boolean label respecting all its signs of confidence above $b$.
In this section $b=45/64$
unless another value is specified, and we retain the definitions of
$P_B,V_B,k_{\rm tilt}$ and $\dim_\eta$ from
Section~\ref{ssec:7.8}.

The definition of the tilt dimension requires a map to a simplicial complex, so a cover has to be converted into such a map. A partition of unity subordinate to a thickened cover only places the regions of Lemma~\ref{res:2.0} in neighborhoods of the cover members, and sign compatibility does not pass to neighborhoods. The next lemma gives a map for which these regions pull back to the members themselves.

\begin{lemma}\label{lem:exact-cover-map}
If $F_1,\ldots,F_s$ is a finite closed cover of a compact metric
space $X$ of multiplicity $L$, there are a finite complex $K$ of
dimension at most $L-1$ and a continuous map $T:X\to|K|$ such that,
for every nonempty member $F_j$,
the region $D_j$ of Lemma~\ref{res:2.0} on which the barycentric coordinate of the vertex $j$ is maximal pulls back to exactly
$F_j$.
\end{lemma}
\begin{proof}
Discard empty members. For every subfamily $J$ of size $L+1$,
$\bigcap_{j\in J}F_j=\varnothing$. The continuous function
$x\mapsto\max_{j\in J}\operatorname{dist}(x,F_j)$ is therefore positive
on the compact space $X$, and it has a
positive minimum. Choose $\varepsilon>0$ smaller than all these
finitely many minima; when $s\le L$ there is no restriction. The open sets
$U_j=\{x:\operatorname{dist}(x,F_j)<\varepsilon\}$ have
multiplicity at most $L$, since a point in $L+1$ of them would give
$\max_{j\in J}\operatorname{dist}(x,F_j)<\varepsilon$ for the corresponding $J$.
Their nerve $K$ has dimension at most
$L-1$. Set
\[
 w_j(x)=(\varepsilon-\operatorname{dist}(x,F_j))_+,
 \qquad T_j(x)=\frac{w_j(x)}{\sum_i w_i(x)}.
\]
Since the $F_j$ cover $X$, at every $x$ some $w_j(x)=\varepsilon$, so the denominator is
positive and $T$ is continuous. The positive coordinates of $T(x)$ are
the indices $j$ with $x\in U_j$, which form a face of the nerve.
Moreover $w_j\le\varepsilon$ everywhere, so $T_j(x)$ is maximal exactly when $w_j(x)=\varepsilon$,
that is, when $\operatorname{dist}(x,F_j)=0$. As $F_j$ is closed,
this is equivalent to $x\in F_j$.
\end{proof}

The next proposition places the sign-compatible dimension in the chain of invariants of Section~\ref{ssec:7.8}. A sign-compatible cover of the vote image yields a parameterization satisfying the low-vote condition, and the dimension itself lies between the metric covering dimensions at the scales $2b$ and $1+b+\delta$.

\begin{proposition}\label{prop:sign-cover}
For every sign matrix $B$,
\[
 \operatorname{lr}(B)-1\le k_{\rm tilt}(B)
 \le d_b^{\rm sign}(V_B)\le d_b^{\rm sign}(P_B)
 \le\dim_{2b}(P_B).
\]
For any compact $X\subseteq[-1,1]^n$ and $\delta>0$,
\[
 \dim_{1+b+\delta}(X)\le d_b^{\rm sign}(X)
 \le\dim_{2b}(X).
\]
\end{proposition}
\begin{proof}
The first inequality is Lemma~\ref{res:7.11}.

\proofstep{From a cover of the vote image to the tilt dimension.}
Let $d=d_b^{\rm sign}(V_B)$ and take a finite closed
cover $F_1,\ldots,F_s$ of $V_B$ by $b$-sign-compatible sets, of
multiplicity at most $d+1$. Apply
Lemma~\ref{lem:exact-cover-map} with $X=V_B$ and $L=d+1$, which gives $K$ and
$T$, and label each vertex $j$ by a compatible Boolean vector $h_j$,
so that $h_j(i)u_i\ge-b$ for $u\in F_j$. Put $F(\nu)=T(g_\nu)$, a continuous
map on $\Delta_C$. If $F(\nu)\in D_j$, then $g_\nu\in F_j$. If
moreover $h_j(i)\ne y$, then $h_j(i)=-y$, so $-yg_\nu(i)\ge-b$, that is,
$yg_\nu(i)\le b$. This is the low-vote condition, and it holds even at atoms
of zero mass. Thus we have a low-vote
parameterization of dimension at most $d$, which proves $k_{\rm tilt}(B)\le d_b^{\rm sign}(V_B)$.

\proofstep{Restriction and small sets.}
Restricting a cover
from $P_B$ to $V_B\subseteq P_B$ proves the next inequality, since a subset of a
$b$-sign-compatible set has the same property. If a set has
$\ell_\infty$-diameter less than $2b$, it contains no pair with
$u_i>b$, $v_i<-b$, because such a pair would be at distance greater than $2b$.
So a cover admissible for $\dim_{2b}$ is $b$-sign-compatible, which proves the rightmost
bounds, for $P_B$ and for every compact $X$.

\proofstep{The diameter of a sign-compatible set.}
On a sign-compatible set each coordinate has
range at most $1+b$: either its maximum is at
most $b$, so its values lie in $[-1,b]$, or its minimum is at least $-b$, so they lie
in $[-b,1]$. Hence such a set has $\ell_\infty$-diameter at most $1+b$, less than $1+b+\delta$.
This proves $\dim_{1+b+\delta}(X)\le d_b^{\rm sign}(X)$.
\end{proof}

Lower bounds for the sign-compatible dimension come from cubes in coordinate projections: coordinate information can force a large cover even when the ambient convex body has a simple linear description. The next lemma gives the form used in parts (ii) and (iii) of Theorem~\ref{thm:walsh-sign-geometry}.
\begin{lemma}\label{lem:projected-cube}
Suppose a compact convex $P\subseteq[-1,1]^n$ projects onto a set
containing $[-t,t]^d$ in $d$ chosen coordinates, where $t>b$.
Then $d_b^{\rm sign}(P)\ge d$. In particular,
$d_b^{\rm sign}(P_B)\ge\operatorname{vc}(C_B\cup-C_B)$.
\end{lemma}
\begin{proof}
If $d=0$ there is nothing to prove, so let $d\ge1$.

\proofstep{A section over the cube.}
For each $\sigma\in\{\pm1\}^d$, choose a lift $p_\sigma\in P$
of $t\sigma$, and define
\[
 f(z)=\sum_{\sigma}\left(\prod_{j=1}^d
            \frac{1+\sigma_j z_j}{2}\right)p_\sigma,
 \qquad z\in[-1,1]^d.
\]
The coefficients are nonnegative and sum to
$\prod_j\bigl(\frac{1+z_j}2+\frac{1-z_j}2\bigr)=1$, so $f$ is continuous and takes values in the convex set $P$.
Summing first over the coordinates of $\sigma$ other than the $k$-th, we see that the $k$-th chosen
coordinate of $f(z)$ is $t\bigl(\frac{1+z_k}2-\frac{1-z_k}2\bigr)=tz_k$. Thus $f$ projects to $tz$.

\proofstep{Pulling back a cover.}
Let $\{F_j\}$ be a finite closed cover of $P$ by $b$-sign-compatible sets.
The sets $f^{-1}(F_j)$ form a finite closed cover of the cube.
No member meets two opposite faces: points $z,z'$ of $f^{-1}(F_j)$ with $z_k=1$ and
$z'_k=-1$ would give two points of $F_j$ whose $k$-th chosen coordinate
equals $t>b$ and $-t<-b$. By Lemma~\ref{lem:cube-cover-multiplicity} this cover of the cube has
multiplicity at least $d+1$. A point lying in several sets $f^{-1}(F_j)$ has its image in the corresponding sets $F_j$, so $\{F_j\}$ also has multiplicity at least $d+1$, and therefore $d_b^{\rm sign}(P)\ge d$.

\proofstep{Shattered coordinates.}
If a coordinate set of size $d$ is shattered by $C_B\cup-C_B$, the
projection of $P_B$ to it contains all $2^d$ corners of the cube. So it is the
full cube $[-1,1]^d$, and we may take $t=1>b$.
\end{proof}

\subsection{A common construction from affine closures}
The upper bounds in parts (i) and (ii) of Theorem~\ref{thm:walsh-sign-geometry} come from one construction, which we describe here.
Write $H=H_{2^\ell}$ for the Sylvester Hadamard matrix, with rows and columns
indexed by $\mathbb F_2^\ell$ and entries
$H_{a,x}=\chi_a(x)=(-1)^{a\cdot x}$. A signed row is an affine
Boolean function $x\mapsto(-1)^{a\cdot x+c}$.
An affine Boolean function is determined on an affine subspace by its values on any set of points whose affine hull is that subspace. So if the signs of a function $g$ of confidence above a level $b_0$ extend to an affine Boolean function, they determine a partial affine function on their affine hull, and sorting these signs by confidence yields a chain of such partial functions. Chains of partial affine functions have at most $\ell+1$ members, however many atoms are involved.
The next lemma turns these chains into sign-compatible covers.

\begin{lemma}\label{lem:affine-confidence-map}
Let $0\le b_0<b<1$, and let $X$ be a compact collection of functions
$g:\mathbb F_2^\ell\to[-1,1]$. Suppose that, for every $g\in X$,
the signed constraints $\{(x,\operatorname{sign}g_x):|g_x|>b_0\}$
extend to an affine Boolean function. Then
$d_b^{\rm sign}(X)\le \ell+1$. If every $g\in X$ has
$\|g\|_\infty>b_0$, then $d_b^{\rm sign}(X)\le \ell$.
\end{lemma}
\begin{proof}
\proofstep{The poset.} Consider the finite poset of affine Boolean functions restricted
to nonempty affine subspaces of $\mathbb F_2^\ell$, ordered by extension. Strict
extension increases the dimension of the underlying affine
subspace, which lies in $\{0,\ldots,\ell\}$, so a chain has at most $\ell+1$ members
and the order complex has dimension at most $\ell$.
For every poset element $U$ choose one full affine extension $h_U$:
extend the linear part from the direction space to all of $\mathbb F_2^\ell$,
keeping the value at a base point.
Put $L=\ell+1$, and choose $\theta>0$ so small that
\[
 \left(\frac{b-b_0}{1-b_0}\right)^\theta>1-\frac1L.
\]
This is possible since the base lies in $(0,1)$, and it also makes sense at $L=1$, where the right side is zero.

\proofstep{The closures.} For a signed atom $z=(x,y)$ and $g\in X$ set $w_z=w_z(g)=(yg_x-b_0)_+^\theta$.
As $b_0\ge0$, at most one of $(x,+),(x,-)$ has positive weight.
For $W=\max_z w_z>0$, list the distinct positive weights as
$\lambda_1>\cdots>\lambda_s>0$, with $\lambda_{s+1}=0$.
The set $S_j=\{z:w_z\ge\lambda_j\}$ consists of constraints with $|g_x|>b_0$, so by the
hypothesis of the lemma it extends to an affine Boolean function. An affine Boolean function is
determined on the affine hull of a point set by its values there, since affine combinations over $\mathbb F_2$
are odd sums. So $S_j$ has a unique affine closure $U_j$, the restriction of any such extension
to the affine hull of the points of $S_j$. As $S_1\subseteq S_2\subseteq\cdots$,
the $U_j$ form a chain, possibly with repetitions. Define
\[
 B(g)=\sum_{j=1}^s\frac{\lambda_j-\lambda_{j+1}}{W}e_{U_j},
\]
adding coefficients when closures coincide. The coefficients are nonnegative and sum
to $\lambda_1/W=1$, so $B(g)$ lies in the order complex.

\proofstep{Continuity where $W>0$.} The coefficient of a poset element $U$ in $WB(g)$ is
\[
 N_U(g)=\int_0^{W}\mathbf 1\bigl\{\operatorname{cl}\{z:w_z\ge\xi\}=U\bigr\}\,d\xi,
\]
where $\operatorname{cl}$ denotes the affine closure, because $\{z:w_z\ge\xi\}=S_j$ for $\xi\in(\lambda_{j+1},\lambda_j]$.
This formula also covers tied weights. If all weights change by at most $\eta$, the set
$\{z:w_z\ge\xi\}$ can change only for $\xi$ within $\eta$ of some $w_z$, and this
includes the $\xi$ between the old and new values of $W$. These $\xi$ form a set of measure at most $2\eta$ times the finite number of atoms.
Hence $N_U$ is continuous in the weights, and so in $g$. Normalization by $W$ is continuous
where $W>0$.

\proofstep{The labels.} Let $W(g)>0$ and $yg_x>b$, and put $z=(x,y)$. We call such a $z$ a $b$-confident sign of $g$. Since $|g_{x'}|\le1$ for all $x'$, we have $W\le(1-b_0)^\theta$,
so $w_z/W>((b-b_0)/(1-b_0))^\theta>1-1/L$.
All levels $\lambda_j\le w_z$ have $z\in S_j$, so their closures contain $z$. If the
chosen label $h_U$ is wrong at $z$, then $U$ does not contain $z$, so $U=U_j$ only for
levels $\lambda_j>w_z$. Thus the total barycentric mass of vertices whose chosen label
is wrong at $z$ is at most $(\lambda_1-w_z)/W=1-w_z/W<1/L$.
The support of $B(g)$ is a chain of at most $L$ vertices, so a maximal coordinate is at least $1/L$.
Therefore, if $B(g)\in D_U$, the label $h_U$ agrees with every $b$-confident sign of $g$,
and so $h_U(x)g_x\ge-b$ for all $x$.

If $\|g\|_\infty>b_0$ for every $g\in X$, then $W>0$ on $X$, and $B$ is continuous on $X$.
The sets $B^{-1}(D_U)$ are closed, cover $X$, have multiplicity
at most $L=\ell+1$ by Lemma~\ref{res:2.0}, and are $b$-sign-compatible.
This proves $d_b^{\rm sign}(X)\le\ell$.

\proofstep{The general case.} Note that $W(g)=0$ exactly when $\|g\|_\infty\le b_0$. Take the cone on the order complex, with
apex $v_0$ carrying any Boolean label, and use
\[
 T(g)=(1-\tau(g))e_{v_0}+\tau(g)B(g),\qquad
 \tau(g)=\min\left\{1,
       \frac{(\|g\|_\infty-b_0)_+}{b-b_0}\right\}.
\]
Set $T(g)=e_{v_0}$ when $W=0$. The function $\tau$ is continuous and vanishes where $W=0$, while the
coordinates of $B$ are bounded, so $T$ is continuous there. Elsewhere it is continuous because $B$ is continuous where $W>0$.

Let $T(g)\in D_v$. If $g$ has a $b$-confident coordinate, then $\|g\|_\infty>b$, so $\tau(g)=1$ and
$T(g)=B(g)$. The apex coordinate is then zero, so $v\ne v_0$, and the
preceding maximal-coordinate argument applies. If there is
no such coordinate, every label $h$ satisfies $h_xg_x\ge-b$ for all $x$. So the sets
$T^{-1}(D_v)$ are $b$-sign-compatible. The cone has
dimension at most $\ell+1$, so these closed sets have multiplicity at most $\ell+2$ by Lemma~\ref{res:2.0}, which proves the bound $d_b^{\rm sign}(X)\le\ell+1$.
\end{proof}

We can now compute the sign-compatible dimensions for these matrices. On the vote image every confident sign agrees with the row that realizes the distribution, so the second assertion of Lemma~\ref{lem:affine-confidence-map} applies, and it meets the lower bound coming from list replicability. On the signed hull the confident signs are consistent at high confidence, while at moderate confidence the hull projects onto a larger cube.

\begin{theorem}\label{thm:walsh-sign-geometry}
For $\ell\ge1$ the following statements hold.
\begin{enumerate}[label=(\roman*)]
\item At $b=45/64$,
$d_b^{\rm sign}(V_H)=k_{\rm tilt}(H)=\ell$.
\item Put $s_\ell=2\lfloor(\ell+2)/2\rfloor$. If
$b>1-2/s_\ell$, then $d_b^{\rm sign}(P_H)=\ell+1$.
\item If $0<b<5/7$, then
$d_b^{\rm sign}(P_H)\ge \ell+\lfloor \ell/6\rfloor+1$.
\end{enumerate}
\end{theorem}
\begin{proof}
\proofstep{The vote of a realizable distribution.}
Let $\nu$ be realized by $\chi_{a_0}$ and let $\mu$ be its marginal
on $\mathbb F_2^\ell$, so that $\nu_{x,\chi_{a_0}(x)}=\mu(x)$. With the unnormalized Fourier transform
$\widehat\mu(a)=\sum_x\mu(x)\chi_a(x)$, put
$r=2\lceil16\ln(2\cdot2^\ell)\rceil$ and
$Z=\sum_a|\widehat\mu(a)|^r$. In the notation of Section~\ref{ssec:2.5} this is $r=2p$ with $p=p_N$, $N=2^\ell$. The profile of $\nu$ is
$\Phi(\nu)_a=\sum_x\mu(x)\chi_{a_0}(x)\chi_a(x)=\widehat\mu(a+a_0)$.
Substituting this in the vote of Section~\ref{ssec:2.5} and using $\chi_a=\chi_{a+a_0}\chi_{a_0}$, we get
\[
 \chi_{a_0}(x)g_\nu(x)=
 \frac{\sum_a\operatorname{sign}(\widehat\mu(a))
               |\widehat\mu(a)|^r\chi_a(x)}{Z}.
\]

\proofstep{Comparison with a convolution power.}
Since $r$ is even, replacing a numerator coefficient by
$\widehat\mu(a)^{r+1}$ changes it by $|\widehat\mu(a)|^r(1-|\widehat\mu(a)|)$. So this replacement changes every coordinate by at most
\[
 \sum_a\pi_a(1-|\widehat\mu(a)|)
 \le\sum_a\pi_a(-\ln|\widehat\mu(a)|)
 =\frac{H(\pi)-\ln Z}{r}\le\frac{\ln(2^\ell)}r<\frac1{32},
 \qquad \pi_a=|\widehat\mu(a)|^r/Z.
\]
Terms with zero weight are omitted. The equality holds because $-\ln|\widehat\mu(a)|=-(\ln\pi_a+\ln Z)/r$,
and $H(\pi)$ is the entropy of $\pi$, which is at most $\ln(2^\ell)$. Further $Z\ge1$ because
$\widehat\mu(0)=1$, and $r\ge32\ln(2^{\ell+1})$.

With the convolution
$(f*f')(x)=\sum_yf(y)f'(x+y)$, the transform turns convolutions into products, and
Fourier inversion reads $\sum_a\widehat f(a)\chi_a(x)=2^\ell f(x)$. As $r$ is even, $Z=\sum_a\widehat\mu(a)^r=2^\ell\mu^{*r}(0)$. Hence the resulting
ratio is
\[
 \frac{\sum_a\widehat\mu(a)^{r+1}\chi_a(x)}{Z}=\frac{\mu^{*(r+1)}(x)}{\mu^{*r}(0)},
\]
which is nonnegative.
It is at most one because its Fourier numerator is bounded in
absolute value by $\sum_a|\widehat\mu(a)|^{r+1}\le Z$. Therefore
$\chi_{a_0}(x)g_\nu(x)>-1/32$ for every $x$.

\proofstep{The vote image: proof of (i).}
Take $b_0=1/16$. If $|g_\nu(x)|>b_0$ and the sign of $g_\nu(x)$ differed from $\chi_{a_0}(x)$, we would have
$\chi_{a_0}(x)g_\nu(x)<-1/16$. So all signs confident above $b_0$ agree with
the same row, which is an affine Boolean function. Also, summing the first display against $\mu$,
\[
 \mathbb E_{x\sim\mu}[\chi_{a_0}(x)g_\nu(x)]
 =\sum_a\frac{\operatorname{sign}(\widehat\mu(a))|\widehat\mu(a)|^r\widehat\mu(a)}Z
 =\sum_a\pi_a|\widehat\mu(a)|
 =1-\sum_a\pi_a(1-|\widehat\mu(a)|)>31/32.
\]
Consequently $\|g_\nu\|_\infty>31/32>b_0$, and
Lemma~\ref{lem:affine-confidence-map}, with $X=V_H$ and $b_0<b=45/64$, gives
$d_b^{\rm sign}(V_H)\le \ell$. The exact value
$\operatorname{lr}(H)=\ell+1$ in Theorem~\ref{res:5.5}, together
with Proposition~\ref{prop:sign-cover}, gives
$\ell\le k_{\rm tilt}(H)\le d_b^{\rm sign}(V_H)\le\ell$, which proves (i).

\proofstep{High confidence in the full hull: proof of (ii).}
Since $s_\ell\ge2$, we can choose $0\le1-2/s_\ell<b_0<b$. A signed constraint $(x,(-1)^e)$ on an affine Boolean
function $(-1)^{\alpha\cdot x+c}$ is the equation $\alpha\cdot x+c=e$ in the unknown $(\alpha,c)\in\mathbb F_2^{\ell+1}$, with
coefficient vector $(x,1)$. A collection of such equations is inconsistent if and only if the
coefficient vectors of a subcollection sum to zero while their right sides sum to one.

Let $\mathcal S$ be a minimally inconsistent
collection, of size $s$. By minimality the coefficient vectors of all of $\mathcal S$ sum to zero, and the right sides sum to one. If a proper nonempty
subcollection $\mathcal S'$ also had coefficient vectors with sum zero, then so would $\mathcal S\setminus\mathcal S'$, and the right
sides of one of the two would sum to one, contradicting minimality. So the
augmented vectors $(x,1)$ of $\mathcal S$ form a
circuit. Its size $s$ is at most $\ell+2$, since the ambient dimension is $\ell+1$, and it is even, since the last coordinates sum to zero.
Hence $s\le s_\ell$.

Every affine Boolean
function violates at least one of these $s$ signed constraints $(x_j,y_j)$,
so its value of $\sum_jy_jf(x_j)$ is at most $s-2$. Thus every $g\in P_H$, being a convex combination of affine Boolean functions, satisfies
$\sum_{j=1}^s y_jg_{x_j}\le s-2$ on that inconsistent collection.
If all $s$ constraints were signs of $g$ confident above $b_0$, the sum would exceed
$sb_0>s(1-2/s_\ell)\ge s-2$, a contradiction. The confident constraints of $g$ are
therefore consistent, and the first assertion of Lemma~\ref{lem:affine-confidence-map}, with $X=P_H$,
gives $d_b^{\rm sign}(P_H)\le \ell+1$.

The $\ell+1$ coordinates $0,e_1,\ldots,e_\ell$ are shattered by affine
Boolean functions, whose values there are $(-1)^c$ and $(-1)^{\alpha_i+c}$. These functions are exactly the
signed rows, so Lemma~\ref{lem:projected-cube} gives equality.

\proofstep{A larger cube at moderate confidence: proof of (iii).}
Write $\ell=6q+r_0$ with $0\le r_0<6$. Select the coordinate $0$,
all $\ell$ basis coordinates, and the sum of the six basis vectors
in each of $q$ disjoint blocks. There are $d=\ell+q+1$ distinct coordinates.
Consider an affine Boolean function $f(x)=(-1)^{\alpha\cdot x+c}$ with sign $s=(-1)^c$ at $0$. On a block with basis vectors $e_i$ and sum $u$,
we have $f(e_i)=s(-1)^{\alpha_i}$ and $f(u)=s(-1)^{\sum_i\alpha_i}$, so the
seven signs in the block have product $s$.
Conversely, for fixed $s$, any vector of seven signs of product $s$ occurs: the six $\alpha_i$ are determined by the six basis signs, and the seventh sign is then correct.
The blocks are disjoint, so the signs on different blocks and the other basis
signs can be chosen independently.

The parity polytope
\[
 Q_s=\operatorname{conv}\{z\in\{\pm1\}^7:\prod_jz_j=s\}
\]
contains $(5/7)[-1,1]^7$. For a vertex $\sigma$ of the cube with the wrong parity, its seven single-coordinate flips have the right parity, and their
average has $j$-th coordinate $(6\sigma_j-\sigma_j)/7$, so it is $5/7$ times that
vertex. For a vertex of the right parity, use its segment to
zero. Zero belongs to $Q_s$ by averaging all its vertices, because flipping a fixed coordinate together with another one is a bijection of these vertices that reverses the sign of the fixed coordinate.
Convexity proves the assertion for the whole scaled cube.

Now let $\tau\in(5/7)[-1,1]^d$ be a target on the selected coordinates. For each fixed value of $s$, choose on each block a
distribution on the vertices of $Q_s$ with mean the block part of $\tau$, and on each of the other basis coordinates a
distribution on $\{\pm1\}$ with the prescribed mean, independently. This gives a distribution on signed rows of sign $s$ at $0$ with these means.
Mixing the two choices of $s$, with probabilities $(1\pm\tau_0)/2$, gives the mean $\tau_0$ at $0$ while retaining the other means. Hence the selected
projection of $P_H$ contains $(5/7)[-1,1]^d$. As $b<5/7$,
Lemma~\ref{lem:projected-cube} proves (iii).
\end{proof}

Since $45/64<5/7$, parts (i) and (iii) apply at the same threshold $b=45/64$. At this threshold the sign-compatible dimension of the vote image is smaller than that of the signed hull by at least $\lfloor\ell/6\rfloor+1$, so the nonlinear vote map has a quantitative effect on this invariant.

The vote of a distribution realized by a row of $H$ has one further exact property: it vanishes outside the affine hull of the support of $\mu$. So all confident signs of the vote lie in that hull, which fits the affine closures of Lemma~\ref{lem:affine-confidence-map}. In the notation of the proof above, if
$\operatorname{supp}\mu\subseteq x_0+L$ for a linear subspace $L$, then
$g_\nu(x)=0$ outside $x_0+L$. Indeed, for
$h\in L^\perp$ we have $\chi_h(x')=\chi_h(x_0)$ on the support of $\mu$, so
$\widehat\mu(a+h)=(-1)^{h\cdot x_0}\widehat\mu(a)$, and the signed power coefficient
$C(a)=\operatorname{sign}(\widehat\mu(a))|\widehat\mu(a)|^r$ transforms in the
same way. Replacing $a$ by $a+h$ in $\Gamma(x)=\sum_aC(a)\chi_a(x)$ gives
$\Gamma(x)=\chi_h(x+x_0)\Gamma(x)$. If $x\notin x_0+L$, then $x+x_0\notin L=(L^\perp)^\perp$, so some
$h\in L^\perp$ has $\chi_h(x+x_0)=-1$, and $\Gamma(x)=0$. By the first display of the proof, $g_\nu(x)=\chi_{a_0}(x)\Gamma(x)/Z=0$.

\subsection{Linear measurements and two successive thresholds}
We now turn from the vote image to the full signed hull, and we ask what direct compressions of the hull must cost. The hull $P_H$ is the image under $H$ of an $\ell_1$ ball, so the most direct compressions are linear measurements and the thresholding of Fourier coefficients. For large $\ell$, both cost far more than the dimension $\ell$ of the nonlinear vote image of Theorem~\ref{thm:walsh-sign-geometry}: linear measurements need rank proportional to $2^\ell$, and two successive thresholds lead to an image of dimension at least $2^{\ell/2}/2$. Throughout this subsection $n=2^\ell$.

A linear measurement $A$ has \emph{$b$-sign-compatible fibers on $P_H$} if every set $A^{-1}(y)\cap P_H$ is $b$-sign-compatible. In that case the measurement alone determines a label that respects all signs of confidence above $b$.

\begin{proposition}\label{prop:linear-sign-measurements}
Let $n=2^\ell$ and $0<b<1$. If a linear measurement $A:\mathbb R^n\to\mathbb R^k$
has $b$-sign-compatible fibers on $P_H$, then
$\operatorname{rank}A\ge n(1-b^2)$. At $b=45/64$ this is
$2071n/4096$.
\end{proposition}
\begin{proof}
Since $H^T=H$ and $H^TH=nI$, a point $g=H\alpha$ determines $\alpha=H^Tg/n$, and
the hull is
$P_H=\{H\alpha:\|\alpha\|_1\le1\}=\{g:\|H^Tg\|_1/n\le1\}$. It contains the Euclidean unit
ball, because Cauchy--Schwarz gives $\|H^Tg\|_1\le\sqrt n\|H^Tg\|_2=n\|g\|_2$.

Let $Q$ be the orthogonal projection onto
$\ker A$. Then $\operatorname{tr}Q=n-\operatorname{rank}A$, so some diagonal entry satisfies
$Q_{ii}\ge\operatorname{tr}Q/n=1-\operatorname{rank}A/n$.
If the asserted rank bound fails, this entry is greater than
$b^2$. Since $Q^2=Q=Q^T$, we have $\|Qe_i\|_2^2=Q_{ii}$, so the vector $x=Qe_i/\sqrt{Q_{ii}}$ is a unit vector in the kernel,
and $x_i=Q_{ii}/\sqrt{Q_{ii}}=\sqrt{Q_{ii}}>b$. Both $x$ and $-x$ lie in $P_H$
and in the same fiber $\ker A\cap P_H$, with $x_i>b$ and $(-x)_i<-b$, contradicting compatibility.
Finally $1-(45/64)^2=2071/4096$.
\end{proof}
For the matrices $H$ this contains the $\varepsilon$-rank bound
of~\cite[Theorem 1.2]{ALSV13}, in fiber form.

A second direct compression thresholds twice. It keeps the Fourier coefficients of $g$ above a level $\delta$, and then the values above a level $c$ of the function left after this first step. For $a\ge0$ write $S_a(t)=\operatorname{sign}(t)(|t|-a)_+$,
applied coordinatewise. Given $g=H\alpha\in P_H$, define
\[
 E_{\delta,c}(g)=\big(S_\delta(\alpha),
               S_c(H(\alpha-S_\delta(\alpha)))\big).
\]
This encoding gives an upper bound of order $\sqrt n$ for the sign-compatible dimension of the hull. The second assertion of the next theorem shows that for no choice of the two levels does its image have dimension of smaller order.
\begin{theorem}\label{thm:two-threshold-geometry}
For $0<b<1$, with $\mathbb N=\{1,2,\ldots\}$,
\[
 d_b^{\rm sign}(P_H)\le
 \min\left\{n,\min_{s\in\mathbb N}
       \left(s-1+\left\lfloor\frac{n}{s b^2}\right\rfloor\right)\right\}
 \le\min\{n,\lceil2\sqrt n/b\rceil\}.
\]
On the other hand, for every $\delta>0$ and $0<c<1$, the polyhedral
image $E_{\delta,c}(P_H)$ has dimension at least $\sqrt n/2$.
\end{theorem}
\begin{proof}
\proofstep{The encoding at suitable levels.}
Fix an integer $s\ge1$, set $\delta=1/s$, and put
$q=\lfloor n/(sb^2)\rfloor$. Since $n/(sb^2)<q+1$, we can choose $0<c<b$ so close to $b$ that
$n/(sc^2)<q+1$. For $g=H\alpha\in P_H$ let $A=S_\delta(\alpha)$ and $\beta=\alpha-A$. A nonzero coordinate of $A$ has $|\alpha_a|>1/s$, and
$\|\alpha\|_1\le1$, so $A$ has fewer than $s$ of them. Moreover $|\beta_a|=\min(|\alpha_a|,\delta)$, so $\beta_a^2\le\delta|\alpha_a|$. Thus
\[
 |\operatorname{supp}A|\le s-1,
 \qquad \|\beta\|_2^2\le\delta\|\alpha\|_1\le1/s.
\]
Since $H^TH=nI$, this gives $\|H\beta\|_2^2\le n/s$. If
$T=S_c(H\beta)$ has $k\ge1$ nonzero coordinates, then $kc^2<\|H\beta\|_2^2\le n/s$, so $k<n/(sc^2)<q+1$, that is, $k\le q$.

The map
$E=E_{\delta,c}$ is continuous and piecewise affine on the polytope $P_H$, because $\alpha=Hg/n$ is linear in $g$ and the thresholds are piecewise affine. So its image
is a finite union of polytopes. It lies in the finite union of coordinate
subspaces given by the two support bounds, of dimension at most $s-1+q$.
The continuous decoder $\mathcal D(A,T)=HA+T$ satisfies
$\mathcal D(E(g))-g=T-H\beta$, whose coordinates have absolute value $\min(|(H\beta)_x|,c)$; so
$\|\mathcal D(E(g))-g\|_\infty\le c$.

\proofstep{A cover from the encoding.}
Triangulate the image so finely that $\mathcal D$ varies by less
than $b-c$ on each simplex. Label each vertex $v$ by the sign vector $h_v$ of $\mathcal D(v)$,
with a fixed convention at zero. If $E(g)$ lies in a simplex with vertex $v$, then
$\|\mathcal D(v)-g\|_\infty<(b-c)+c=b$. Hence, if $yg_x>b$, then $y\mathcal D(v)_x>0$ and $h_v(x)=y$.
The regions $D_v$ of Lemma~\ref{res:2.0} lie in the closed stars, so the sets
$E^{-1}(D_v)$ are closed, cover $P_H$, have multiplicity at most $s+q$, and are $b$-sign-compatible, with $h_v$ as a compatible label. This gives
$d_b^{\rm sign}(P_H)\le s-1+q$.

\proofstep{The bound $n$ and the last inequality.}
The bound $d_b^{\rm sign}(P_H)\le n$ follows from a triangulation of the polytope $P_H\subseteq\mathbb R^n$ whose closed vertex stars have diameter less than $2b$:
the regions $D_v$ have multiplicity at most $n+1$, and a set of diameter less than $2b$ is $b$-sign-compatible. Taking $s=\lceil\sqrt n/b\rceil$ gives
$s-1<\sqrt n/b$ and $n/(sb^2)\le\sqrt n/b$, hence $s-1+\lfloor n/(sb^2)\rfloor<2\sqrt n/b$, which implies the last inequality.

\proofstep{Many independently varying Fourier coordinates.}
For the lower bound on the image, let $\delta>0$ and $0<c<1$ be arbitrary.
Set $k=\min\{n,\max\{0,\lceil1/\delta\rceil-1\}\}$.
When $k\ge1$ we have $k<1/\delta$, that is, $k\delta<1$. Choose $k$ Fourier coordinates,
set each to $a=(\delta+1/k)/2$, and vary them independently by
less than $(1-k\delta)/(4k)$, with all other coefficients zero. Since $a-\delta=(1-k\delta)/(2k)$, they stay above $\delta$. Their
sum stays below $k\bigl(a+\frac{1-k\delta}{4k}\bigr)=\frac34+\frac{k\delta}4<1$, so $g=H\alpha\in P_H$. The first
component $S_\delta(\alpha)$ equals $\alpha-\delta$ on these coordinates and varies in a $k$-dimensional box,
so the image has dimension at least $k$.

\proofstep{Many independently varying residual coordinates.}
Suppose $\delta\ge1/n$. Choose the least power of two $s$ with
$s\ge\max\{1,1/\delta\}$, so that $s\le n$, and a subgroup
$K\subseteq\mathbb F_2^\ell$ of size $s$. Its annihilator $L$ has
size $n/s$. Put $t=(1+c)/2$ and
$\alpha_0=(t/s)\mathbf1_K$. Then $H\alpha_0=t\mathbf1_L$, by orthogonality of characters on $K$.
Perturb this vector independently on $L$ by values $u_x$ of
absolute value less than
\[
 \varepsilon=\min\{(t-c)/2,(1-t)s/(2n),(s\delta-t)/2\},
\]
which is positive because $t-c=1-t=(1-c)/2>0$ and $s\delta\ge1>t$.

The inverse Fourier perturbation $a\mapsto\frac1n\sum_{x\in L}u_x\chi_a(x)$ has each coefficient at most
$|L|\varepsilon/n=\varepsilon/s$ in absolute value, and $\ell_1$ norm at most
$n\varepsilon/s\le(1-t)/2$. The perturbed coefficient vector $\alpha$ consequently
has $\|\alpha\|_1\le t+(1-t)/2<1$, so $g=H\alpha\in P_H$, and every coefficient has magnitude at most $(t+\varepsilon)/s$, which is less
than $\delta$ because $t+\varepsilon<s\delta$. So the first threshold $S_\delta(\alpha)$ is zero, $\beta=\alpha$, and $H\beta=g$. On $L$ we have
$g_x=t+u_x>c$, and off $L$ we have $g_x=0$. So the second threshold
equals $t-c+u_x$ on $L$ and zero elsewhere. This supplies an
$(n/s)$-dimensional box in the image.

\proofstep{Conclusion.}
If $\delta<1/n$, the first construction gives $k=n$. If
$\delta\ge1$, the second gives $s=1$ and dimension $n$. In the
range $1/n\le\delta<1$, put $x=1/\delta\in(1,n]$. Then $k=\lceil x\rceil-1\ge x/2$: for $x\le2$ we have $k=1$, and for $x>2$ we have $k\ge x-1\ge x/2$. So
$k\ge1/(2\delta)$. Further $s\ge2$ and $s/2<1/\delta$, so $s<2/\delta$, hence
$n/s>n\delta/2$. The maximum of these two lower bounds is at
least their geometric mean $\sqrt n/2$. This proves the image-dimension assertion.
\end{proof}

\subsection{An entropy bound on coordinate cube obstructions}
The lower bounds for $d_b^{\rm sign}(P_H)$ in Theorem~\ref{thm:walsh-sign-geometry} come from cubes in coordinate projections, whereas the upper bound of Theorem~\ref{thm:two-threshold-geometry} has order $\sqrt n$. The next bound limits the dimension detectable by the coordinate
cube argument, independently of any particular threshold construction. A map $f$ as in its hypothesis forces $d_b^{\rm sign}(P_B)\ge d$, by the pullback argument in the proof of Lemma~\ref{lem:projected-cube}, so the proposition says how far such an argument can reach.
\begin{proposition}\label{prop:coordinate-cube-bound}
Let $B$ have $m$ rows and let $d\ge1$. Suppose a continuous map
$f:[-1,1]^d\to P_B$ and distinct coordinates $i_1,\ldots,i_d$
satisfy $f_{i_j}(z)>b$ on $z_j=1$ and
$f_{i_j}(z)<-b$ on $z_j=-1$. Then
\[
 d<\frac{2\ln(2m)}{b^2},\qquad
 2m\sum_{j=0}^{\lceil(1-b)d/2\rceil-1}\binom dj\ge2^d.
\]
\end{proposition}
\begin{proof}
\proofstep{Signed rows near the corners.}
Each of the $2d$ faces of the cube is compact and carries a strict inequality
for a continuous function, so there is a common $t>b$ with
$f_{i_j}\ge t$ on $z_j=1$ and $f_{i_j}\le-t$ on $z_j=-1$, for all $j$. A corner $\sigma$ lies on the
face $z_j=\sigma_j$ for every $j$, so
$\sum_j\sigma_jf_{i_j}(\sigma)\ge td$. Since $f(\sigma)$ is a
convex combination of signed rows, some signed row $v$ satisfies
$\sum_j\sigma_jv_{i_j}\ge td$.

\proofstep{The first bound.}
For a uniform random corner $\sigma$ and any $\lambda>0$,
\[
 \mathbb E\max_{v\in C_B\cup-C_B}\sum_j\sigma_jv_{i_j}
 \le\frac{\ln(2m)}\lambda+\frac{d\lambda}{2}.
\]
Indeed, by Jensen's inequality the exponential of $\lambda$ times the left side is at most the
expectation of $\exp(\lambda\max_v\sum_j\sigma_jv_{i_j})$. That exponential is bounded by the sum of the
exponentials over the at most $2m$ signed rows, and each of these has expectation
$\prod_j\cosh(\lambda v_{i_j})=(\cosh\lambda)^d\le e^{d\lambda^2/2}$. Taking logarithms gives the display. With
$\lambda=\sqrt{2\ln(2m)/d}$ the right side equals $\sqrt{2d\ln(2m)}$. As the maximum is at least $td$ at
every corner, $td\le\sqrt{2d\ln(2m)}$, that is, $d\le2\ln(2m)/t^2$, and $t>b$ gives the strict bound.

\proofstep{The covering count.}
Let $\sigma$ be a corner and $v$ a signed row with $\sum_j\sigma_jv_{i_j}\ge td$, and let $e$ be the number of $j$ with
$v_{i_j}\ne\sigma_j$. Then $d-2e\ge td$, so $e\le(1-t)d/2<(1-b)d/2$. As $e$ is an integer,
$e\le\lceil(1-b)d/2\rceil-1$, a nonnegative number since $d\ge1$. So the Hamming balls of this radius
centered at the restrictions of the at most $2m$ signed rows to the
chosen coordinates cover all $2^d$ corners.
Counting their sizes proves the second assertion.
\end{proof}

For $H=H_{2^\ell}$ we have $m=2^\ell$, so the first bound gives $d<2(\ell+1)\ln2/b^2$. Thus, for fixed $b$, the coordinate cube argument cannot give more than order $\ell$ on these matrices, and the lower bound of Theorem~\ref{thm:walsh-sign-geometry}(iii) has the best order that this argument can reach.

\appendix
\addtocontents{toc}{\protect\medskip}
\section{Computational verification}\label{sec:computation}
The proof of Theorem~\ref{thm:main} and all results not listed below are
independent of computer calculation. The statements listed below rest on, or are
confirmed by, finite computations, which are carried out in exact integer or rational
arithmetic, or over $\mathbb F_2$. The programs and certificate files accompany this article in
the directory \texttt{verification} of its source; a file \texttt{README.txt}
there gives the commands and requirements.

\begin{description}[leftmargin=1.5em,style=unboxed]
\item[The class $C_{K_4}$ (Lemma~\ref{lem:k4-links}).]
Among the $64$ orientations of $K_4$ we select the $32$ source-free ones,
enumerate the $603$ faces of the literal complex, including the empty face, and
group them into $33$ orbits under the permutations of the vertices of $K_4$. For a
representative of each orbit, the integral homology of the link is computed from
the Smith normal forms of all its boundary matrices, including torsion.
Program \texttt{check\_c32.py}; record \texttt{c32\_integral\_check.json}.

\item[The Fano plane (Proposition~\ref{res:5.7}).]
The proof in the text can be checked by hand, and the computation confirms it.
We check the signs of the $49$ integer inner products of the rank-four sign
representation, that the $12$ displayed tetrahedra are simplices of the sign
complex, that the displayed chain is a cycle with pairing $1$ against $w_1^3$, the
values of $\dim C_2$, $\dim C_3$, $\operatorname{rank}\partial_2$, and
$\operatorname{rank}\partial_3$ over $\mathbb F_2$, and that the cochain
of $w_1^3$ is not a coboundary.
Program \texttt{check\_finite\_certificates.py}; record
\texttt{finite\_certificates.json.gz}.

\item[The Hadamard matrices $H_8$ and $H_{16}$ (Theorem~\ref{thm:h8-om}).]
We check the three-term Grassmann--Pl\"ucker relations for the displayed
chirotope and enumerate its $128$ topes, which contain all $16$ affine sign
functions. The $13\,888$ assignments compatible with these functions are
enumerated by backtracking with unit propagation, using no solver, and again by
each of three satisfiability solvers with blocking clauses. All four lists agree.
We check that the group of symmetries preserves the list and has seven
orbits of the stated sizes, and we replay the seven rational certificates, checking
their positivity, normalization, and exact identities. The integer factors displayed in the proof, and the rational factors recorded
in the certificate files from which they are obtained by scaling (by $6$ and by $40$) and
rounding, are multiplied out exactly.
Programs \texttt{check\_hadamard\_om.py}, \texttt{check\_h8\_enumeration.py}, and
\texttt{om\_tools.py}; certificates \texttt{h8\_nonrealizability.json},
\texttt{h8\_signrank5\_realization.json}, and
\texttt{h16\_signrank7\_realization.json}.

\item[The Paley matrix $P_{13}$ (Proposition~\ref{res:8.9}).]
We evaluate the $169$ integer inner products of the five-dimensional
representation and locate the four equality cases, and we check the $16$ signed
patterns on the columns $\{0,1,2,4\}$. The integer three-cochain $b$ is
checked against the Euler cocycle on all $13\,572$ quotient four-simplices. The
$42\,970$ antipodal pairs of elementary collapses are replayed one at a time against
the current complex. The final complex has dimension four and $15\,970$ nonempty faces.
Programs \texttt{check\_finite\_certificates.py} and
\texttt{check\_paley\_integral.py}; certificates
\texttt{finite\_certificates.json.gz} and
\texttt{paley13\_integral\_certificate.json.gz}.

\item[Tables for small primes (Section~\ref{ssec:8.6}).]
For $p=5,13,17,29,37$ we enumerate all nonempty signed closures of $P_p$, and
compute their number, the height, the largest switchable square, and the numbers
$m_a$ (Table~\ref{tab:paley-exact}); there are $1\,304\,842$ closures at $p=37$.
For the $21$ primes of Table~\ref{tab:paley-bounds} we check the values of $C_p$ and
$r_2$ and the integer inequalities that exclude each listed rectangle. The signed
patterns of the shattered column sets are checked directly.
Programs \texttt{check\_paley\_tables.py}, \texttt{check\_vote\_geometry.py}, and
\texttt{check\_finite\_certificates.py}.
\end{description}

The directory also contains programs that test statements proved in the text on
further examples: the low-dual-dimension priority lists of Section~\ref{app:static},
the sign patterns of Hamming balls and the row vectors of punctured cubes of
Section~\ref{app:families}, the families of Section~\ref{app:low-index}, and the character
identities, encodings, and parity polytope of Section~\ref{app:vote-geometry}
(\texttt{check\_structural\_families.py} and \texttt{check\_vote\_geometry.py}).
The proofs do not depend on these runs.

\end{document}